%% file: siggraph.tex
\documentclass[acmtog]{acmart}
\acmSubmissionID{307}

\usepackage{appendix}

\usepackage{booktabs}
\usepackage{soul}
\usepackage{caption}
\usepackage{graphicx}
\usepackage{amsmath}
\usepackage{multirow}

\makeatletter
\let\BaseCaption\@makecaption
\makeatother
\usepackage{subcaption}
\makeatletter
\let\@makecaption\BaseCaption
\makeatother

\usepackage[ruled]{algorithm2e} %

\SetAlFnt{\small}
\SetAlCapFnt{\small}
\SetAlCapNameFnt{\small}
\SetAlCapHSkip{0pt}

\newcommand*\concat{\mathbin{\|}}
\newcommand{\etal}{\textit{et al}.}

\setcopyright{acmcopyright}\acmJournal{TOG}
\acmYear{2021}\acmVolume{40}\acmNumber{4}\acmArticle{45}\acmMonth{8} \acmDOI{10.1145/3450626.3459805}

\begin{document}

\title{Only a Matter of Style: Age Transformation Using a Style-Based Regression Model}

\author{Yuval Alaluf}
\affiliation{%
 \institution{Tel-Aviv University}
}
\email{yuvalalaluf@gmail.com}
\author{Or Patashnik}
\affiliation{%
 \institution{Tel-Aviv University}
}
\author{Daniel Cohen-Or}
\affiliation{%
 \institution{Tel-Aviv University}
}

\begin{abstract}
\input{abstract}
\end{abstract}

\begin{CCSXML}
<ccs2012>
<concept>
<concept_id>10010147.10010371</concept_id>
<concept_desc>Computing methodologies~Computer graphics</concept_desc>
<concept_significance>300</concept_significance>
</concept>
<concept>
<concept_id>10010147.10010371.10010382</concept_id>
<concept_desc>Computing methodologies~Image manipulation</concept_desc>
<concept_significance>500</concept_significance>
</concept>
<concept>
<concept_id>10010147.10010257.10010293.10010294</concept_id>
<concept_desc>Computing methodologies~Neural networks</concept_desc>
<concept_significance>100</concept_significance>
</concept>
</ccs2012>
\end{CCSXML}

\ccsdesc[500]{Computing methodologies~Image manipulation}
\ccsdesc[300]{Computing methodologies~Computer graphics}
\ccsdesc[100]{Computing methodologies~Neural networks}

\keywords{Generative Adversarial Networks, Age Transformation, Image Editing}

\begin{teaserfigure}
    \centering
    \vspace{-0.2cm}
    \includegraphics[width=\linewidth]{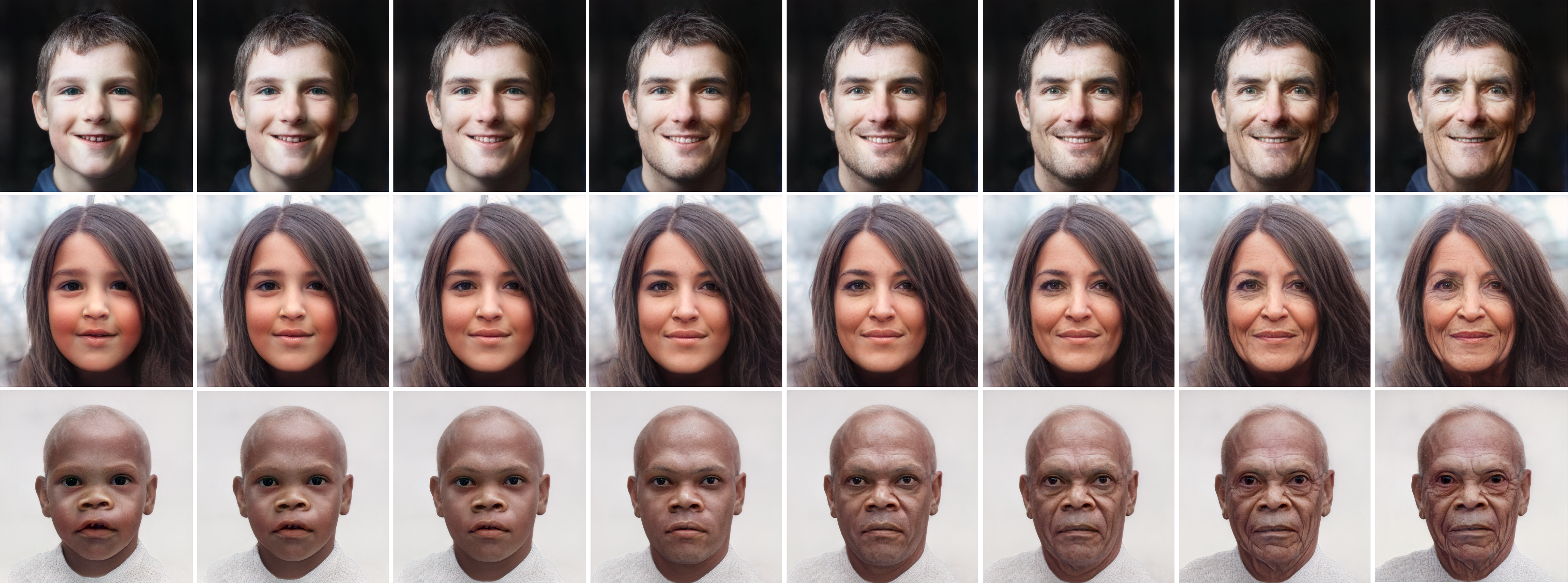}
    \vspace{-0.65cm}
    \caption{Modeling fine-grained lifelong age transformation using the style-based SAM method.}
    \vspace{0.1cm}
    \label{fig:teaser}
\end{teaserfigure}

\maketitle

\input{intro}
\input{related_work}
\input{method}

\input{results}
\input{conclusions}

\begin{acks}
    We would like to thank Elad Richardson, Kfir Goldberg, Ohad Fried, Yotam Nitzan, and Zongze Wu for their fruitful discussions and early feedback. We would also like to thank the anonymous reviewers for their insightful comments and constructive remarks. This work was supported in part by the 
    Israel Science Foundation under Grant No.~\grantnum{}{2366/16} and Grant No.~\grantnum{}{2492/20}.
\end{acks}

{\small
\bibliographystyle{ACM-Reference-Format}
\bibliography{siggraph}
}

\clearpage
\appendix
\appendixpage
\input{appendix}

\end{document}

%% file: abstract.tex
The  task of age transformation illustrates the change of an individual's appearance over time. Accurately modeling this complex transformation over an input facial image is extremely challenging as it requires making convincing, possibly large changes to facial features and head shape, while still preserving the input identity. In this work, we present an image-to-image translation method that learns to directly encode real facial images into the latent space of a pre-trained unconditional GAN (e.g., StyleGAN) subject to a given aging shift. We employ a pre-trained age regression network to explicitly guide the encoder in generating the latent codes corresponding to the desired age. In this formulation, our method approaches the continuous aging process as a regression task between the input age and desired target age, providing fine-grained control over the generated image. Moreover, unlike approaches that operate solely in the latent space using a prior on the path controlling age, our method learns a more disentangled, non-linear path. Finally, we demonstrate that the end-to-end nature of our approach, coupled with the rich semantic latent space of StyleGAN, allows for further editing of the generated images. Qualitative and quantitative evaluations show the advantages of our method compared to state-of-the-art approaches. Code is available at our project page:  \url{https://yuval-alaluf.github.io/SAM}.

%% file: intro.tex
\section{Introduction}
Age transformation is the process of representing the change in a person's appearance across different ages while preserving their identity. 
Recently, this task has received increased attention with the rise of applications allowing users to perform facial editing, and age transformation in particular.
To model the aging process over a single input facial image one must capture both the change in head shape and texture while faithfully preserving the identity and other key facial attributes of the input face. This becomes increasingly challenging when modeling \textit{lifelong} aging where the desired change in age becomes significant (e.g., from ages 5 to 85). 

To bypass explicitly modeling age transformation, data-driven techniques have recently been explored. Due to their phenomenal realism, Generative Adversarial Networks (GANs) have been heavily used for synthesizing images in a data-driven fashion, particularly on facial images. These works can typically be classified into two methodologies: image-to-image translation and latent space manipulation.

To model age transformation as an image-to-image problem, most works~\cite{antipov2017face, zhang2017age, 9009536, orel2020lifespan, 8578926, 7780630, yang2019learning} use age-annotated data and learn a mapping between pre-defined age groups. As the age groups are highly correlated, these methods struggle to model meaningful changes between different ages. Moreover, collecting age-annotated data is often tedious at large scale. 

Other works~\cite{harkonen2020ganspace, abdal2020styleflow, shen2020interpreting, liu2020style, wu2020stylespace} have approached the age transformation task by exploring the semantics of the latent space of a well-trained GAN, such as StyleGAN~\cite{karras2019style, karras2020analyzing}, and perform a latent space traversal to obtain the desired transformed image.
These methods often assume the existence of a corresponding linear path in the latent space controlling the attribute of interest. This, however, relies on the existence of a well-behaved, fully disentangled and linear latent space which is difficult to obtain. 
In addition, while such methods have shown promising editing results on synthetic images generated by StyleGAN, they often struggle to make realistic transformations on \textit{real} images.

In this work, we present a novel method for learning a conditional image generation function capable of capturing the desired change in age while faithfully preserving identity. We approach the age transformation task as an image-to-image translation problem by pairing the expressiveness of a \textit{pre-trained, fixed} StyleGAN generator with an encoder architecture. The encoder is tasked with directly encoding an input facial image into a series of style vectors subject to the desired age change.
These style vectors are then fed into StyleGAN to generate the output image representing the desired age transformation. 
This allows us to easily leverage the state-of-the-art image quality achieved by StyleGAN.
To explicitly guide the encoder in generating the corresponding latent codes, we utilize a fixed, pre-trained age regression network to serve as an additional constraint during the training process.
We name our method \textit{SAM --- Style-based Age Manipulation}, as our age transformation is controlled via the learned intermediate style representation.

Rather than using labeled data directly, our method attains supervision only through the use of readily available pre-trained networks: (i) a StyleGAN2~\cite{karras2020analyzing} generator network trained on facial images, (ii) a pre-trained encoder network from Richardson \etal~\shortcite{richardson2020encoding} trained to encode real face images into the $\mathcal{W}+$ latent space, (iii) an ArcFace~\cite{deng2019arcface} facial recognition network for identity regression, and (iv) a VGG~\cite{simonyan2015deep} network for age regression. Note that our approach does not assume the existence of multiple age classes or domains. Instead, given a single input image and desired target age, we show that our approach can successfully generate the corresponding image, see Figure~\ref{fig:teaser}. Compared to multi-domain approaches that rely on pre-defined age groups~\cite{choi2020stargan, liu2019few} or anchor classes~\cite{orel2020lifespan}, viewing human aging as a continuous \textit{regression} process allows for more fine-grained control over the desired transformation. 

Furthermore, we analyze the latent path learned by SAM and show it results in a more precise, \textit{non-linear} path that is less entangled with other attributes and conforms well to StyleGAN's latent space manifold. Finally, we demonstrate how the learned disentangled path and the StyleGAN latent space, together with the end-to-end nature of SAM, allow for additional fine-grained editing on the generated aging results (e.g. hair color and expression).

Qualitative and quantitative evaluations show that our style-based regression method outperforms current state-of-the-art methods.
The main contributions of this paper are: 
(i) A novel style-based regression approach for fine-grained modeling of the age transformation process; and (ii) an analysis of the non-linear latent path learned by our approach showing the benefits of an end-to-end method for modeling age transformation on real facial images.

%% file: related_work.tex
\section{Related Work}
Age transformation has been an extensively studied topic in computer graphics. Early works either explicitly modeled the face transformation over time or sought a prototype face for each age group. We refer the reader to \cite{duong2018longitudinal, age_synthesis_survey, RAMANATHAN2009131} for a comprehensive survey of such approaches. 
More recent works~\cite{li2019uva, orel2020lifespan, yao2020high, he2018attgan, lample2018fader} take a data-driven approach for modeling face aging by using deep neural networks for image synthesis. 

\subsection{Image-to-Image Translation}
Image-to-image translation techniques aim at translating a given image of a source domain to a corresponding image of a target domain. Isola \etal~\shortcite{isola2018imagetoimage} first introduced the use of conditional GANs~\cite{mirza2014conditional} for solving various image-to-image translation tasks. As corresponding pairs of source domain and target domain images are not always available or are overly tedious to collect, unpaired approaches have recently been developed for solving such tasks~\cite{unit, CycleGAN2017, dualgan}.
Some methods~\cite{huang2018munit, DRIT_plus} translate a given image to a diverse set of corresponding output images, often referred to as multi-modal image synthesis.
Furthermore, \cite{choi2018stargan, choi2020stargan, huang2018munit, liu2019few, zhu2017toward} generalize image-to-image translation into a \textit{multi-domain} translation and train a single network to translate between multiple domains. 

Motivated by the early successes of these works, many methods~\cite{triple-gan, Georgopoulos_2020_CVPR_Workshops, orel2020lifespan, li2020eccv, viazovetskyi2020stylegan2} have approached the task of face aging as an image-to-image translation between multiple age groups. These works associate each image with an age label and perform translation between pre-defined age groups. 
However, as the age groups are highly correlated, these methods often struggle to disentangle age and other attributes and may struggle to faithfully preserve identity when modeling lifelong aging. In contrast to these works, we view age transformation as a \textit{regression problem} by using a pre-trained age classifier to estimate age during training without the need for direct supervision. 
While Yao \etal~\shortcite{yao2020high} use a similar approach, they are limited to translating images within the $20-70$ age range. Moreover, the encoder-decoder architecture of Yao \etal~\shortcite{yao2020high} introduces a wide bottleneck ($256\times256$). As shall be shown, although this assists in reconstructing the input image, they are limited to modeling slight changes in texture and struggle to capture more global changes such as changes in head shape.

\begin{figure*}
    \centering
    \includegraphics[width=0.925\linewidth]{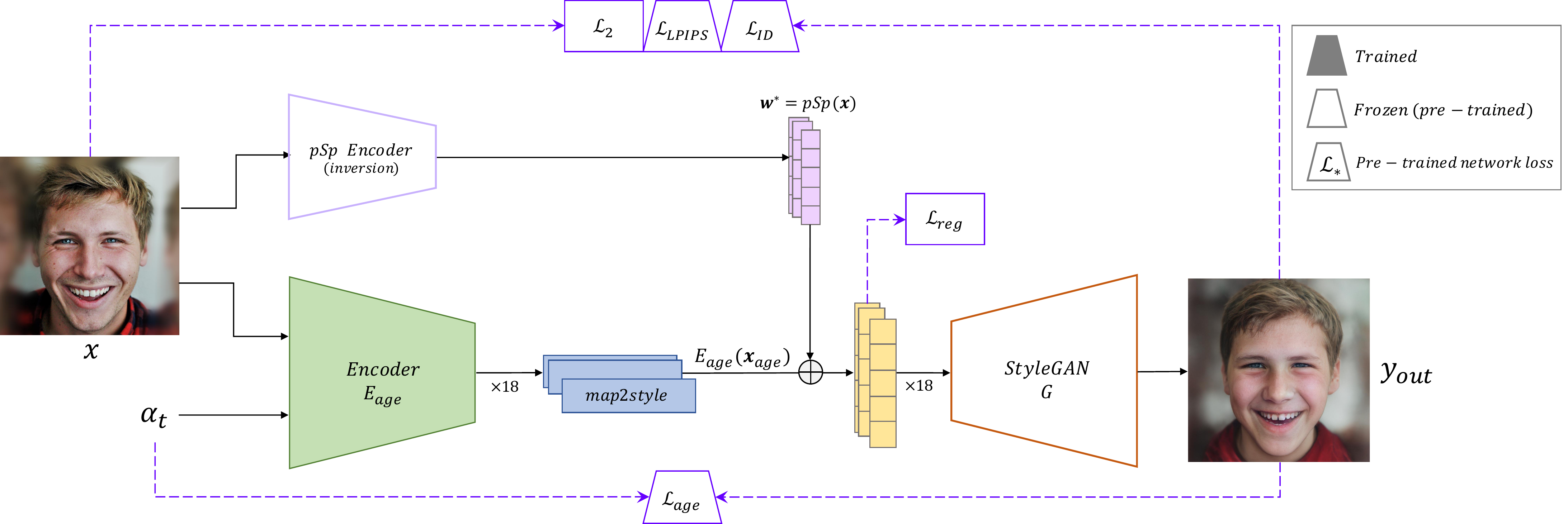}
    \vspace{-0.25cm}
    \caption{Our SAM architecture. The network receives an input face image and a desired target age $\alpha_t$. First, the aging encoder $E_{age}$ is tasked with extracting  feature maps at $3$ different spatial scales. Then, $18$ \textit{map2style} blocks, introduced in~\cite{richardson2020encoding}, are used to gradually down-sample the $3$ feature maps into $18$ different $512$-dimensional style vectors, thereby encoding the input image into the $\mathcal{W}+$ StyleGAN latent space.
    We additionally employ a fixed, pre-trained pSp~\cite{richardson2020encoding} encoder to extract the $\mathcal{W}+$ latent code of $\textbf{x}$, denoted $\textbf{w}^*$, which is then added to the age-transformed latent code, denoted $E_{age}(\textbf{x}_{age})$.
    A pre-trained StyleGAN is then used to generate the desired age-transformed image using the aggregated latent code. During training $\mathcal{L}_{2}, \mathcal{L}_{LPIPS}$ and $\mathcal{L}_{ID}$ ensure visual similarity and identity preservation while $\mathcal{L}_{reg}$ encourages the learned latent codes to be closer to the average latent code. Finally, $\mathcal{L}_{age}$ guides the encoder in generating the desired age-transformed latent code. Observe that during training, only the aging encoder and map2style blocks are trained. Moreover, $\mathcal{L}_{LPIPS}$, $\mathcal{L}_{ID}$, and $\mathcal{L}_{age}$ are computed via fixed, pre-trained networks as described in Section~\ref{losses}.}
    \vspace{-0.2cm}
    \label{fig:architecture}
\end{figure*}

\vspace{-0.1cm}
\subsection{The Latent Space of GANs}
Recently, many works have explored performing semantically meaningful manipulations in the latent space of a well-trained GAN generator, either by directly learning a disentangled mapping of an attribute of interest~\cite{nitzan2020dis} or by a latent space traversal~\cite{goetschalckx2019ganalyze, denton2019detecting, voynov2020unsupervised, harkonen2020ganspace, shen2020interpreting, abdal2020styleflow, shen2020closedform, wu2020stylespace}. Most notably, given its state-of-the-art image quality and disentangled latent space, StyleGAN has been widely used for this task. 

Such latent space methods first perform \textit{GAN inversion}, first introduced by Zhu \etal~\shortcite{zhu2016generative}, in which real images are projected into the latent space of the GAN by finding the latent vector representation that best approximates the input image. Recent years have seen diverse approaches for performing such an inversion~\cite{creswell2018inverting,abdal2019image2stylegan,bau2020semantic,pidhorskyi2020adversarial,abdal2020image2stylegan++,tewari2020pie,zhu2020domain,richardson2020encoding,tov2021designing, xia2021gan}.
Given the inversion, the resulting latent codes are then edited in a semantically meaningful manner by traversing the latent space to obtain a new latent code that is used to generate the edited image. While these manipulations allow for extensive editing on real images, they suffer from several drawbacks that should be considered. First, most works rely on the existence of a disentangled linear latent space path controlling age, which is hard to achieve in practice. Second, they are unable to directly generate an image at a desired age. Instead, one must manually traverse the fixed latent path in search of the desired age-transformed image.

This differs from our approach which can directly encode a real face image, conditioned on a desired target age, into its corresponding latent representation in the StyleGAN domain. Moreover, while previous works assume a prior on the latent path that can be traversed to control age, we train our network to learn this path with no prior assumptions. By doing so, our method learns a non-linear traversal path that is less sensitive to the entanglement of other attributes in the latent space.  

%% file: method.tex
\section{Method}
\subsection{Overview}
In this section, we present our approach for modeling the age transformation process. Given a source facial image $\textbf{x}$ at age $\alpha_s$ and a desired target age $\alpha_t$, our goal is to transform $\textbf{x}$ to an image $\textbf{x}' = SAM(\textbf{x}, \alpha_t)$ representing the source identity at age $\alpha_t$.

To model the aging process, we introduce a complete image-to-image translation architecture by pairing an encoder network and a \textit{fixed}, pre-trained unconditional image generator. The encoder directly encodes a given image and desired target age to a set of style vectors that capture the desired transformation. Given these style vectors, the generator is then used to generate the desired output image. An overview of our architecture is presented in Figure~\ref{fig:architecture}.

Collecting a series of images of the same person over many years is extremely challenging and therefore we cannot directly rely on pairs of corresponding images. To address this challenge, we apply a cycle consistency loss during training, which has been shown to be effective in unpaired image-to-image translation tasks~\cite{unit, tang2020attentiongan, dualgan, CycleGAN2017}.
To guide the encoder in generating the appropriate style vectors, we utilize a pre-trained age regression network that serves as an additional loss constraint during the training process.

\vspace{-0.2cm}
\subsection{Training}~\label{training}
\begin{figure}
    \centering
    \includegraphics[width=0.95\linewidth]{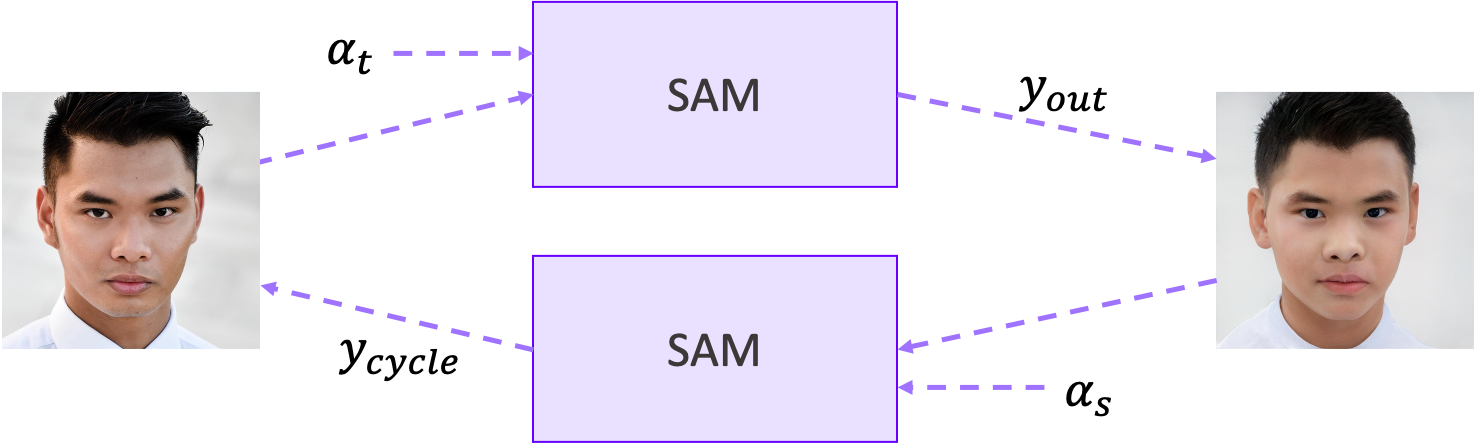}
    \caption{To address the challenge of the unsupervised setting, a cycle consistency pass is performed to recover the input at the source age $\alpha_s$.}
    \label{fig:cycle}
\end{figure}

As the age of a given facial image can be estimated using a well-trained age classifier, only the desired target age must be specified. During training, before feeding an image $\textbf{x}$ through our encoder, the desired target age is randomly generated as
\vspace{-0.1cm}
\begin{equation}
    \alpha_t \sim \mathcal{U}(5, 100).
\end{equation}
That is, a target age between $5$ and $100$ is sampled uniformly at random. The sampled age is then added as a constant-valued channel to the input image, $\textbf{x}$, resulting in a 4-channel input tensor,  which we denote by
\begin{equation}~\label{eq:x_age}
    \textbf{x}_{age} = \textbf{x} \concat \alpha_t. 
\end{equation}
We additionally use a fixed, pre-trained pixel2style2pixel~\cite{richardson2020encoding} encoder trained to encode real face images into the $\mathcal{W+}$ latent space of a pre-trained StyleGAN2 generator. In particular, given an input image $\textbf{x}$, we first compute, 
\begin{equation}~\label{eq:w*}
    \textbf{w}^* := pSp(\textbf{x}) \in \mathbb{R}^{18 \times 512}.
\end{equation}
Using Equations~(\ref{eq:x_age})-(\ref{eq:w*}) we then define the output of our model as
\begin{equation}
    SAM(\textbf{x}_{age}) := G(E_{age}(\textbf{x}_{age}) + \textbf{w}^*)),
\end{equation}
where $E_{age}(\cdot)$ and $G(\cdot)$ denote our aging encoder and the StyleGAN generator, respectively. During training, we perform two passes:
\begin{equation}~\label{eq:forward}
    \textbf{y}_{out} =  SAM(\textbf{x}_{age}),
\end{equation}
\vspace{-0.4cm}
\begin{equation}~\label{eq:cycle}
    \textbf{y}_{cycle} = SAM(\textbf{y}_{out} \concat \alpha_s), 
\end{equation}
where $\textbf{y}_{out}$ is the generated image representing the age transformation. We then apply a cycle consistency pass to recover the original image and set the target age equal to the source age $\alpha_s$ (see Figure~\ref{fig:cycle}).

Observe that we employ two encoders: $pSp(\cdot)$ and $E_{age}(\cdot)$. Here, $pSp(\cdot)$ is used to invert the given image for creating the initial latent embedding of the input. $E_{age}(\cdot)$ is then trained to learn the residual between this initial latent embedding and the latent embedding representing the age-transformed image. Note that the pre-trained pSp encoder remains \textit{fixed} throughout training while $E_{age}(\cdot)$ is trained using the loss objectives described below.

\subsection{Architecture}
The architecture of our aging encoder $E_{age}$ is based on the pSp encoder presented in Richardson \etal ~\shortcite{richardson2020encoding}. Here, we extend it to an unsupervised setting for modeling face aging. 
To form a complete image-to-image translation architecture we combine the encoder with the representative power of a pre-trained StyleGAN~\cite{karras2019style, karras2020analyzing} generator.
Given a 4-channel input described in Section~\ref{training}, the encoder, based on a Feature Pyramid Network~\cite{lin2017feature} architecture, is tasked with generating $18$ unique style vectors corresponding to the $18$ inputs of StyleGAN. 
To do so, the encoder first extracts feature maps at three different spatial scales. Then, $18$ \textit{map2style} blocks, are used to gradually down-sample each feature map to obtain a $512$-dimensional style vector. 
Finally, the $18$ style vectors are fed into the pre-trained StyleGAN generator to obtain the output image representing the desired transformation.
Note that both $pSp(\cdot)$ and $E_{age}$ share the same FPN-based architecture from Richardson \etal ~\shortcite{richardson2020encoding}.

\subsection{Losses}~\label{losses}
Our encoder is trained using a weighted combination of several loss objectives. 
The same set of losses introduced below is used both for the forward pass (Eq.~\ref{eq:forward}) and cycle pass (Eq.~\ref{eq:cycle}). Specifically, the input image $\textbf{x}$ and transformed image $\textbf{y}_{out}$ are compared on the forward pass, while $\textbf{x}$ is compared to the recovered image $\textbf{y}_{cycle}$ in the cycle pass. For conciseness, we describe only the loss objectives for the forward pass.

First, we use the $\mathcal{L}_2$ loss to learn pixel-wise similarities and the LPIPS~\cite{zhang2018unreasonable} loss to learn perceptual similarities:
\begin{equation}
    \mathcal{L_{\text{2}}}\left ( \textbf{x}_{age} \right ) = || \textbf{x} - SAM(\textbf{x}_{age}) ||_2
\end{equation}
\begin{equation}
    \mathcal{L_{\text{LPIPS}}}\left ( \textbf{x}_{age} \right ) = || F(\textbf{x}) - F(SAM(\textbf{x}_{age}))||_2,
\end{equation}
where $F(\cdot)$ denotes the perceptual feature extractor. 
As humans age, their head shape naturally changes over time. Motivated by this, we apply a higher weight on the $\mathcal{L_{\text{2}}}$ and $\mathcal{L_{\text{LPIPS}}}$ losses in the center region of the image and a lower weight on the outer region. 

We additionally adopt a regularization loss~\cite{pbayliesstyleganencoder,richardson2020encoding} that encourages the encoder to output style vectors closer to the average latent vector. We find that this regularization improves image quality by removing unwanted image artifacts without harming the fidelity of the desired age transformation.

\vspace{0.15cm}
\paragraph{Identity Loss.}~\label{identity_loss} 
The authors in~\cite{richardson2020encoding} show the importance of the identity loss in obtaining accurate reconstructions of real facial images. As identity-preservation is a key challenge in modeling the age transformation process, we incorporate this identity loss into our training process by measuring the cosine similarity between the output image and its source image. While distinctive facial features are preserved as we age, our perceived identity may change over time (e.g., a person looks different at age $5$ and age $55$)~\cite{pmid25824013, MILEVA2020101260}. To capture this observation, we compute an identity loss weighted by the change in age such that a larger age change corresponds to a smaller weight. That is, we compute:
\begin{equation}
    \mathcal{L}_{\text{ID}}\left (\textbf{x}_{age} \right ) = w(\Delta_{age}) \cdot (1-\left \langle R(\textbf{x}),R(SAM(\textbf{x}_{age})) \right \rangle ),
\end{equation}
where $R$ is a pre-trained ArcFace~\cite{deng2019arcface} network. 
Estimating the age $\alpha_s$ of the input image using the pre-trained age regression network, we define
$\Delta_{age} = \frac{1}{100}|\alpha_s - \alpha_t |$. The weight function $w(\cdot)$ is then defined by,
\begin{equation}
    w(\Delta_{age}) = 0.25 \cdot \cos({\pi \cdot \Delta_{age}}) + 0.75.
\end{equation}
Here, a weight of $1$ corresponds to no change in age, while an increase in $\Delta_{age}$ corresponds to a monotonic decrease in $w(\cdot)$ with a minimum weight of $0.5$. The motivation behind the weighted identity loss is further discussed in Section~\ref{age-invariance}.

\input{figures/experiment_results_teaser}

\vspace{0.15cm}
\paragraph{Aging Loss.} To measure the accuracy of the age transformation, we use a pre-trained age predictor network~\cite{Rothe-ICCVW-2015, Rothe-IJCV-2018}, denoted $A$. Given an input image $\textbf{x}$ and desired target age $\alpha_t$ the aging loss is  computed as the $\mathcal{L}_2$ loss between $\alpha_t$ and the age of the generated image,
\begin{equation}
    \mathcal{L_{\text{age}}}\left (\textbf{x}_{age} \right ) = || \alpha_t - A(SAM(\textbf{x}_{age})) ||_2.
\end{equation}

\vspace{0.5cm}
In summary, the objective of the forward pass is given by: 
\vspace{0.15cm}
\begin{align}
\begin{split}
\small
    \mathcal{L}_{forward}(\textbf{x}_{age}) = {}& \lambda_{l2} \mathcal{L}_2(\textbf{x}_{age}) +
    \lambda_{lpips} \mathcal{L}_{\text{LPIPS}}(\textbf{x}_{age}) + \\
    {}& \lambda_{reg} \mathcal{L}_{\text{reg}}(\textbf{x}_{age}) +
    \lambda_{id} \mathcal{L}_{\text{ID}}(\textbf{x}_{age}) + \\
    {}& \lambda_{age} \mathcal{L}_{\text{age}}(\textbf{x}_{age})
\end{split}
\end{align}
where $\lambda_{l2}$, $\lambda_{lpips}$, $\lambda_{reg}$, $\lambda_{id}$, $\lambda_{age}$ are constants defining the loss weights.
Combining both forward and cycle passes, the full objective is then given by, \vspace{0.15cm}
\begin{align}
\begin{split}
\small
    \mathcal{L}(\textbf{x}_{age}, \textbf{y}_{out} \concat \alpha_s) = {}& \mathcal{L}_{forward}(\textbf{x}_{age}) + \\
                                                                       {}& \lambda_{cycle} \mathcal{L}_{cycle}(\textbf{y}_{out} \concat \alpha_s),
\end{split}
\end{align}
where $\lambda_{cycle}$ defines the weight of the cycle loss. Recall that in the cycle pass the recovered image, $SAM(\textbf{y}_{out} \concat \alpha_s)$, is compared to the original input image $\textbf{x}$. Additional implementation details and loss weights can be found in Appendix~\ref{implementation_details}.

%% file: figures/experiment_results_teaser.tex
\begin{figure*}
    \centering
    \includegraphics[width=0.975\linewidth]{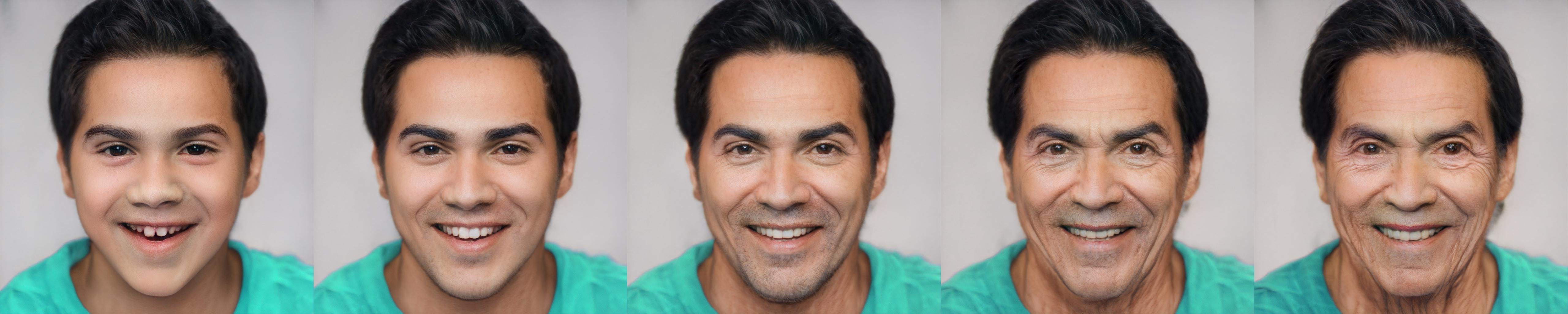} \\
    \includegraphics[width=0.975\linewidth]{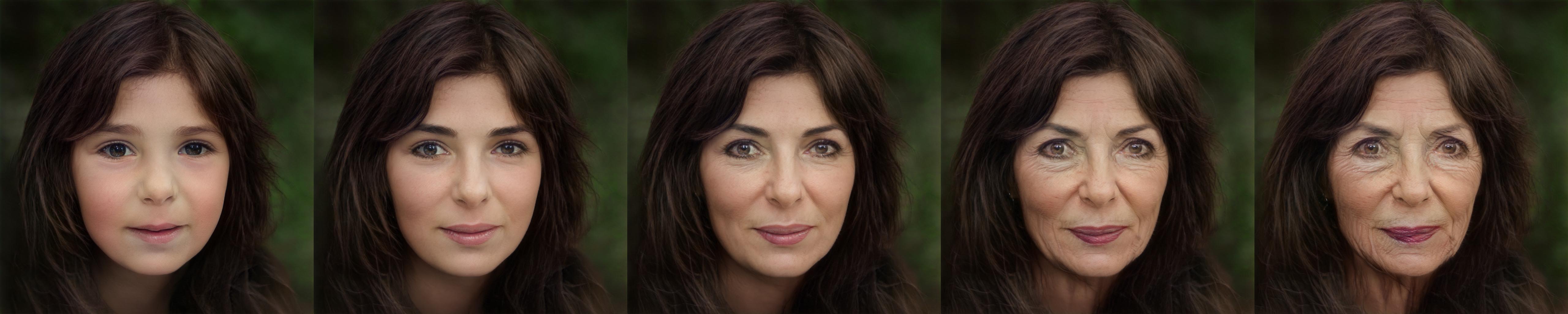}
    \caption{Aging results generated using SAM. Observe the state-of-the-art image quality achieved by leveraging a fixed, pretrained StyleGAN generator.}
    \label{fig:limitations}
\end{figure*}

%% file: results.tex
\section{Experiments}~\label{experiments}

In this section, we perform extensive experimentation to explore the effectiveness of our method. In particular, we compare our approach to state-of-the-art age transformation and latent space methods. We then show the benefits of using our end-to-end approach for learning the latent path controlling age. All evaluations are conducted on the CelebA-HQ~\cite{karras2017progressive} test dataset. 

A comparison to multi-domain image-to-image translation approaches as well as an ablation study of SAM can be found in the Appendix~\ref{multi_domain_comp} and Appendix~\ref{ablation}. Full lifespan animations and an accompanying video are also provided in the project page. Please note that due to the licensing agreement of the CelebA~\cite{liu2015faceattributes} dataset, we cannot present the original input images. As an alternative, we display each image's inversion obtained using the pSp encoder~\cite{richardson2020encoding}.

Effectively evaluating an age transformation method is extremely challenging. For a given input image, we wish to generate an image that faithfully resembles the same individual. At the same time, however, it is necessary to perform meaningful changes to the input image that accurately reflect the complex human aging process. 
Therefore, before evaluating the various aging methods, we explore the effectiveness of current state-of-the-art facial recognition networks in recognizing the same individual at different ages.
In doing so, we show that measuring identity preservation of aging methods should be done with great care.

\input{figures/human_aging_identity}

\input{figures/aging_methods_comparison}

\subsection{The Age-Invariance of Facial Recognition Networks}~\label{age-invariance}
Numerous experimental psychological studies~\cite{MILEVA2020101260, pmid25824013, JENKINS2011313, doi:10.1080/713756750} have shown that although people typically recognize individuals they know across their lifetime, current facial recognition systems struggle in doing so. 
We perform two experiments to evaluate the age-invariance of the current state-of-the-art facial recognition network. First, we collect $50$ pairs of individuals at different ages.
We then use the ArcFace~\cite{deng2019arcface} recognition network to compute the cosine similarity of each pair of images. We obtain an average similarity score of $0.45$ with a standard deviation of $\pm0.10$. Next, we collect multiple images of the same individual at different ages. Setting aside one of the images as a query image, we measure the cosine similarity of the remaining images to the query. As shown in Figure~\ref{fig:real_aging_identity}, as we move away from the age of the query, the similarity monotonically decreases, indicating that identity, as measured by facial recognition systems, does not remain fixed with age.

We conclude that relying on facial recognition systems may be ineffective when faced with a large variations in age. Following the above experiments and the conclusions from~\cite{JENKINS2011313, pmid25824013, MILEVA2020101260}, we find that currently, the most effective approach for quantitatively measuring identity across one's lifespan is through human perceptual evaluation. Moreover, such an evaluation should be performed on well-known individuals as viewers are significantly less accurate in identifying unfamiliar faces~\cite{Megreya2006, Megreya2008, doi:10.1068/p3335}. In addition, motivated by these findings, we employ a weighted identity loss to model this observation (see Section~\ref{losses}).

\vspace{-0.4cm}
\subsection{Comparison with Age Transformation Methods}~\label{compare_aging_methods}
We begin our evaluation by comparing our proposed method to two state-of-the-art age transformation methods from Or-El \etal~\shortcite{orel2020lifespan} and Yao \etal~\shortcite{yao2020high}, which we refer to as \textit{LIFE} and \textit{HRFAE}, respectively. 
For both LIFE and HRFAE we use their official implementations and pre-trained models for evaluations.
Note that each method supports a different age range ($0-70$ for LIFE, $20-70$ for HRFAE, and $5-100$ for SAM). Therefore, when comparing two methods, we consider only ages supported by both methods.

\paragraph{Qualitative Evaluation.}
We provide a visual comparison with LIFE in Figure~\ref{fig:comparison_lifespan}. One can see that although LIFE is able to successfully capture the change in head shape across the different ages (see row $2$), the resulting images contain unwanted artifacts.
We observe the ability of SAM to more naturally alter facial features that change as we age, e.g., facial hair and wrinkles (see row $2$). In Figure~\ref{fig:comparison_hrfae} we compare our method with HRFAE. 
While HRFAE is able to generate high-resolution images, they are limited to generating subtle changes in texture between the different ages due to the wide bottleneck present in their architecture. In contrast, our approach is able to achieve visually pleasing results, while better modeling the change in head shape and texture. For example, observe the change in the jaw line of SAM's outputs. 

Notice that all methods generally focus on modeling changes in the face regions. For example, all methods struggle in modeling changes in hair color and capturing receding hair lines with the progression of age.
Specifically, we note that while SAM preserves various features (e.g., hair color) across the different target ages, this may not always be desirable, since often an individual's hair color change over time.
To address this, we can apply simple editing techniques to generate multiple outputs for a single input image illustrating these possible changes over time. These techniques are explored in Section~\ref{additional_editing}.

\input{figures/quantitative_aging}

\paragraph{Quantitative Evaluation.}
We note that since LIFE operates using pre-defined age groups, we are unable to compare with it by directly translating images to specific target ages. However, LIFE is able to interpolate between the different age groups to illustrate a continuous age progression.
Although this interpolation can generate a full lifespan of images, the exact age of each can not be guaranteed. 
Therefore, we choose to generate a full set of $80$ images for each source image and select the generated image whose predicted age is closest to a desired target age. 
This process is repeated for multiple target ages.
We note that although HRFAE and SAM generate images using a specified target age, for a fair comparison of the three methods, we use the same selection protocol for all three. 
Moreover, for each method we select images only within the method's supported age range.

Having performed the above selection process, we now evaluate each method's ability to accurately generate images for a wide-range of ages. To do so, for each target age, we measure the average difference between the target age and the predicted ages of the images chosen by the above selection process.
To ensure the predicted ages are independent of our losses, we use the Microsoft Azure Face API for age estimation.
All metrics are computed using $1,000$ test samples taken from the CelebA-HQ~\cite{karras2017progressive} test dataset.

Note that approximately
$80\%$ of the test set falls within the $20-40$ age range. We therefore expect the age difference to increase as we move away from this range.
Figure~\ref{fig:aging_accuracy} presents the comparison of the aging accuracy of the three methods. One may notice LIFE's ability to more accurately generate images with a target age of $5-10$. We note this to be a limitation of SAM, which is discussed further in Section~\ref{limitations}.
Nevertheless, for the remaining target ages we find that SAM out-performs both LIFE and HRFAE. This difference is most notable when translating to the older age range ($60+$). 

\paragraph{Human Evaluation.}
Following the observations in Section~\ref{age-invariance}, to more reliably quantify the performance of each method evaluated above, we also perform a human evaluation. 
For measuring the aging accuracy, we show the outputs of the three methods side-by-side and ask which method best portrays an individual at the desired target age. Similarly, to measure overall image quality, we ask the worker to select the output that is most visually appealing. For a complete analysis, we repeat the above scenario three times and translate the test images to target ages of $5$, $30$, and $65$. The results are shown in Table~\ref{tb:human_evaluation}. For each of the three target ages, a total of $150$ responses were recorded for each of the two aspects (i.e. $300$ responses for each target age). Note, all images were randomly selected for this evaluation.

To evaluate whether the identity of the images generated by the three methods are well-preserved and recognizable, we take a different approach. Here, we collect images of well-known celebrities and perform age transformation to multiple target ages using the three methods. We then ask each worker to identify the individual shown in the transformed image. To increase the difficulty, the question is asked as an open question (i.e., the worker is not given any choices to select from). In each cell of Table~\ref{tb:human_evaluation_identity} we show the percent of queries that were correctly identified out of the $150$ responses for the corresponding method and target age. Taking into account that some respondents simply do not know the queried individual, we find that the three methods are similar in their ability to faithfully preserve the input identity.

\input{figures/human_evaluation}

\input{figures/human_evaluation_identity}

\subsection{Comparison with Latent Space Methods}
With the recent emergence of strong image synthesis models such as StyleGAN, latent space traversal has become a popular approach for editing on real images. While these works have shown potential, they typically assume some prior, such as linearity, on the latent path controlling an attribute of interest.
Previous works~\cite{karras2019style, shen2020interpreting, collins2020editing, yang2020semantic} have demonstrated the disentanglement of StyleGAN's latent space. However, since this latent space is a manifold, a path that controls a specific attribute is not necessarily linear. It is also not guaranteed that such a linear path remains within the manifold. Such approaches may therefore fail to fully capture the disentanglement property of the latent space and may generate samples outside the true data distribution.

\input{figures/interfacegan_comparison}

In contrast, we train an encoder to directly embed real images into the latent space of a pre-trained StyleGAN. As a result, one may view our method as an approach to \textit{learn} a latent space path without explicitly needing to model it. Here, we show the advantages of using an end-to-end approach for learning the latent path without requiring any prior assumptions (see Figure~\ref{fig:pca_latent_path}).

\input{figures/latent_path}

We compare our results with two state-of-the-art latent space approaches: InterFaceGAN~\cite{shen2020interpreting} and StyleFlow~\cite{abdal2020styleflow}. InterFaceGAN assumes that the latent path controlling a given attribute (e.g., age) is linear. StyleFlow learns non-linear paths in the latent space by using normalizing flows conditioned on the attribute of interest. We show that compared to both of these methods that operate specifically in the latent path, our end-to-end approach obtains improved disentanglement and superior visual quality on real facial images.

\paragraph{Qualitative Evaluation.}
As InterFaceGAN is originally trained using StyleGAN1 generators, we retrain their method using the same StyleGAN2 generator used by our method and follow the same training procedure as described in their paper to obtain the linear direction. 
For StyleFlow, we use their official implementation for generating the transformed images. To edit a given \textit{real} image, we first encode the image into the $\mathcal{W+}$ latent space using pSp~\cite{richardson2020encoding}, the state-of-the-art StyleGAN encoder, and then interpolate the obtained latent code using the latent direction of each method. A visual comparison is provided in Figure~\ref{fig:interface_comparison}.

For completeness, we provide additional results for InterFaceGAN in Appendix~\ref{additional_results} using their official implementation and age boundary. As their original paper and official implementation uses a StyleGAN1~\cite{karras2019style} generator, we use IDInvert~\cite{zhu2020domain} for inverting the real input image.

Although InterFaceGAN is able to generate high-quality images, it can be seen that they fail to disentangle age and other facial attributes. For example, one can observe that eye glasses and hair color are heavily entangled with the progression in age.
This may be explained by the bias of the training data as older people are more likely to wear glasses. This potential data bias, coupled with the linearity of their approach, results in poor lifelong age transformation results. Notice that while StyleFlow obtains better disentanglement of age compared to InterFaceGAN, they struggle to faithfully transform the input images into the older age range ($60+$) and obtain lower-quality results.
In contrast, our approach achieves superior disentanglement between age and other facial attributes while faithfully preserving identity across various ages.

\paragraph{Exploring the Latent Path.}
To better understand the latent path learned by our approach, we use PCA to project the $\mathcal{W+}$ age-transformed latent codes obtained for various real face images. To emphasize the non-linearity of the learned paths, we compute the projection plane using the age-transformed latent codes of a single input image.
As can be seen in Figure~\ref{fig:pca_multiple_paths}, the paths of different images are similar, strengthening the claim that the latent space of StyleGAN is a well-behaved manifold that can be traversed to edit a particular attribute of interest (e.g. age) in a disentangled manner.
Note that while these paths are similar, they are each different in nature, illustrating the non-trivial solution learned by SAM.

Although the StyleGAN latent space is well-behaved, we observe that a fully disentangled path is hard to achieve when trying to explicitly model it using some prior on the latent path. To show this, we project the age transformation latent paths learned by InterFaceGAN and SAM for the same real input face image. As can be seen in Figure~\ref{fig:pca_latent_path_interfacegan}, the path learned by InterFaceGAN is indeed linear and results in a strong entanglement between age and other facial attributes. Also, notice that explicitly modeling the learned path using our approach is not trivial as the path's behavior changes at different parts of the age progression.

\begin{figure}
    \centering
    \includegraphics[width=0.925\linewidth]{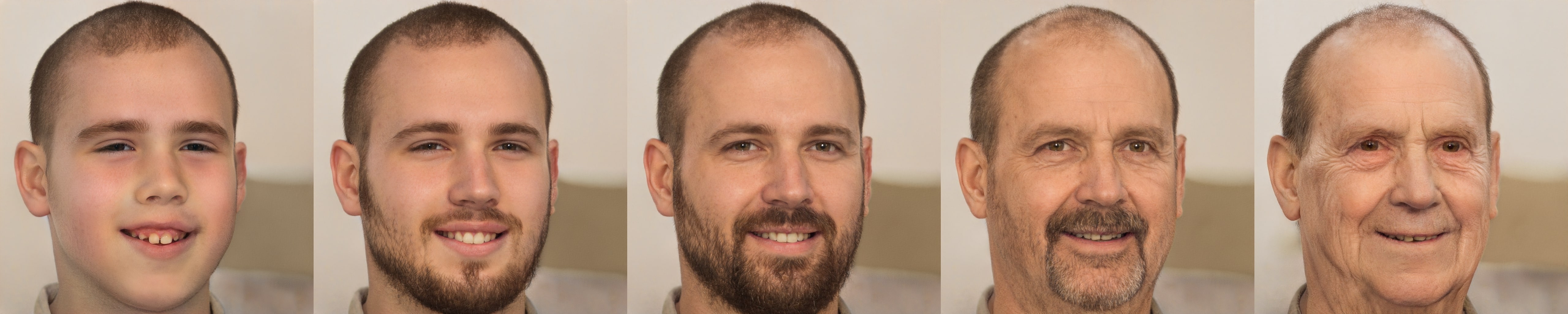} \\
    \includegraphics[width=0.925\linewidth]{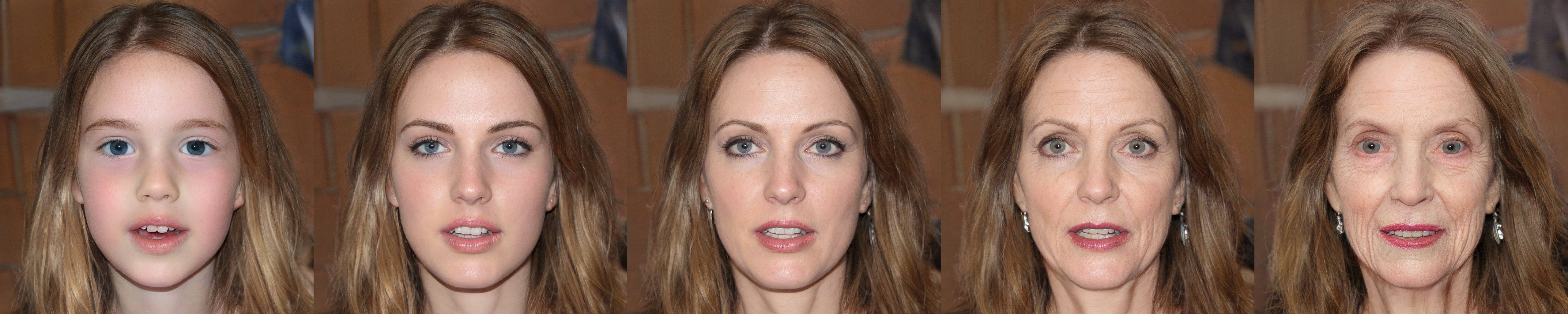} \\
    \includegraphics[width=0.925\linewidth]{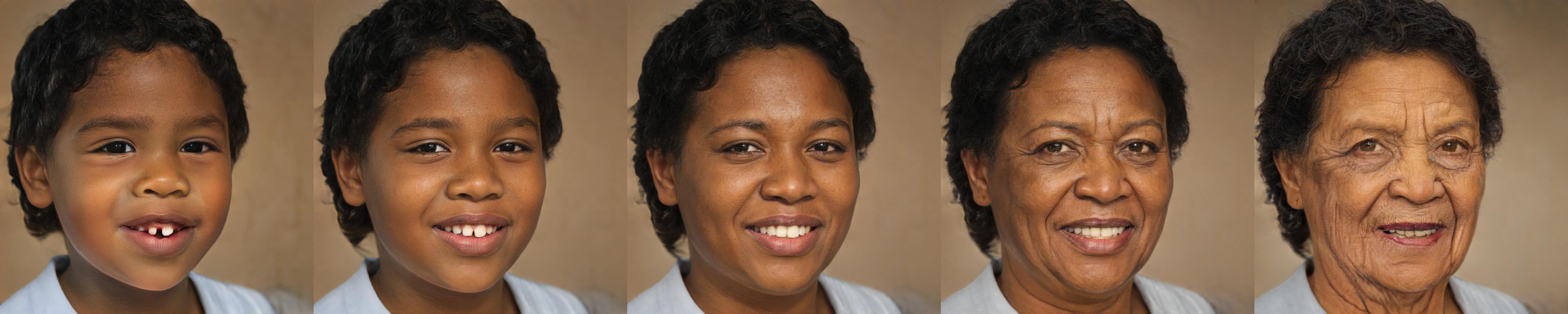}
    \setlength{\belowcaptionskip}{-10pt}
    \caption{Images generated from randomly sampled $w$ vectors that are traversed along the age manifold direction learned by SAM.}
    \label{fig:w_traversal}
\end{figure}

To validate the learned manifold direction that controls age, we randomly sample an image from the $\mathcal{W}$ latent space of StyleGAN and perform age transformation by traversing along a path lying on the learned manifold.
We illustrate several examples in Figure~\ref{fig:w_traversal}. As can be seen, we get a smooth and disentangled age progression, which shows the generalization of our method in learning a latent path corresponding to an age shift. This may raise a question as to the benefits of our end-to-end approach. Although it is possible to operate solely in the latent space and apply a ``general" learned path to a given image, the end-to-end nature of SAM allows one to achieve more fine-grained control over the resulting age. For example, with SAM one can directly specify the desired target age, which cannot be done when operating solely in the latent space. 

\input{figures/patch_editing}

\subsection{Additional Editing}~\label{additional_editing}
As demonstrated in the previous section, our method is able to successfully disentangle age from other facial attributes. That is, all attributes except for age are faithfully preserved from the input image. As we age, however, attributes such as hair style and hair color naturally change (e.g., the hair of most individuals naturally becomes gray or recedes with age). 
It is therefore desirable to be able to control such attributes when modeling age progression. Here, we show that by using simple editing techniques, SAM provides additional control over the resulting image.

\paragraph{Patch Editing.}
Patch editing provides a simple and intuitive editing approach where one can ``paste'' a particular attribute such as glasses, facial hair, and bangs onto the input source image. The edited image is then passed through SAM to obtain the age-transformed edited image. As shown in Figure~\ref{fig:patch_editing}, SAM is able to seamlessly fuse the patch into the source image, while preserving the aging accuracy of our approach. The ability of SAM to successfully perform this task lies in the end-to-end nature of our approach. In particular, given some source image, edited or not, our encoder is tasked with encoding the input into a latent code of a realistic face image. Therefore, even when the patch is imprecisely pasted, SAM is able to merge the crop into its surrounding context while transforming the input image to the desired age.

\input{figures/style_mixing}

\paragraph{Style Mixing.}
While patch editing allows for fine-grained control over specific attributes, it is typically limited to local edits of the input image. For example, controlling lighting and hair color is harder to achieve through patch editing due to the more global nature of the required edit. Since our approach maps a given input image into StyleGAN's $\mathcal{W}+$ latent space, we can leverage the inherent editing capabilities it offers. As shown in~\cite{karras2019style}, the fine styles mostly control the lighting and color of the generated image. Motivated by this, after transforming a given source image to the desired target age, we can additionally perform style-mixing on layers $8-9$ with a given reference image. Multi-modal synthesis is then naturally supported by performing style-mixing on multiple reference images. As shown in Figure~\ref{fig:style_mixing} doing so enables fine-grained control in generating multiple plausible age transformation results.

%% file: figures/human_aging_identity.tex
\begin{figure}
    \setlength{\tabcolsep}{1pt}
    \centering
        \begin{tabular}{c c c c c}
        \includegraphics[width=0.085\textwidth]{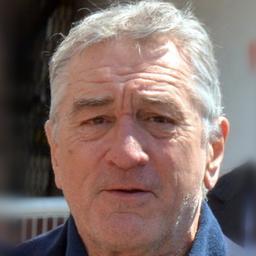} &
        \includegraphics[width=0.085\textwidth]{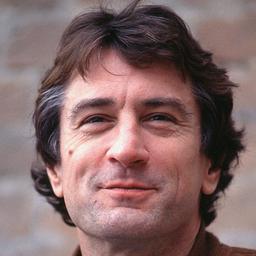} &
        \includegraphics[width=0.085\textwidth]{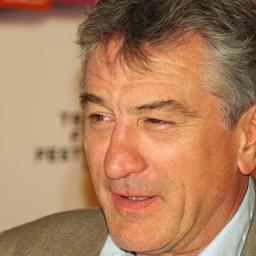} &
        \includegraphics[width=0.085\textwidth]{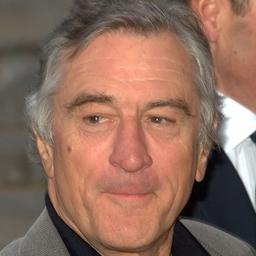} &
        \includegraphics[width=0.085\textwidth]{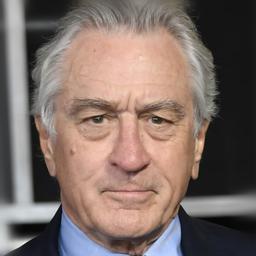}
        \tabularnewline
        Query & $0.505$ & $0.503$ & $0.571$ & $0.619$ \\
        2016 & 1990 & 2008 & 2010 & 2019 \\
        \includegraphics[width=0.085\textwidth]{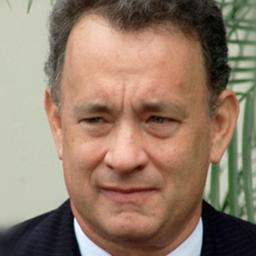} &
        \includegraphics[width=0.085\textwidth]{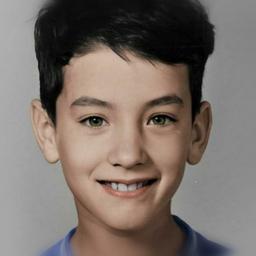} &
        \includegraphics[width=0.085\textwidth]{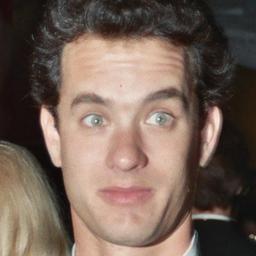} &
        \includegraphics[width=0.085\textwidth]{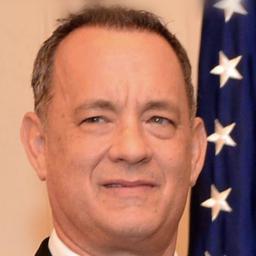} &
        \includegraphics[width=0.085\textwidth]{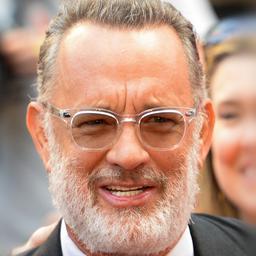}
        \tabularnewline
        Query & $0.378$ & $0.464$ & $0.724$ & $0.517$ \\
        2012 & Childhood & 1989 & 2014 & 2019 \\
        \end{tabular}
    \vspace{-0.2cm}
    \setlength{\belowcaptionskip}{-12pt}  
    \caption{Identity similarity results using the ArcFace~\cite{deng2019arcface} recognition network. For each row, we compute the cosine similarity between the query and the remaining images. Image credits in order:
    \cite{deniro_1},\cite{deniro_3},\cite{deniro_2008},\cite{deniro_4},\cite{deniro_5},\cite{hanks_1},\cite{hanks_2},\cite{hanks_3},\cite{hanks_4},\cite{hanks_5} 
    }
    \vspace{-0.4cm}
    \label{fig:real_aging_identity}
\end{figure}

%% file: figures/aging_methods_comparison.tex
\begin{figure*}
    \centering
    \setlength{\belowcaptionskip}{-7.5pt}
    \begin{subfigure}{0.5\textwidth}
        \setlength{\tabcolsep}{1pt}
        \centering
            \begin{tabular}{c c c c c c c c c}
            Inversion & & & 3-6 & 7-9 & 15-19 & 30-39 & 50-69 \\
            \includegraphics[width=0.13\textwidth]{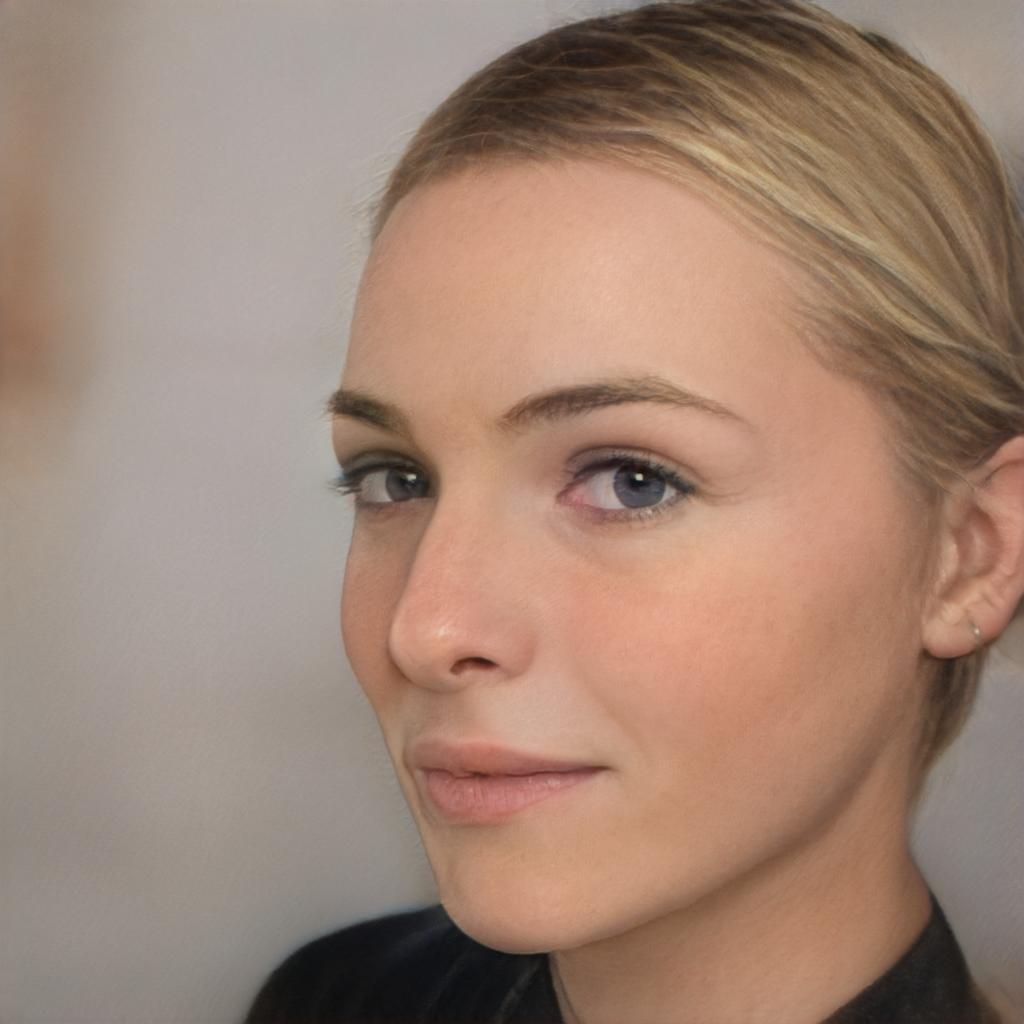} &
                  & \raisebox{0.15in}{\rotatebox[origin=t]{90}{LIFE}} & 
                    \includegraphics[width=0.13\textwidth]{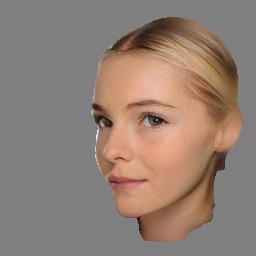} &
                    \includegraphics[width=0.13\textwidth]{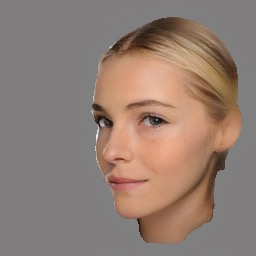} &
                    \includegraphics[width=0.13\textwidth]{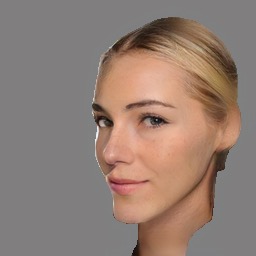} &
                    \includegraphics[width=0.13\textwidth]{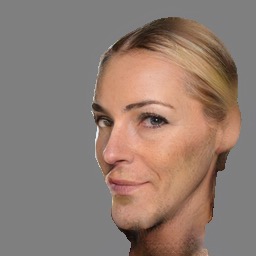} &
                    \includegraphics[width=0.13\textwidth]{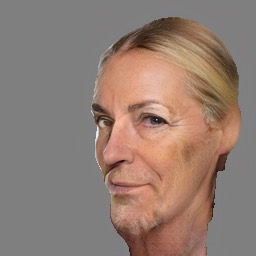} \\
                  & & \raisebox{0.15in}{\rotatebox[origin=t]{90}{SAM}} &
                    \includegraphics[width=0.13\textwidth]{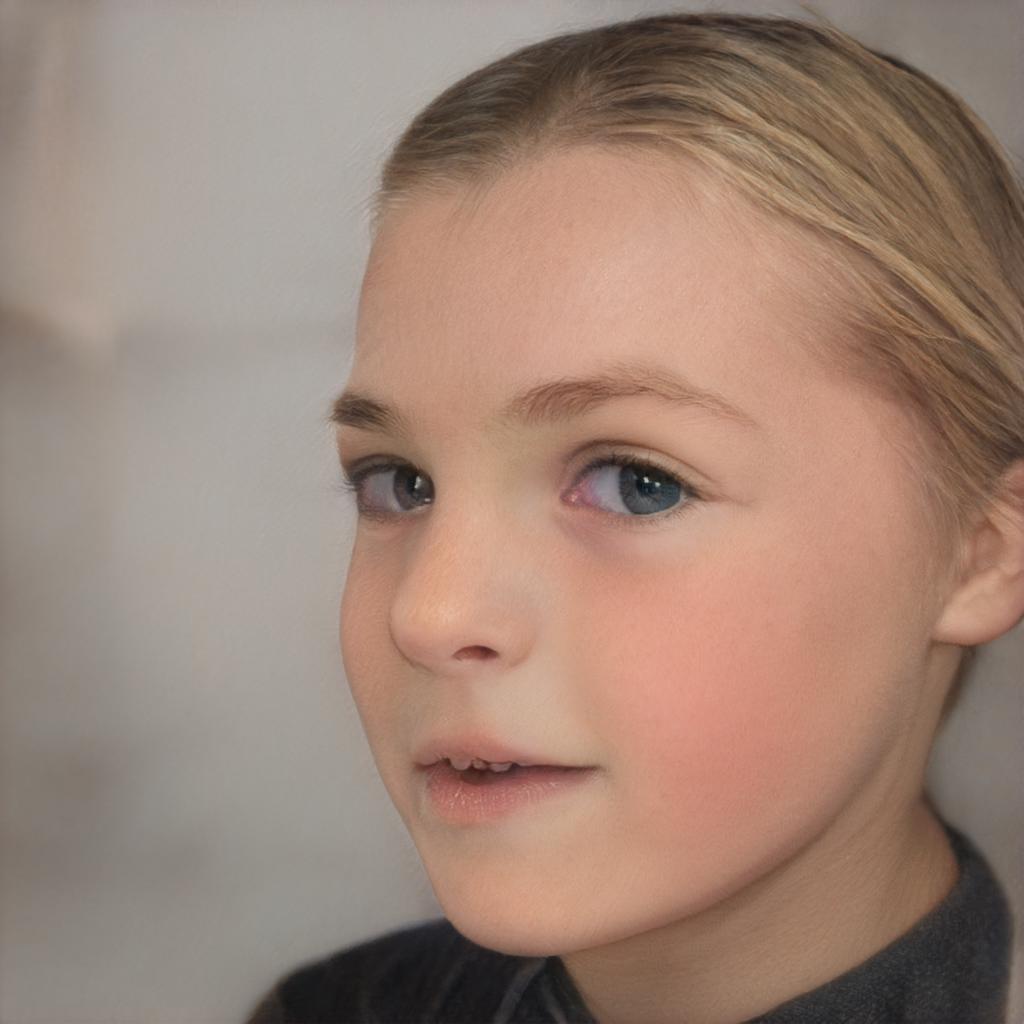} &
                    \includegraphics[width=0.13\textwidth]{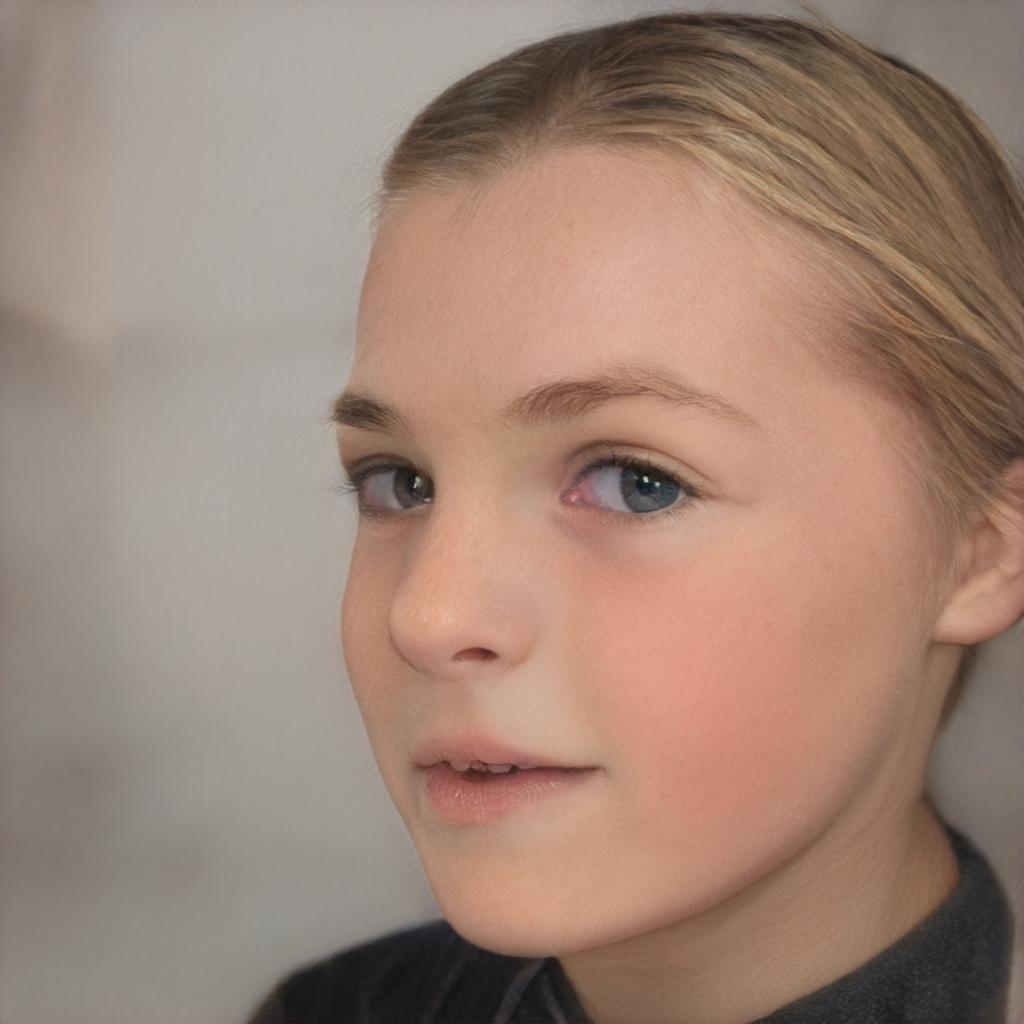} &
                    \includegraphics[width=0.13\textwidth]{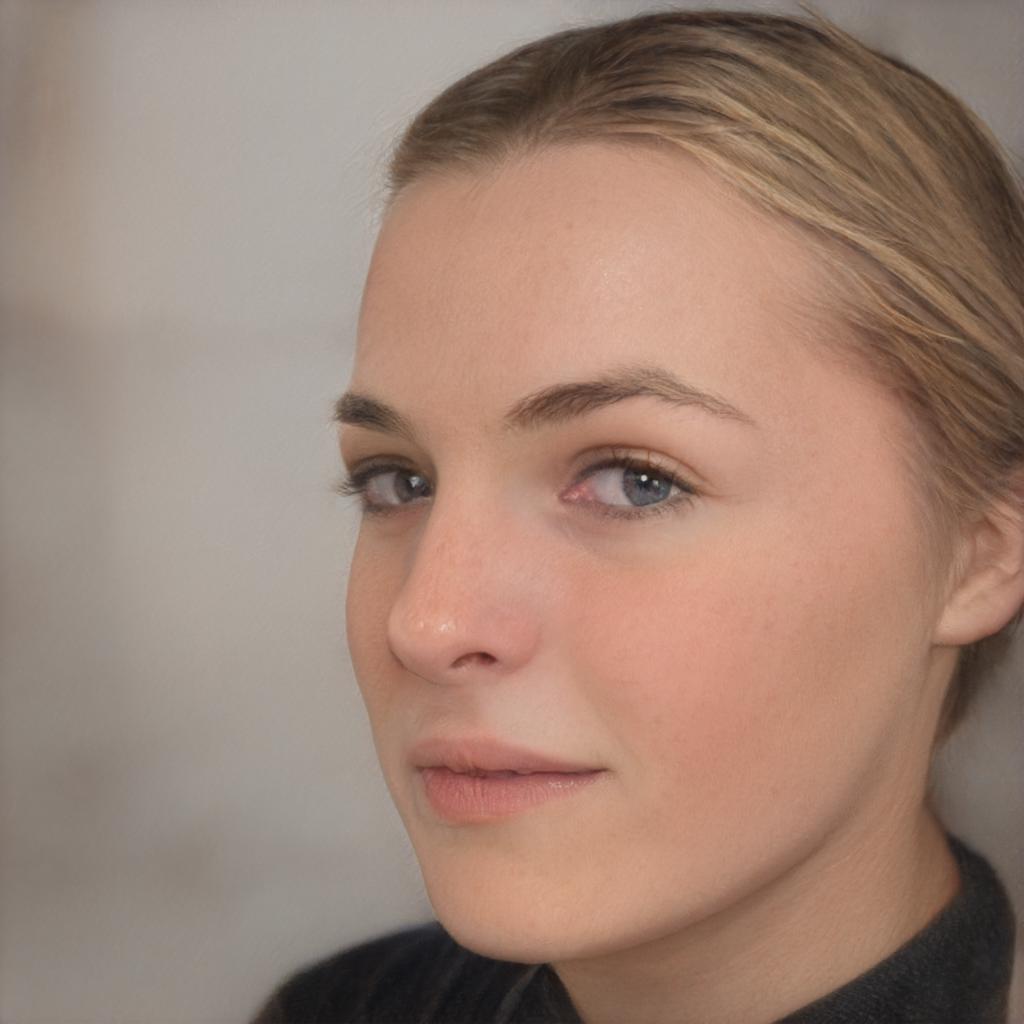} &
                    \includegraphics[width=0.13\textwidth]{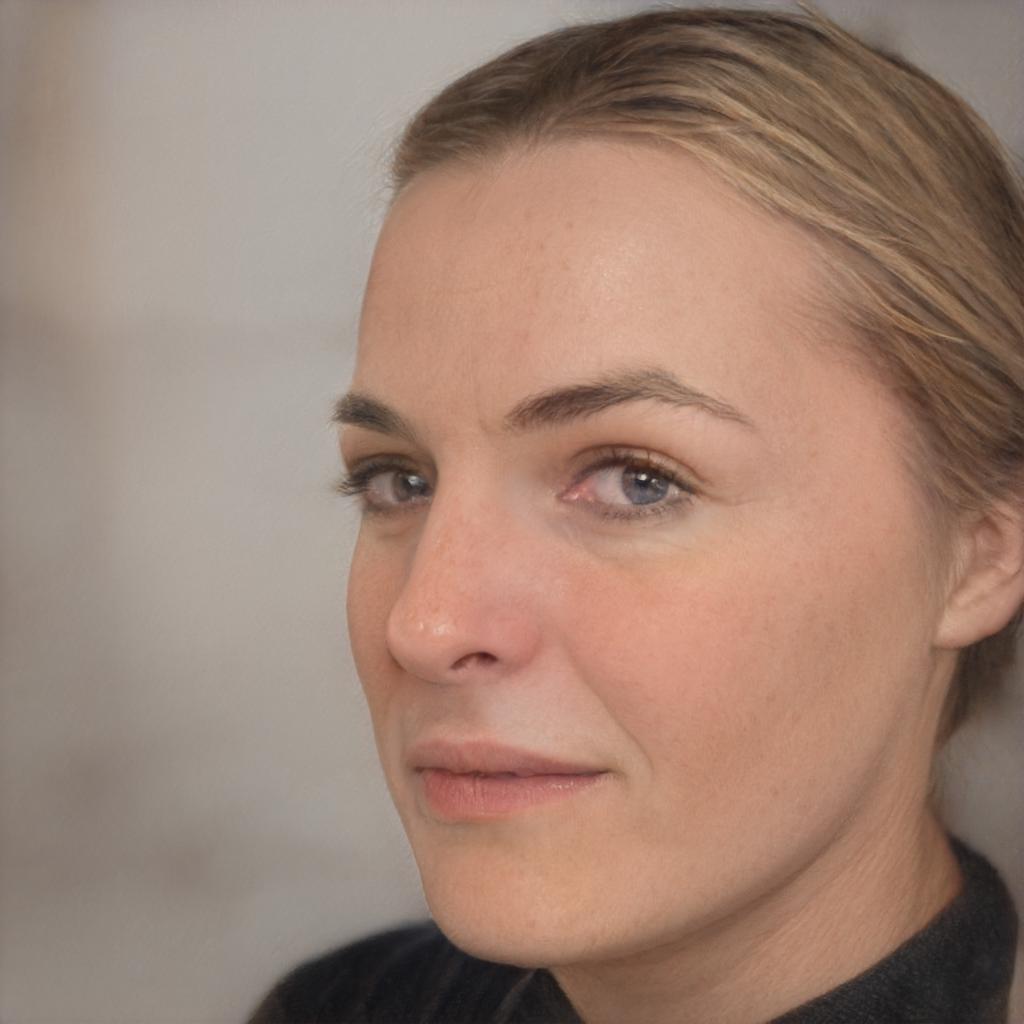} &
                    \includegraphics[width=0.13\textwidth]{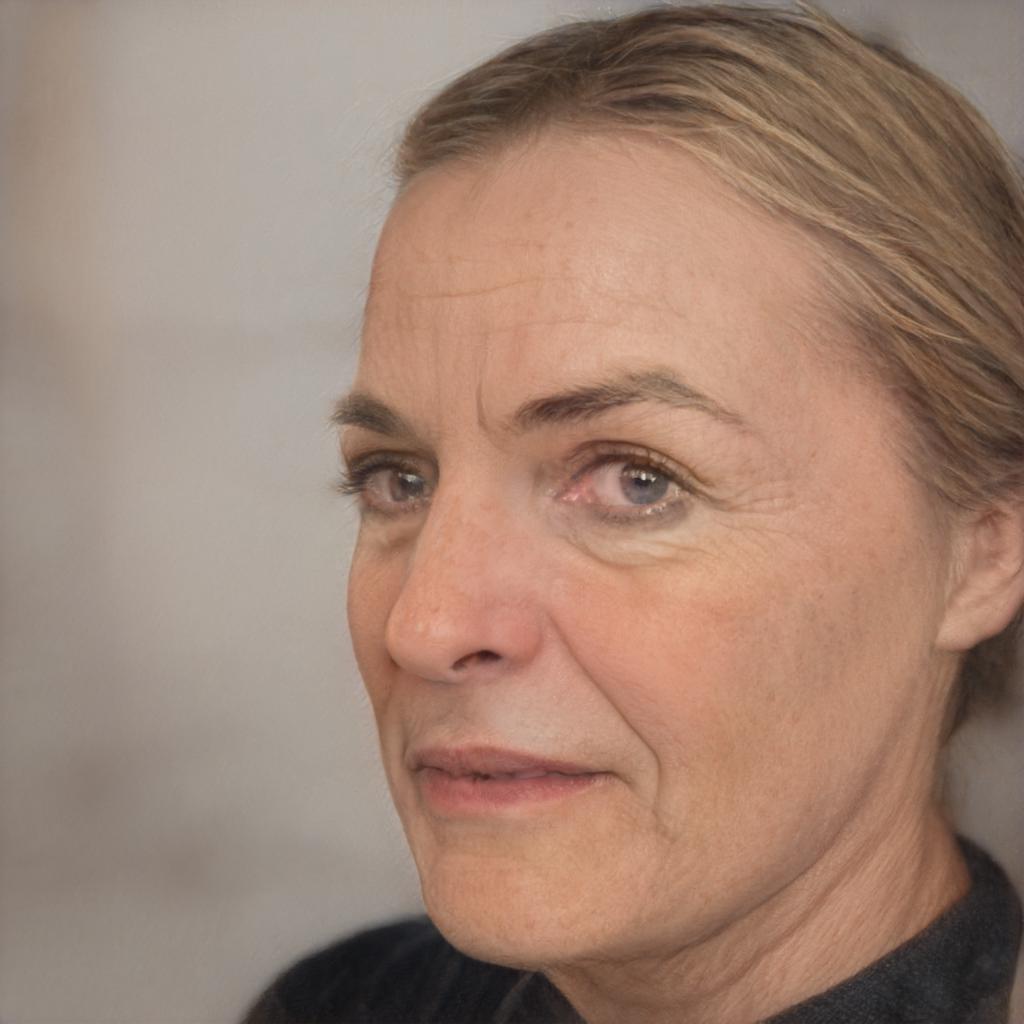}
            \tabularnewline
    
            \includegraphics[width=0.13\textwidth]{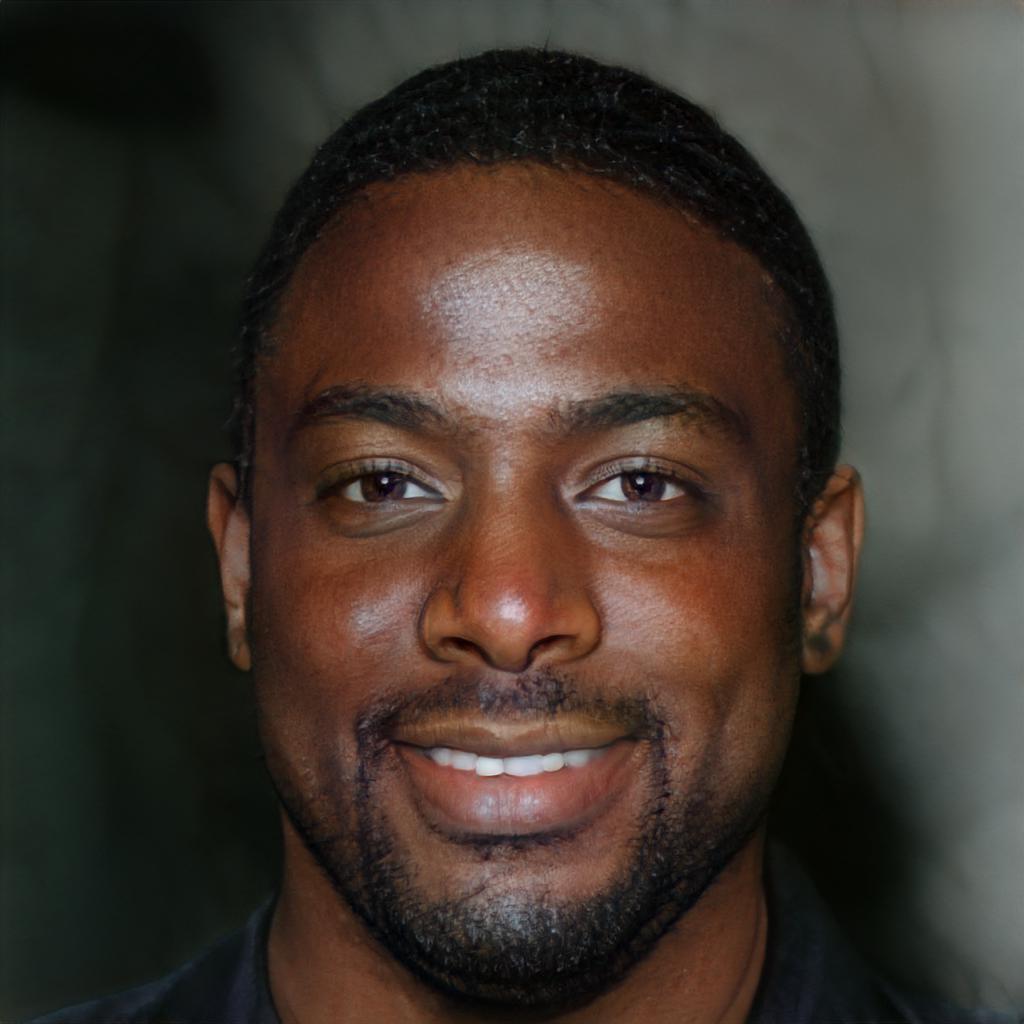} &
                  & \raisebox{0.15in}{\rotatebox[origin=t]{90}{LIFE}} & 
                    \includegraphics[width=0.13\textwidth]{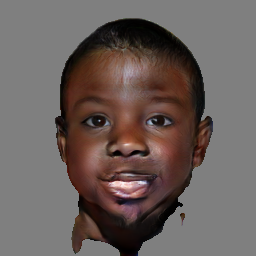} &
                    \includegraphics[width=0.13\textwidth]{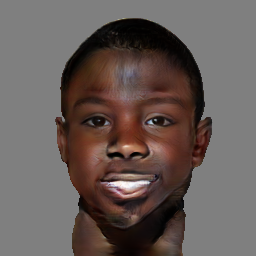} &
                    \includegraphics[width=0.13\textwidth]{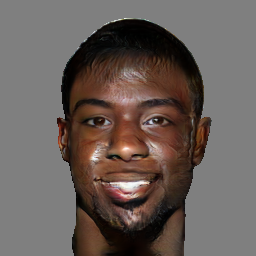} &
                    \includegraphics[width=0.13\textwidth]{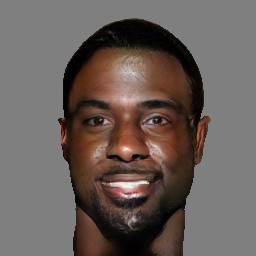} &
                    \includegraphics[width=0.13\textwidth]{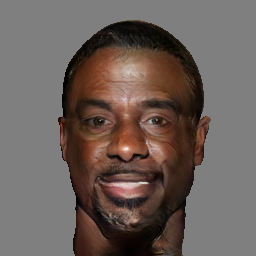} \\
                  & & \raisebox{0.15in}{\rotatebox[origin=t]{90}{SAM}} &
                    \includegraphics[width=0.13\textwidth]{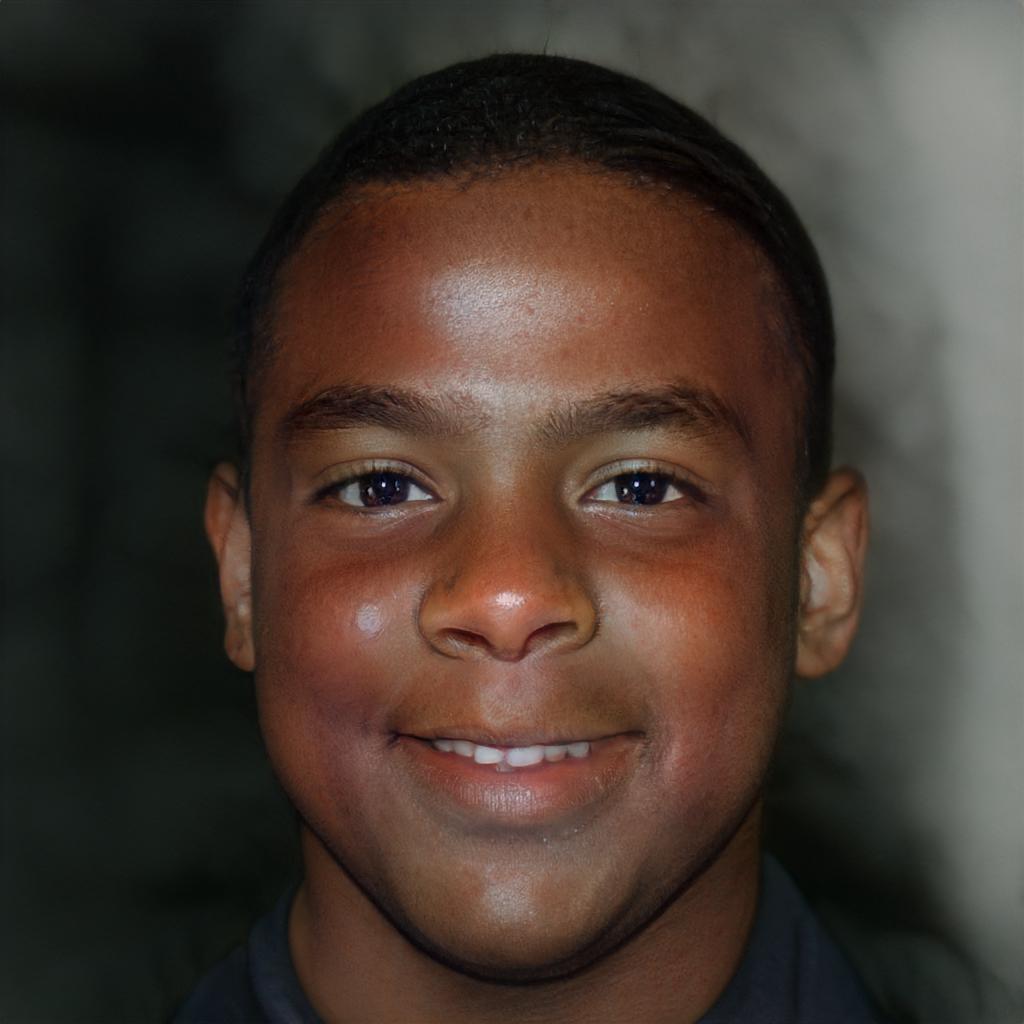} &
                    \includegraphics[width=0.13\textwidth]{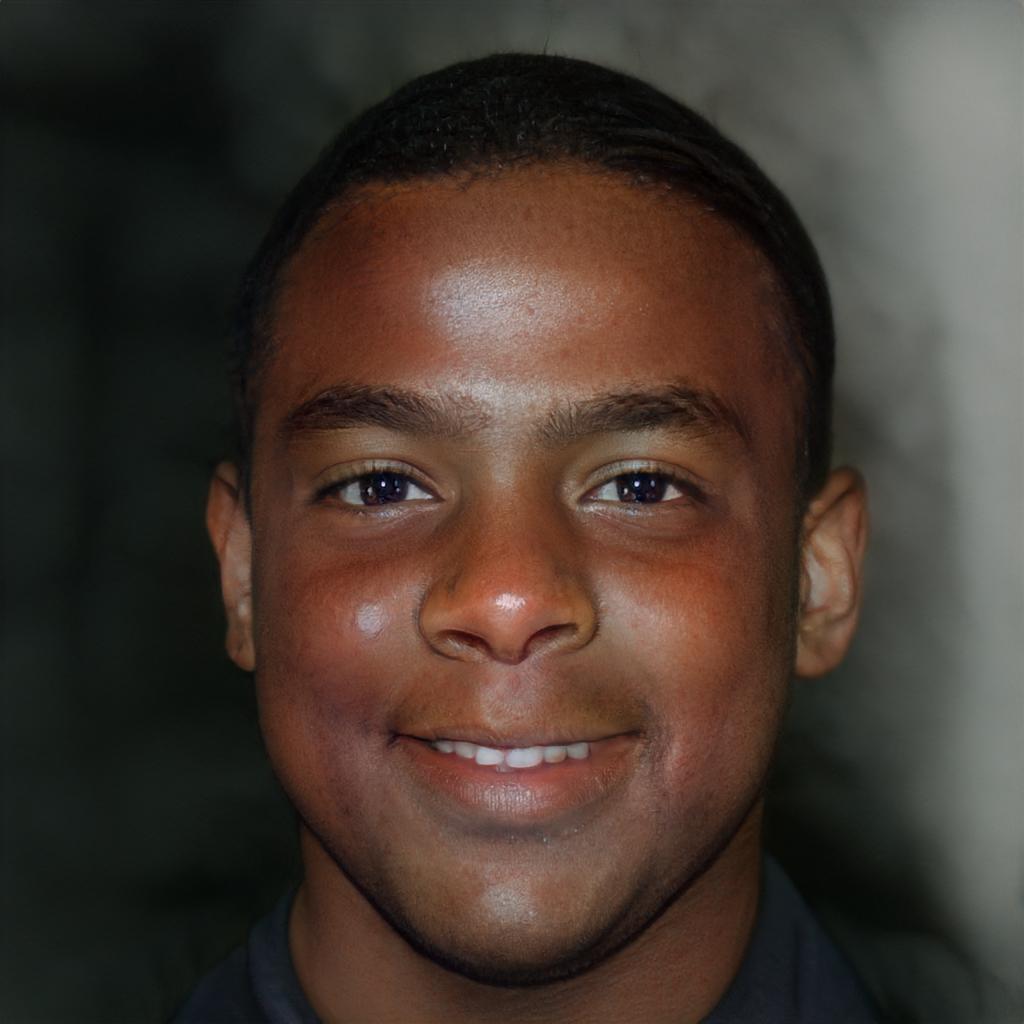} &
                    \includegraphics[width=0.13\textwidth]{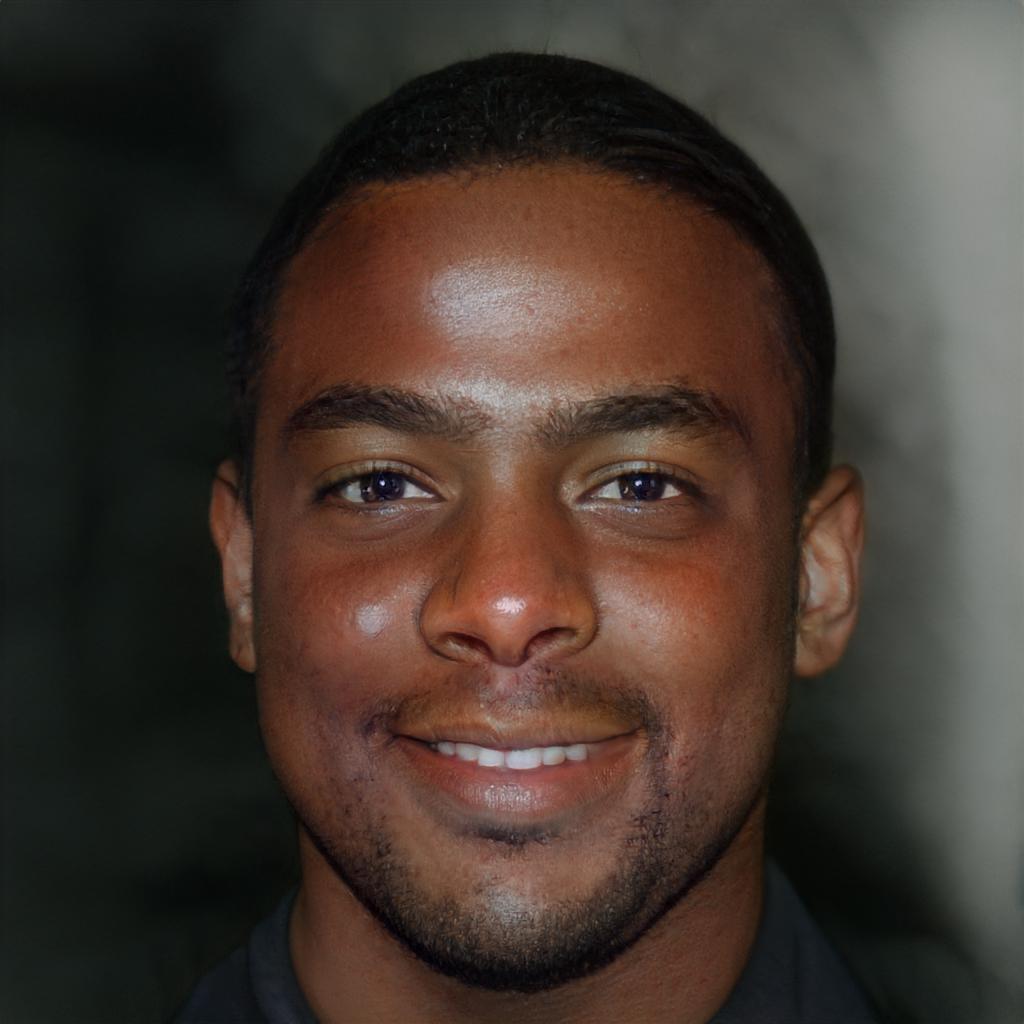} &
                    \includegraphics[width=0.13\textwidth]{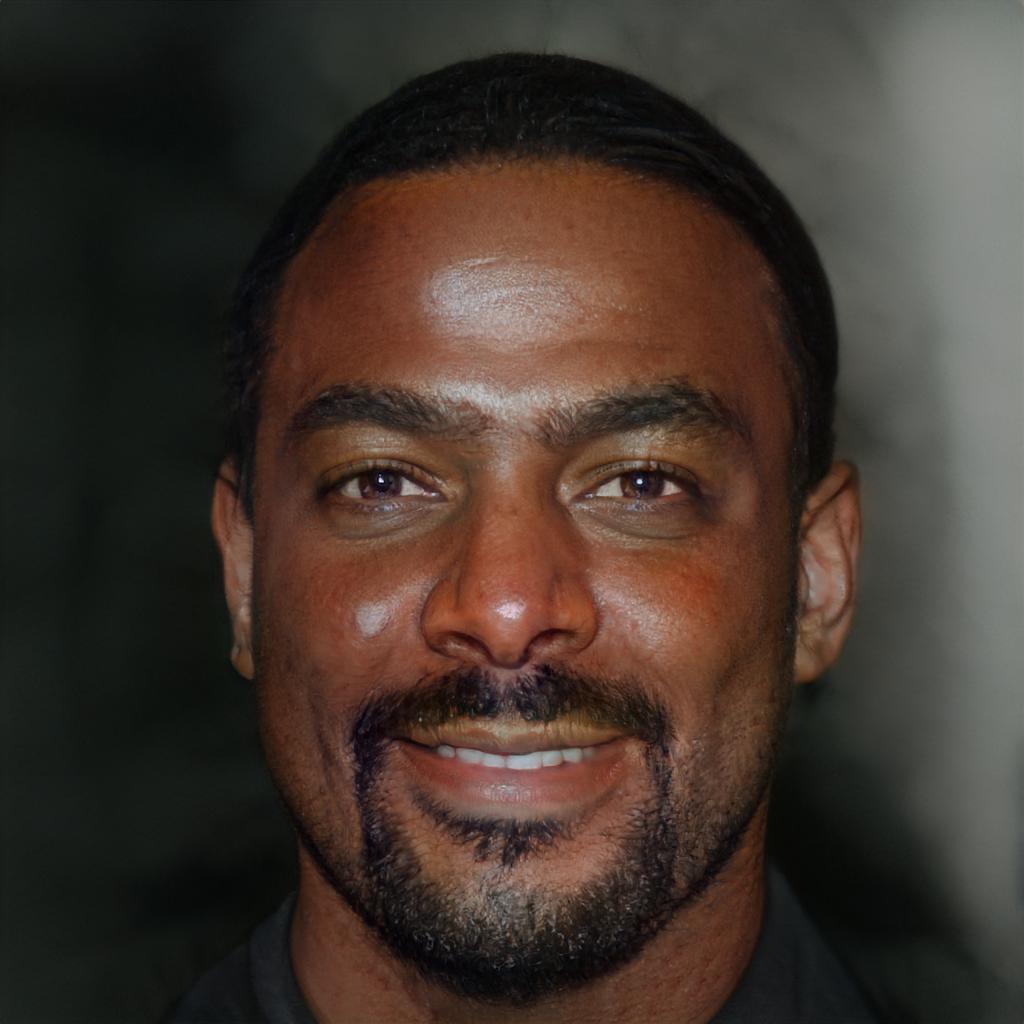} &
                    \includegraphics[width=0.13\textwidth]{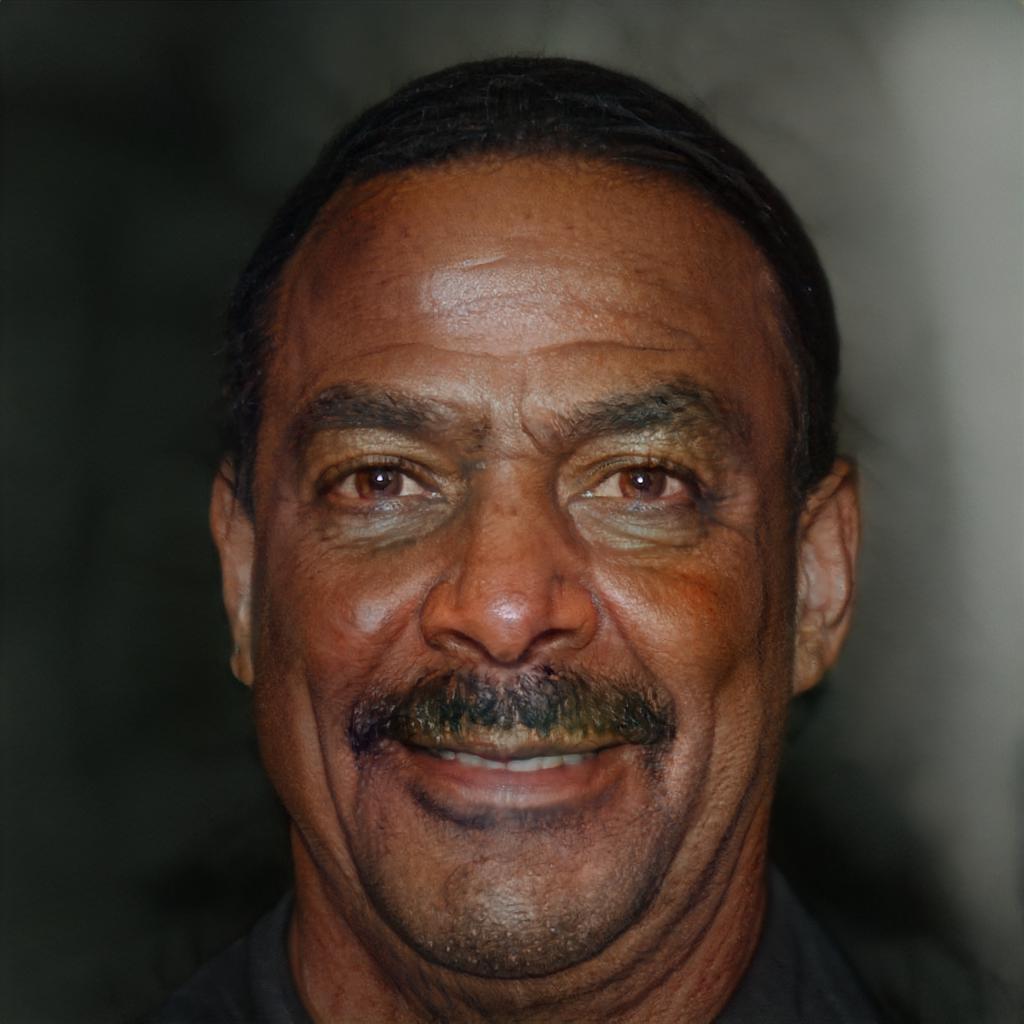}
    
            \tabularnewline
            \includegraphics[width=0.13\textwidth]{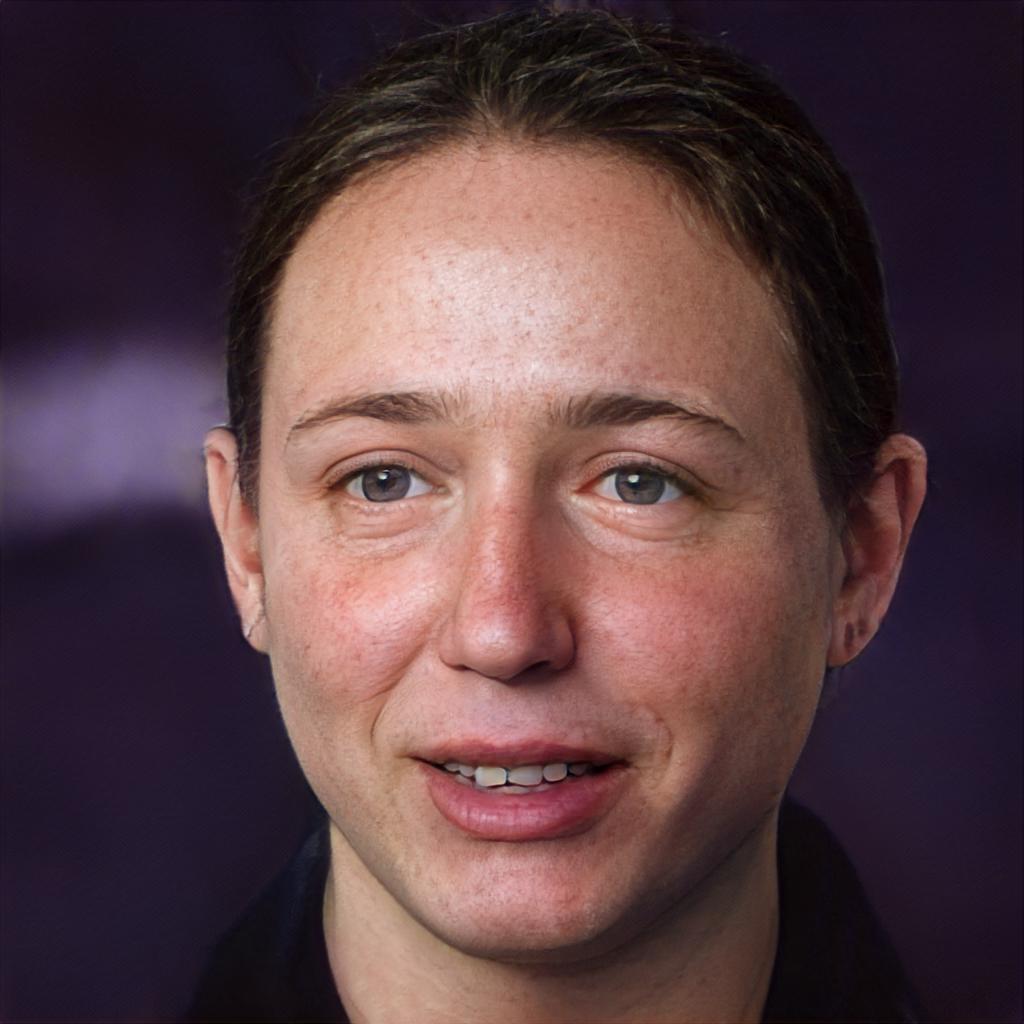} &
                  & \raisebox{0.15in}{\rotatebox[origin=t]{90}{LIFE}} & 
                    \includegraphics[width=0.13\textwidth]{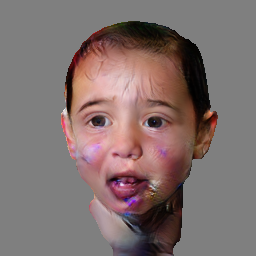} &
                    \includegraphics[width=0.13\textwidth]{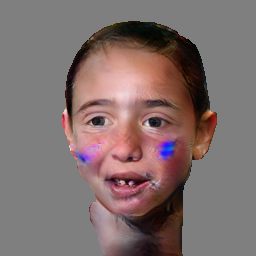} &
                    \includegraphics[width=0.13\textwidth]{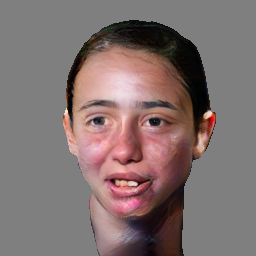} &
                    \includegraphics[width=0.13\textwidth]{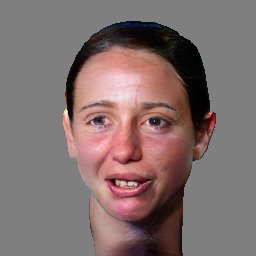} &
                    \includegraphics[width=0.13\textwidth]{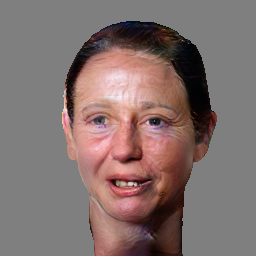} \\
                  & & \raisebox{0.15in}{\rotatebox[origin=t]{90}{SAM}} &
                    \includegraphics[width=0.13\textwidth]{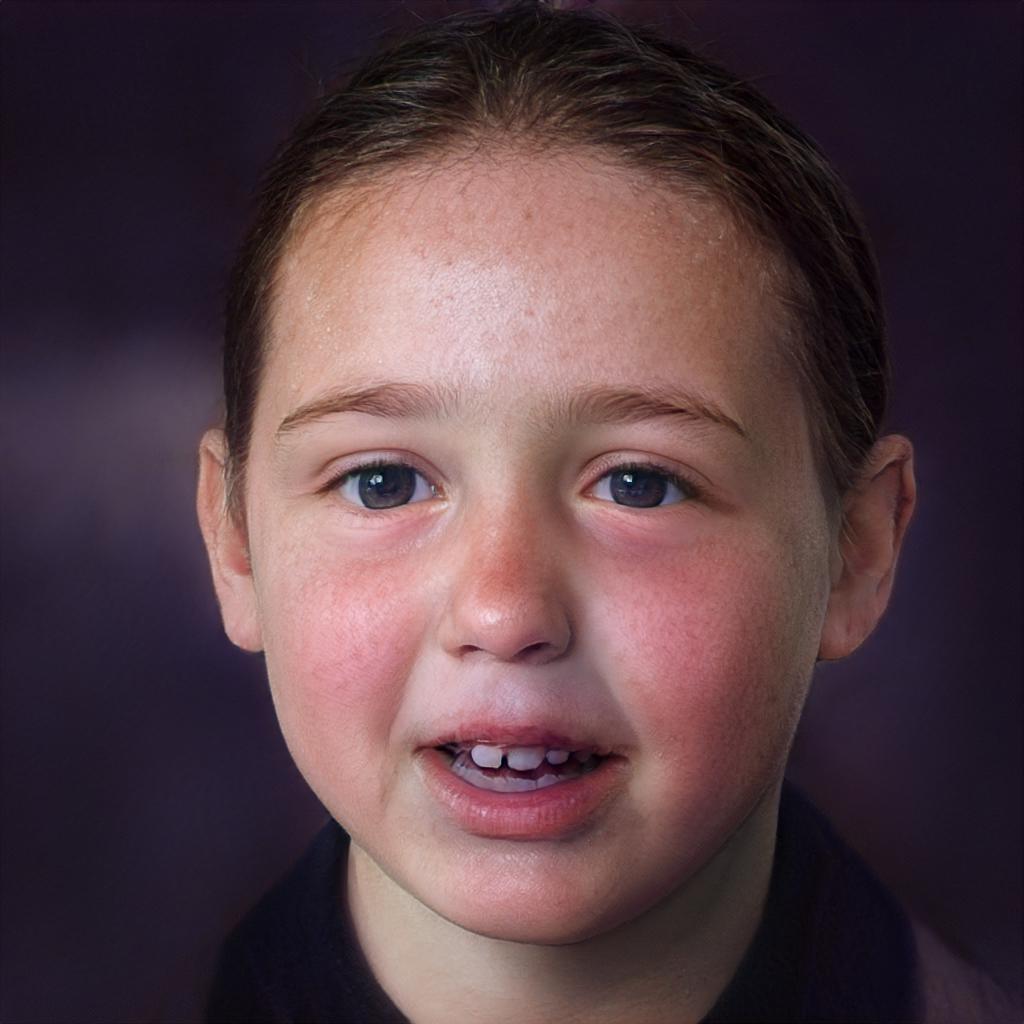} &
                    \includegraphics[width=0.13\textwidth]{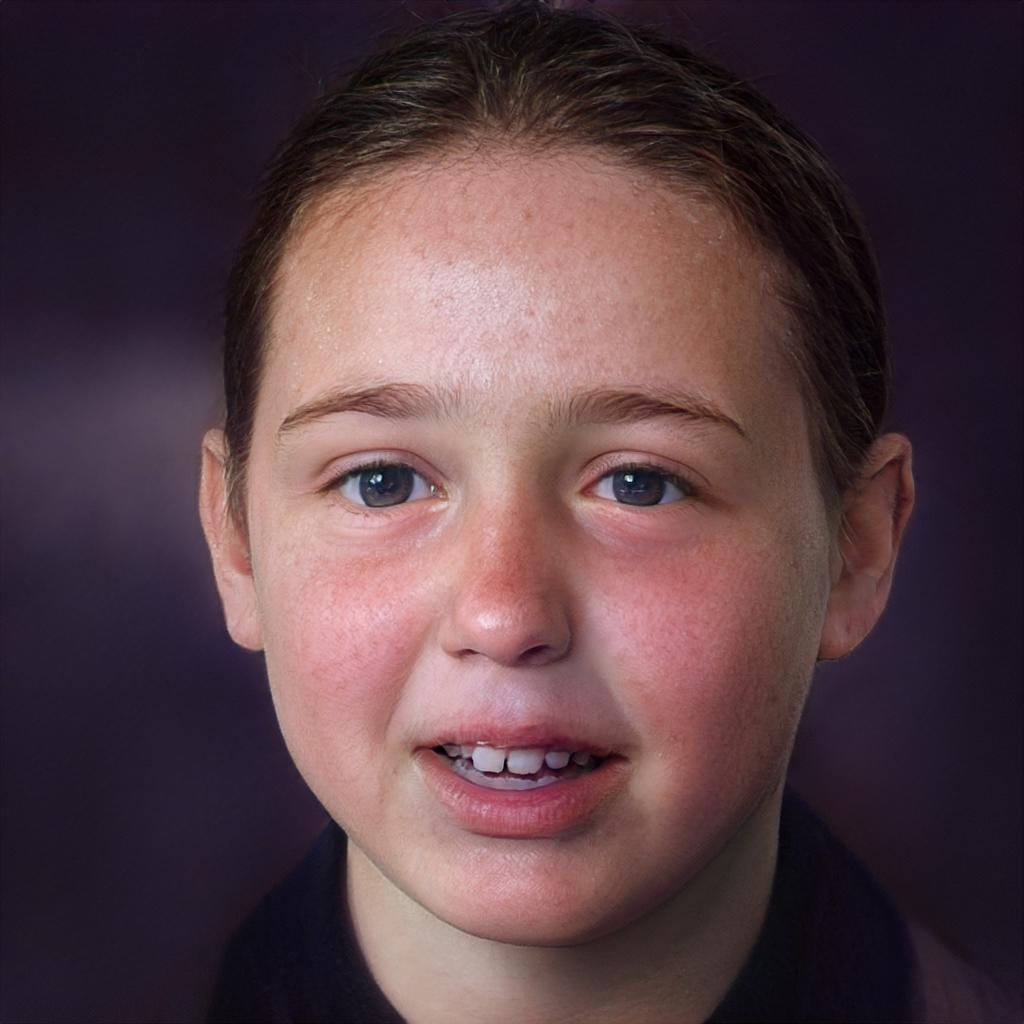} &
                    \includegraphics[width=0.13\textwidth]{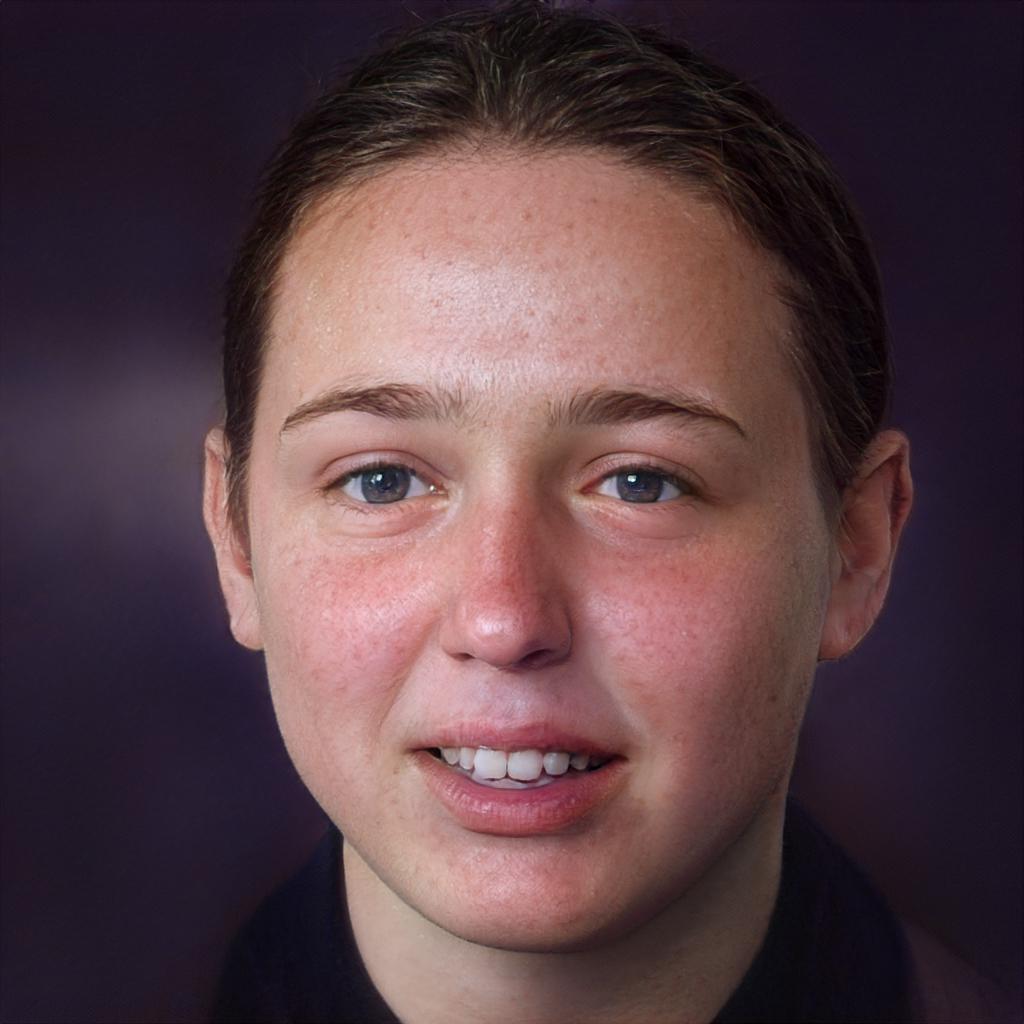} &
                    \includegraphics[width=0.13\textwidth]{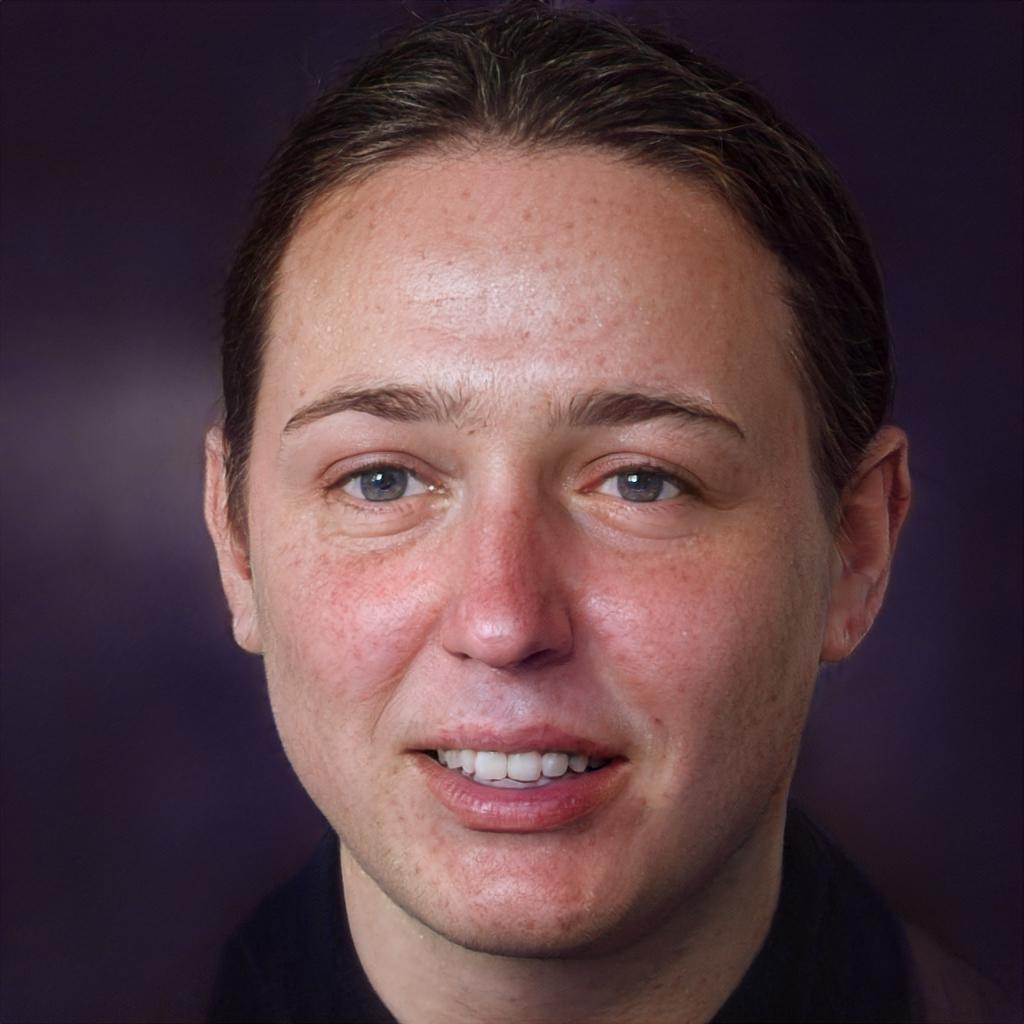} &
                    \includegraphics[width=0.13\textwidth]{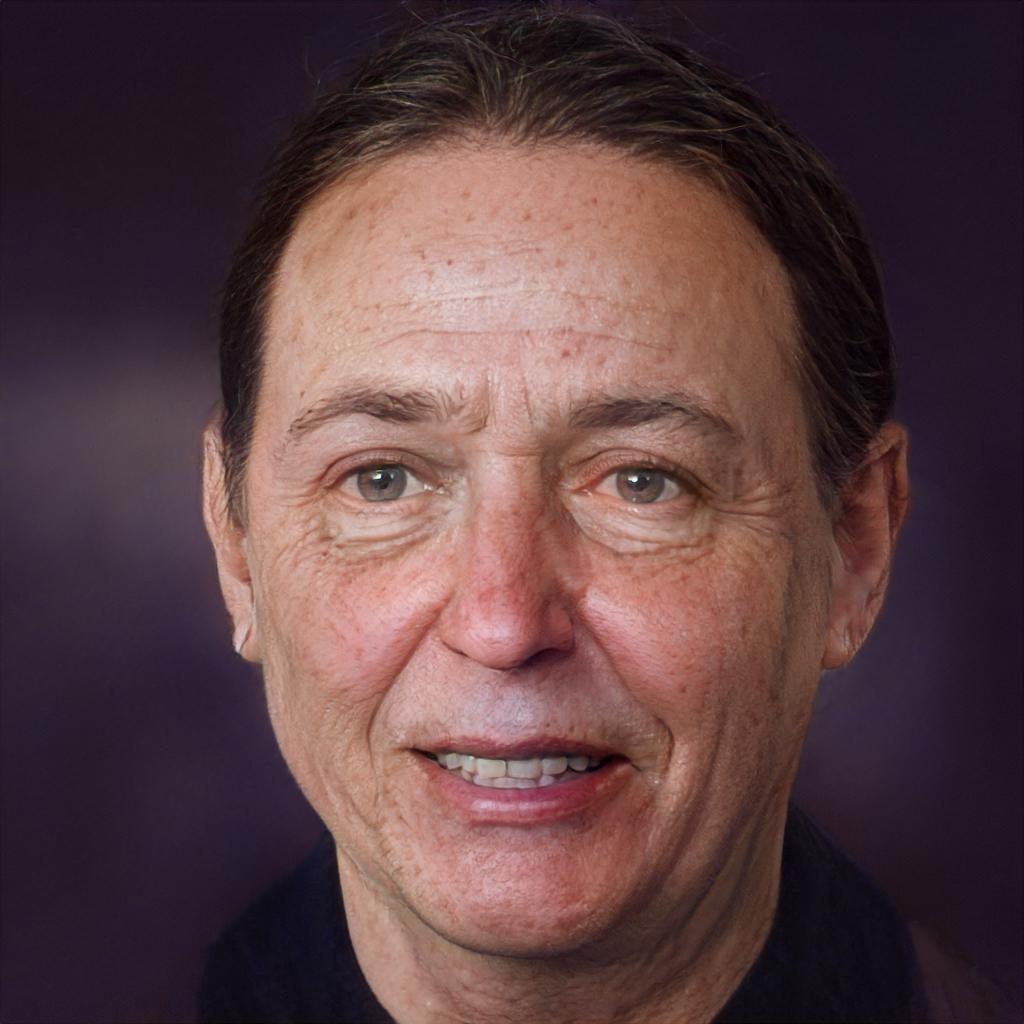}
    
            \end{tabular}
        \vspace{-0.1cm}
        \caption{}
        \label{fig:comparison_lifespan}
    \end{subfigure}%
    \begin{subfigure}{0.5\textwidth}
        \setlength{\tabcolsep}{1pt}
        \centering
            \begin{tabular}{c c c c c c c c c}
            Inversion & & & 25 & 35 & 45 & 55 & 65 \\

            \includegraphics[width=0.13\textwidth]{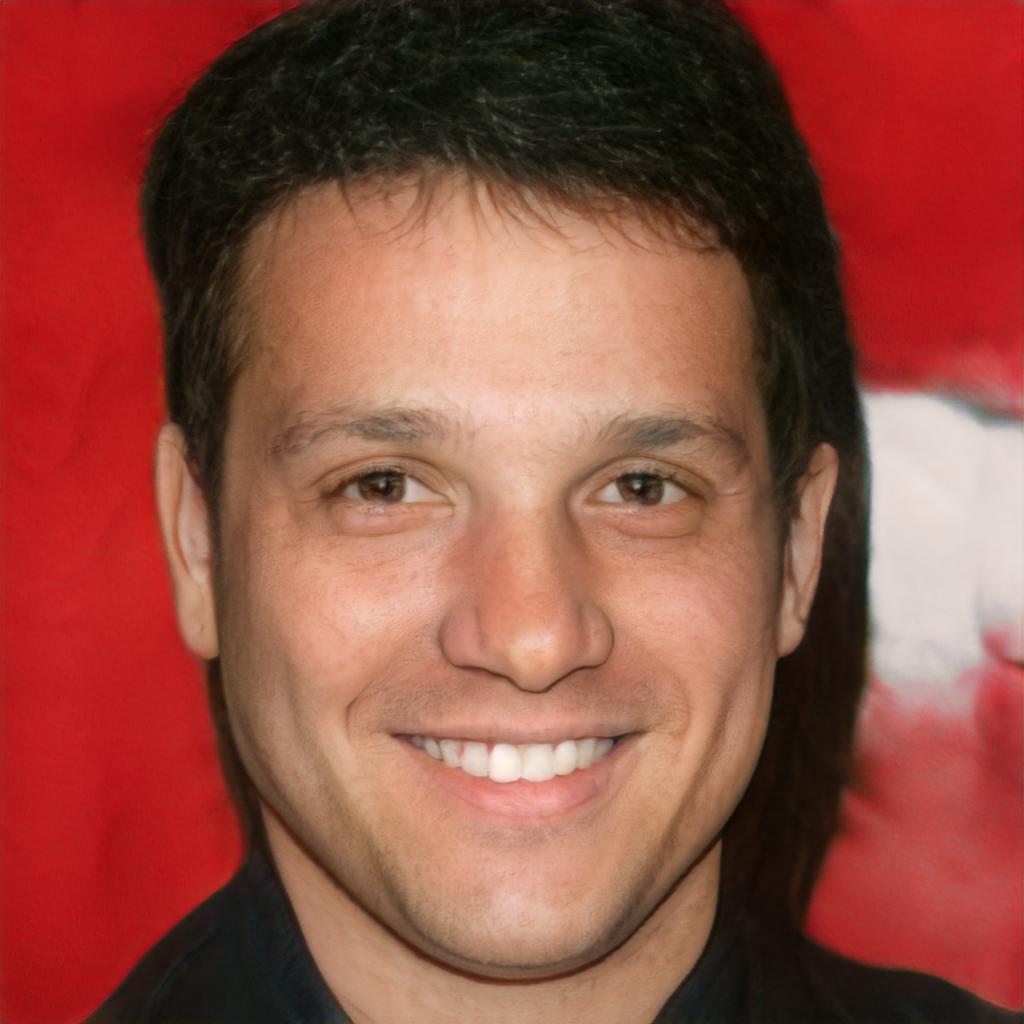} &
                  & \raisebox{0.15in}{\rotatebox[origin=t]{90}{HRFAE}} & 
                    \includegraphics[width=0.13\textwidth]{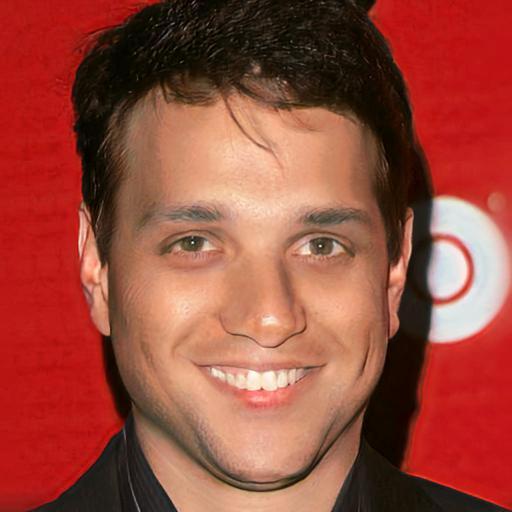} &
                    \includegraphics[width=0.13\textwidth]{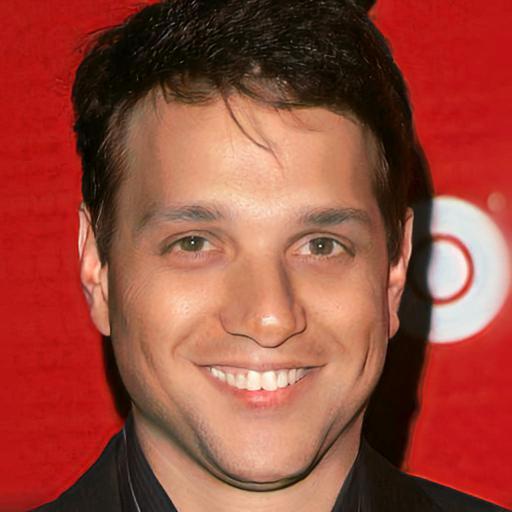} &
                    \includegraphics[width=0.13\textwidth]{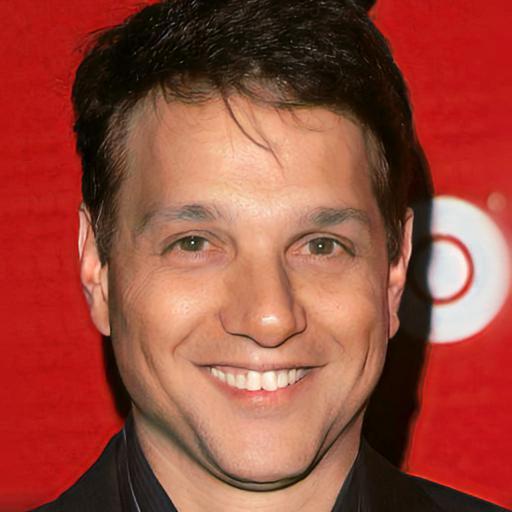} &
                    \includegraphics[width=0.13\textwidth]{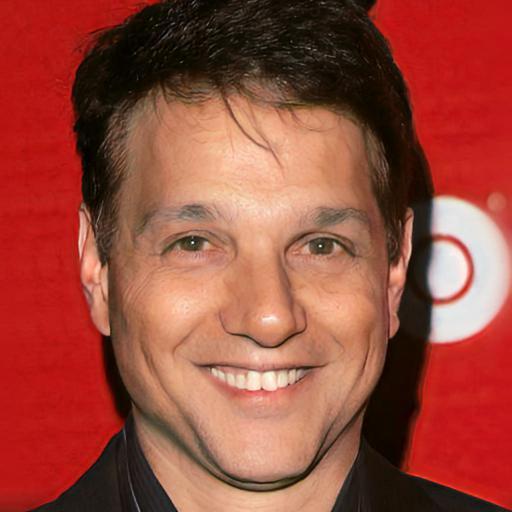} &
                    \includegraphics[width=0.13\textwidth]{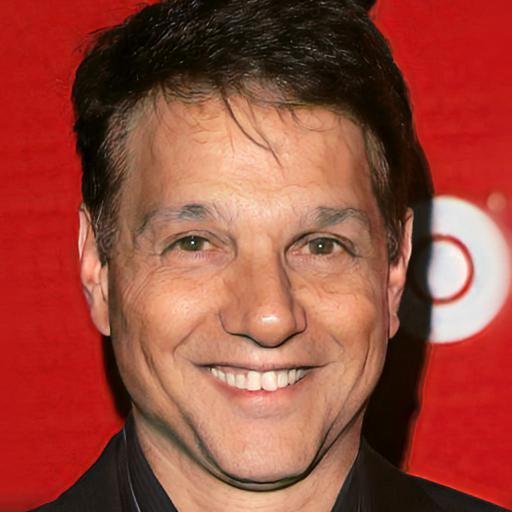} \\
                  & & \raisebox{0.15in}{\rotatebox[origin=t]{90}{SAM}} &
                    \includegraphics[width=0.13\textwidth]{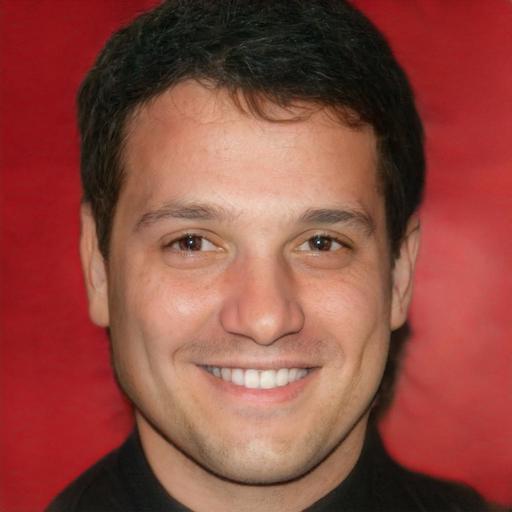} &
                    \includegraphics[width=0.13\textwidth]{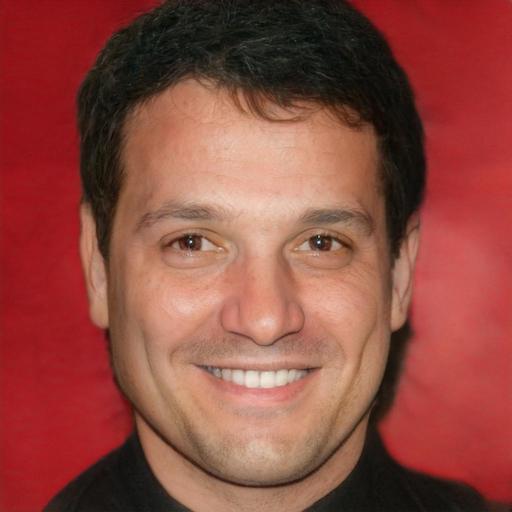} &
                    \includegraphics[width=0.13\textwidth]{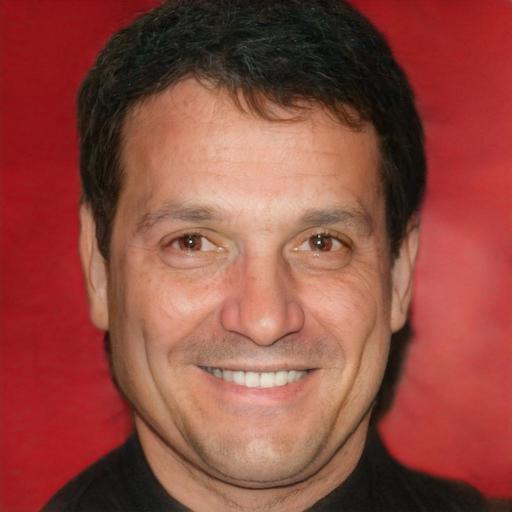} &
                    \includegraphics[width=0.13\textwidth]{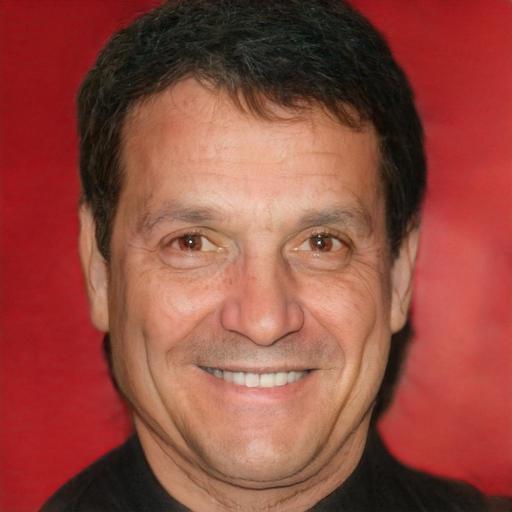} &
                    \includegraphics[width=0.13\textwidth]{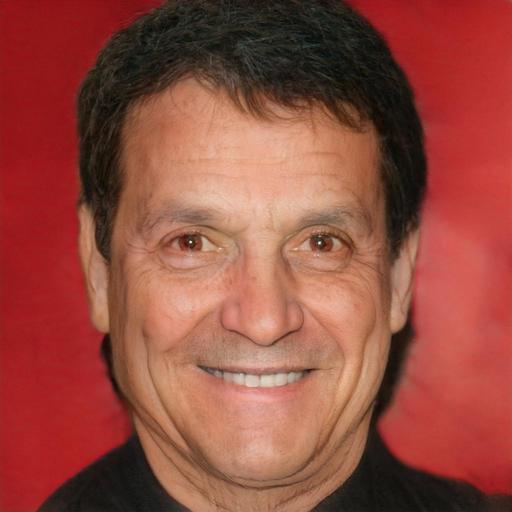}

            \tabularnewline
            \includegraphics[width=0.13\textwidth]{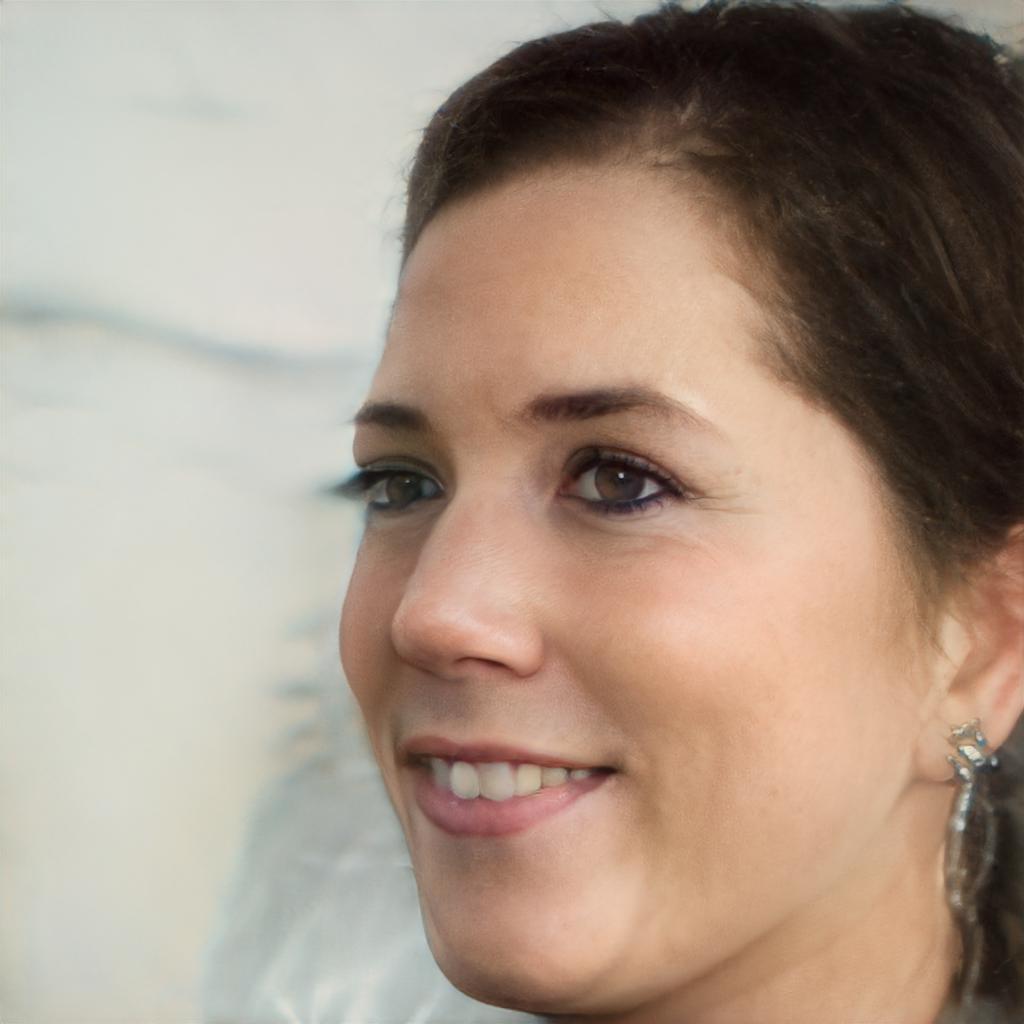} &
                  & \raisebox{0.15in}{\rotatebox[origin=t]{90}{HRFAE}} & 
                    \includegraphics[width=0.13\textwidth]{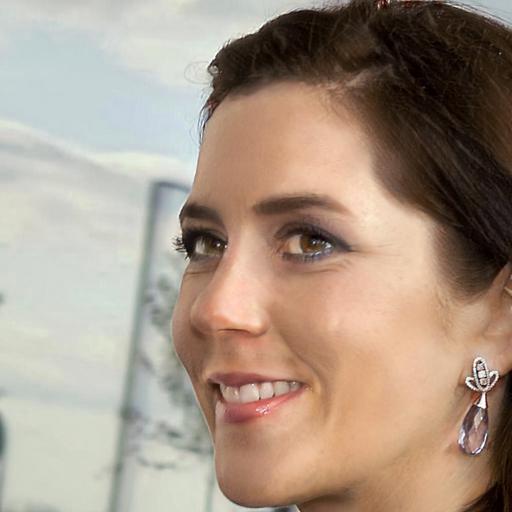} &
                    \includegraphics[width=0.13\textwidth]{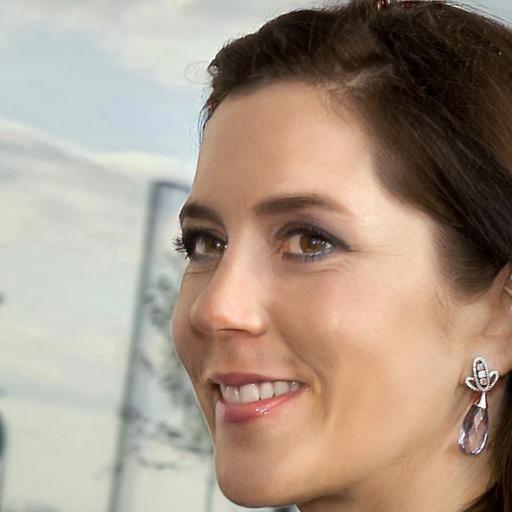} &
                    \includegraphics[width=0.13\textwidth]{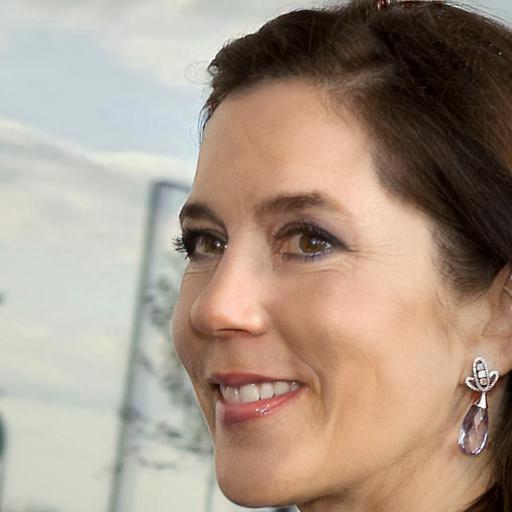} &
                    \includegraphics[width=0.13\textwidth]{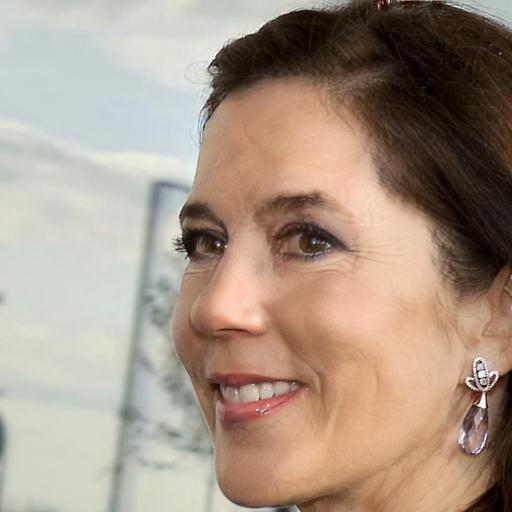} &
                    \includegraphics[width=0.13\textwidth]{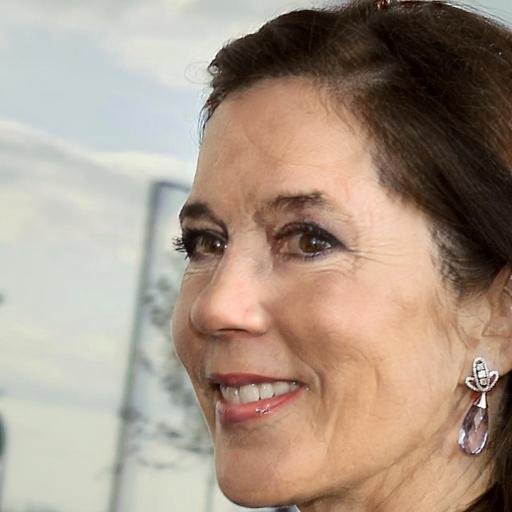} \\
                  & & \raisebox{0.15in}{\rotatebox[origin=t]{90}{SAM}} &
                    \includegraphics[width=0.13\textwidth]{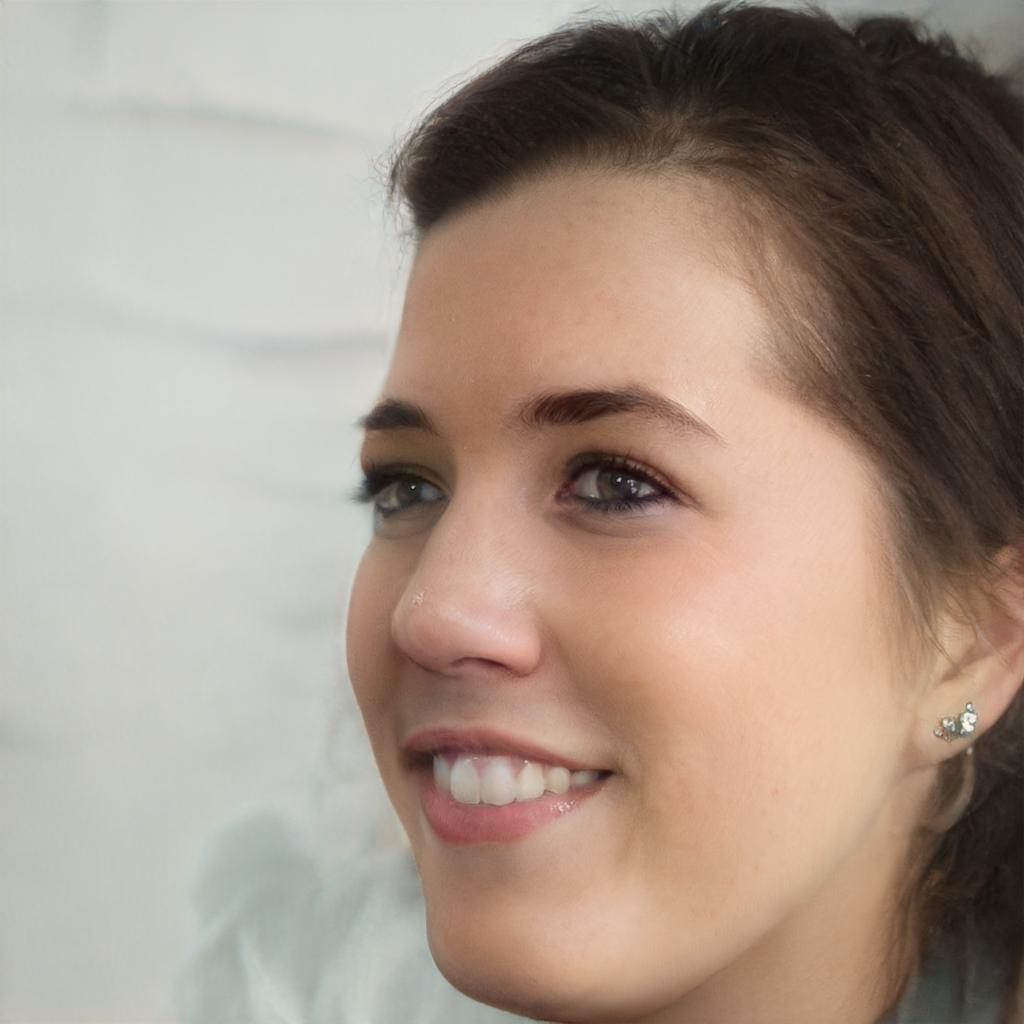} &
                    \includegraphics[width=0.13\textwidth]{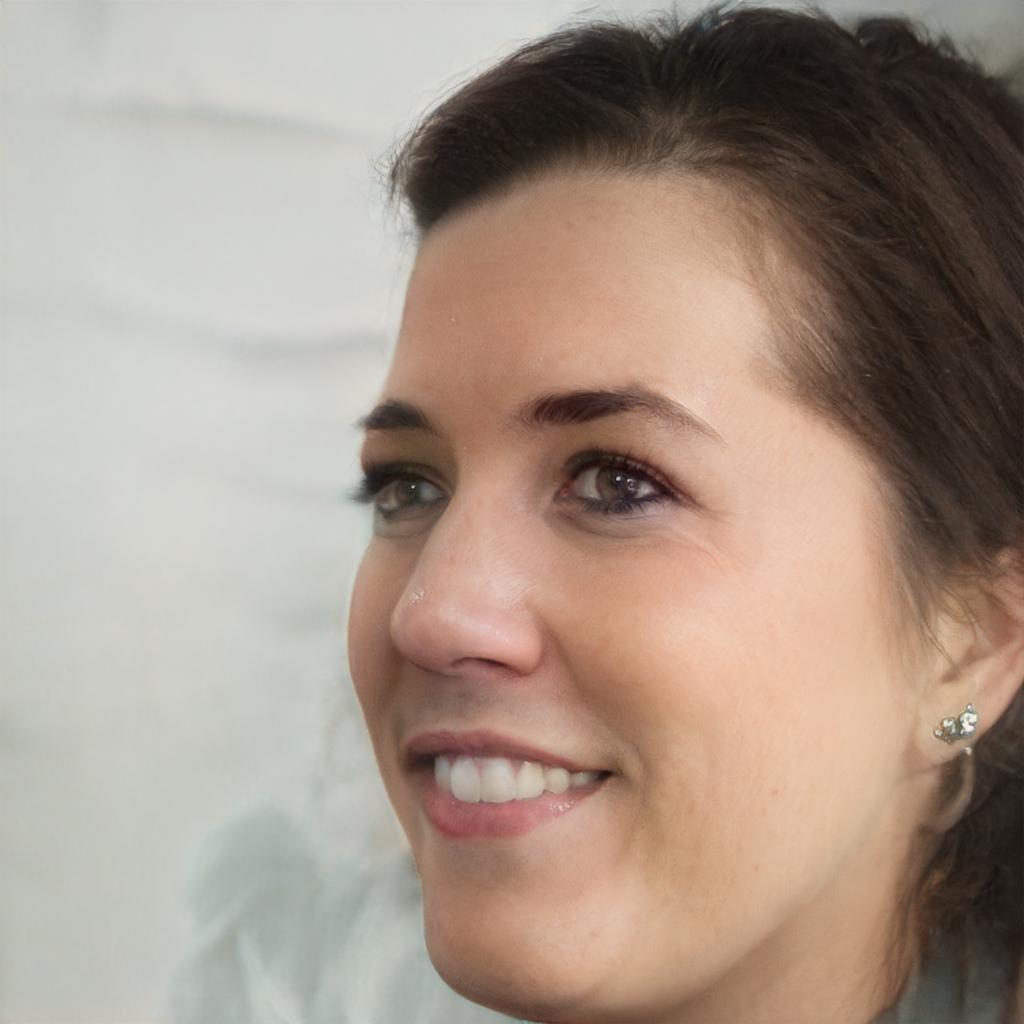} &
                    \includegraphics[width=0.13\textwidth]{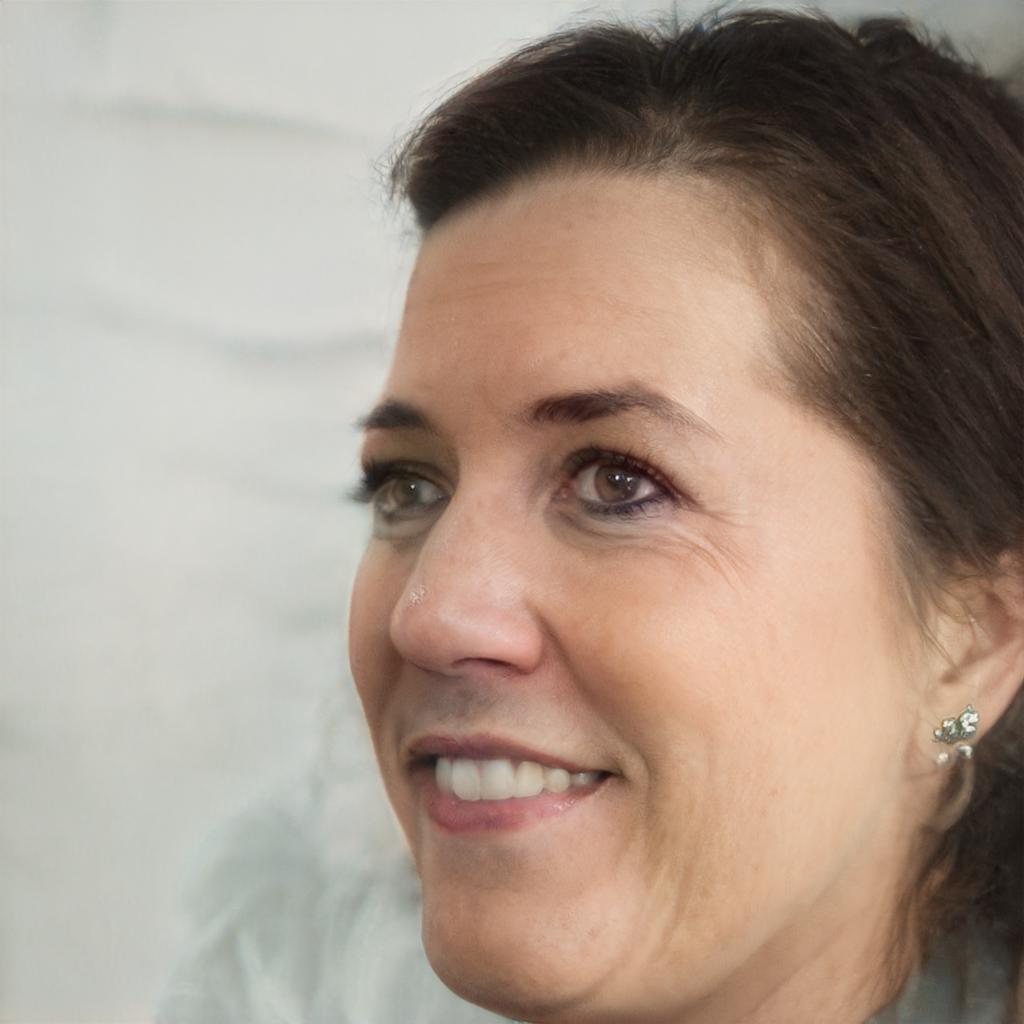} &
                    \includegraphics[width=0.13\textwidth]{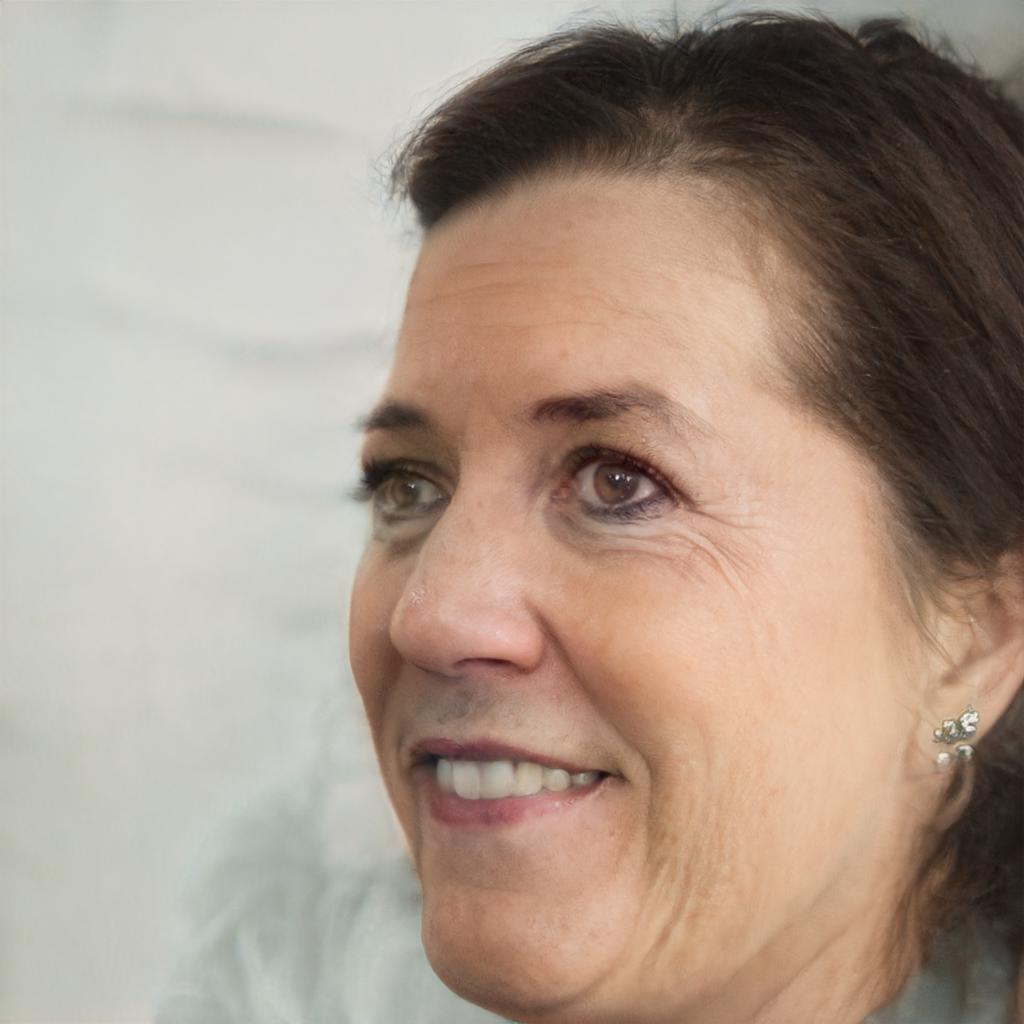} &
                    \includegraphics[width=0.13\textwidth]{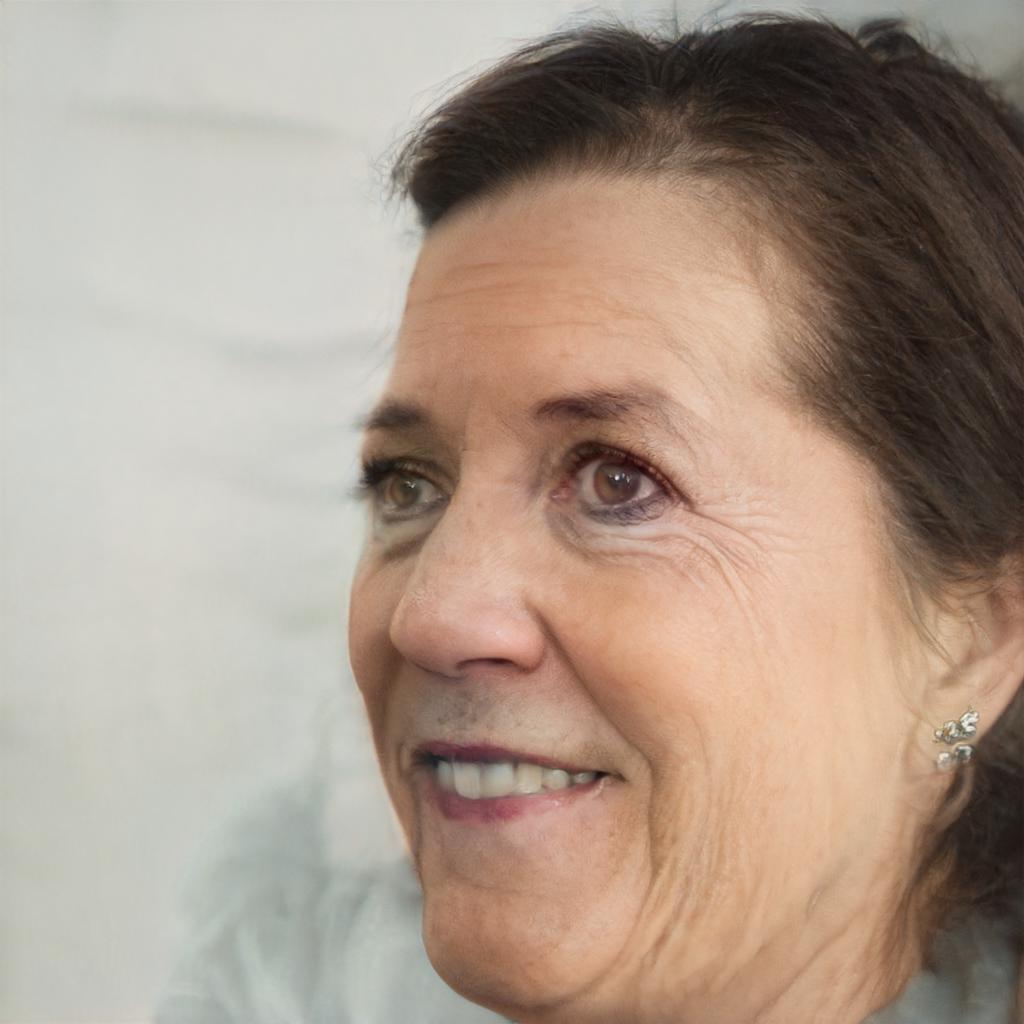}

            \tabularnewline
                \includegraphics[width=0.13\textwidth]{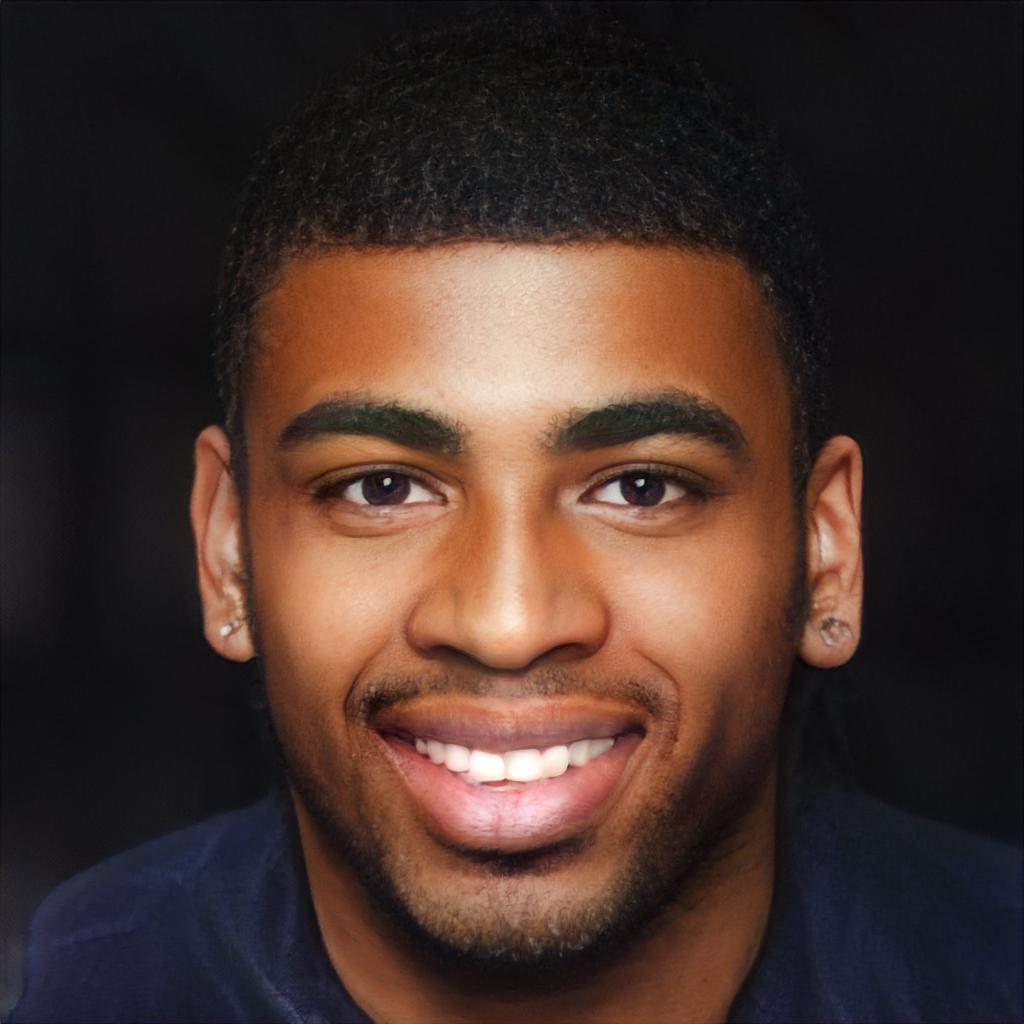} &
                  & \raisebox{0.15in}{\rotatebox[origin=t]{90}{HRFAE}} & 
                    \includegraphics[width=0.13\textwidth]{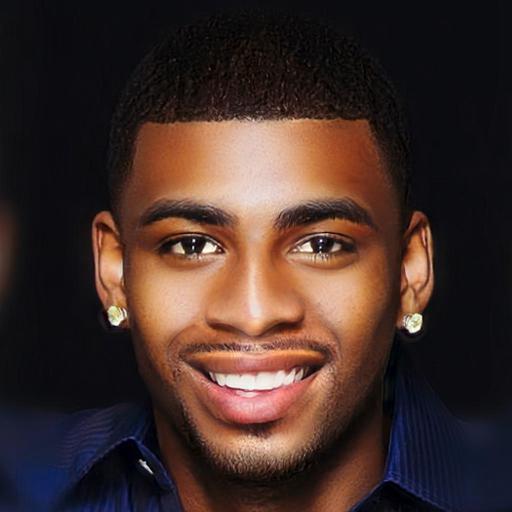} &
                    \includegraphics[width=0.13\textwidth]{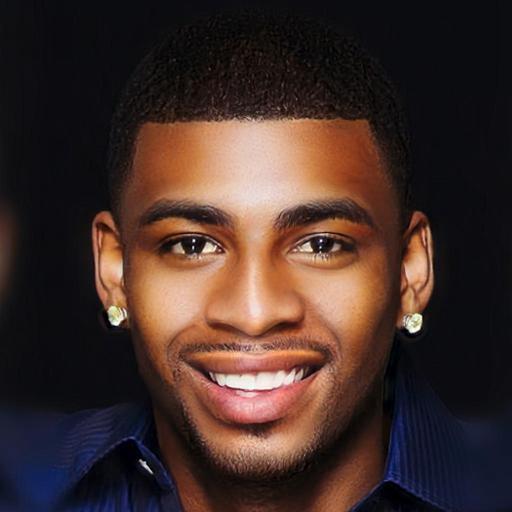} &
                    \includegraphics[width=0.13\textwidth]{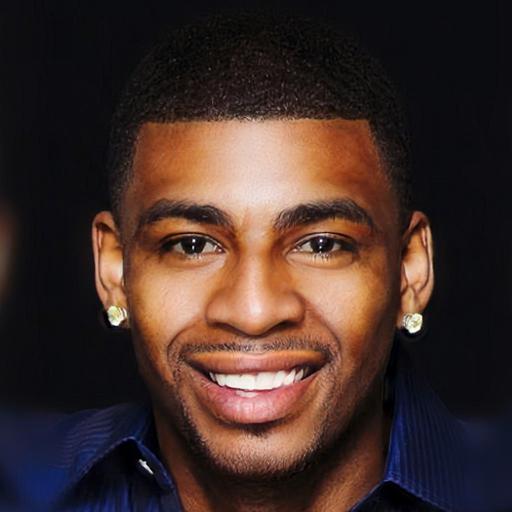} &
                    \includegraphics[width=0.13\textwidth]{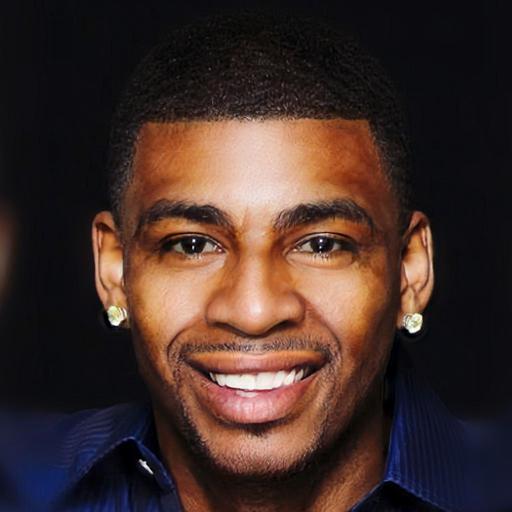} &
                    \includegraphics[width=0.13\textwidth]{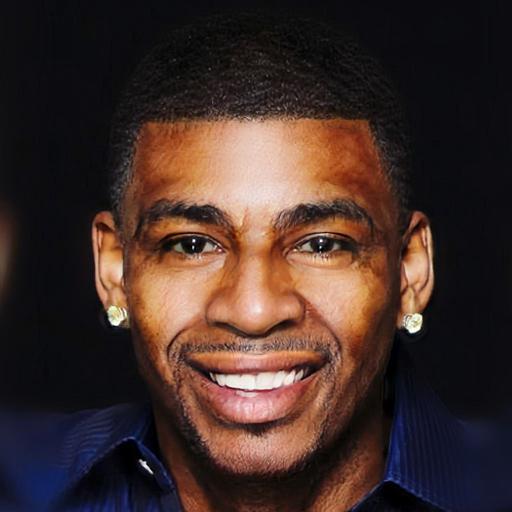} \\
                  & & \raisebox{0.15in}{\rotatebox[origin=t]{90}{SAM}} &
                    \includegraphics[width=0.13\textwidth]{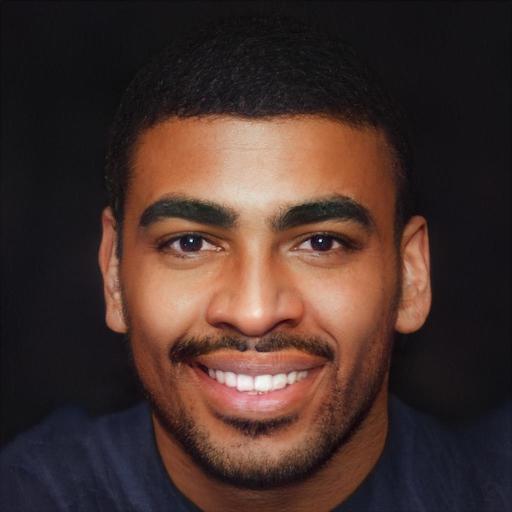} &
                    \includegraphics[width=0.13\textwidth]{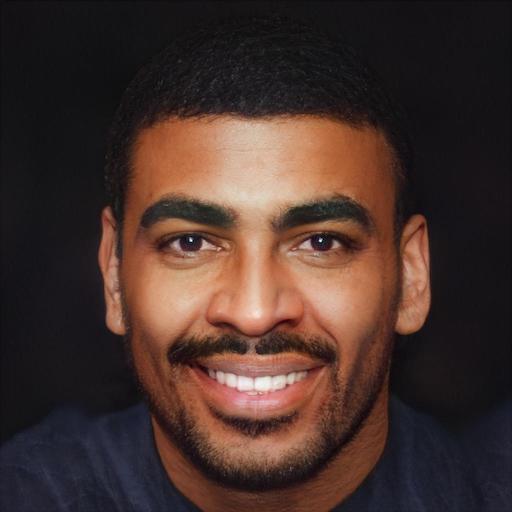} &
                    \includegraphics[width=0.13\textwidth]{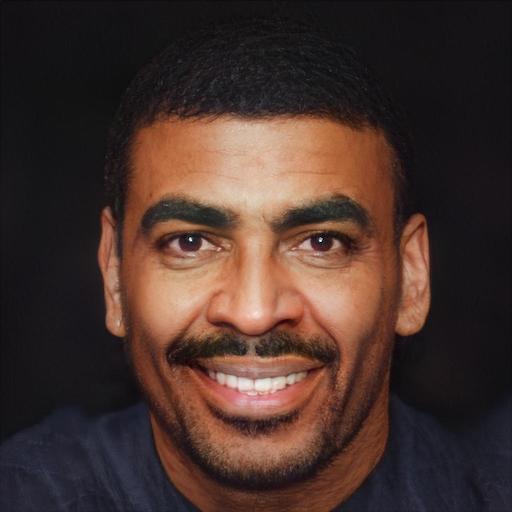} &
                    \includegraphics[width=0.13\textwidth]{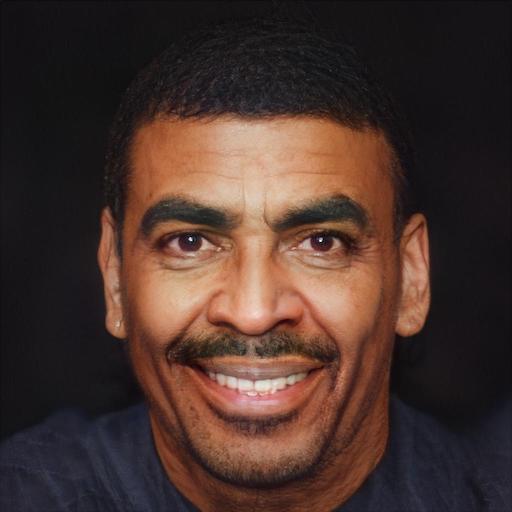} &
                    \includegraphics[width=0.13\textwidth]{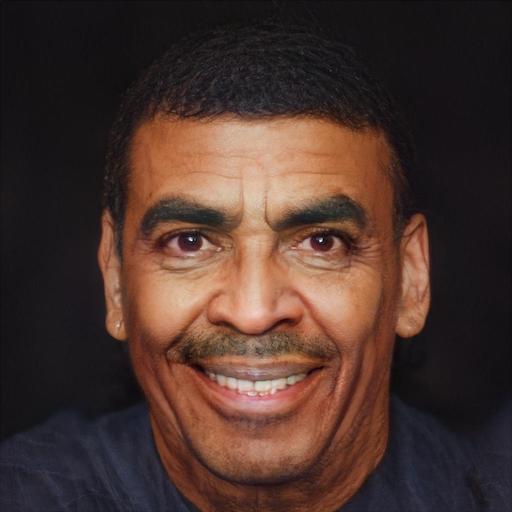}
    
            \end{tabular}
        \vspace{-0.1cm}
        \caption{}
        \label{fig:comparison_hrfae}
    \end{subfigure}
    \vspace{-0.25cm}
    \caption{Qualitative comparison of age transformation results with (a) LIFE~\cite{orel2020lifespan} and (b) HRFAE~\cite{yao2020high} on the CelebA-HQ~\cite{karras2017progressive} test set. For translating our images to the age groups in LIFE, we set the target age equal to the median age of each group. Best viewed zoomed-in. Additional results can be found in Appendix~\ref{additional_results}.}
\end{figure*}

%% file: figures/quantitative_aging.tex
\begin{figure}
    \centering
    \setlength{\belowcaptionskip}{-3pt}
    \setlength{\tabcolsep}{1pt}
    \centering
        \begin{tabular}{c}
        \includegraphics[width=0.95\linewidth]{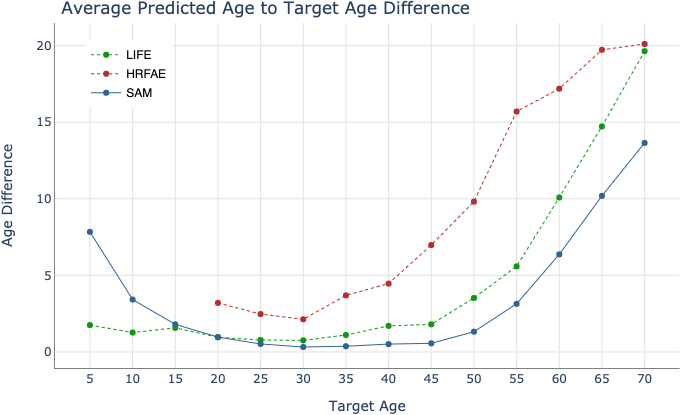}
        \end{tabular}
    \vspace{-0.25cm}
    \caption{Here, we examine each method's ability to 
    generate a full lifespan of images. 
    Note that the lower the better.}
    \label{fig:aging_accuracy}
\end{figure}

%% file: figures/human_evaluation.tex
\begin{table}
    \centering
    \caption{Each cell indicates the percent of respondents who preferred the corresponding method for the evaluated metric. When compared to LIFE~\cite{orel2020lifespan} and HRFAE~\cite{yao2020high}, SAM achieves comparable or superior aging accuracy and image quality across the different target ages.}
    \vspace{-0.2cm}
    \begin{tabular}{lcccccc}
    \toprule Target &
    \multicolumn{2}{c}{5} & 
    \multicolumn{2}{c}{30} &
    \multicolumn{2}{c}{65} \\ 
    \cmidrule(r){1-1}
    \cmidrule(r){2-3}
    \cmidrule(r){4-5}
    \cmidrule(r){6-7}
    Method & 
    {Age} & {Quality} & 
    {Age} & {Quality} & 
    {Age} & {Quality} \\
    \midrule
    HRFAE & - & - & 12.56 & 17.45 & 9.24  & 15.97 \\
    LIFE  & \textbf{55.1} & 18.88 & 7.25  & 12.11 & 11.14 & 11.64 \\
    SAM   & 44.9 & \textbf{81.12} & \textbf{80.19} & \textbf{70.44} & \textbf{79.62} & \textbf{72.39} \\
    \bottomrule
    \end{tabular}
    \caption*{\textbf{Human Evaluation I}}
    \vspace{-0.5cm}
    \label{tb:human_evaluation}
\end{table}

%% file: figures/human_evaluation_identity.tex
\begin{table}
    \centering
    \vspace{-0.2cm}
    \caption{We asked respondents to identity images of celebrities transformed to the different target ages using each of the three methods. Each cell indicates the percent of queries correctly identified at the corresponding target age. As shown, all three methods perform comparably.}
    \vspace{-0.2cm}
    \begin{tabular}{lcccccc}
    \toprule Target &
    \multicolumn{1}{c}{5} & 
    \multicolumn{1}{c}{30} &
    \multicolumn{1}{c}{65} \\ 
    \cmidrule(r){1-1}
    \cmidrule(r){2-2}
    \cmidrule(r){3-3}
    \cmidrule(r){4-4}
    Method & 
    {ID Recall} & 
    {ID Recall} & 
    {ID Recall} \\
    \midrule
    HRFAE & - & 90.1 & 88.9 \\
    LIFE  & 76.3 & 86.2 & 80.4 \\
    SAM   & 89.6 & 91.1 & 86.7 \\
    \bottomrule
    \end{tabular}
    \caption*{\textbf{Human Evaluation II}}
    \vspace{-0.6cm}
    \label{tb:human_evaluation_identity}
\end{table}

%% file: figures/interfacegan_comparison.tex
\begin{figure*}
    \centering
    \setlength{\belowcaptionskip}{-2.5pt}
    \setlength{\tabcolsep}{1pt}
    \begin{tabular}{c c c}
        Inversion & & \\
        \includegraphics[width=0.09\textwidth]{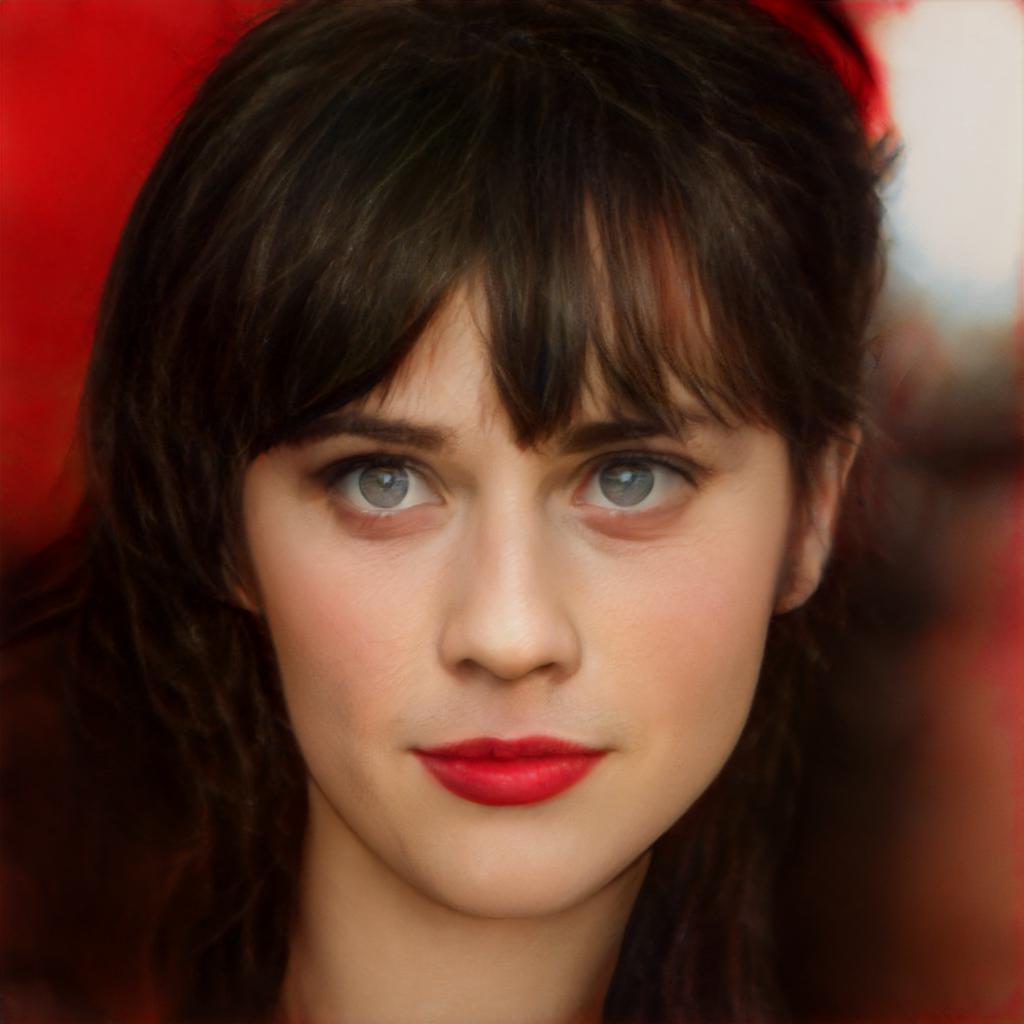} &
        \raisebox{0.25in}{\rotatebox[origin=t]{90}{InterFace}} & 
        \includegraphics[width=0.8\textwidth]{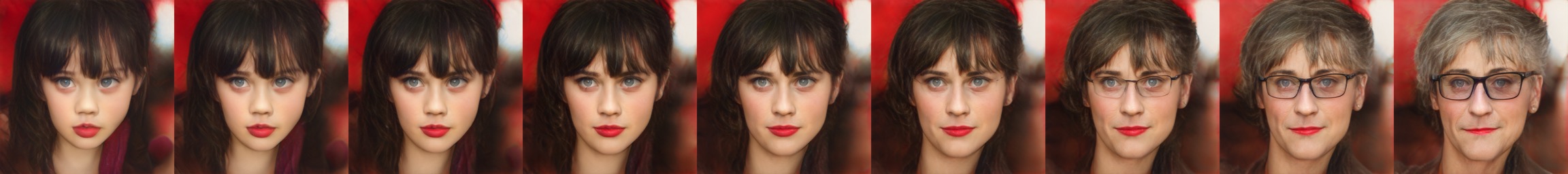} \\
        & \raisebox{0.225in}{\rotatebox[origin=t]{90}{StyleFlow}} & 
        \includegraphics[width=0.8\textwidth]{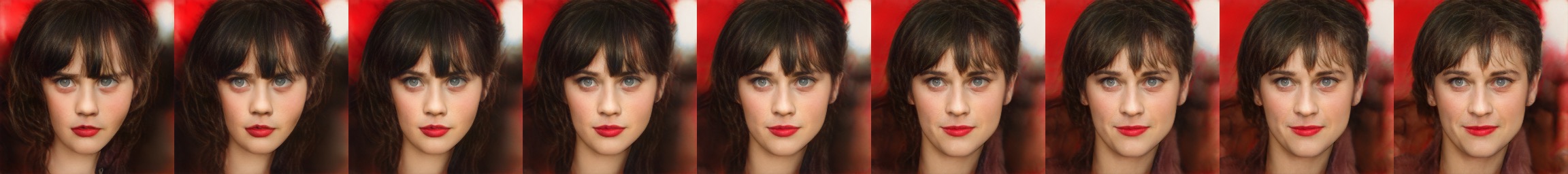} \\
        & \raisebox{0.25in}{\rotatebox[origin=t]{90}{SAM}} &
        \includegraphics[width=0.8\textwidth]{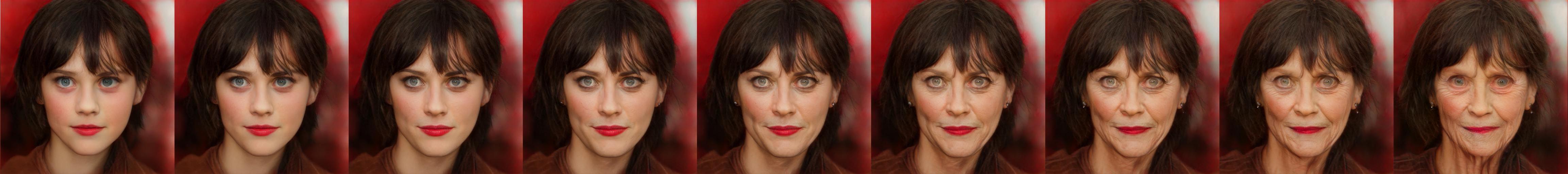}
        \tabularnewline

        \includegraphics[width=0.09\textwidth]{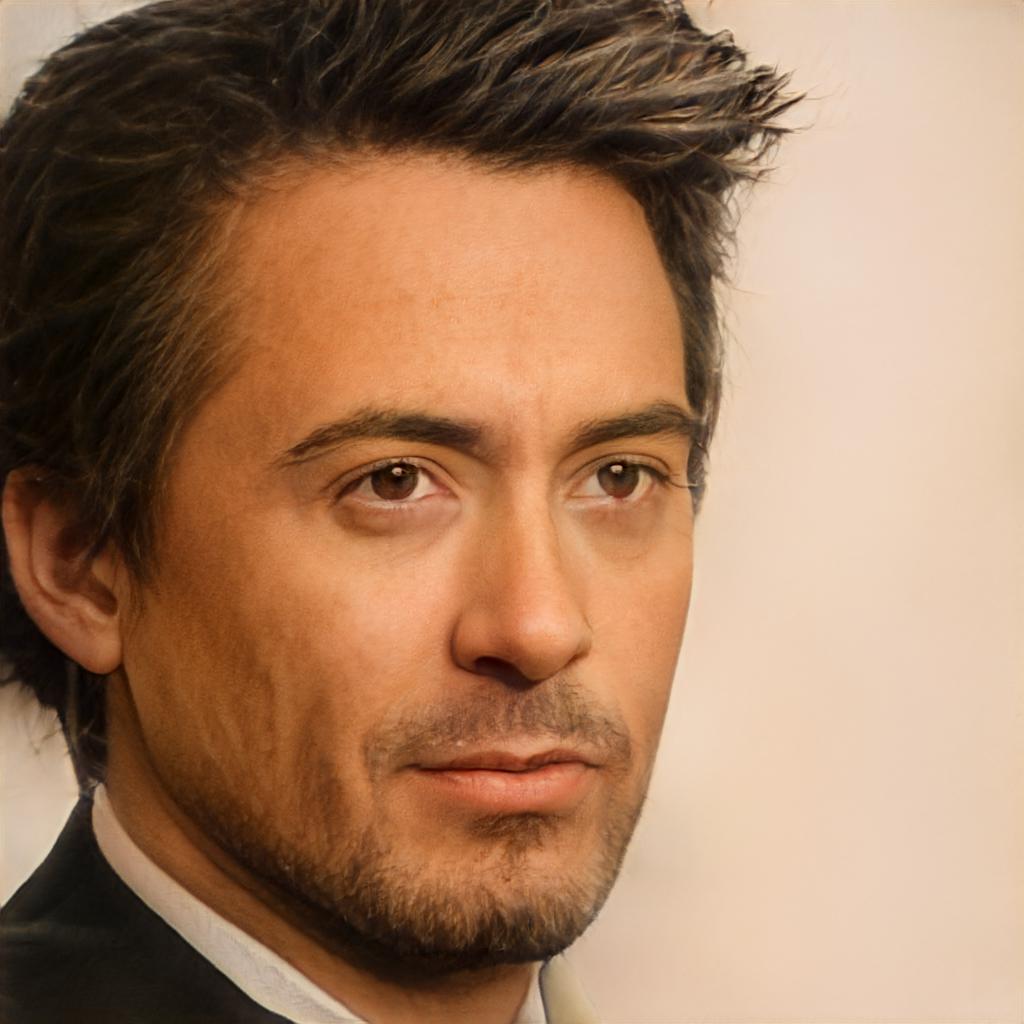} &
        \raisebox{0.25in}{\rotatebox[origin=t]{90}{InterFace}} & 
        \includegraphics[width=0.8\textwidth]{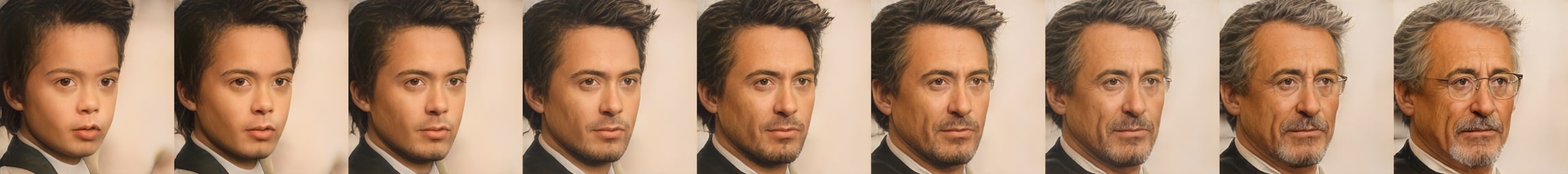} \\
        & \raisebox{0.225in}{\rotatebox[origin=t]{90}{StyleFlow}} & 
        \includegraphics[width=0.8\textwidth]{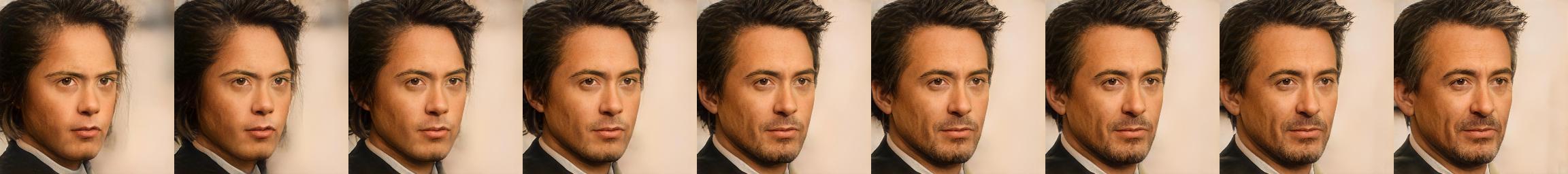} \\
        & \raisebox{0.25in}{\rotatebox[origin=t]{90}{SAM}} &
        \includegraphics[width=0.8\textwidth]{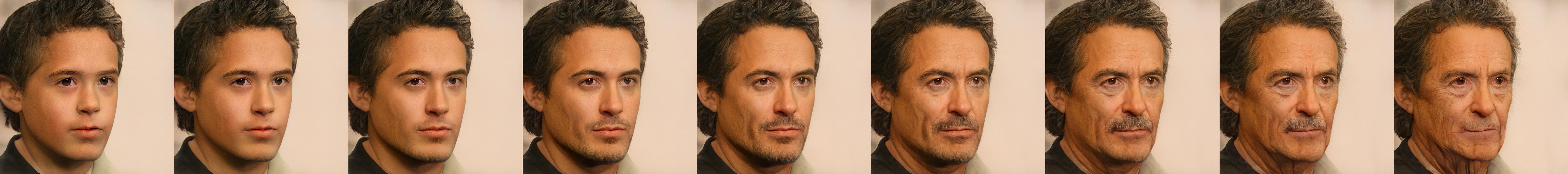}
        \tabularnewline

        \includegraphics[width=0.09\textwidth]{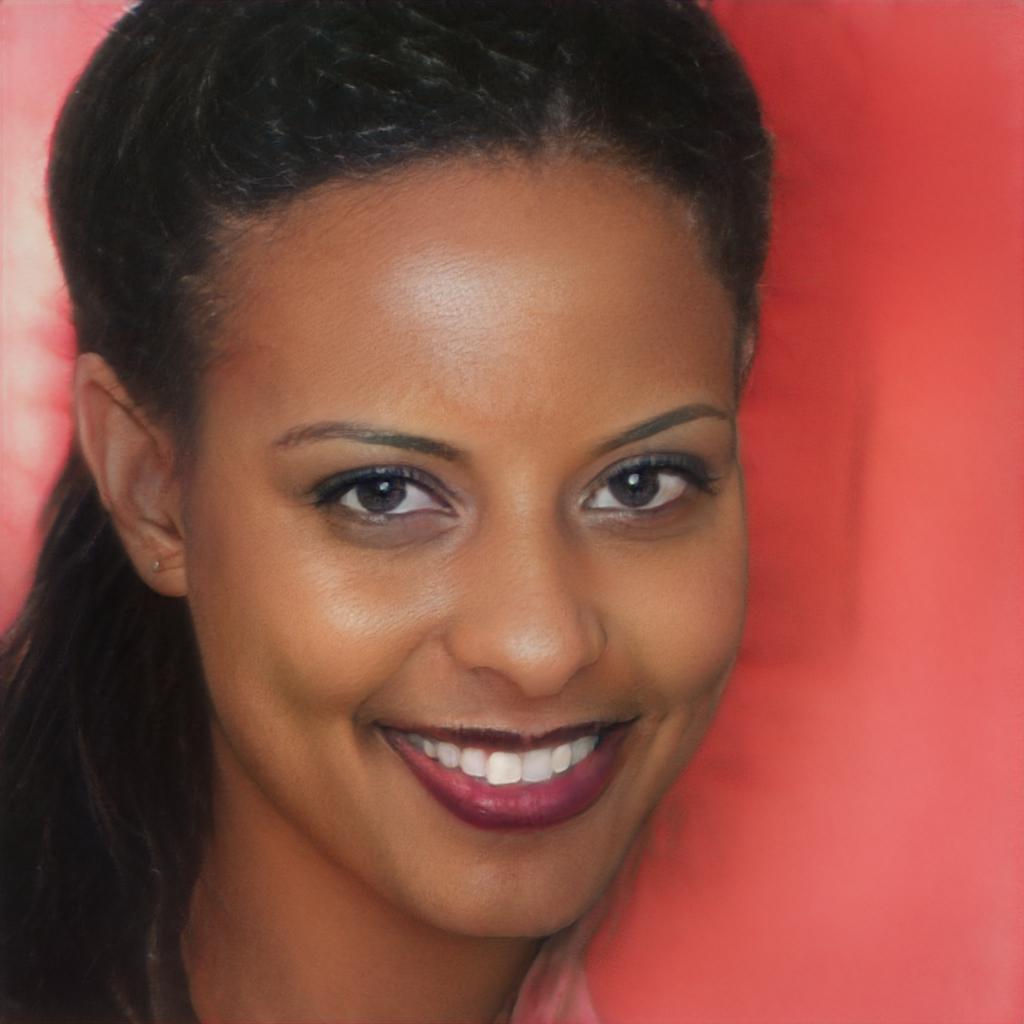} &
        \raisebox{0.25in}{\rotatebox[origin=t]{90}{InterFace}} & 
        \includegraphics[width=0.8\textwidth]{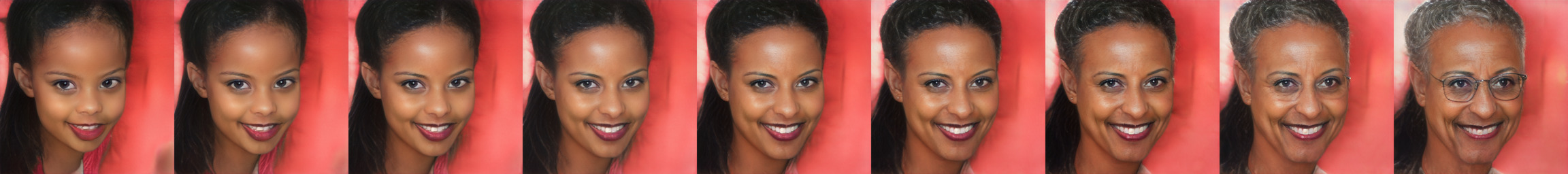} \\        
        & \raisebox{0.225in}{\rotatebox[origin=t]{90}{StyleFlow}} & 
        \includegraphics[width=0.8\textwidth]{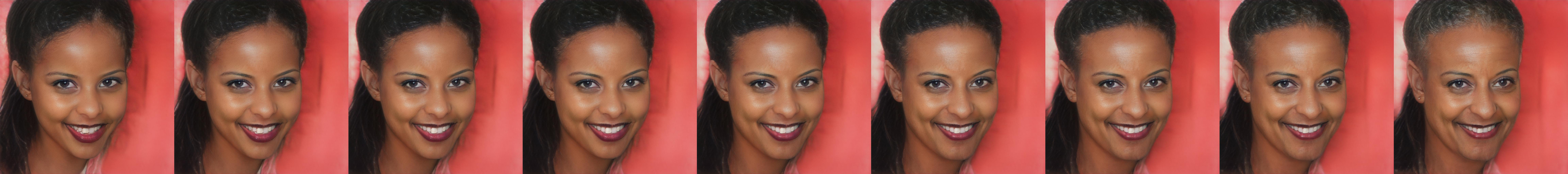} \\
        & \raisebox{0.25in}{\rotatebox[origin=t]{90}{SAM}} &
        \includegraphics[width=0.8\textwidth]{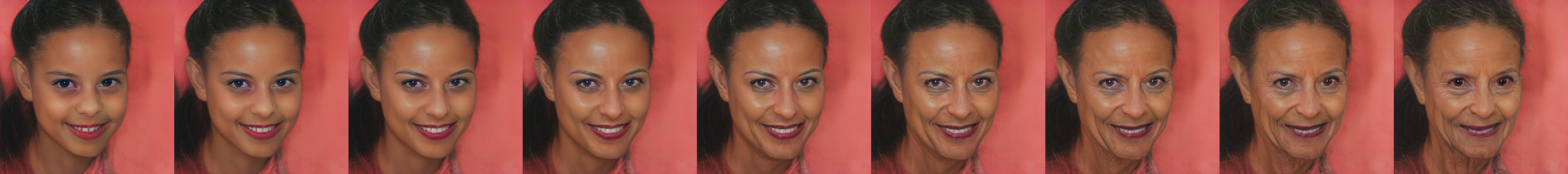}
    \end{tabular}
    \vspace{-0.15cm}
    \caption{
    A visual comparison with InterFaceGAN~\cite{shen2020interpreting} and StyleFlow~\cite{abdal2020styleflow} on real face images. Results for InterFaceGAN and StyleFlow are obtained by inverting the input into StyleGAN's latent space using pSp~\cite{richardson2020encoding} and traversing along the learned age path.
    }
    \label{fig:interface_comparison}
\end{figure*}

%% file: figures/latent_path.tex
\begin{figure*}
    \centering
    \begin{subfigure}{0.49\textwidth}
        \centering
        \includegraphics[width=0.95\linewidth]{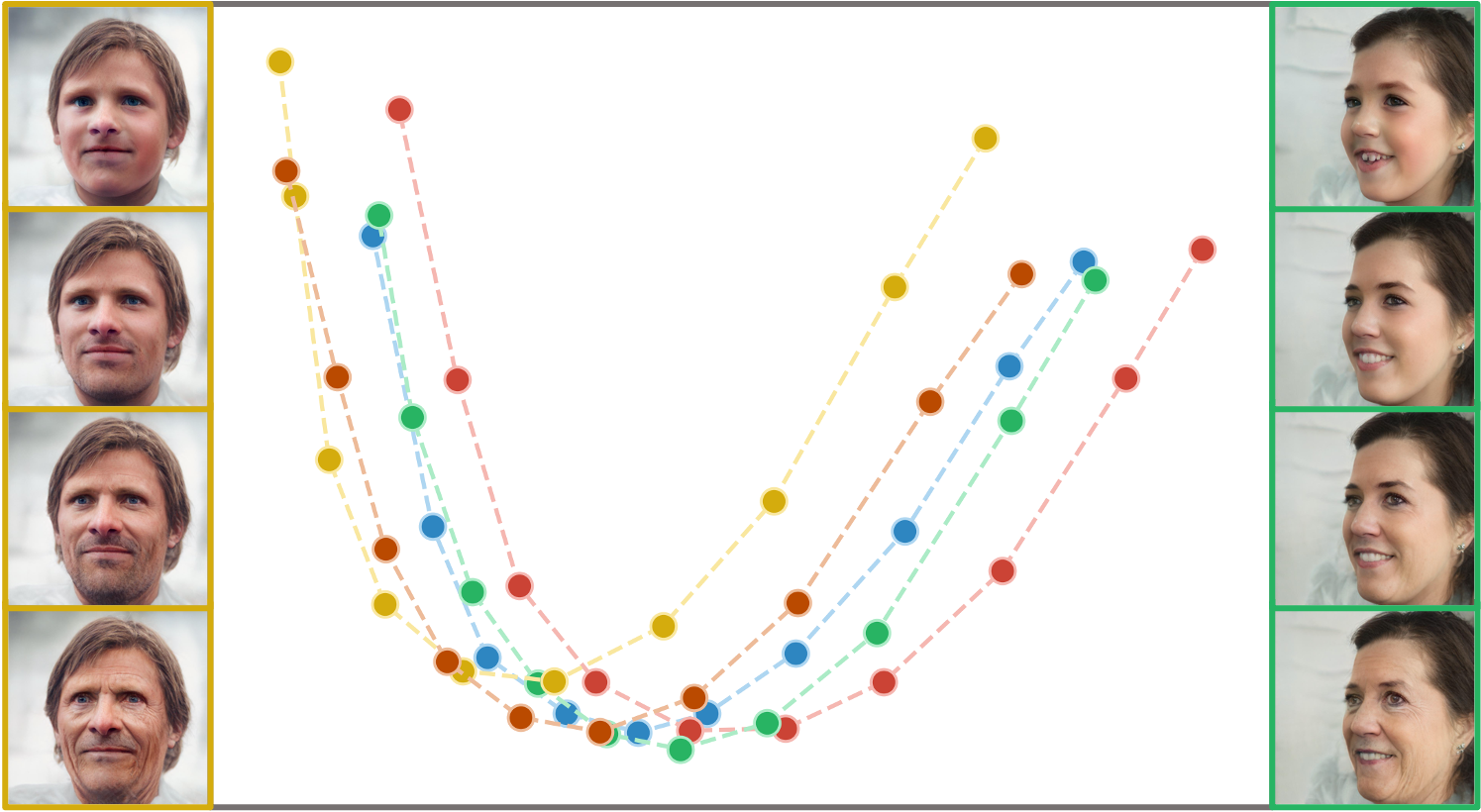}
        \caption{}
        \label{fig:pca_multiple_paths}
    \end{subfigure}%
    \hspace{0.05cm}
    \begin{subfigure}{0.49\textwidth}
        \centering
        \includegraphics[width=0.95\linewidth]{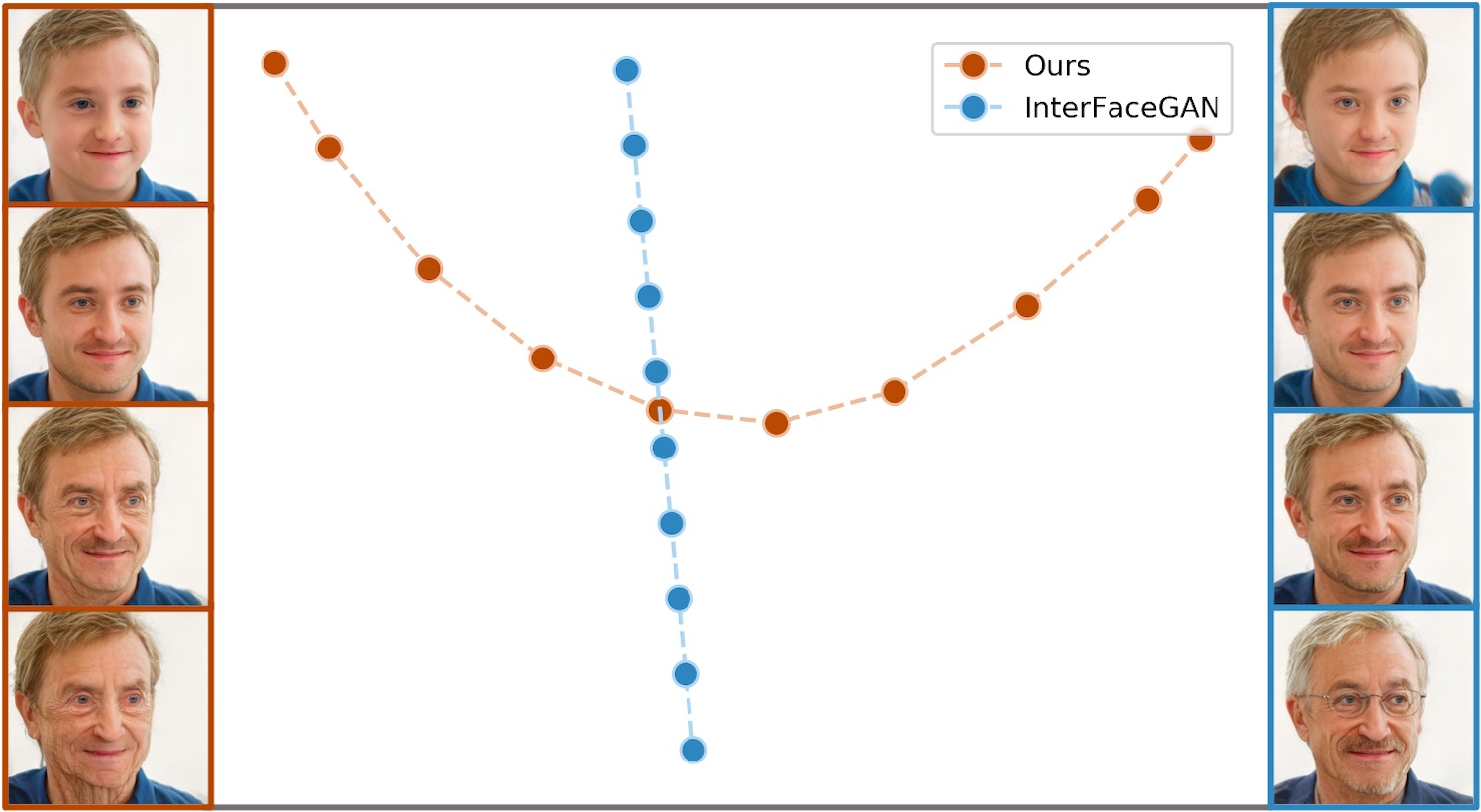}
        \caption{} 
        \label{fig:pca_latent_path_interfacegan}
    \end{subfigure}
    \caption{In (\subref{fig:pca_multiple_paths}) we project the learned latent paths of five different images using PCA. Illustrated on the sides are intermediate outputs of two of the five images obtained by traversing along the corresponding colored path.
    As can be seen, the non-linearity of the learned paths are better suited to the complex nature of StyleGAN's latent space manifold. In (\subref{fig:pca_latent_path_interfacegan}) we compare the linear nature of InterFaceGAN (shown in blue) with the non-linearity of SAM (shown in red).}
    \label{fig:pca_latent_path}
\end{figure*}

%% file: figures/patch_editing.tex
\begin{figure}
    \centering
    \setlength{\belowcaptionskip}{-2.5pt}
    \setlength{\tabcolsep}{2pt}
    \begin{tabular}{cccc}
        \includegraphics[width=0.0900\textwidth]{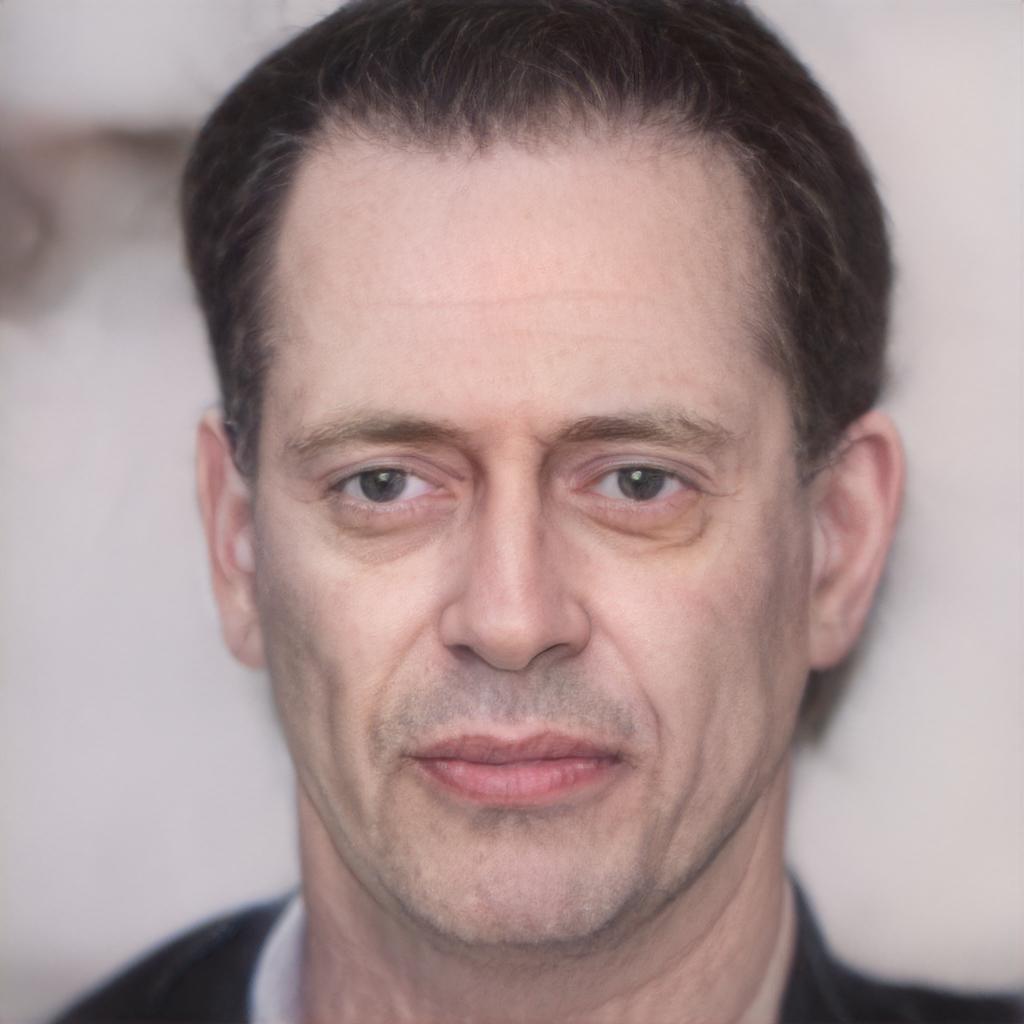} &
        \includegraphics[width=0.0900\textwidth]{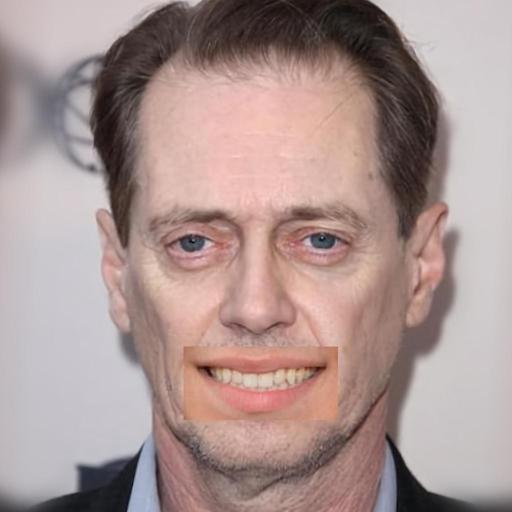} &
        \includegraphics[width=0.0900\textwidth]{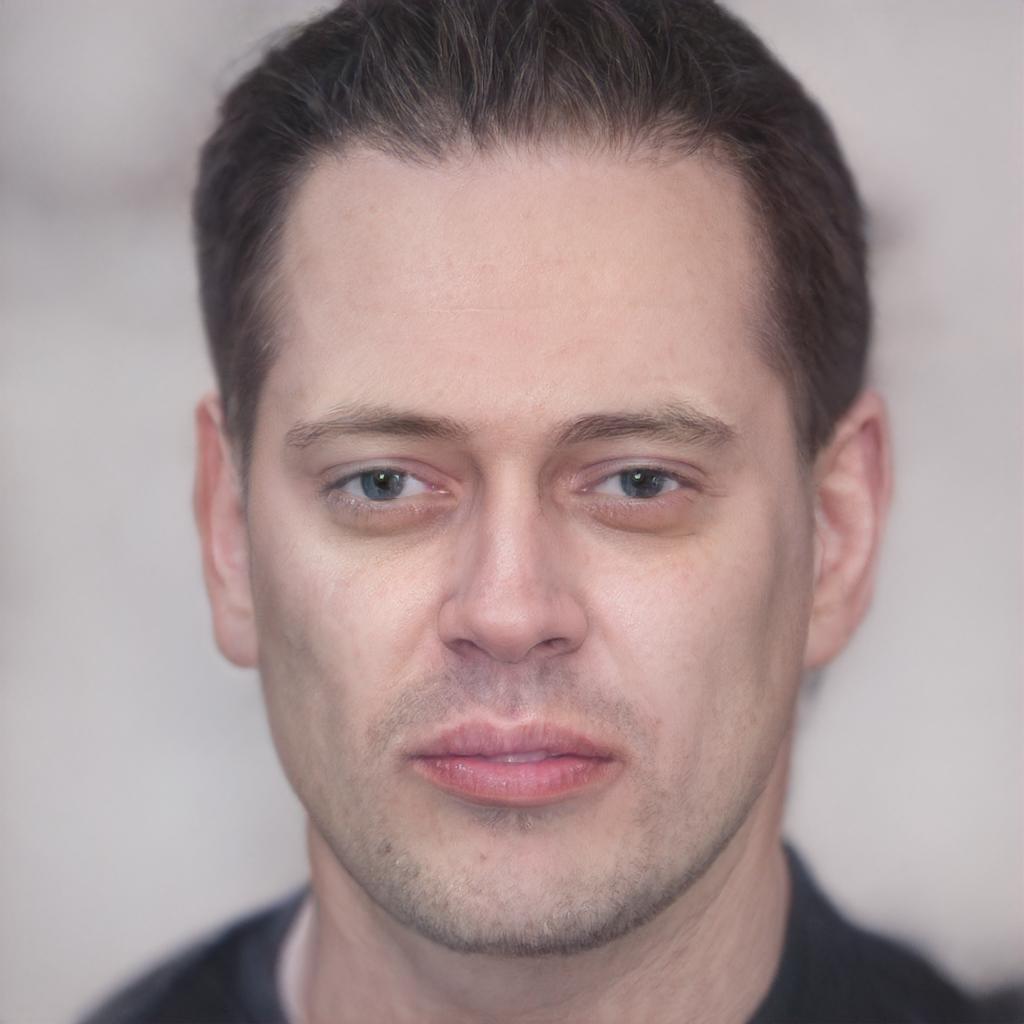} &
        \includegraphics[width=0.0900\textwidth]{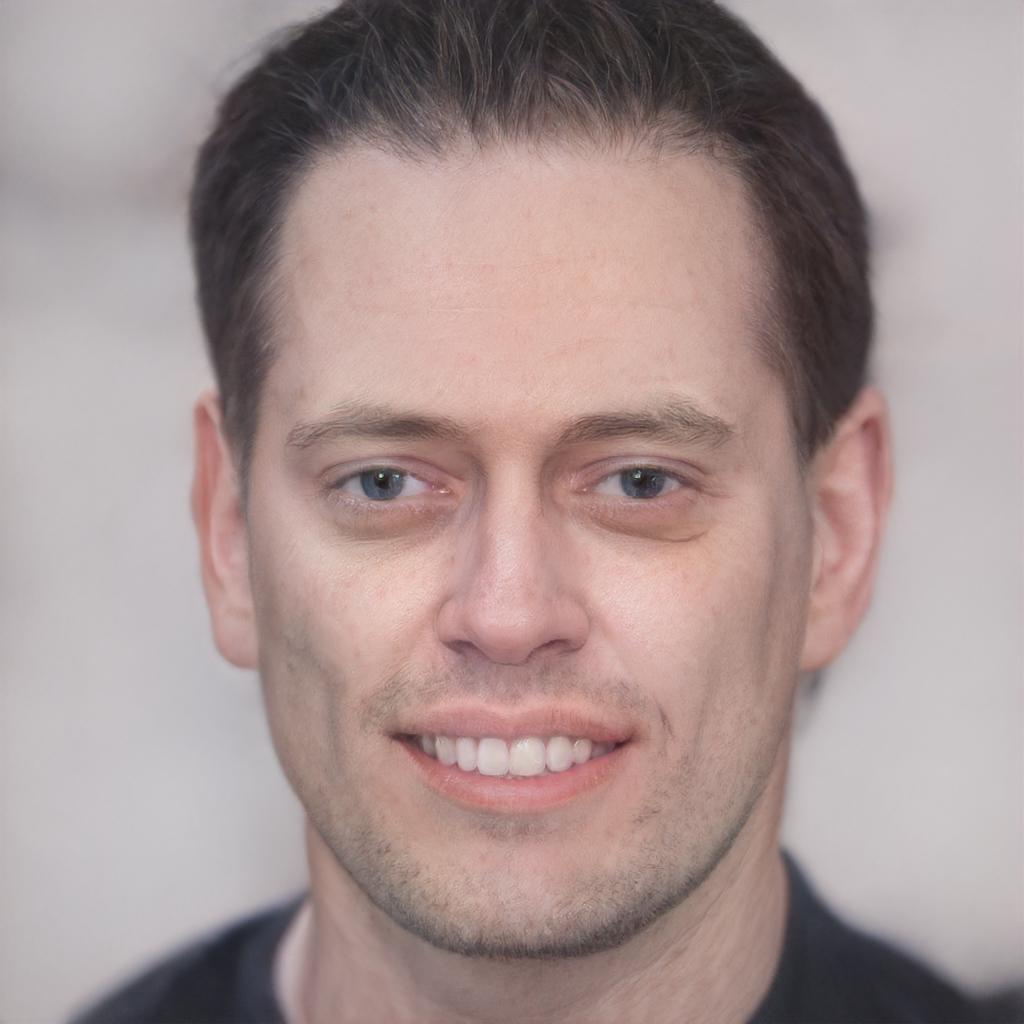}
        \tabularnewline
        \includegraphics[width=0.0900\textwidth]{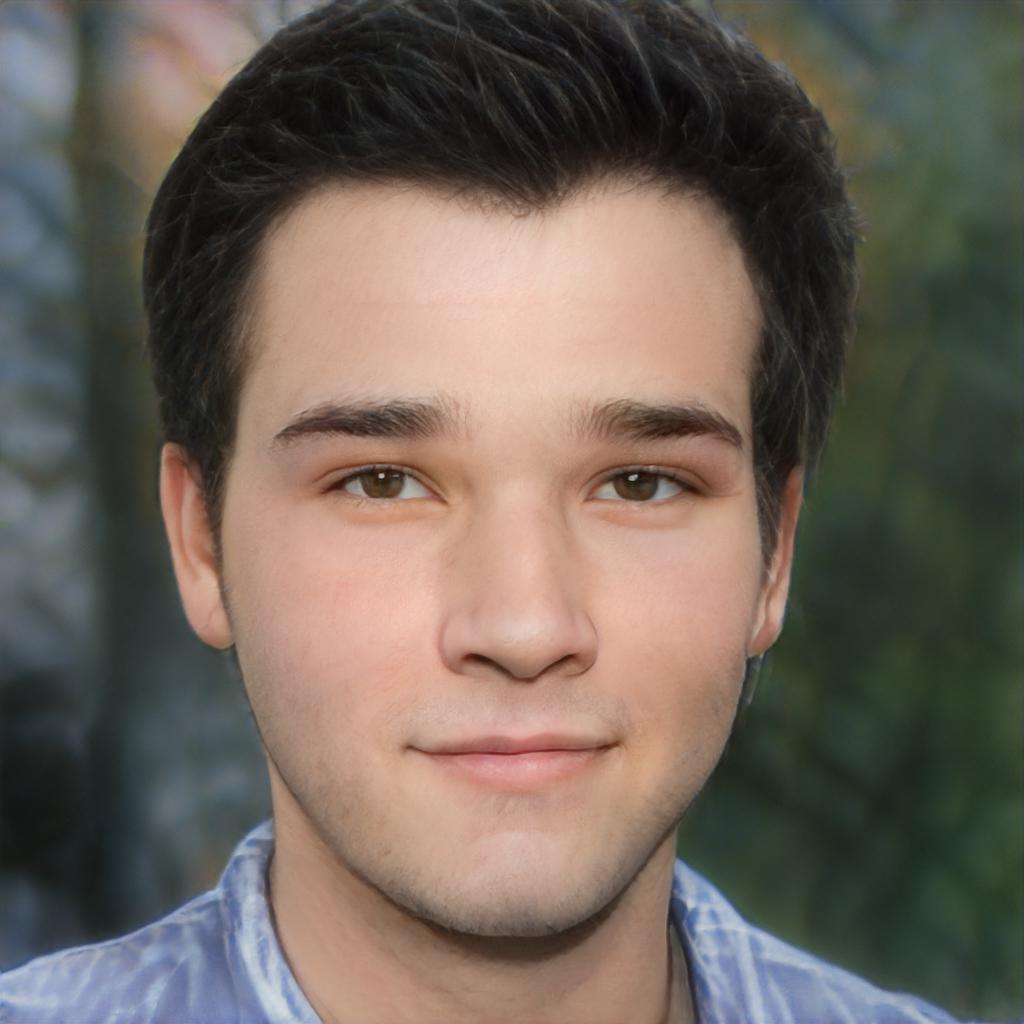} &
        \includegraphics[width=0.0900\textwidth]{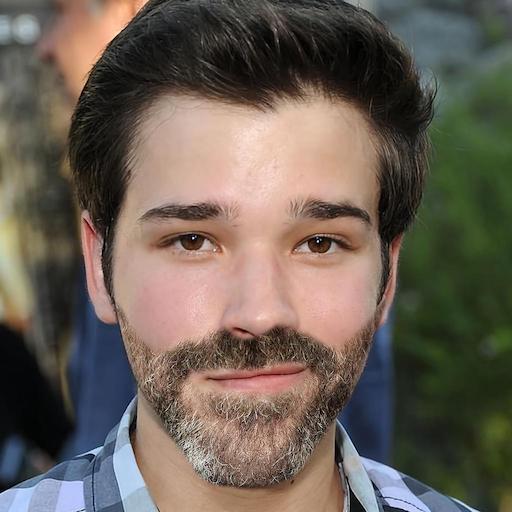} &
        \includegraphics[width=0.0900\textwidth]{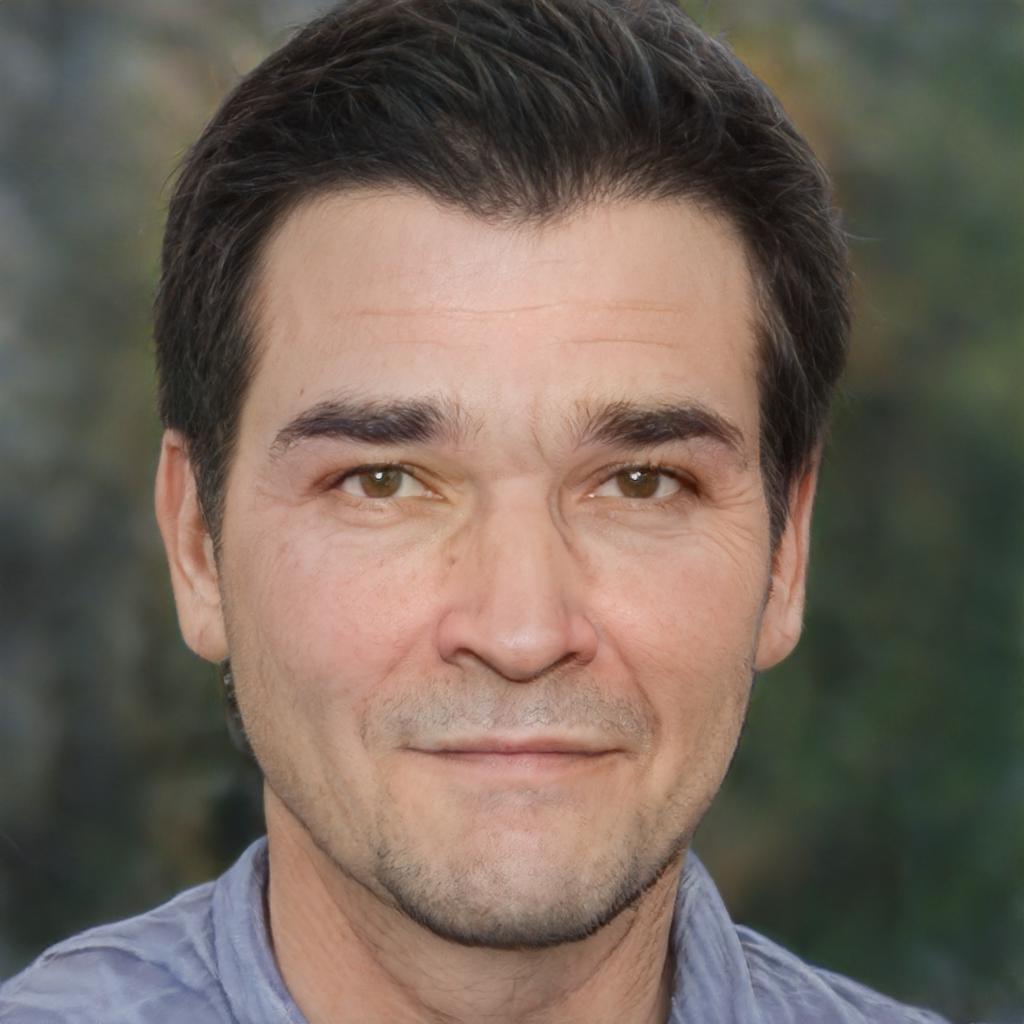} &
        \includegraphics[width=0.0900\textwidth]{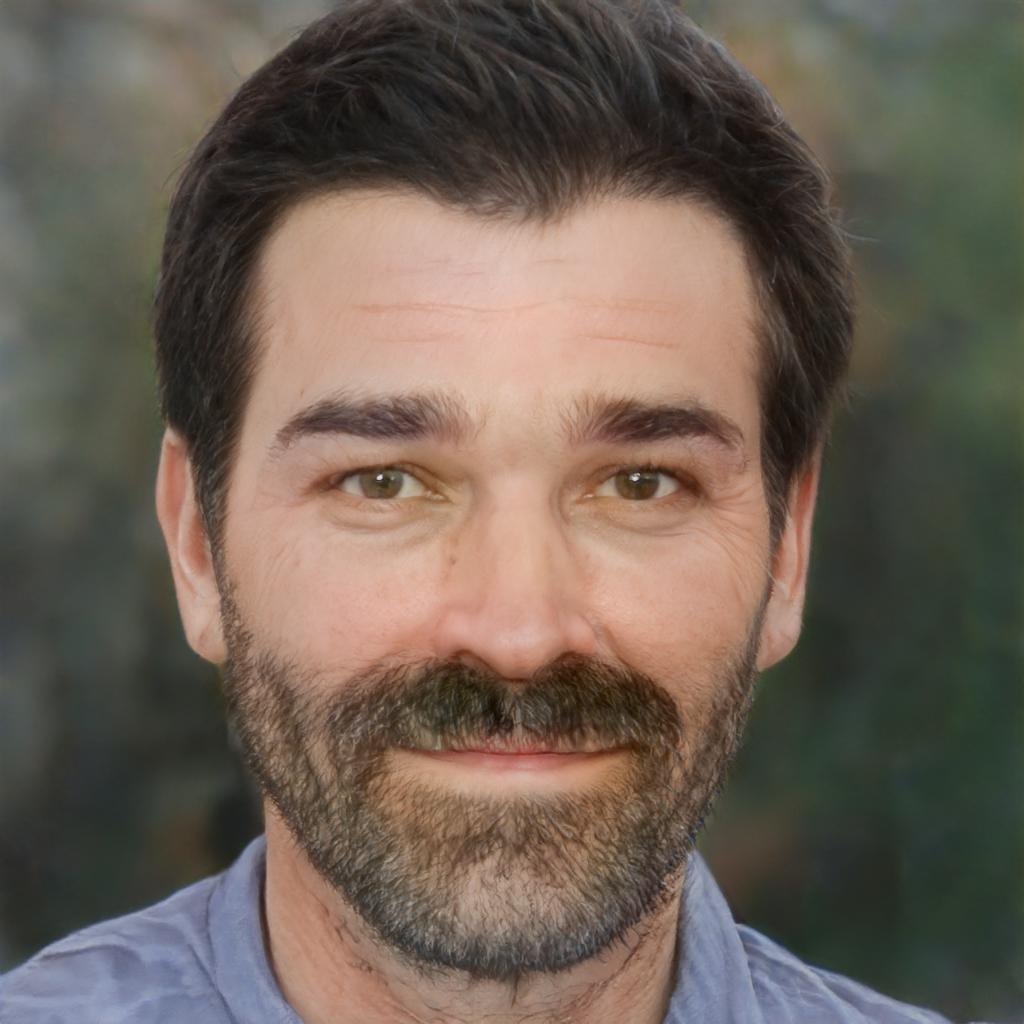}
        \tabularnewline
        \includegraphics[width=0.0900\textwidth]{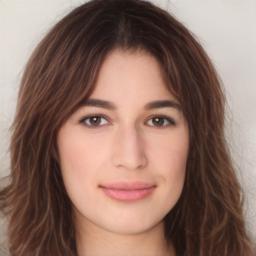} &
        \includegraphics[width=0.0900\textwidth]{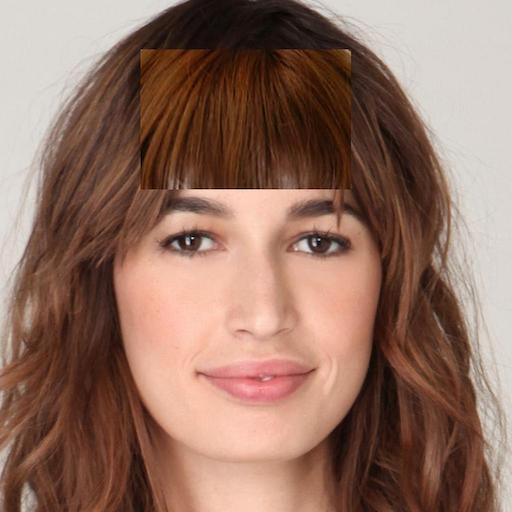} &
        \includegraphics[width=0.0900\textwidth]{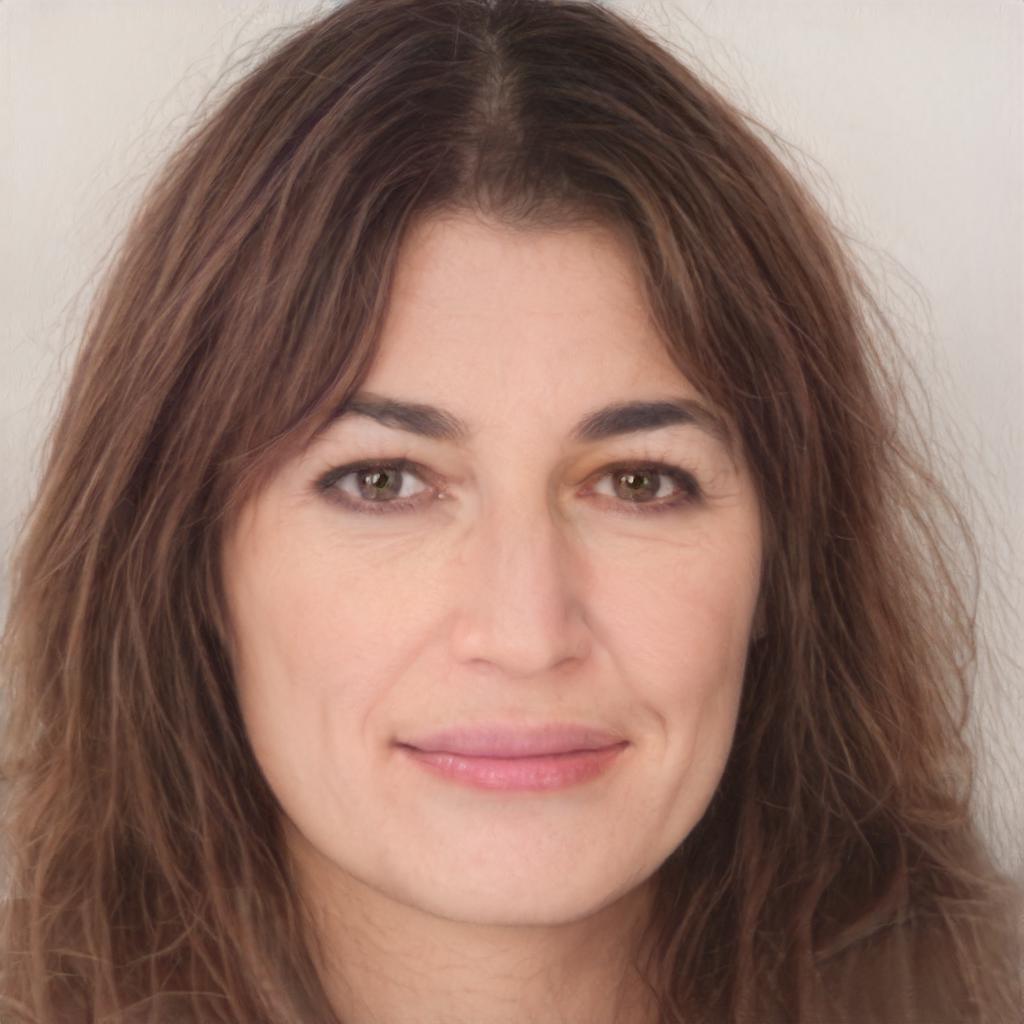} &
        \includegraphics[width=0.0900\textwidth]{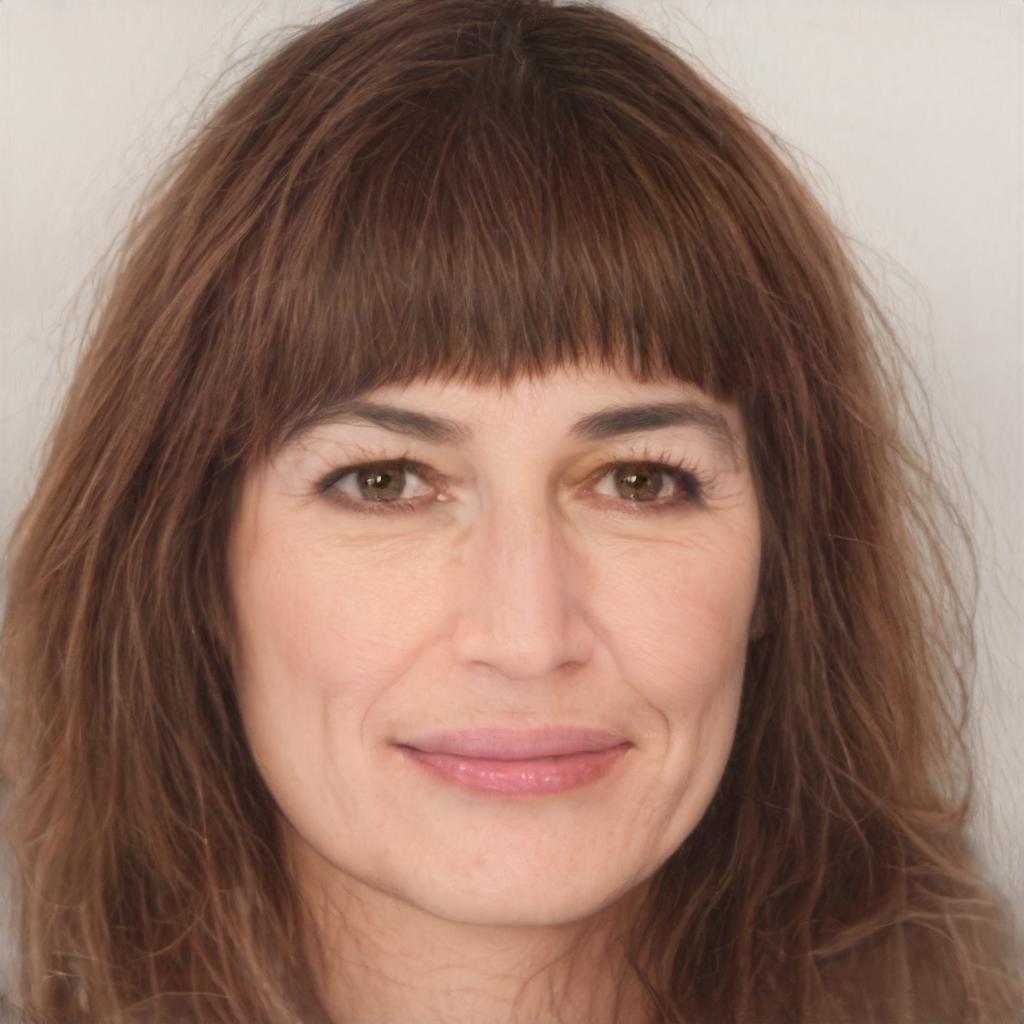}
        \tabularnewline
        \includegraphics[width=0.0900\textwidth]{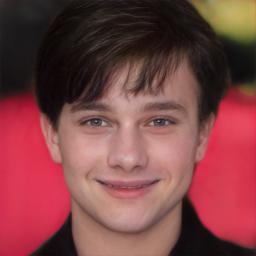} &
        \includegraphics[width=0.0900\textwidth]{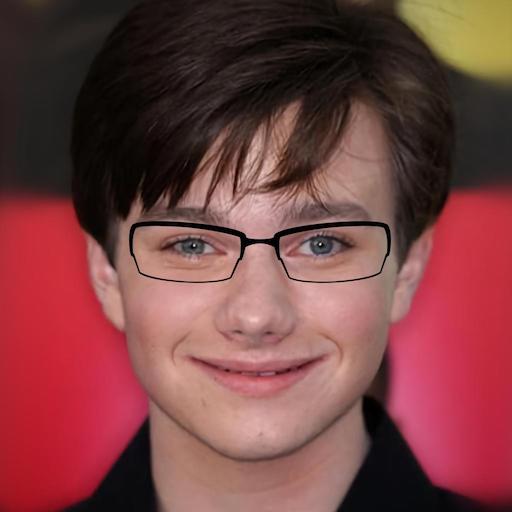} &
        \includegraphics[width=0.0900\textwidth]{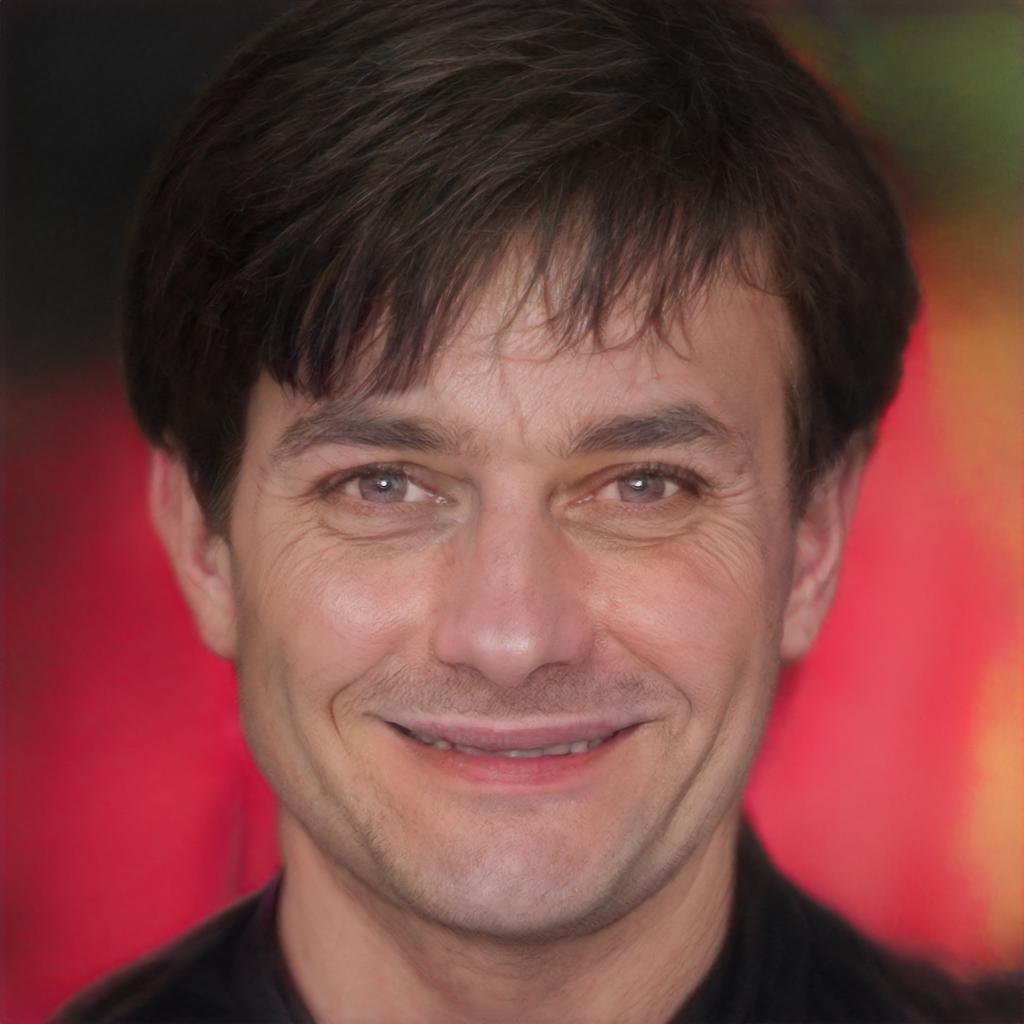} &
        \includegraphics[width=0.0900\textwidth]{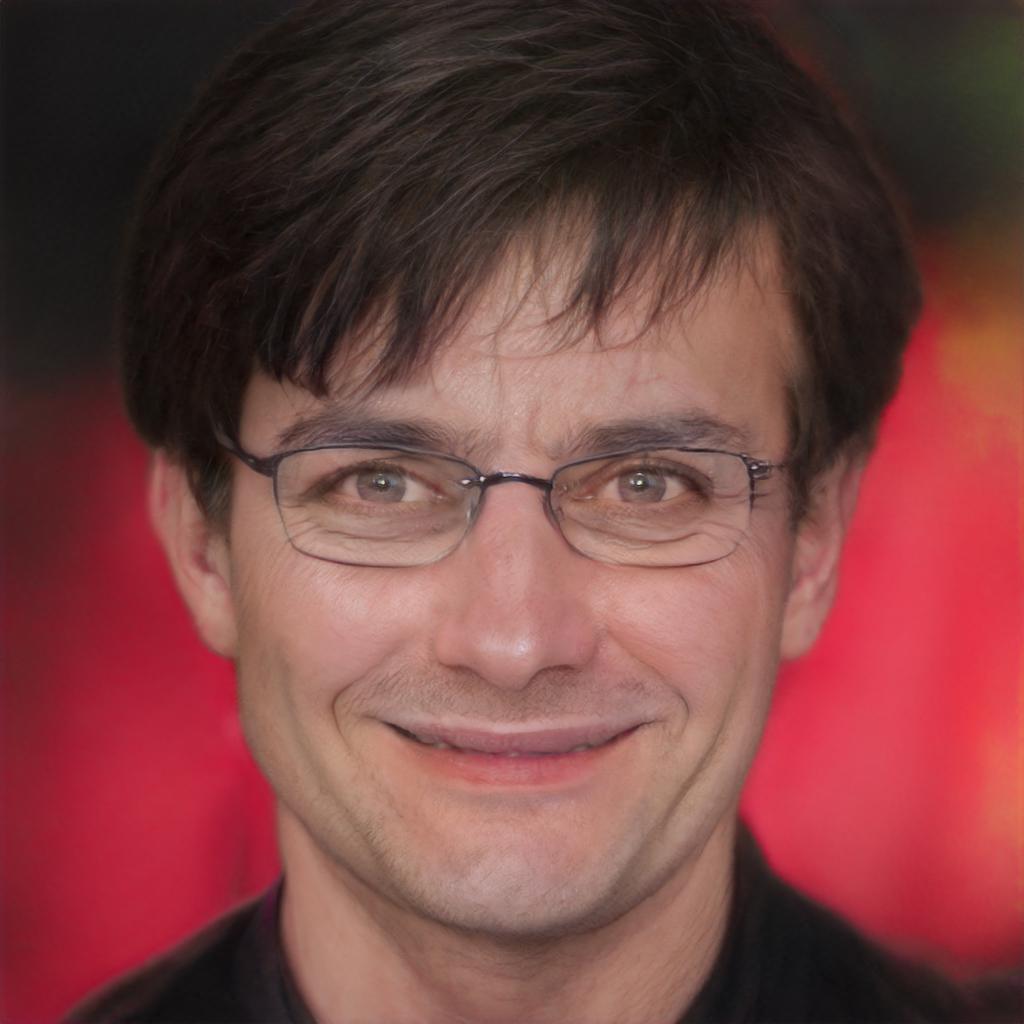}
        \tabularnewline
        
        \begin{tabular}{@{}c@{}}Inversion \end{tabular} &
        \begin{tabular}{@{}c@{}}Edited \\ Input\end{tabular} &
        \begin{tabular}{@{}c@{}}Inversion \\ Transformed\end{tabular} &
        \begin{tabular}{@{}c@{}}Edited \\ Transformed\end{tabular}
    \end{tabular}
    \setlength{\belowcaptionskip}{-10pt}
    \caption{SAM with patch editing allows for more fine-grained control over specific features such as expression, facial hair, hair style, and glasses.
    }
    \label{fig:patch_editing}
\end{figure}

%% file: figures/style_mixing.tex
\begin{figure}
    \centering
    \setlength{\belowcaptionskip}{-2.5pt}
    \setlength{\tabcolsep}{1pt}
    \begin{tabular}{cc}
        \includegraphics[width=0.09\textwidth]{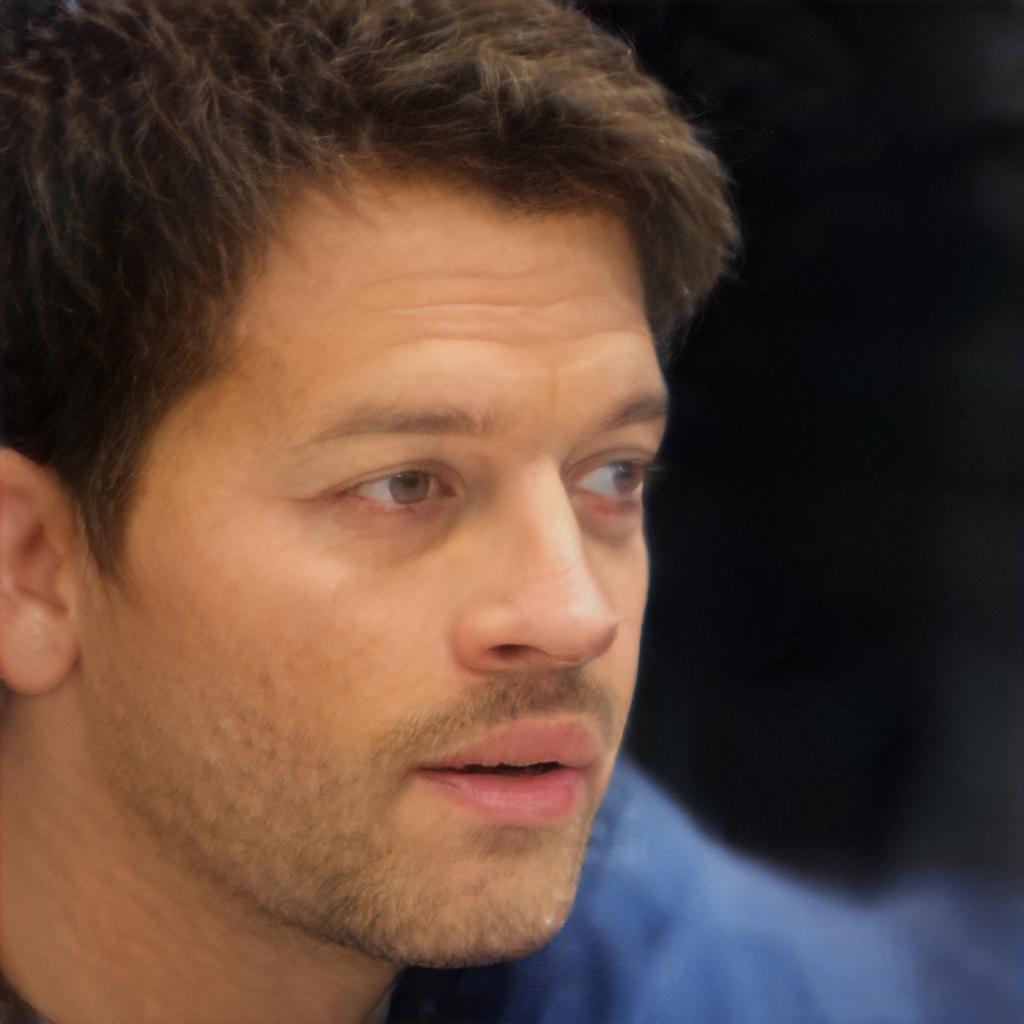} &
        \includegraphics[width=0.36\textwidth]{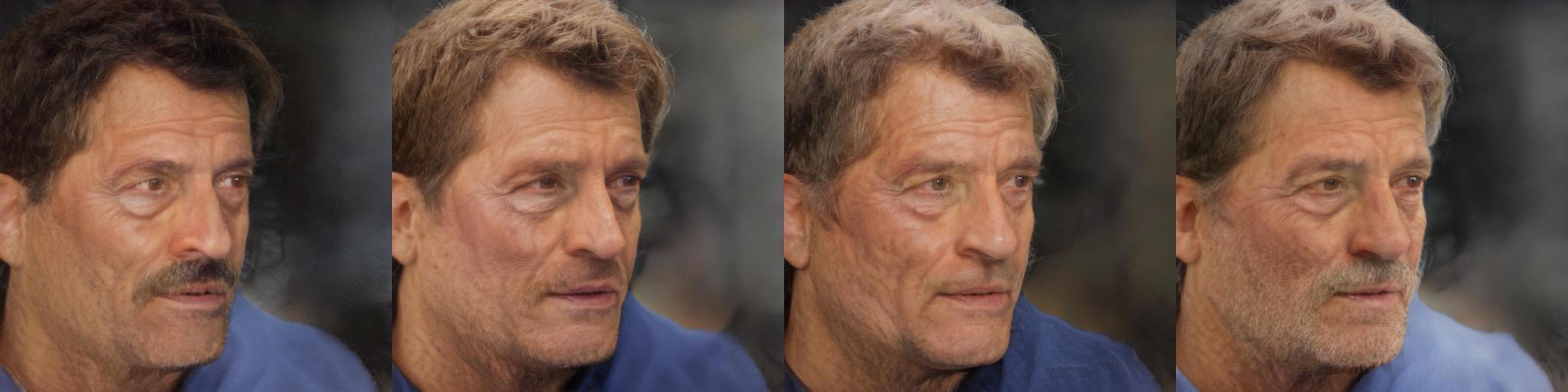}
        \tabularnewline
        \includegraphics[width=0.09\textwidth]{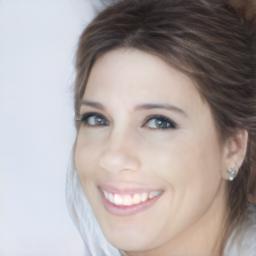} &
        \includegraphics[width=0.36\textwidth]{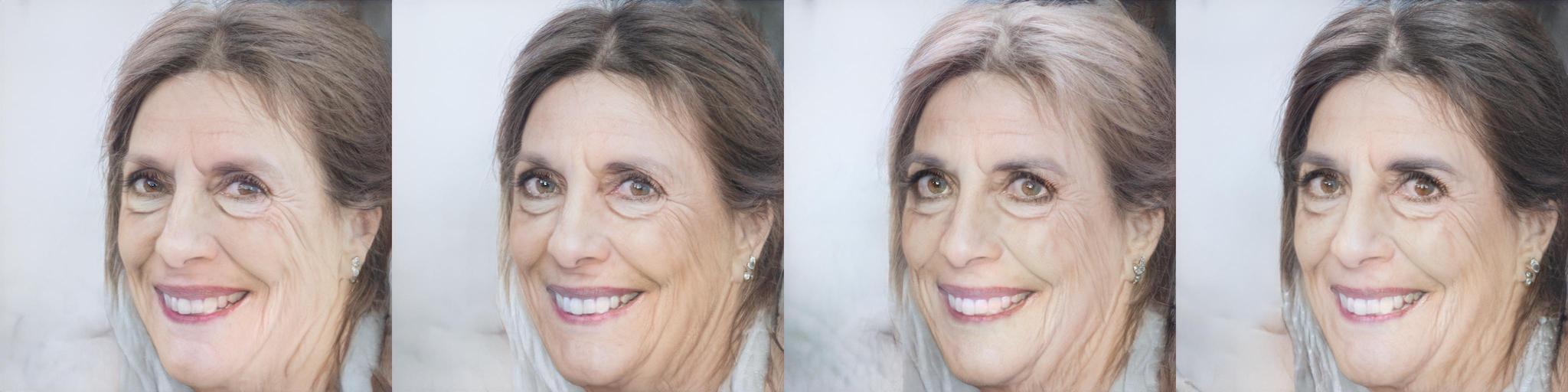}
        \tabularnewline
        \includegraphics[width=0.09\textwidth]{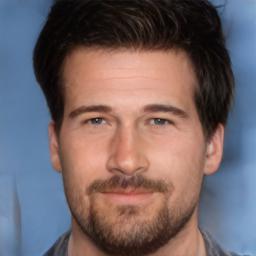} &
        \includegraphics[width=0.36\textwidth]{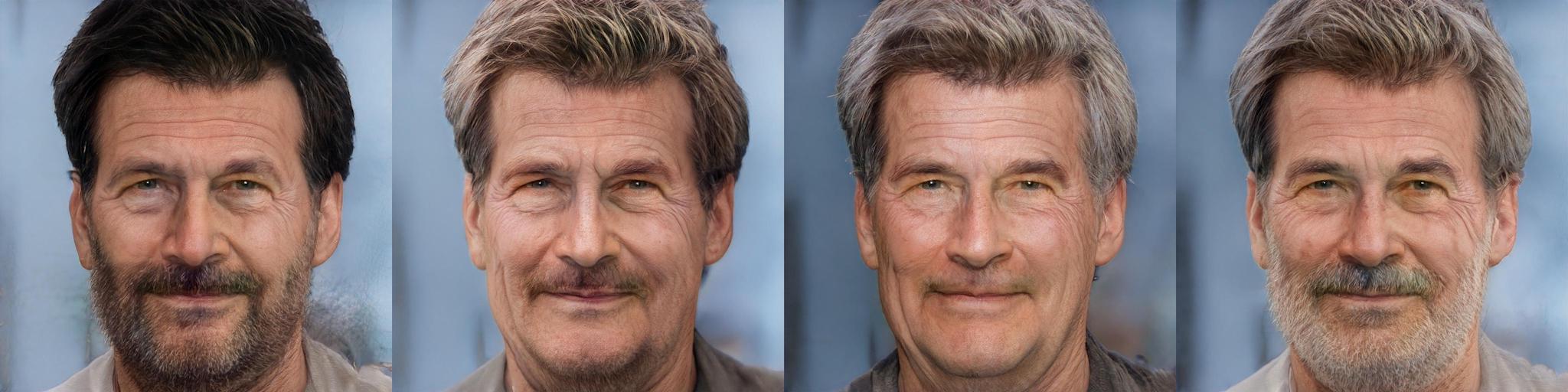}
        \tabularnewline
        \includegraphics[width=0.09\textwidth]{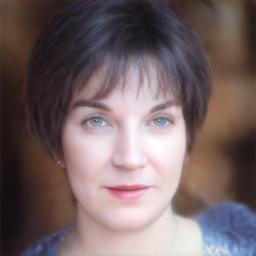} &
        \includegraphics[width=0.36\textwidth]{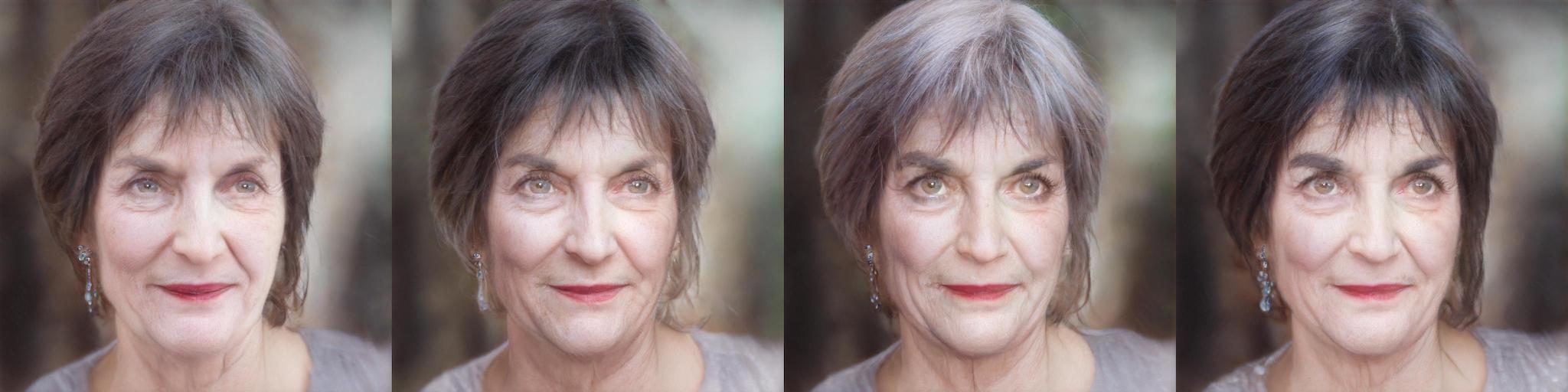}
        \tabularnewline
        \begin{tabular}{@{}c@{}}Inversion\end{tabular} &
        \begin{tabular}{@{}c@{}}Multi-Modal Outputs \end{tabular}
    \end{tabular}
    \setlength{\belowcaptionskip}{-10pt}
    \caption{Performing style-mixing on the age-transformed outputs of SAM allows additional editing control over features such as hair color and facial hair. Here, for each input image we perform style-mixing with four references images on layers $8-9$ to obtain multi-modal results.}
    \label{fig:style_mixing}
\end{figure}

%% file: conclusions.tex
\section{Discussion and Limitations}~\label{limitations}
Although our suggested approach is effective in representing the age transformation process, there are several limitations that should be considered. Although using a fixed, pre-trained StyleGAN generator simplifies the training process and allows for generating high-quality images, doing so may make it challenging to effectively model extreme poses, challenging expressions, and accessories. 
Furthermore, as our results are governed by the style representation, our method is limited to images that can be accurately embedded into StyleGAN's latent space. Thus, modeling faces that reside outside the StyleGAN domain, such as that of Uncle Sam, may be challenging. Likewise, embedding an image into a set of vectors may make it more difficult to faithfully preserve input features such as the image background.

In our evaluations, we showed how our proposed method successfully disentangles age and other attributes such as hair color and hair style. Yet, such attributes naturally change as we age. To model these changes, we therefore proposed two editing techniques for controlling global changes (e.g., hair color) and local changes (e.g., the presence of eye glasses and facial hair). With that, capturing more complex changes such as receding hair lines and changes in skin color still remain a challenge with methods available today in part because they may be more difficult to capture. 

Another important assumption of our method is the existence of an age predictor that is able to generalize well to all age groups. As most age-annotated datasets are heavily biased toward images of adults, achieving high age prediction accuracy for young children may be challenging. As a result, our method may struggle to generate images of children under the age of 5. This limitation may be resolved by using a more accurate age predictor for computing the aging loss presented in Section~\ref{losses}. We present several challenging cases in Figure~\ref{fig:limitations}.

\input{figures/limitations}

\section{Conclusion}
This work presented a novel approach for modeling the age transformation task using a single input facial image.
Our end-to-end approach maps a given input image and desired target age to the latent space of a fixed StyleGAN generator. To model the aging process, we employ an age predictor network that guides the encoder in generating latent codes corresponding to the desired transformation, resulting in a \textit{learned} latent path representing age progression. We showed that treating age progression as a regression problem via the age predictor provides more fine-grained control over the output age. Further, unlike previous latent space approaches that control age using a prior assumption on the latent path, our approach encourages the model to learn a \textit{non-linear} path more suitable for disentangling age from other facial attributes. We then demonstrate how this improved disentanglement allows for further fine-grained editing on the generated aging results.
We believe that the key insights provided in this work can be extended for additional editing applications. We leave this exploration as a potential direction for future work.

%% file: figures/limitations.tex
\begin{figure}
    \centering
    \includegraphics[width=\linewidth]{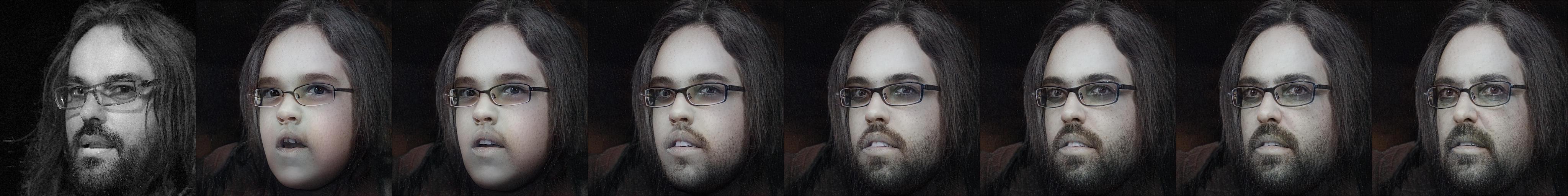} \\
    \includegraphics[width=\linewidth]{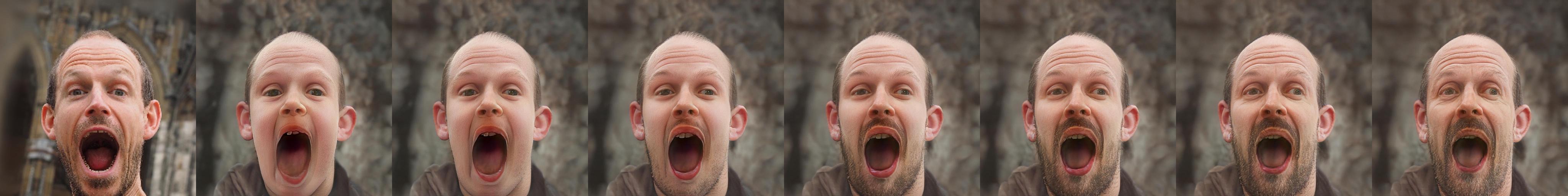} \\
    \includegraphics[width=\linewidth]{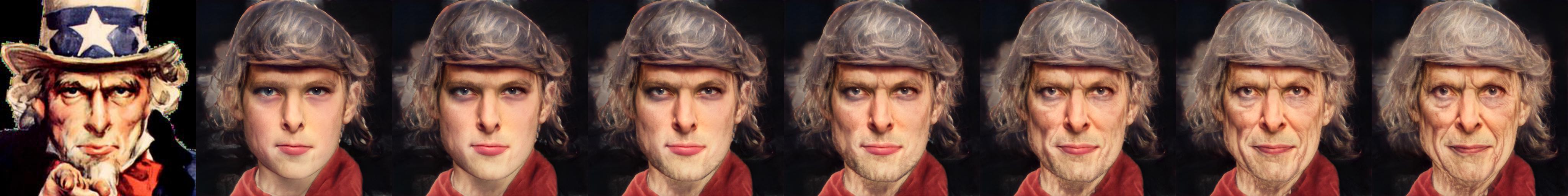} 
    \caption{Limitations of SAM. Our approach may struggle when faced with extreme expressions or out-of-domain inputs not seen during the training of StyleGAN.}
    \label{fig:limitations}
\end{figure}

%% file: appendix.tex
\section{Implementation Details}~\label{implementation_details}

\paragraph{Architectures} As in~\cite{richardson2020encoding}, we use a pre-trained ResNet-IR architecture from~\cite{deng2019arcface} for our encoder backbone. We use a \textit{fixed} StyleGAN2 generator pre-trained on the FFHQ~\cite{karras2019style} dataset. Identity embeddings are extracted using a pre-trained ArcFace network from~\cite{deng2019arcface}. As in~\cite{yao2020high}, we use the age classifier from~\cite{Rothe-ICCVW-2015, Rothe-IJCV-2018} pre-trained on the IMDB-WIKI dataset. As the IMDB-WIKI dataset has very few images below the age of $10$ or above the age of $60$, we fine-tune the age classifier using the FFHQ-Aging dataset~\cite{orel2020lifespan}. 

\paragraph{Training}
For each batch, we randomly set the target ages equal to the estimated source ages with probability $0.33$, focusing the network on reconstructing the input images. For training the encoder, we use the Ranger optimizer, a combination of Rectified Adam~\cite{liu2019variance} with Lookahead~\cite{zhang2019lookahead}, with a constant learning rate of $0.001$. The input image resolution is $256\times256$. Finally, only horizontal flips are used for augmentations. All experiments are performed using a single NVIDIA Tesla P40 GPU with a batch size of $6$.

\input{figures/appendix/multi_domain_comparison}

\paragraph{Losses} Before computing the loss functions, the generated $1024\times1024$ images are resized to $256\times256$. For $\mathcal{L}_{\text{ID}}$, the images are cropped around the face region and resized to $112\times112$ before being fed into the facial recognition network. Similarly, for $\mathcal{L}_{\text{age}}$, the images are resized to $224\times224$ before being fed into the age classifier. 
Further, we set the loss lambdas as follows: $\lambda_{l2}$ and $\lambda_{lpips}$ are set to $\lambda_{l2}=1$ and $\lambda_{lpips}=0.6$ in the center region of the image and $\lambda_{l2}=0.25$ and $\lambda_{lpips}=0.1$ in the outer region.
We additionally set $\lambda_{reg}=0.005$, $\lambda_{id}=0.1$, and $\lambda_{age}=5$. Finally, the cycle loss coefficient is set to $\lambda_{cycle}=1$. 

\vspace{0.5cm}
\section{Comparison with Multi-Domain Methods}~\label{multi_domain_comp}
Here, we compare our method with the current state-of-the-art in multi-domain image-to-image translation: FUNIT~\cite{liu2019few} and StarGANv2~\cite{choi2020stargan}. These methods work by translating a source image based on a reference image taken from the target domain. In the case of aging, the reference image should guide the network in translating only the age of the source to resemble an image from the target domain. 
We retrain each of the alternative methods using the same age domains used in LIFE~\cite{orel2020lifespan} and use the default training settings. 

\input{figures/appendix/ablation_qualitative}

\paragraph{Qualitative Evaluation.}
We qualitatively compare the three methods in Figure~\ref{fig:comparison_multi_domain}. For performing inference on FUNIT and StarGANv2, we randomly sample a reference image from the FFHQ dataset and translate the source image using the selected reference image.
When performing inference using our approach, we set the target age equal to the median age of each group.
As can be observed, in both methods the reference image has a large effect on the translated image and alters various attributes besides the age. For example, one can see that the texture of the translated images is mostly taken from the reference image, resulting in unrealistic changes in skin-tone. As a result, since a different reference image is used for each age group, the generated images may vary significantly. This is undesirable when modeling a continuous process such as human aging. 
This behavior may be explained by the design of these two methods. Both are built upon the AdaIN~\cite{huang2017arbitrary} layer, which mostly alters the texture of the source image based on the chosen reference image. Although this is desirable in tasks such as style transfer, the opposite is desired here where the texture, pose, lighting, etc. should be taken from the source. 
Our method is able to better preserve the input identity and is able to better handle non-frontal images. Overall, our results have a much higher visual quality with substantially fewer artifacts.

\vspace{0.65cm}
\section{Ablation Study}~\label{ablation}
In this section we perform an ablation study on the training formulation of SAM and the loss objectives employed during training. 

\vspace{0.15cm}
\paragraph{Training Formulation.}
We begin by validating the training formulation of SAM.
Recall that the output of SAM is defined as
\begin{equation}
    SAM(\textbf{x}_{age}) := G(E(\textbf{x}_{age}) + \textbf{w}^*)).
\end{equation}
Here, $\textbf{x}_{age} = \textbf{x} \concat \alpha_t$ is a 4-channel input defining the desired age transformation on input image $\textbf{x}$ and $\textbf{w}^*$ is the inversion of $\textbf{x}$ obtained using a fixed, pre-trained Pixel2Style2Pixel~\cite{richardson2020encoding} encoder. That is,
\begin{equation}
    \textbf{w}^* := pSp(\textbf{x}) \in \mathbb{R}^{18 \times 512}.
\end{equation}
In this formulation, we train our SAM encoder to learn the \textit{residual} vector between the latent code of the original input image and the latent code of the age-transformed image. In a sense, SAM is tasked with learning a \textit{shift} in the latent space with respect to $\textbf{w}^*$.

Another possible approach for modeling age transformation is to directly learn the age-transformed image without relying on the inversion of the input image. That is, one can directly output
\begin{equation}
    SAM_{direct}(\textbf{x}_{age}) := G(E(\textbf{x}_{age}))).
\end{equation}
Note that for simplicity, we denote the two variants by $SAM$ and $SAM_{direct}$, respectively. 

\input{figures/appendix/ablation_aging_accuracy}

Here, we show that learning an ``age shift" in the latent space results in significant improvement in the identity preservation of the input image across the various target ages without harming the aging accuracy of our method. 

In Figure~\ref{ablation_qualitative} we provide a visual comparison of lifelong aging results generated using both of the approaches. Observe SAM's ability to better capture non-frontal poses, hair styles, and facial expressions. In particular, notice SAM's ability to more accurately retain the eye gaze direction of the bottom left input image and better model the hair style of the top right image.

To verify that our residual-based approach does not harm the aging accuracy of the transformations we repeat the quantitative evaluation performed in Section 4.2 and compare SAM and $SAM_{direct}$ with respect to five target ages: $5$,$20$,$35$,$50$, and $65$. For each target age, we compute the average difference between the desired age and the predicted age of each generated image. Full results are provided in Table~\ref{tb:ablation_aging_acc}. As shown, our residual-based approach does not restrict the SAM's ability to generate images across a wide range of ages. 

\input{figures/appendix/ablation_losses}

\paragraph{Loss Objectives.}
We now move to explore the importance of each of the loss objectives used in the training of SAM.
Specifically, we show the importance of introducing a cycle pass during training. We additionally show that the $\mathcal{L}_2$ and LPIPS losses should be employed in the forward pass as well as in the cycle pass. Finally, we explore the influence of the $w$-regularization on the generated outputs. 

In Figure~\ref{fig:ablation_losses}, we show a qualitative comparison comparing SAM when trained with and without each of the above losses. For each input image, we translate the image to three different target ages: $5$, $45$, and $85$. As can be seen, no one loss objective has a significant impact on the generated results. We first observe that when the $\mathcal{L}_2$ and LPIPS losses are not employed in the forward passes, SAM is better able to model changes in hair color (e.g., see example $2$).
However, we do find that our final configuration is able to better capture the younger age groups (see examples $2$ and $3$) and model changes to facial features such as wrinkles (see examples $1$ and $2$). Overall, we find that SAM is not overly sensitive to any one loss objective and the selected hyper-parameters as the results of all variants are plausible.

\section{Additional Results}~\label{additional_results}
The remainder of this document contains additional visual comparisons and results generated by SAM in the following Figures. Note that, as in the main manuscript, we are unable to display the original input images from the CelebA-HQ~\cite{liu2015faceattributes,karras2017progressive} dataset due to their licensing agreement. For each input, we therefore display the reconstruction image obtained by inverted each input using pSp~\cite{richardson2020encoding}.

\input{figures/appendix/appendix_life_comparison}
\input{figures/appendix/appendix_hrfae_comparison}
\input{figures/appendix/appendix_latent_space}
\input{figures/appendix/appendix_style_mixing}
\input{figures/appendix/appendix_patch_editing}
\input{figures/appendix/appendix_celebs}

\input{figures/appendix/appendix_full_lifespan}

%% file: figures/appendix/multi_domain_comparison.tex
\begin{figure}
    \centering
    \setlength{\belowcaptionskip}{-7.5pt}
    \setlength{\tabcolsep}{1pt}
    \begin{tabular}{c c c c c c c c c}
            Inversion & & & 3-6 & 15-19 & 30-39 & 50-69 & 70-120 \\
            \includegraphics[width=0.07\textwidth]{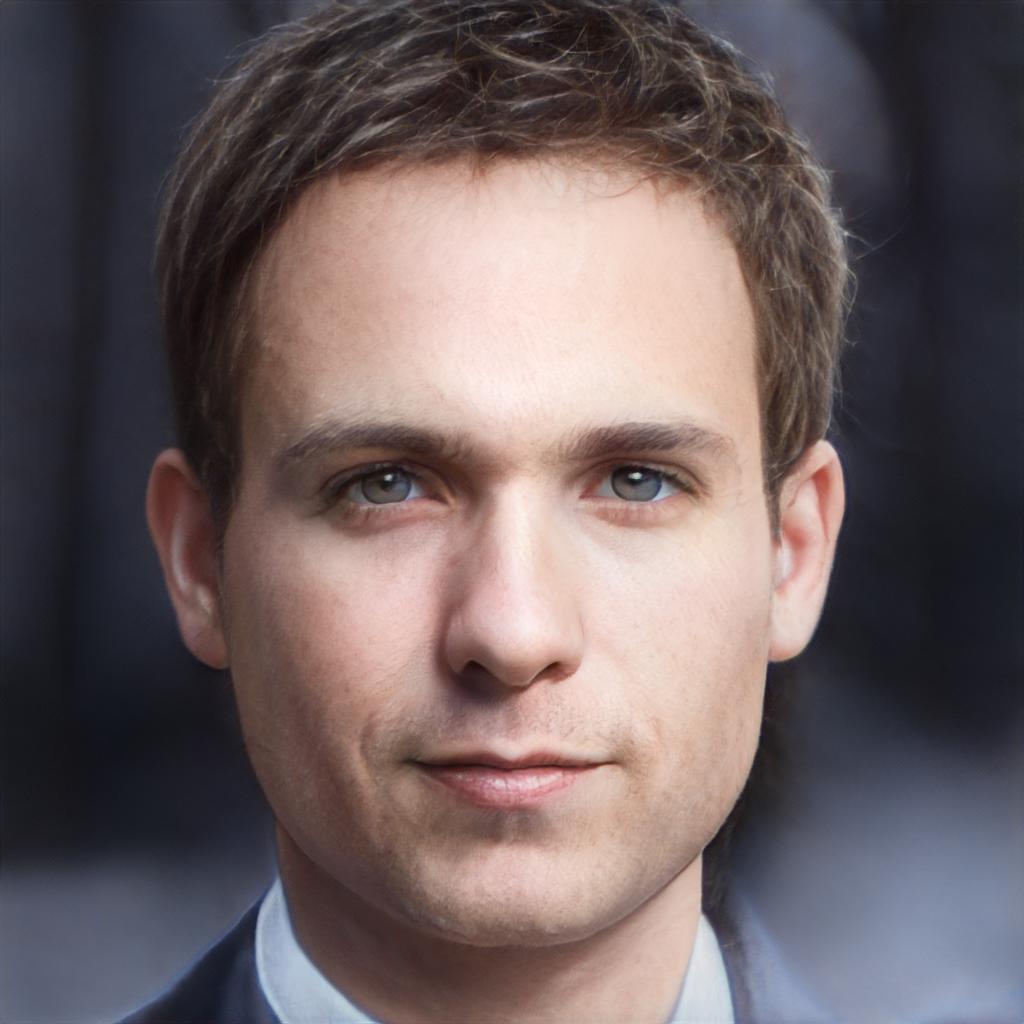} &
            & \raisebox{0.15in}{\rotatebox[origin=t]{90}{STAR}} & 
            \includegraphics[width=0.07\textwidth]{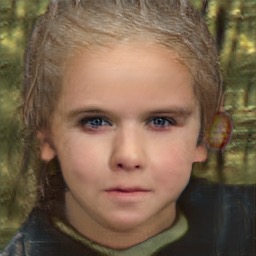} &
            \includegraphics[width=0.07\textwidth]{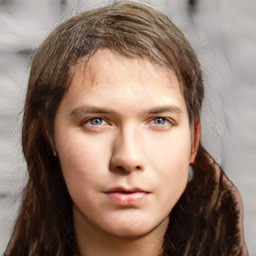} &
            \includegraphics[width=0.07\textwidth]{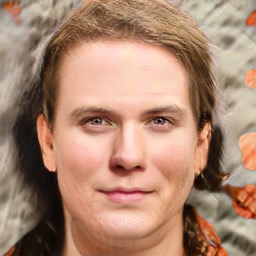} &
            \includegraphics[width=0.07\textwidth]{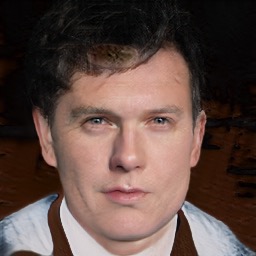} &
            \includegraphics[width=0.07\textwidth]{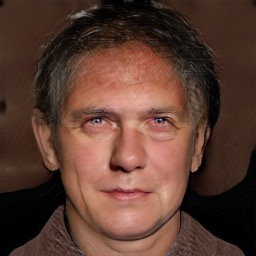} \\
            & & \raisebox{0.15in}{\rotatebox[origin=t]{90}{FUNIT}} & 
            \includegraphics[width=0.07\textwidth]{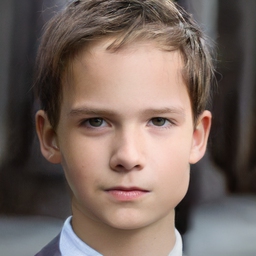} &
            \includegraphics[width=0.07\textwidth]{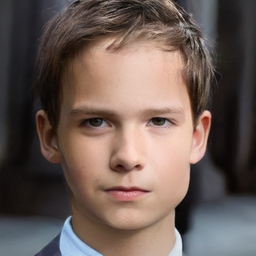} &
            \includegraphics[width=0.07\textwidth]{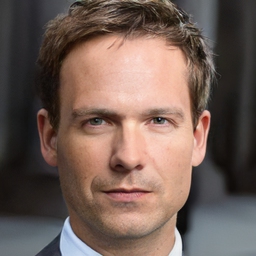} &
            \includegraphics[width=0.07\textwidth]{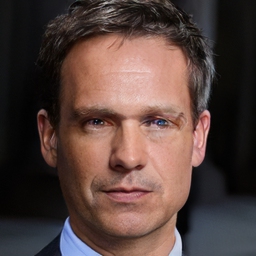} &
            \includegraphics[width=0.07\textwidth]{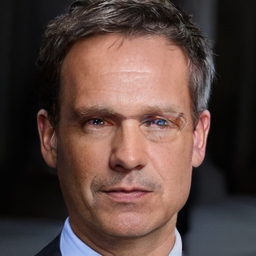} \\
            & & \raisebox{0.15in}{\rotatebox[origin=t]{90}{SAM}} &
            \includegraphics[width=0.07\textwidth]{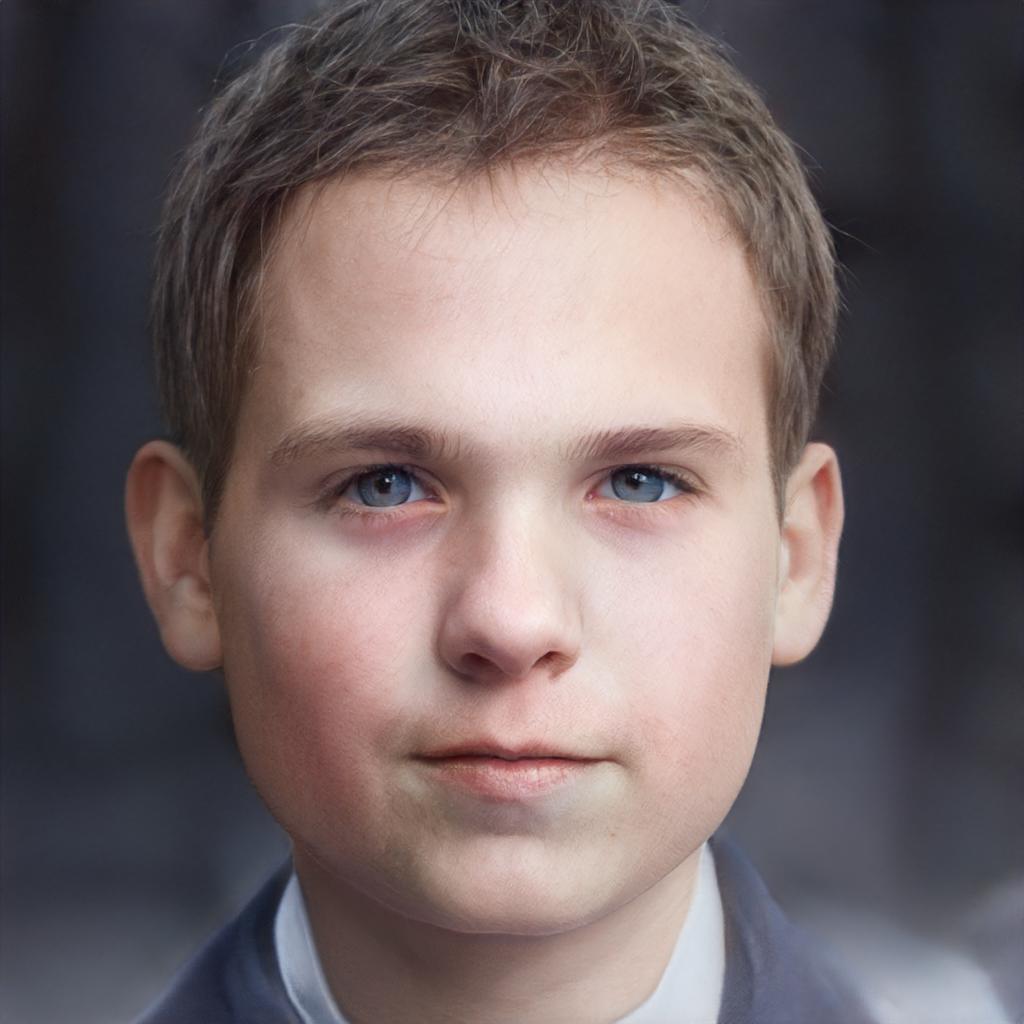} &
            \includegraphics[width=0.07\textwidth]{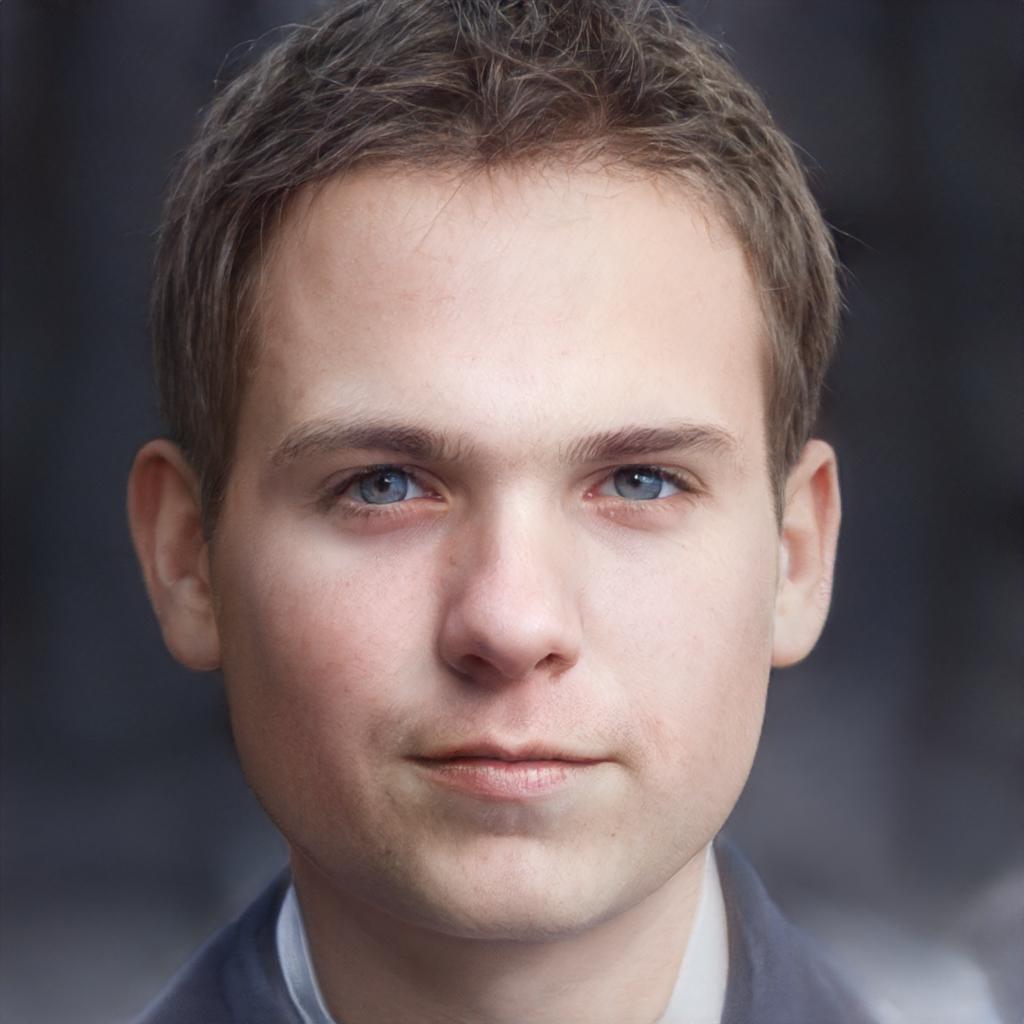} &
            \includegraphics[width=0.07\textwidth]{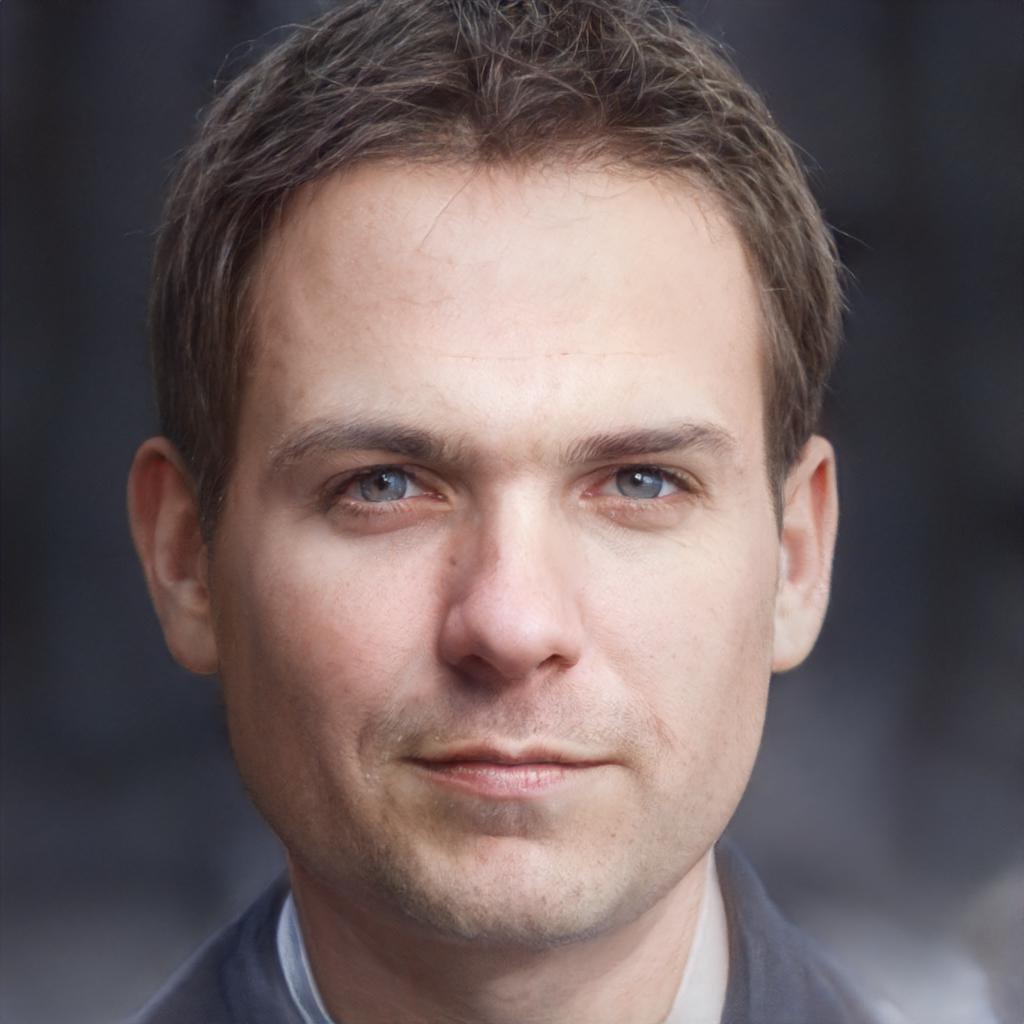} &
            \includegraphics[width=0.07\textwidth]{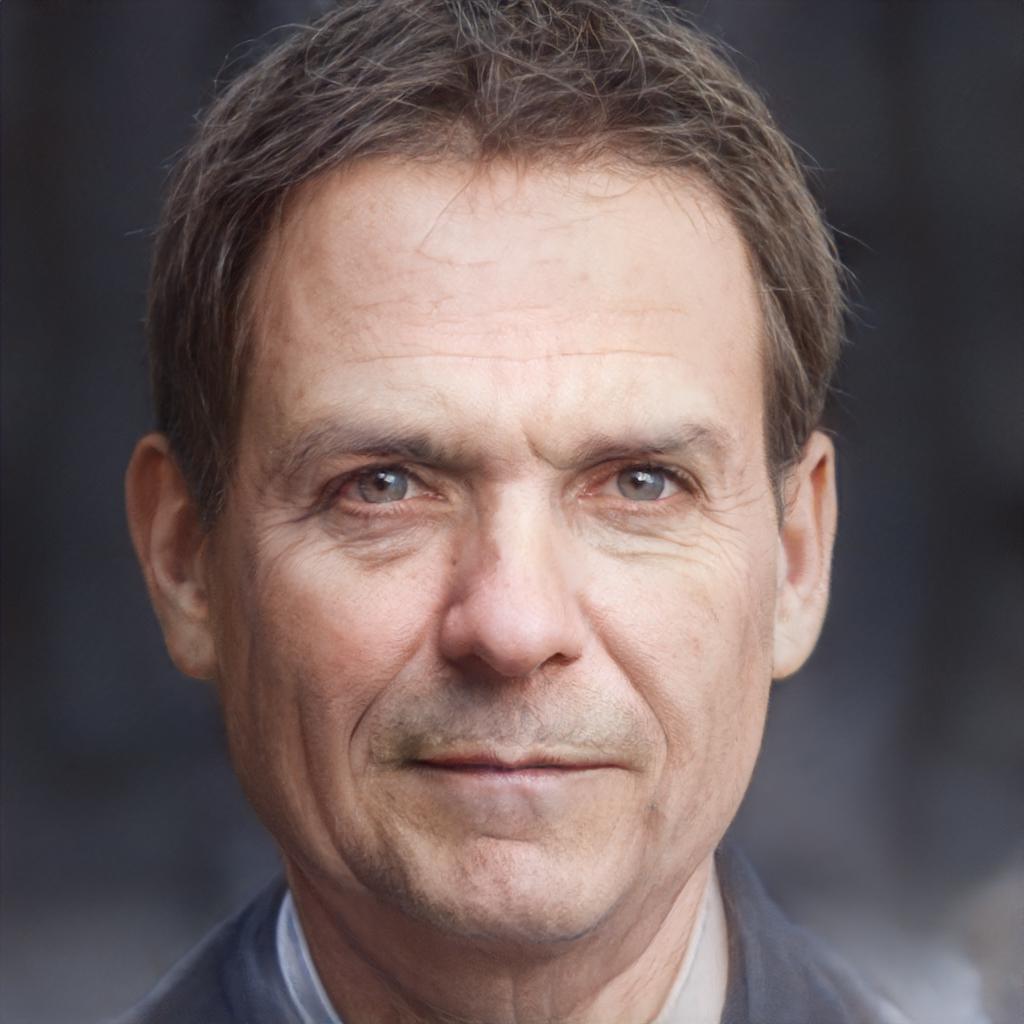} &
            \includegraphics[width=0.07\textwidth]{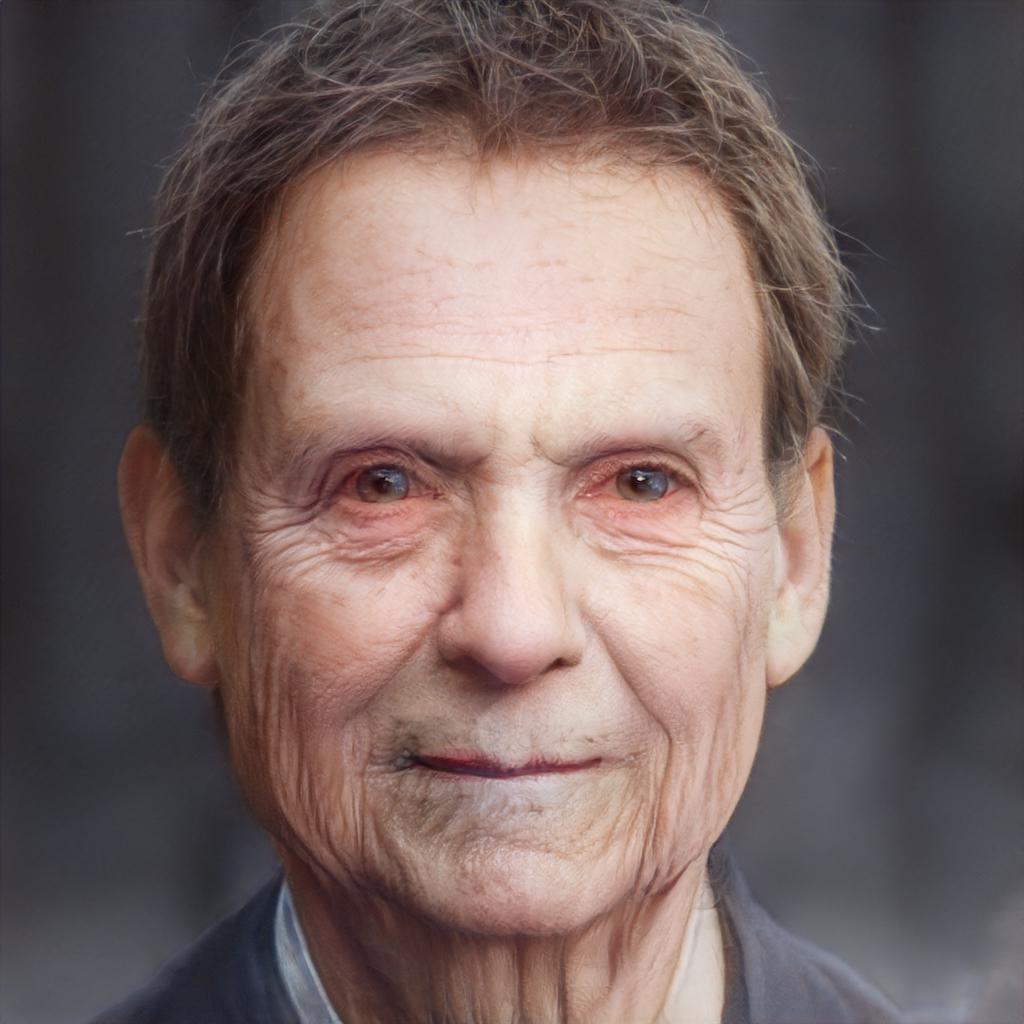}
            \tabularnewline

            \includegraphics[width=0.07\textwidth]{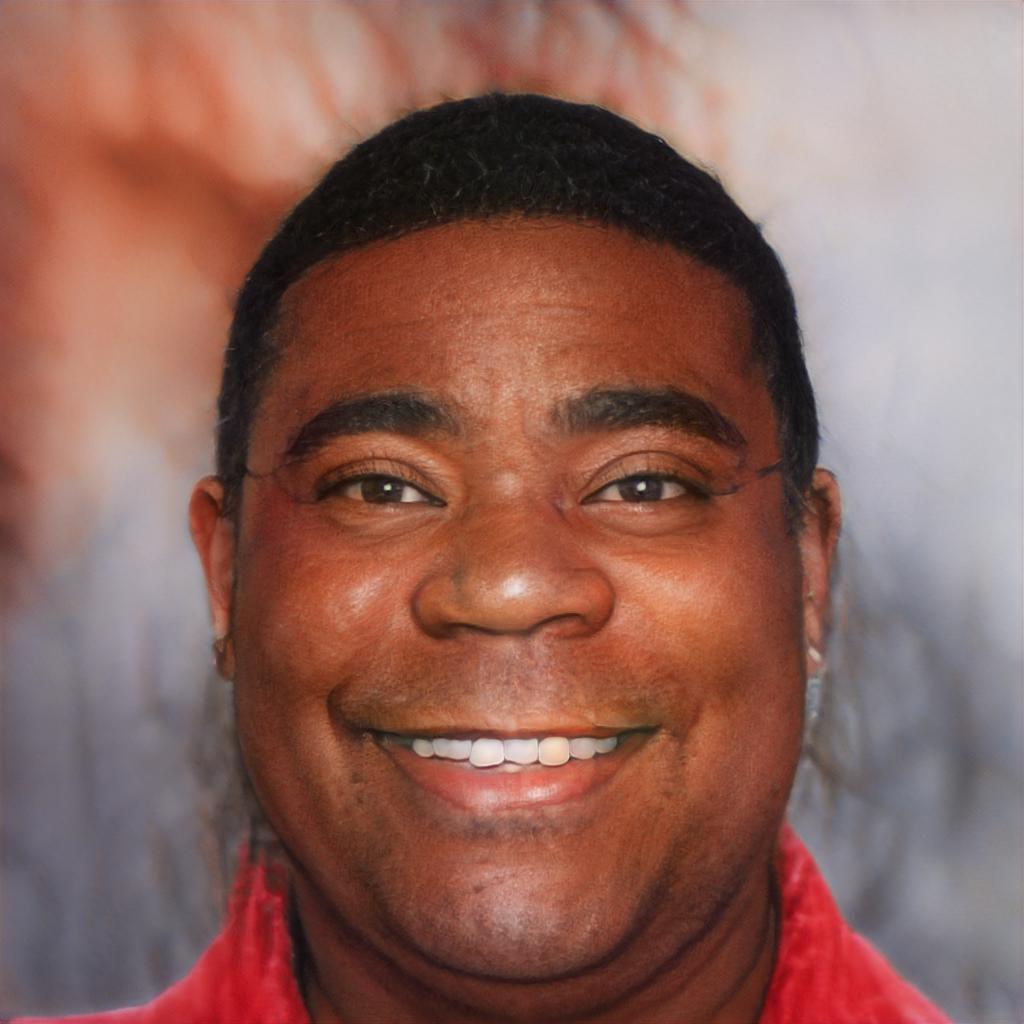} &
            & \raisebox{0.15in}{\rotatebox[origin=t]{90}{STAR}} & 
            \includegraphics[width=0.07\textwidth]{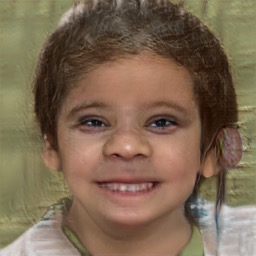} &
            \includegraphics[width=0.07\textwidth]{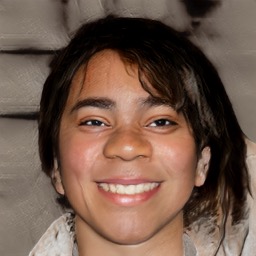} &
            \includegraphics[width=0.07\textwidth]{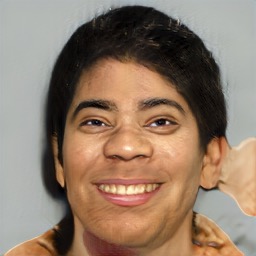} &
            \includegraphics[width=0.07\textwidth]{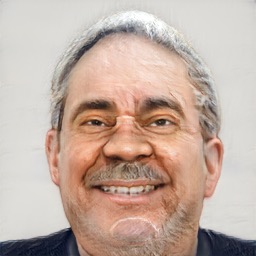} &
            \includegraphics[width=0.07\textwidth]{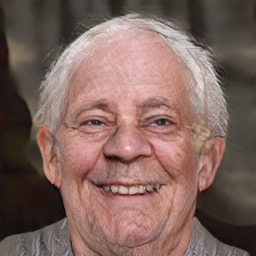} \\
            & & \raisebox{0.15in}{\rotatebox[origin=t]{90}{FUNIT}} & 
            \includegraphics[width=0.07\textwidth]{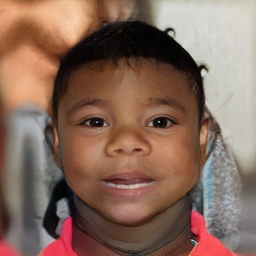} &
            \includegraphics[width=0.07\textwidth]{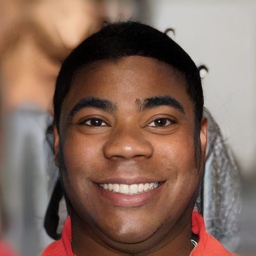} &
            \includegraphics[width=0.07\textwidth]{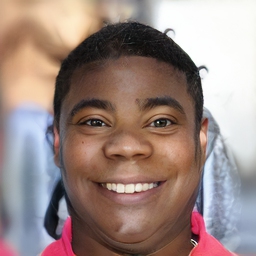} &
            \includegraphics[width=0.07\textwidth]{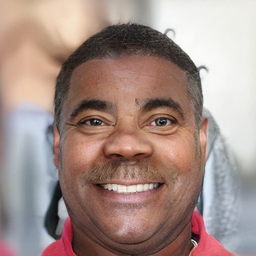} &
            \includegraphics[width=0.07\textwidth]{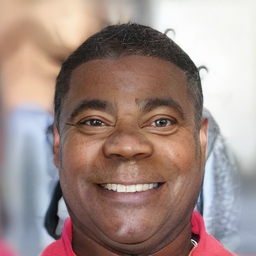} \\
            & & \raisebox{0.15in}{\rotatebox[origin=t]{90}{SAM}} &
            \includegraphics[width=0.07\textwidth]{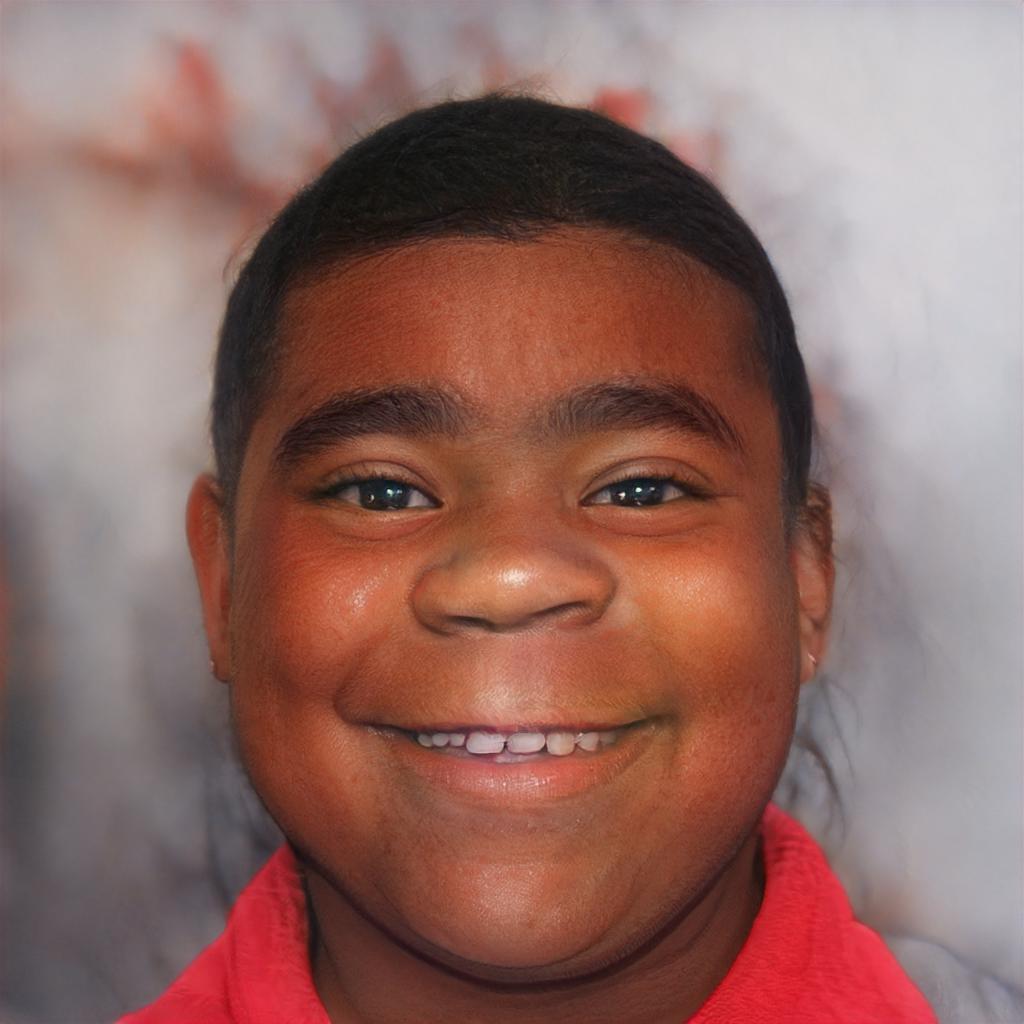} &
            \includegraphics[width=0.07\textwidth]{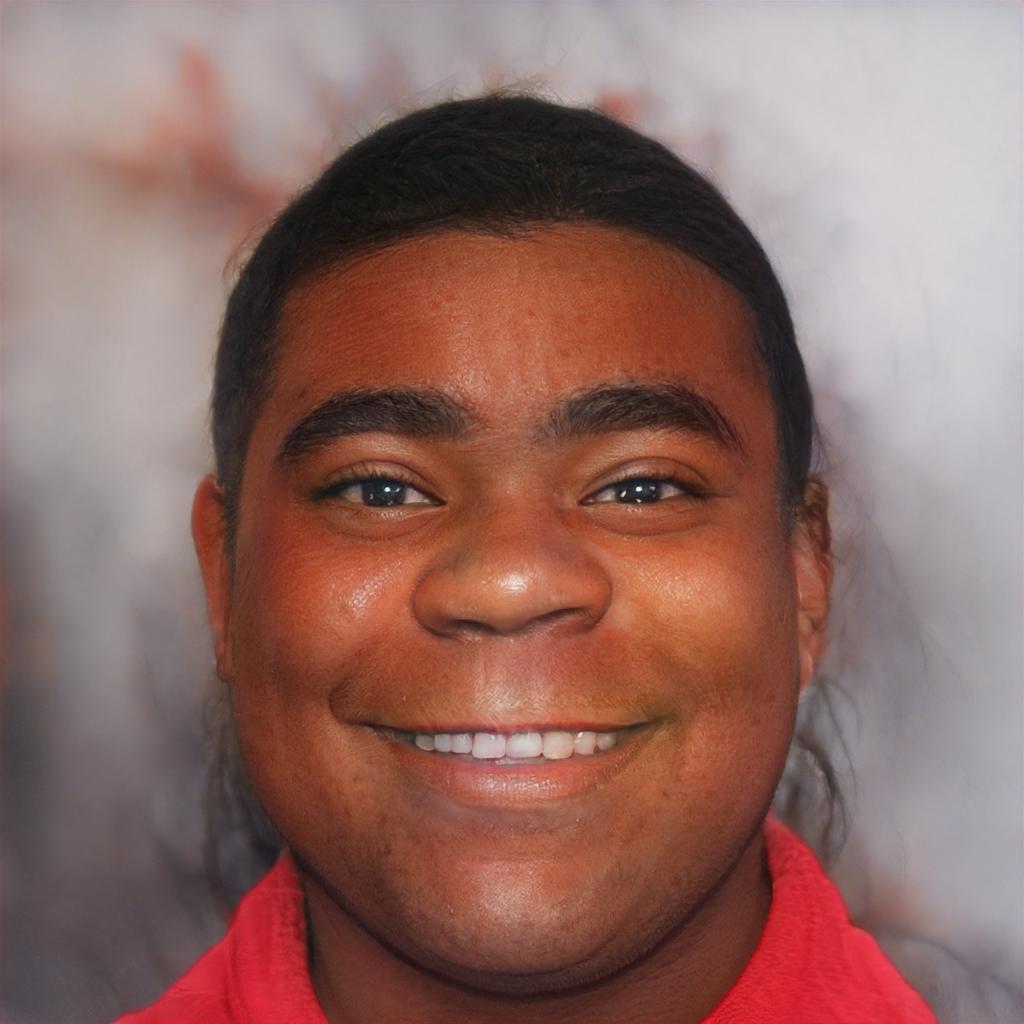} &
            \includegraphics[width=0.07\textwidth]{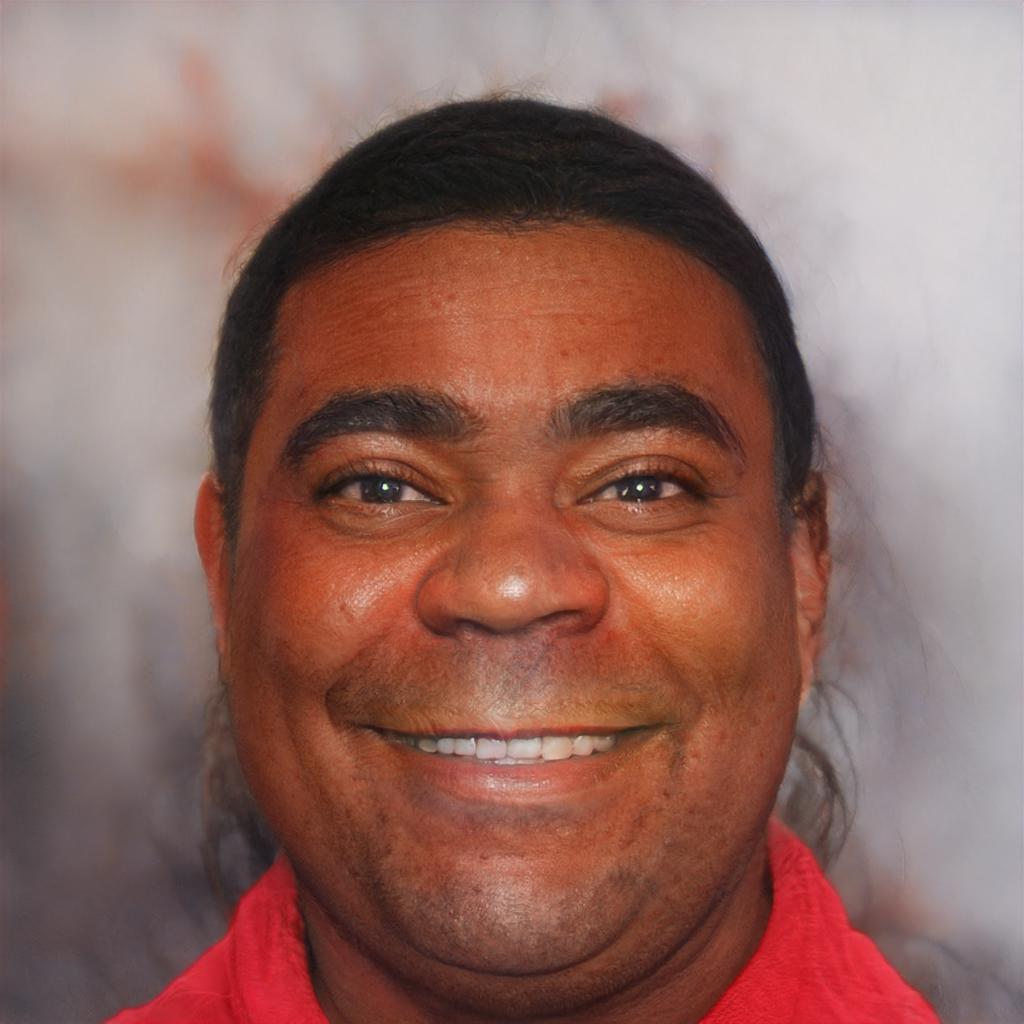} &
            \includegraphics[width=0.07\textwidth]{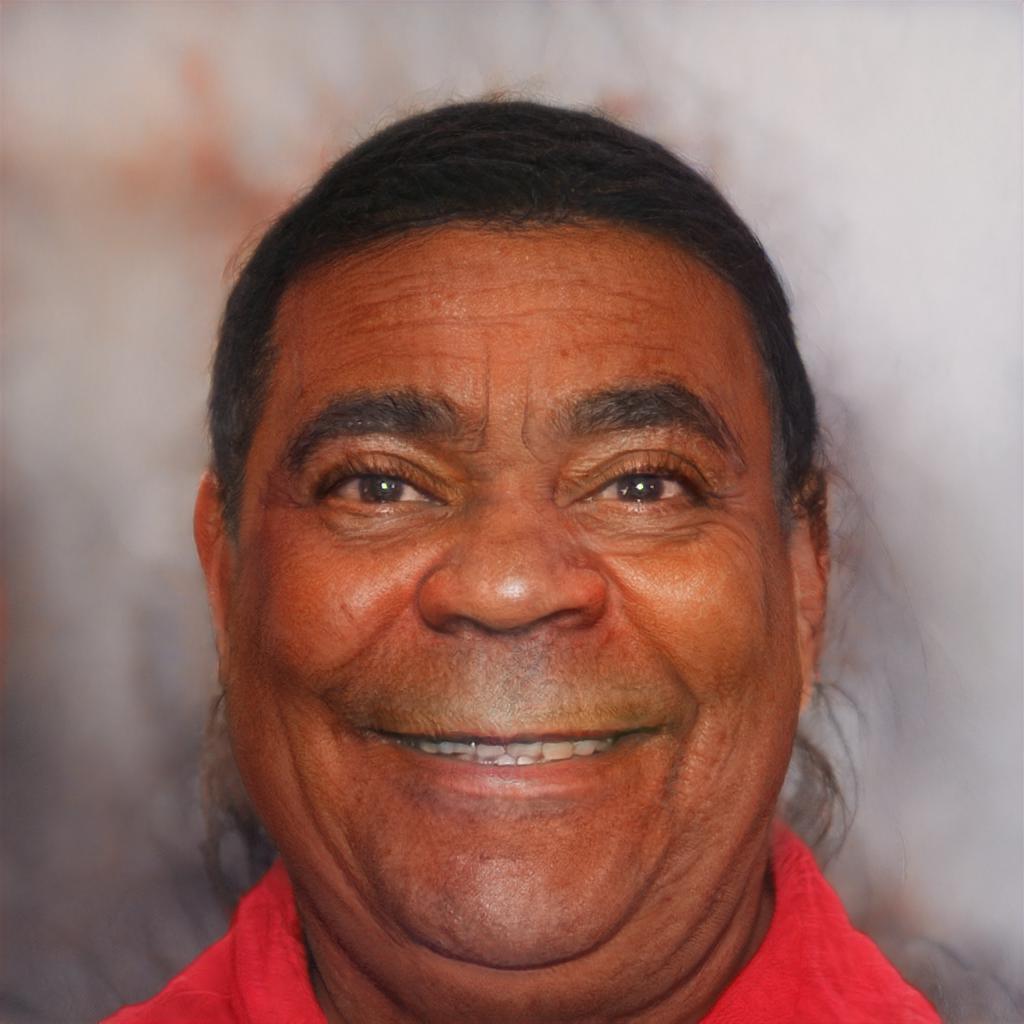} &
            \includegraphics[width=0.07\textwidth]{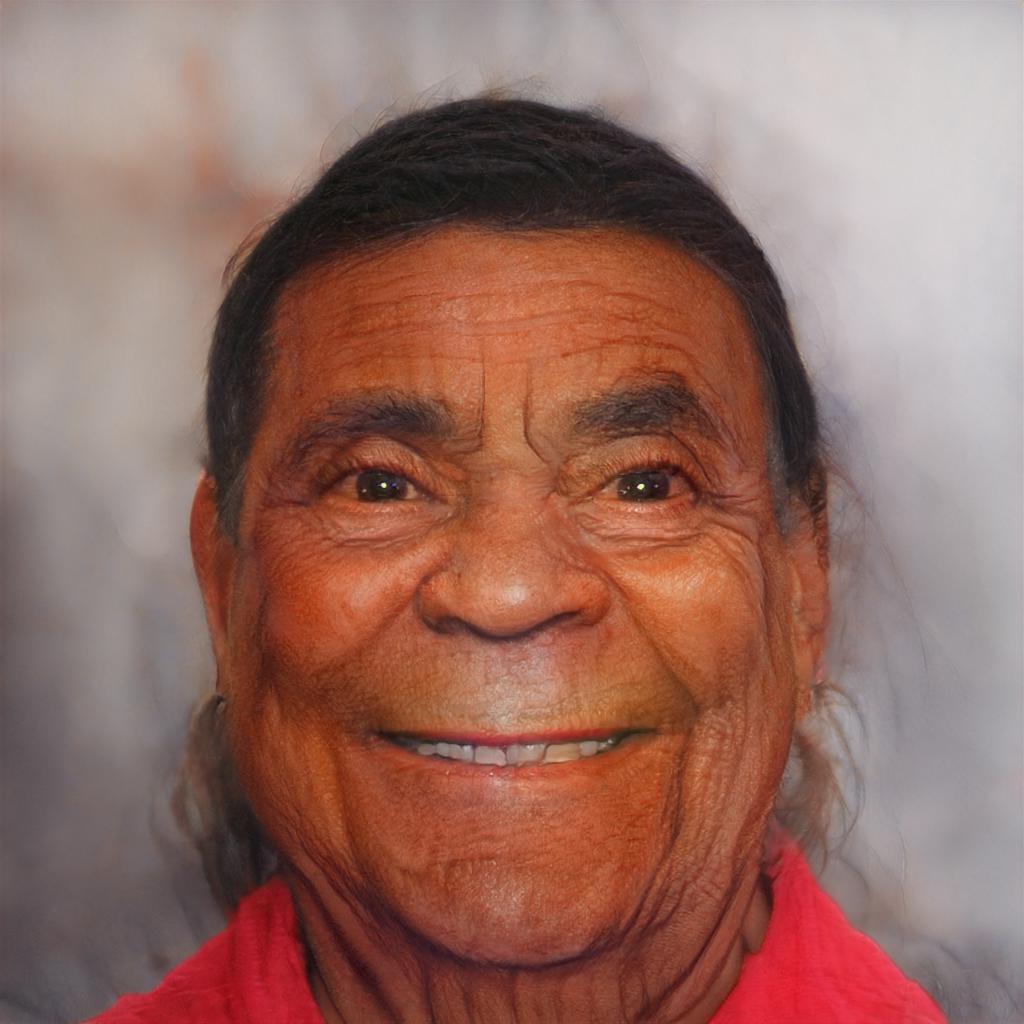}
            
            \tabularnewline
            \includegraphics[width=0.07\textwidth]{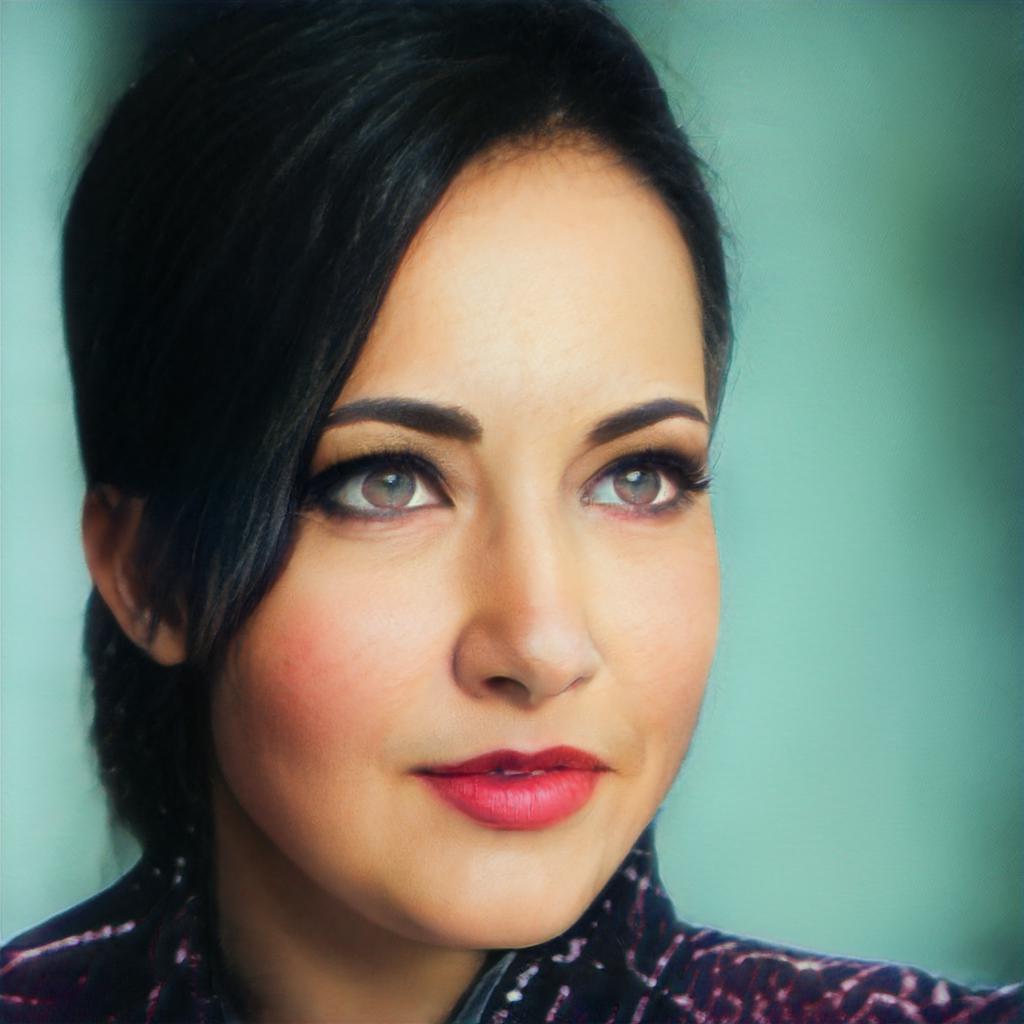} &
            & \raisebox{0.15in}{\rotatebox[origin=t]{90}{STAR}} & 
            \includegraphics[width=0.07\textwidth]{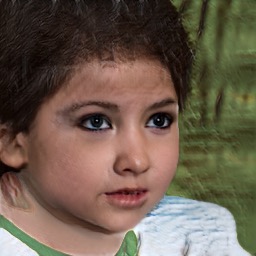} &
            \includegraphics[width=0.07\textwidth]{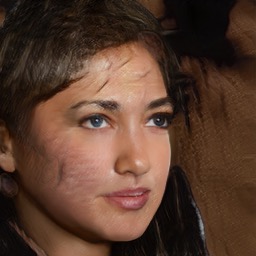} &
            \includegraphics[width=0.07\textwidth]{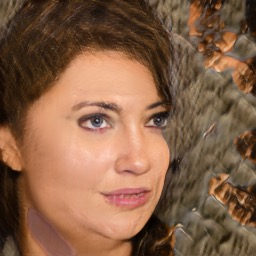} &
            \includegraphics[width=0.07\textwidth]{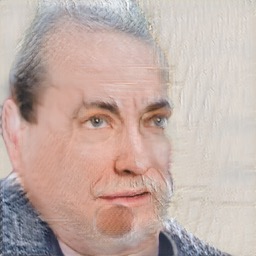} &
            \includegraphics[width=0.07\textwidth]{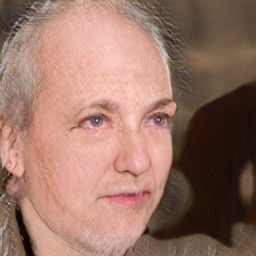} \\
            & & \raisebox{0.15in}{\rotatebox[origin=t]{90}{FUNIT}} & 
            \includegraphics[width=0.07\textwidth]{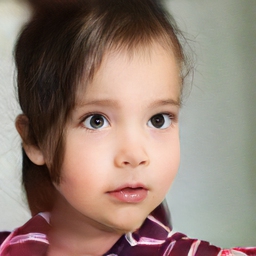} &
            \includegraphics[width=0.07\textwidth]{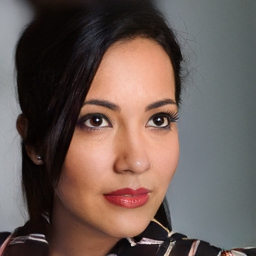} &
            \includegraphics[width=0.07\textwidth]{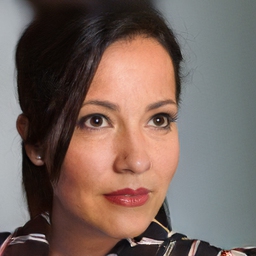} &
            \includegraphics[width=0.07\textwidth]{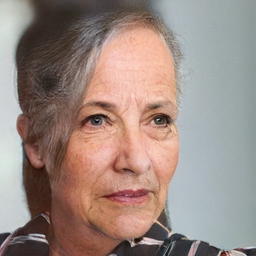} &
            \includegraphics[width=0.07\textwidth]{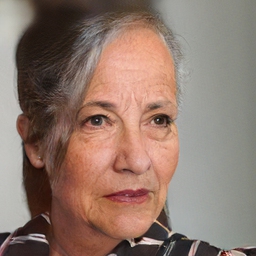} \\
            & & \raisebox{0.15in}{\rotatebox[origin=t]{90}{SAM}} &
            \includegraphics[width=0.07\textwidth]{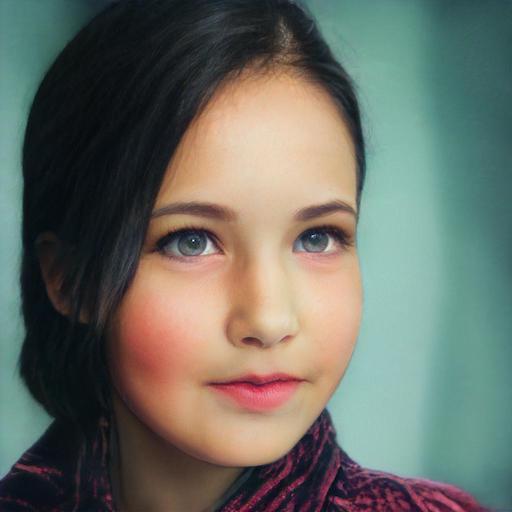} &
            \includegraphics[width=0.07\textwidth]{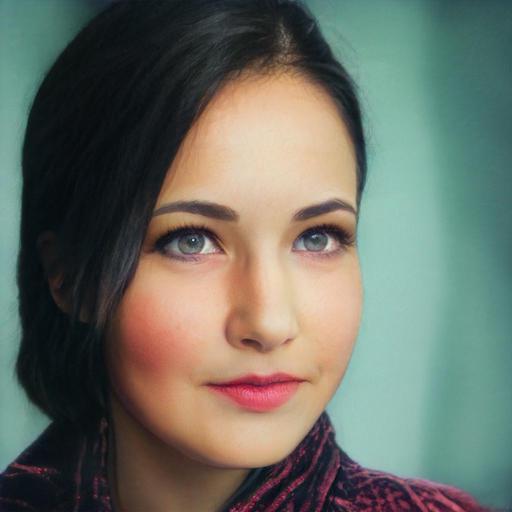} &
            \includegraphics[width=0.07\textwidth]{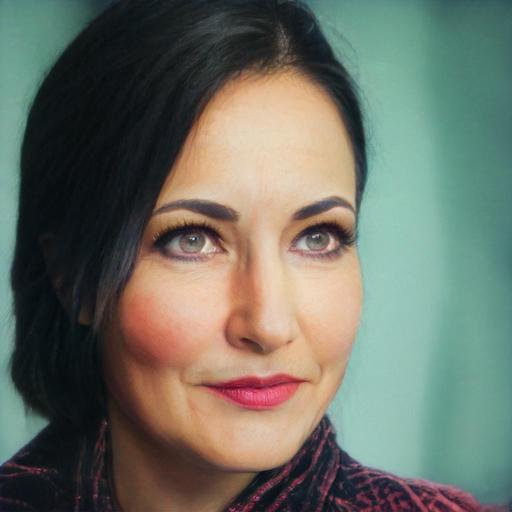} &
            \includegraphics[width=0.07\textwidth]{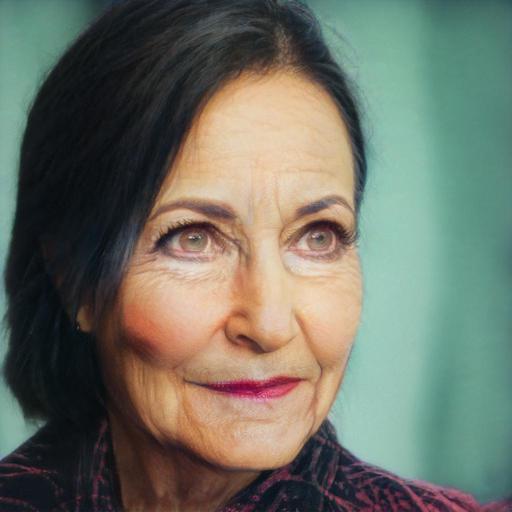} &
            \includegraphics[width=0.07\textwidth]{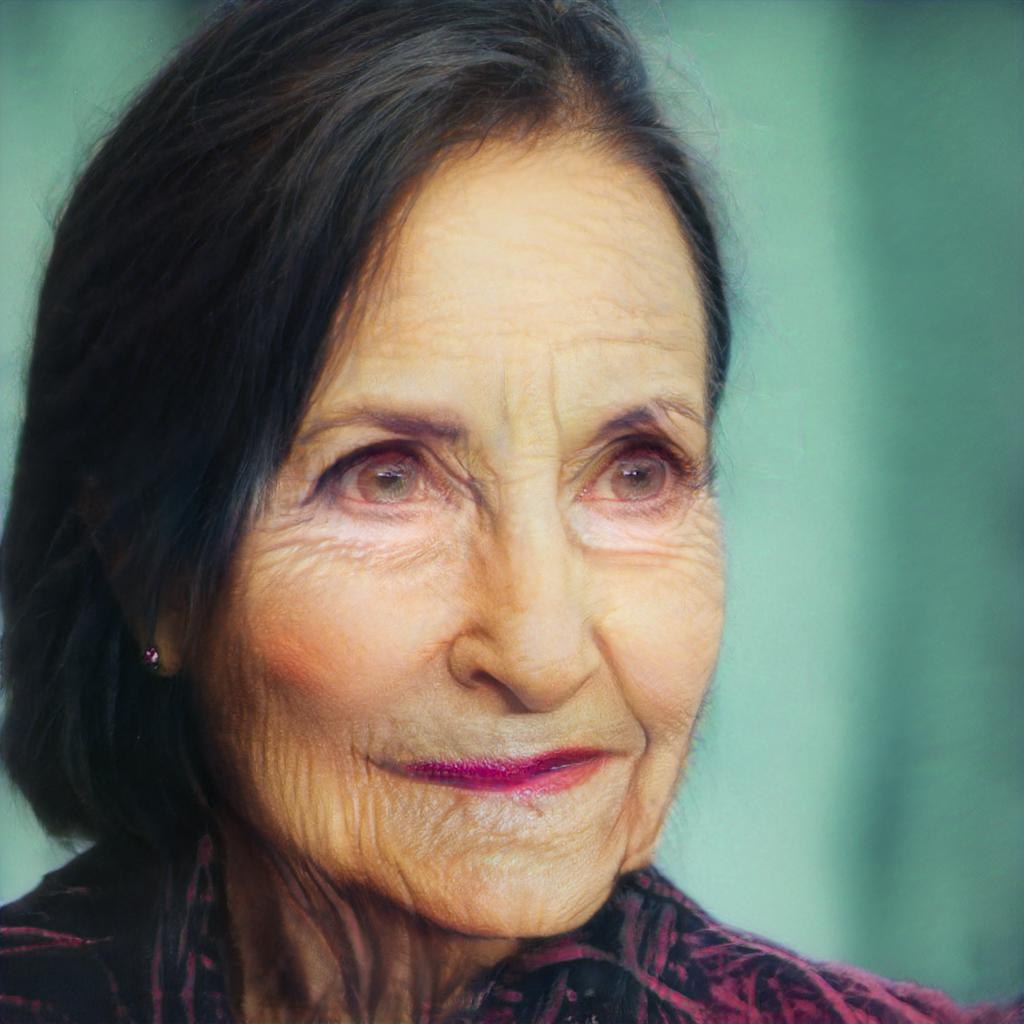}
    \end{tabular}
    \vspace{-0.2cm}
    \caption{Qualitative Comparison with FUNIT~\cite{liu2019few} and StarGANv2~\cite{choi2020stargan} (STAR). Note that FUNIT and StarGANv2 translate each source image using the same reference image. For translating images using SAM we set the target age equal to the the median age of each group.}
    \label{fig:comparison_multi_domain}
\end{figure}

%% file: figures/appendix/ablation_qualitative.tex
\begin{figure*}[t!]
    \centering
    \setlength{\belowcaptionskip}{-2.5pt}
    \begin{minipage}{0.5\textwidth}
        \setlength{\tabcolsep}{1pt}
        \begin{tabular}{c c c}
            Inversion & & \\
            \includegraphics[width=0.15\textwidth]{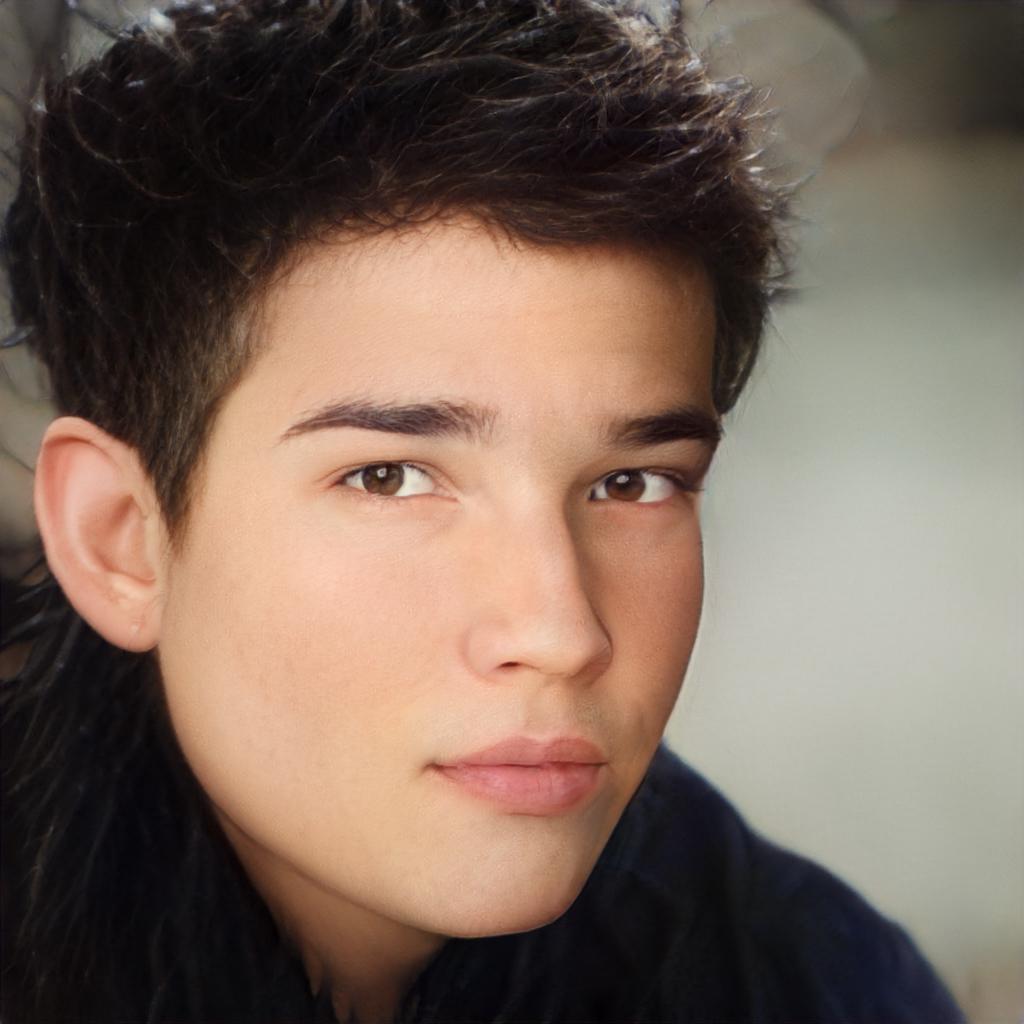} &
            \raisebox{0.185in}{\rotatebox[origin=t]{90}{SAM}} & 
            \includegraphics[width=0.75\textwidth]{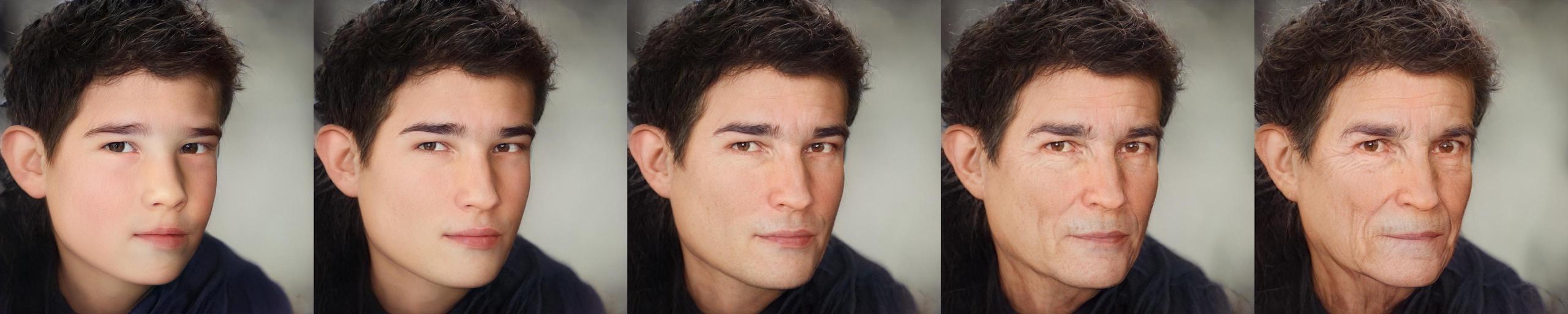} \\
            & \raisebox{0.185in}{\rotatebox[origin=t]{90}{$SAM_{direct}$}} & 
            \includegraphics[width=0.75\textwidth]{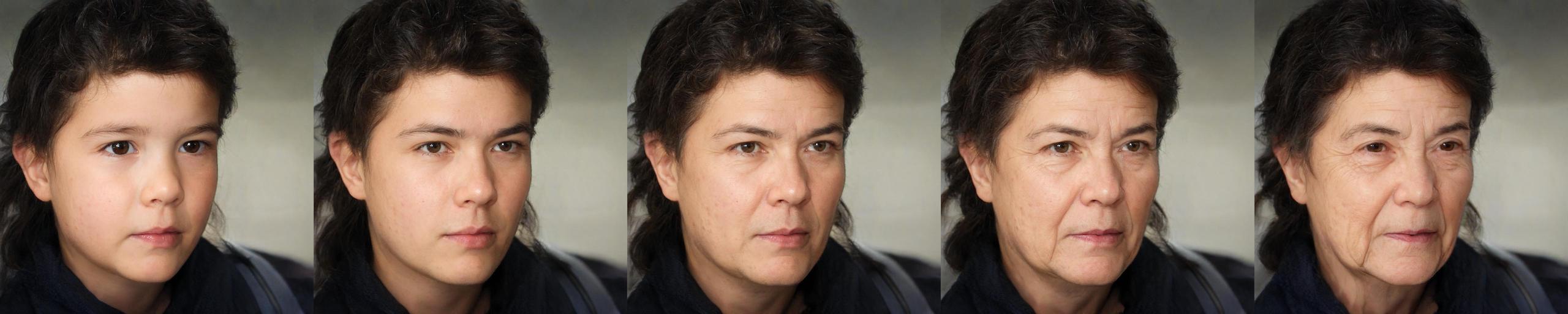}
            \tabularnewline
            \includegraphics[width=0.15\textwidth]{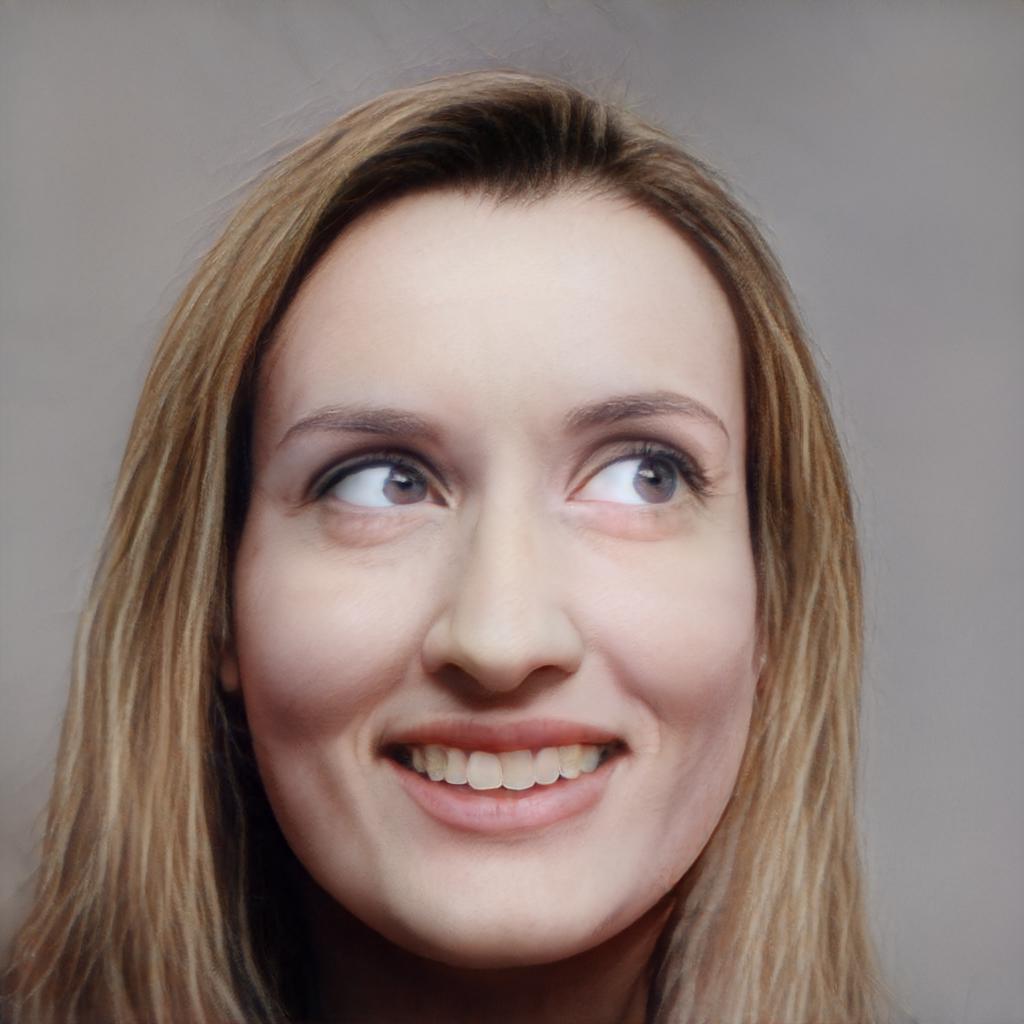} &
            \raisebox{0.185in}{\rotatebox[origin=t]{90}{SAM}} & 
            \includegraphics[width=0.75\textwidth]{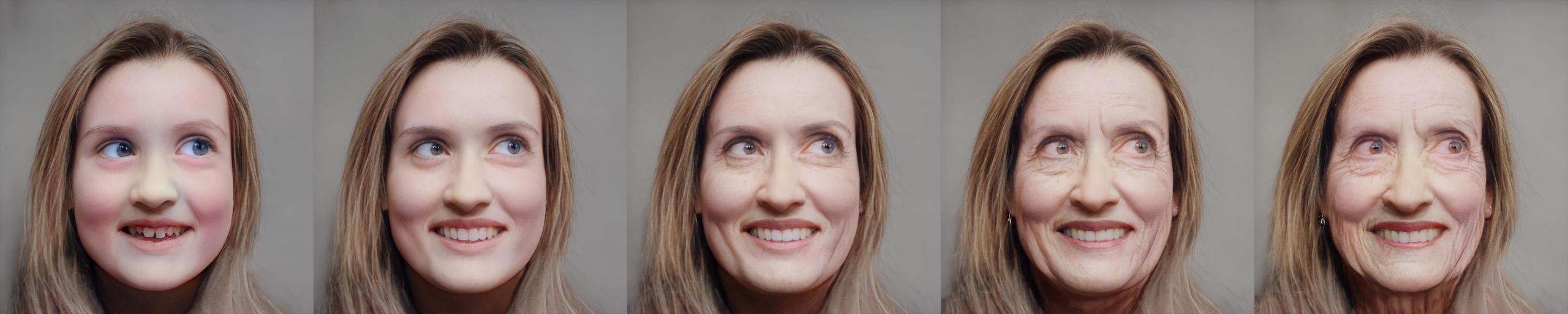} \\
            & \raisebox{0.185in}{\rotatebox[origin=t]{90}{$SAM_{direct}$}} & 
            \includegraphics[width=0.75\textwidth]{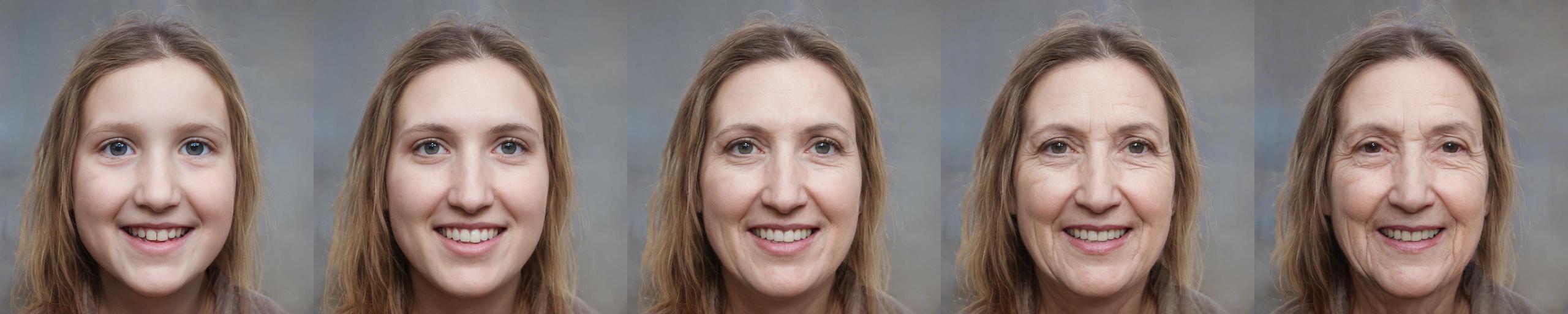}
        \end{tabular}
    \end{minipage}%
    \begin{minipage}{0.5\textwidth}
        \setlength{\tabcolsep}{1pt}
        \begin{tabular}{c c c}
            Inversion & & \\
            \includegraphics[width=0.15\textwidth]{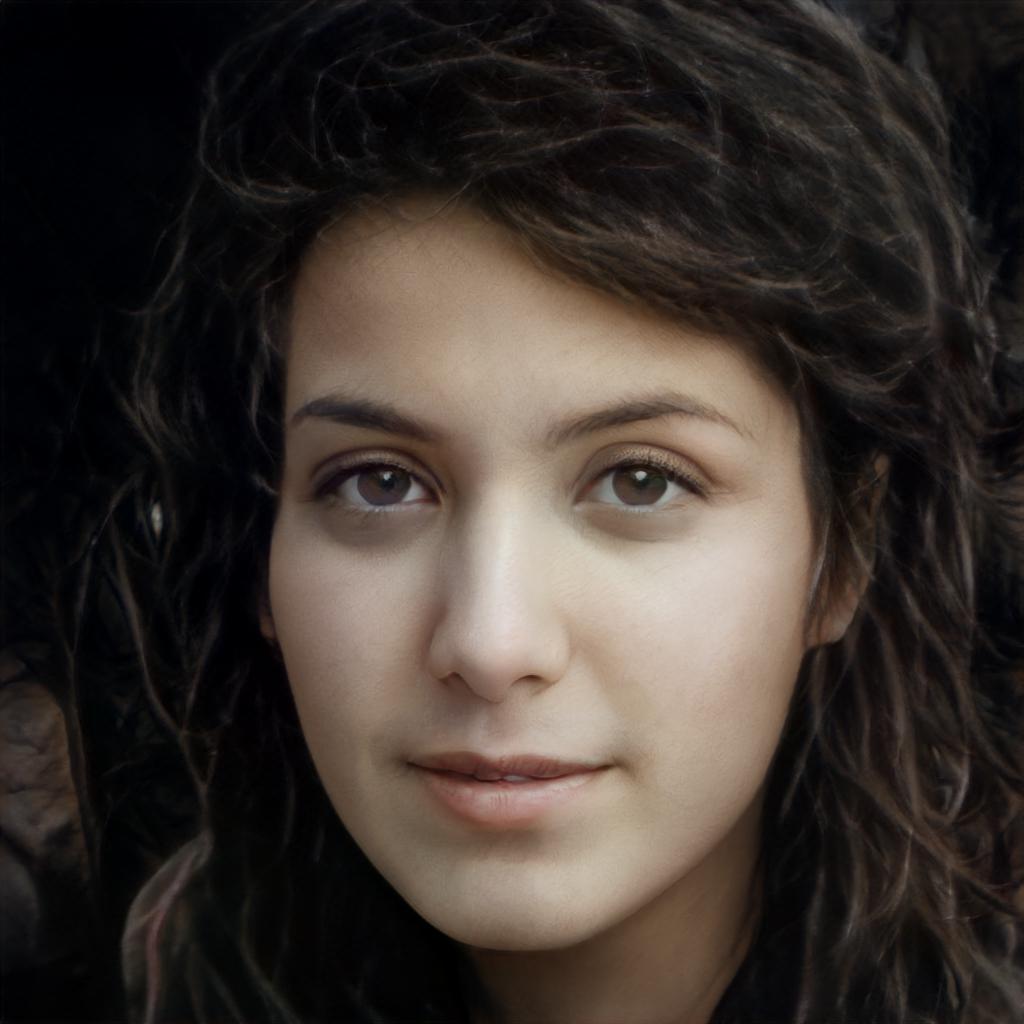} &
            \raisebox{0.185in}{\rotatebox[origin=t]{90}{SAM}} & 
            \includegraphics[width=0.75\textwidth]{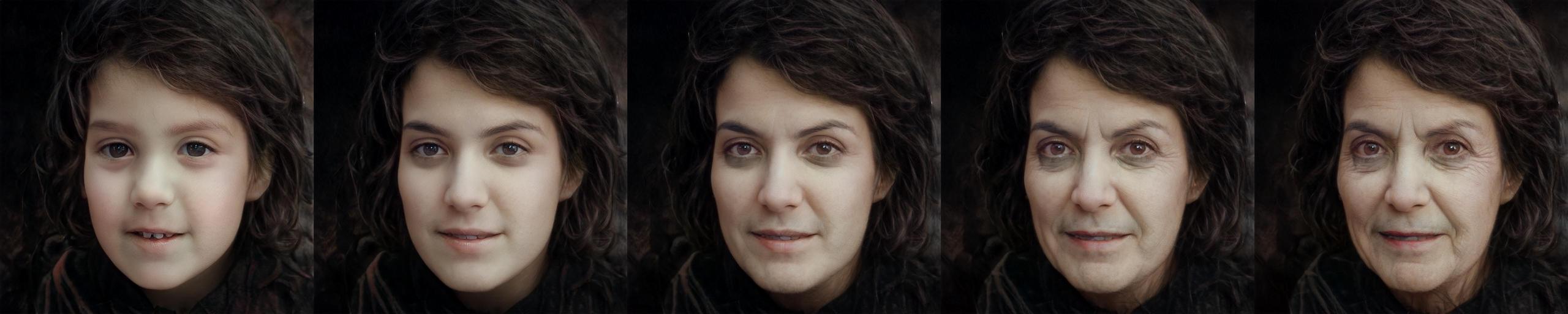} \\
            & \raisebox{0.185in}{\rotatebox[origin=t]{90}{$SAM_{direct}$}} & 
            \includegraphics[width=0.75\textwidth]{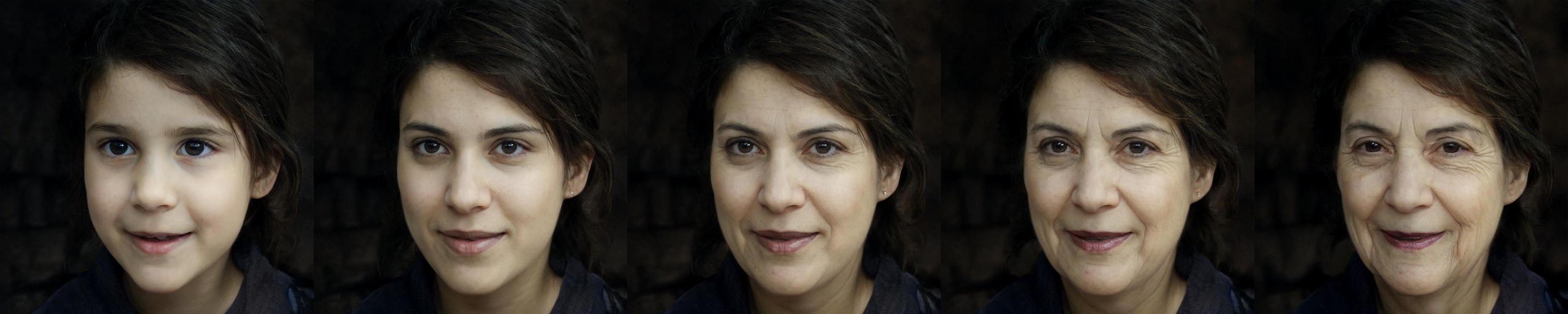}
            \tabularnewline
            \includegraphics[width=0.15\textwidth]{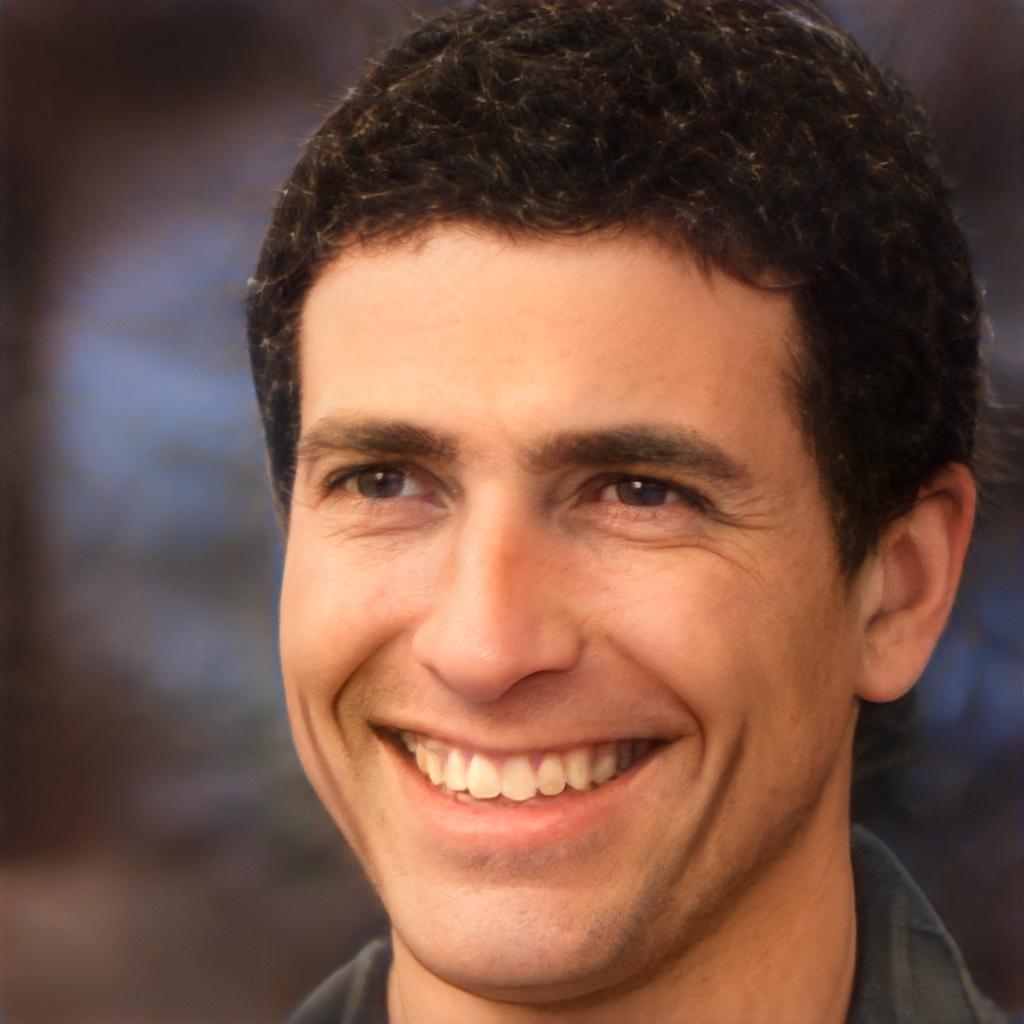} &
            \raisebox{0.185in}{\rotatebox[origin=t]{90}{SAM}} & 
            \includegraphics[width=0.75\textwidth]{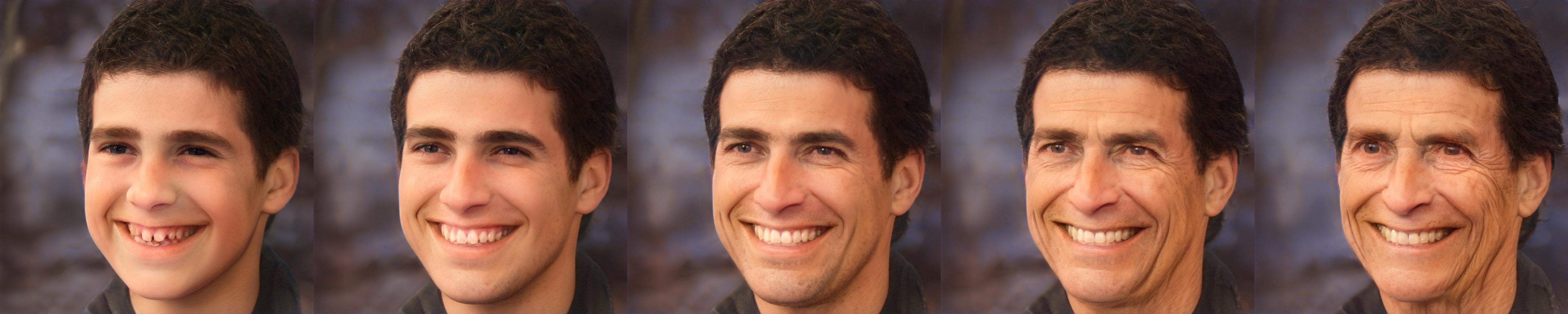} \\
            & \raisebox{0.185in}{\rotatebox[origin=t]{90}{$SAM_{direct}$}} & 
            \includegraphics[width=0.75\textwidth]{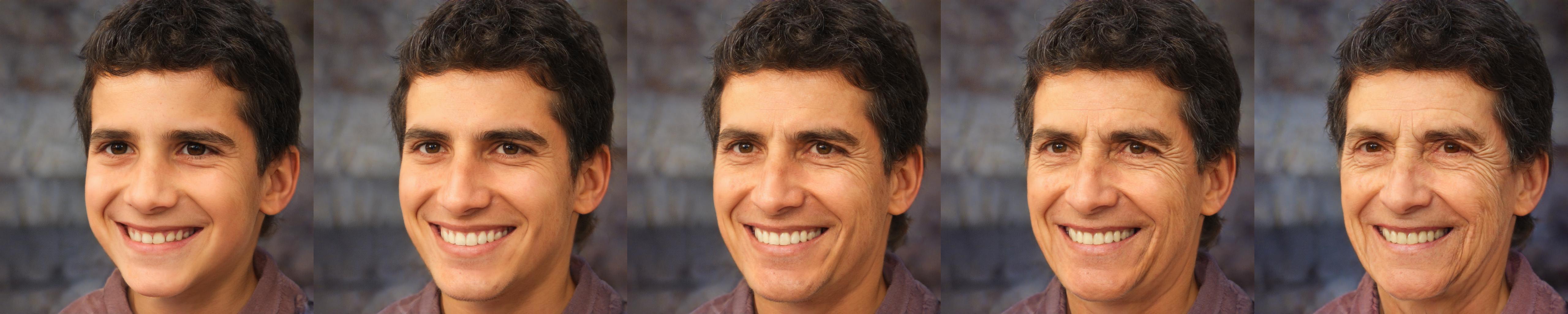}
        \end{tabular}
    \end{minipage}
    \vspace{-0.25cm}
    \caption{Qualitative comparison between our SAM method and the $SAM_{direct}$ variant. As can be seen, our SAM method is able to more accurately preserve identity and key facial features while faithfully modeling lifelong age transformation.}
    \label{ablation_qualitative}
\end{figure*}

%% file: figures/appendix/ablation_aging_accuracy.tex
\begin{table}
    \centering
    \caption{Quantitative evaluation of the two SAM variants. Our residual-based SAM compares favorably with $SAM_{direct}$. Note, lower is better.}
    \vspace{-0.4cm}
    \begin{tabular}{l c c c c c}
        \toprule 
        Target Age & 5 & 20 & 35 & 50 & 65 \\ 
        \cmidrule(r){1-1}
        \cmidrule(r){2-2}
        \cmidrule(r){3-3}
        \cmidrule(r){4-4}
        \cmidrule(r){5-5}
        \cmidrule(r){6-6}
        $SAM$  & \textbf{5.79} & 0.74 & 0.39 & \textbf{0.36} & 7.26 \\
        $SAM_{direct}$  & 6.23 & \textbf{0.69} & \textbf{0.37} & 0.41 &  \textbf{6.83} \\
        \bottomrule
    \end{tabular}
    \vspace{0.05cm}
    \caption*{\textbf{Ablation Study: Average Age Difference}}
    \vspace{-0.65cm}
    \label{tb:ablation_aging_acc}
\end{table}

%% file: figures/appendix/ablation_losses.tex
\begin{figure}
    \centering
    \setlength{\belowcaptionskip}{-7.5pt}
    \setlength{\tabcolsep}{1pt}
    \begin{tabular}{c c c c c c c c}
            
            \includegraphics[width=0.085\textwidth]{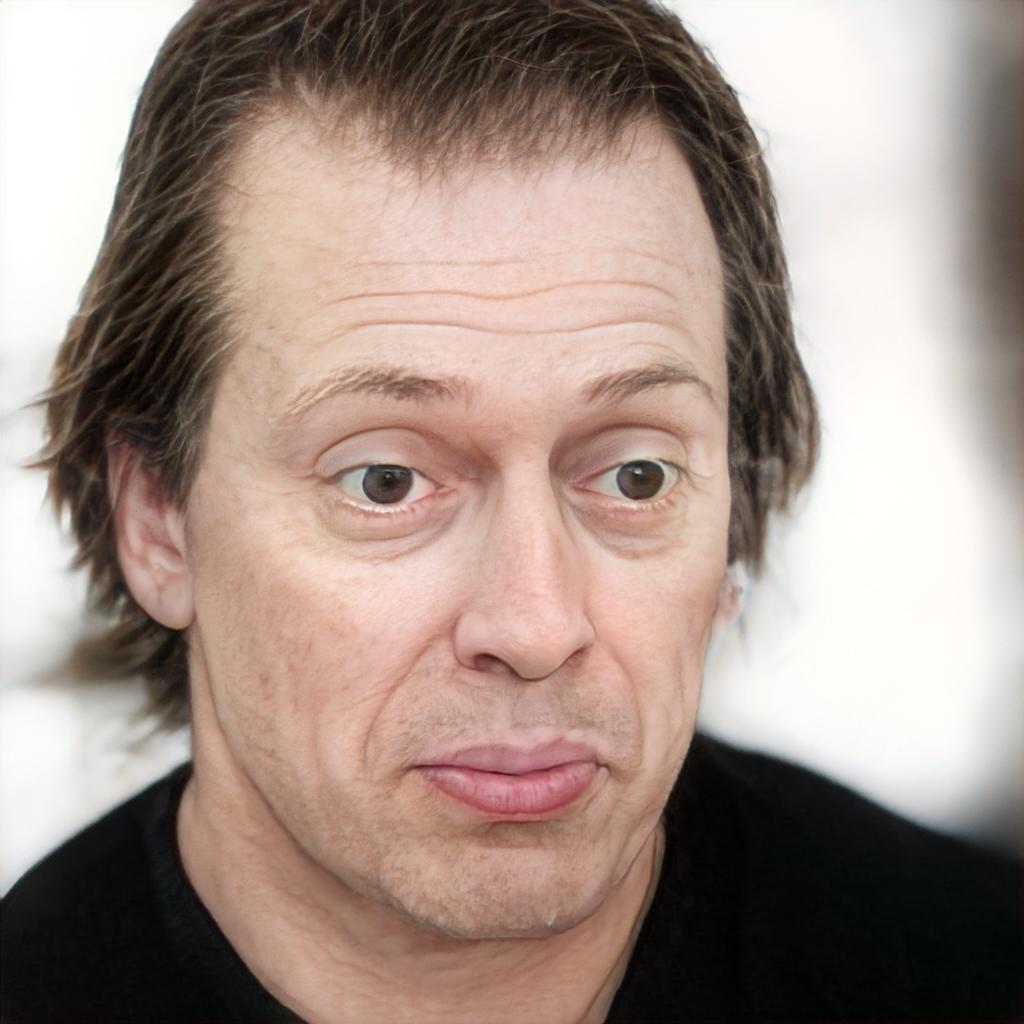} &
            \raisebox{0.2in}{\rotatebox[origin=t]{90}{5}} & 
            \includegraphics[width=0.085\textwidth]{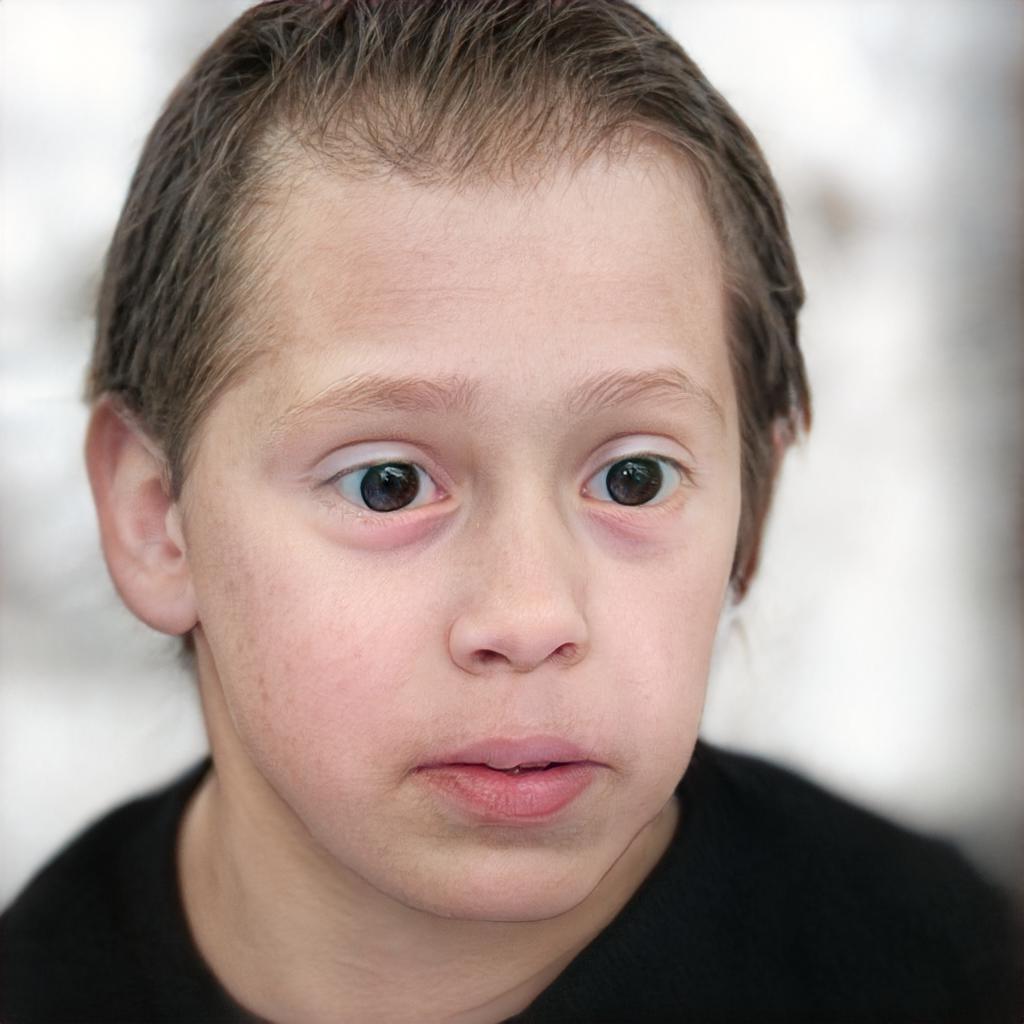} &
            \includegraphics[width=0.085\textwidth]{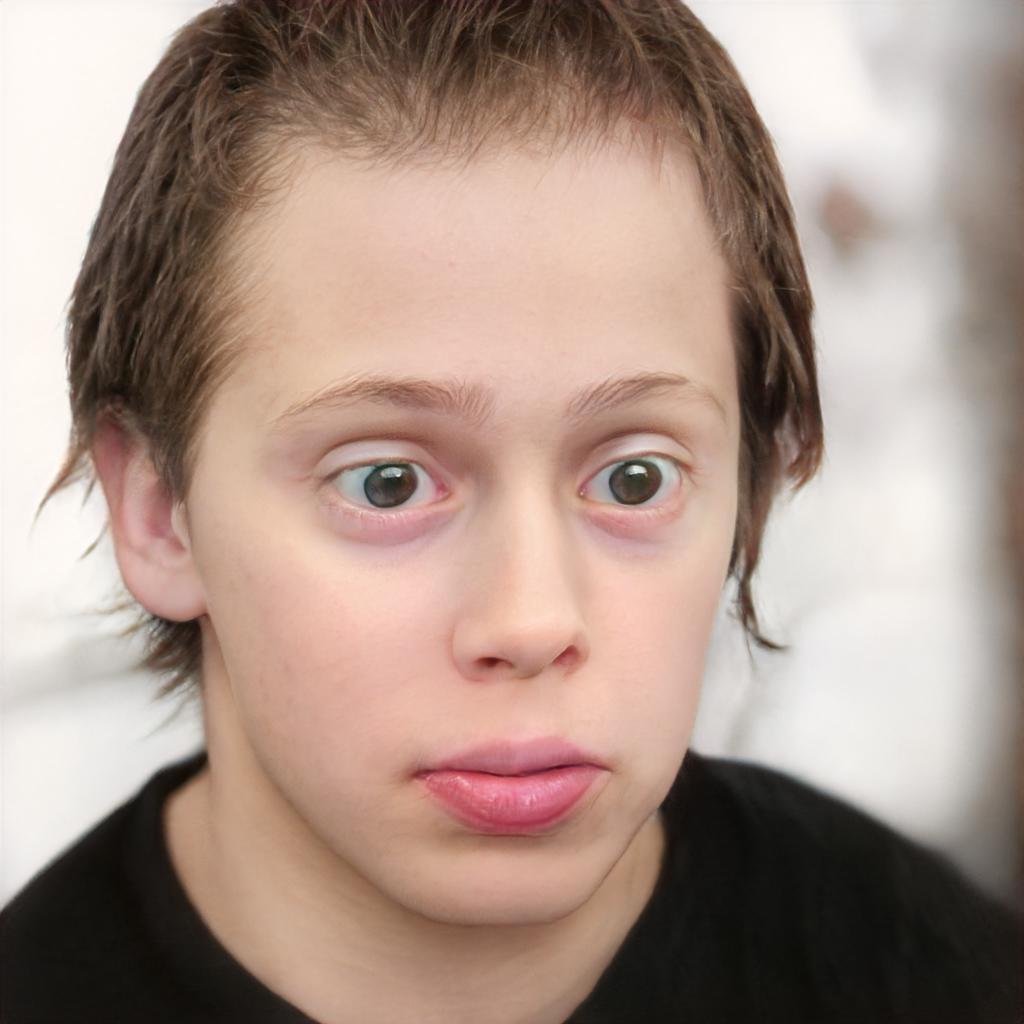} &
            \includegraphics[width=0.085\textwidth]{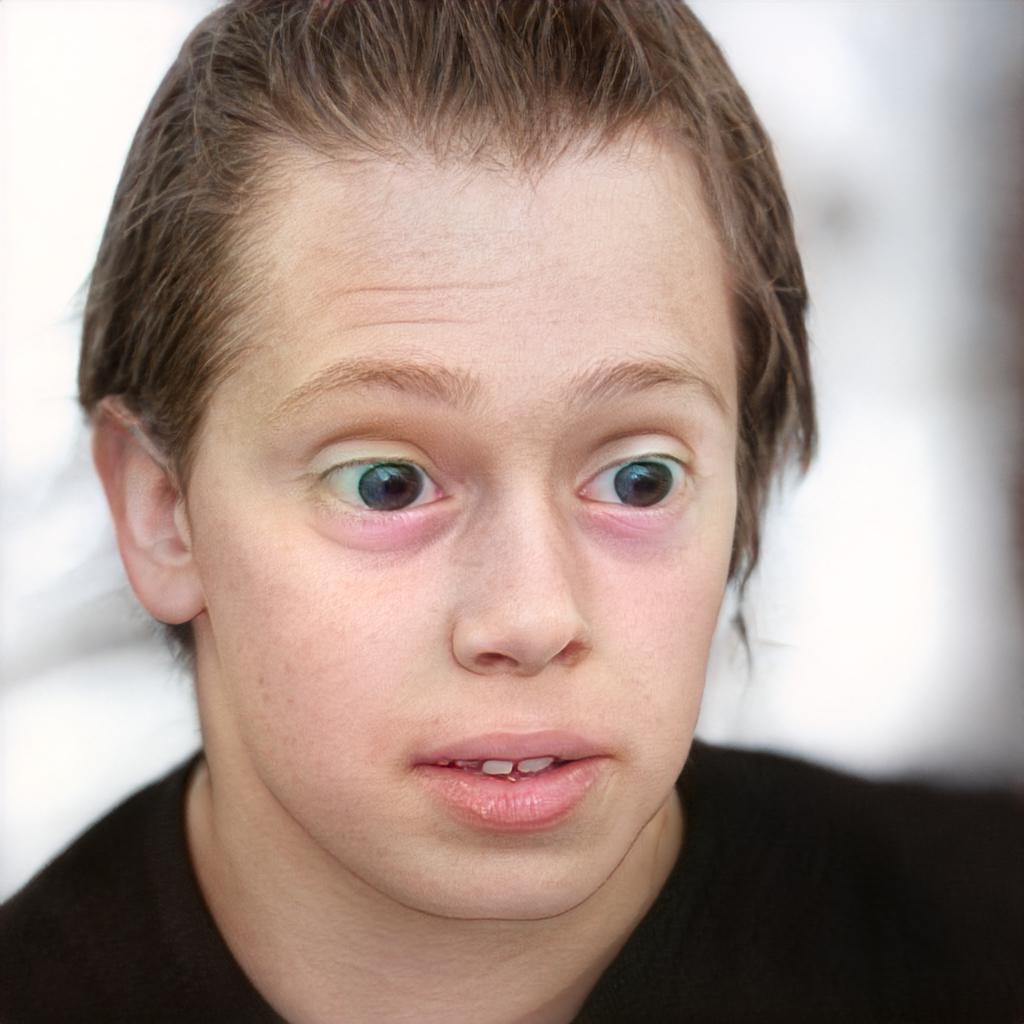} &
            \includegraphics[width=0.085\textwidth]{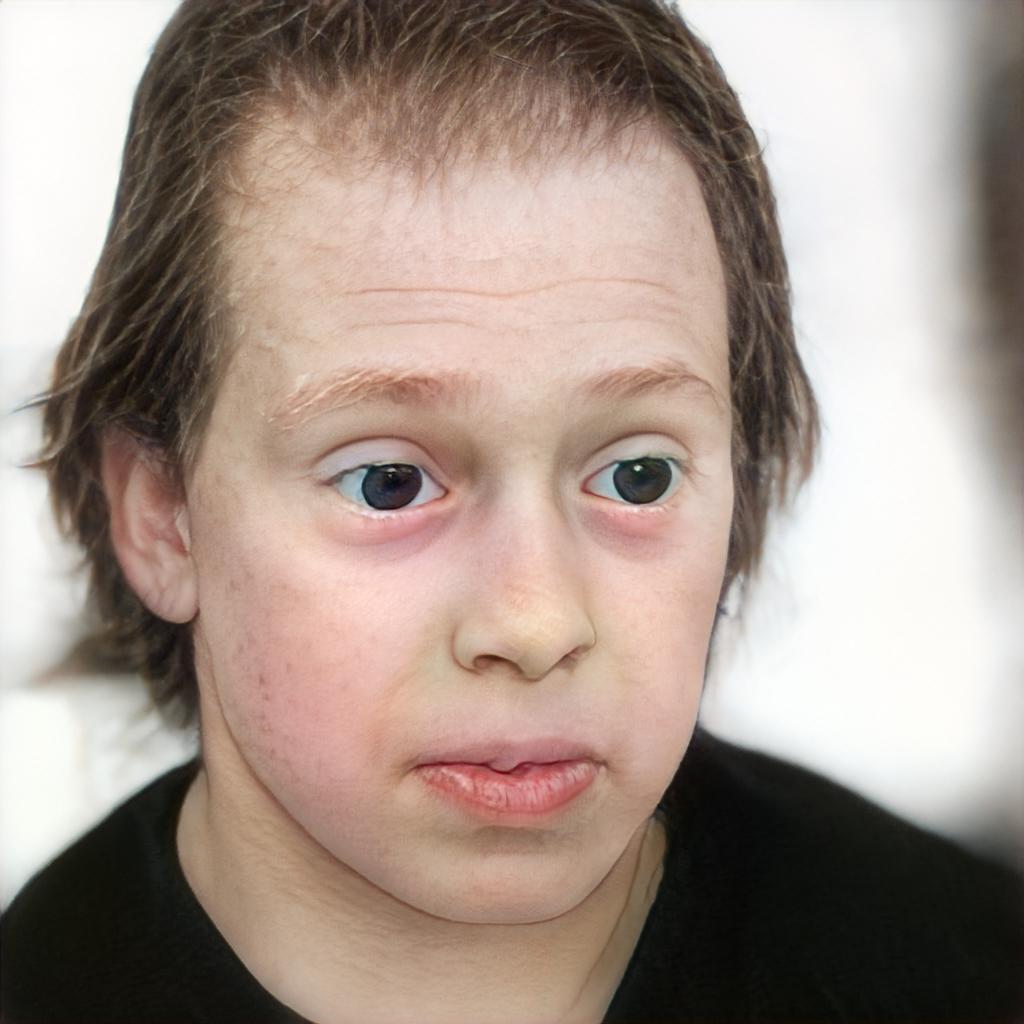} \\
            & \raisebox{0.2in}{\rotatebox[origin=t]{90}{45}} & 
            \includegraphics[width=0.085\textwidth]{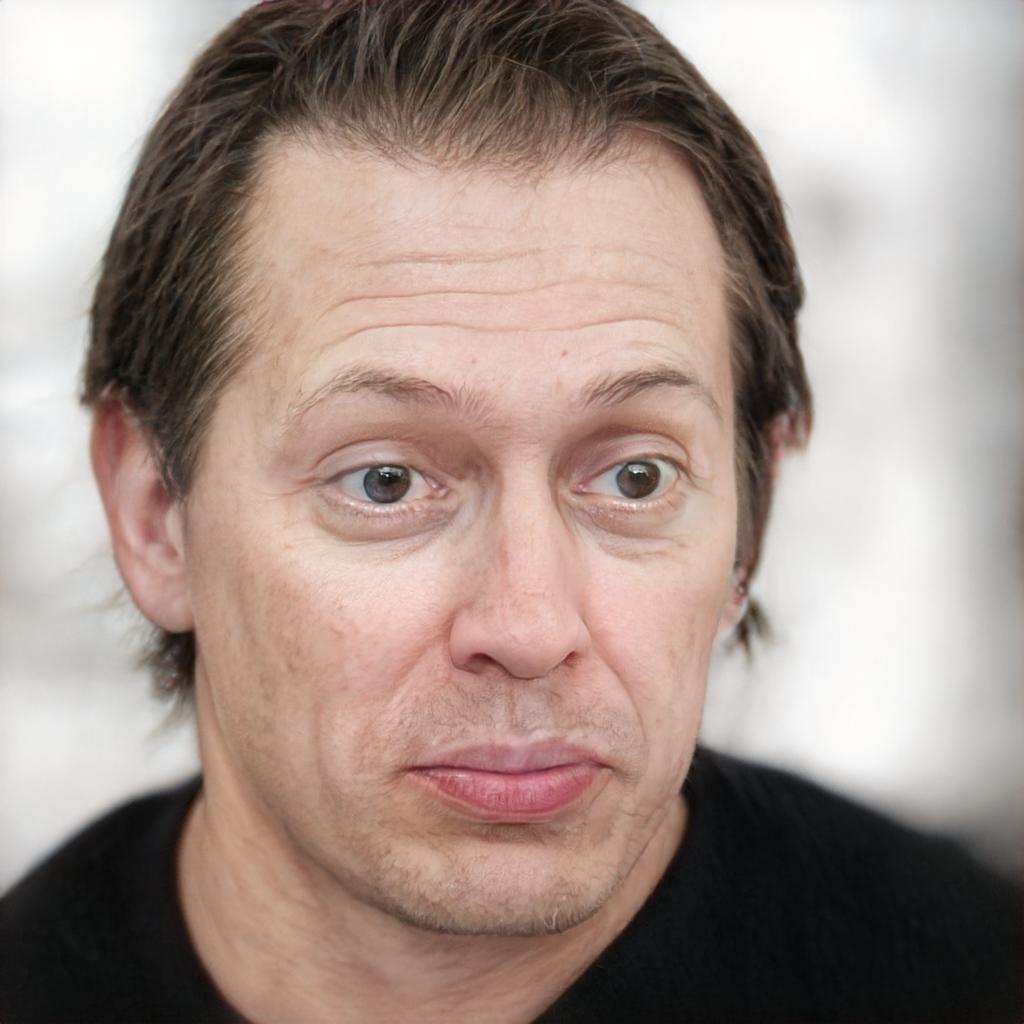} &
            \includegraphics[width=0.085\textwidth]{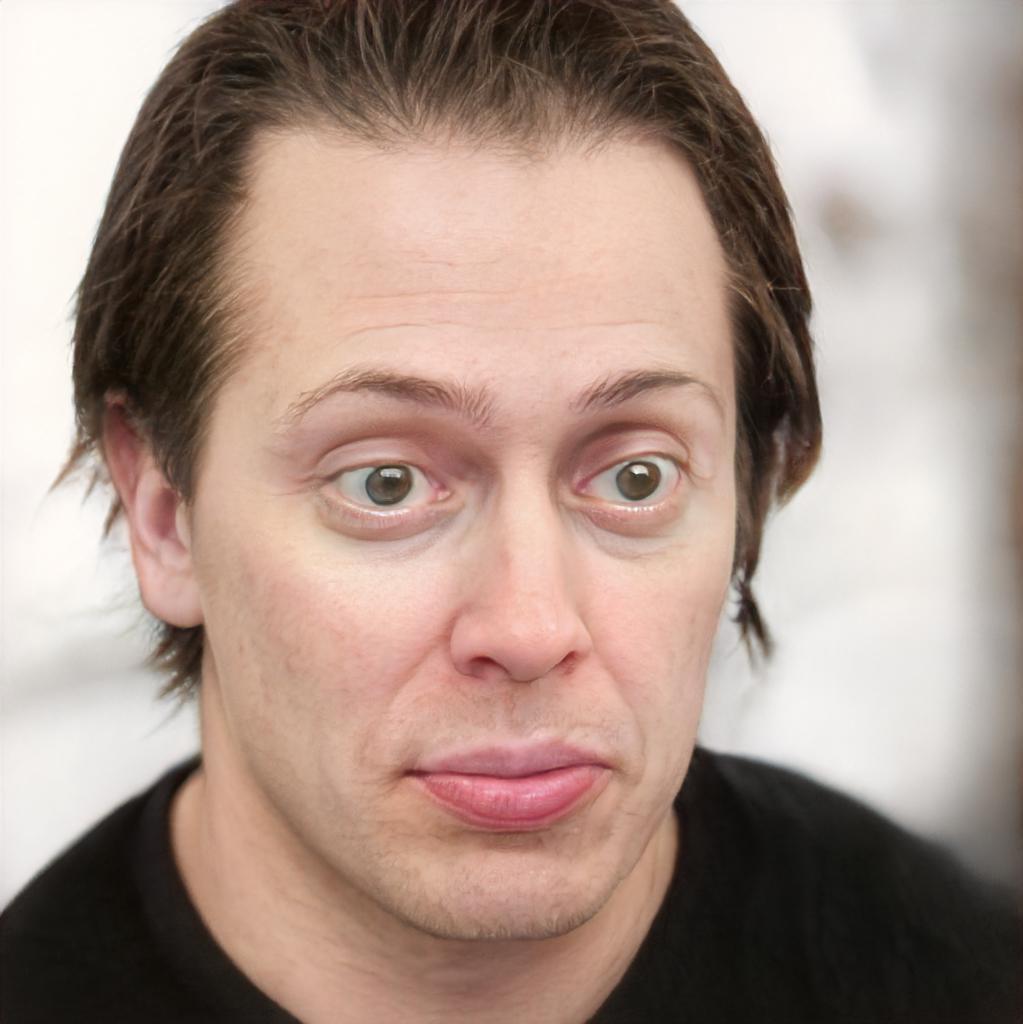} &
            \includegraphics[width=0.085\textwidth]{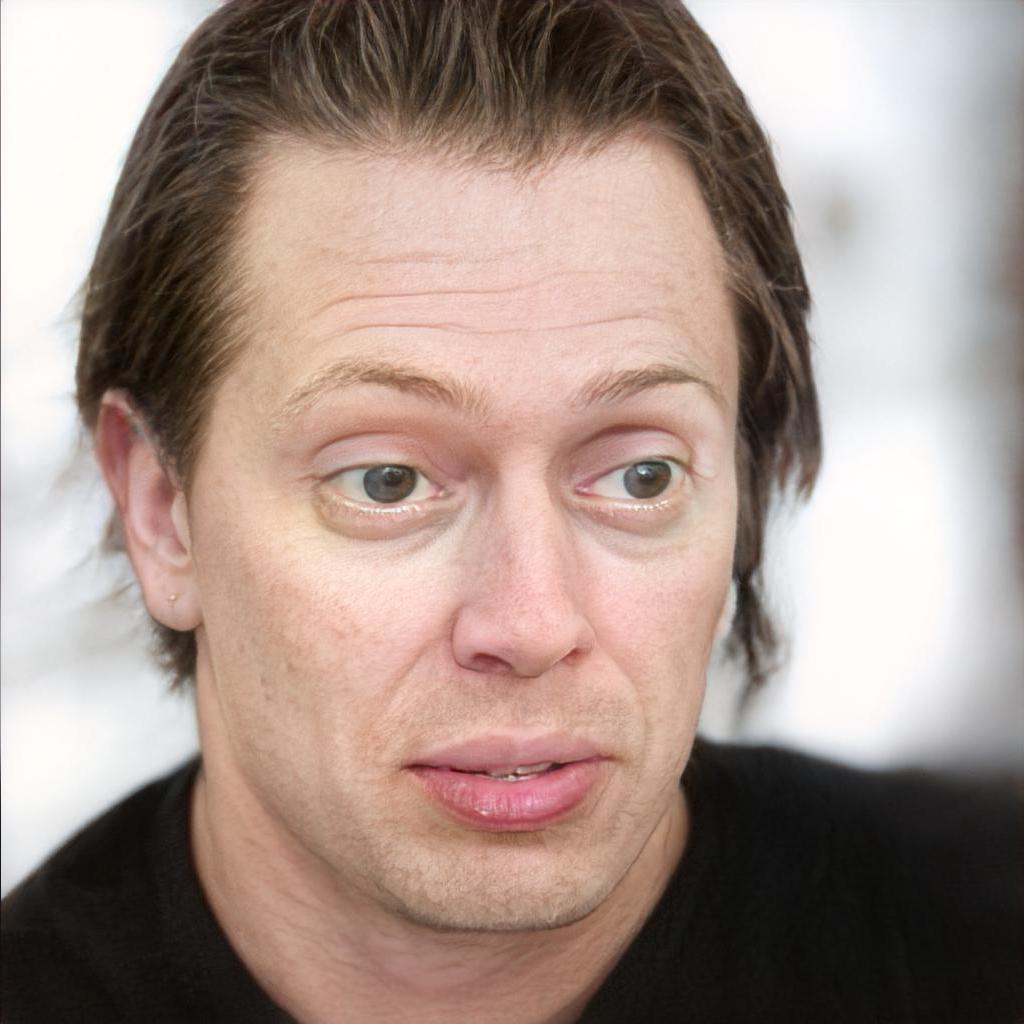} &
            \includegraphics[width=0.085\textwidth]{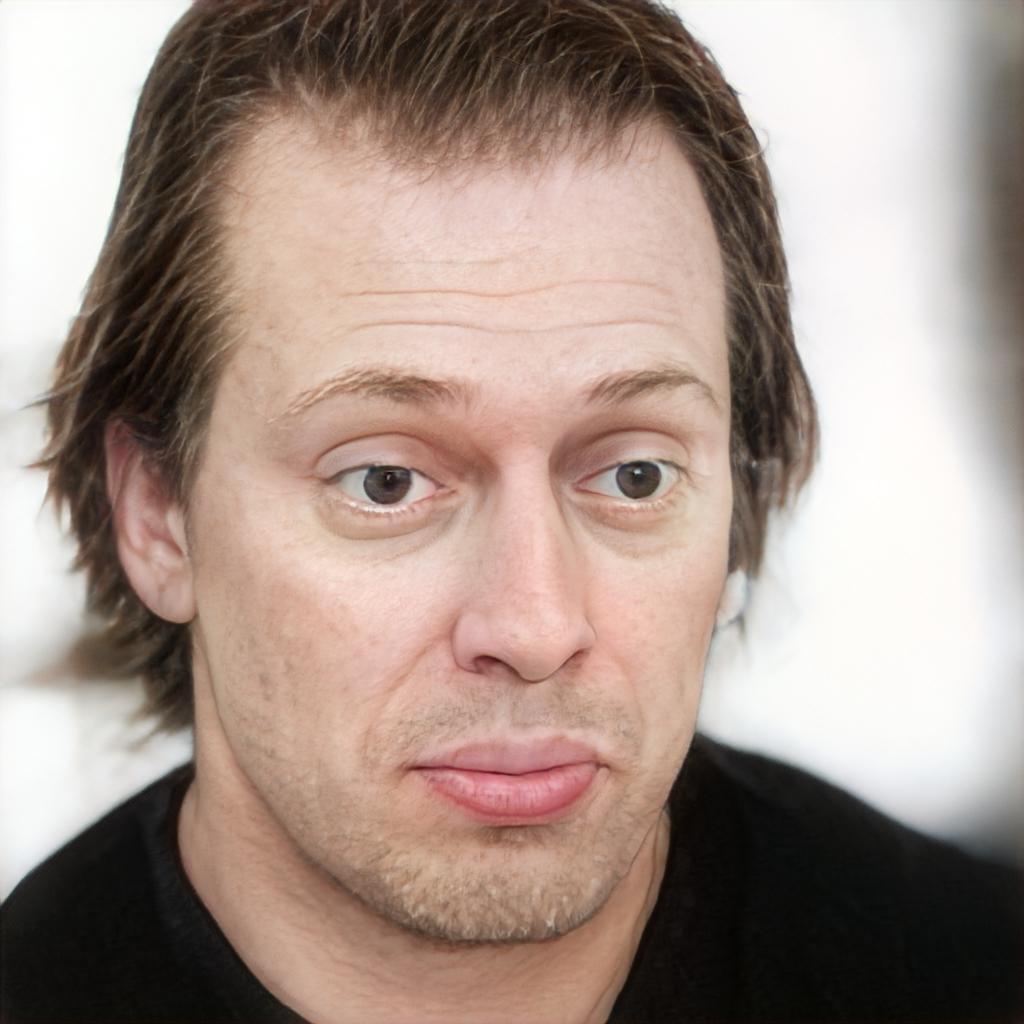} \\
            & \raisebox{0.2in}{\rotatebox[origin=t]{90}{85}} &
            \includegraphics[width=0.085\textwidth]{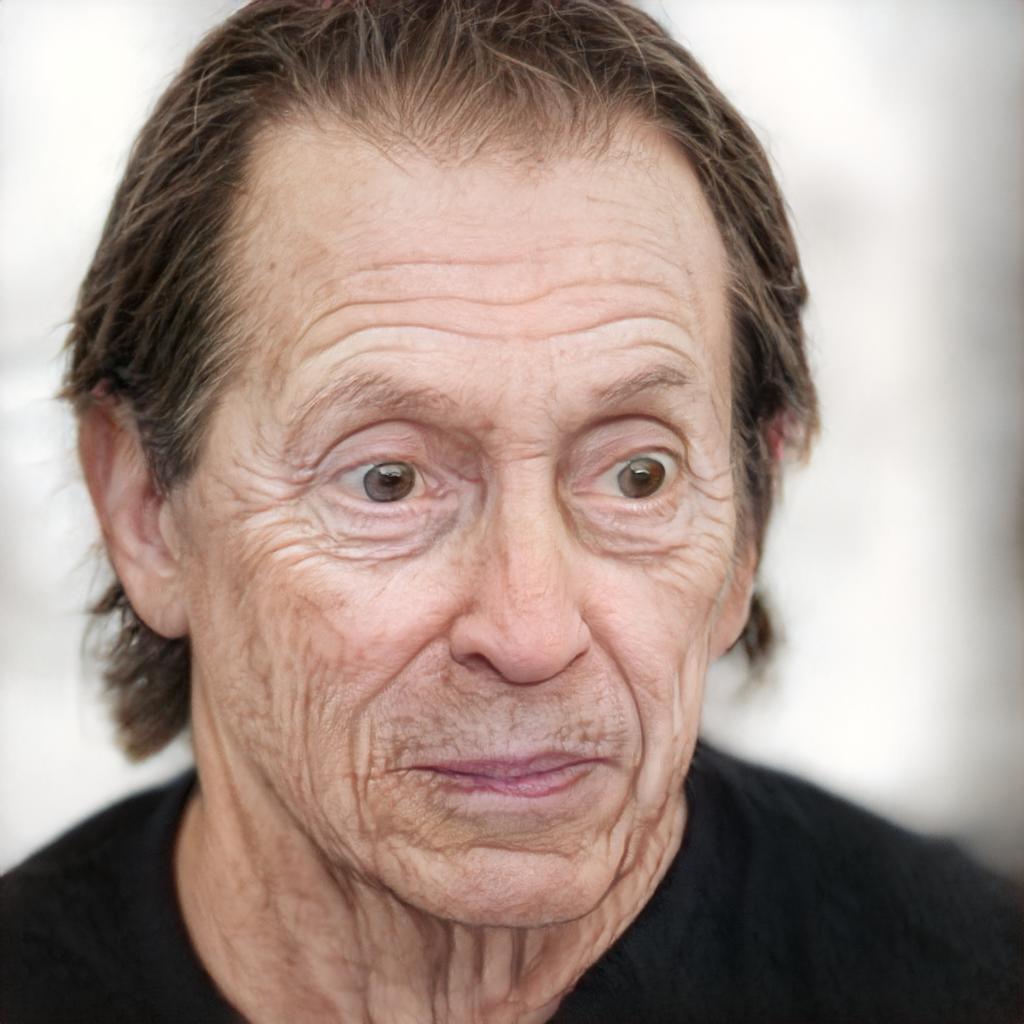} &
            \includegraphics[width=0.085\textwidth]{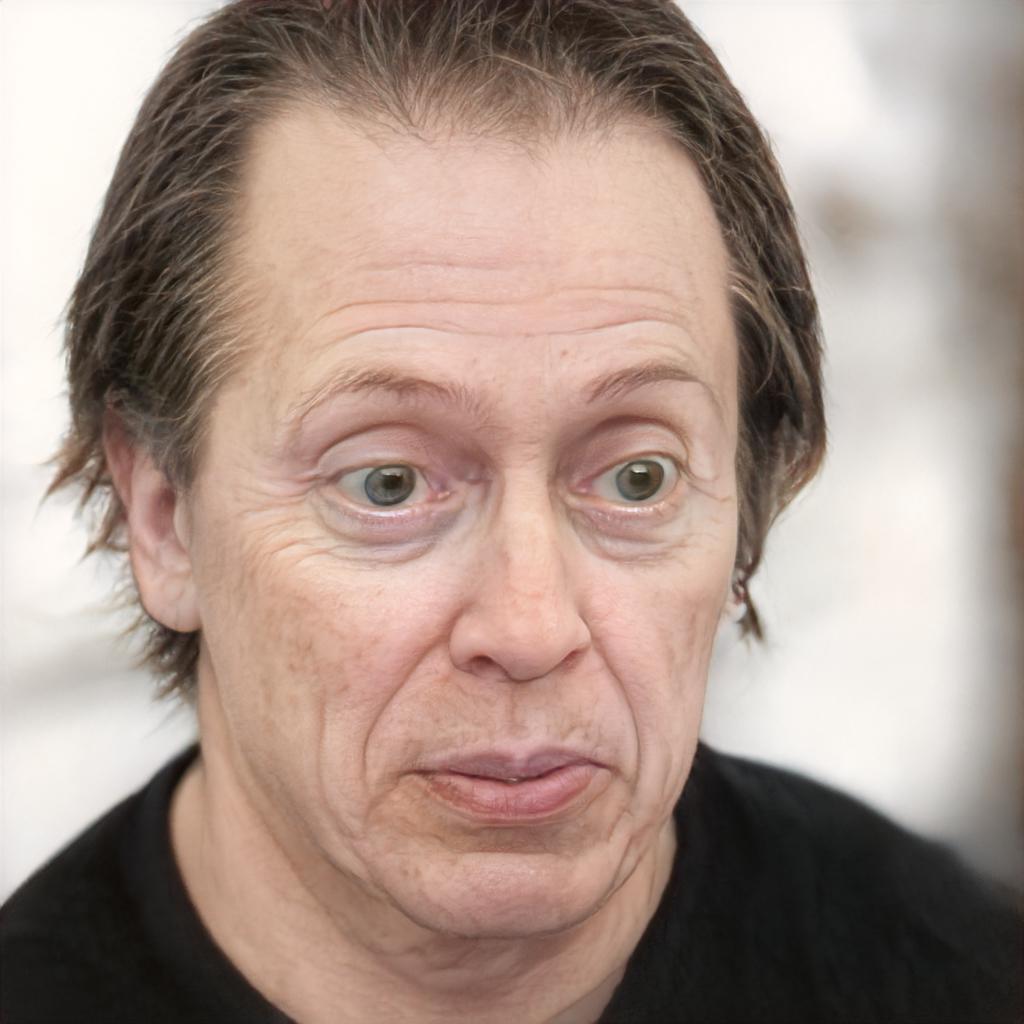} &
            \includegraphics[width=0.085\textwidth]{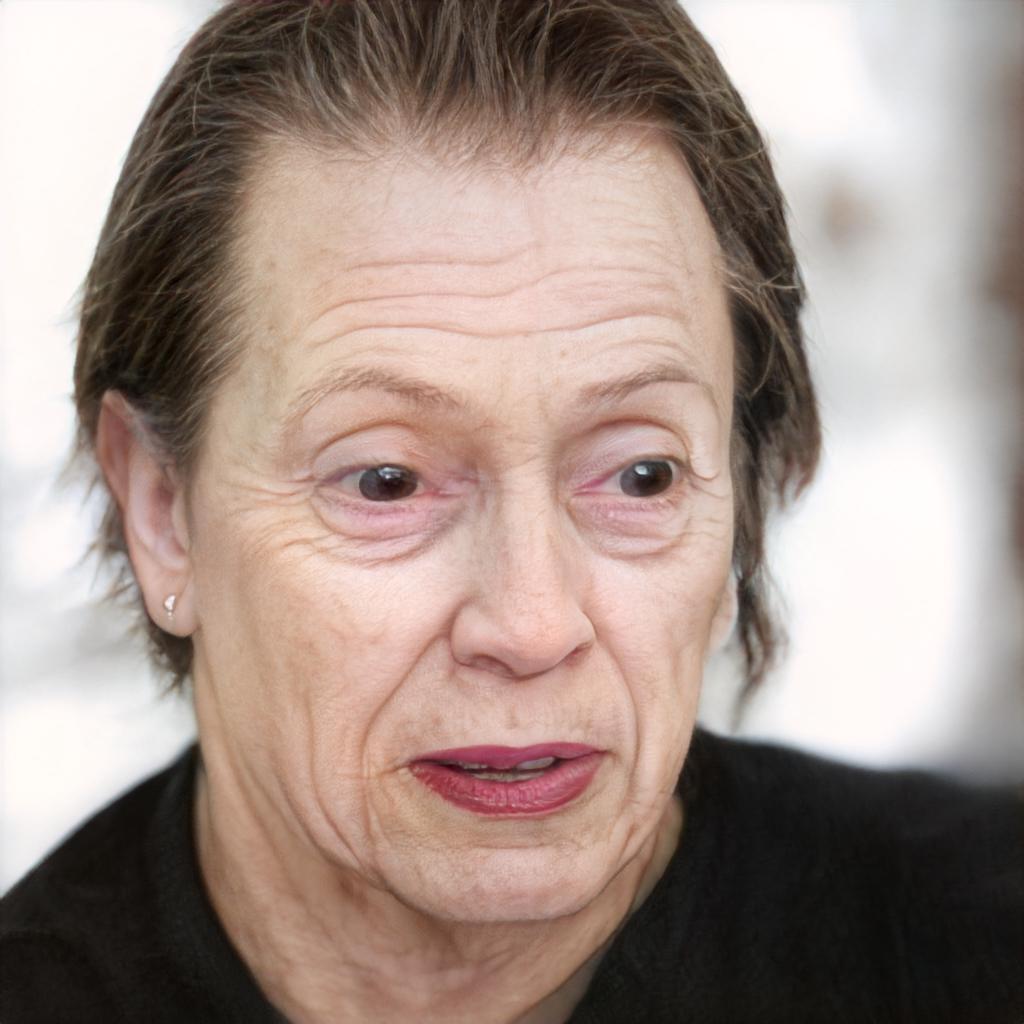} &
            \includegraphics[width=0.085\textwidth]{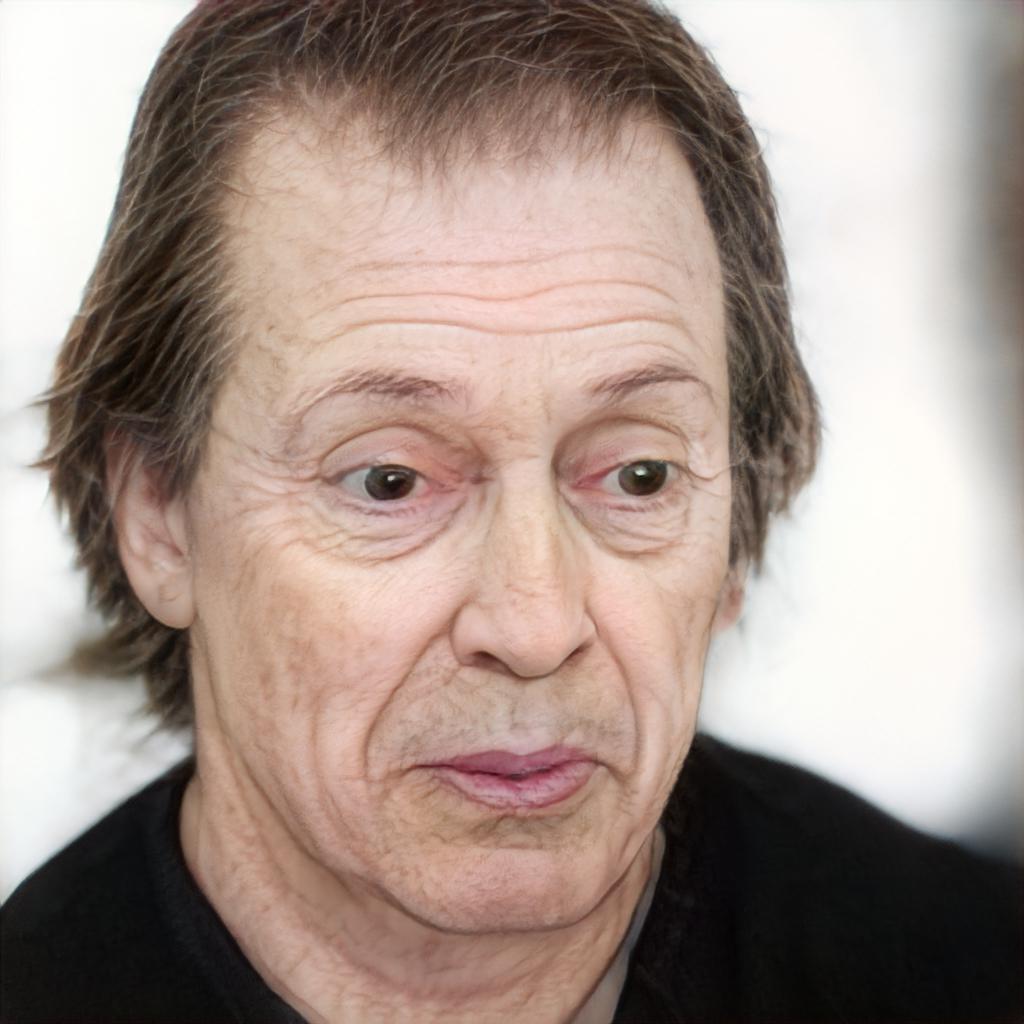} 
            \tabularnewline

            \includegraphics[width=0.085\textwidth]{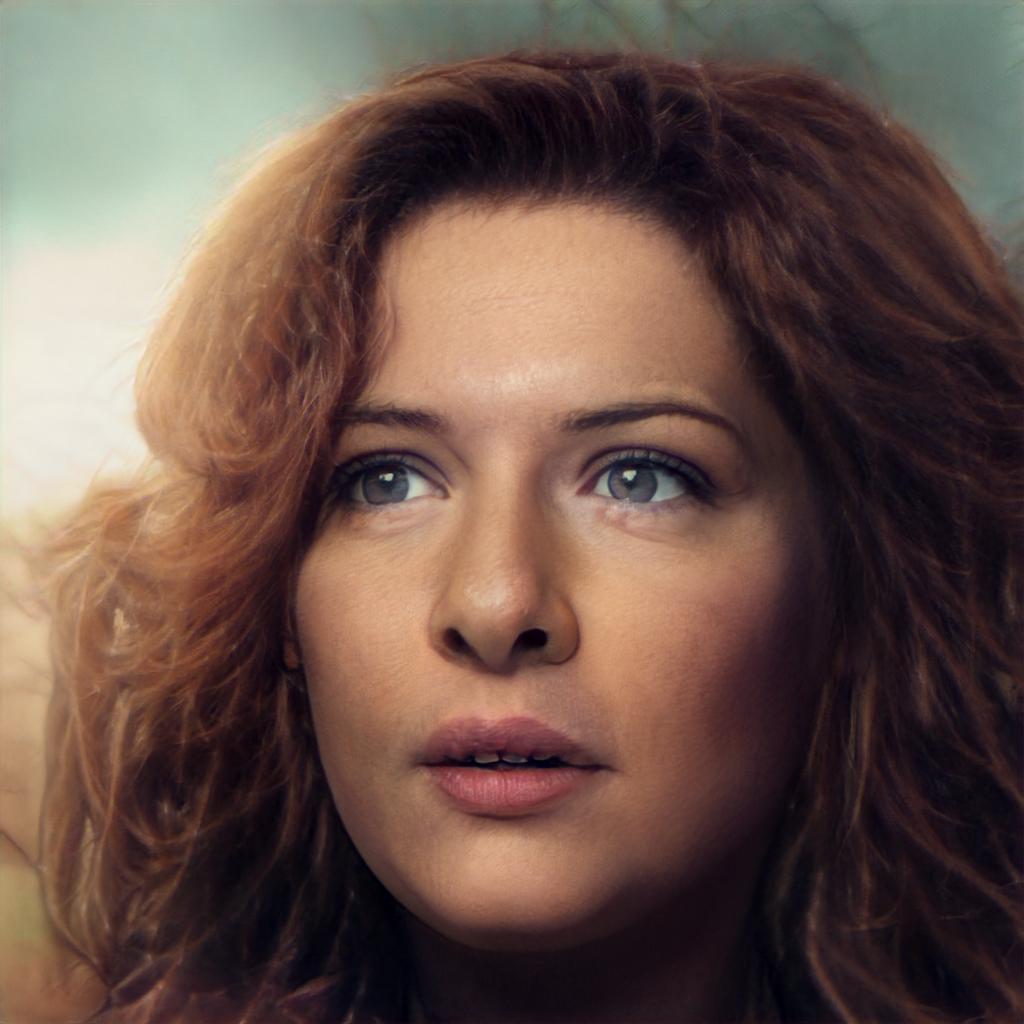} &
            \raisebox{0.2in}{\rotatebox[origin=t]{90}{5}} & 
            \includegraphics[width=0.085\textwidth]{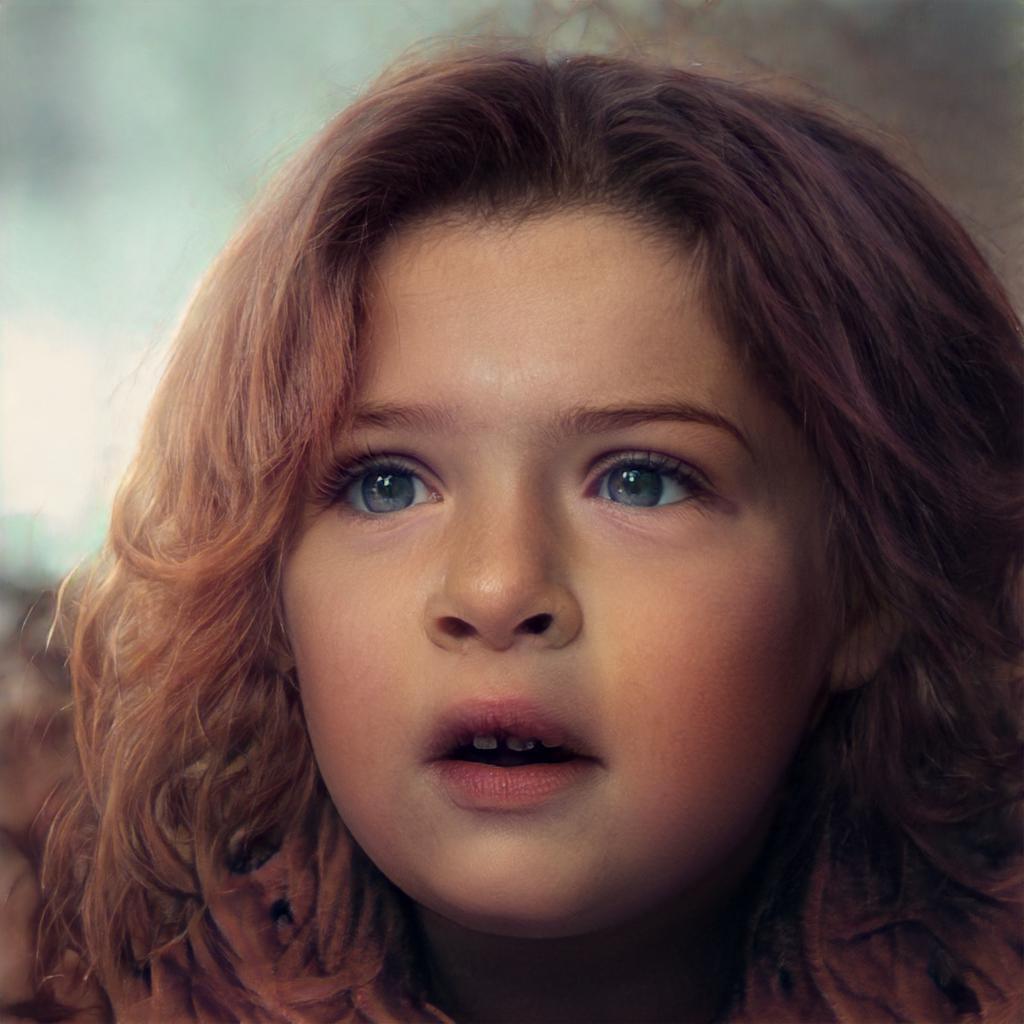} &
            \includegraphics[width=0.085\textwidth]{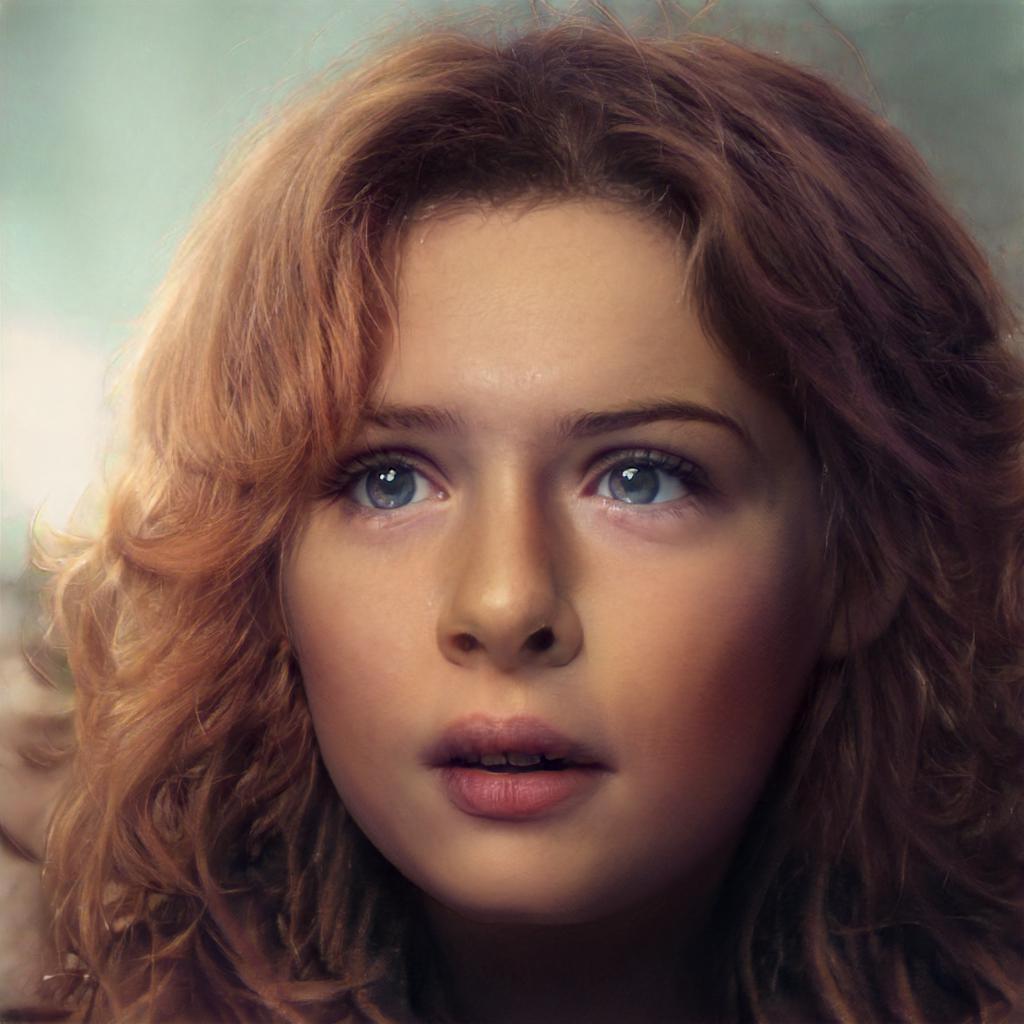} &
            \includegraphics[width=0.085\textwidth]{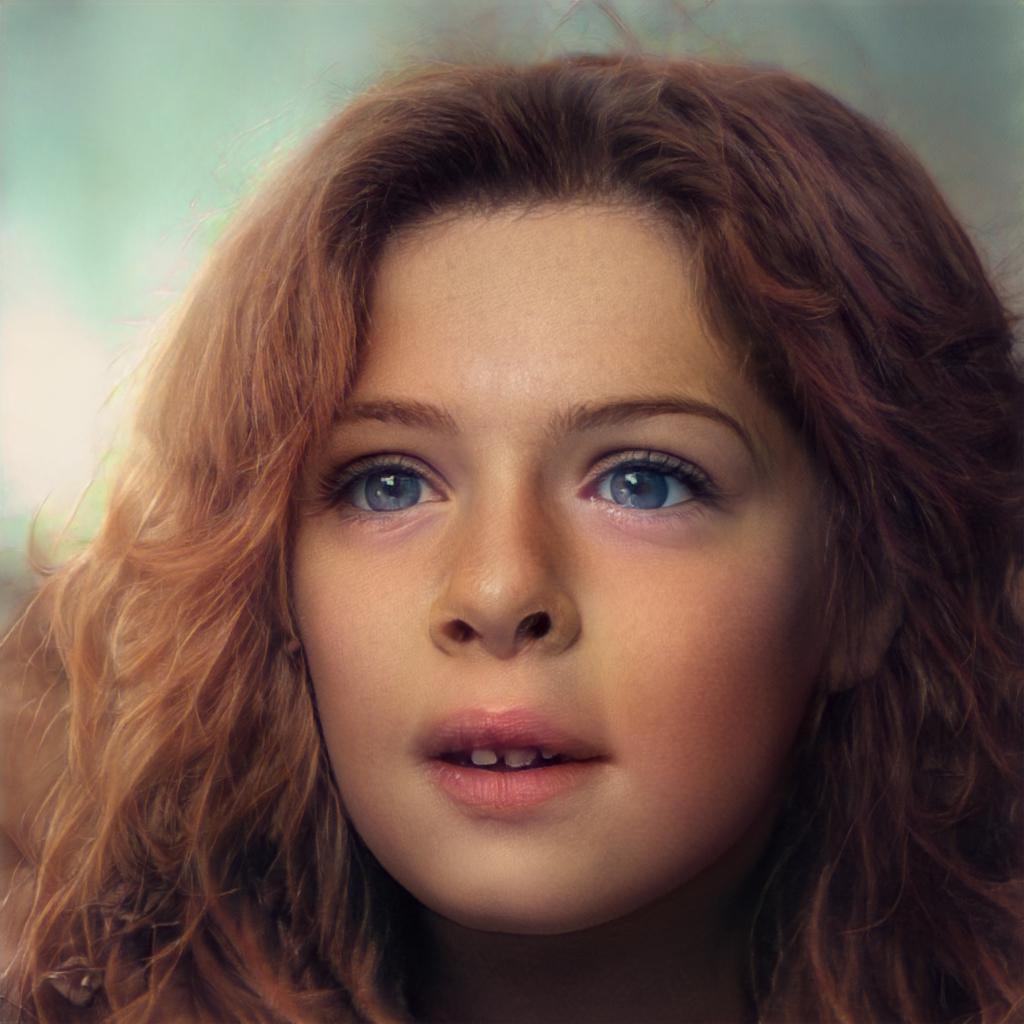} &
            \includegraphics[width=0.085\textwidth]{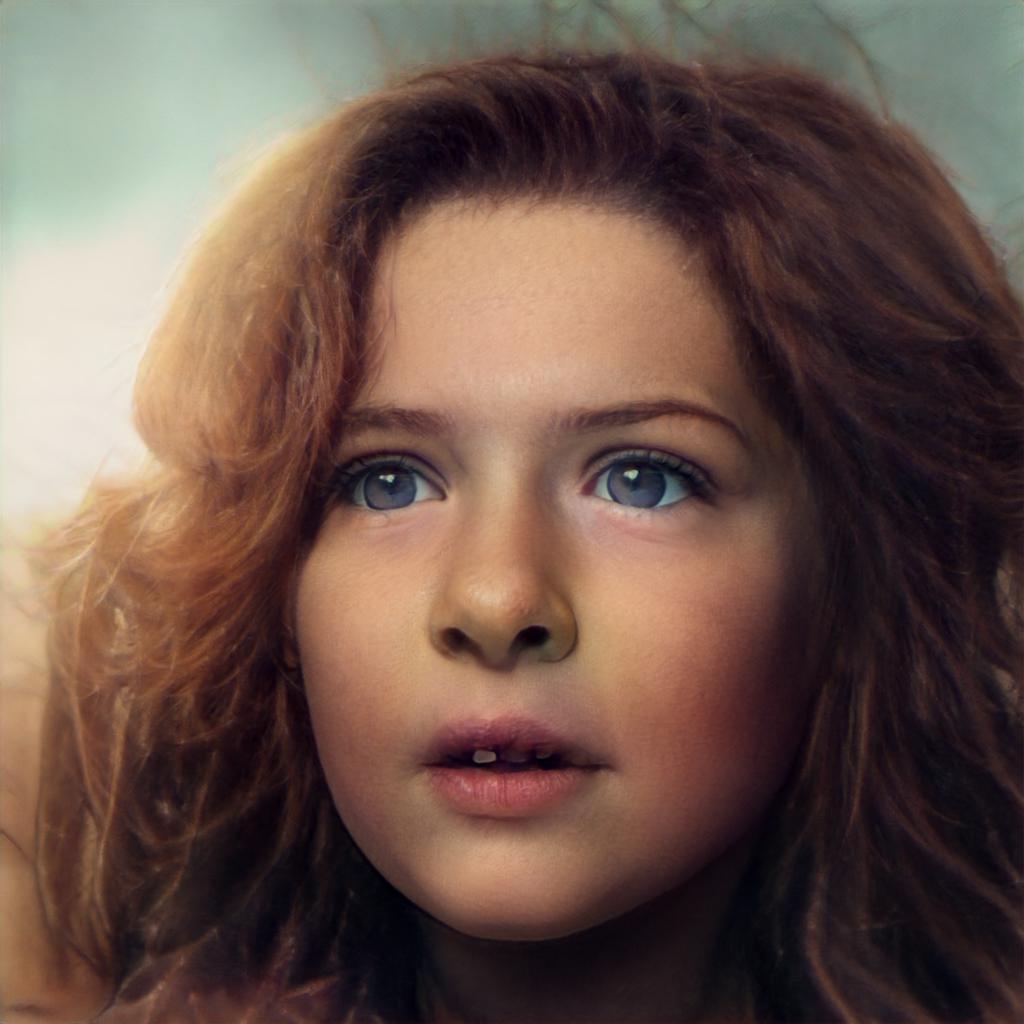} \\
            & \raisebox{0.2in}{\rotatebox[origin=t]{90}{45}} & 
            \includegraphics[width=0.085\textwidth]{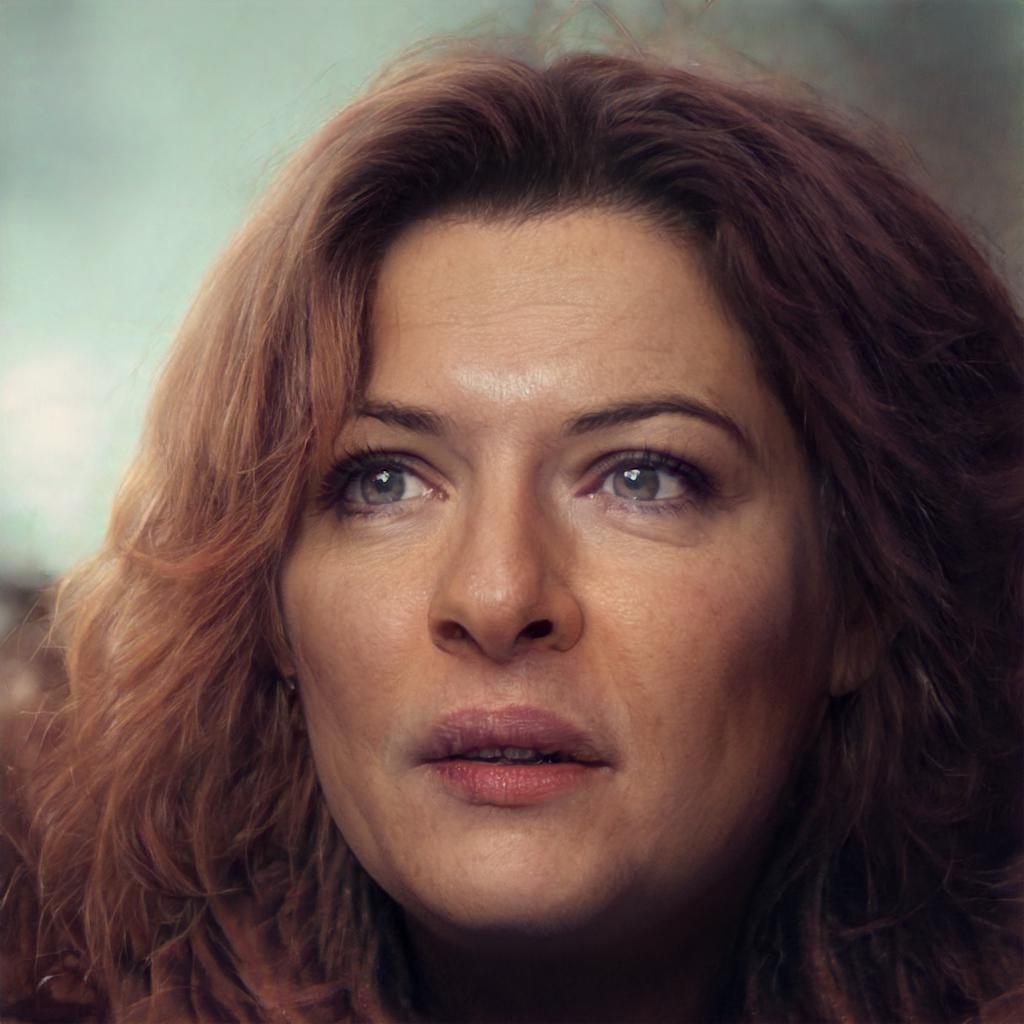} &
            \includegraphics[width=0.085\textwidth]{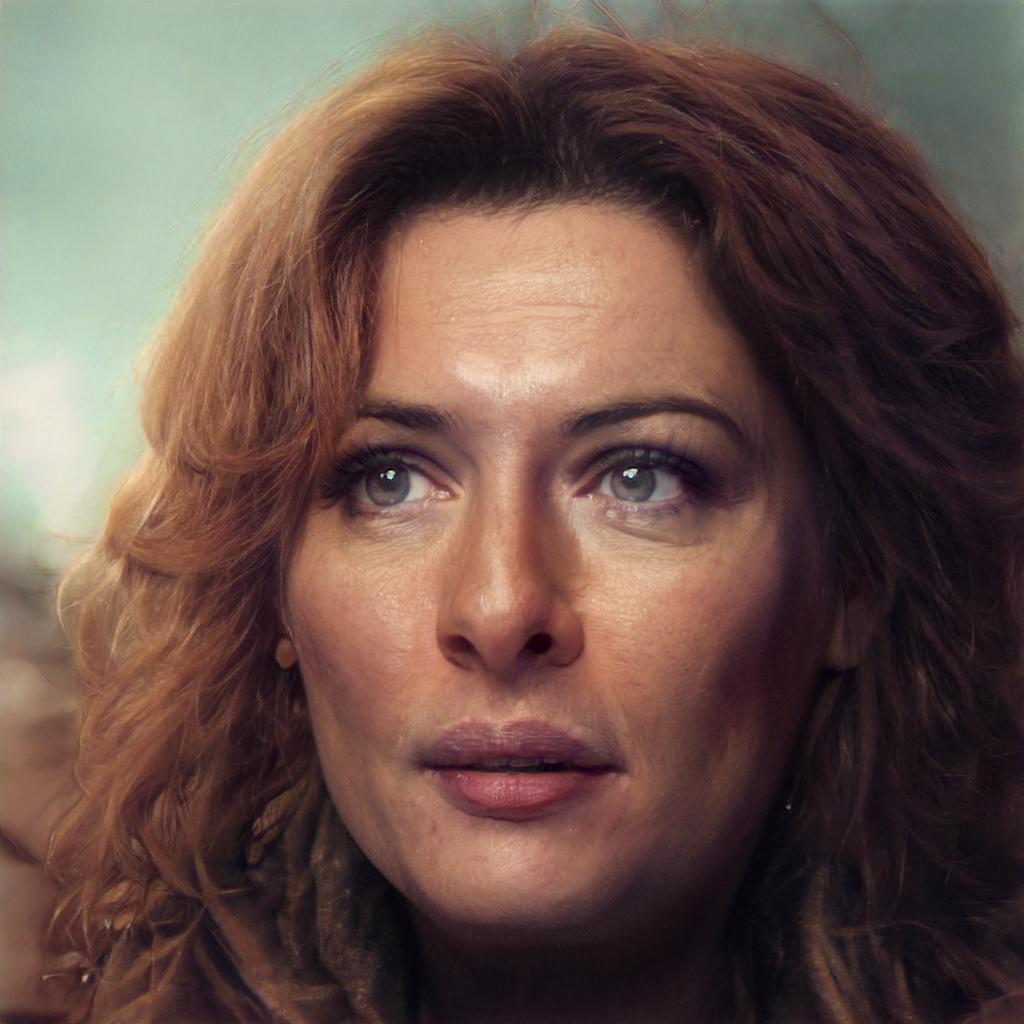} &
            \includegraphics[width=0.085\textwidth]{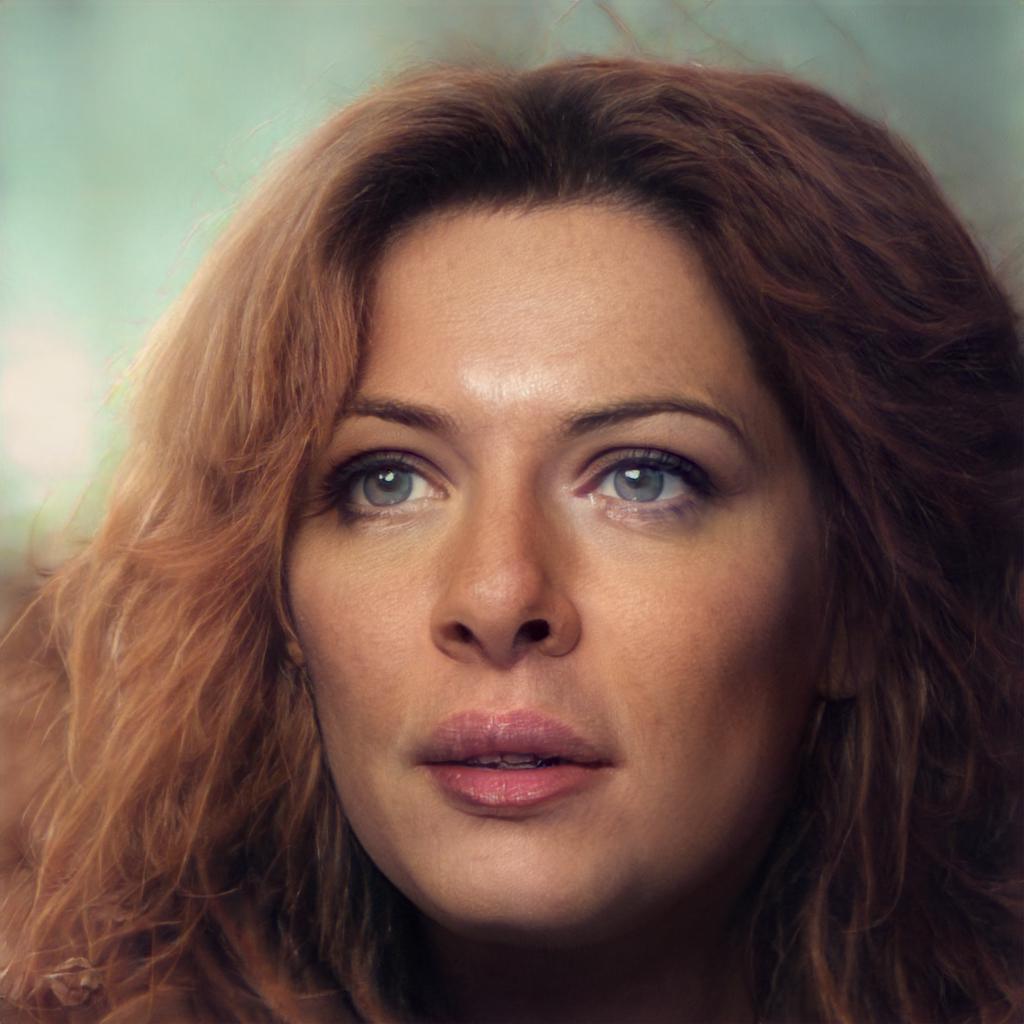} &
            \includegraphics[width=0.085\textwidth]{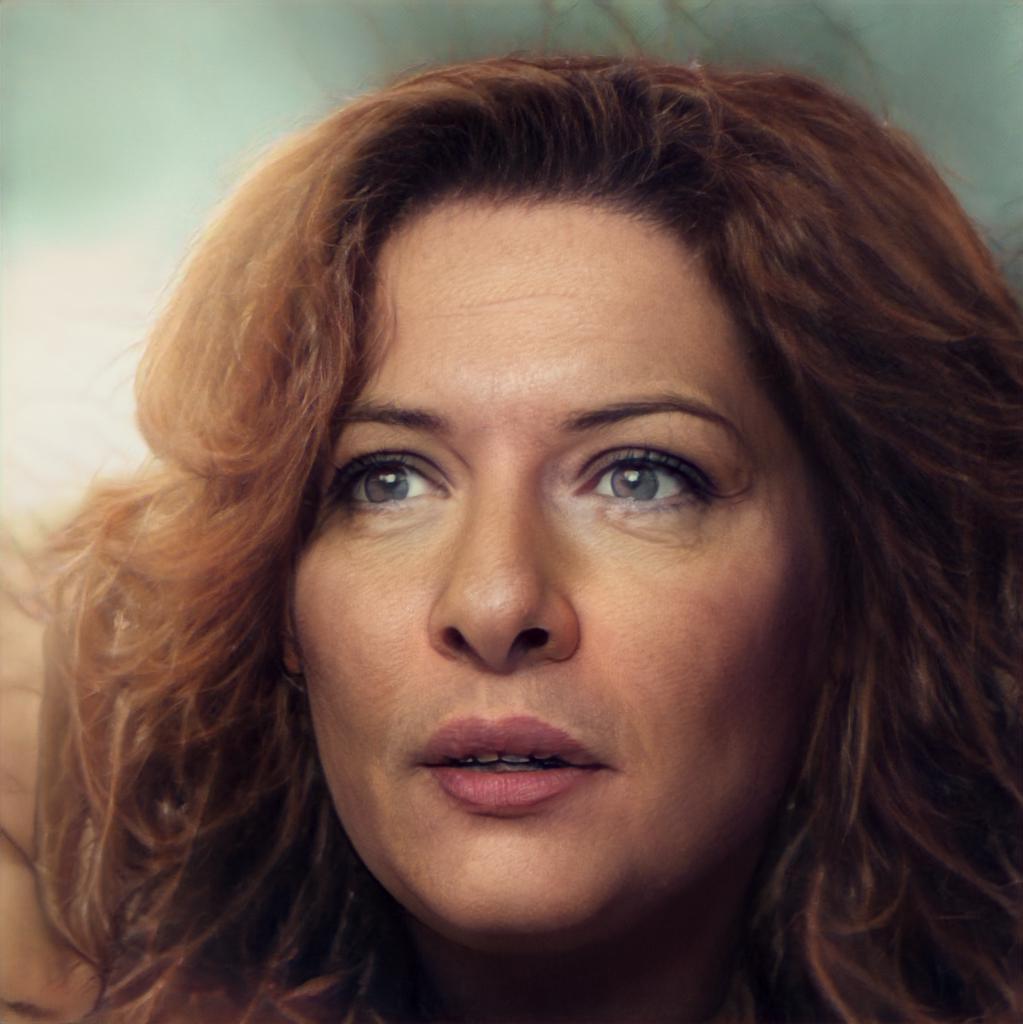} \\
            & \raisebox{0.2in}{\rotatebox[origin=t]{90}{85}} &
            \includegraphics[width=0.085\textwidth]{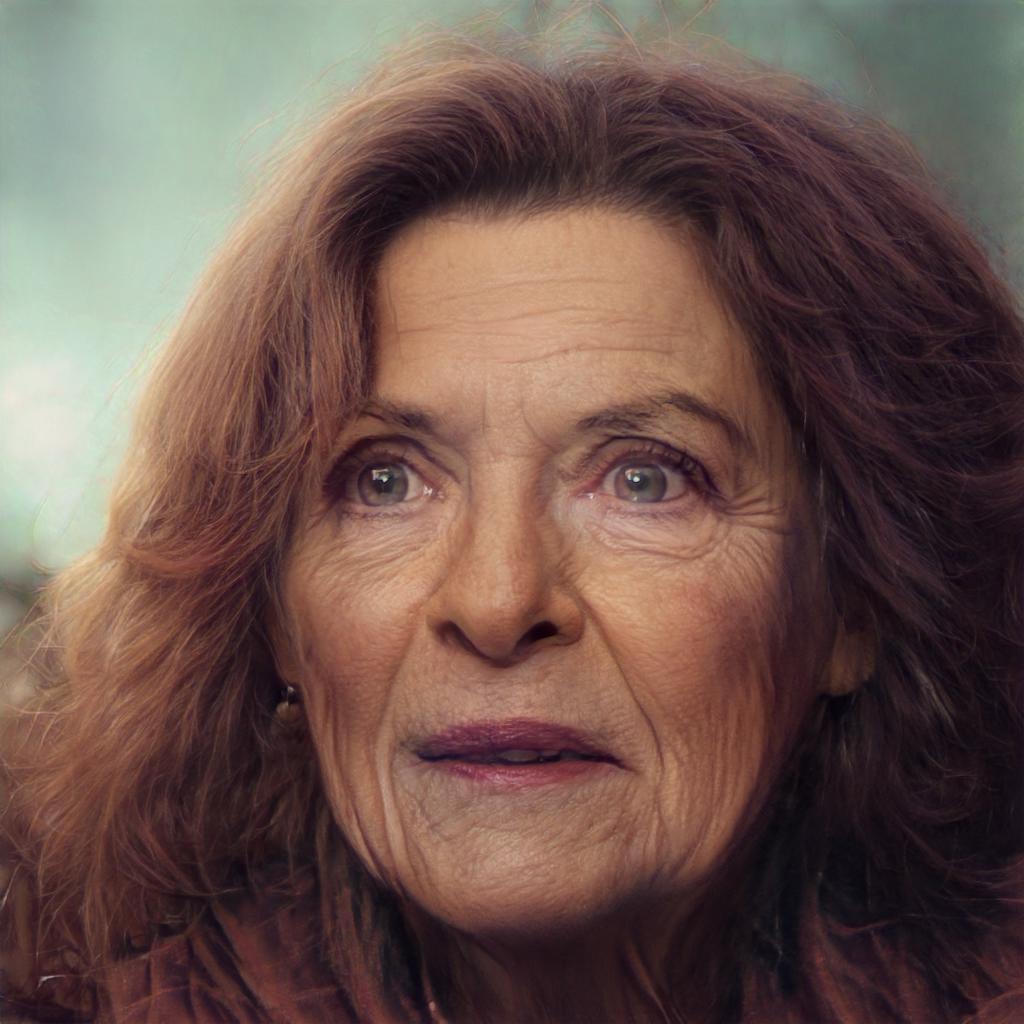} &
            \includegraphics[width=0.085\textwidth]{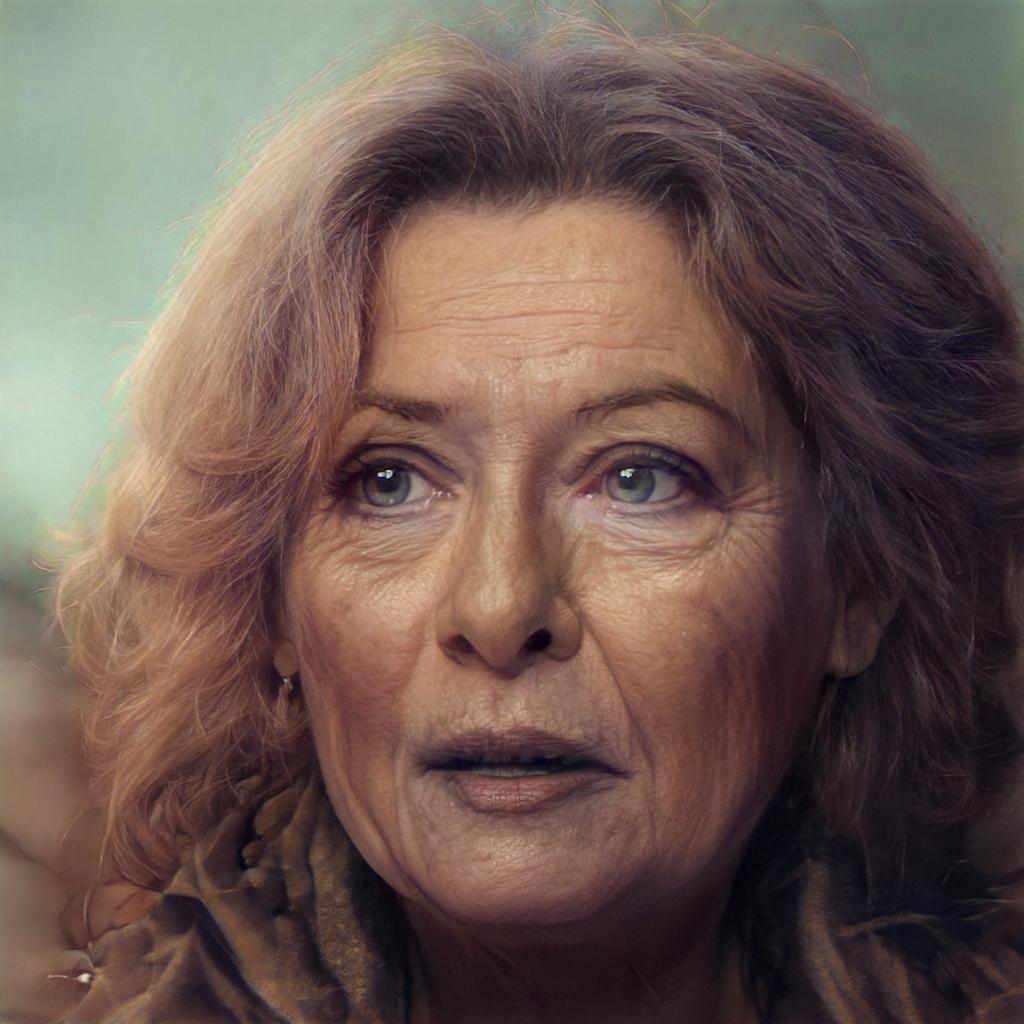} &
            \includegraphics[width=0.085\textwidth]{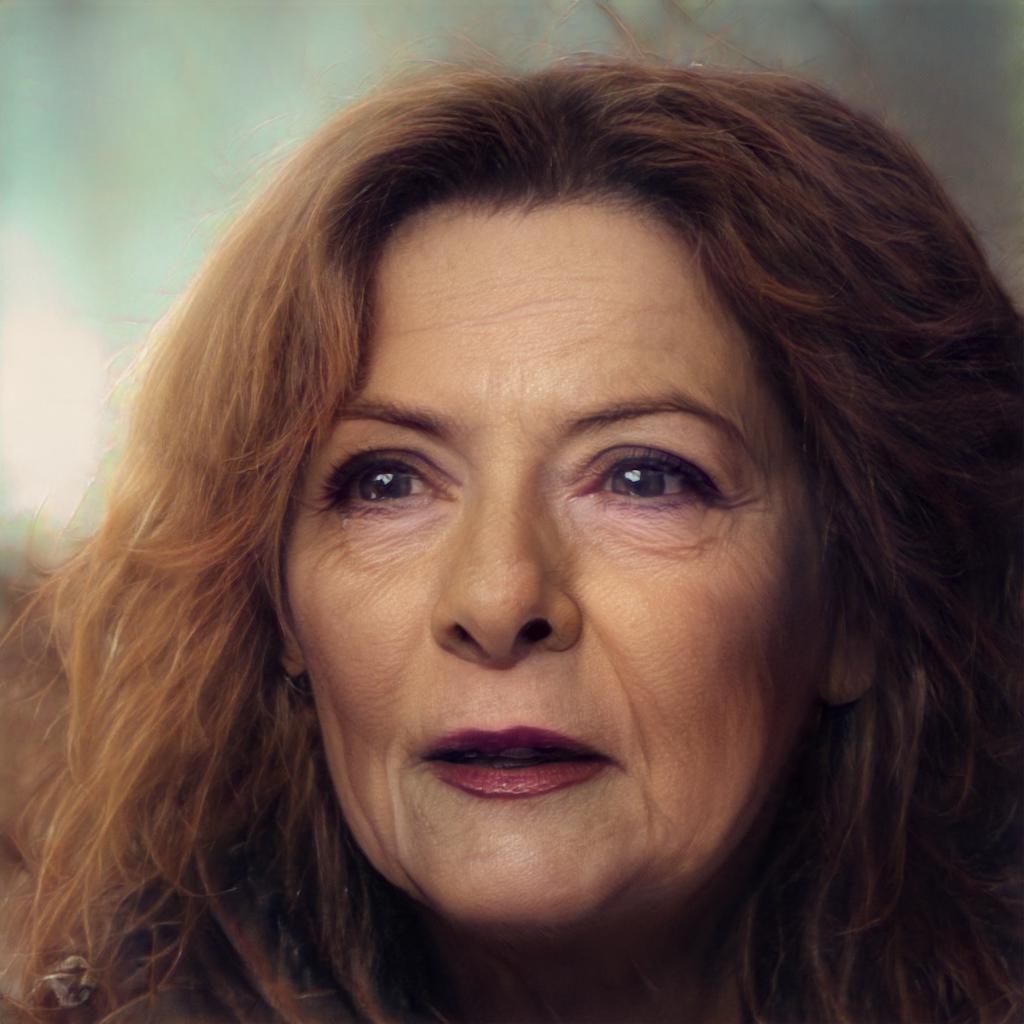} &
            \includegraphics[width=0.085\textwidth]{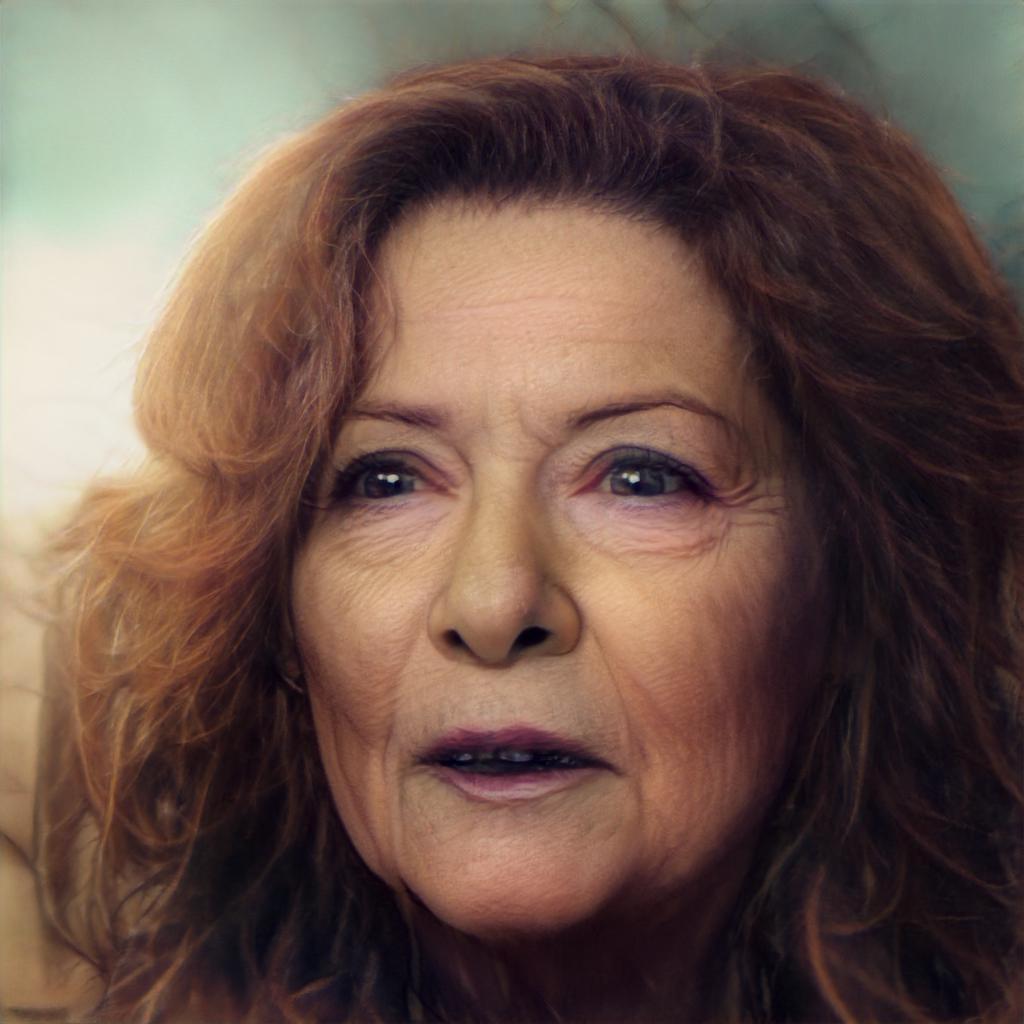} 
            \tabularnewline
            
            \includegraphics[width=0.085\textwidth]{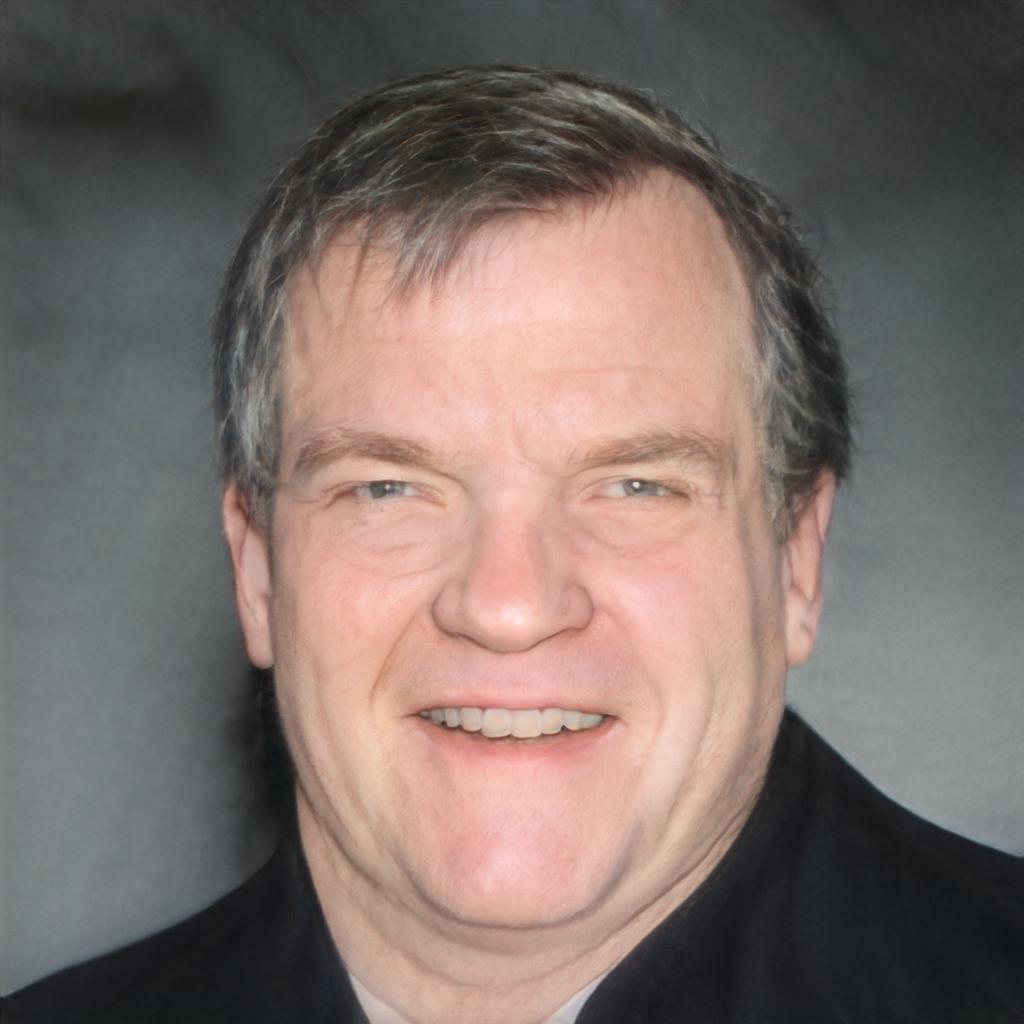} &
            \raisebox{0.2in}{\rotatebox[origin=t]{90}{5}} & 
            \includegraphics[width=0.085\textwidth]{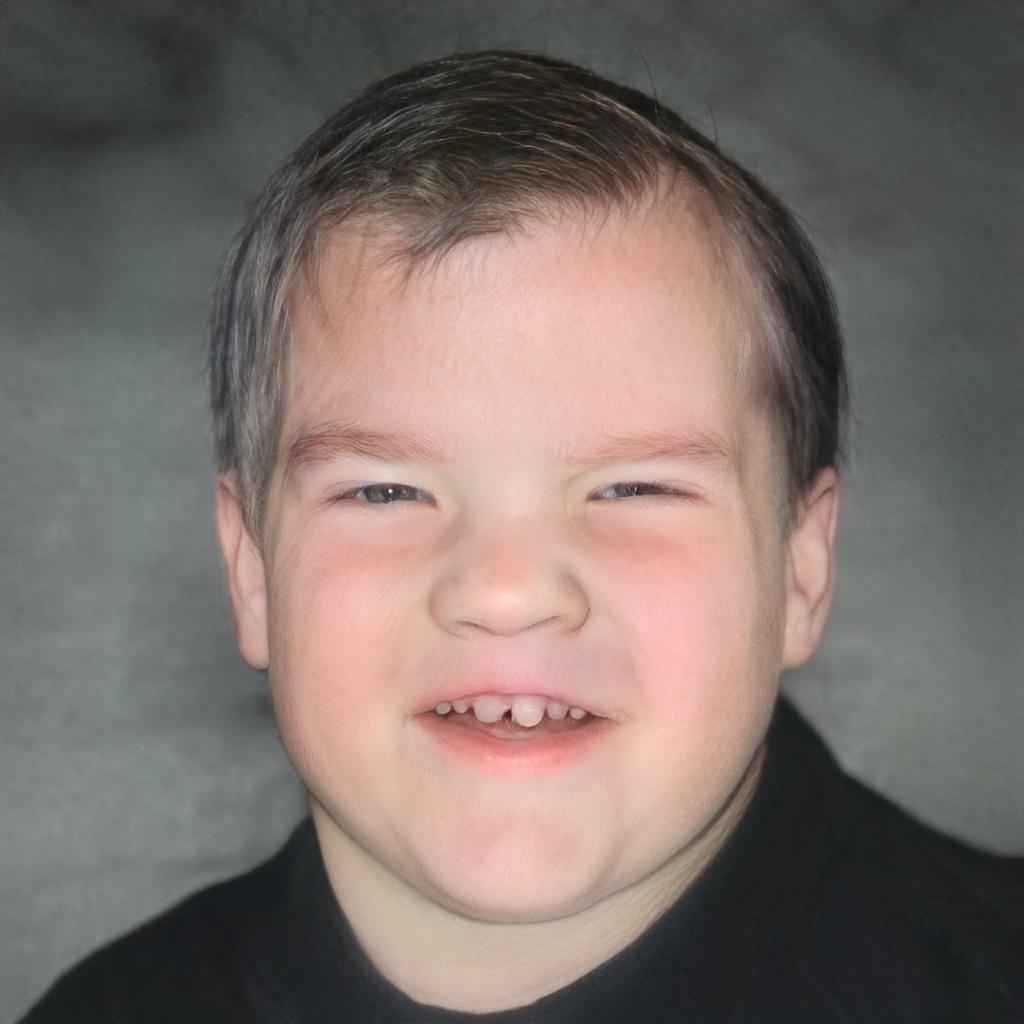} &
            \includegraphics[width=0.085\textwidth]{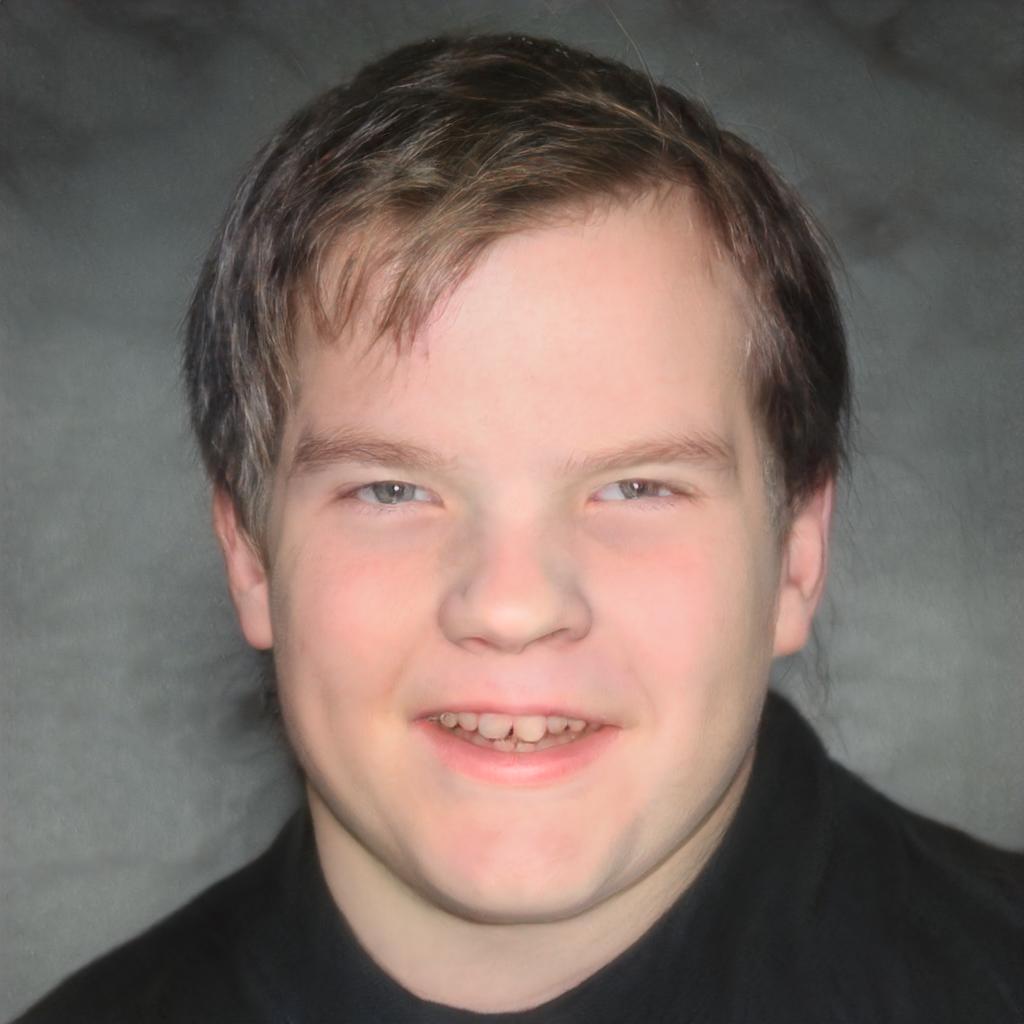} &
            \includegraphics[width=0.085\textwidth]{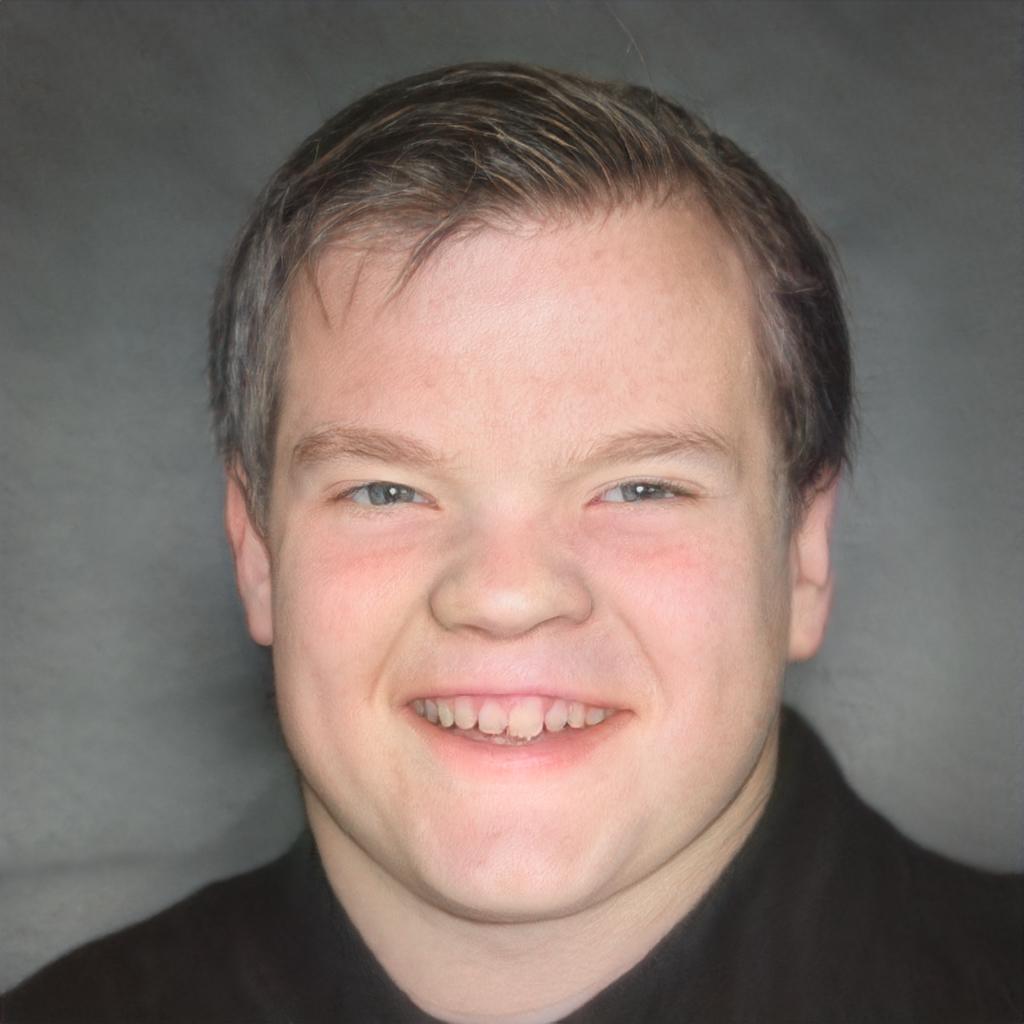} &
            \includegraphics[width=0.085\textwidth]{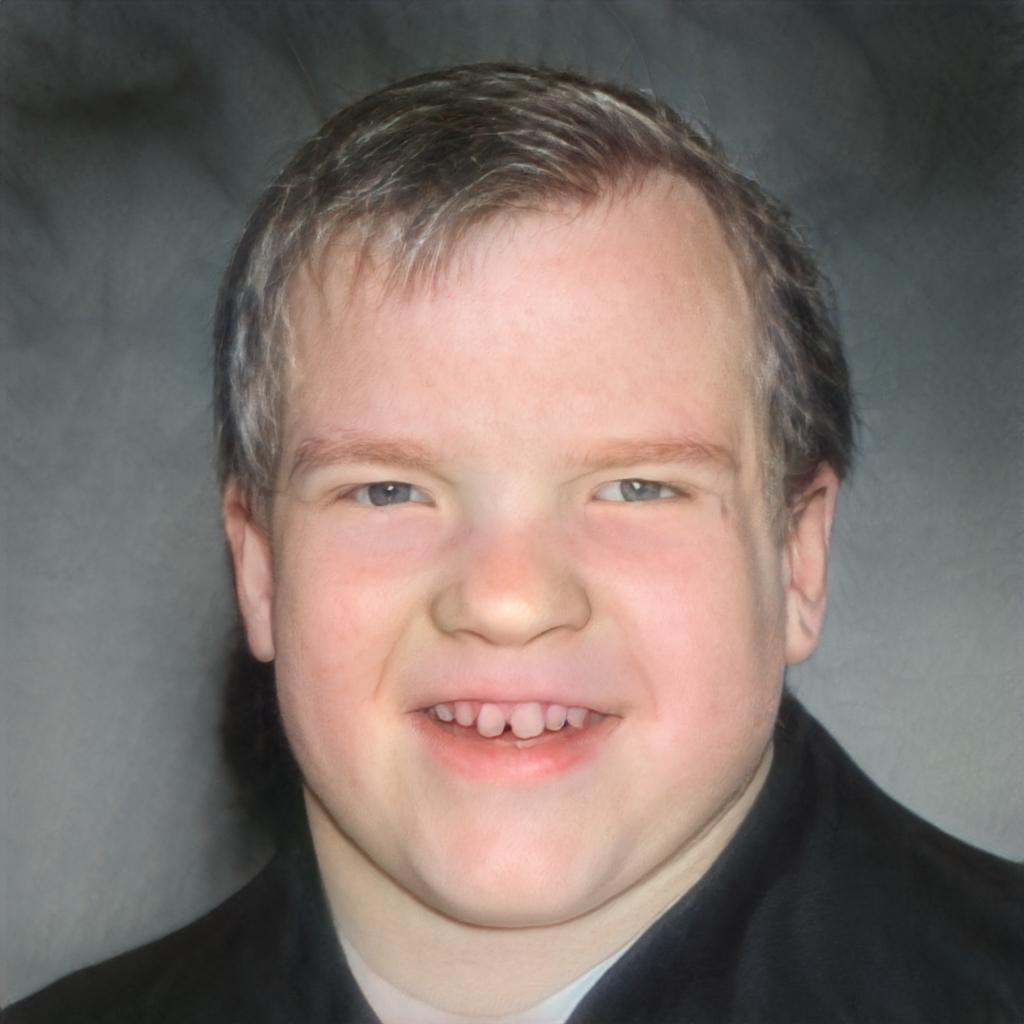} \\
            & \raisebox{0.2in}{\rotatebox[origin=t]{90}{45}} & 
            \includegraphics[width=0.085\textwidth]{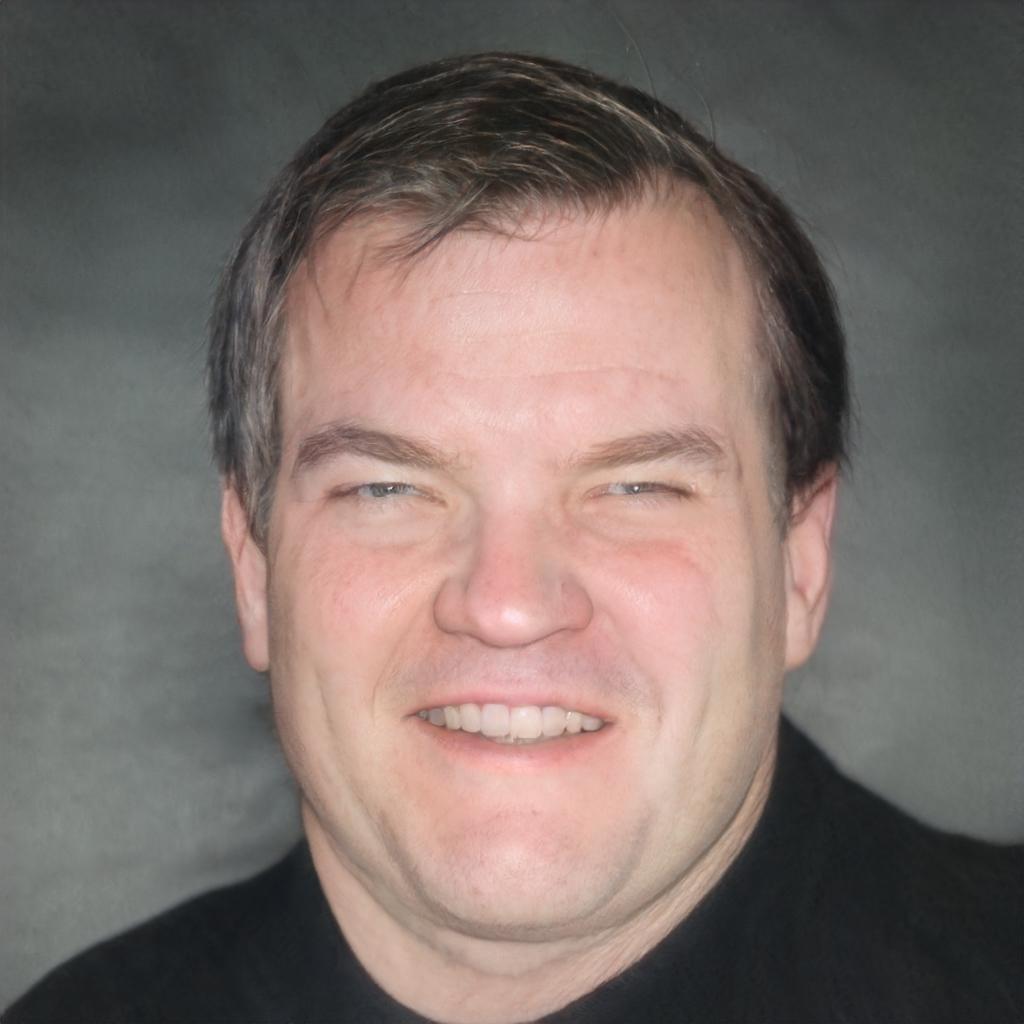} &
            \includegraphics[width=0.085\textwidth]{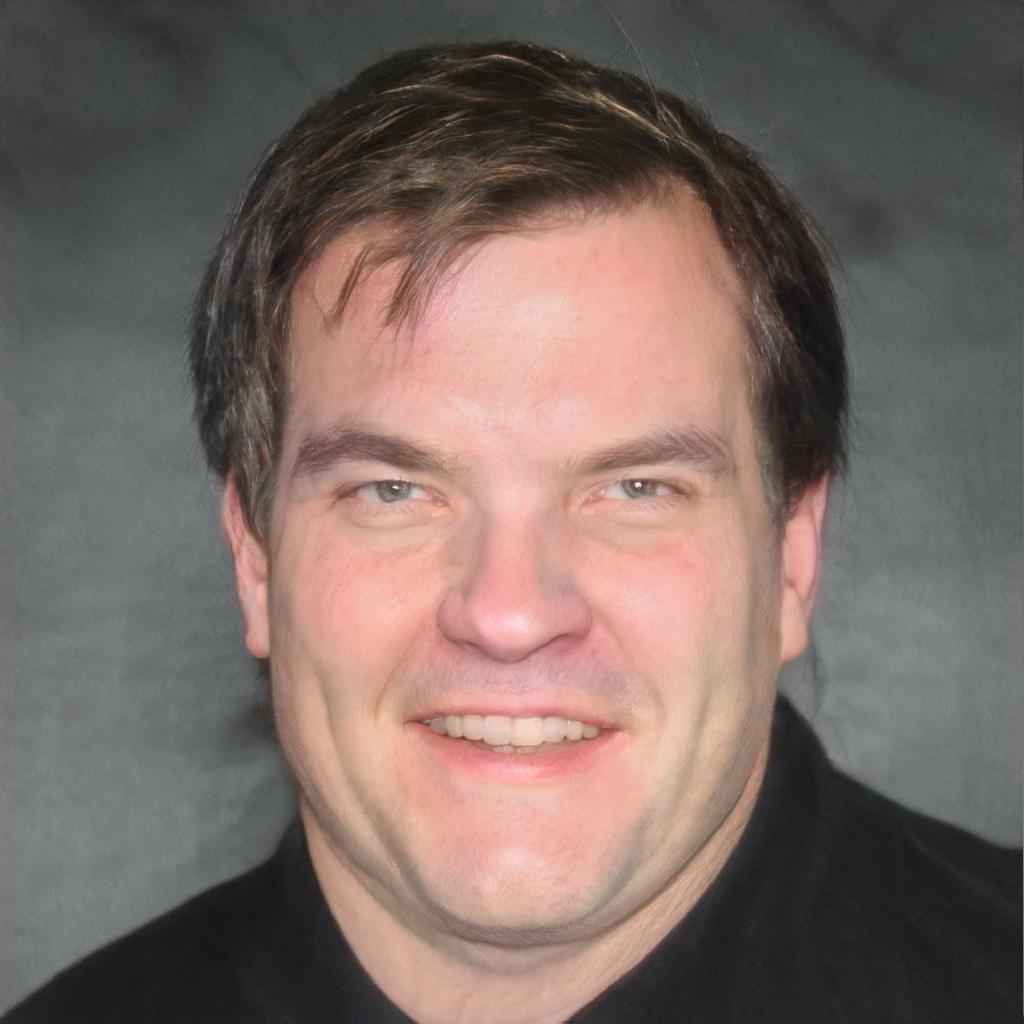} &
            \includegraphics[width=0.085\textwidth]{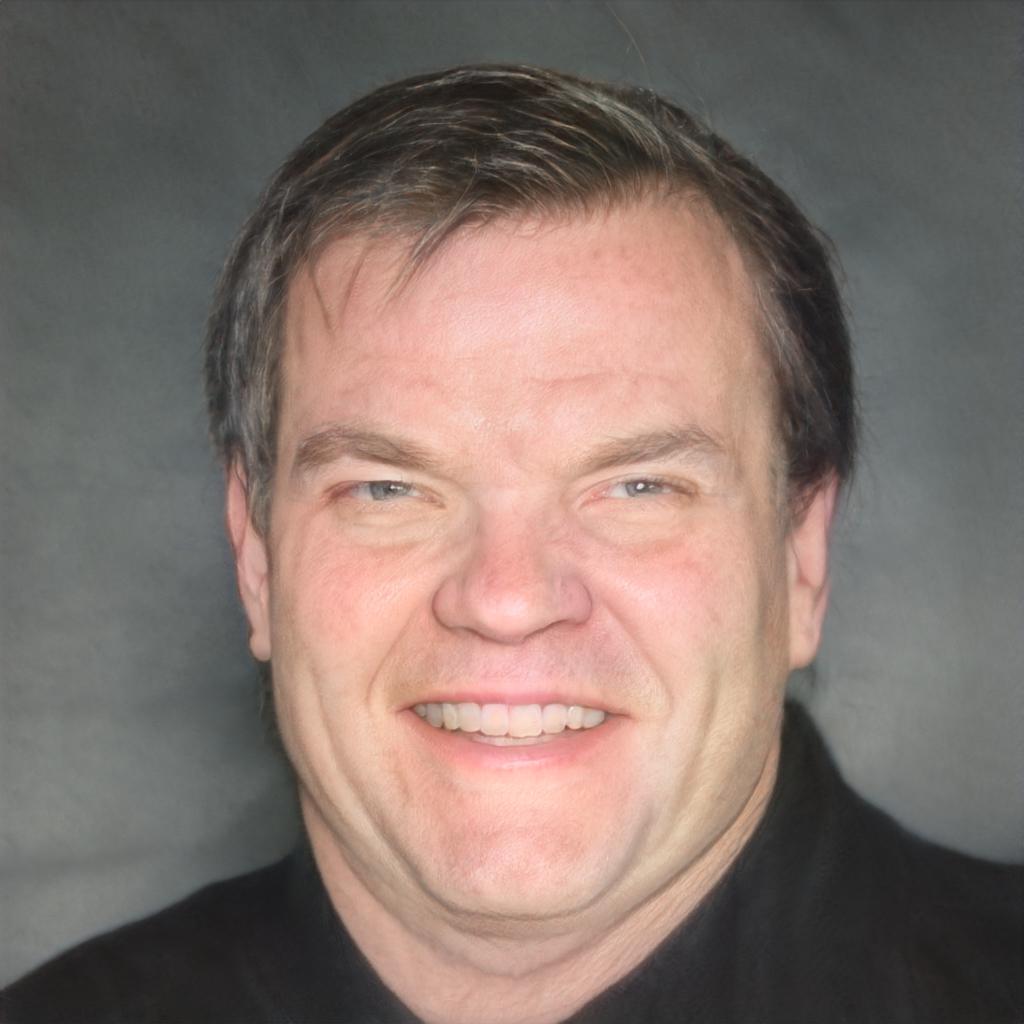} &
            \includegraphics[width=0.085\textwidth]{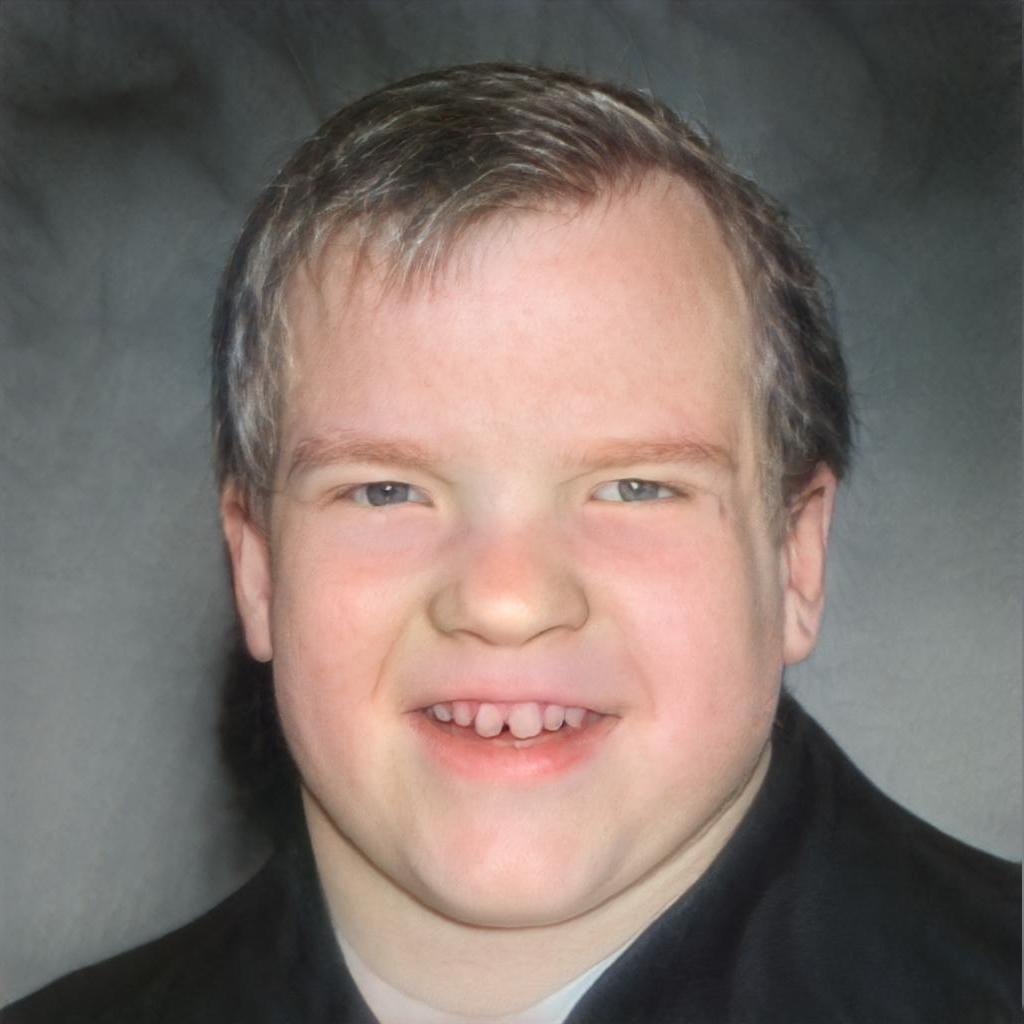} \\
            & \raisebox{0.2in}{\rotatebox[origin=t]{90}{85}} &
            \includegraphics[width=0.085\textwidth]{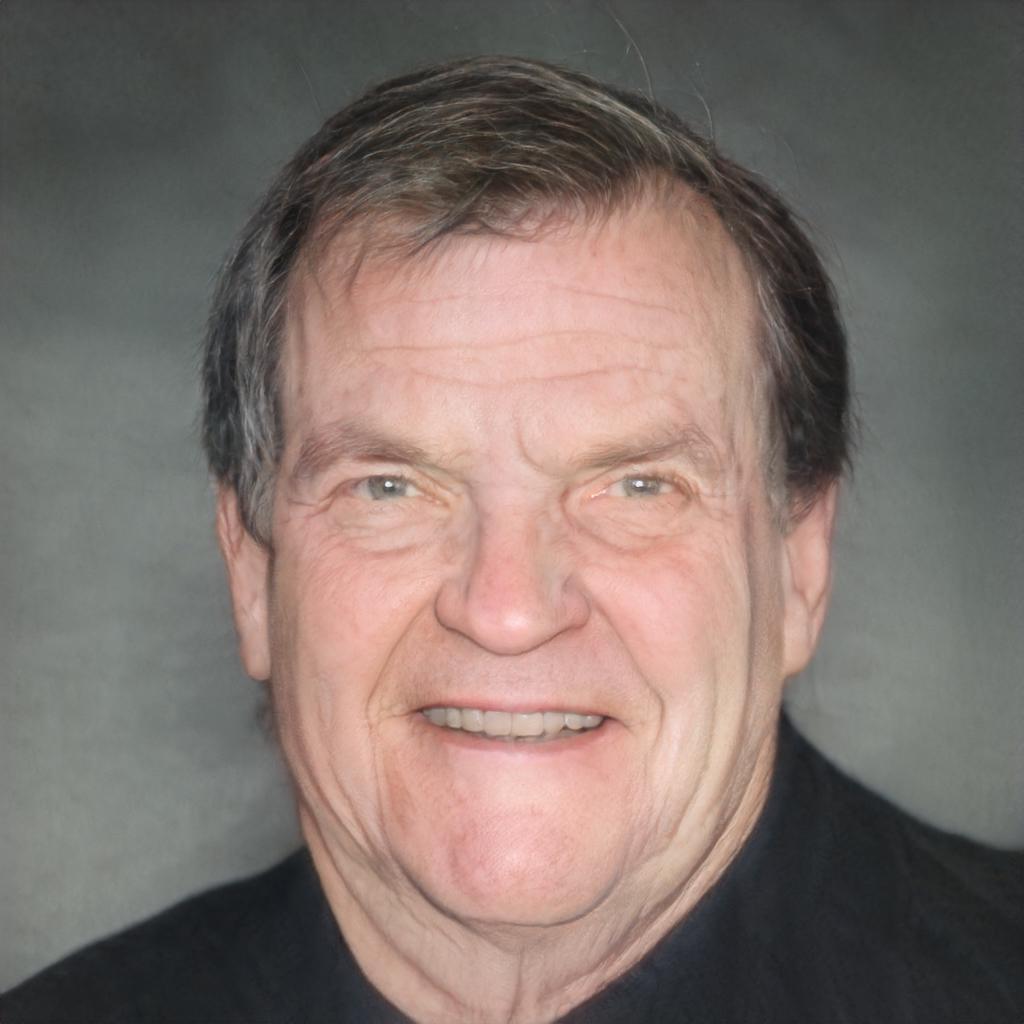} &
            \includegraphics[width=0.085\textwidth]{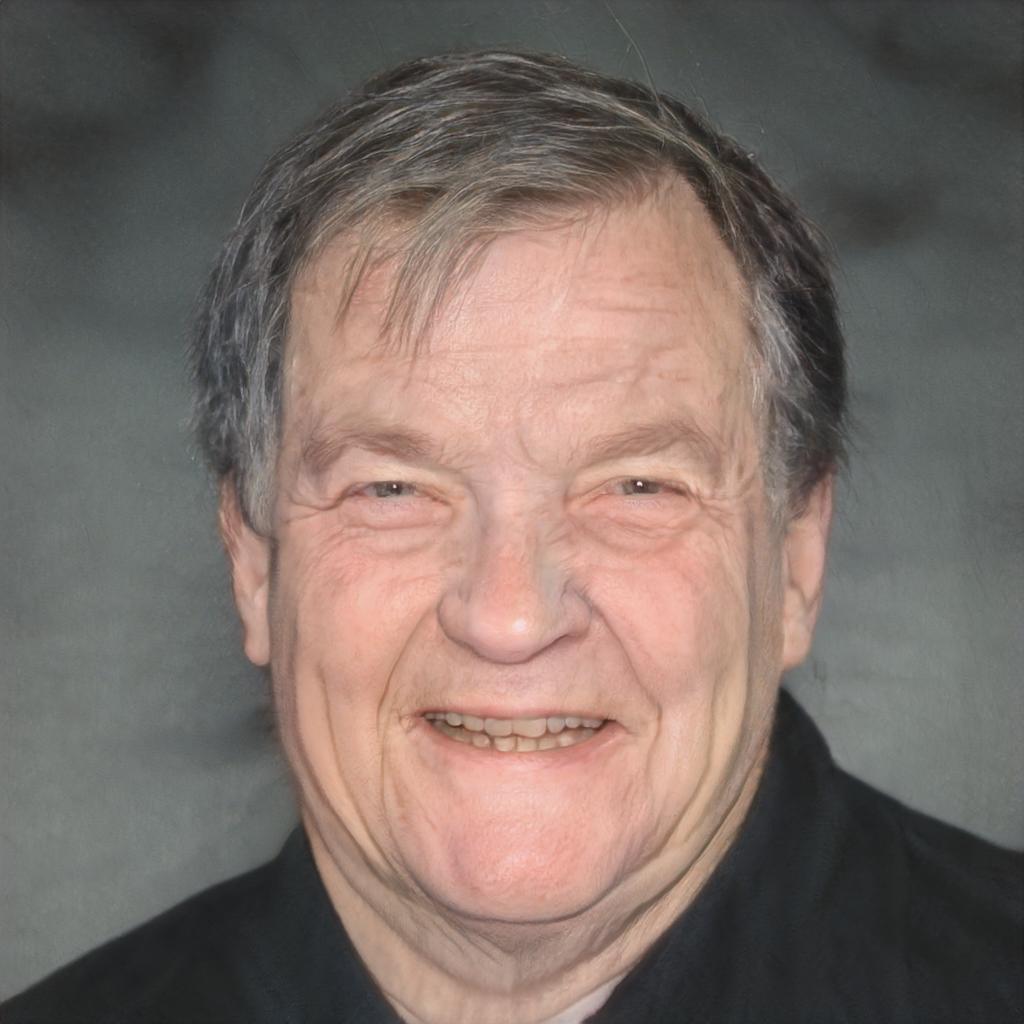} &
            \includegraphics[width=0.085\textwidth]{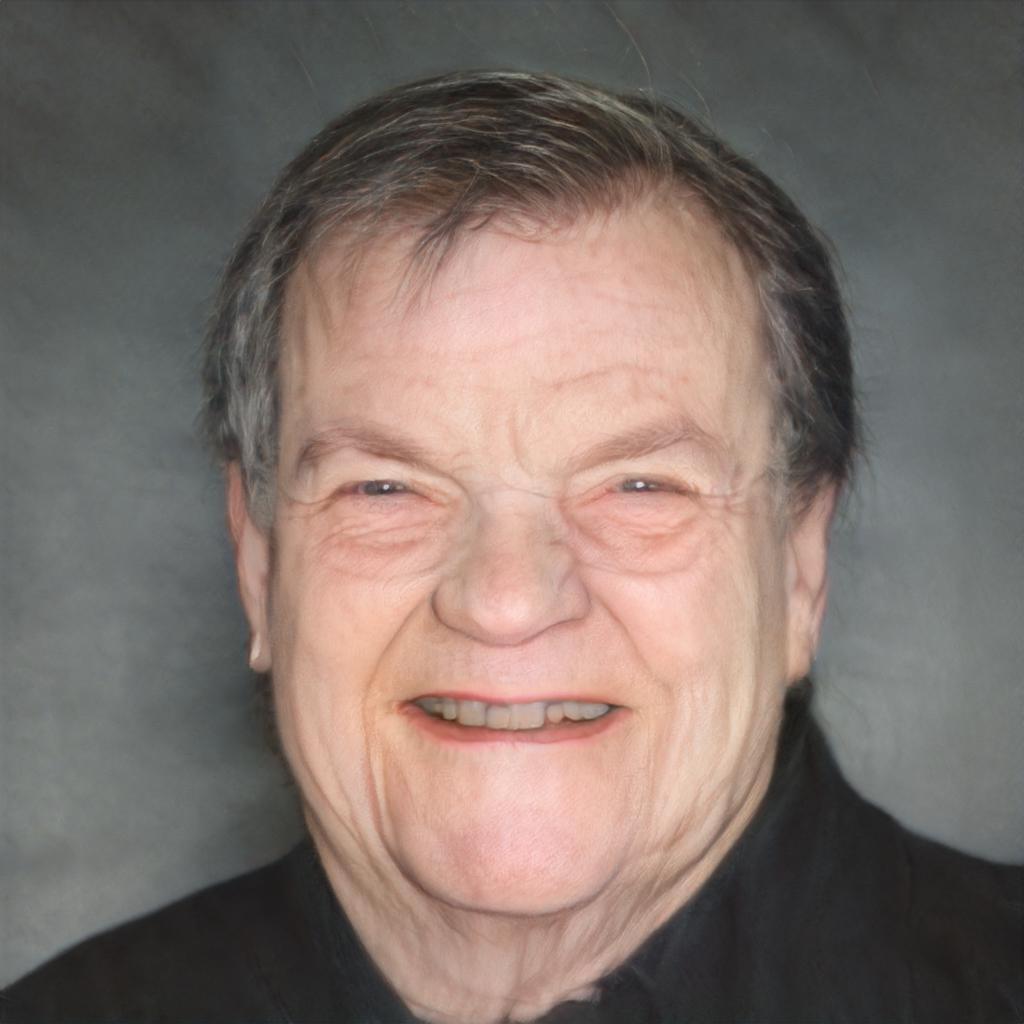} &
            \includegraphics[width=0.085\textwidth]{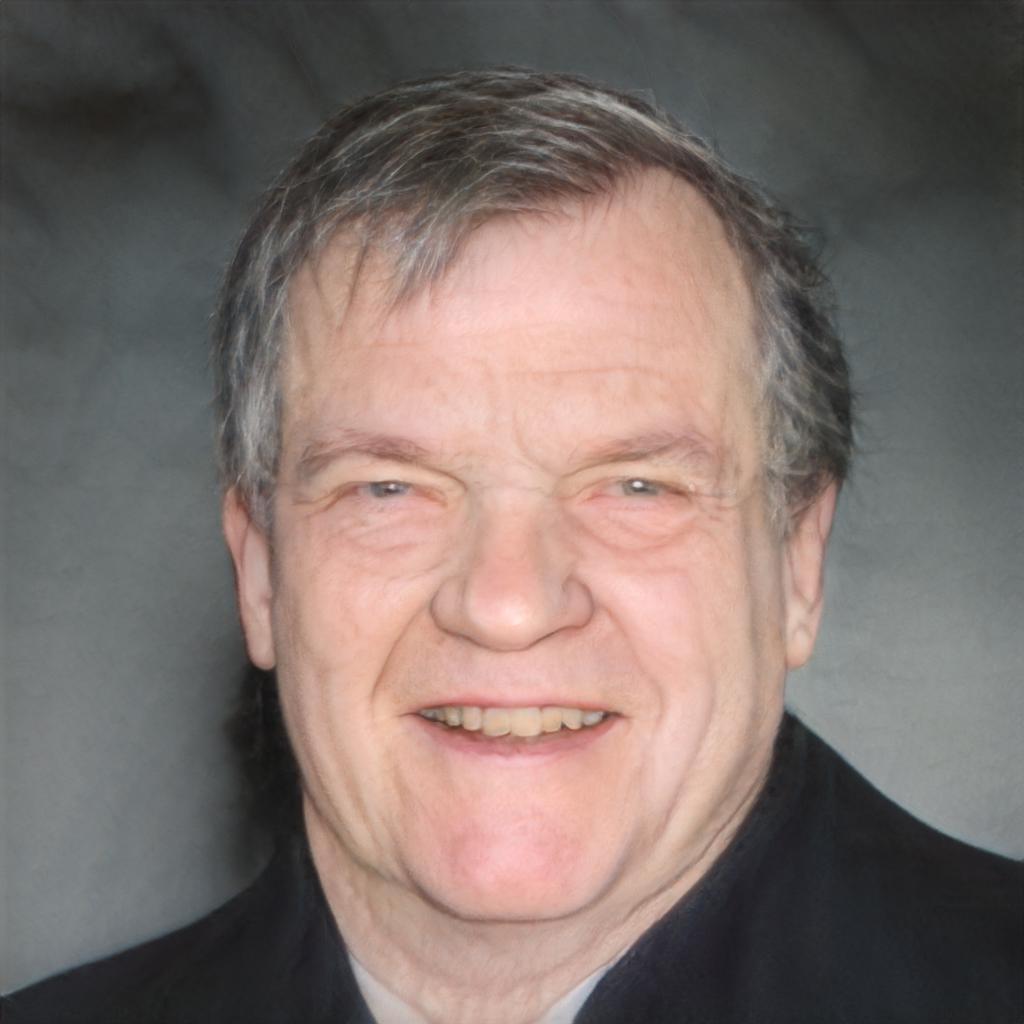} 
            \tabularnewline
            
            \begin{tabular}{@{}c@{}}Inversion \end{tabular} & &
            \begin{tabular}{@{}c@{}}SAM \end{tabular} &
            \begin{tabular}{@{}c@{}}w/o $L_2$ \& \\ LPIPS \end{tabular} &
            \begin{tabular}{@{}c@{}}w/o \\ Cycle \end{tabular} &
            \begin{tabular}{@{}c@{}}w/o \\ $w$-Reg \end{tabular} 
            
    \end{tabular}
    \vspace{-0.2cm}
    \caption{Ablation study on the loss objectives employed during training of SAM. For each column, we train SAM using the same configuration without the specified loss objective. As shown, although no one variant is significantly better than the others, we find that our final configuration is able to better capture the target age of $5$ and changes in face wrinkles when translating to the age of $85$.}
    \label{fig:ablation_losses}
\end{figure}

%% file: figures/appendix/appendix_life_comparison.tex
\begin{figure*}
    \centering
    \setlength{\belowcaptionskip}{-2.5pt}
    \setlength{\tabcolsep}{1pt}
    \centering
        \begin{tabular}{c c c c c c c c c}
        Inversion & & & 3-6 & 7-9 & 15-19 & 30-39 & 50-69 \\
        \includegraphics[width=0.10\textwidth]{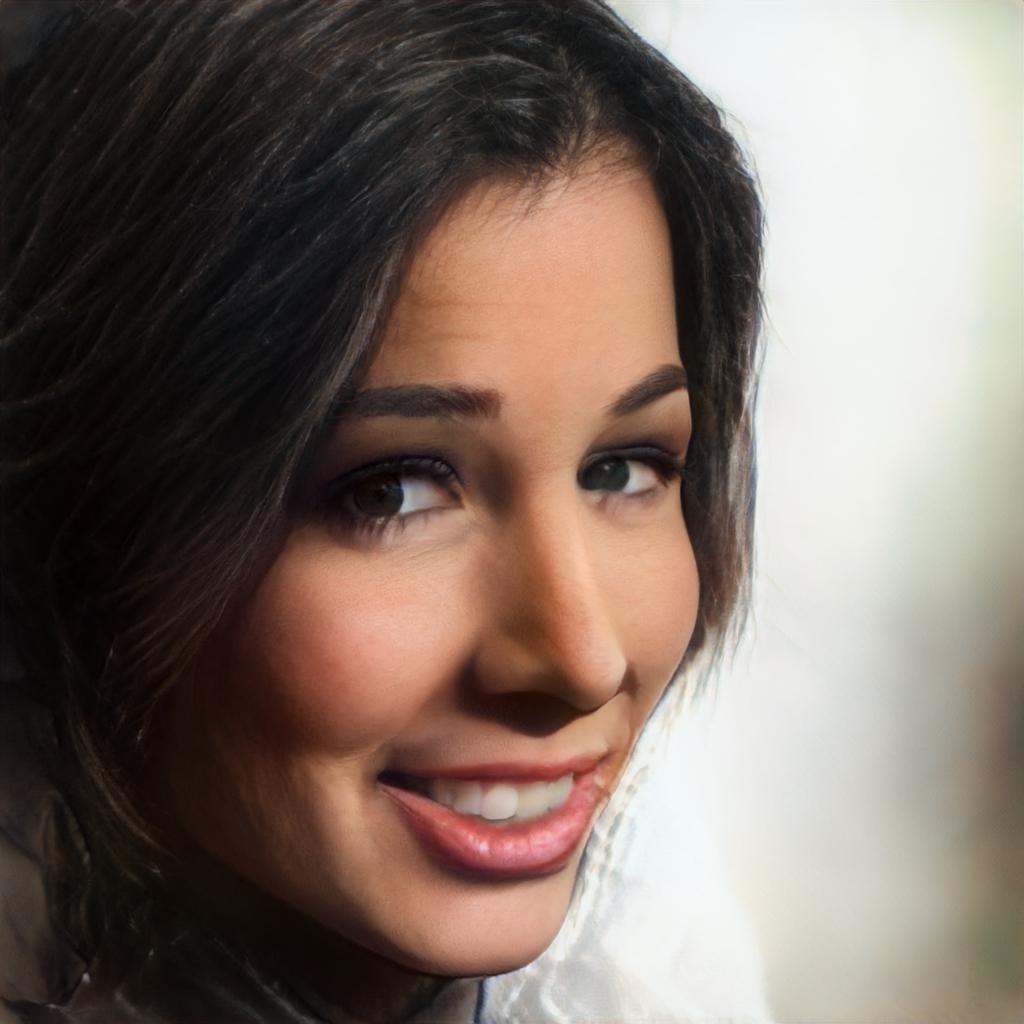} &
              & \raisebox{0.3in}{\rotatebox[origin=t]{90}{LIFE}} & 
                \includegraphics[width=0.10\textwidth]{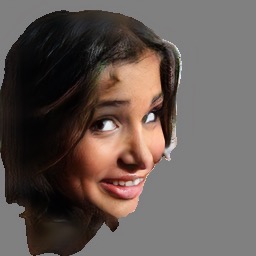} &
                \includegraphics[width=0.10\textwidth]{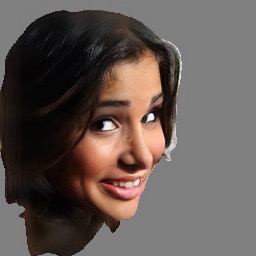} &
                \includegraphics[width=0.10\textwidth]{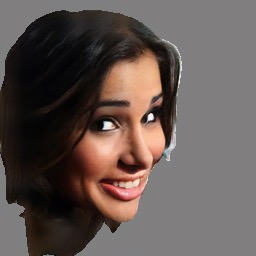} &
                \includegraphics[width=0.10\textwidth]{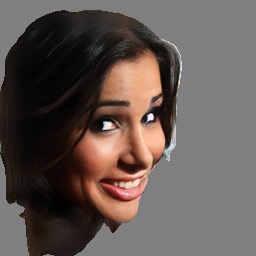} &
                \includegraphics[width=0.10\textwidth]{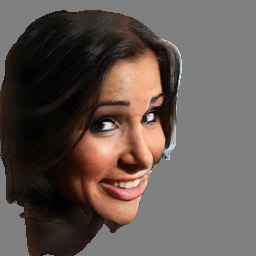} \\
              & & \raisebox{0.3in}{\rotatebox[origin=t]{90}{SAM}} &
                \includegraphics[width=0.10\textwidth]{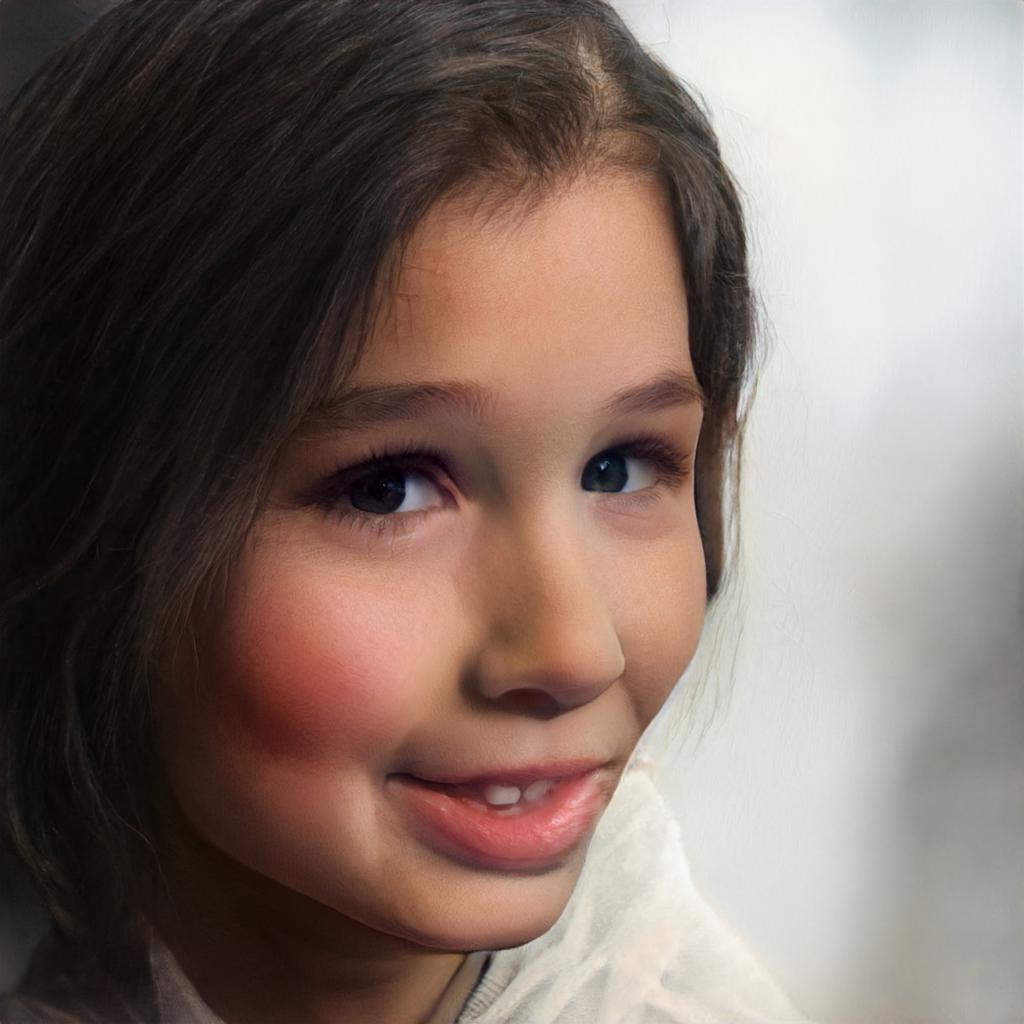} &
                \includegraphics[width=0.10\textwidth]{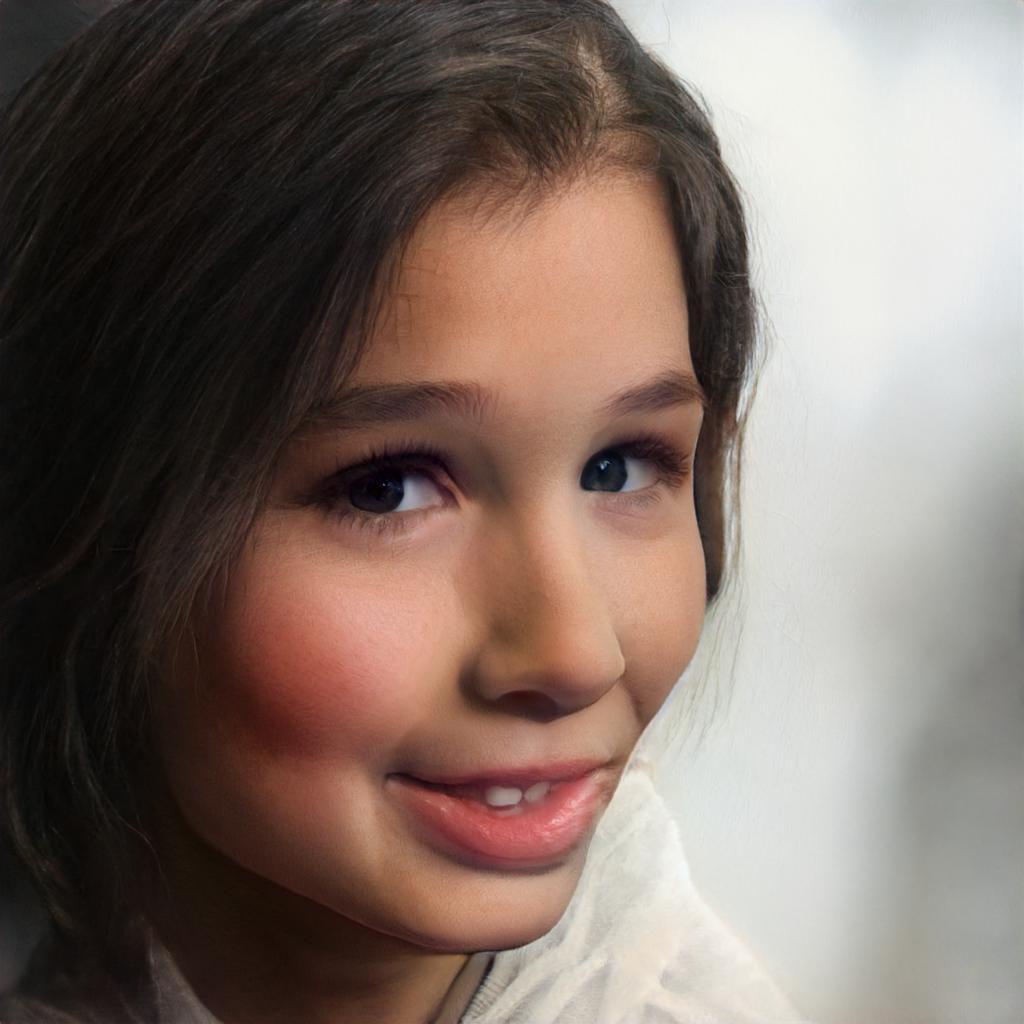} &
                \includegraphics[width=0.10\textwidth]{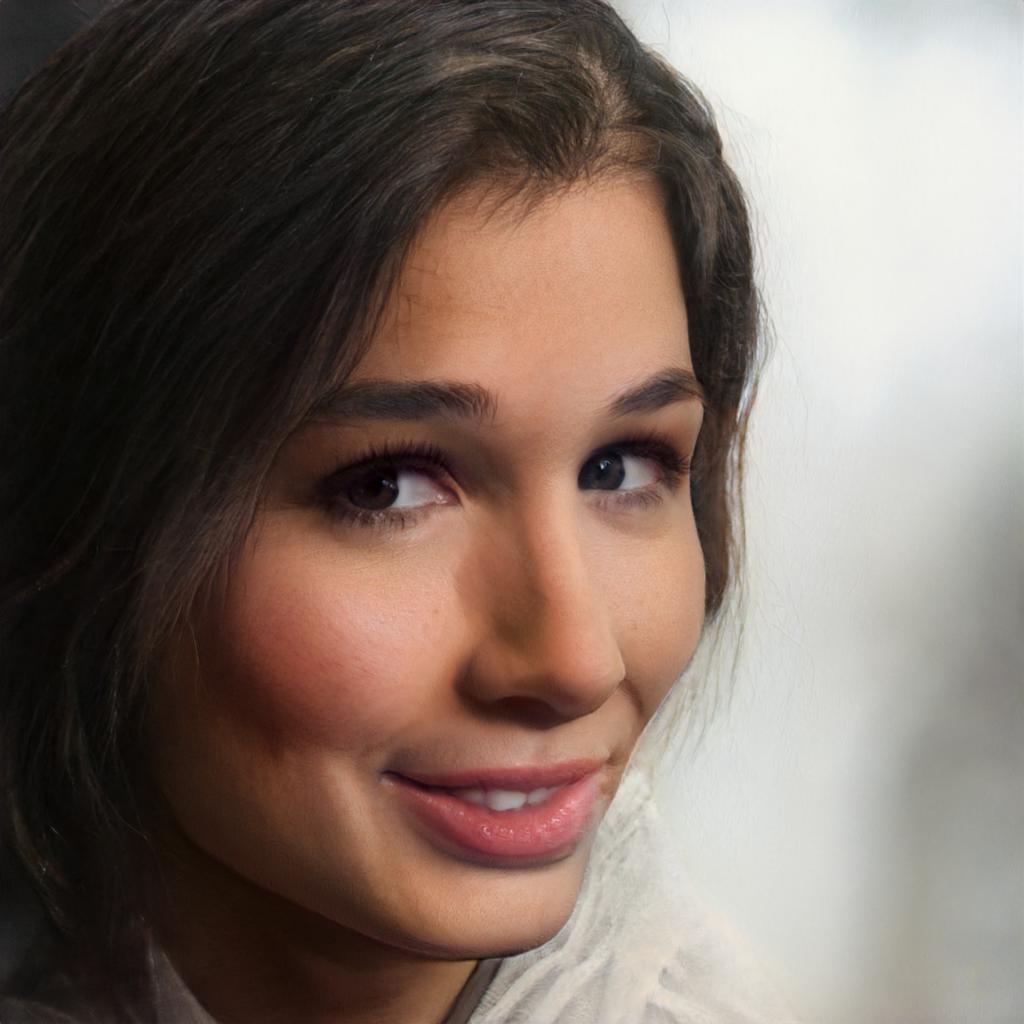} &
                \includegraphics[width=0.10\textwidth]{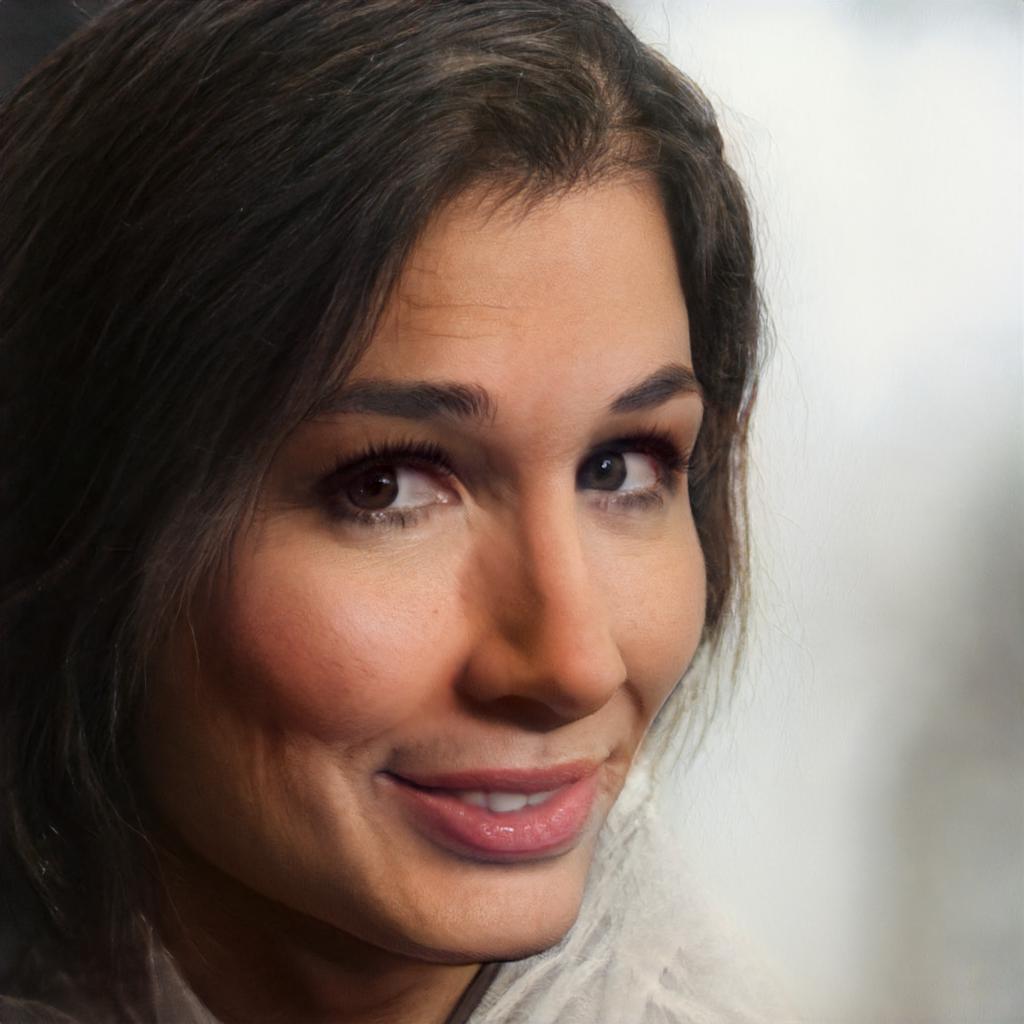} &
                \includegraphics[width=0.10\textwidth]{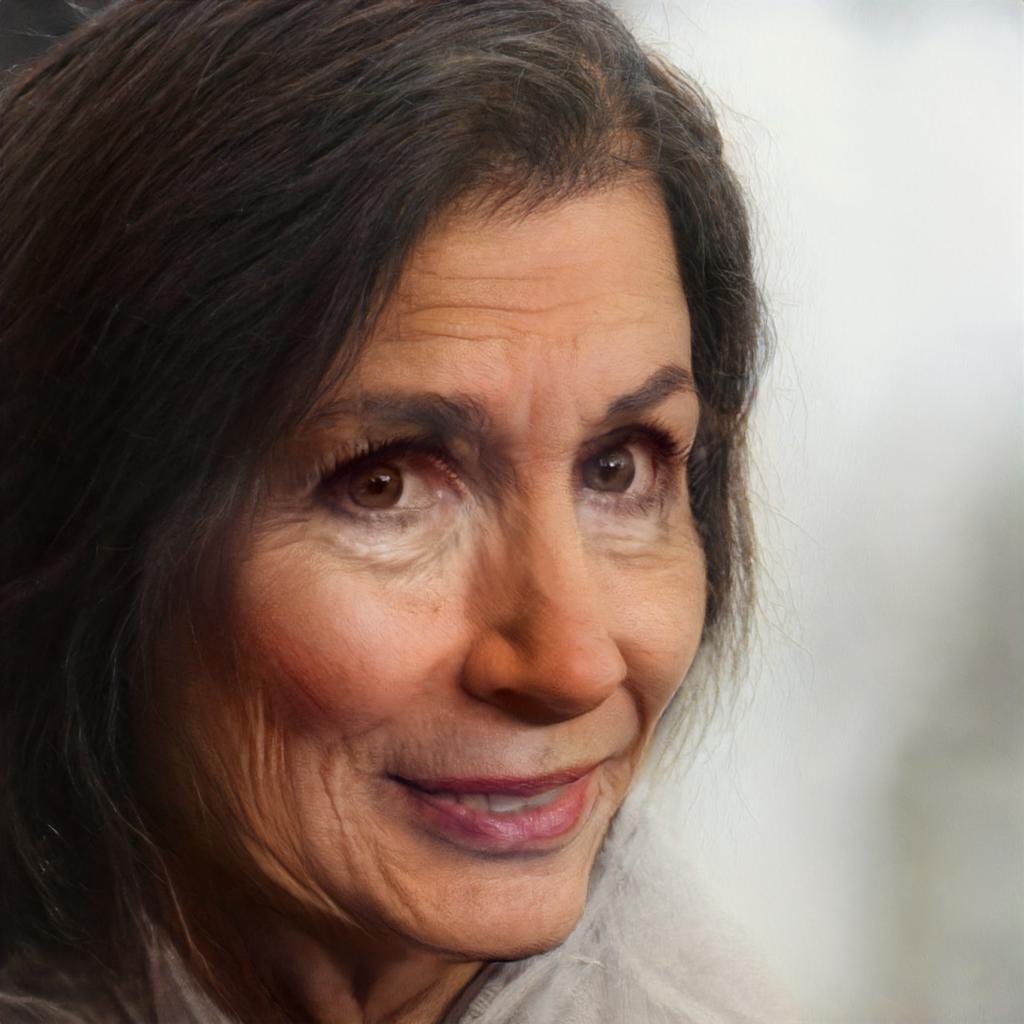}
        \tabularnewline
        
        \includegraphics[width=0.10\textwidth]{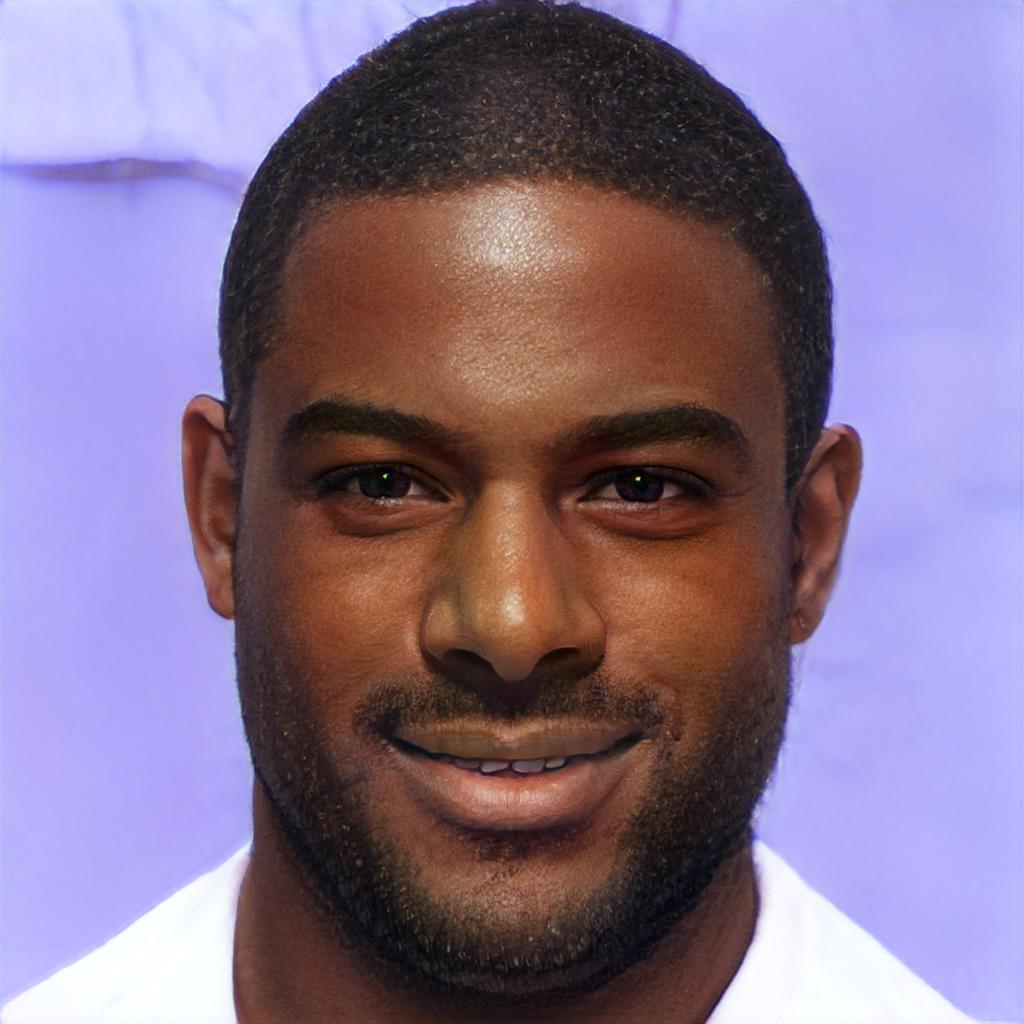} &
              & \raisebox{0.3in}{\rotatebox[origin=t]{90}{LIFE}} & 
                \includegraphics[width=0.10\textwidth]{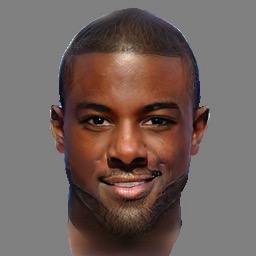} &
                \includegraphics[width=0.10\textwidth]{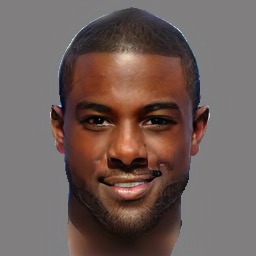} &
                \includegraphics[width=0.10\textwidth]{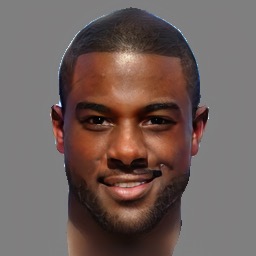} &
                \includegraphics[width=0.10\textwidth]{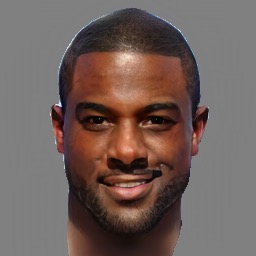} &
                \includegraphics[width=0.10\textwidth]{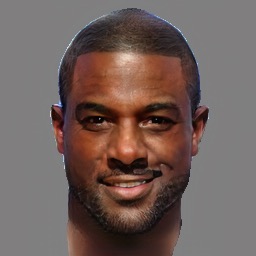} \\
              & & \raisebox{0.3in}{\rotatebox[origin=t]{90}{SAM}} &
                \includegraphics[width=0.10\textwidth]{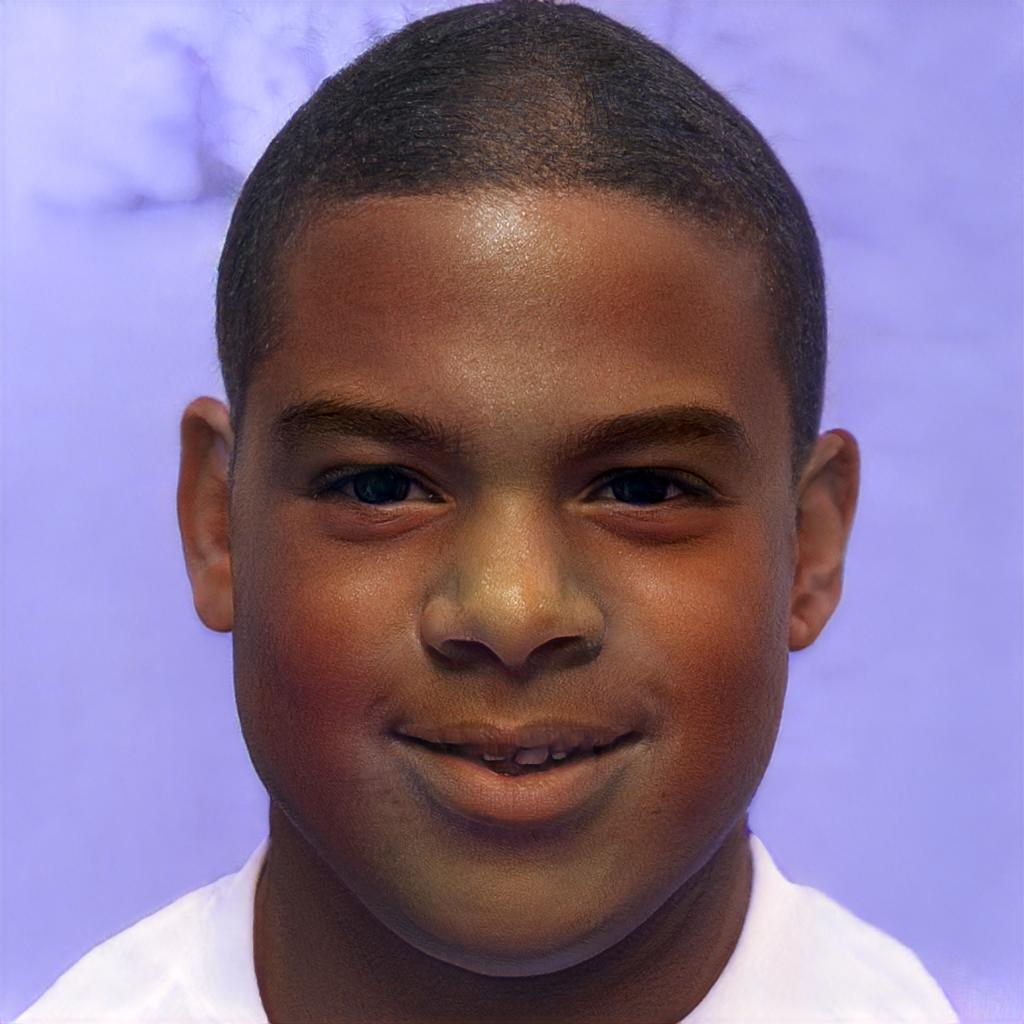} &
                \includegraphics[width=0.10\textwidth]{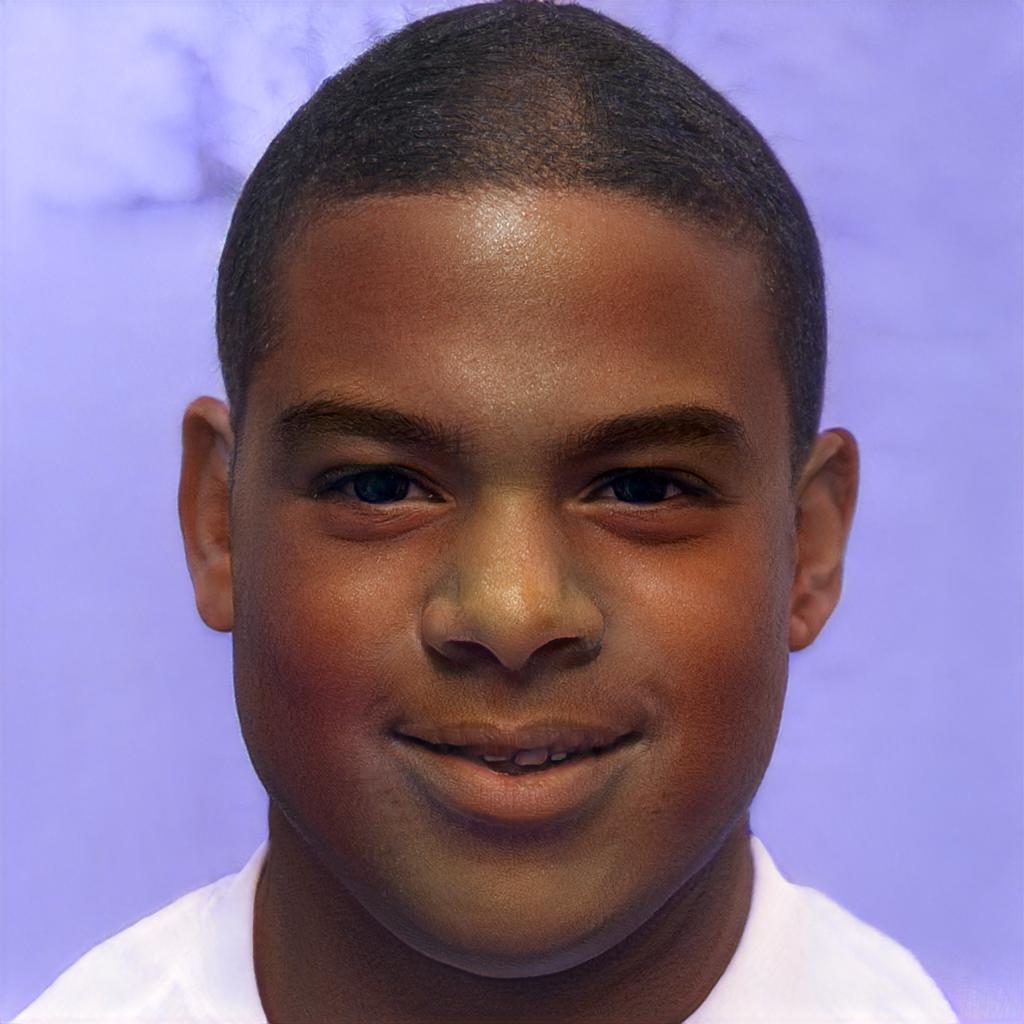} &
                \includegraphics[width=0.10\textwidth]{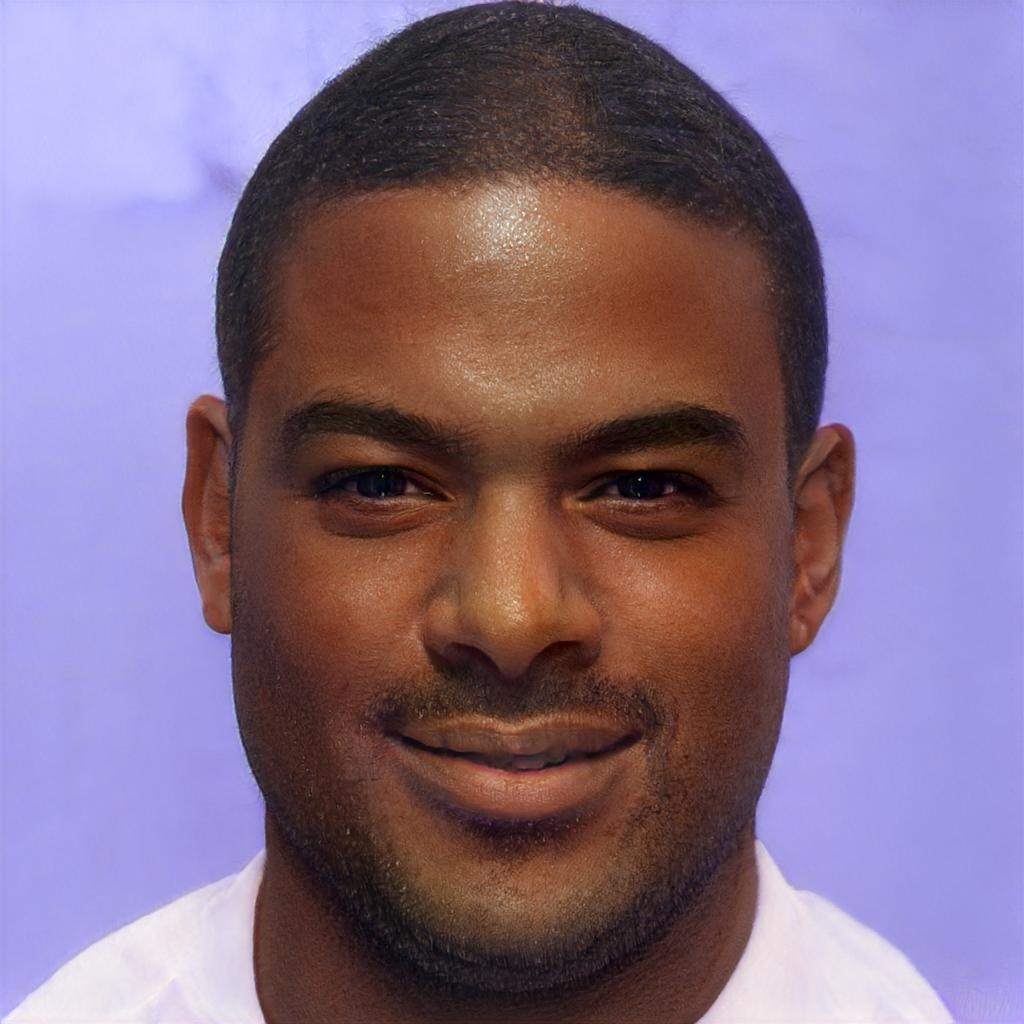} &
                \includegraphics[width=0.10\textwidth]{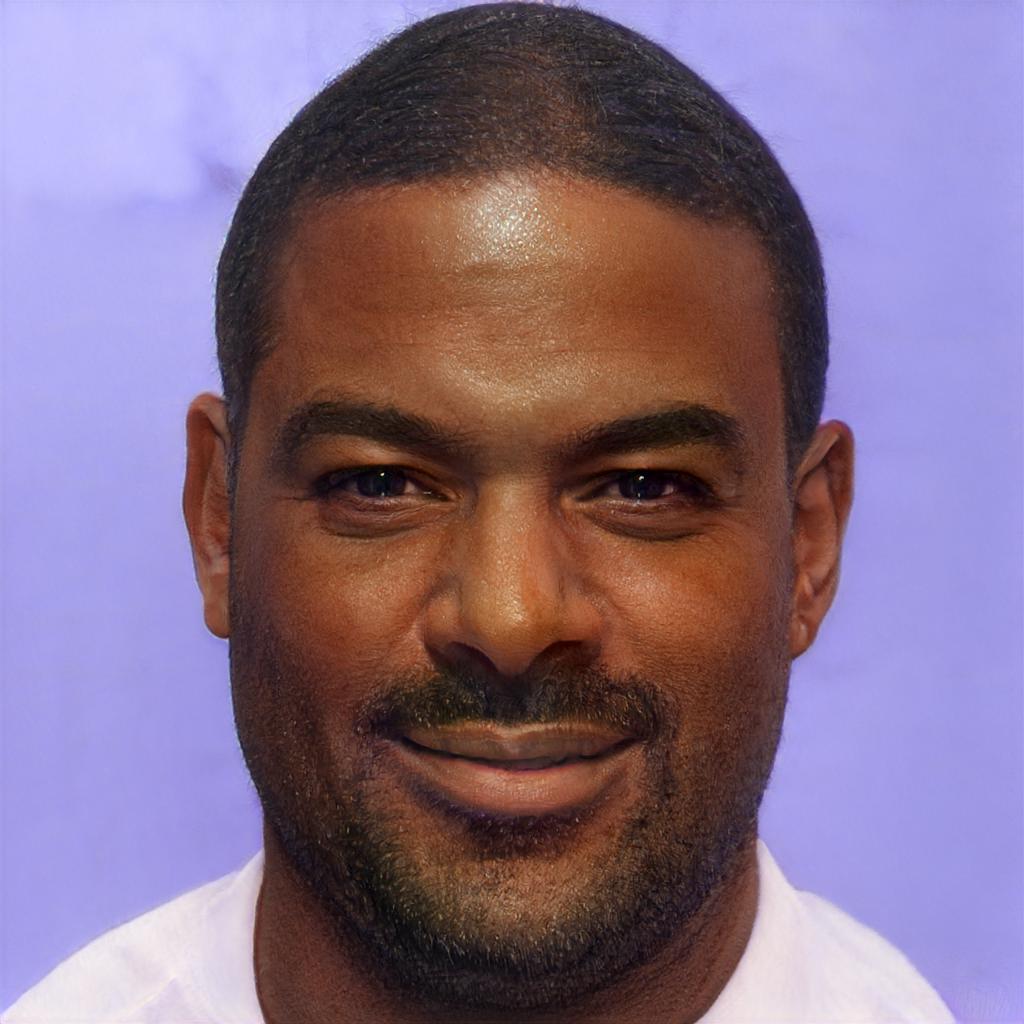} &
                \includegraphics[width=0.10\textwidth]{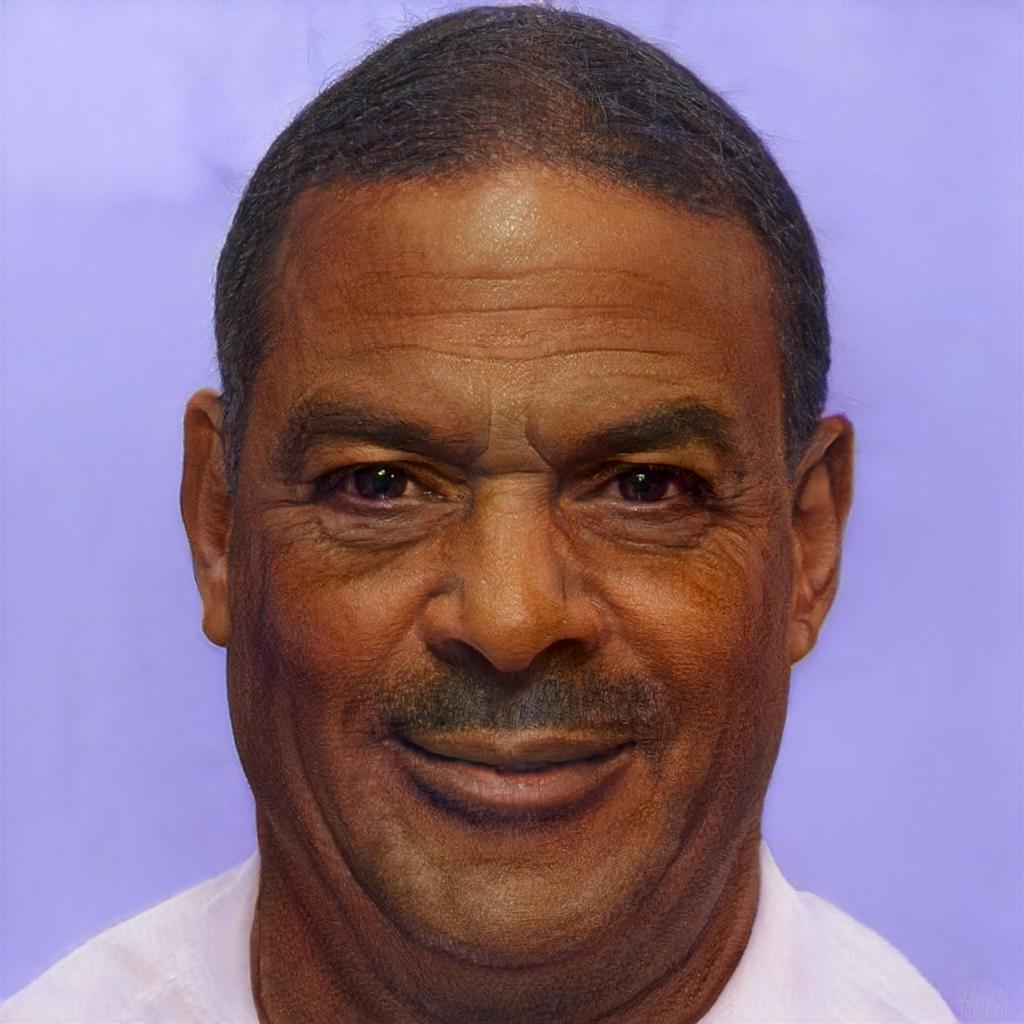}
        \tabularnewline
        
        \includegraphics[width=0.10\textwidth]{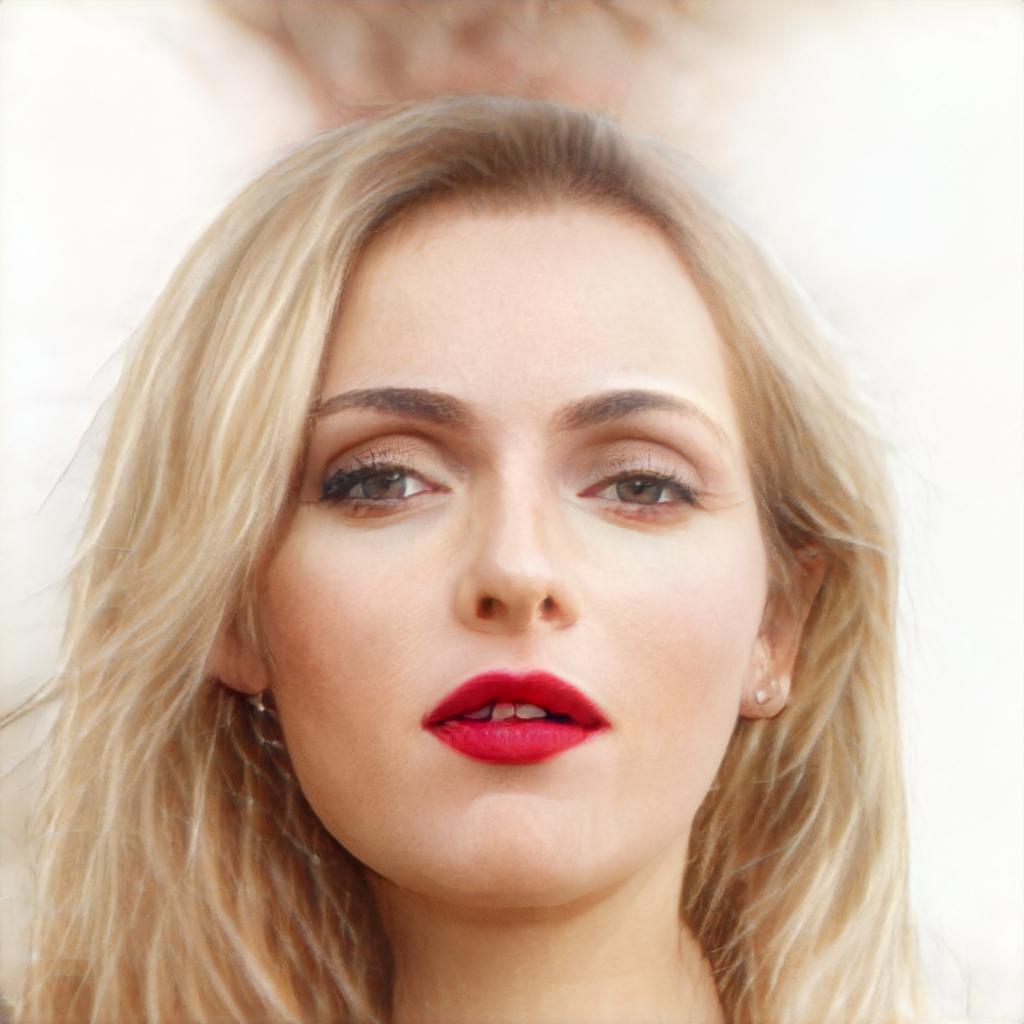} &
              & \raisebox{0.3in}{\rotatebox[origin=t]{90}{LIFE}} & 
                \includegraphics[width=0.10\textwidth]{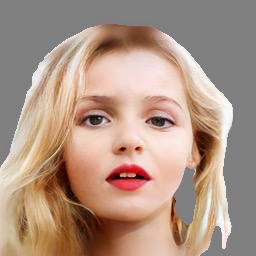} &
                \includegraphics[width=0.10\textwidth]{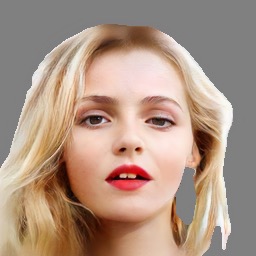} &
                \includegraphics[width=0.10\textwidth]{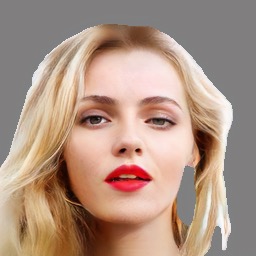} &
                \includegraphics[width=0.10\textwidth]{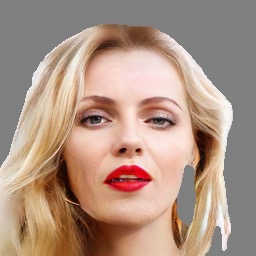} &
                \includegraphics[width=0.10\textwidth]{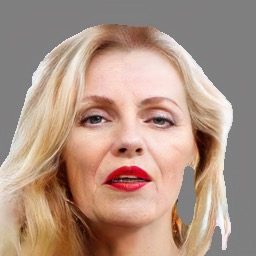} \\
              & & \raisebox{0.3in}{\rotatebox[origin=t]{90}{SAM}} &
                \includegraphics[width=0.10\textwidth]{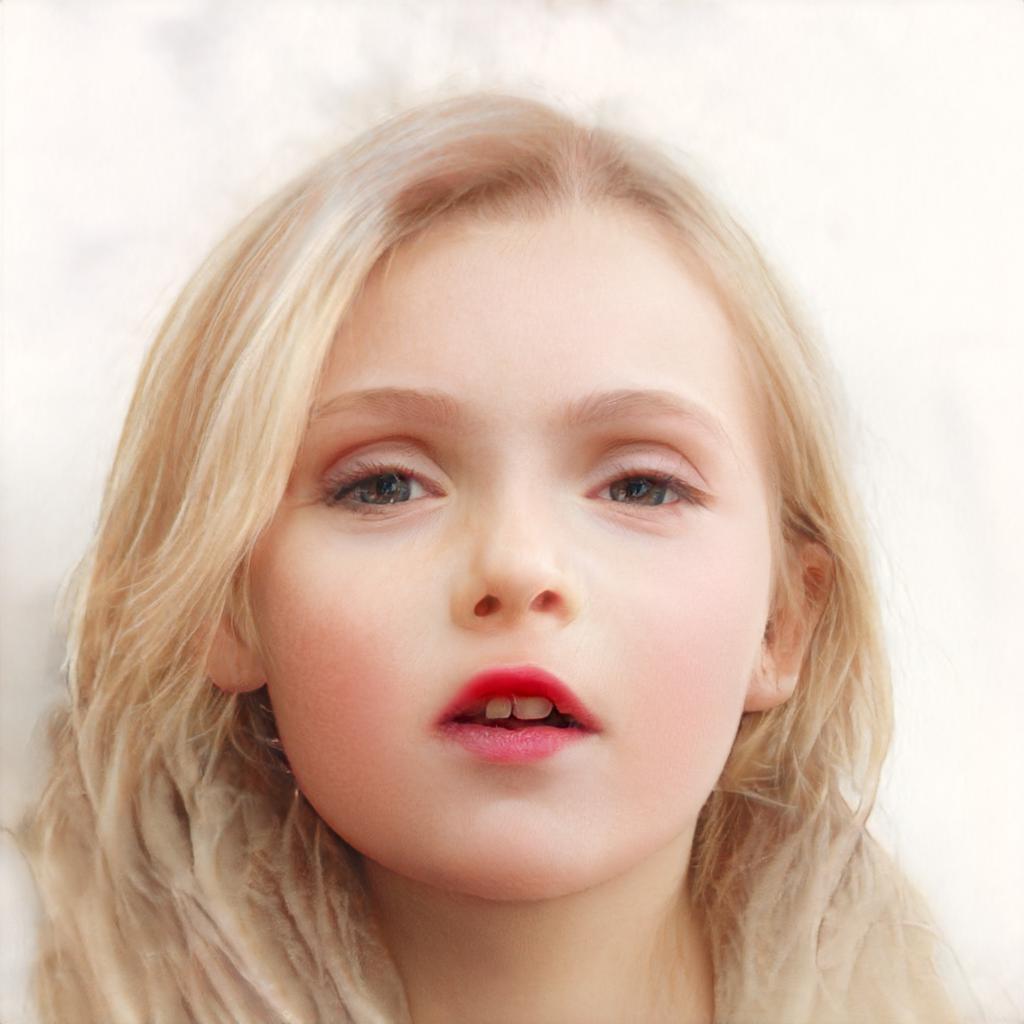} &
                \includegraphics[width=0.10\textwidth]{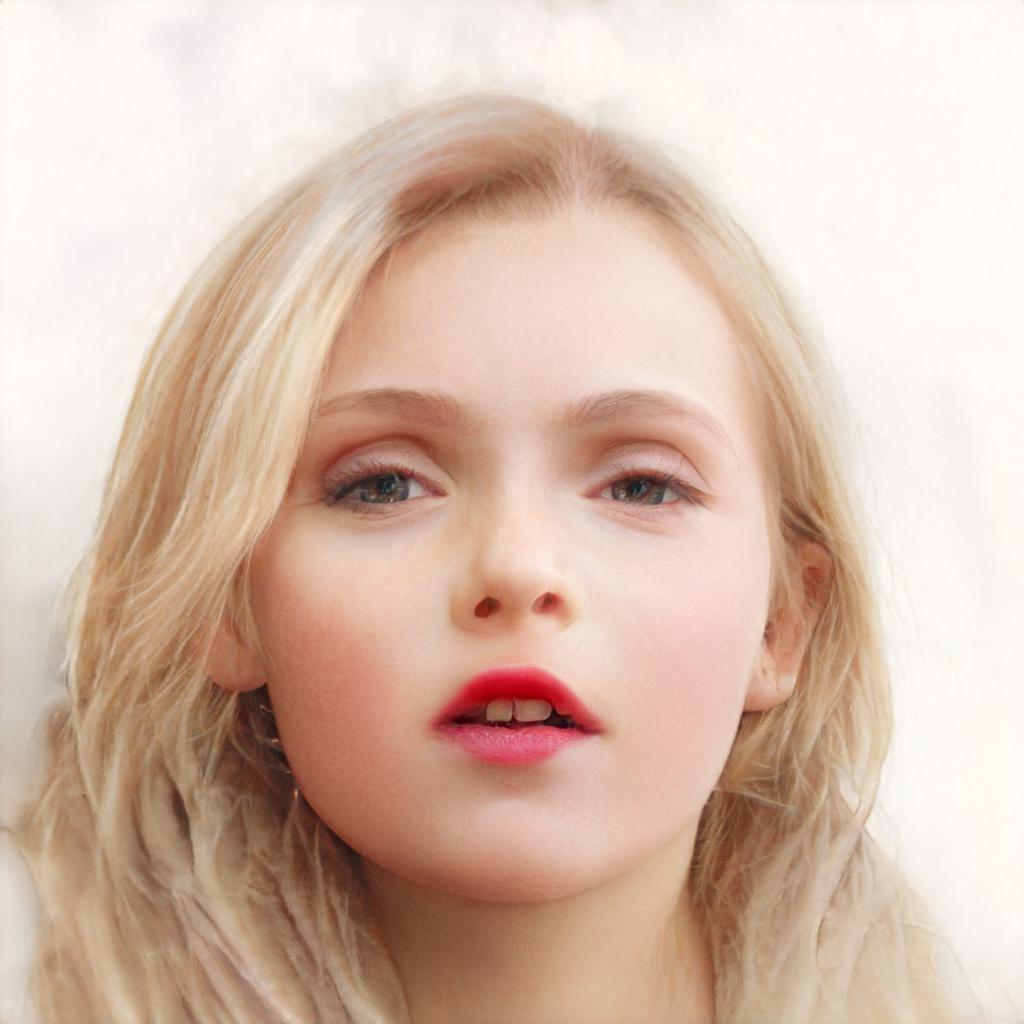} &
                \includegraphics[width=0.10\textwidth]{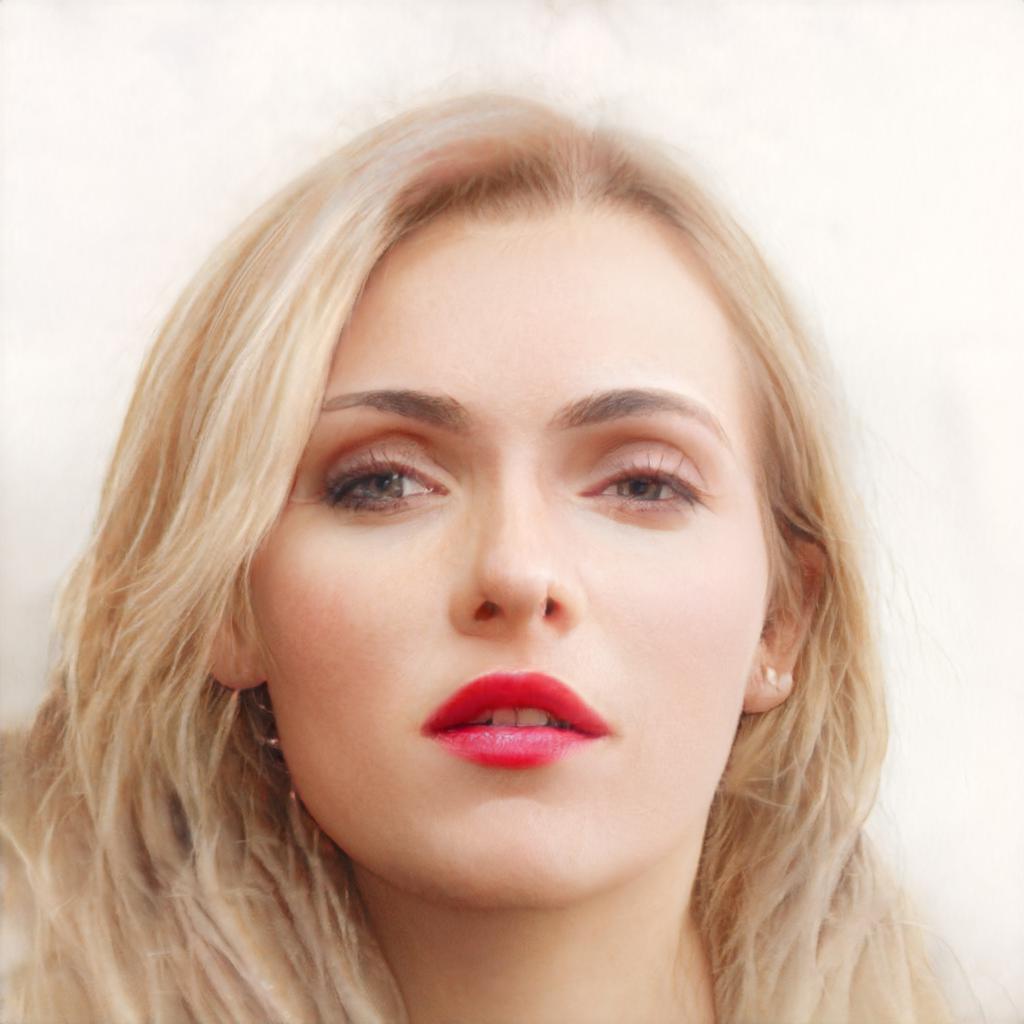} &
                \includegraphics[width=0.10\textwidth]{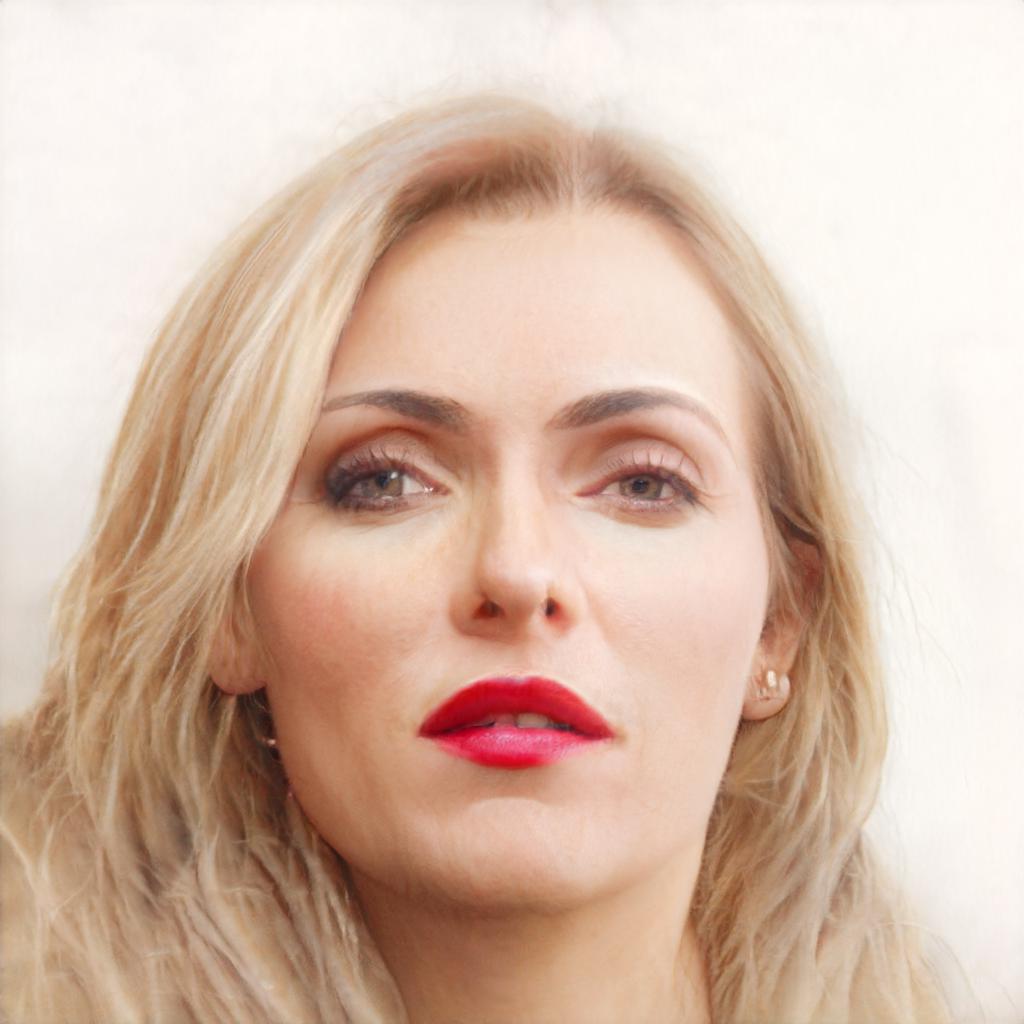} &
                \includegraphics[width=0.10\textwidth]{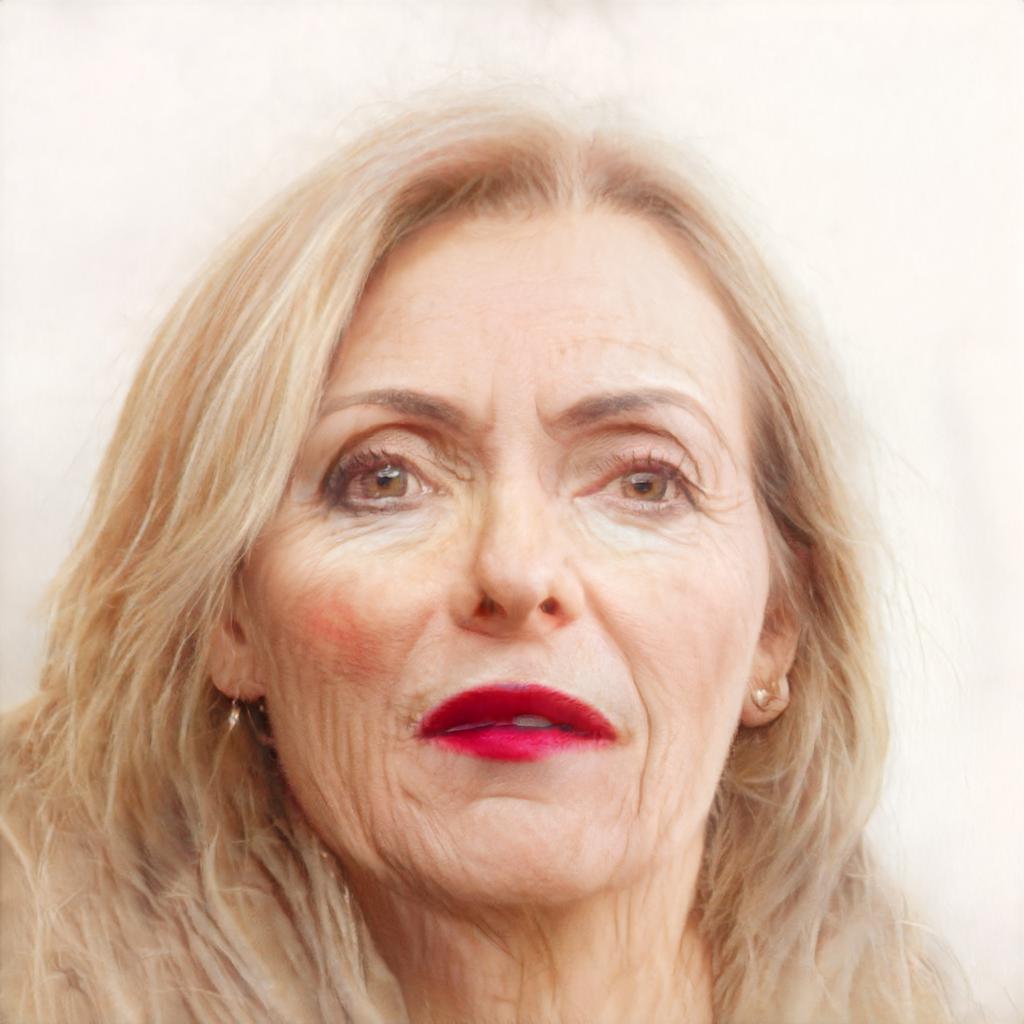}
        \tabularnewline
        
        \includegraphics[width=0.10\textwidth]{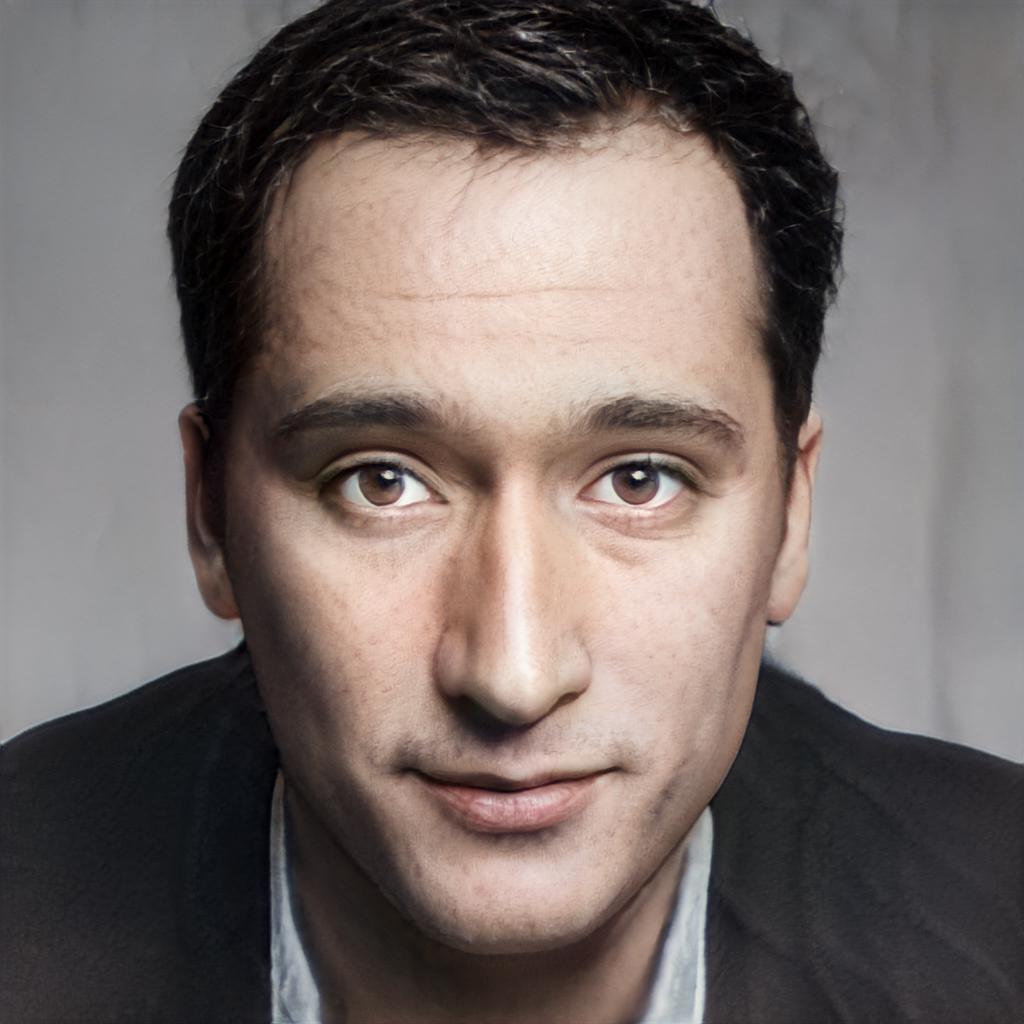} &
              & \raisebox{0.3in}{\rotatebox[origin=t]{90}{LIFE}} & 
                \includegraphics[width=0.10\textwidth]{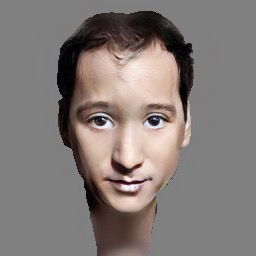} &
                \includegraphics[width=0.10\textwidth]{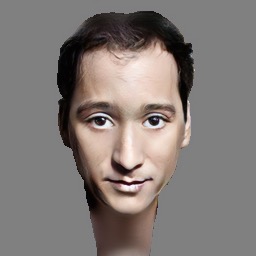} &
                \includegraphics[width=0.10\textwidth]{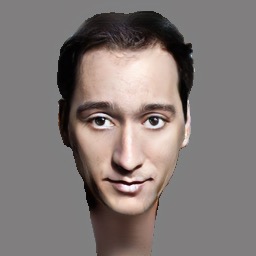} &
                \includegraphics[width=0.10\textwidth]{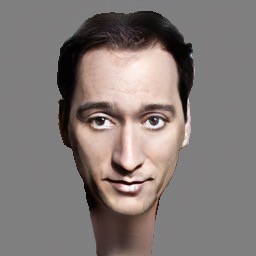} &
                \includegraphics[width=0.10\textwidth]{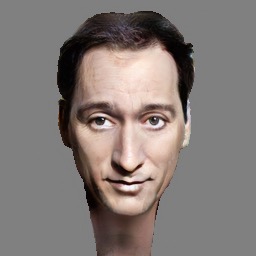} \\
              & & \raisebox{0.3in}{\rotatebox[origin=t]{90}{SAM}} &
                \includegraphics[width=0.10\textwidth]{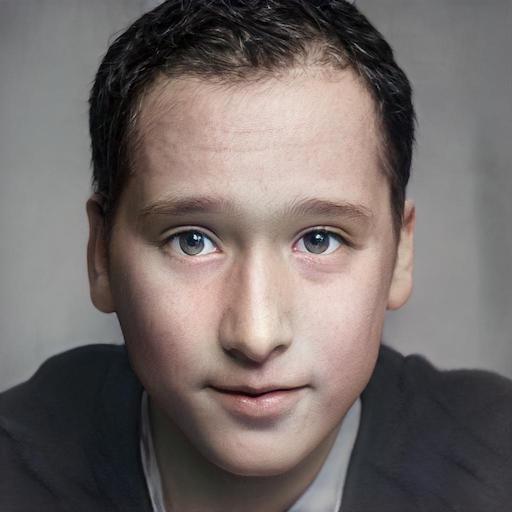} &
                \includegraphics[width=0.10\textwidth]{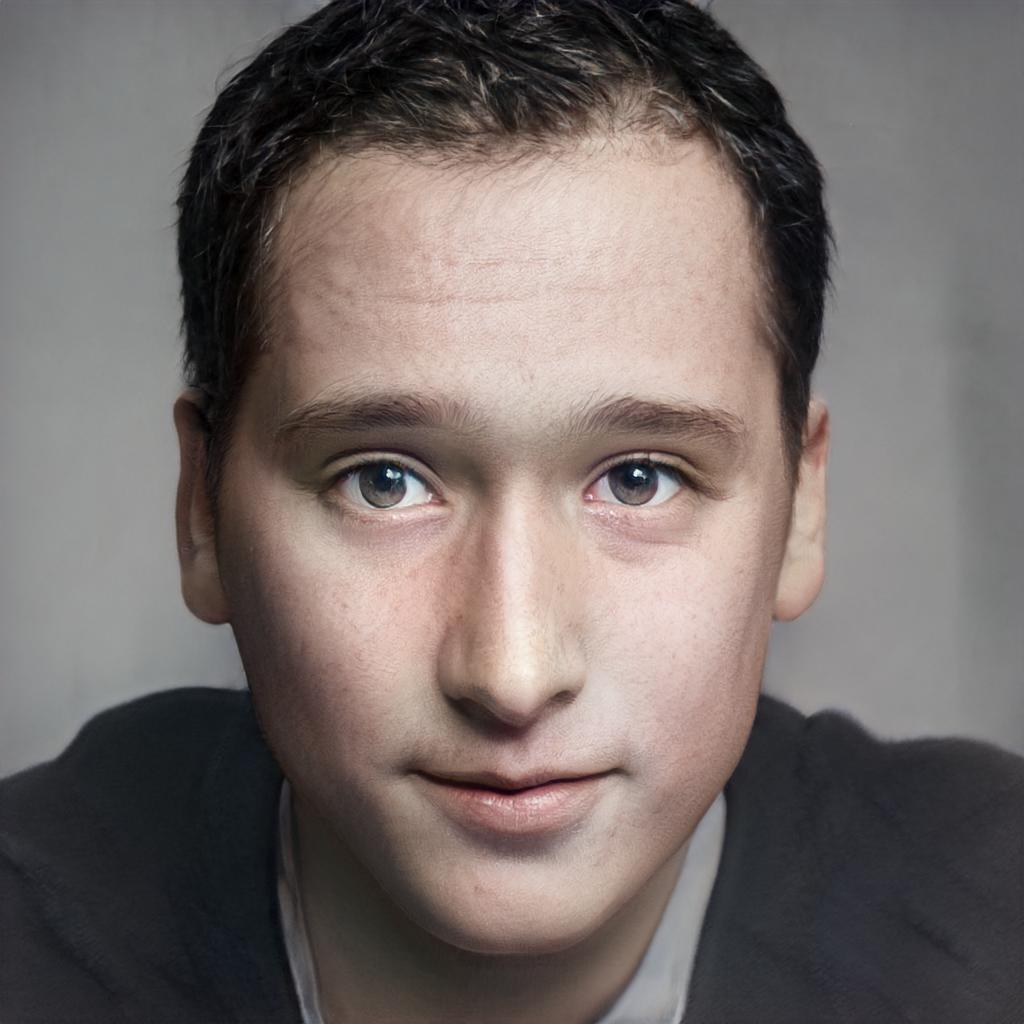} &
                \includegraphics[width=0.10\textwidth]{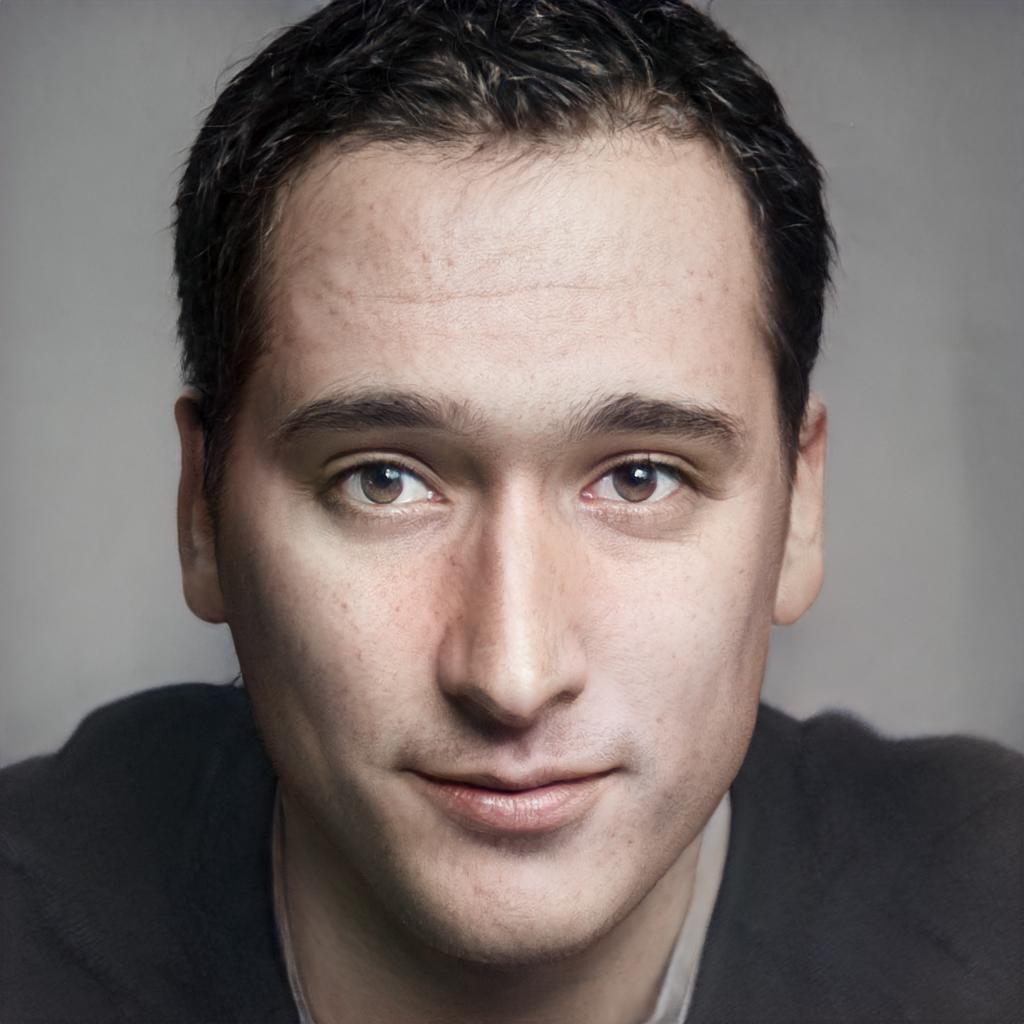} &
                \includegraphics[width=0.10\textwidth]{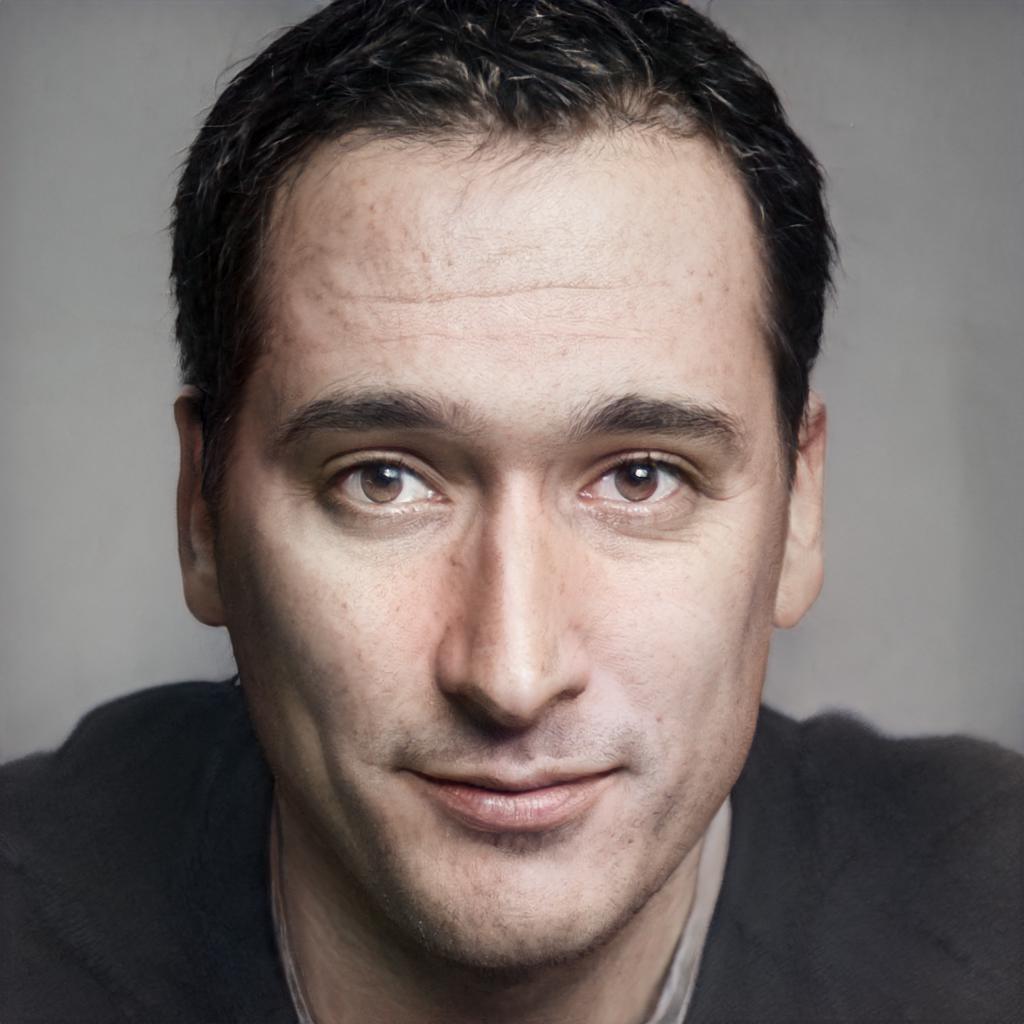} &
                \includegraphics[width=0.10\textwidth]{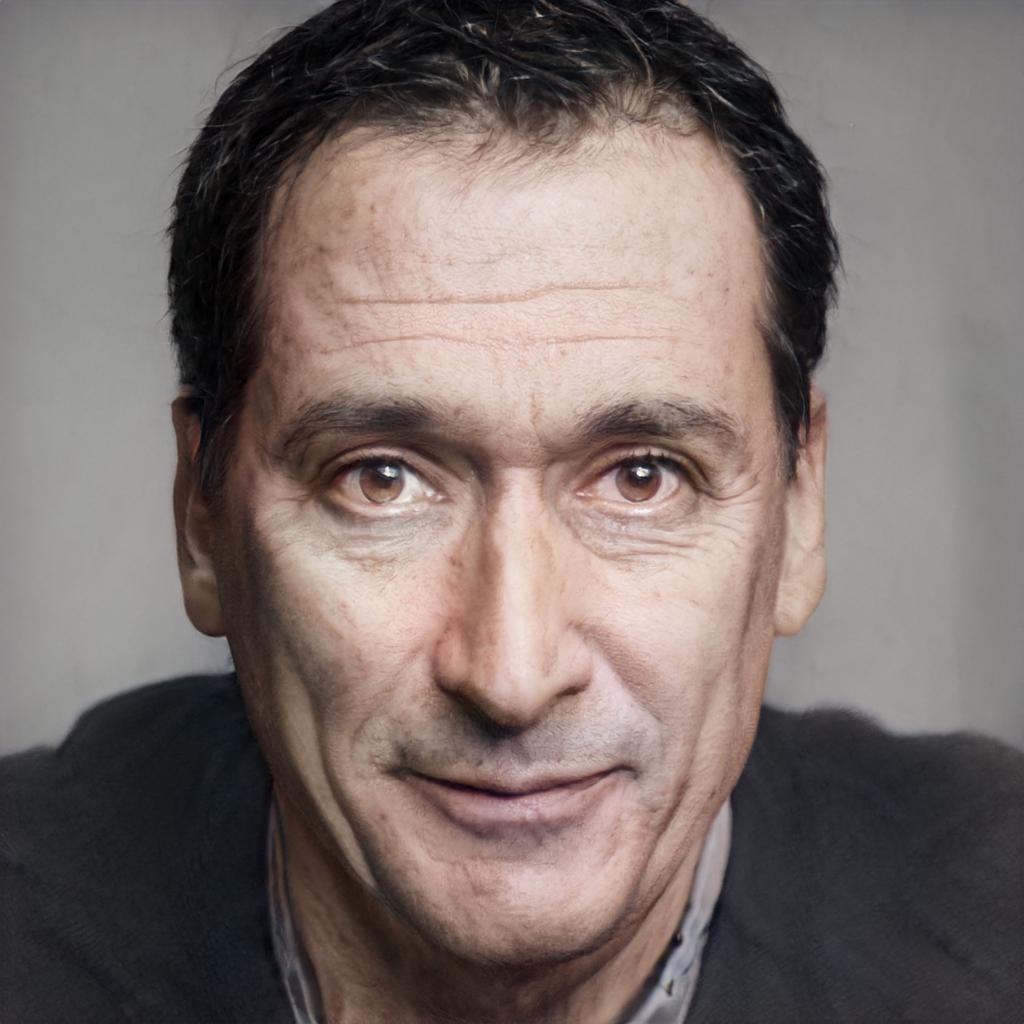}
        \tabularnewline

        \includegraphics[width=0.10\textwidth]{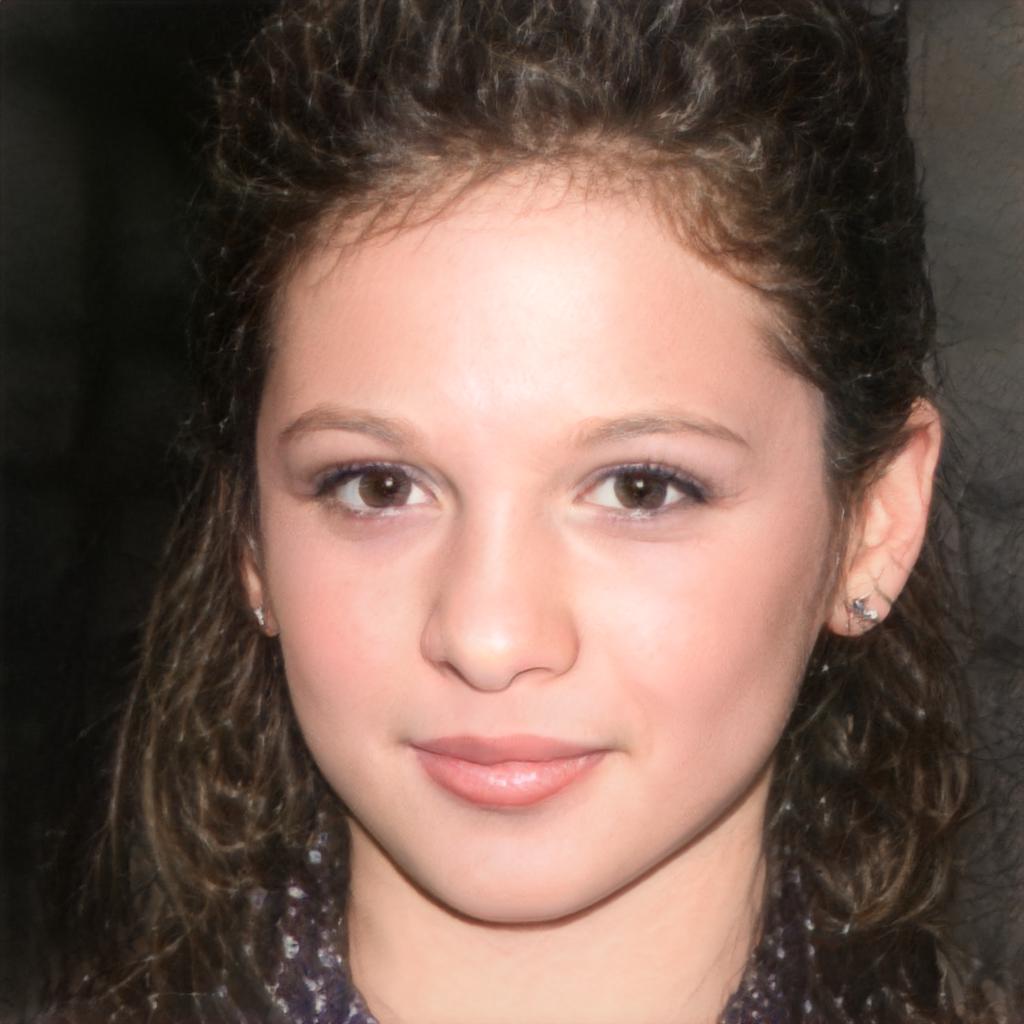} &
              & \raisebox{0.3in}{\rotatebox[origin=t]{90}{LIFE}} & 
                \includegraphics[width=0.10\textwidth]{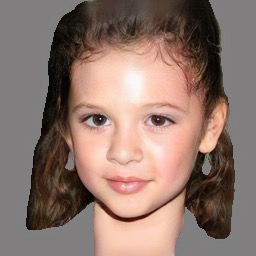} &
                \includegraphics[width=0.10\textwidth]{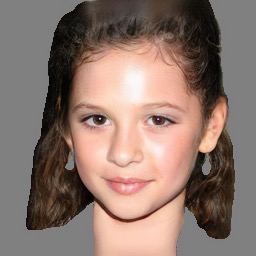} &
                \includegraphics[width=0.10\textwidth]{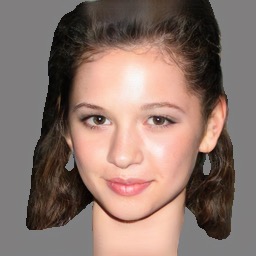} &
                \includegraphics[width=0.10\textwidth]{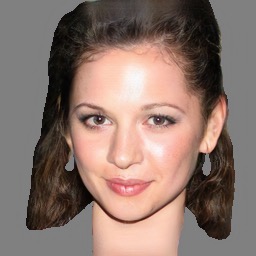} &
                \includegraphics[width=0.10\textwidth]{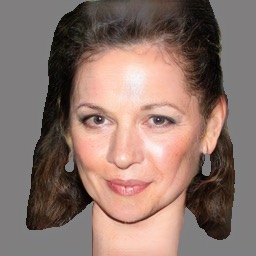} \\
              & & \raisebox{0.3in}{\rotatebox[origin=t]{90}{SAM}} &
                \includegraphics[width=0.10\textwidth]{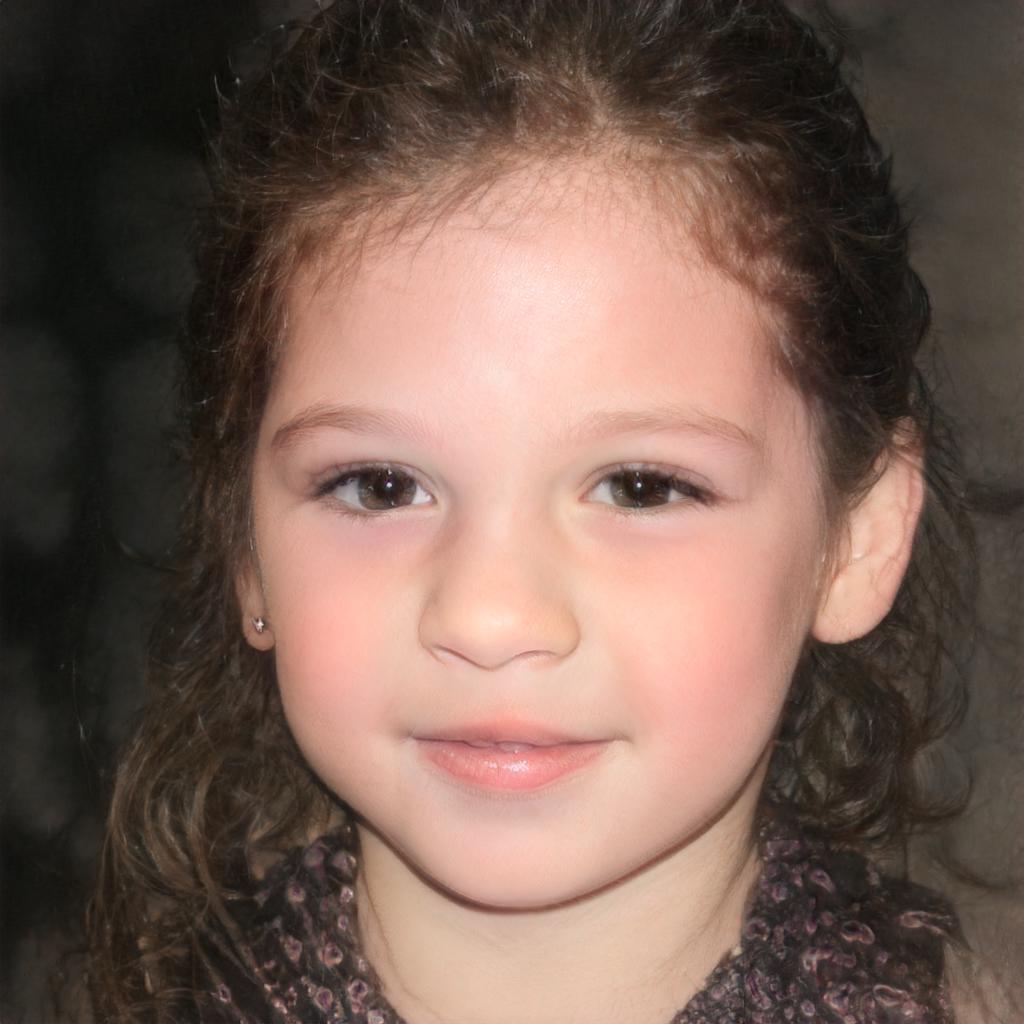} &
                \includegraphics[width=0.10\textwidth]{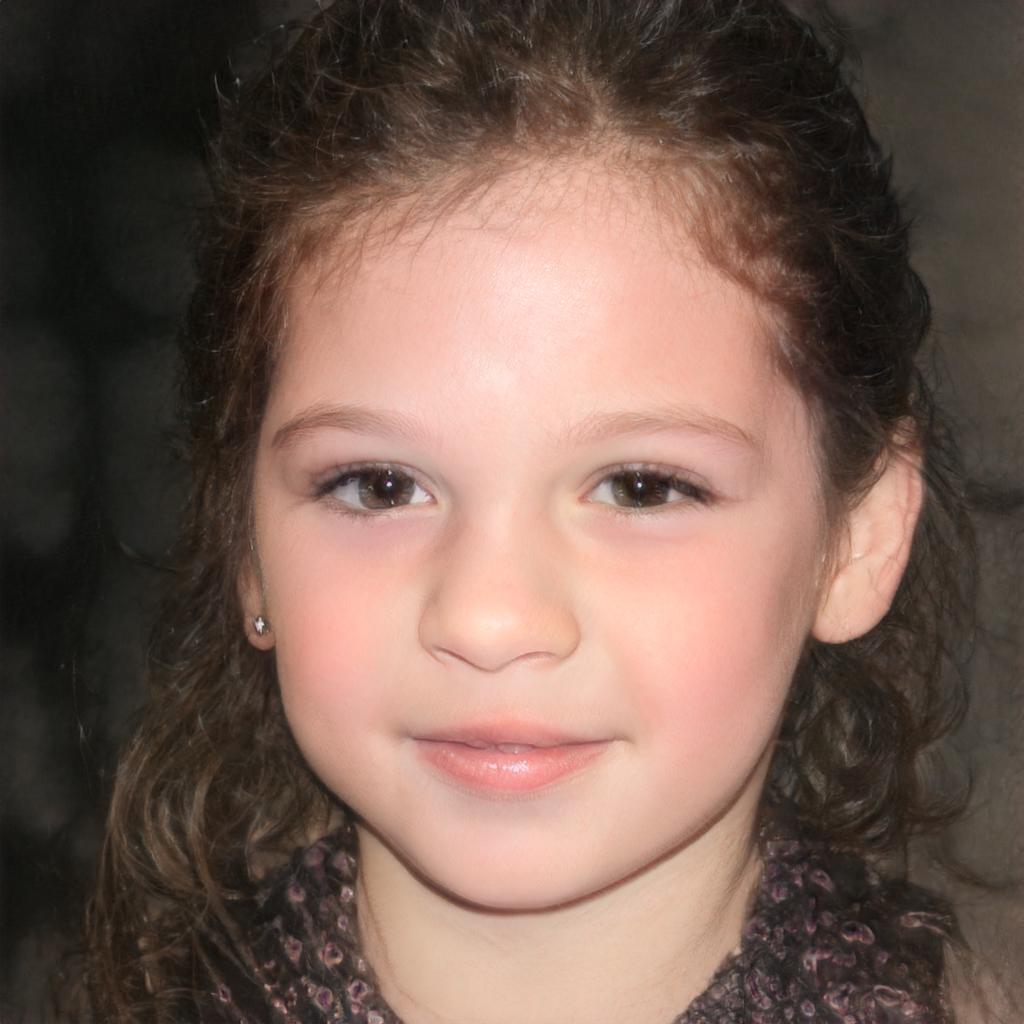} &
                \includegraphics[width=0.10\textwidth]{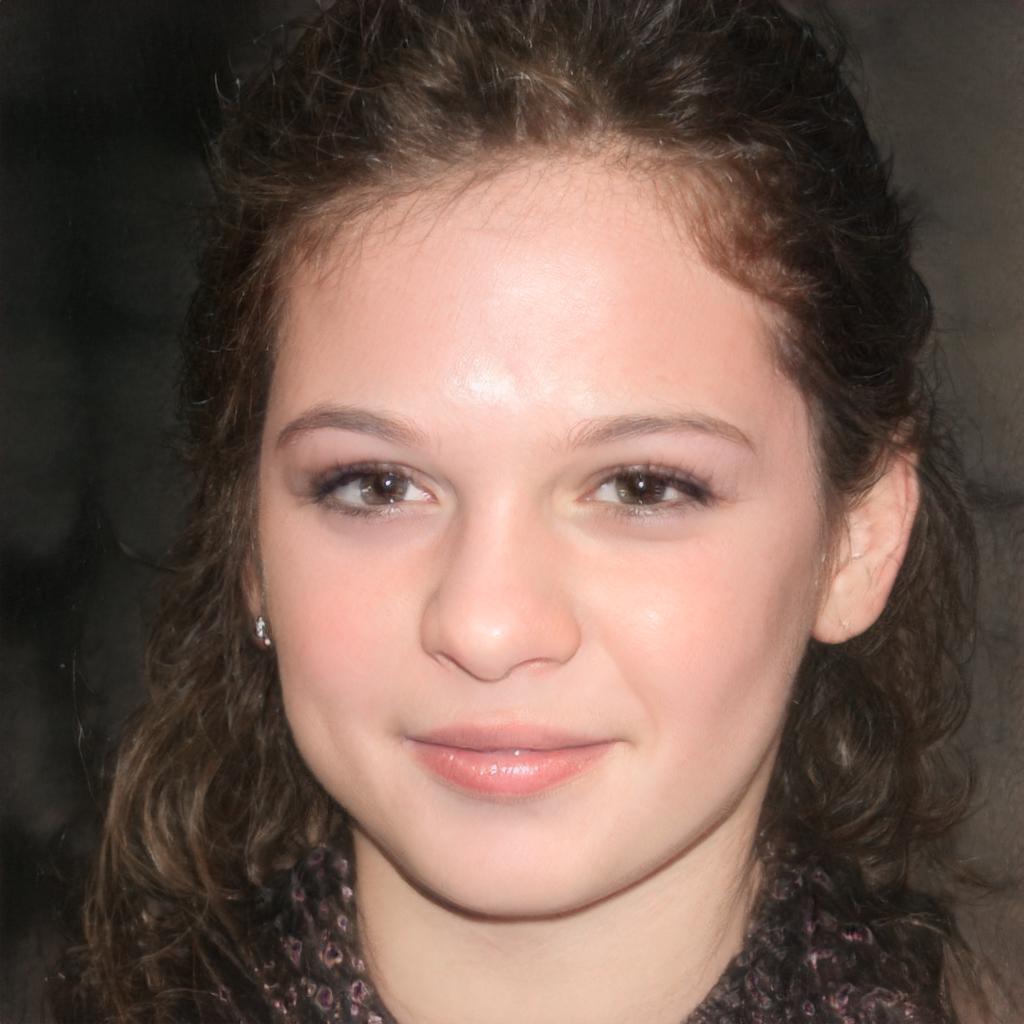} &
                \includegraphics[width=0.10\textwidth]{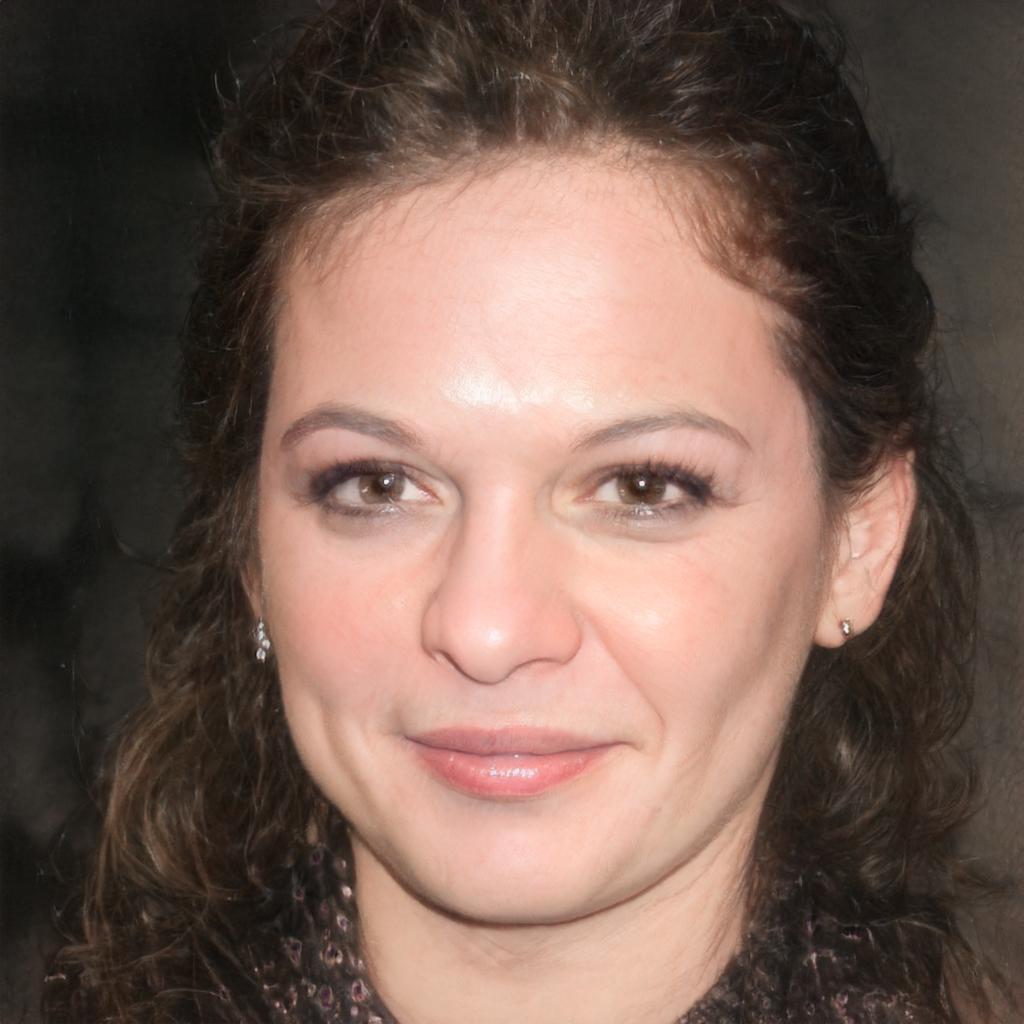} &
                \includegraphics[width=0.10\textwidth]{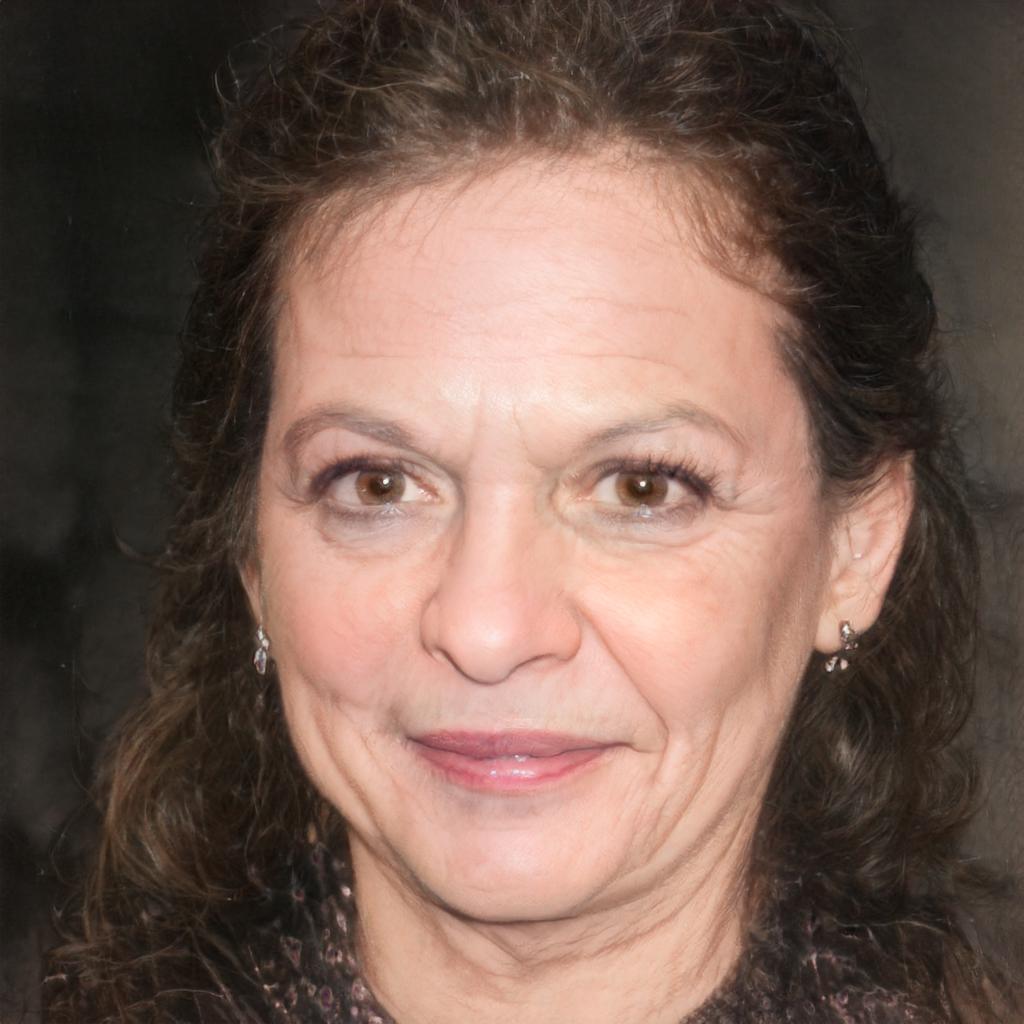}
        \tabularnewline

    \end{tabular}
    \label{fig:appendix_comparison_lifespan}
    \caption{Additional qualitative comparisons of age transformation results with  LIFE~\cite{orel2020lifespan} on the CelebA-HQ~\cite{karras2017progressive} test set. For translating our images to the age groups in~\cite{orel2020lifespan}, we set the target age equal to the the median age of each group.}
\end{figure*}

%% file: figures/appendix/appendix_hrfae_comparison.tex
\begin{figure*}
    \centering
    \setlength{\belowcaptionskip}{-2.5pt}
    \setlength{\tabcolsep}{1pt}
    \centering
        \begin{tabular}{c c c c c c c c c}
        Inversion & & & 25 & 35 & 45 & 55 & 65 \\
        \includegraphics[width=0.10\textwidth]{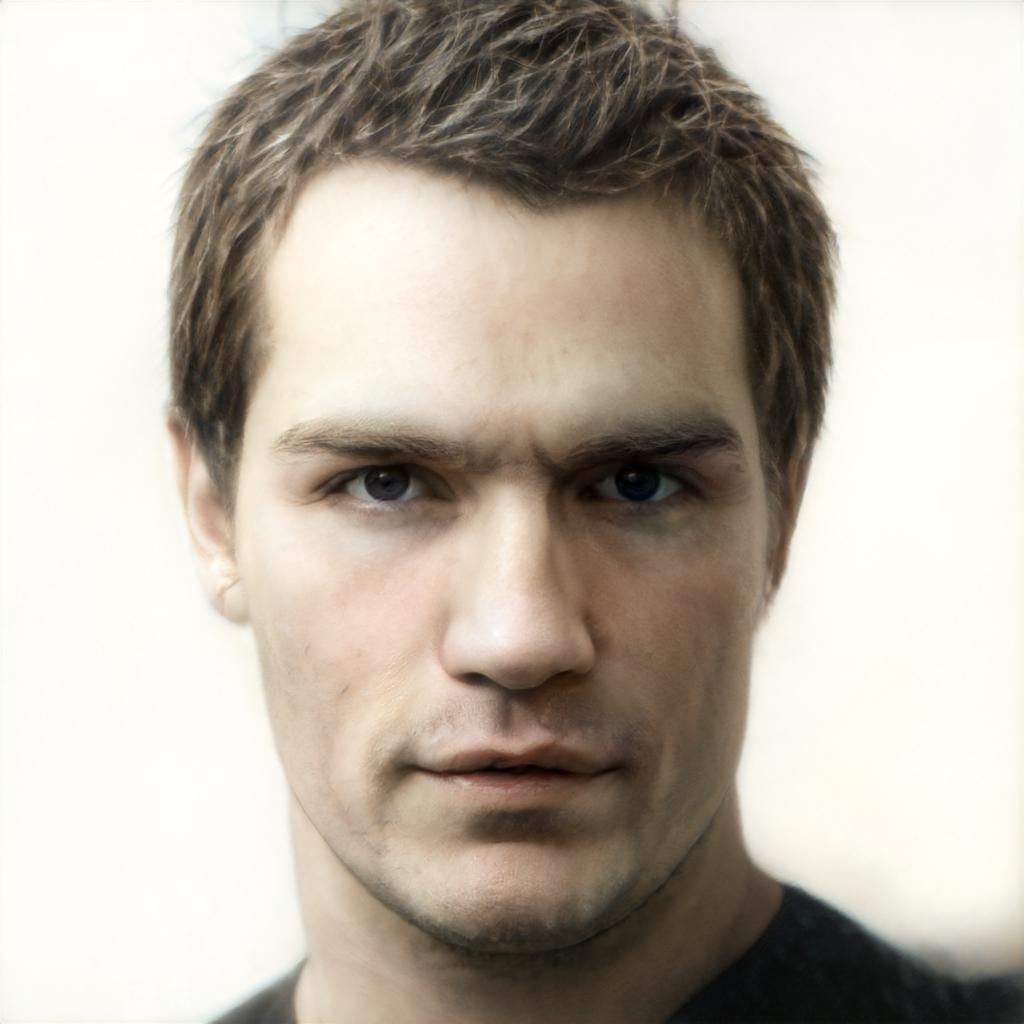} &
              & \raisebox{0.3in}{\rotatebox[origin=t]{90}{HRFAE}} & 
                \includegraphics[width=0.10\textwidth]{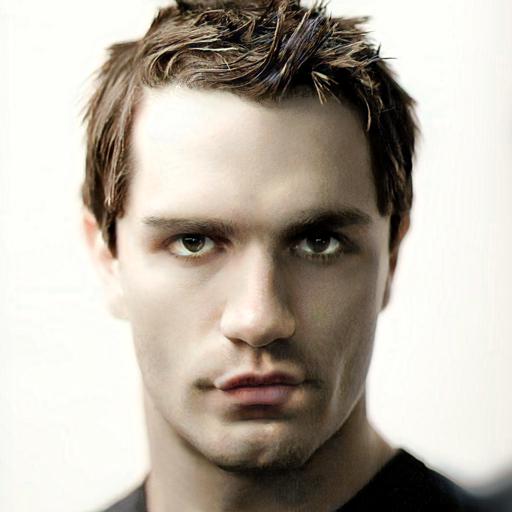} &
                \includegraphics[width=0.10\textwidth]{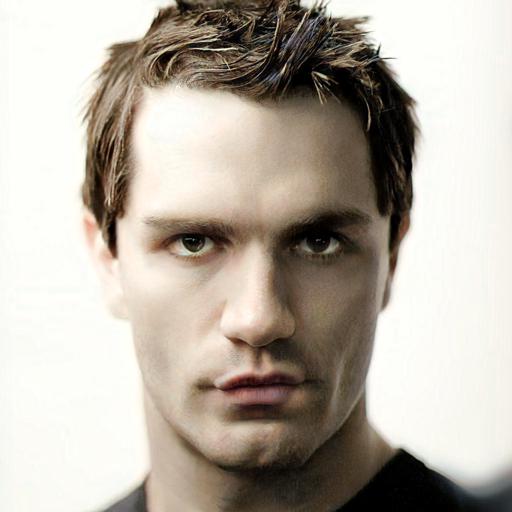} &
                \includegraphics[width=0.10\textwidth]{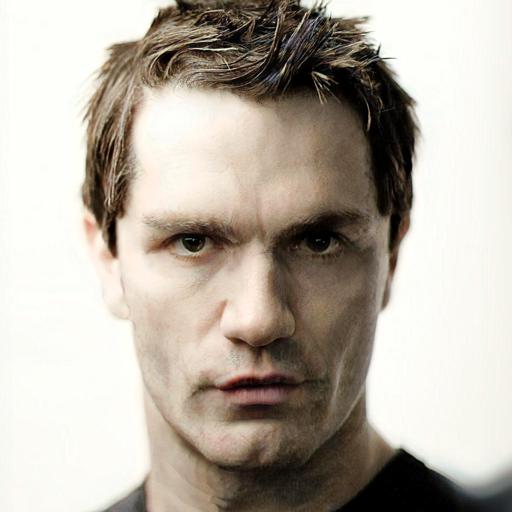} &
                \includegraphics[width=0.10\textwidth]{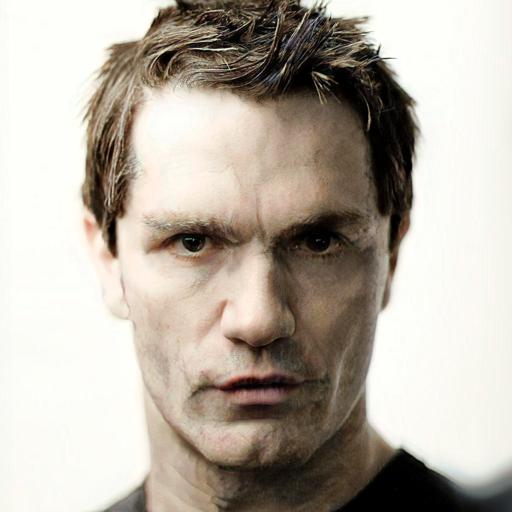} &
                \includegraphics[width=0.10\textwidth]{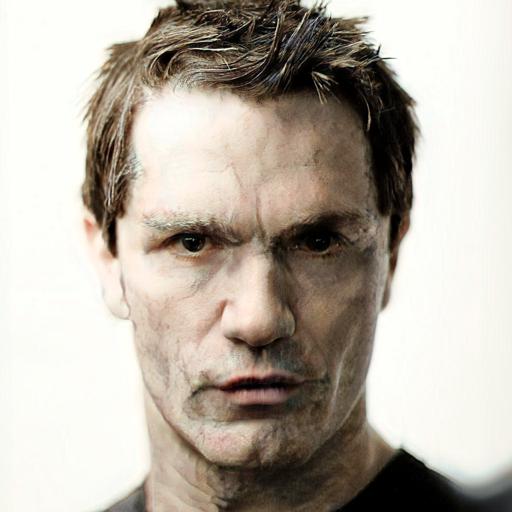} \\
              & & \raisebox{0.3in}{\rotatebox[origin=t]{90}{SAM}} &
                \includegraphics[width=0.10\textwidth]{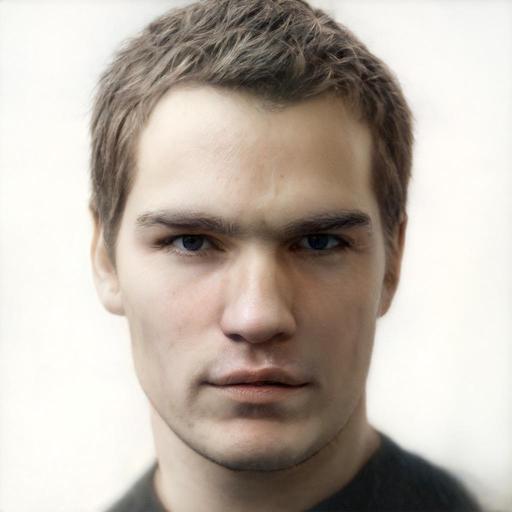} &
                \includegraphics[width=0.10\textwidth]{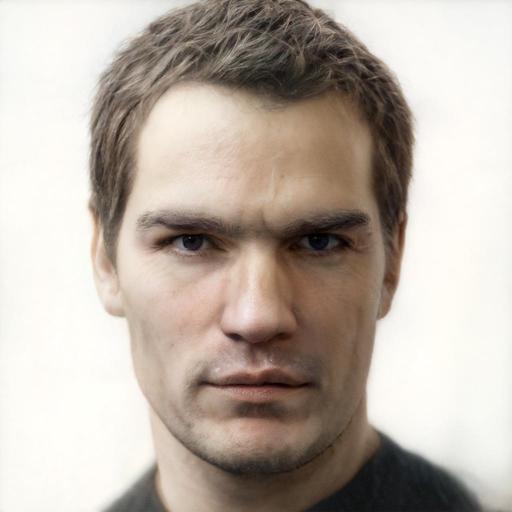} &
                \includegraphics[width=0.10\textwidth]{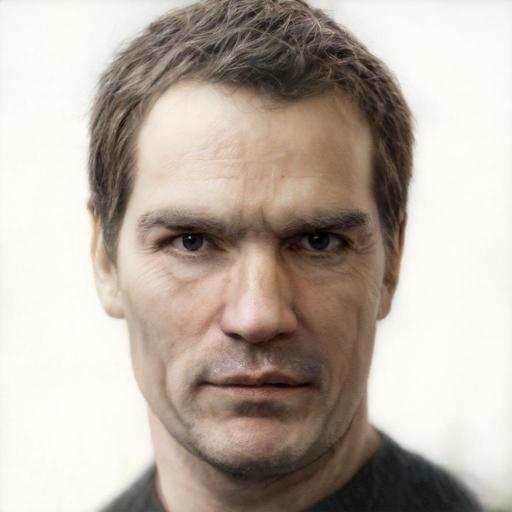} &
                \includegraphics[width=0.10\textwidth]{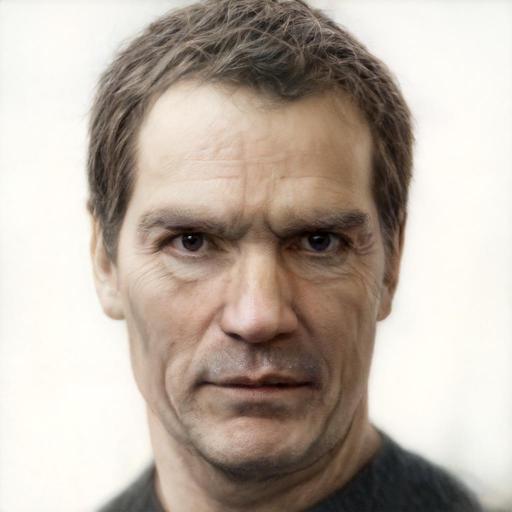} &
                \includegraphics[width=0.10\textwidth]{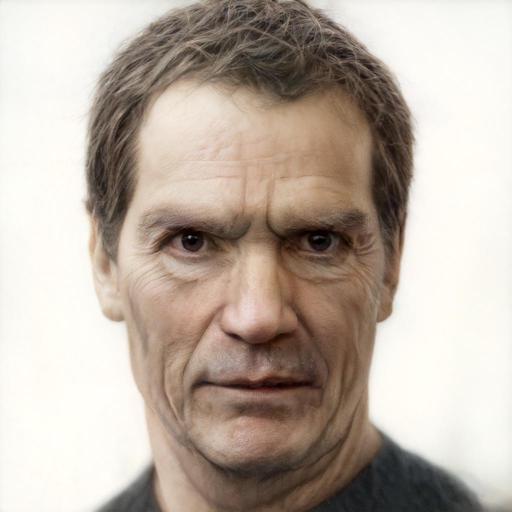}
        \tabularnewline
        
        \includegraphics[width=0.10\textwidth]{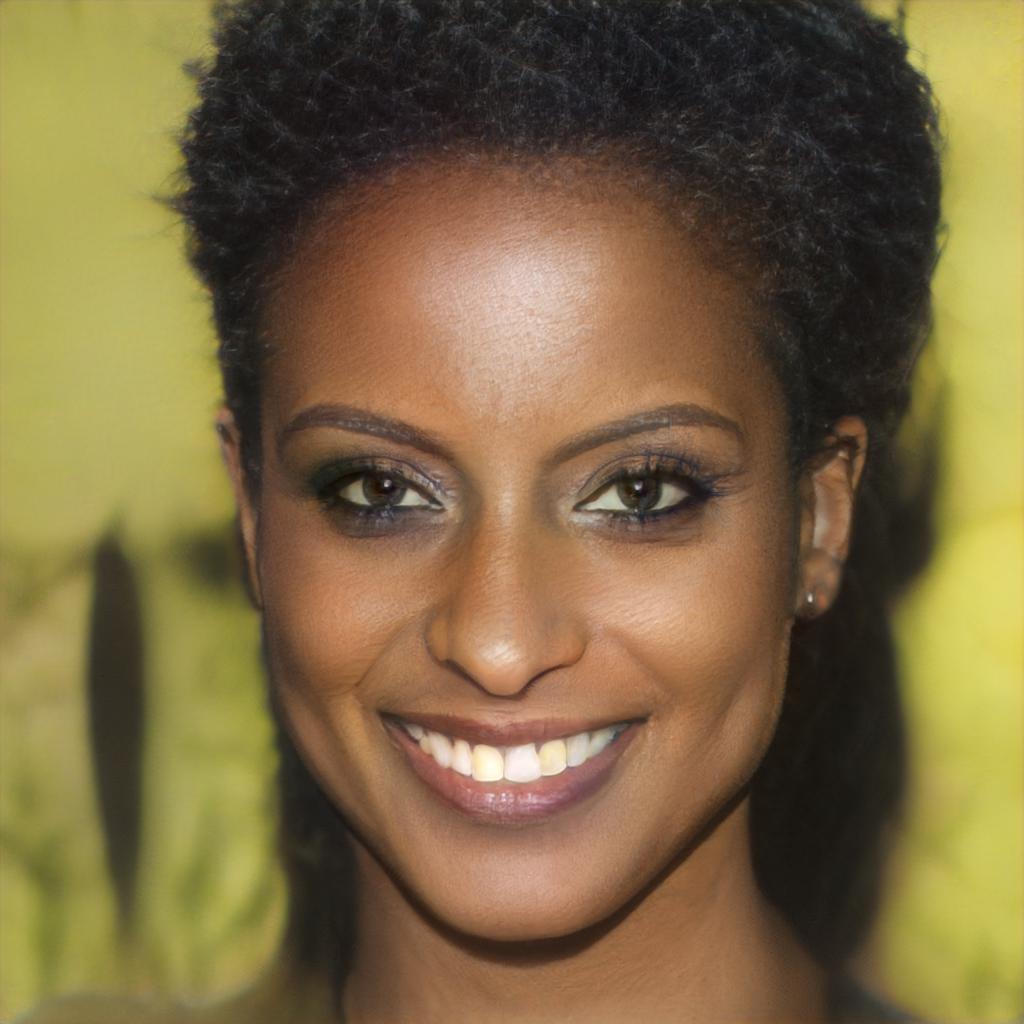} &
              & \raisebox{0.3in}{\rotatebox[origin=t]{90}{HRFAE}} & 
                \includegraphics[width=0.10\textwidth]{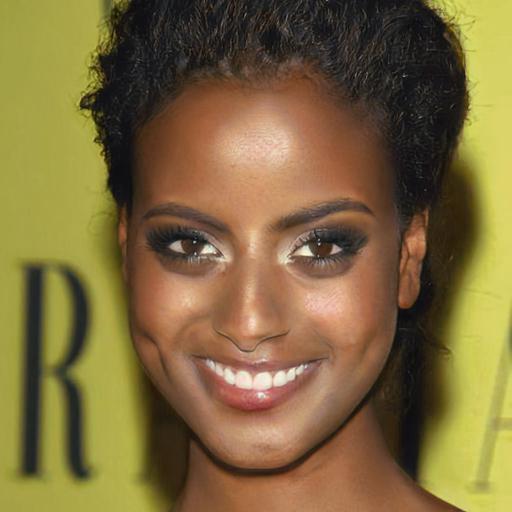} &
                \includegraphics[width=0.10\textwidth]{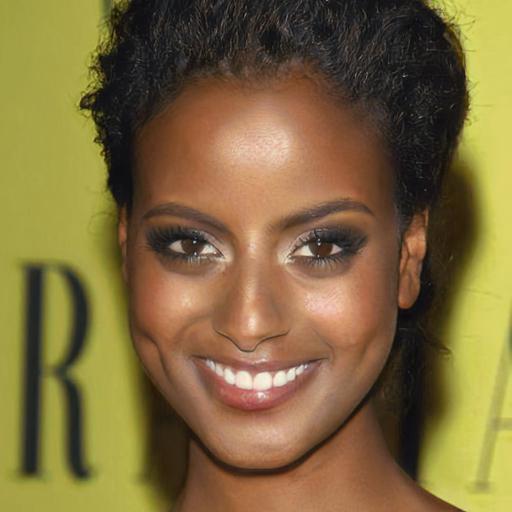} &
                \includegraphics[width=0.10\textwidth]{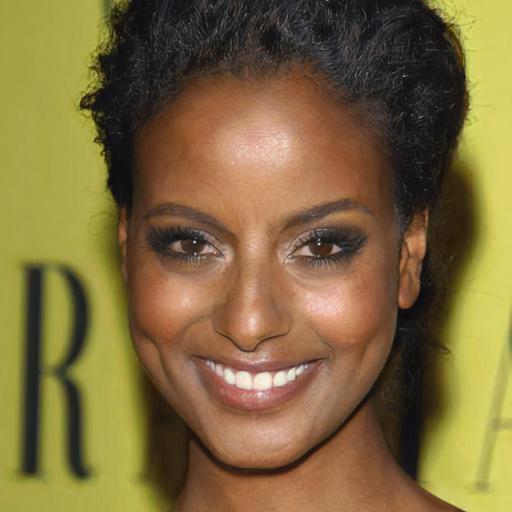} &
                \includegraphics[width=0.10\textwidth]{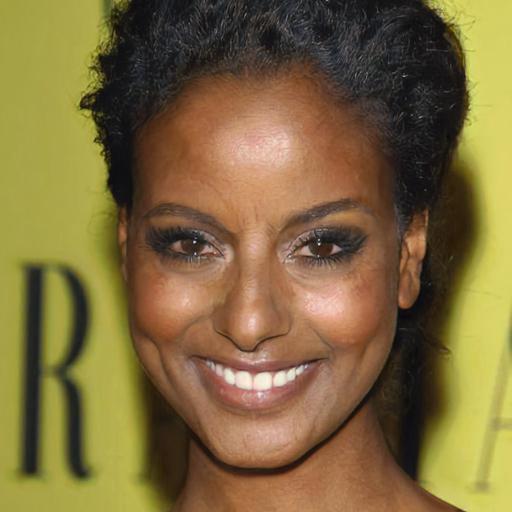} &
                \includegraphics[width=0.10\textwidth]{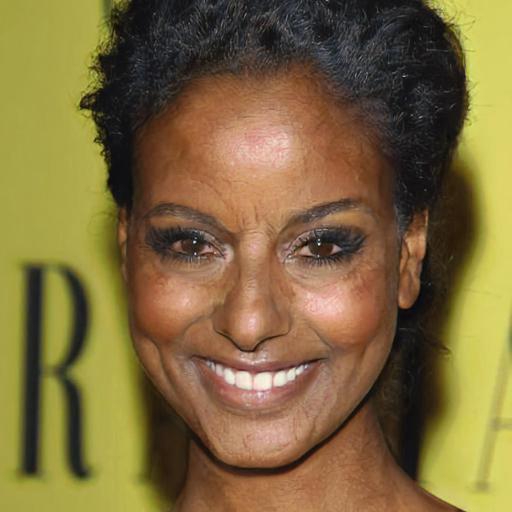} \\
              & & \raisebox{0.3in}{\rotatebox[origin=t]{90}{SAM}} &
                \includegraphics[width=0.10\textwidth]{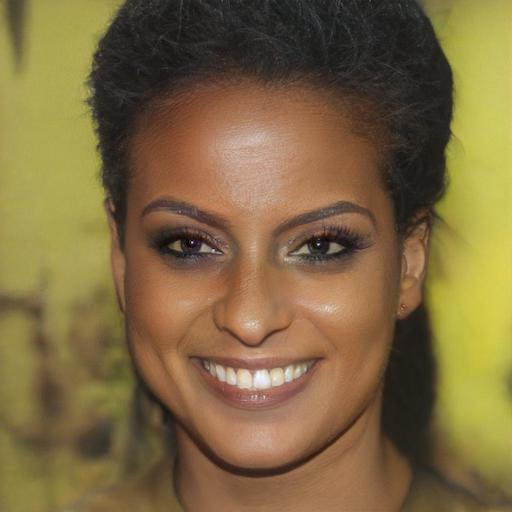} &
                \includegraphics[width=0.10\textwidth]{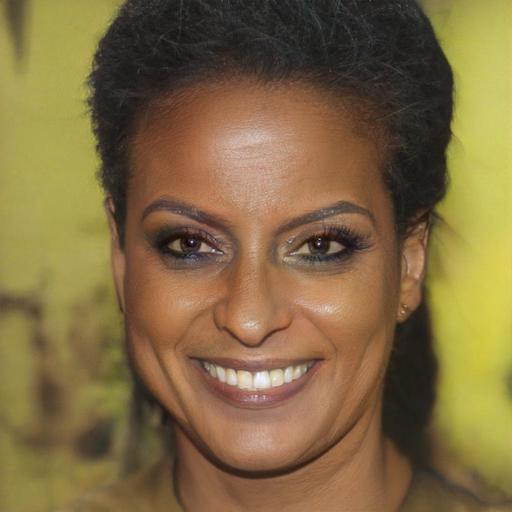} &
                \includegraphics[width=0.10\textwidth]{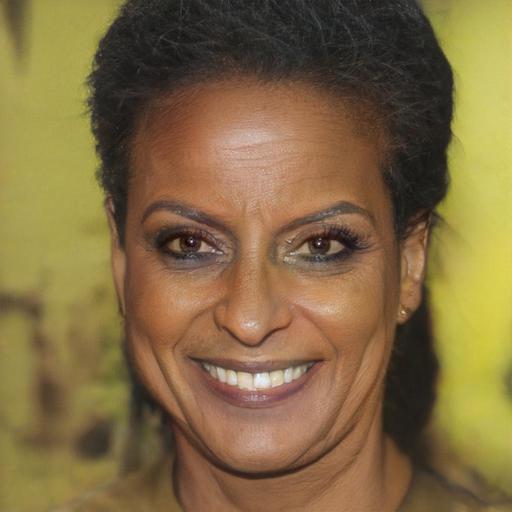} &
                \includegraphics[width=0.10\textwidth]{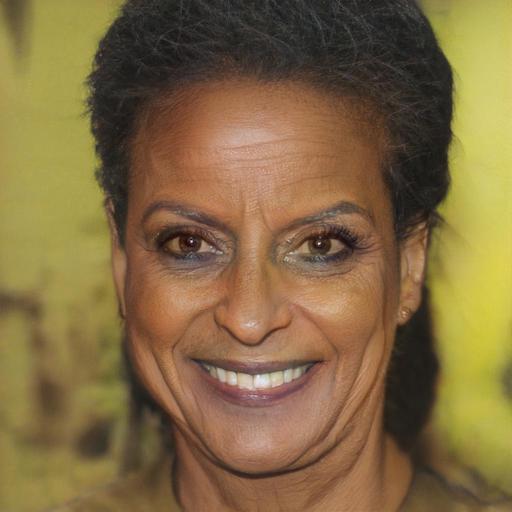} &
                \includegraphics[width=0.10\textwidth]{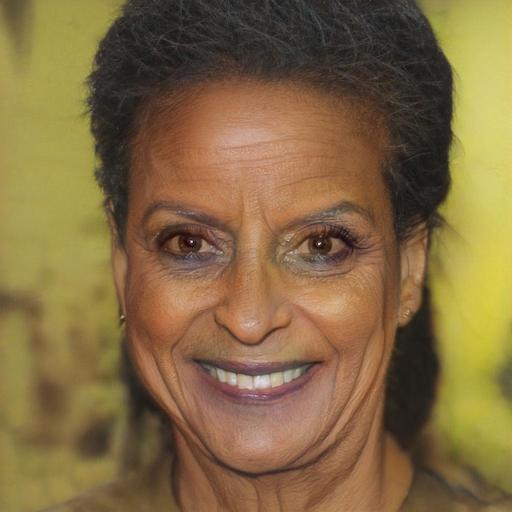}
        \tabularnewline
        
        \includegraphics[width=0.10\textwidth]{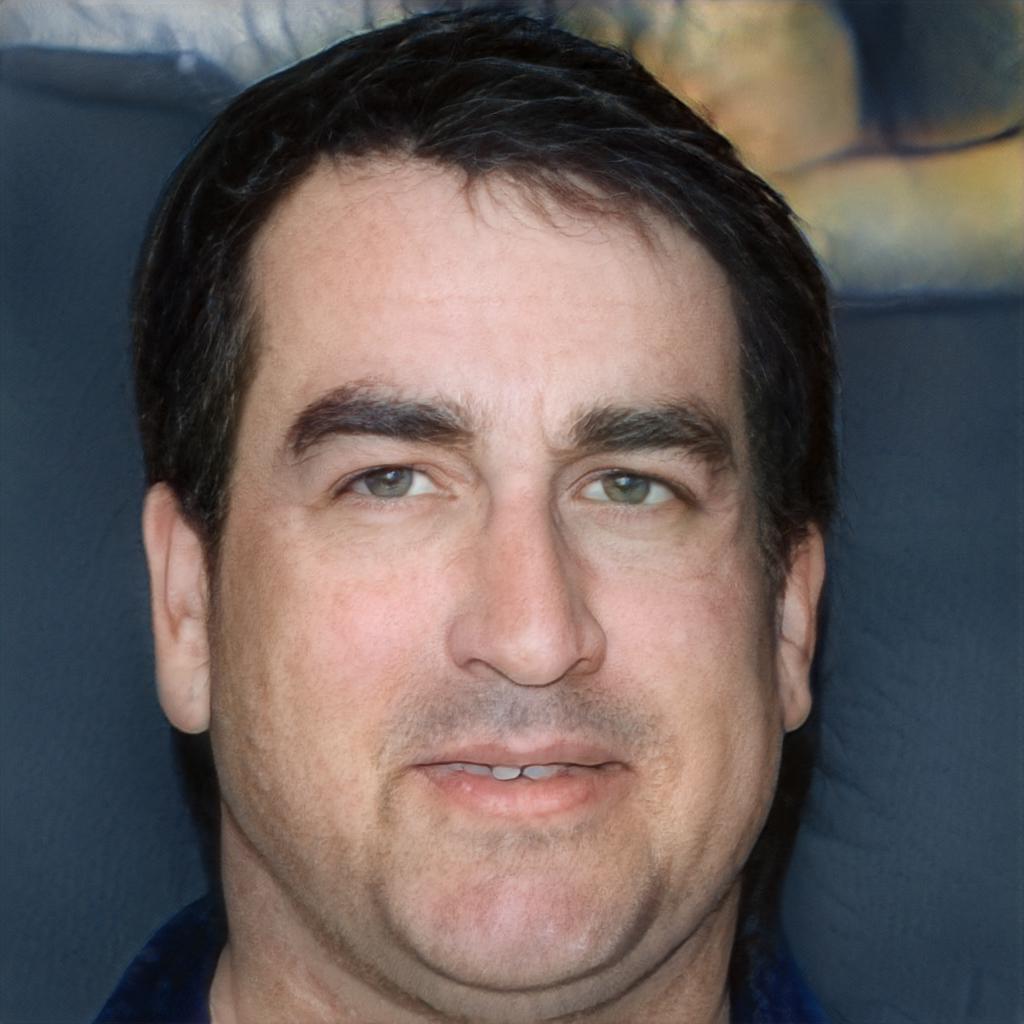} &
              & \raisebox{0.3in}{\rotatebox[origin=t]{90}{HRFAE}} & 
                \includegraphics[width=0.10\textwidth]{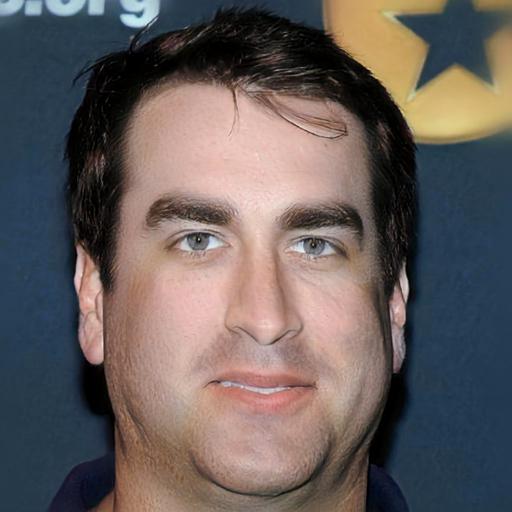} &
                \includegraphics[width=0.10\textwidth]{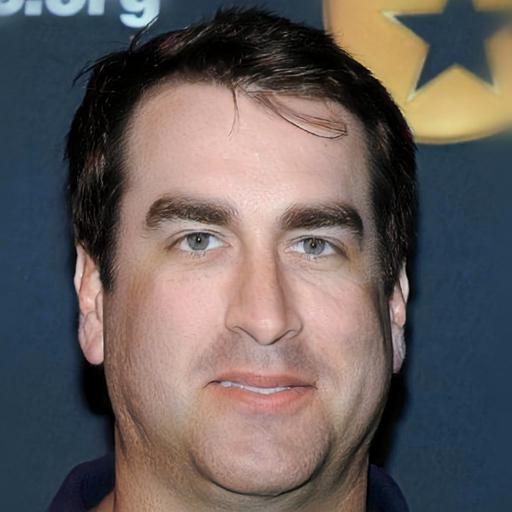} &
                \includegraphics[width=0.10\textwidth]{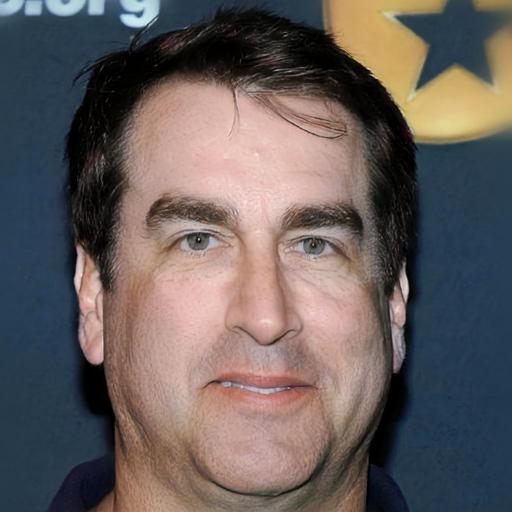} &
                \includegraphics[width=0.10\textwidth]{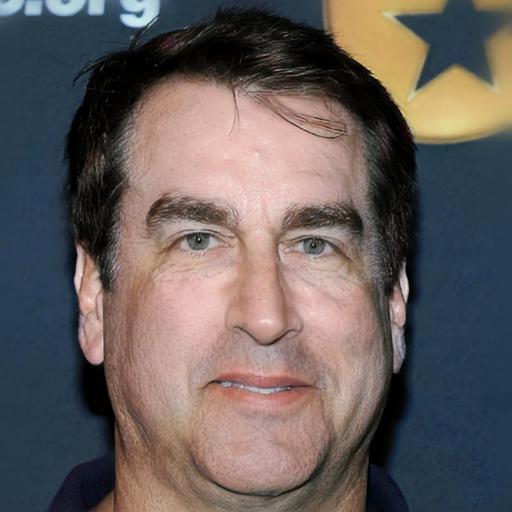} &
                \includegraphics[width=0.10\textwidth]{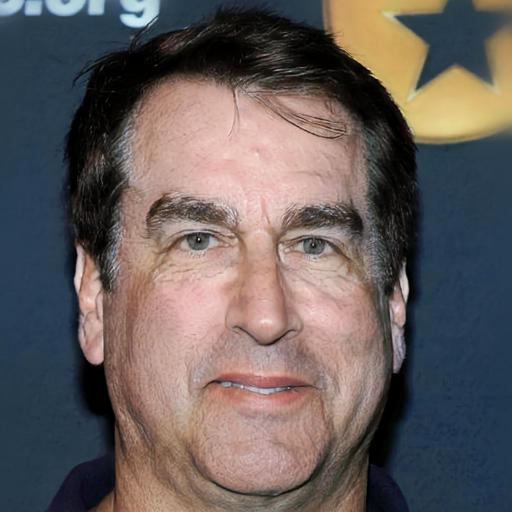} \\
              & & \raisebox{0.3in}{\rotatebox[origin=t]{90}{SAM}} &
                \includegraphics[width=0.10\textwidth]{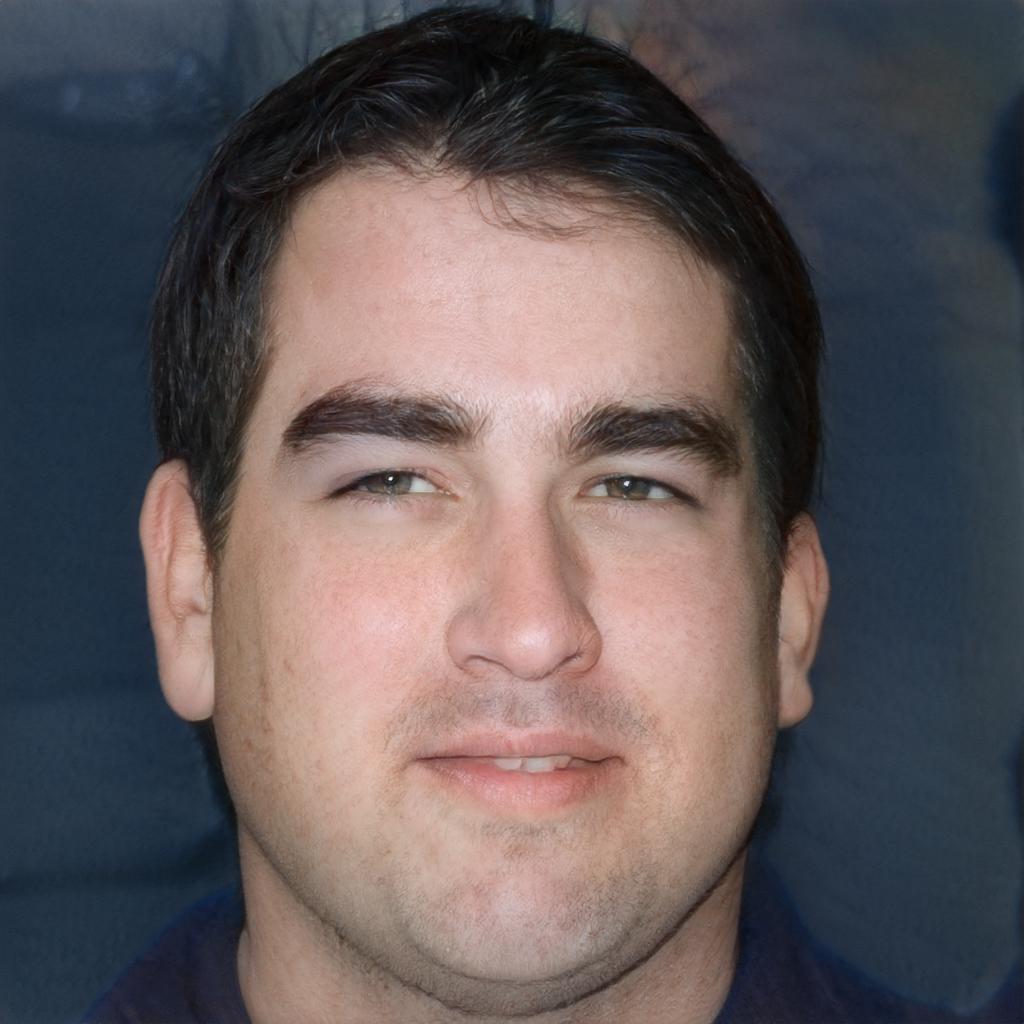} &
                \includegraphics[width=0.10\textwidth]{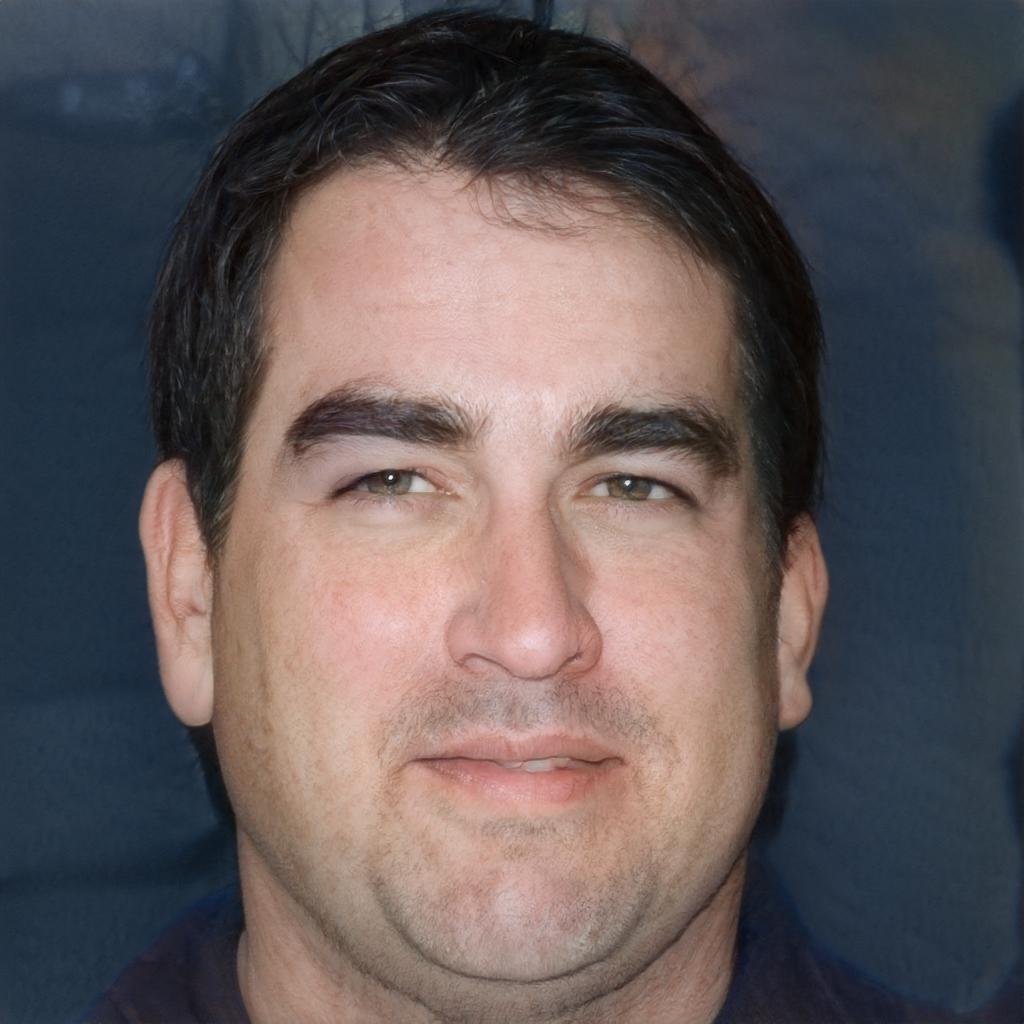} &
                \includegraphics[width=0.10\textwidth]{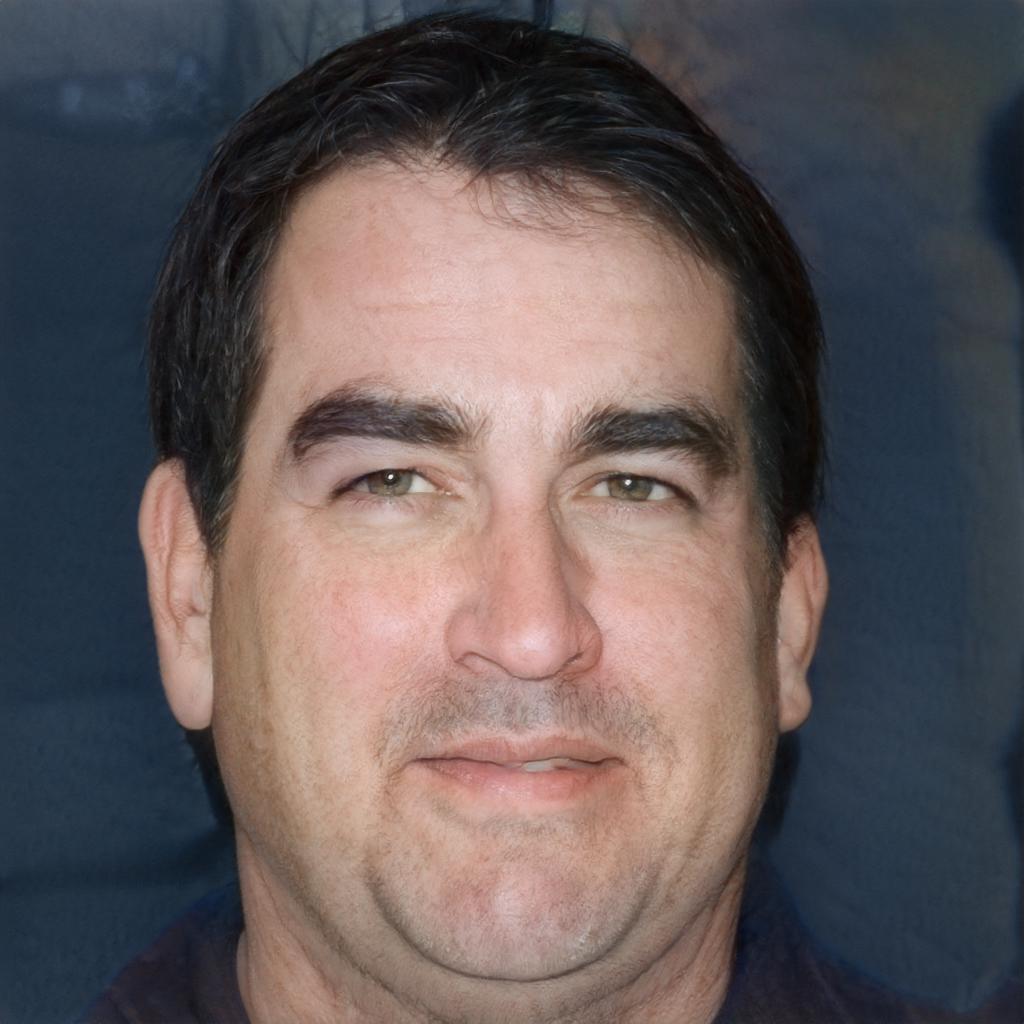} &
                \includegraphics[width=0.10\textwidth]{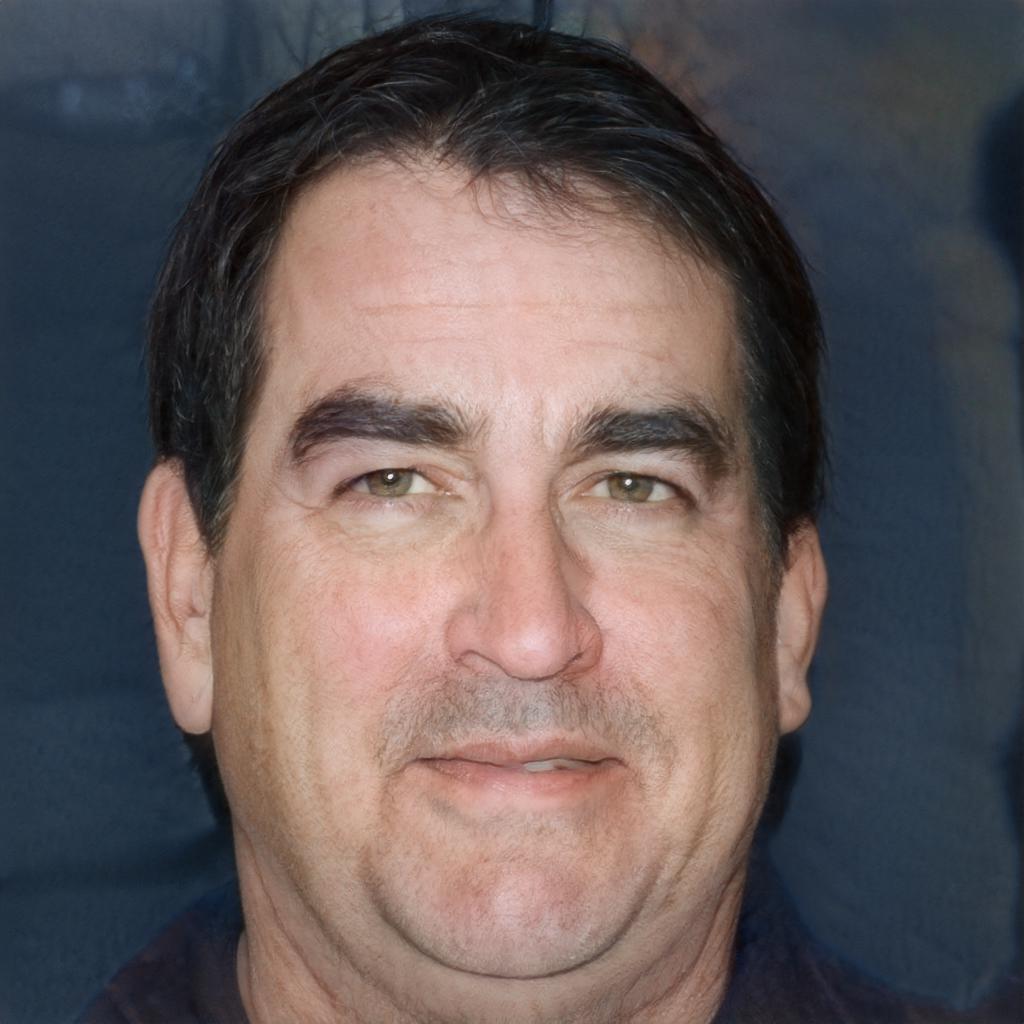} &
                \includegraphics[width=0.10\textwidth]{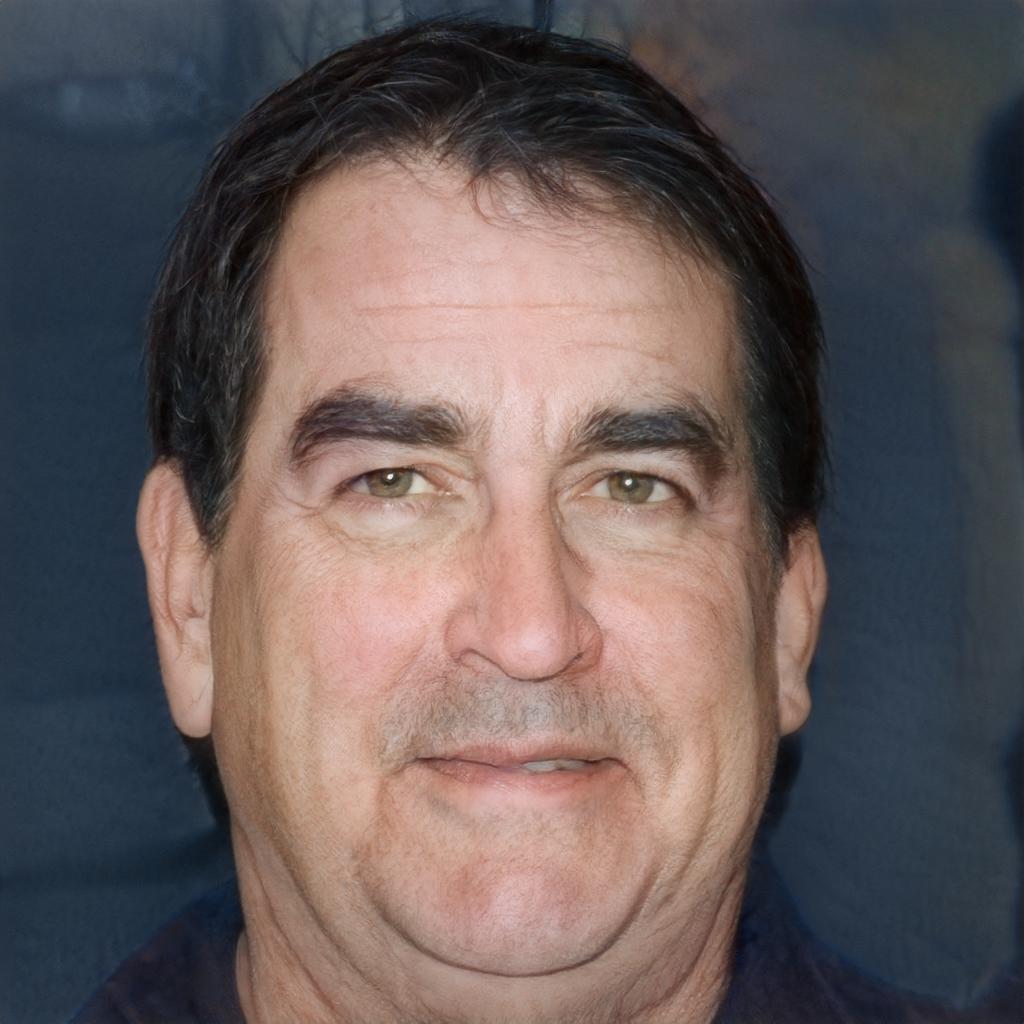}
        \tabularnewline
        
        \includegraphics[width=0.10\textwidth]{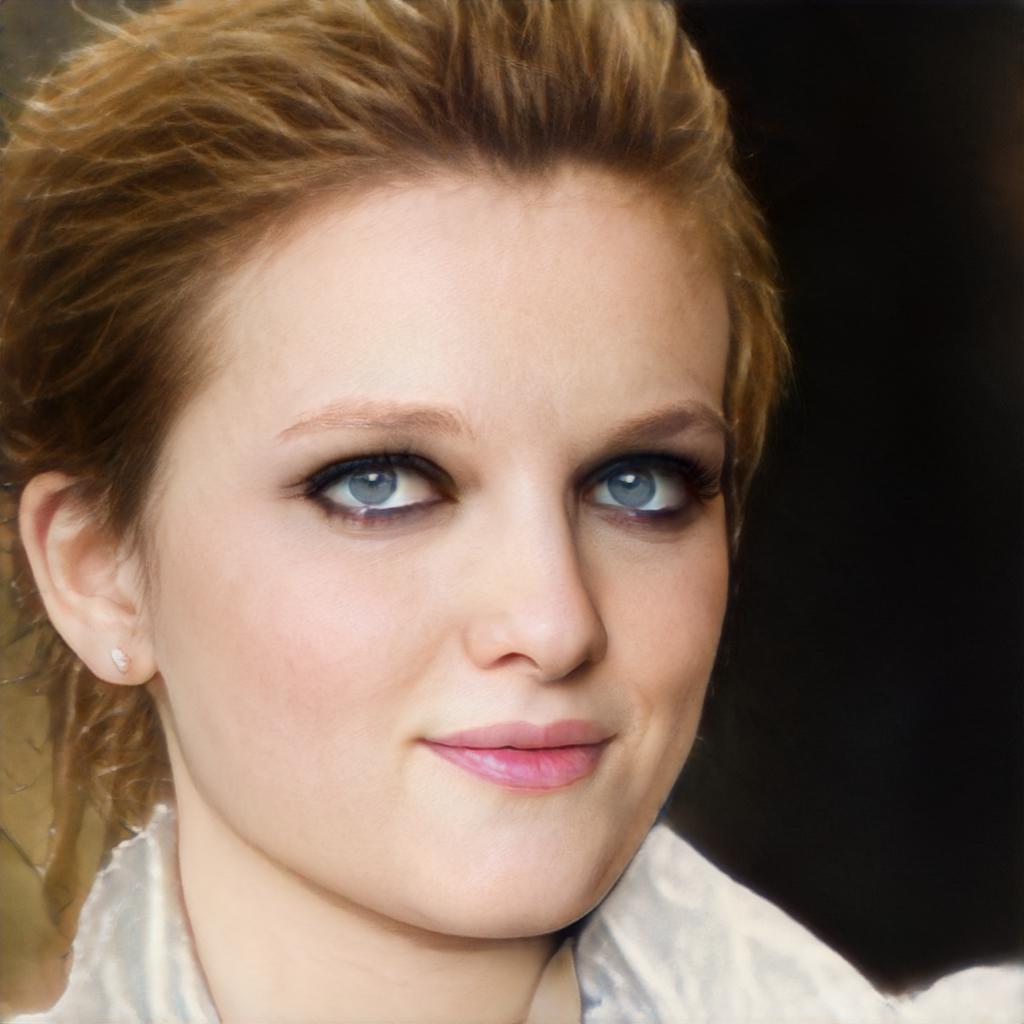} &
              & \raisebox{0.3in}{\rotatebox[origin=t]{90}{HRFAE}} & 
                \includegraphics[width=0.10\textwidth]{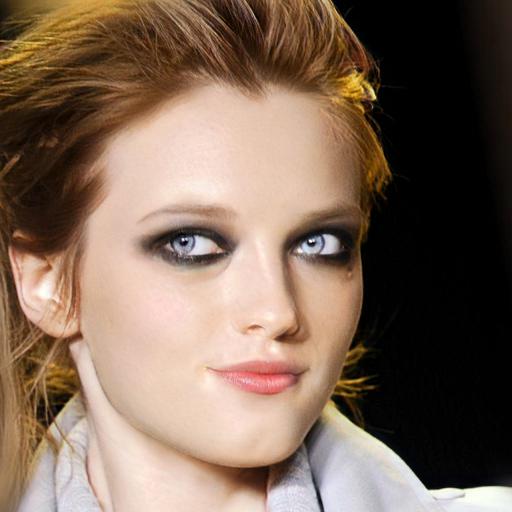} &
                \includegraphics[width=0.10\textwidth]{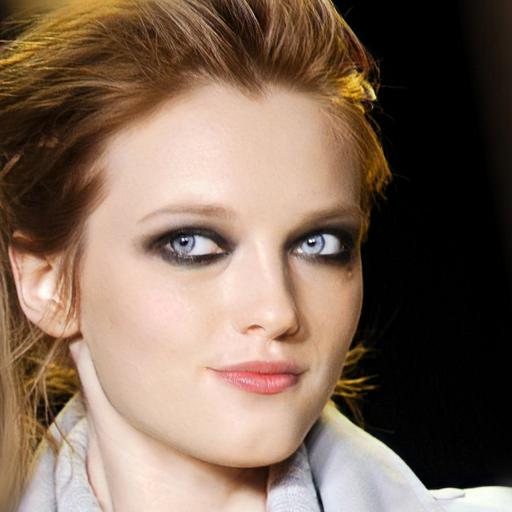} &
                \includegraphics[width=0.10\textwidth]{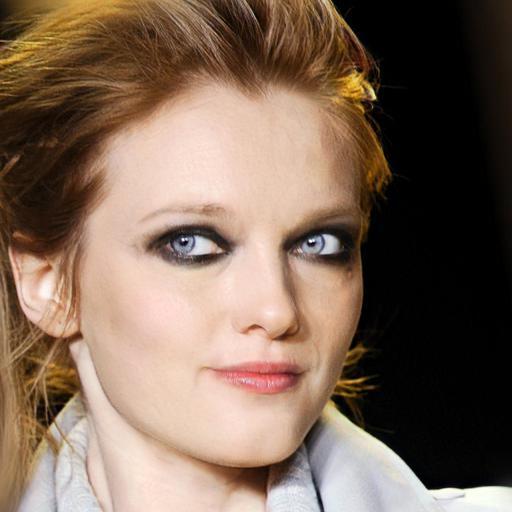} &
                \includegraphics[width=0.10\textwidth]{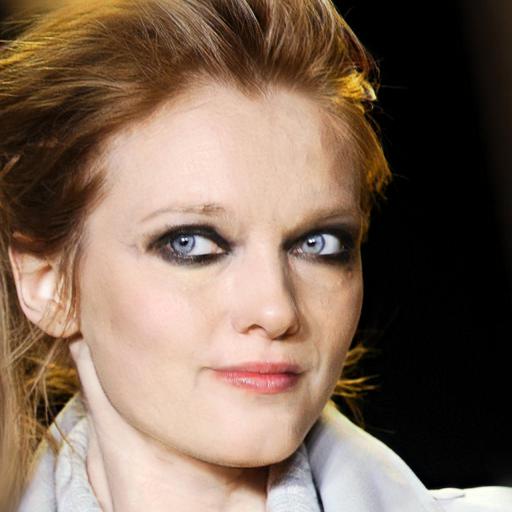} &
                \includegraphics[width=0.10\textwidth]{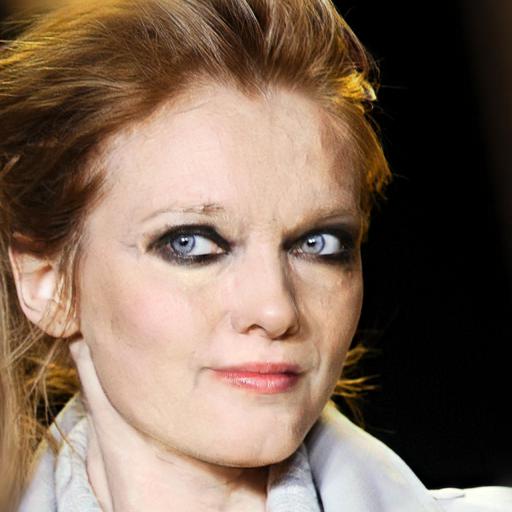} \\
              & & \raisebox{0.3in}{\rotatebox[origin=t]{90}{SAM}} &
                \includegraphics[width=0.10\textwidth]{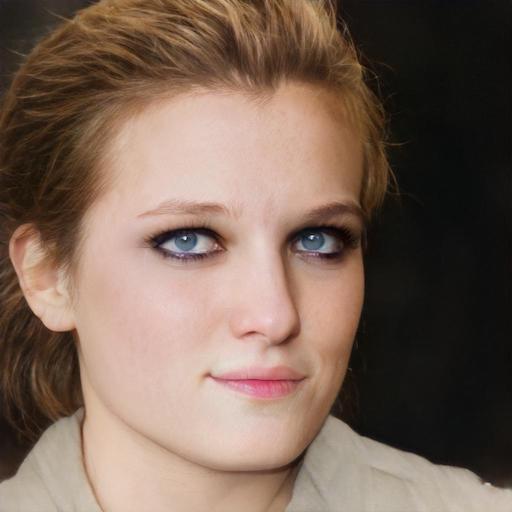} &
                \includegraphics[width=0.10\textwidth]{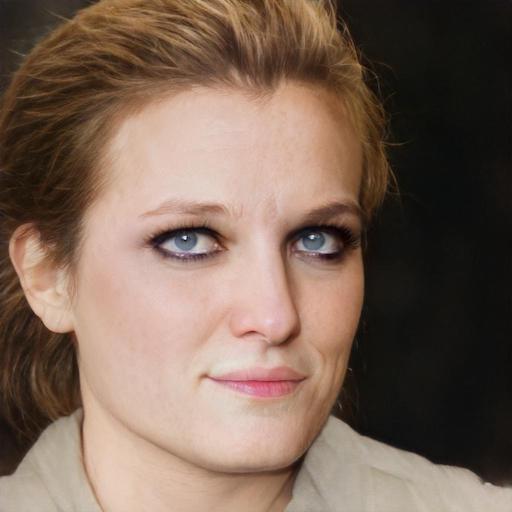} &
                \includegraphics[width=0.10\textwidth]{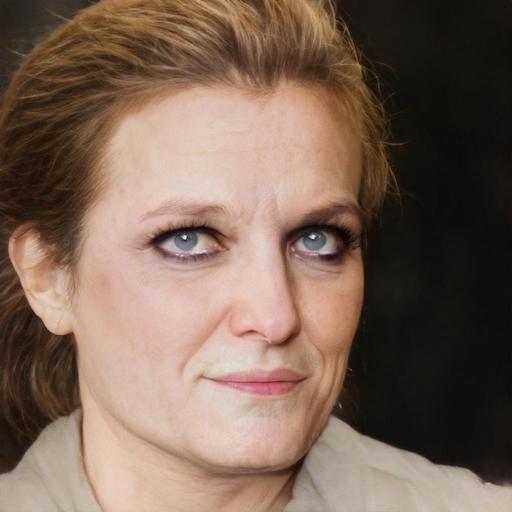} &
                \includegraphics[width=0.10\textwidth]{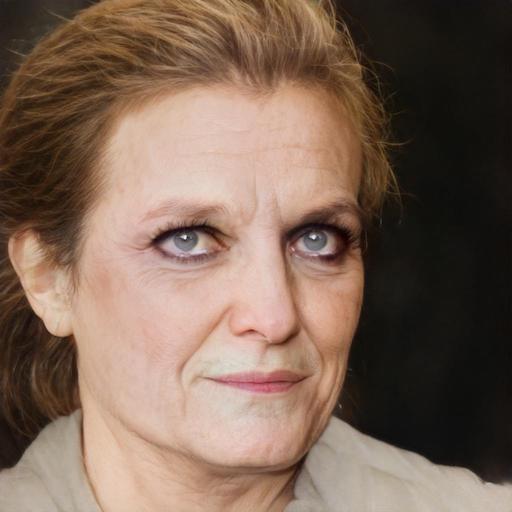} &
                \includegraphics[width=0.10\textwidth]{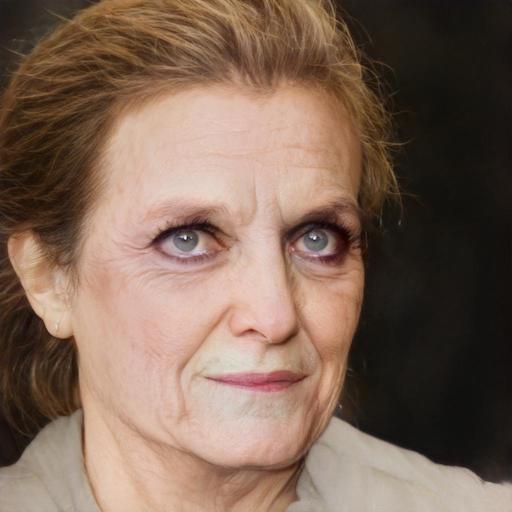}
        \tabularnewline

        \includegraphics[width=0.10\textwidth]{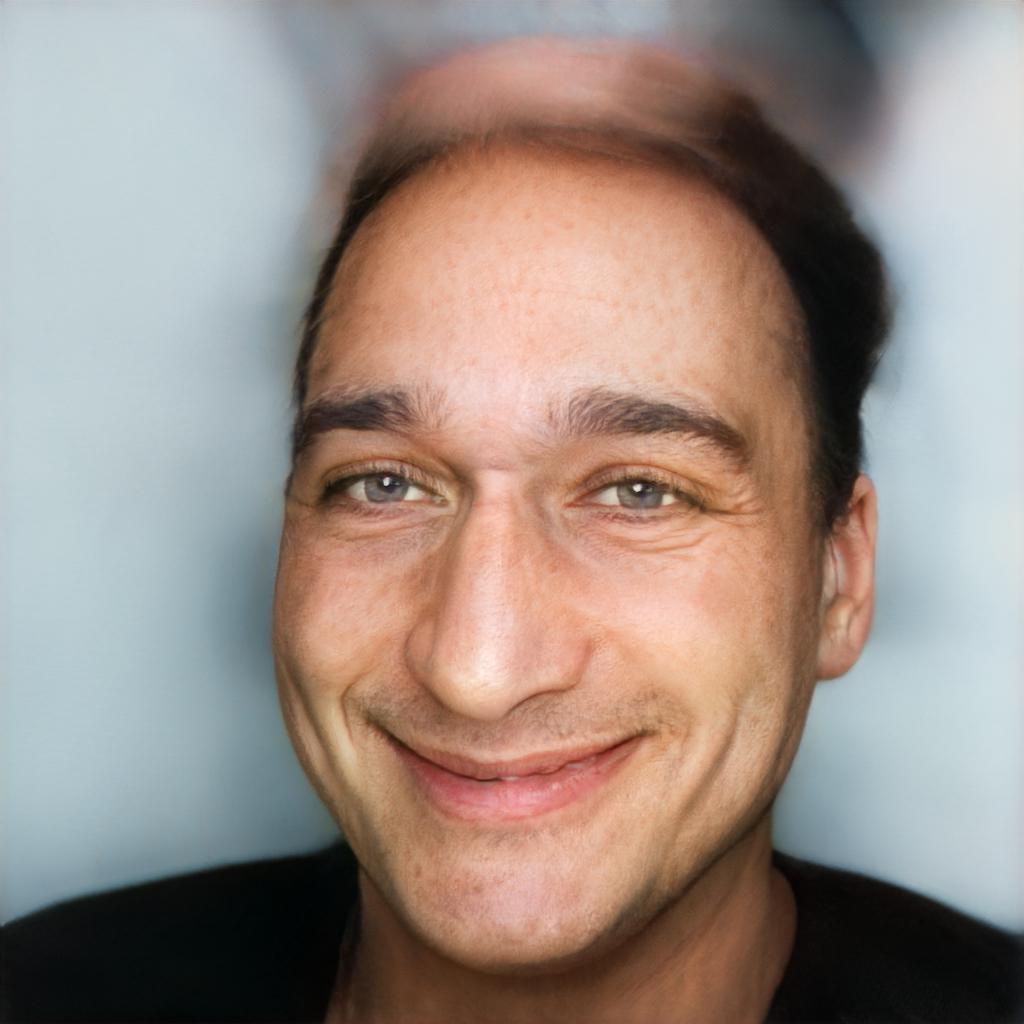} &
              & \raisebox{0.3in}{\rotatebox[origin=t]{90}{HRFAE}} & 
                \includegraphics[width=0.10\textwidth]{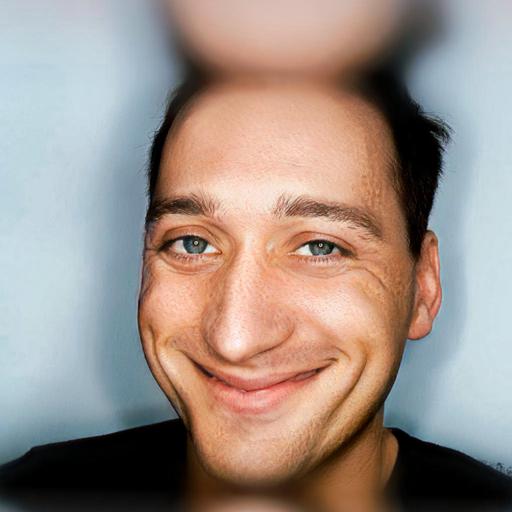} &
                \includegraphics[width=0.10\textwidth]{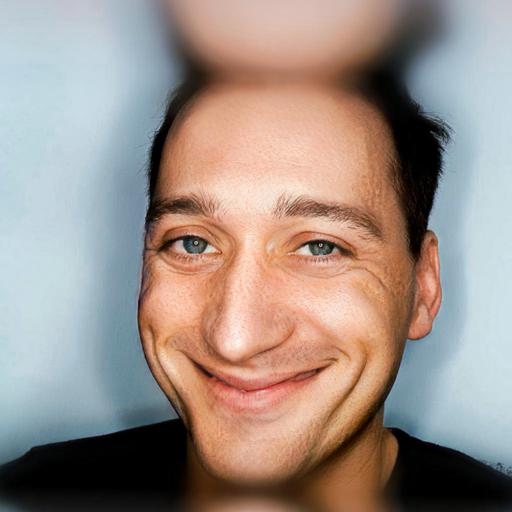} &
                \includegraphics[width=0.10\textwidth]{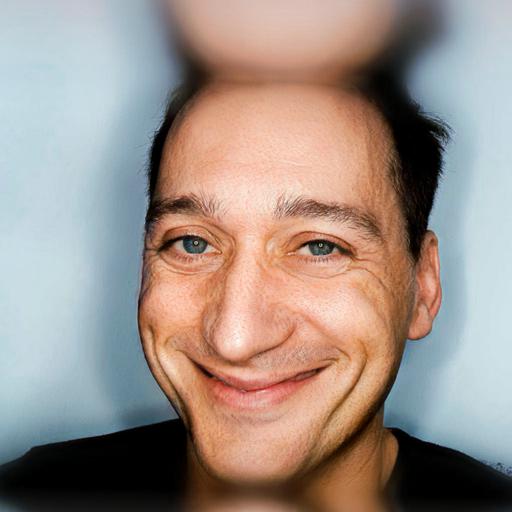} &
                \includegraphics[width=0.10\textwidth]{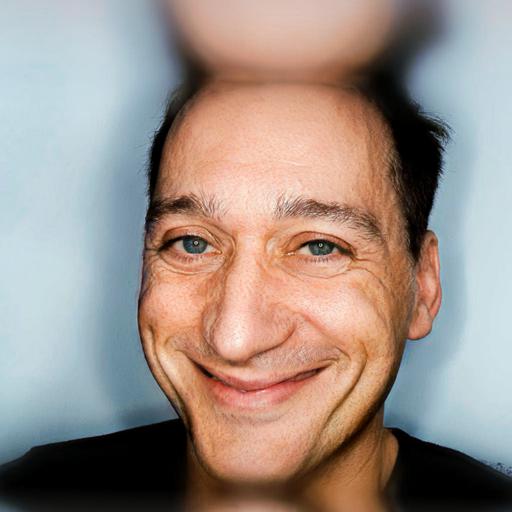} &
                \includegraphics[width=0.10\textwidth]{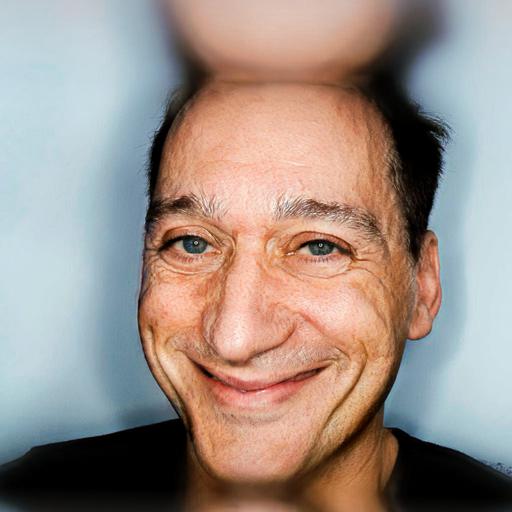} \\
              & & \raisebox{0.3in}{\rotatebox[origin=t]{90}{SAM}} &
                \includegraphics[width=0.10\textwidth]{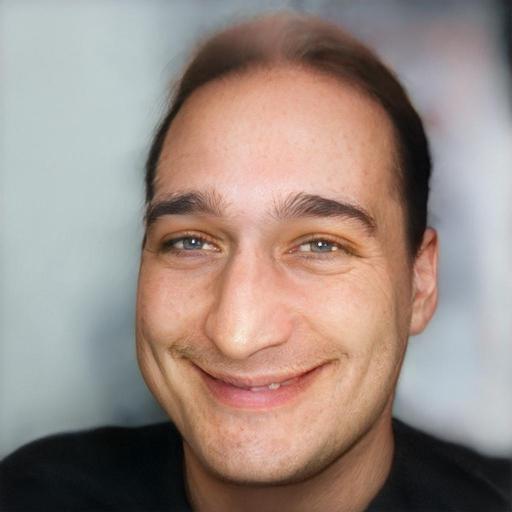} &
                \includegraphics[width=0.10\textwidth]{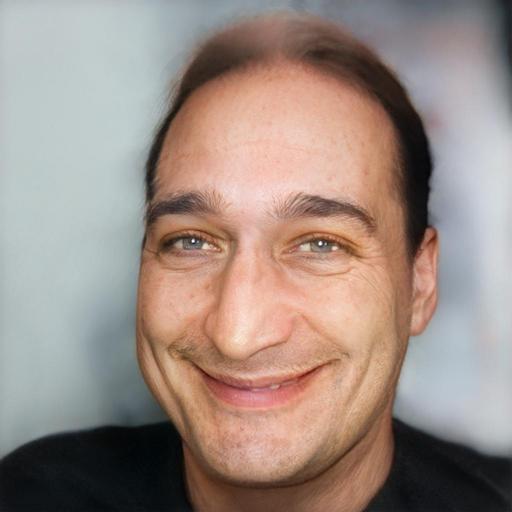} &
                \includegraphics[width=0.10\textwidth]{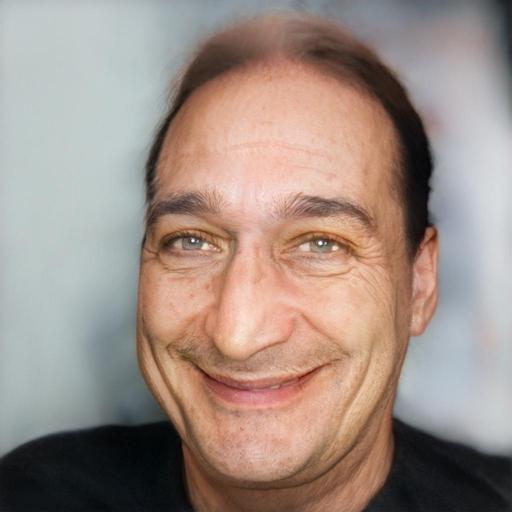} &
                \includegraphics[width=0.10\textwidth]{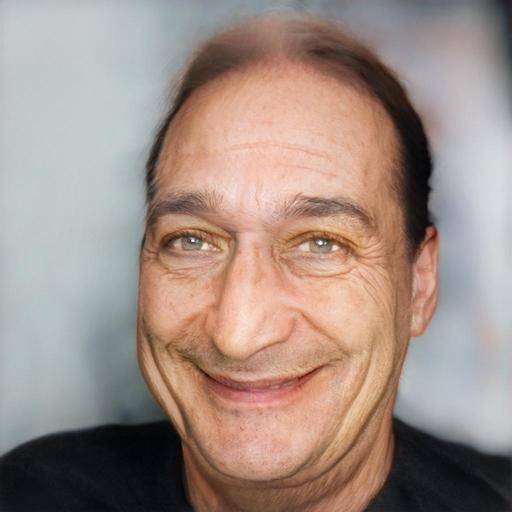} &
                \includegraphics[width=0.10\textwidth]{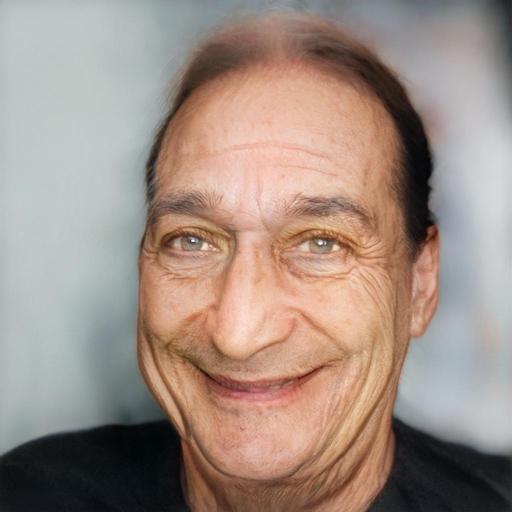}
        \tabularnewline

    \end{tabular}
    \label{fig:appendix_comparison_HRFAEspan}
    \caption{Additional qualitative comparisons of age transformation results with  HRFAE~\cite{yao2020high} on the CelebA-HQ~\cite{karras2017progressive} test set.}
\end{figure*}

%% file: figures/appendix/appendix_latent_space.tex
\begin{figure*}
    \centering
    \setlength{\belowcaptionskip}{-2.5pt}
    \setlength{\tabcolsep}{1pt}
    \begin{tabular}{c c c}
        Inversion & & \\
        \includegraphics[width=0.09\textwidth]{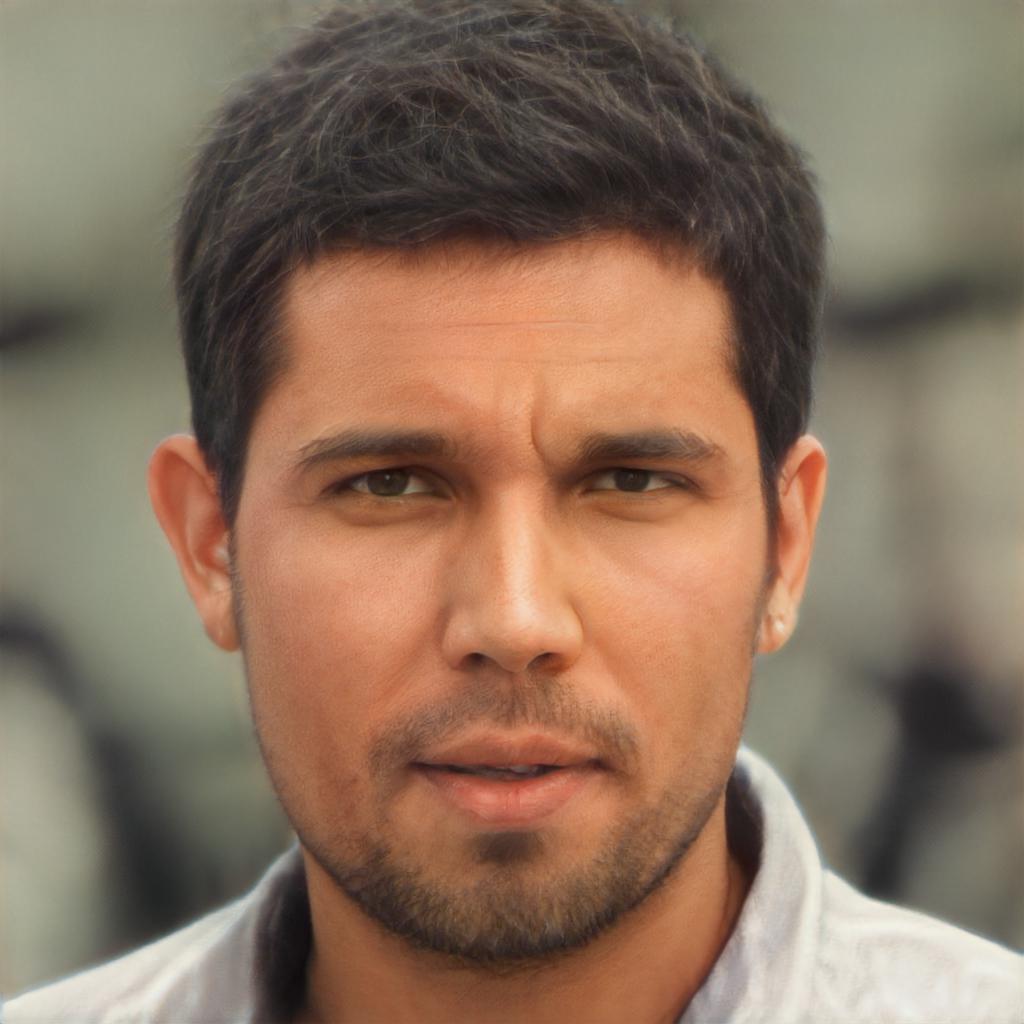} &
        \raisebox{0.25in}{\rotatebox[origin=t]{90}{InterFace SG1}} & 
        \includegraphics[width=0.8\textwidth]{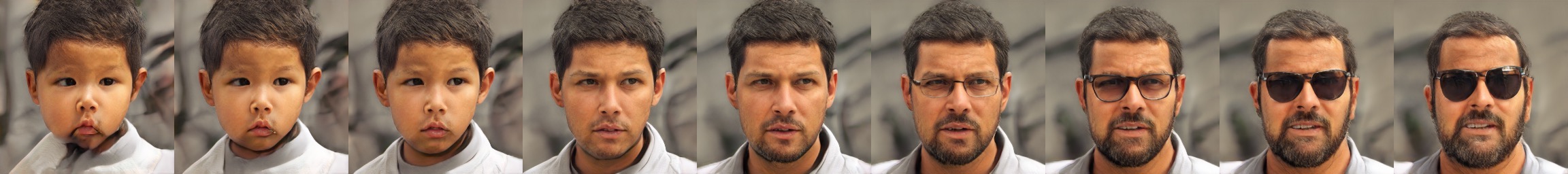} \\
        & \raisebox{0.225in}{\rotatebox[origin=t]{90}{StyleFlow}} & 
        \includegraphics[width=0.8\textwidth]{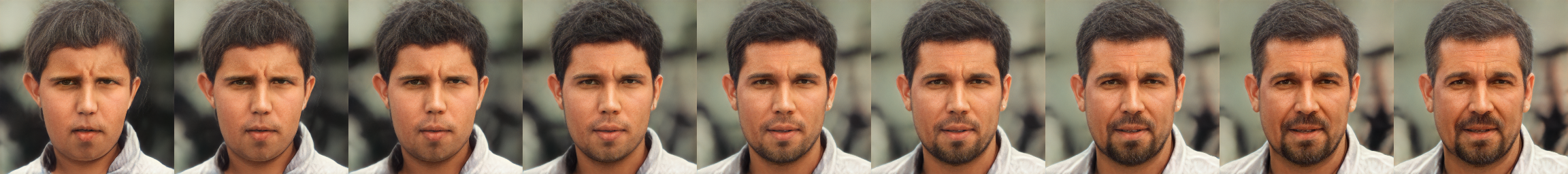} \\
        & \raisebox{0.225in}{\rotatebox[origin=t]{90}{SAM}} &
        \includegraphics[width=0.8\textwidth]{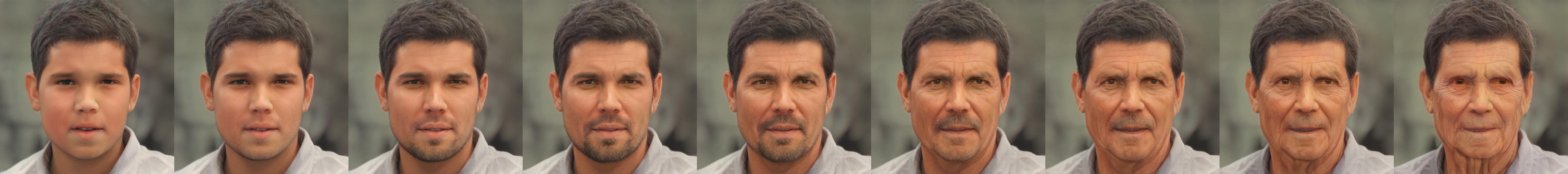}
        \tabularnewline

        \includegraphics[width=0.09\textwidth]{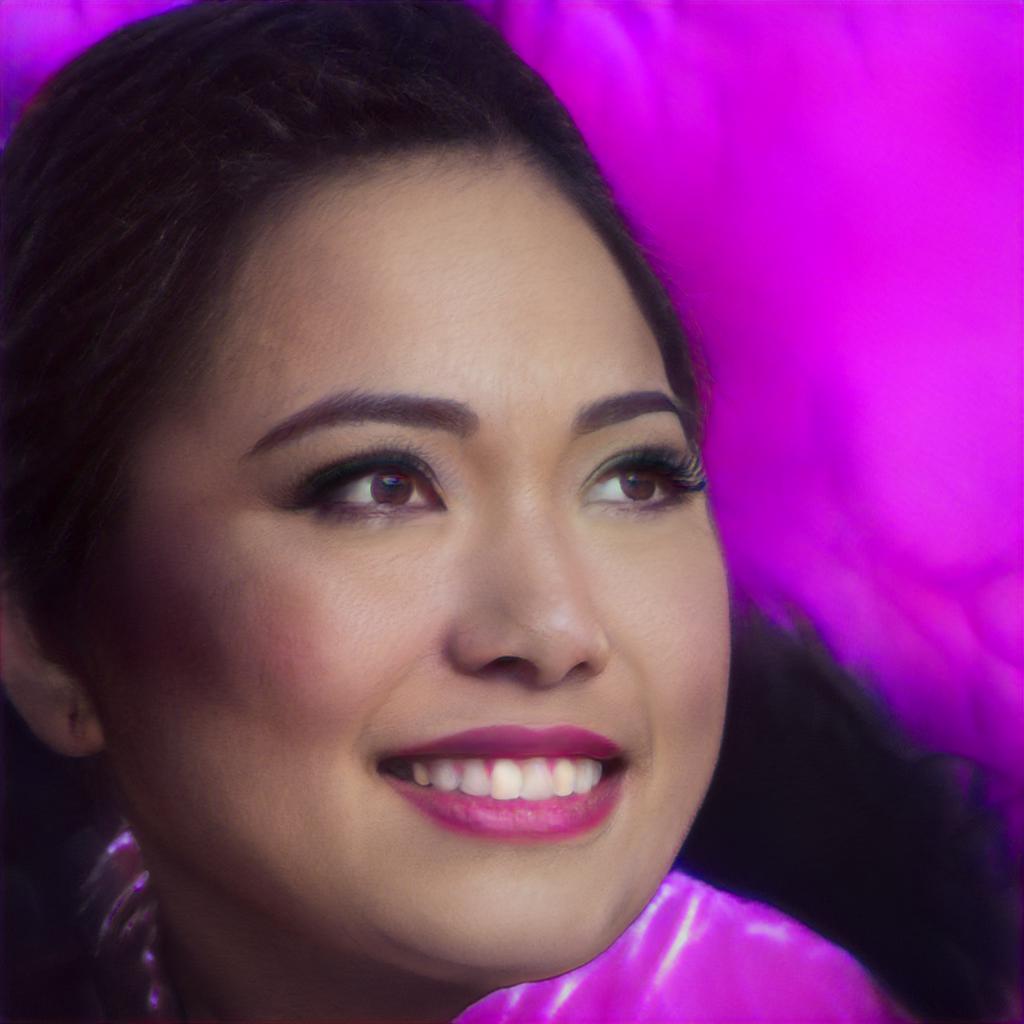} &
        \raisebox{0.25in}{\rotatebox[origin=t]{90}{InterFace SG1}} & 
        \includegraphics[width=0.8\textwidth]{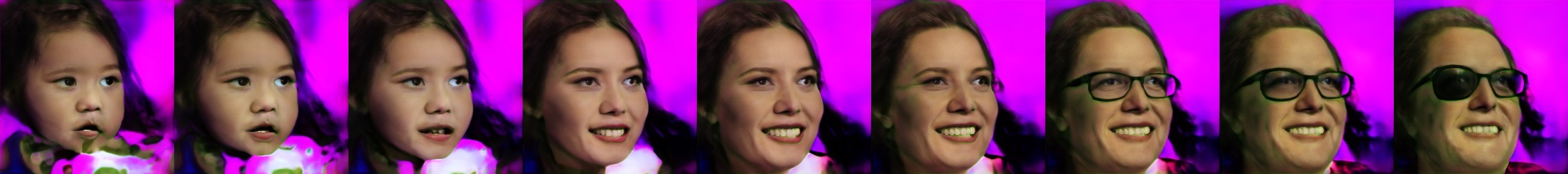} \\
        & \raisebox{0.225in}{\rotatebox[origin=t]{90}{StyleFlow}} & 
        \includegraphics[width=0.8\textwidth]{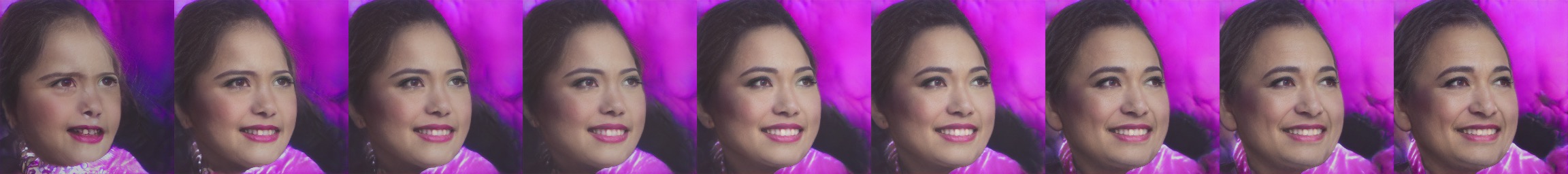} \\
        & \raisebox{0.225in}{\rotatebox[origin=t]{90}{SAM}} &
        \includegraphics[width=0.8\textwidth]{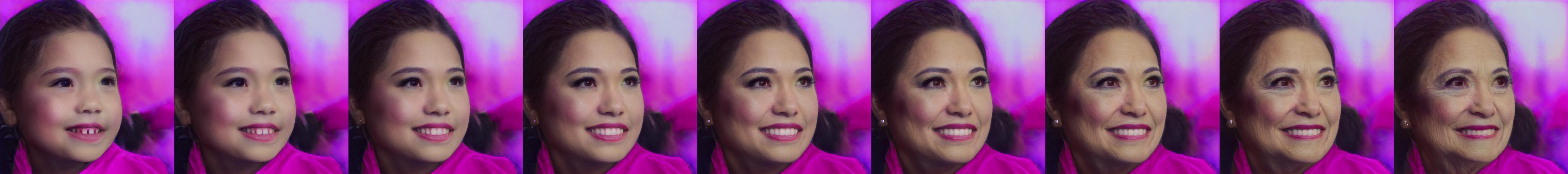}
        \tabularnewline

        \includegraphics[width=0.09\textwidth]{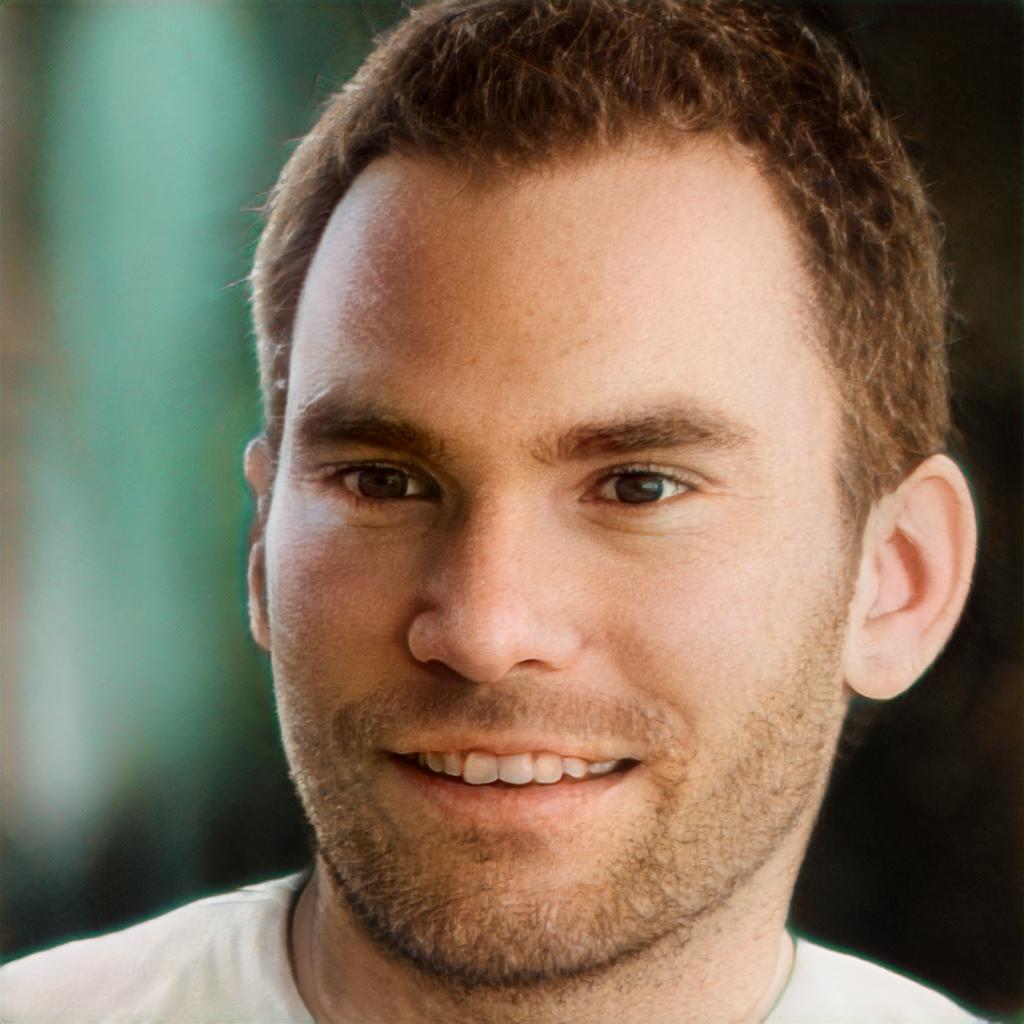} &
        \raisebox{0.25in}{\rotatebox[origin=t]{90}{InterFace SG1}} & 
        \includegraphics[width=0.8\textwidth]{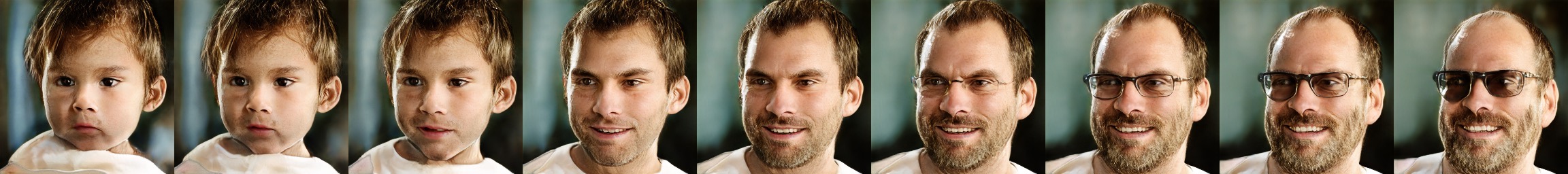} \\        
        & \raisebox{0.225in}{\rotatebox[origin=t]{90}{StyleFlow}} & 
        \includegraphics[width=0.8\textwidth]{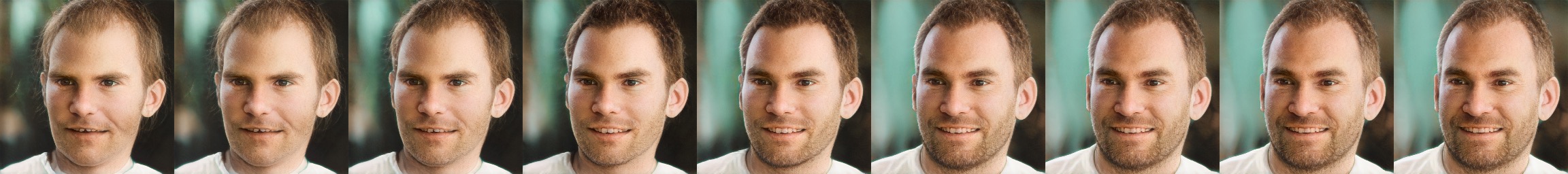} \\
        & \raisebox{0.225in}{\rotatebox[origin=t]{90}{SAM}} &
        \includegraphics[width=0.8\textwidth]{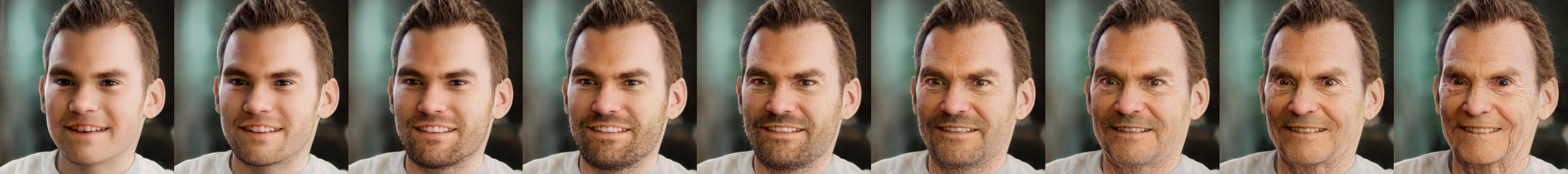}
    \end{tabular}
    \vspace{-0.15cm}
    \caption{Additional visual comparisons with InterFaceGAN~\cite{shen2020interpreting} and StyleFlow~\cite{abdal2020styleflow} on real face images. Results for InterFaceGAN are obtained using IDInvert~\cite{zhu2020domain} for inversion into the StyleGAN1 latent space. StyleFlow is performed on latents obtained by inverting the input using a pSp~\cite{richardson2020encoding} encoder into the StyleGAN2 latent space.
    }
    \label{fig:appendix_interface_comparison}
\end{figure*}

%% file: figures/appendix/appendix_style_mixing.tex
\begin{figure*}
    \centering
    \setlength{\belowcaptionskip}{-2.5pt}
    \setlength{\tabcolsep}{1pt}
    \begin{tabular}{cc}
        \includegraphics[width=0.125\textwidth]{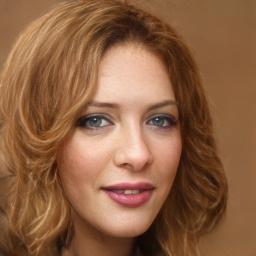} &
        \includegraphics[width=0.625\textwidth]{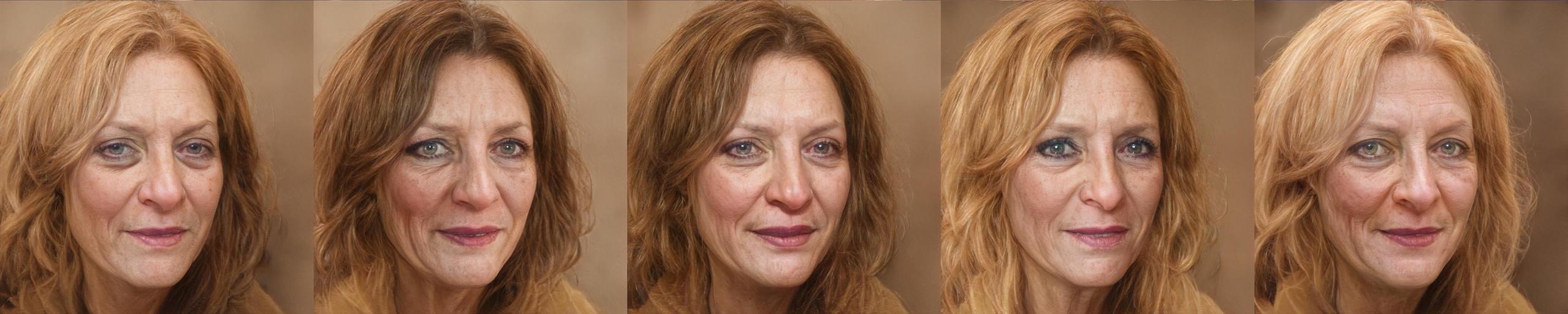}
        \tabularnewline
        \includegraphics[width=0.125\textwidth]{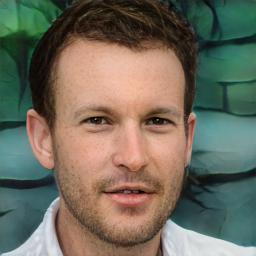} &
        \includegraphics[width=0.625\textwidth]{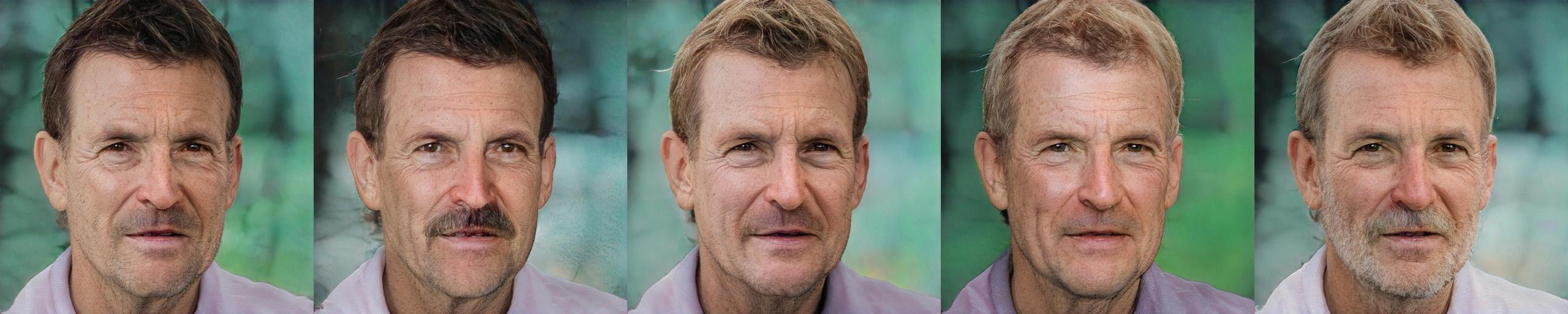}
        \tabularnewline
        \includegraphics[width=0.125\textwidth]{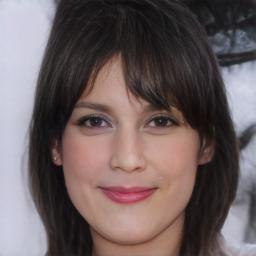} &
        \includegraphics[width=0.625\textwidth]{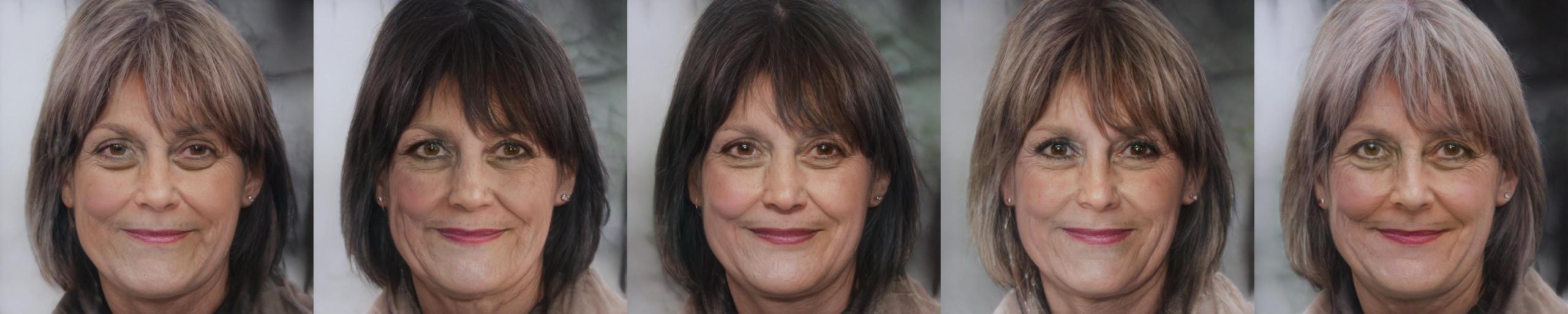}
        \tabularnewline
        \begin{tabular}{@{}c@{}}Inversion \end{tabular} &
        \begin{tabular}{@{}c@{}}Multi-Modal Outputs\end{tabular}
    \end{tabular}
    \setlength{\belowcaptionskip}{-10pt}
    \caption{Additional multi-modal results generated by SAM by performing style-mixing on the age-transformed outputs. Here, for each input image we perform style-mixing with five references images on layers $8-9$ to obtain multiple transformation results.}
    \label{fig:appendix_style_mixing*}
\end{figure*}

%% file: figures/appendix/appendix_patch_editing.tex
\begin{figure*}
    \centering
    \setlength{\belowcaptionskip}{-2.5pt}
    \setlength{\tabcolsep}{1pt}
    \vspace{0.75cm}
    \begin{tabular}{cccc}
        \includegraphics[width=0.15\textwidth]{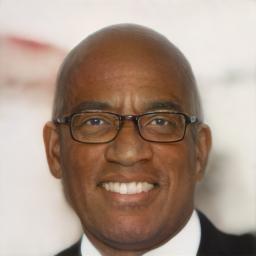} &
        \includegraphics[width=0.15\textwidth]{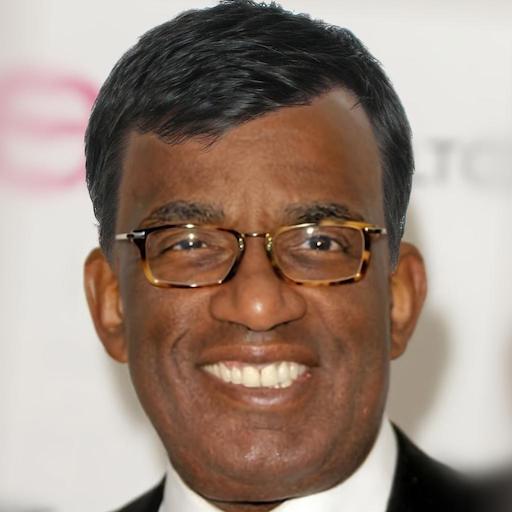} &
        \includegraphics[width=0.15\textwidth]{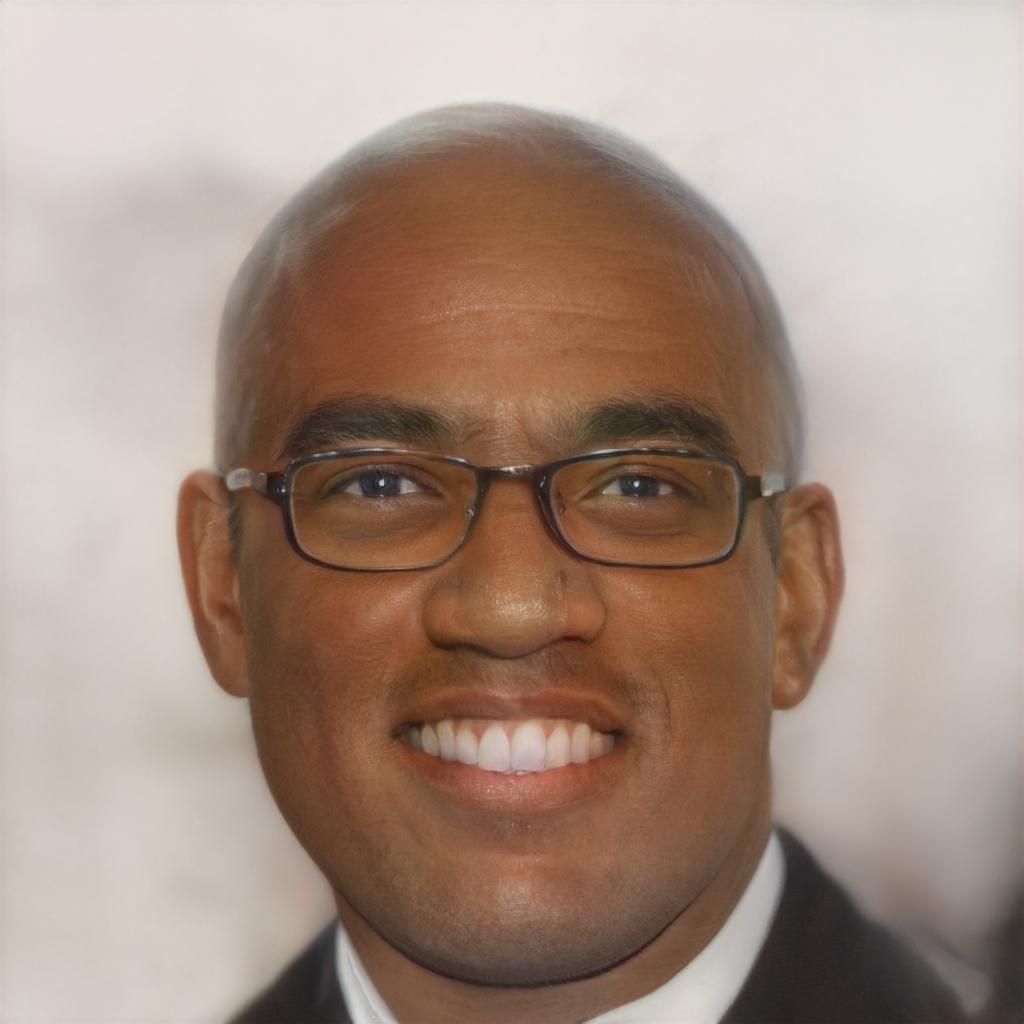} &
        \includegraphics[width=0.15\textwidth]{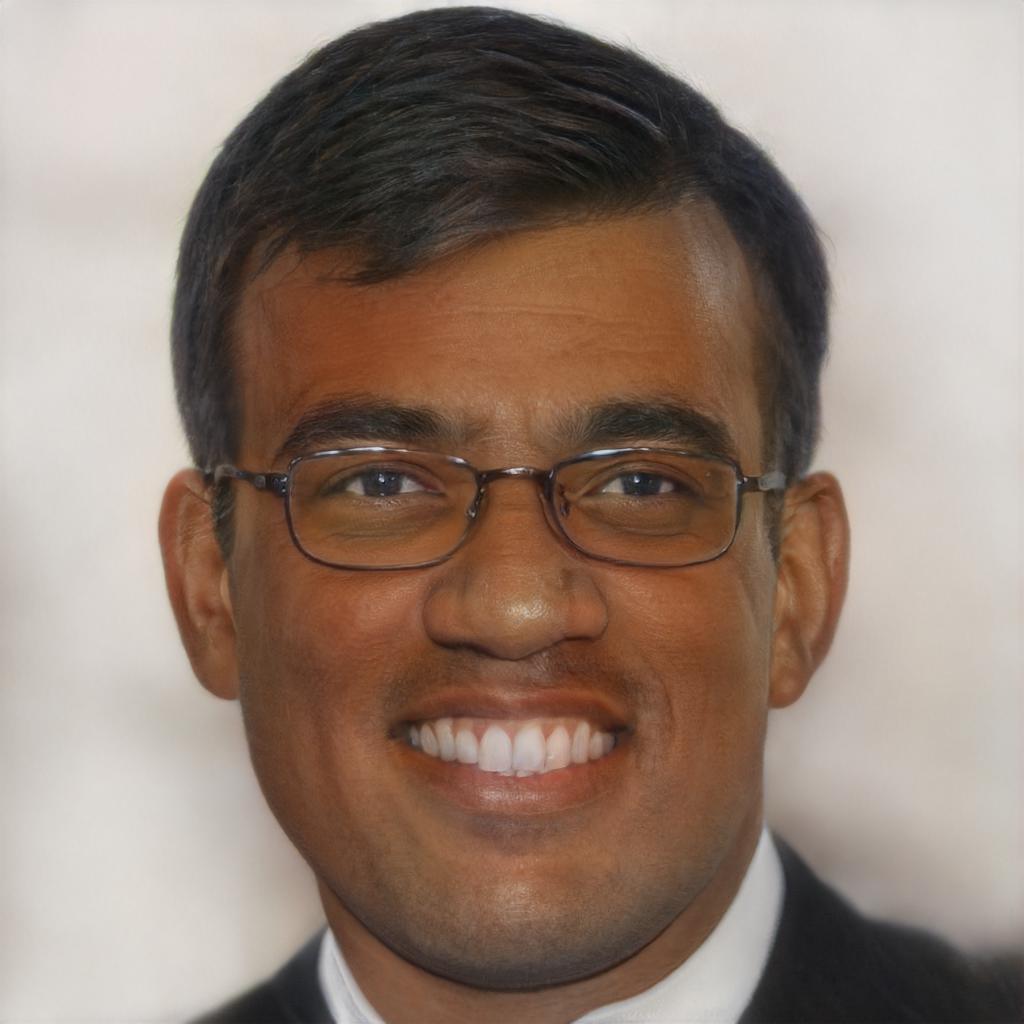}
        \tabularnewline
        \includegraphics[width=0.15\textwidth]{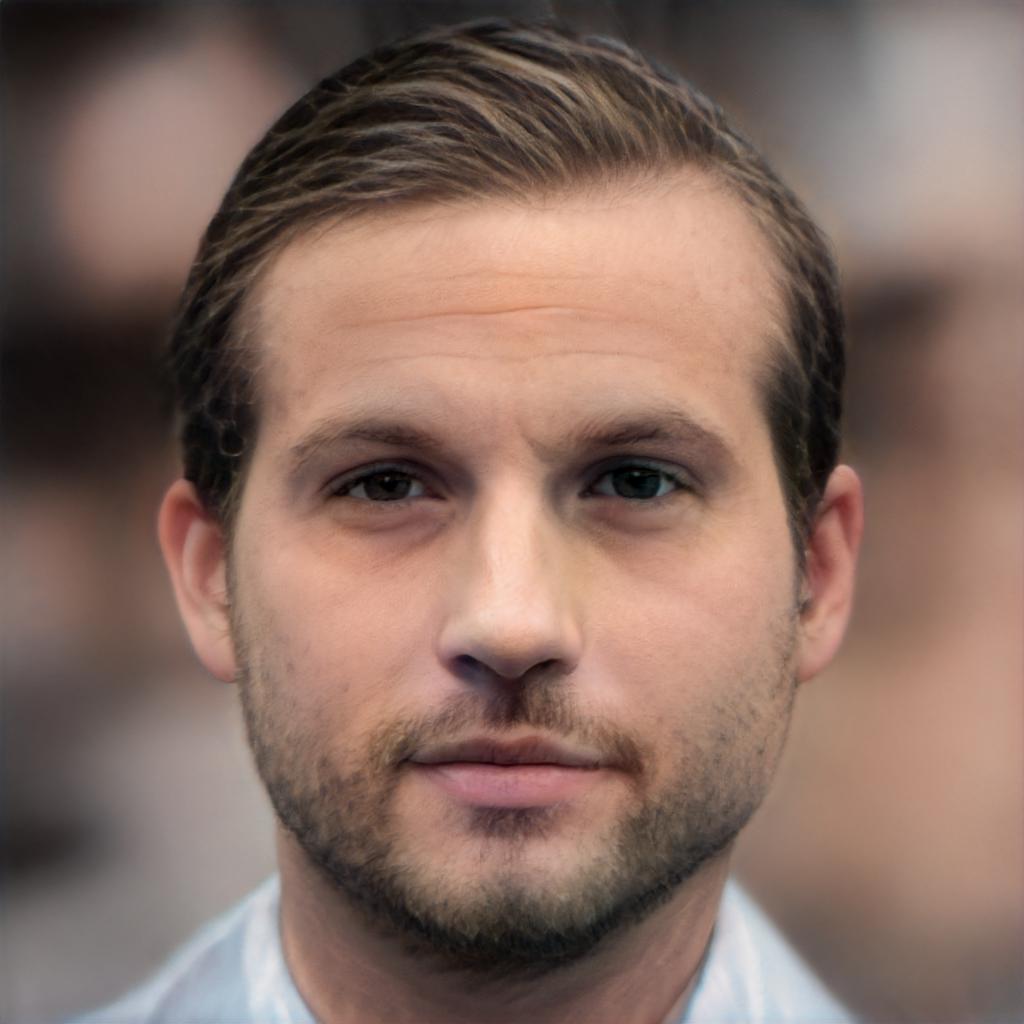} &
        \includegraphics[width=0.15\textwidth]{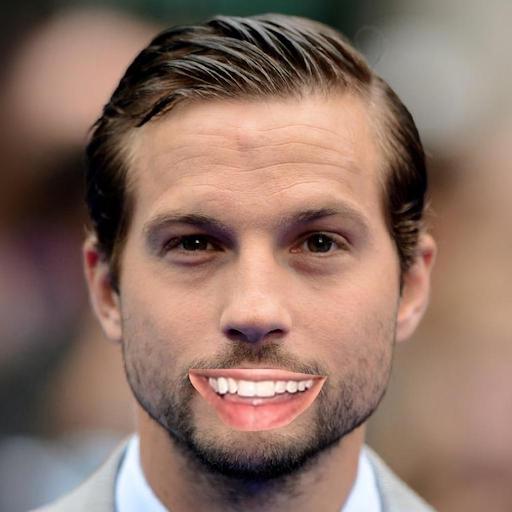} &
        \includegraphics[width=0.15\textwidth]{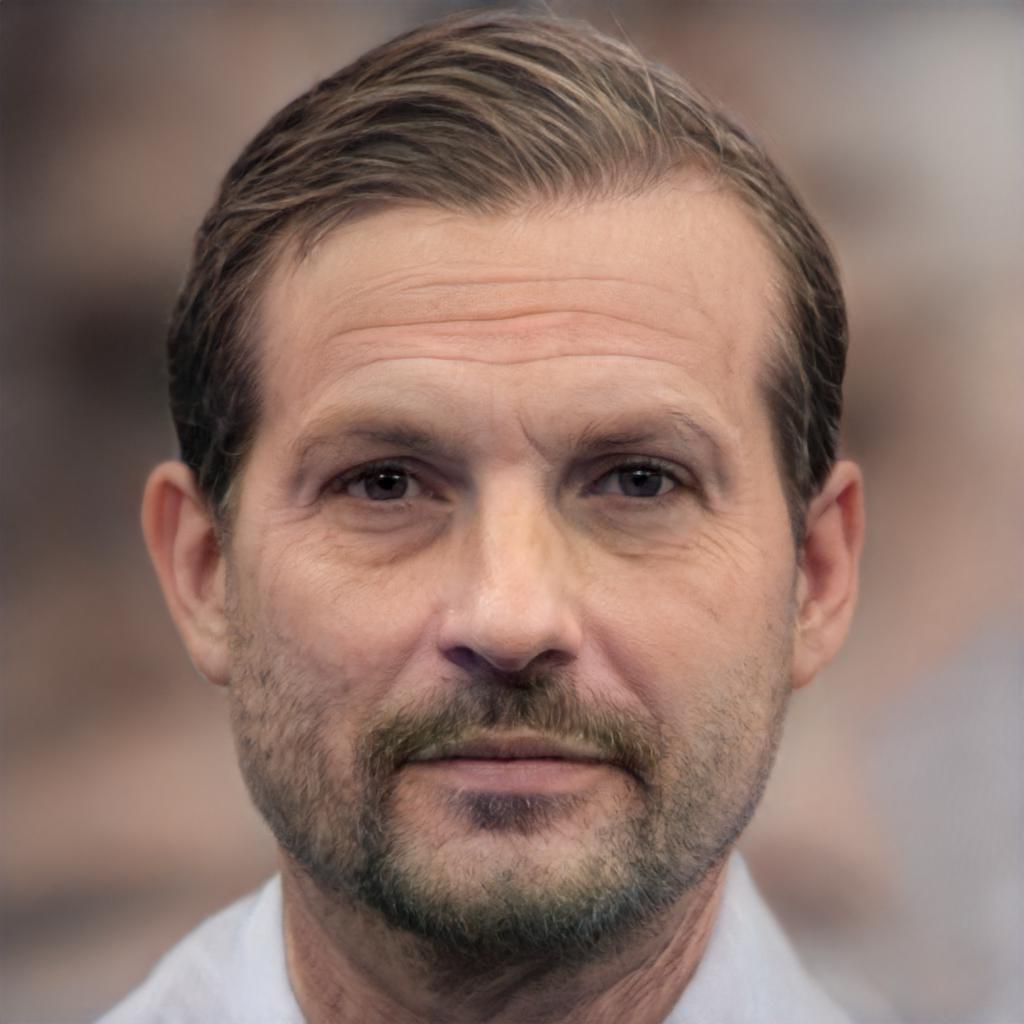} &
        \includegraphics[width=0.15\textwidth]{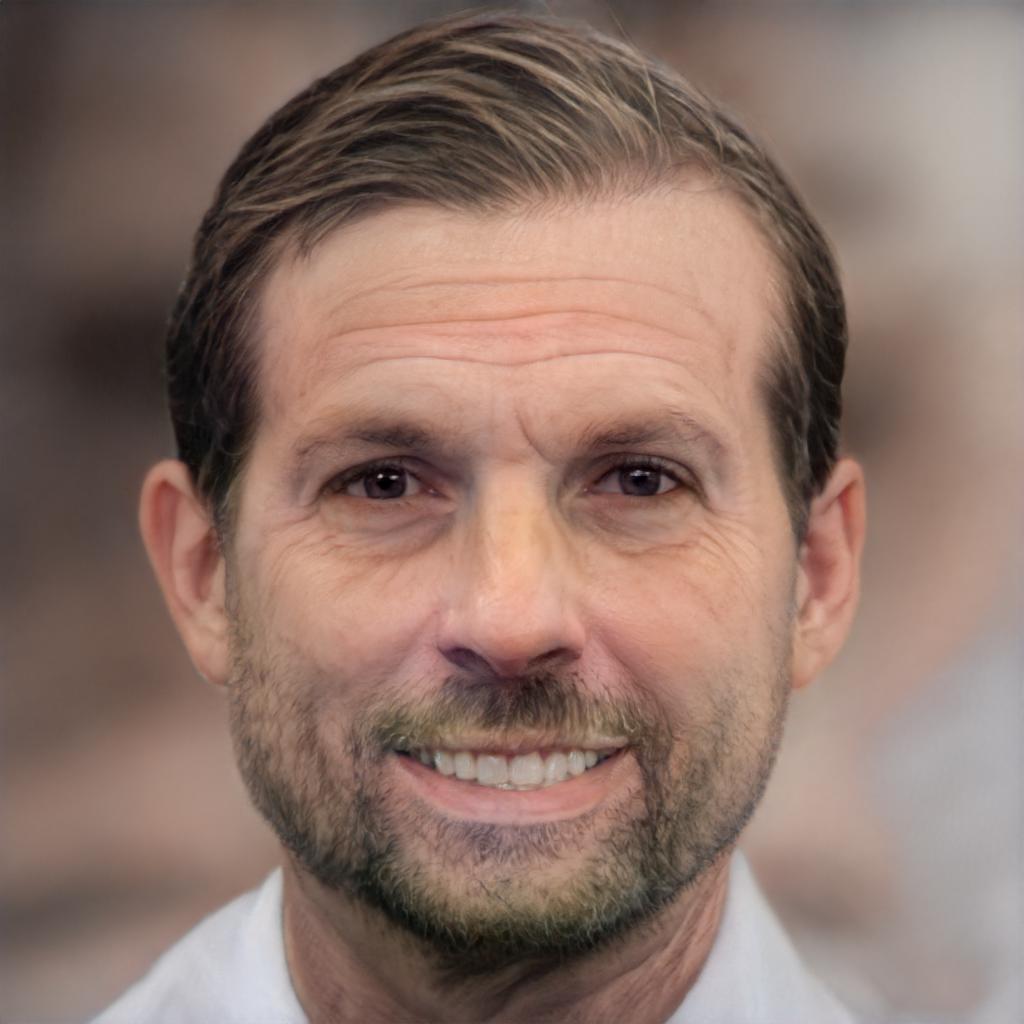}
        \tabularnewline
        \includegraphics[width=0.15\textwidth]{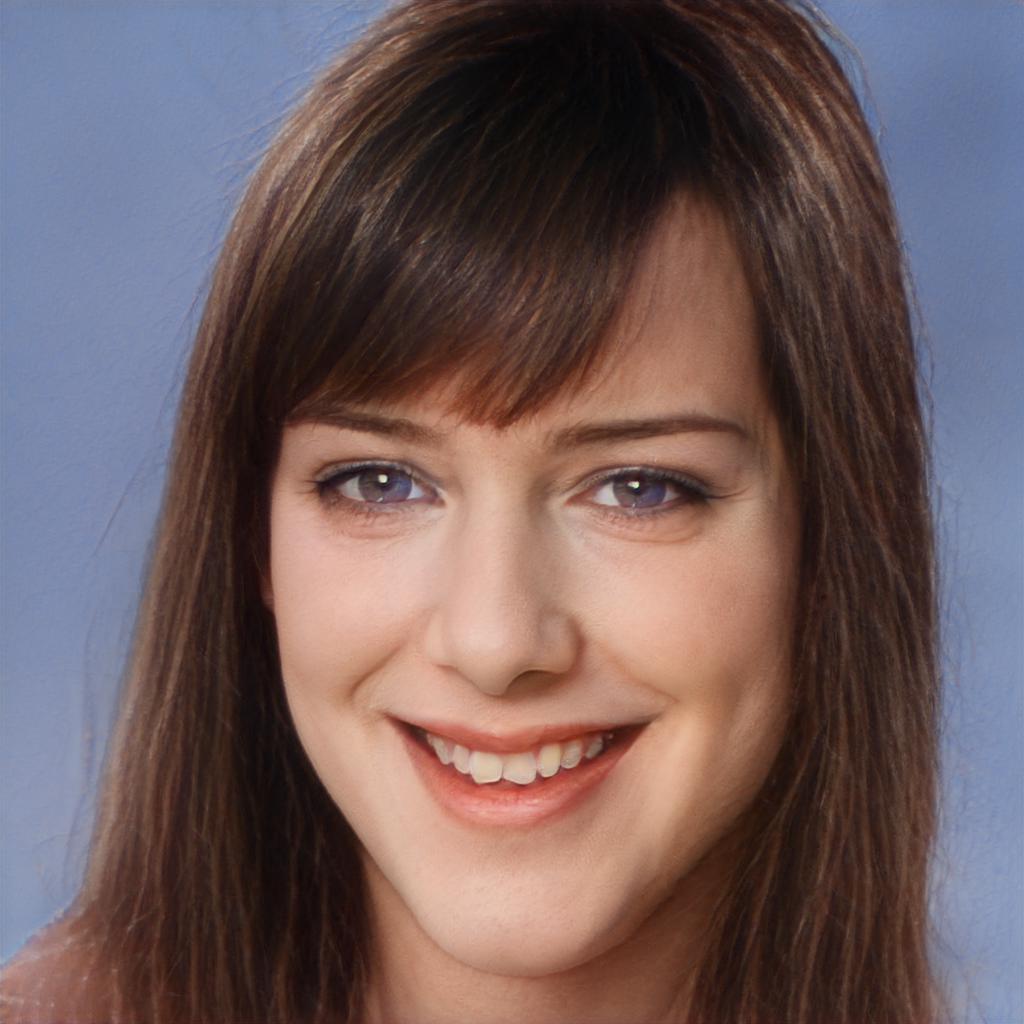} &
        \includegraphics[width=0.15\textwidth]{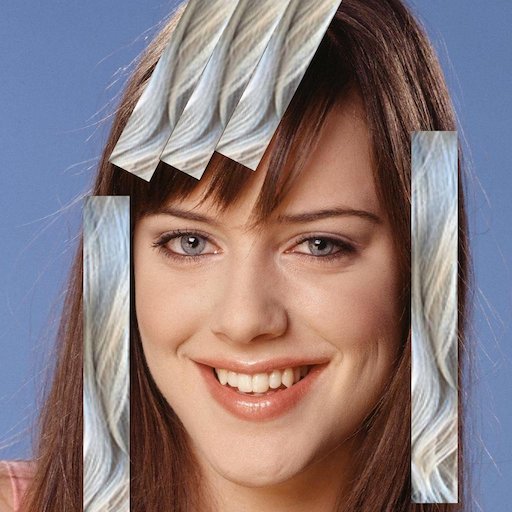} &
        \includegraphics[width=0.15\textwidth]{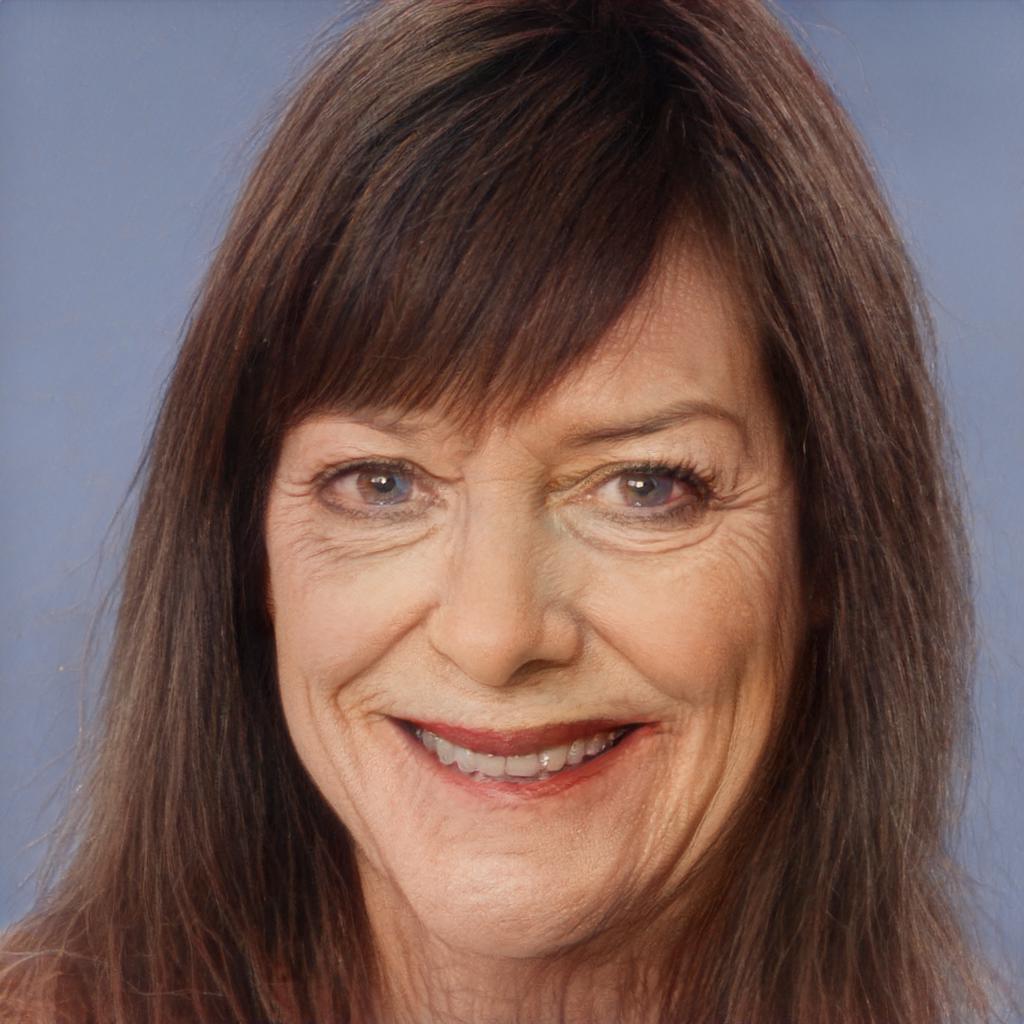} &
        \includegraphics[width=0.15\textwidth]{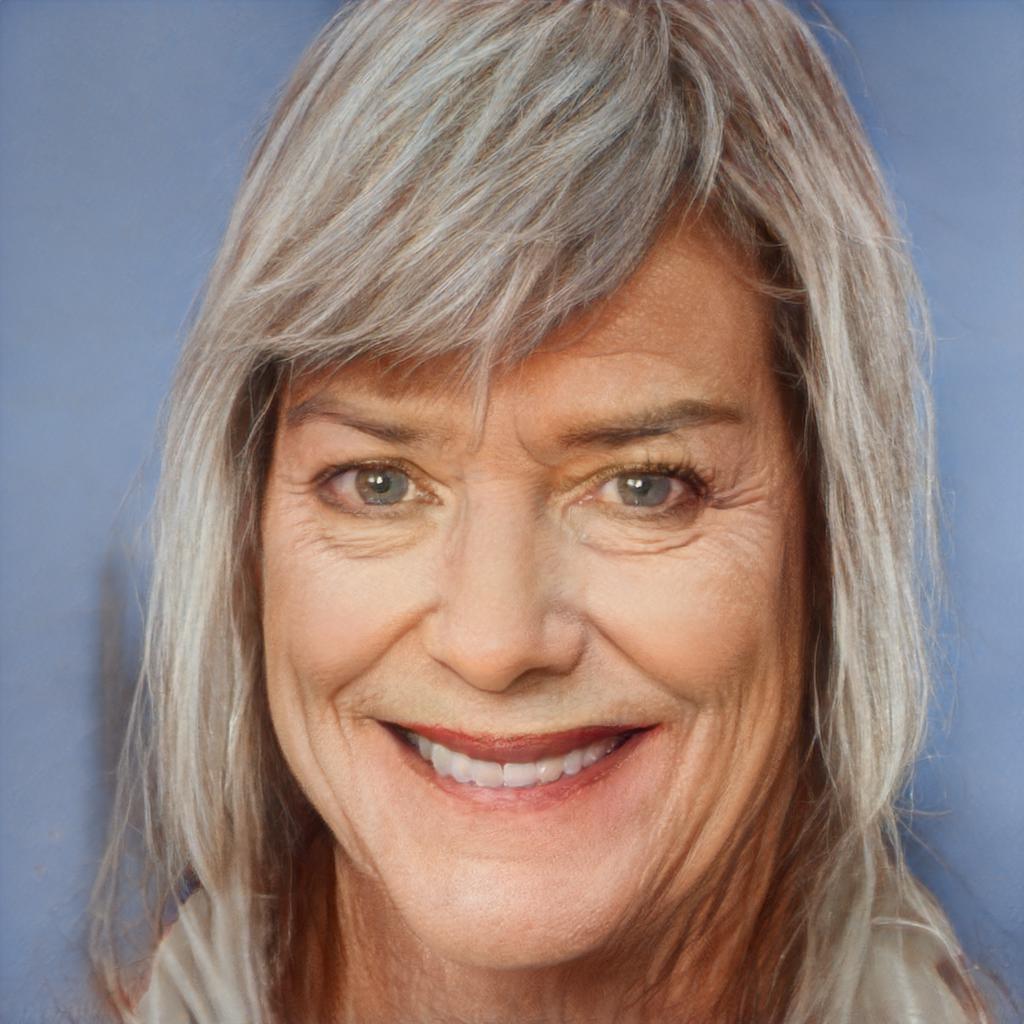}
        \tabularnewline
        \begin{tabular}{@{}c@{}}Inversion \end{tabular} &
        \begin{tabular}{@{}c@{}}Edited \\ Input\end{tabular} &
        \begin{tabular}{@{}c@{}}Inversion \\ Transformed\end{tabular} &
        \begin{tabular}{@{}c@{}}Edited \\ Transformed\end{tabular}
    \end{tabular}
    \setlength{\belowcaptionskip}{-10pt}
    \caption{Additional patch editing results generated using SAM.}
    \label{fig:patch_editing}
\end{figure*}

%% file: figures/appendix/appendix_celebs.tex
\begin{figure*}
    \centering
    \setlength{\tabcolsep}{1pt}
    \includegraphics[width=0.9\textwidth]{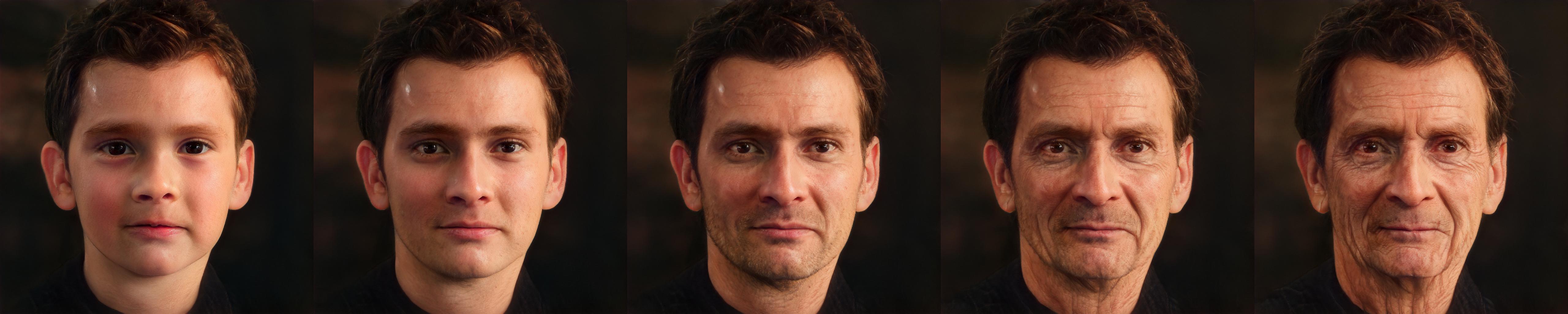} \\
    \vspace{0.1cm}
    \includegraphics[width=0.9\textwidth]{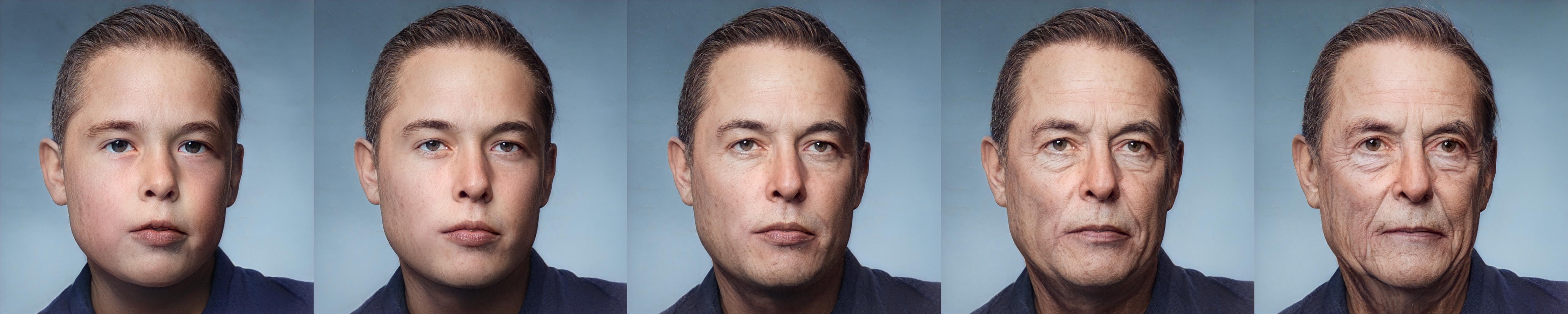} \\
    \vspace{0.1cm}
    \includegraphics[width=0.9\textwidth]{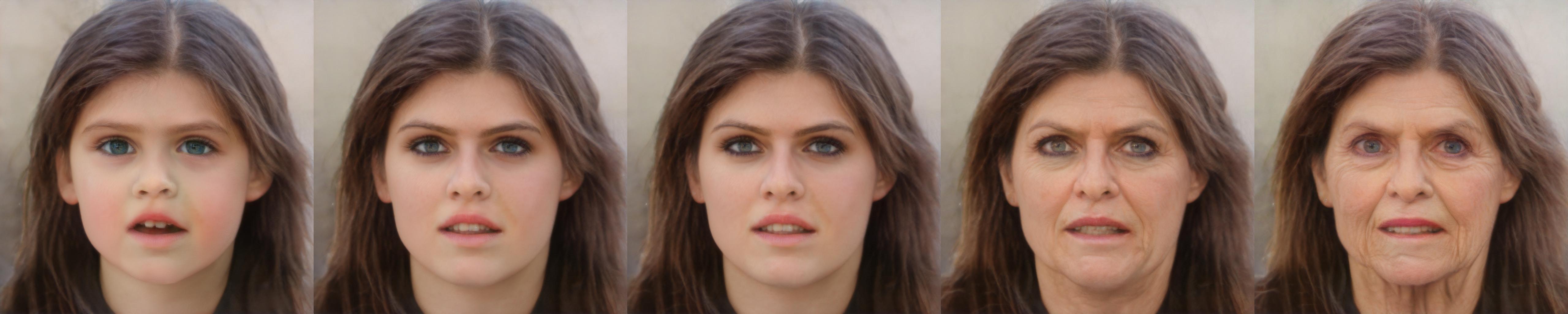} \\
    \vspace{0.1cm}
    \includegraphics[width=0.9\textwidth]{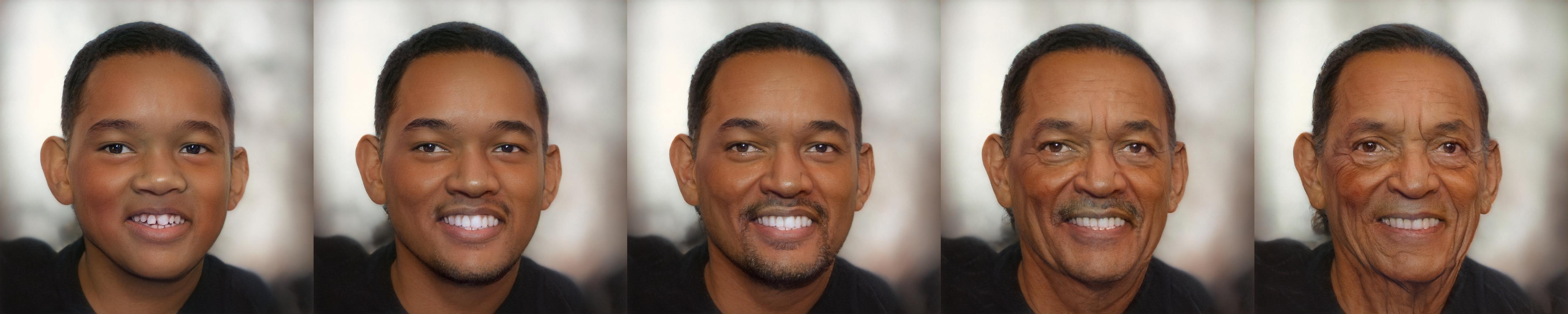} \\ 
    \vspace{0.1cm}
    \includegraphics[width=0.9\textwidth]{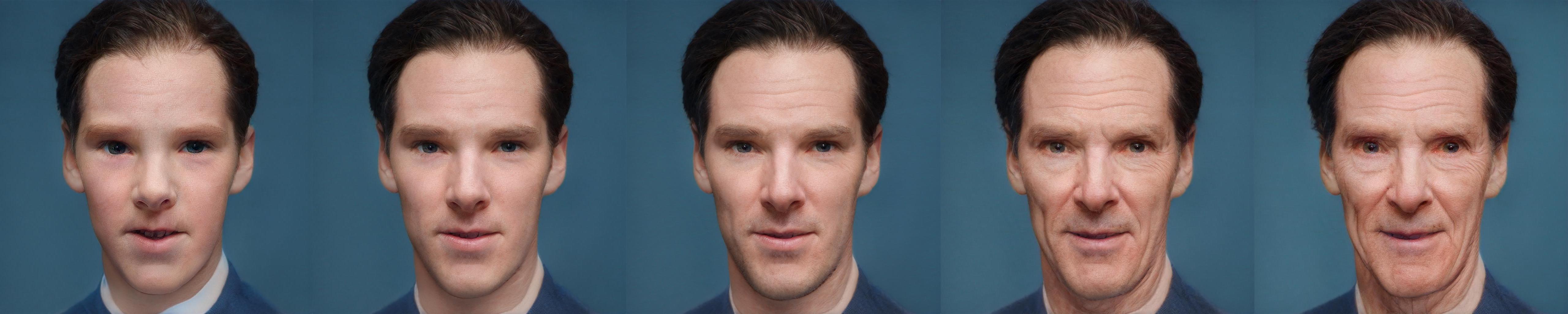} \\
    \vspace{0.1cm}
    \includegraphics[width=0.9\textwidth]{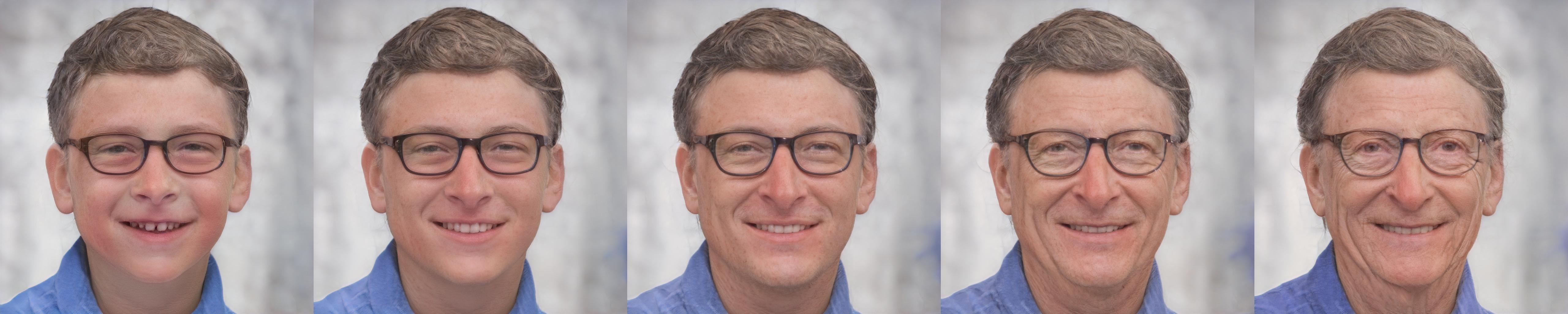}
    \setlength{\belowcaptionskip}{-10pt}
    \caption{Additional age transformation results generated using SAM.}
    \label{fig:appendix_celebs}
\end{figure*}

%% file: figures/appendix/appendix_full_lifespan.tex
\begin{figure*}
    \centering
    \setlength{\belowcaptionskip}{-2.5pt}
    \setlength{\tabcolsep}{1pt}
    \begin{tabular}{cccccc}
        \includegraphics[width=0.135\textwidth]{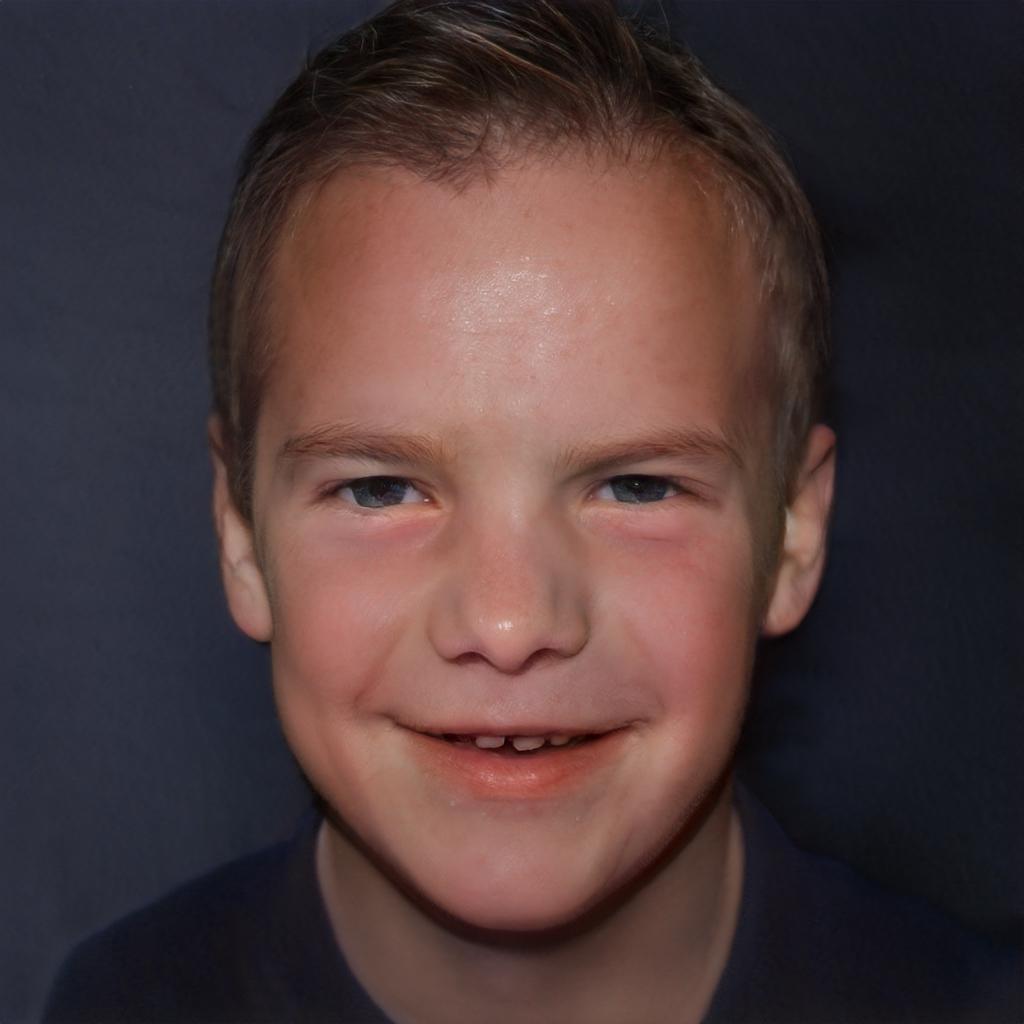} &
        \includegraphics[width=0.135\textwidth]{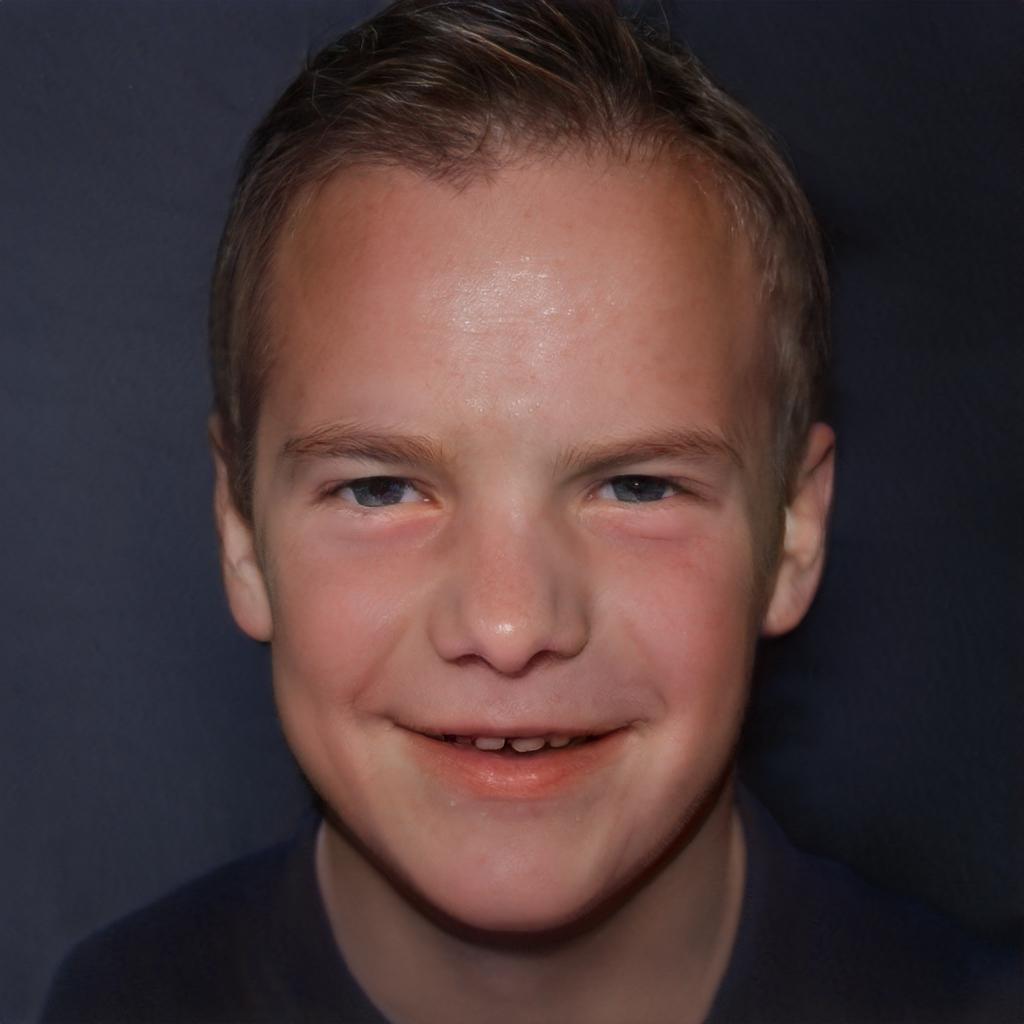} &
        \includegraphics[width=0.135\textwidth]{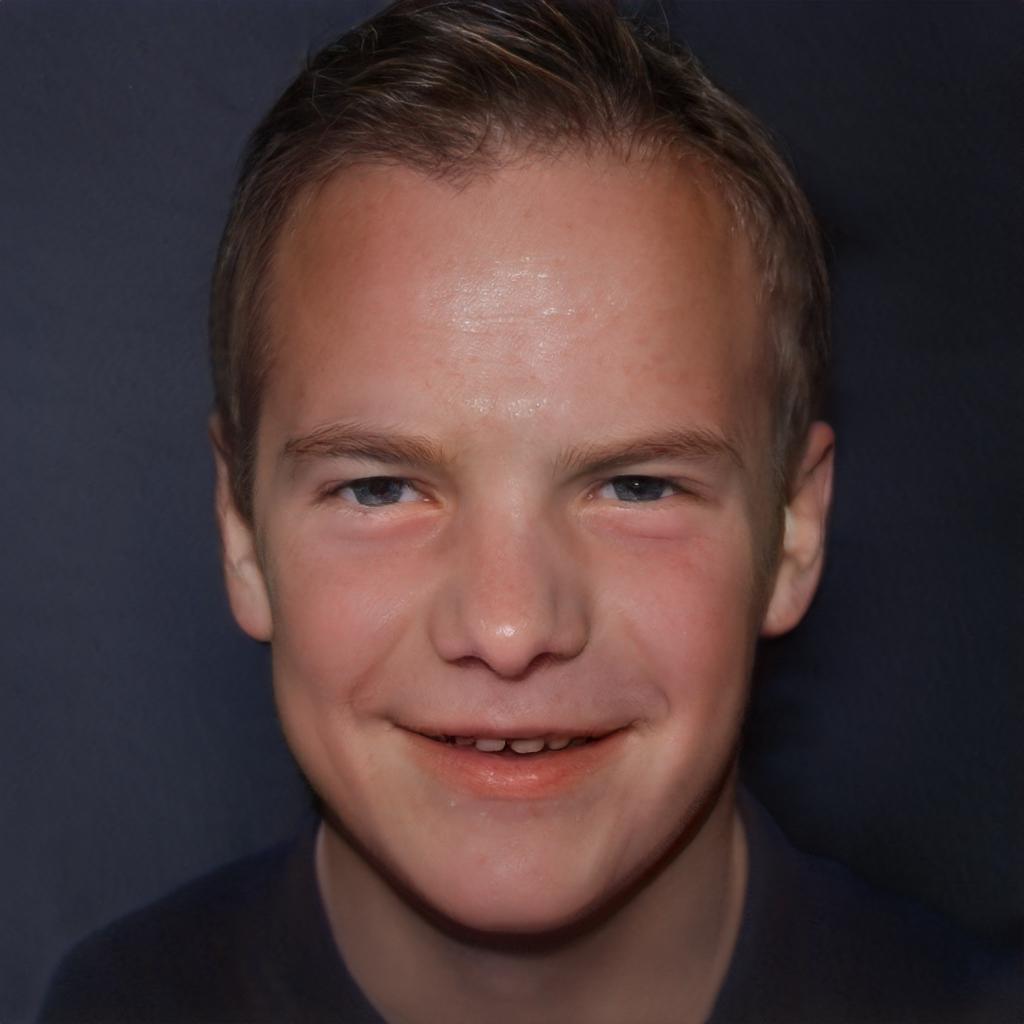} &
        \includegraphics[width=0.135\textwidth]{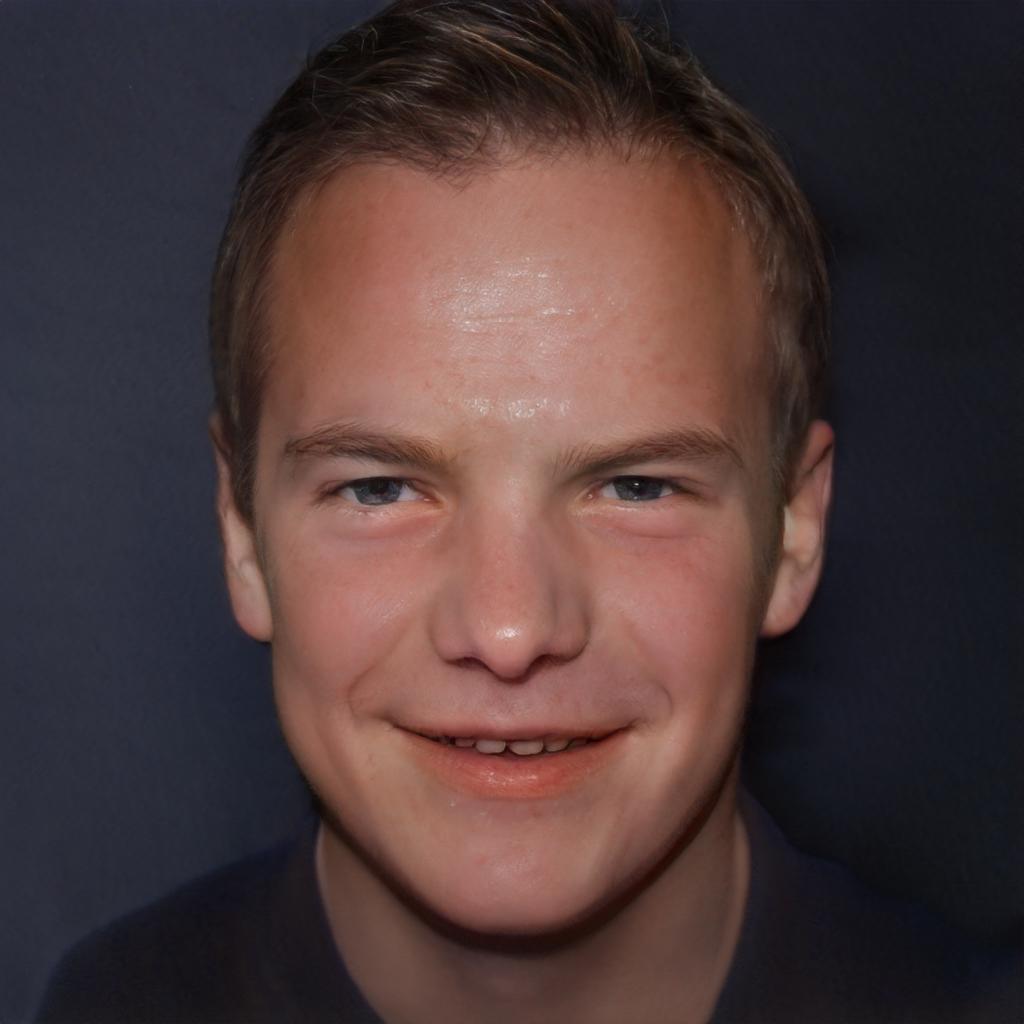} &
        \includegraphics[width=0.135\textwidth]{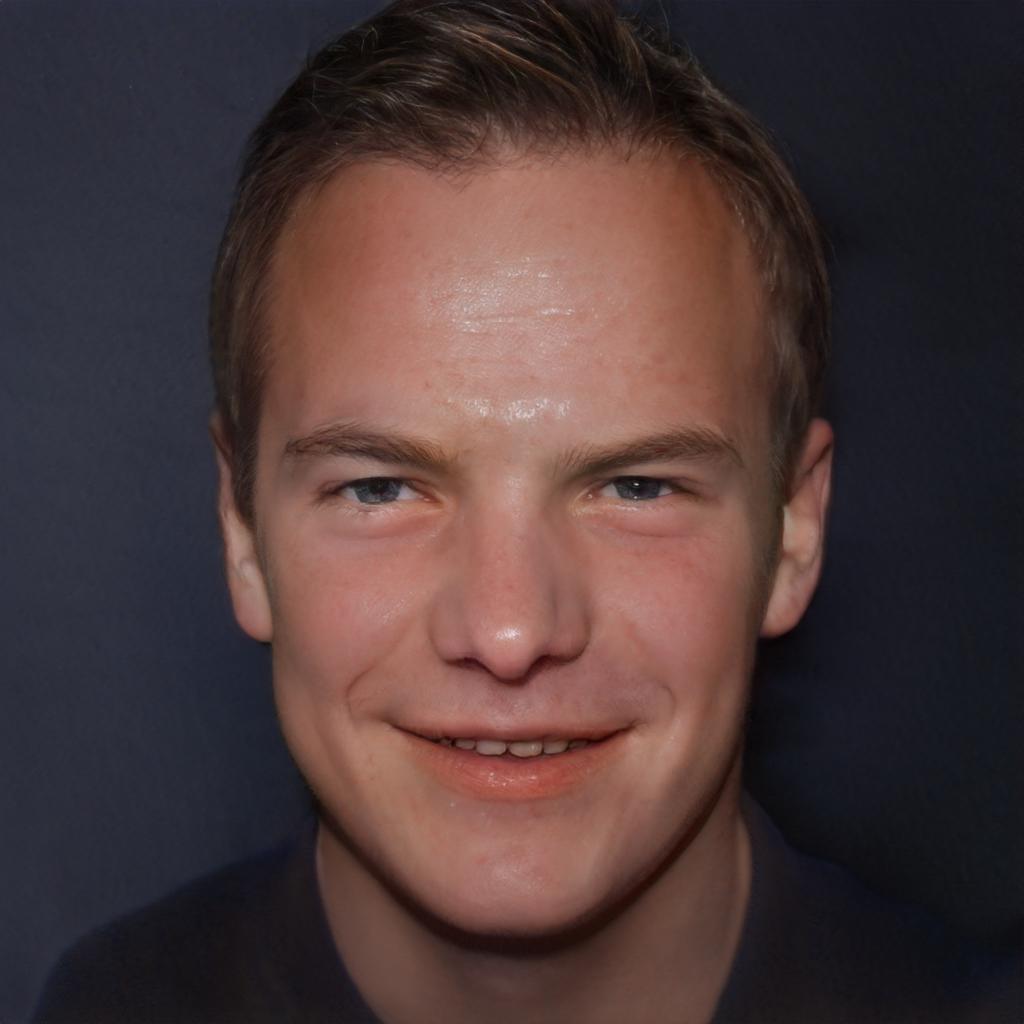} &
        \includegraphics[width=0.135\textwidth]{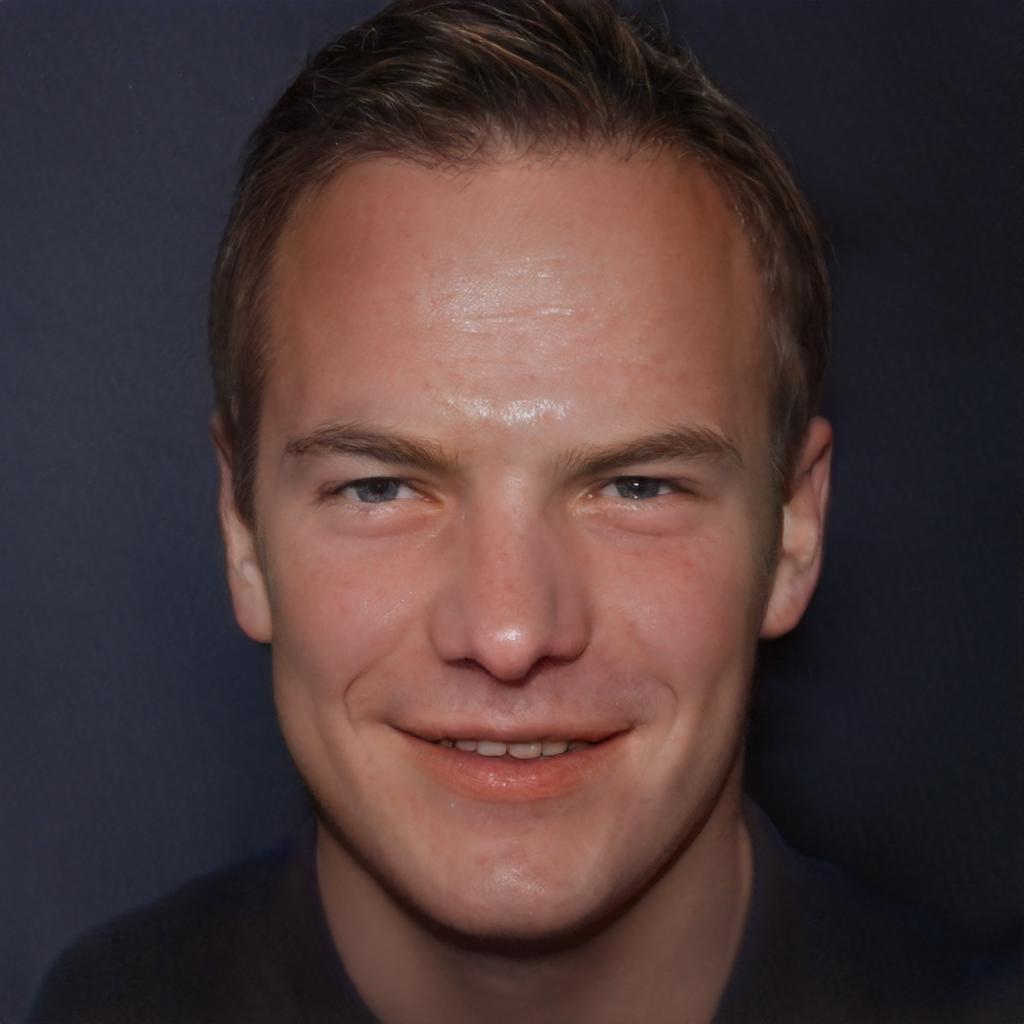} 
        \tabularnewline
        \includegraphics[width=0.135\textwidth]{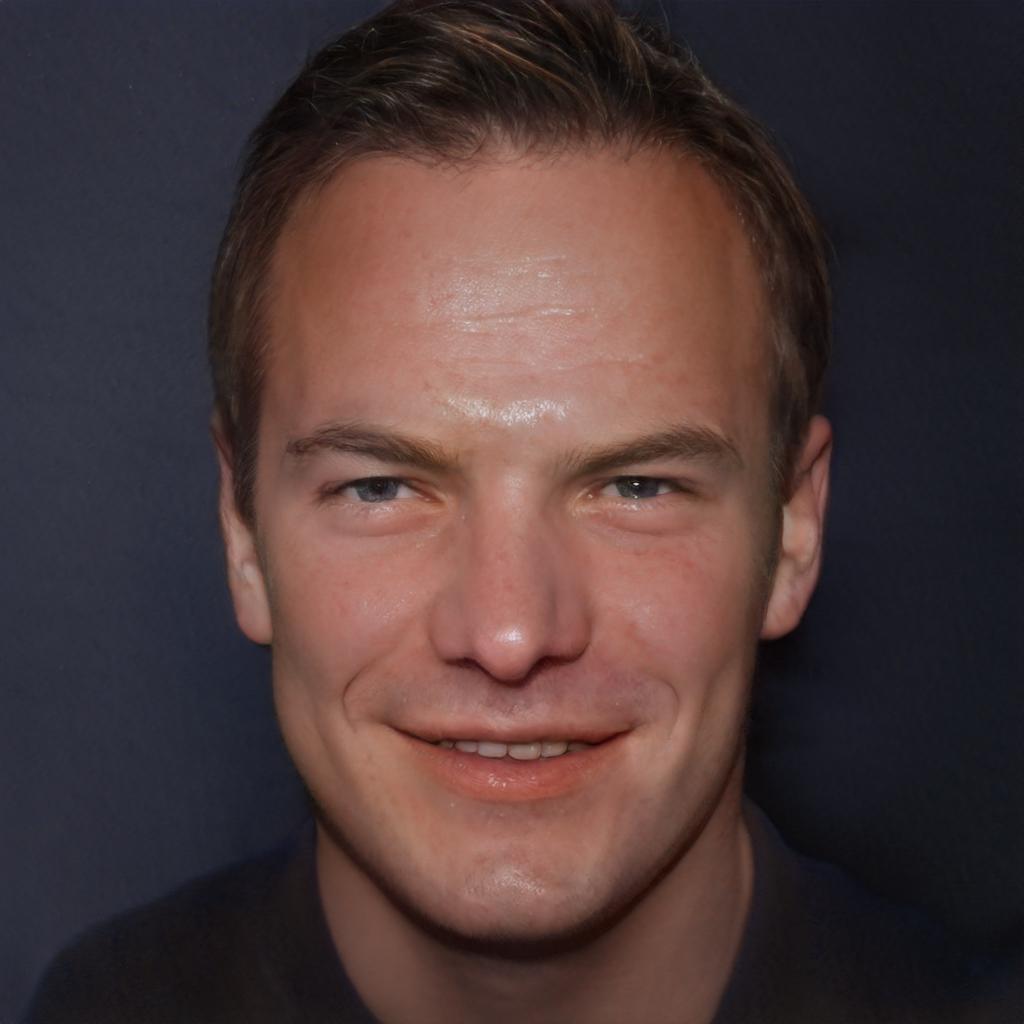} &
        \includegraphics[width=0.135\textwidth]{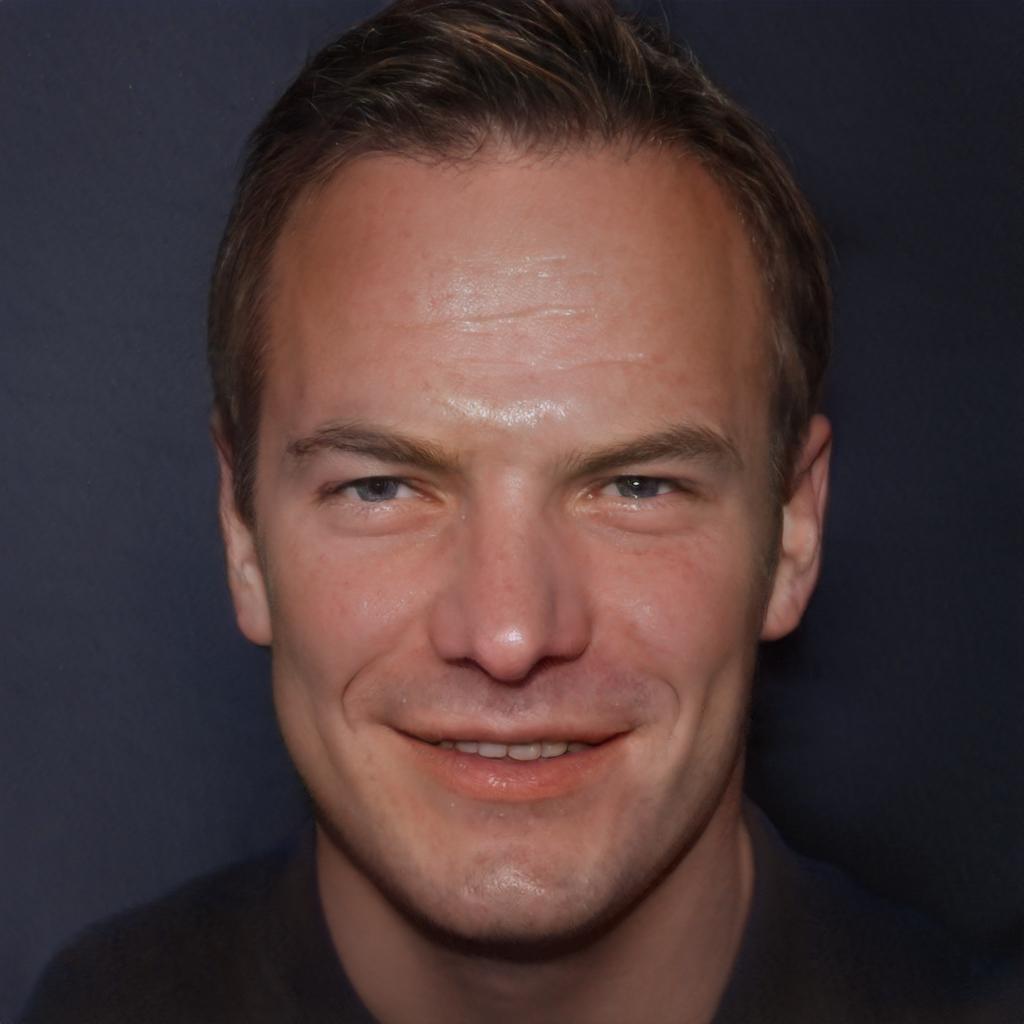} &
        \includegraphics[width=0.135\textwidth]{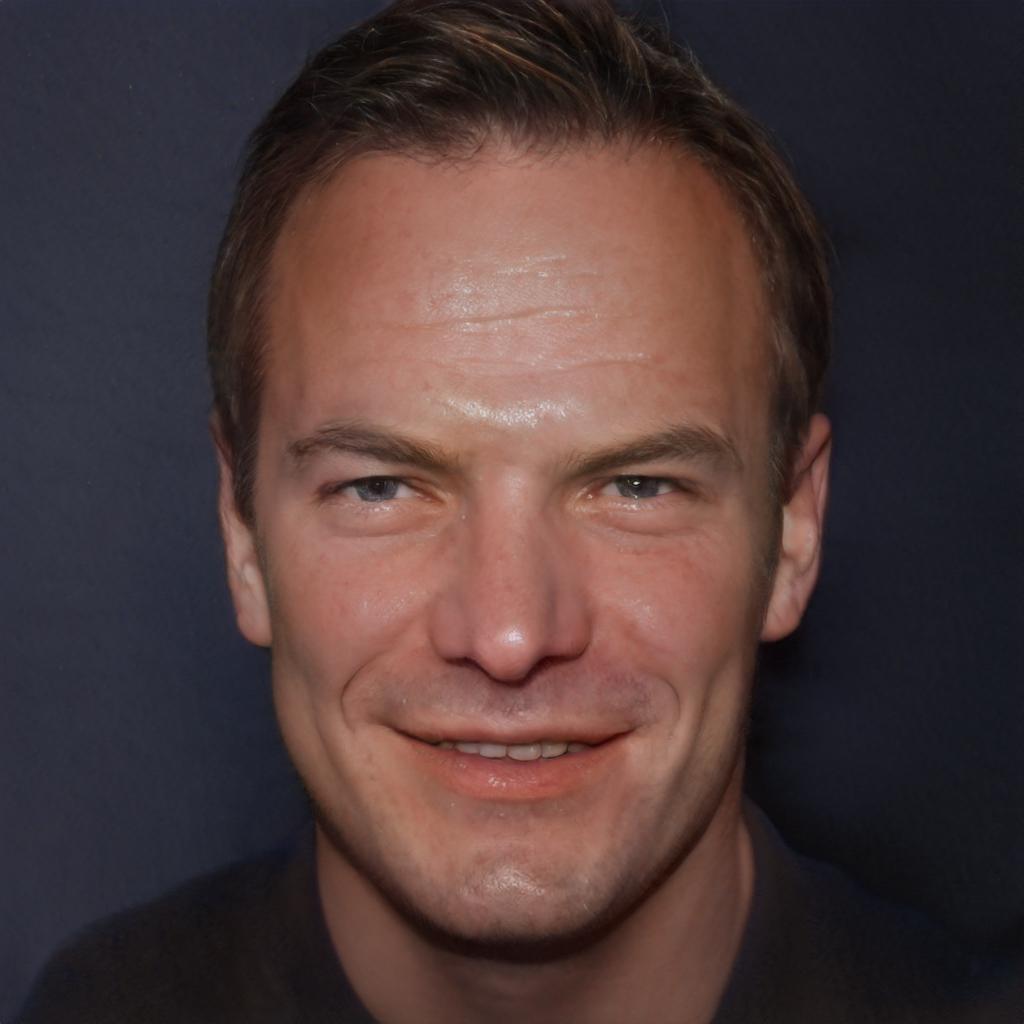} &
        \includegraphics[width=0.135\textwidth]{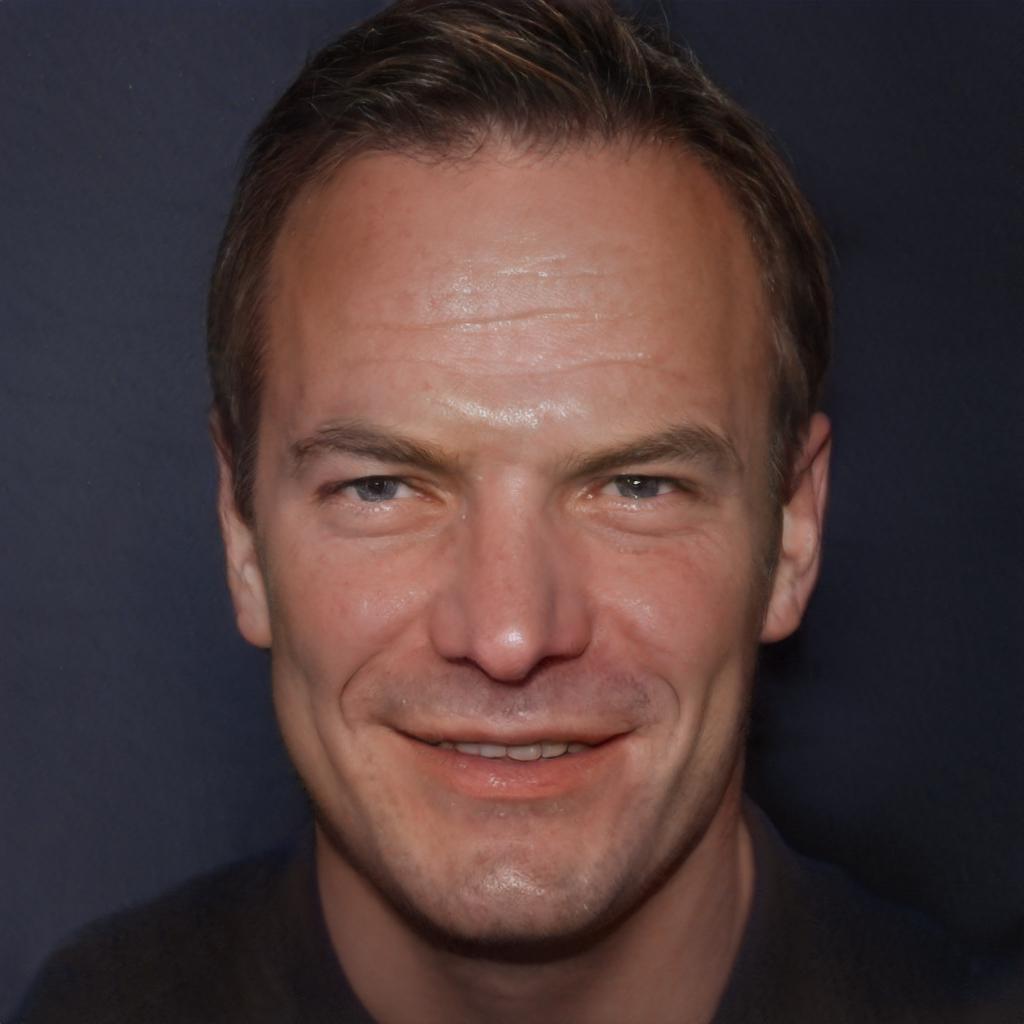} &
        \includegraphics[width=0.135\textwidth]{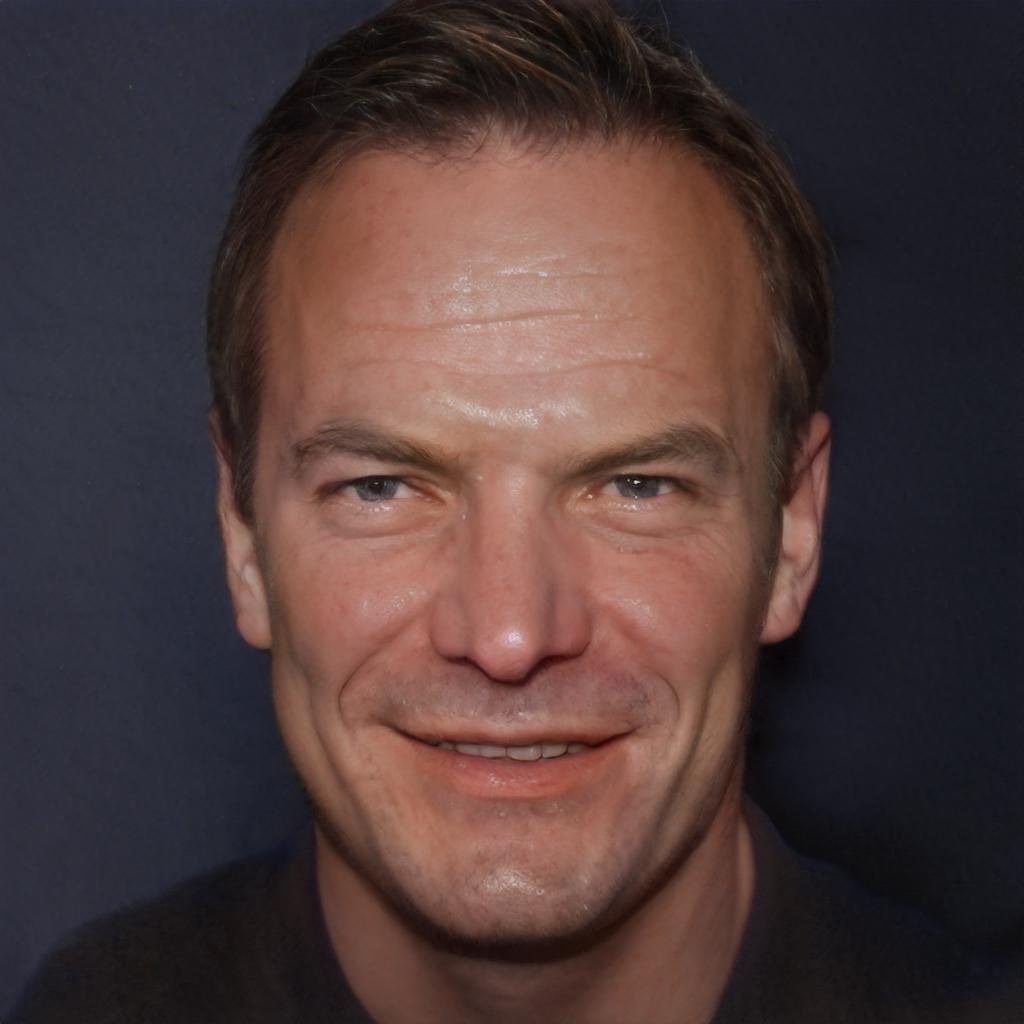} &
        \includegraphics[width=0.135\textwidth]{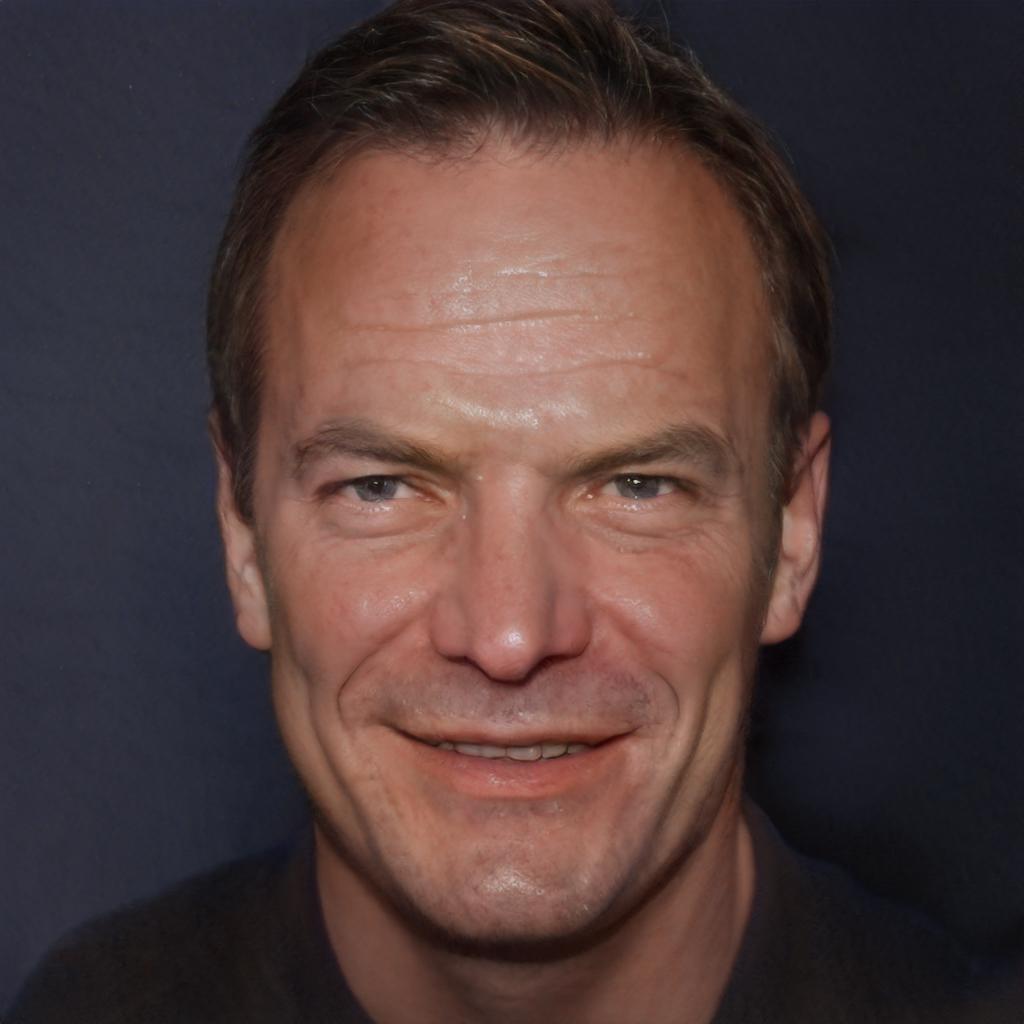} 
        \tabularnewline
        \includegraphics[width=0.135\textwidth]{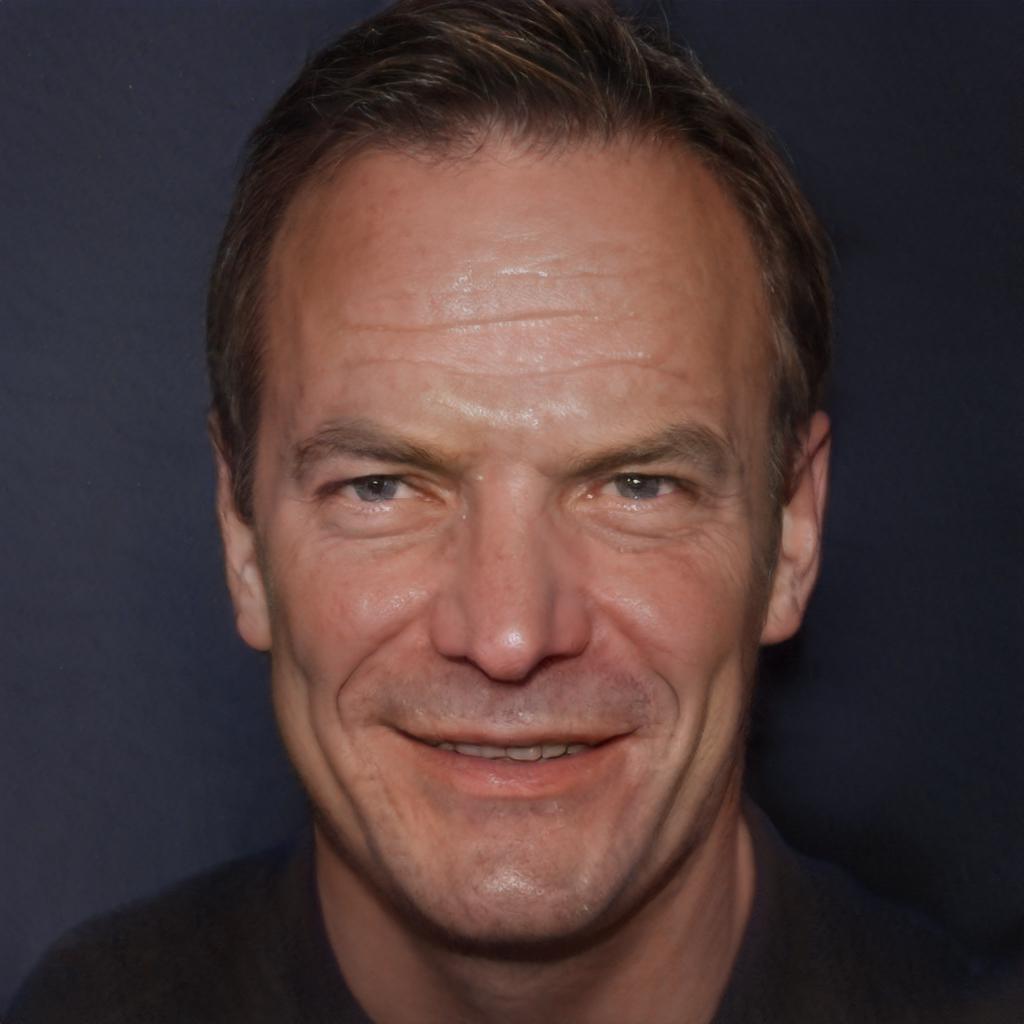} &
        \includegraphics[width=0.135\textwidth]{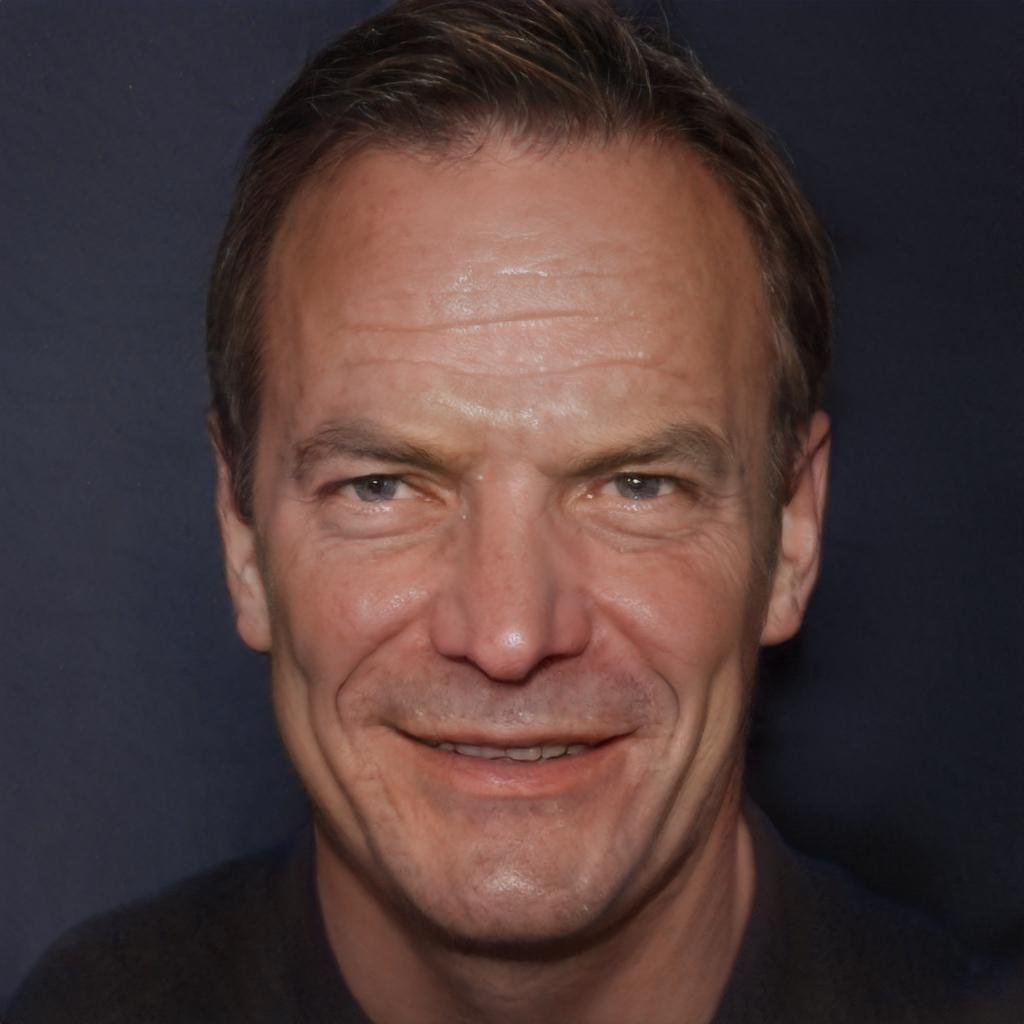} &
        \includegraphics[width=0.135\textwidth]{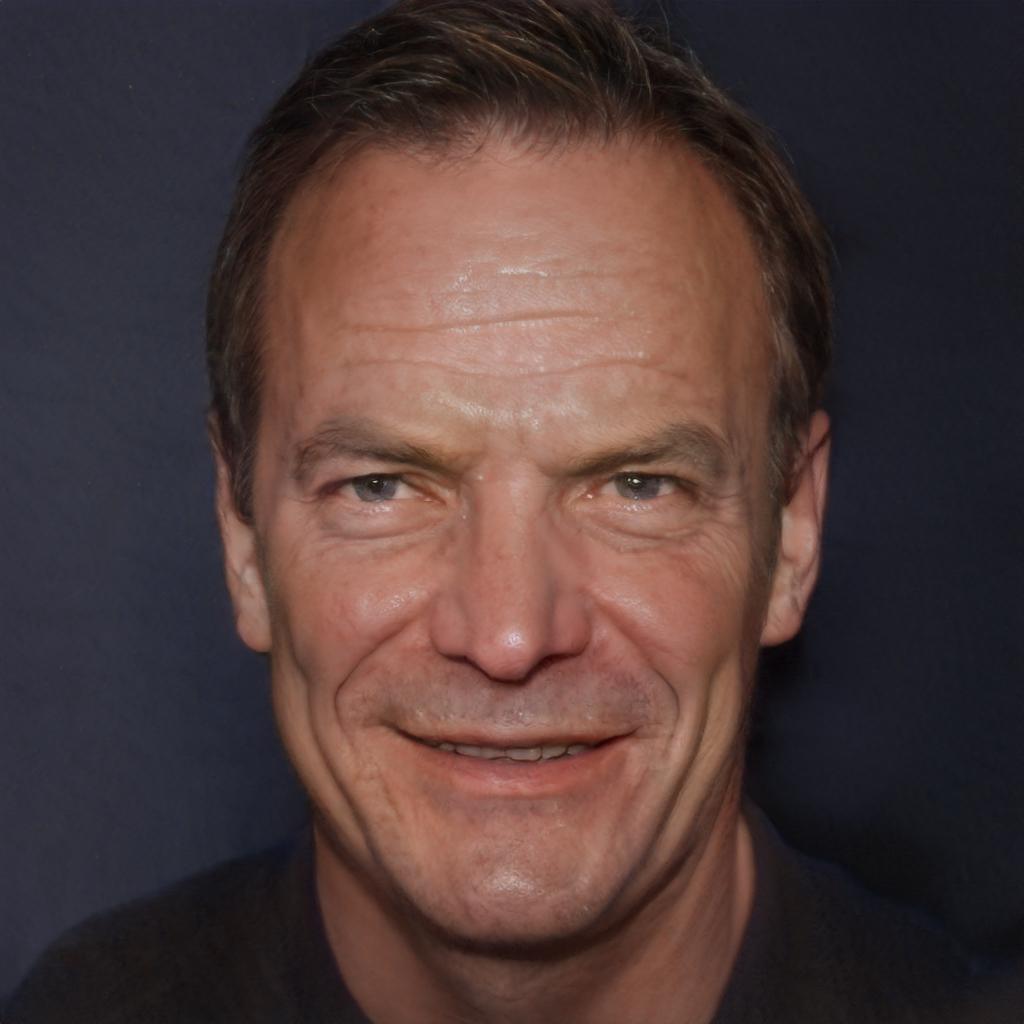} &
        \includegraphics[width=0.135\textwidth]{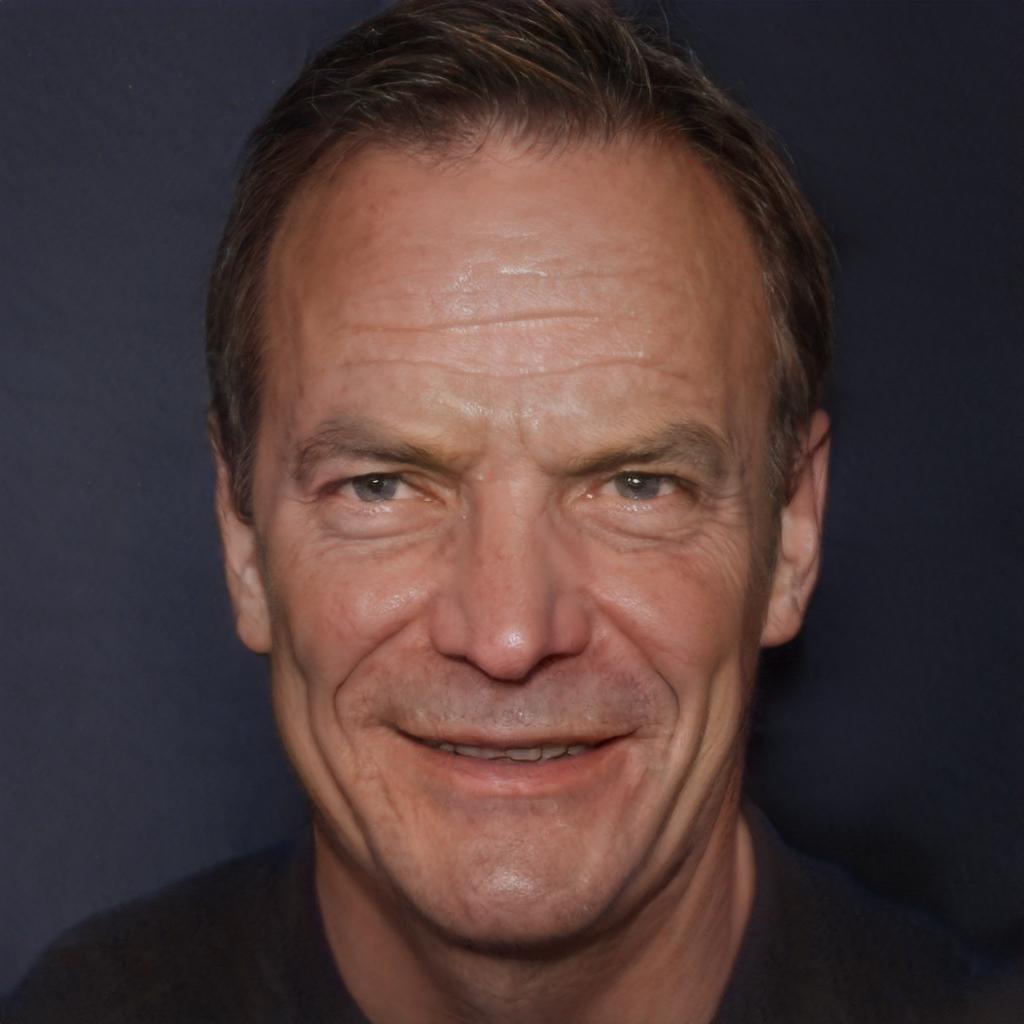} &
        \includegraphics[width=0.135\textwidth]{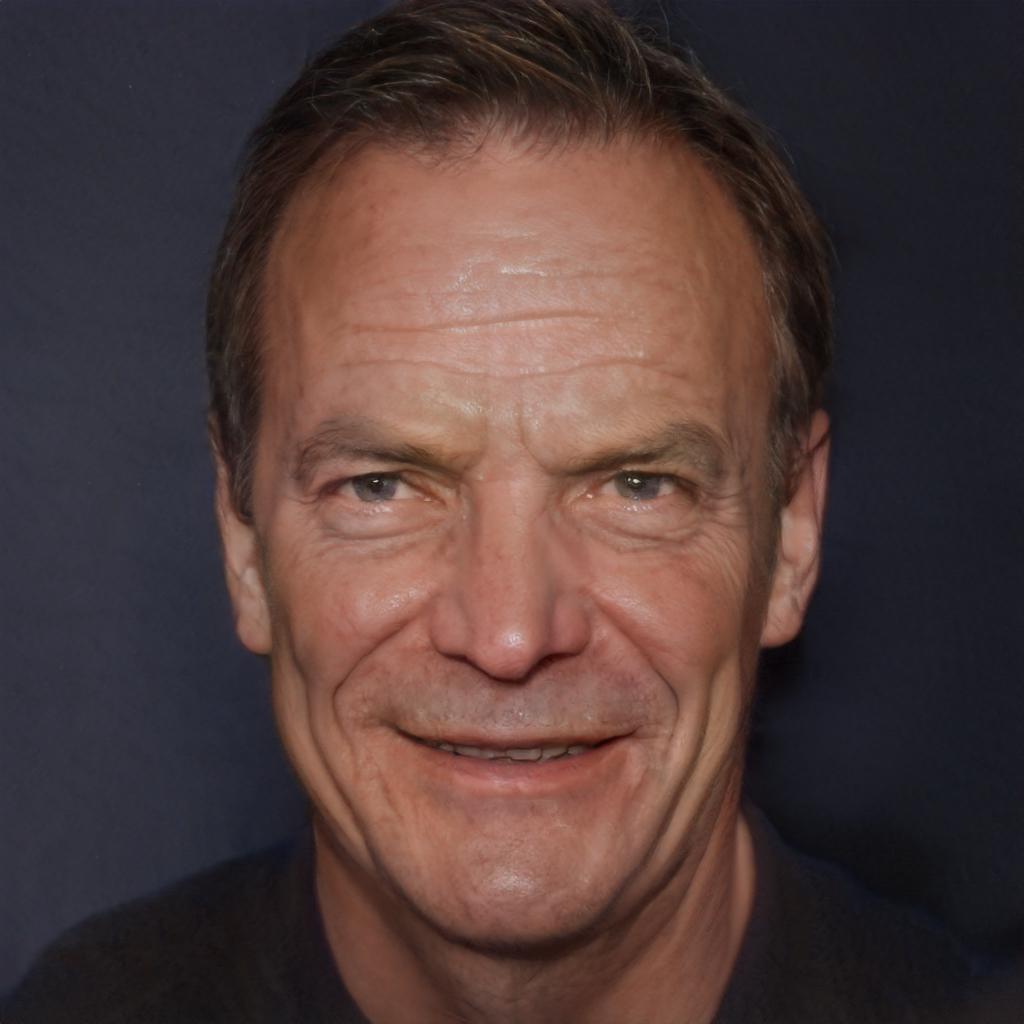} &
        \includegraphics[width=0.135\textwidth]{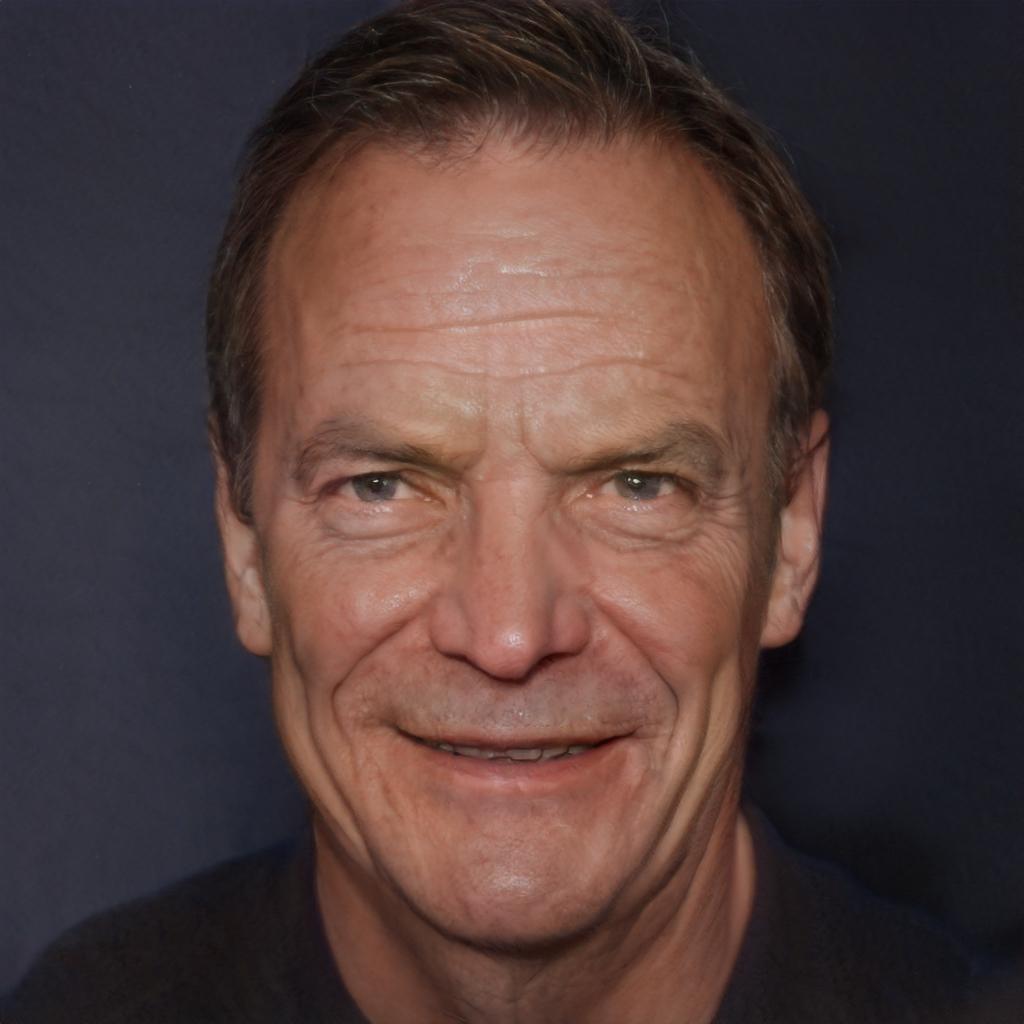} 
        \tabularnewline
        \includegraphics[width=0.135\textwidth]{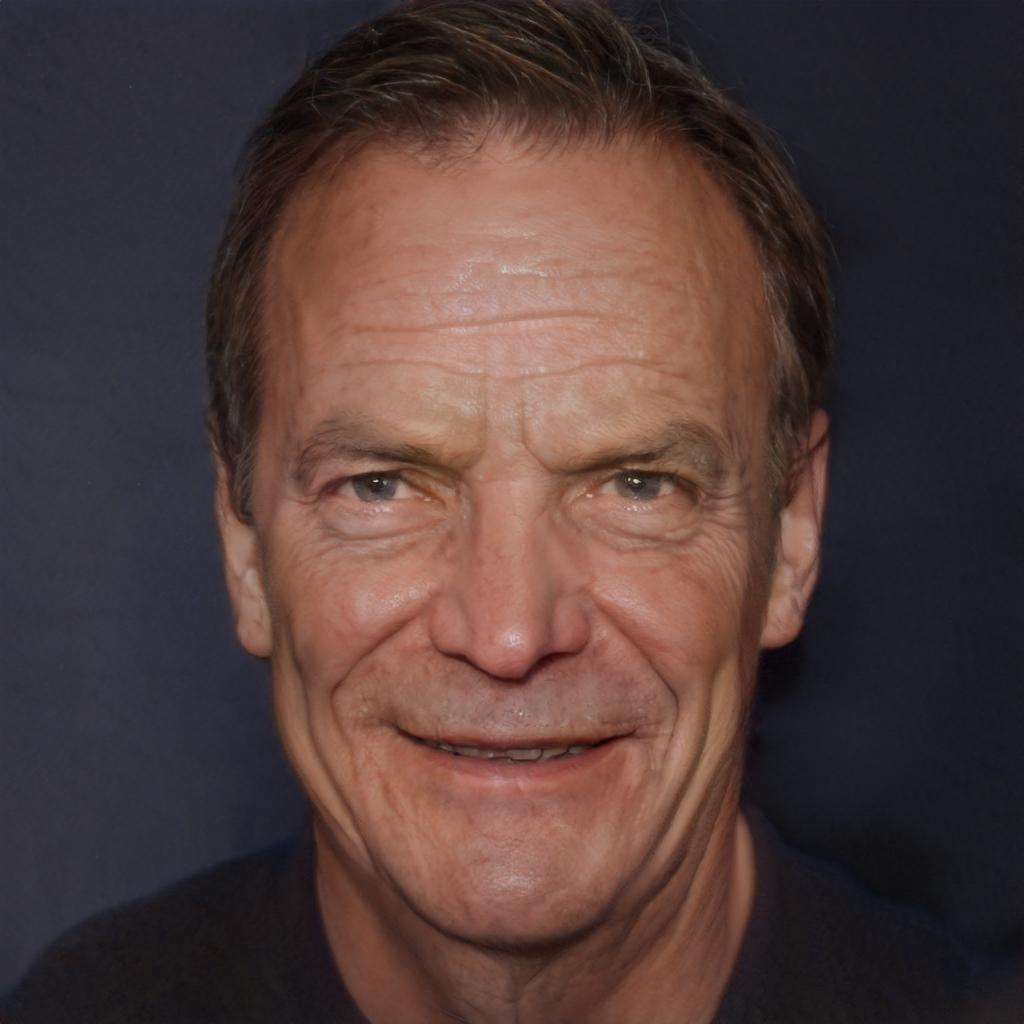} &
        \includegraphics[width=0.135\textwidth]{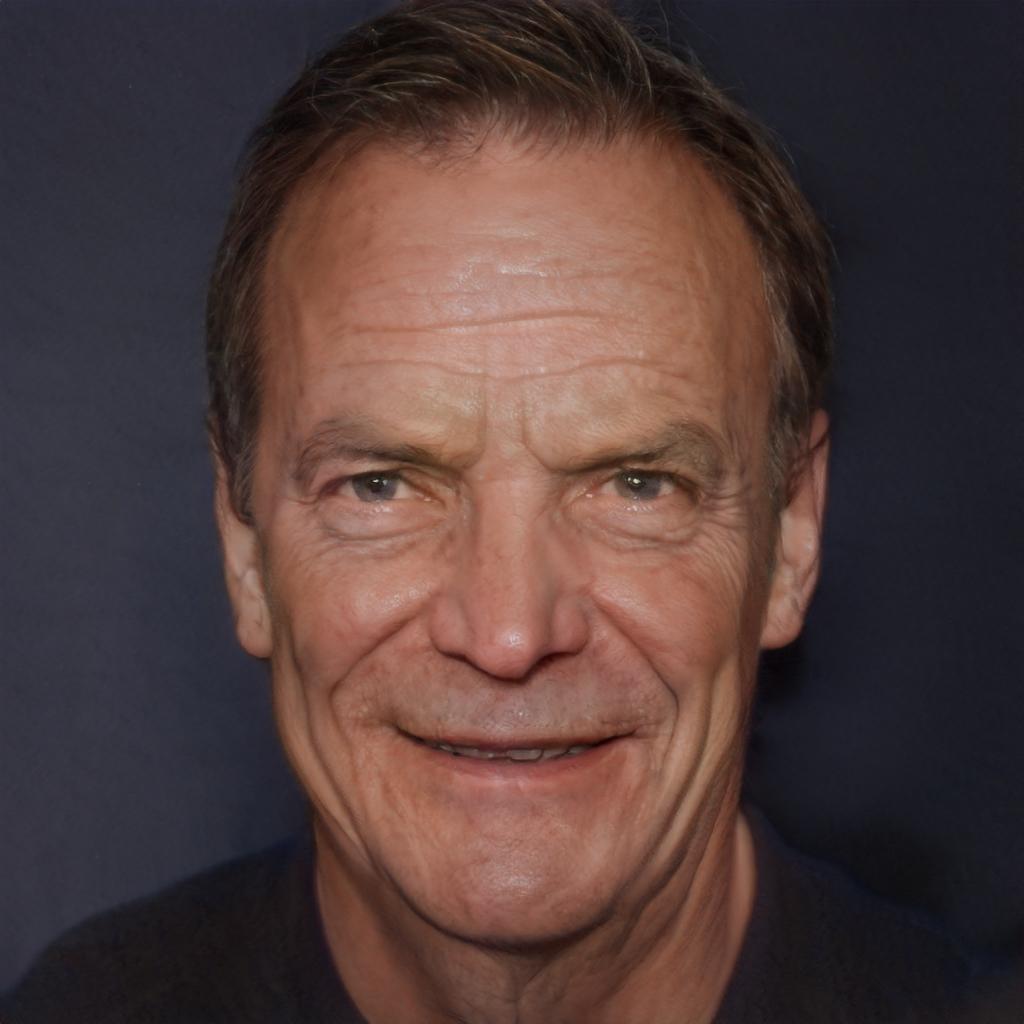} &
        \includegraphics[width=0.135\textwidth]{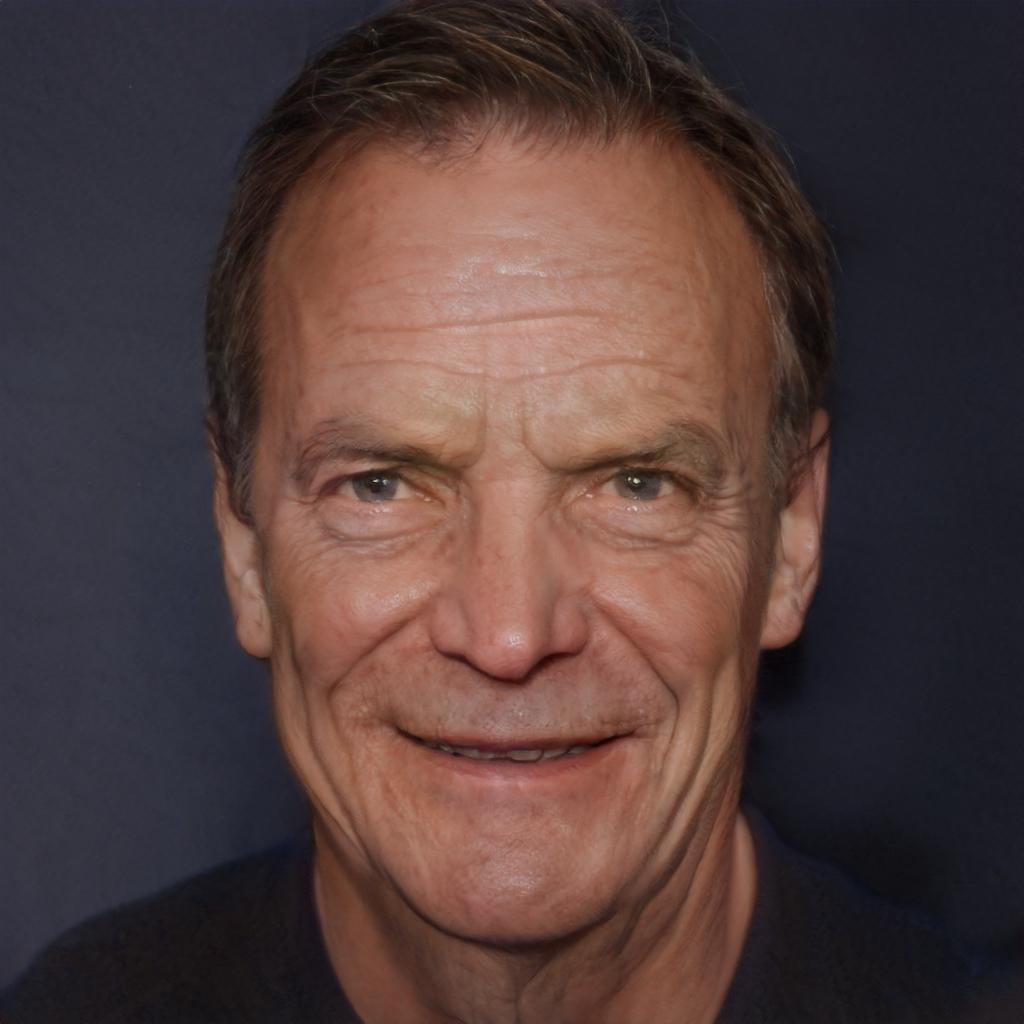} &
        \includegraphics[width=0.135\textwidth]{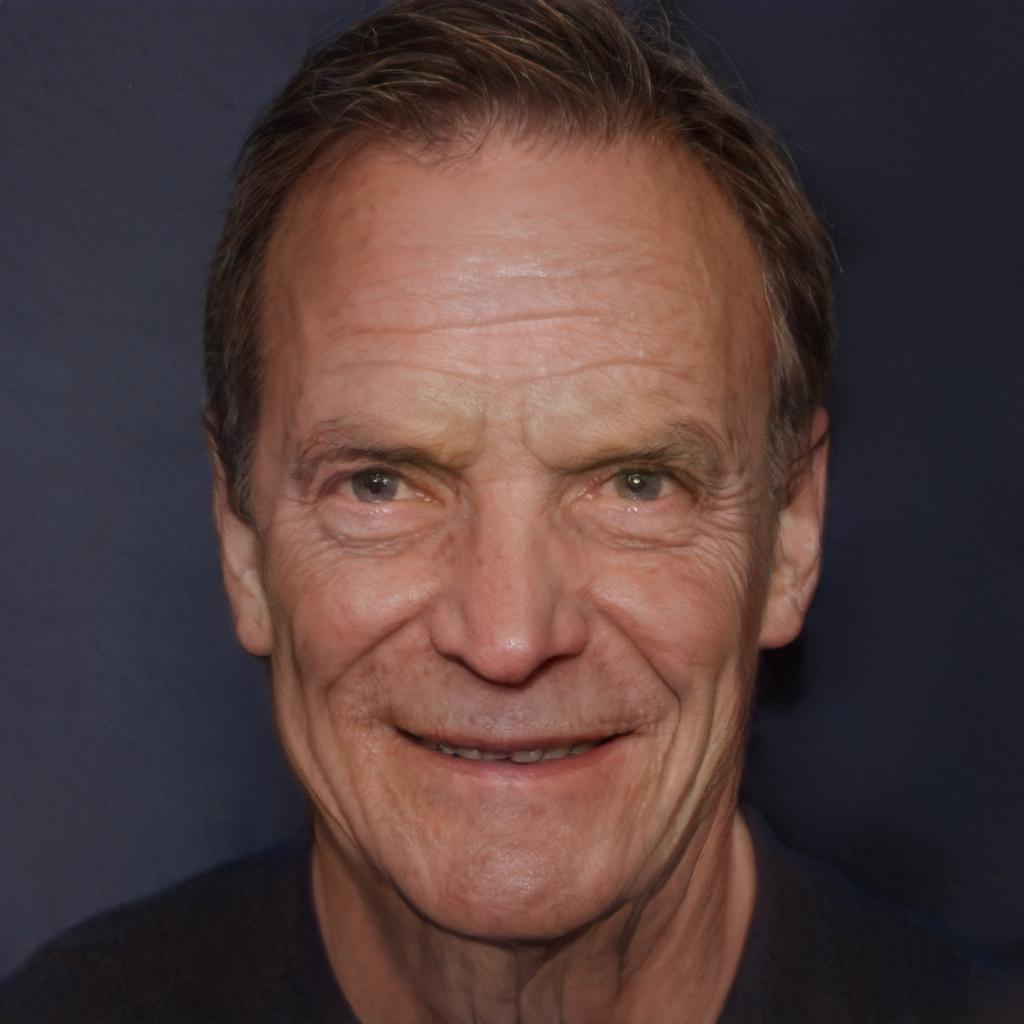} &
        \includegraphics[width=0.135\textwidth]{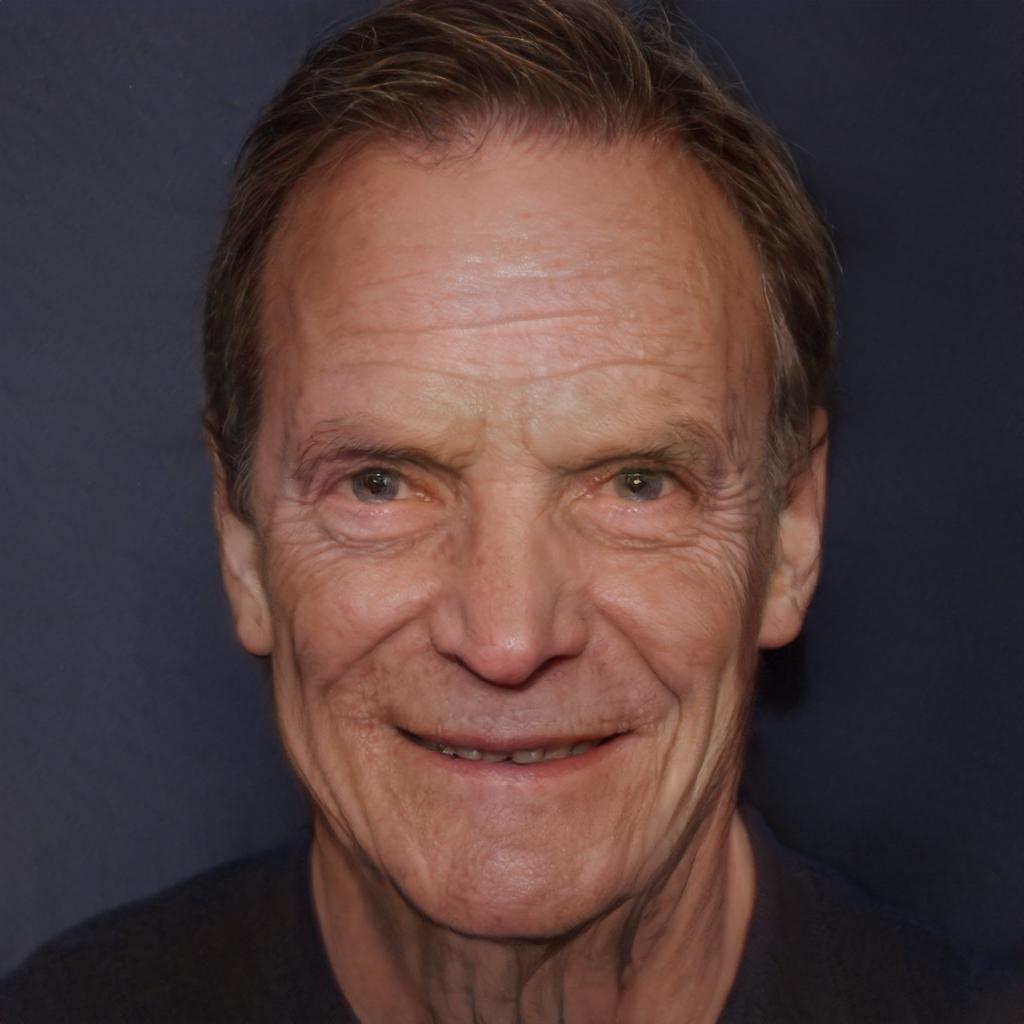} &
        \includegraphics[width=0.135\textwidth]{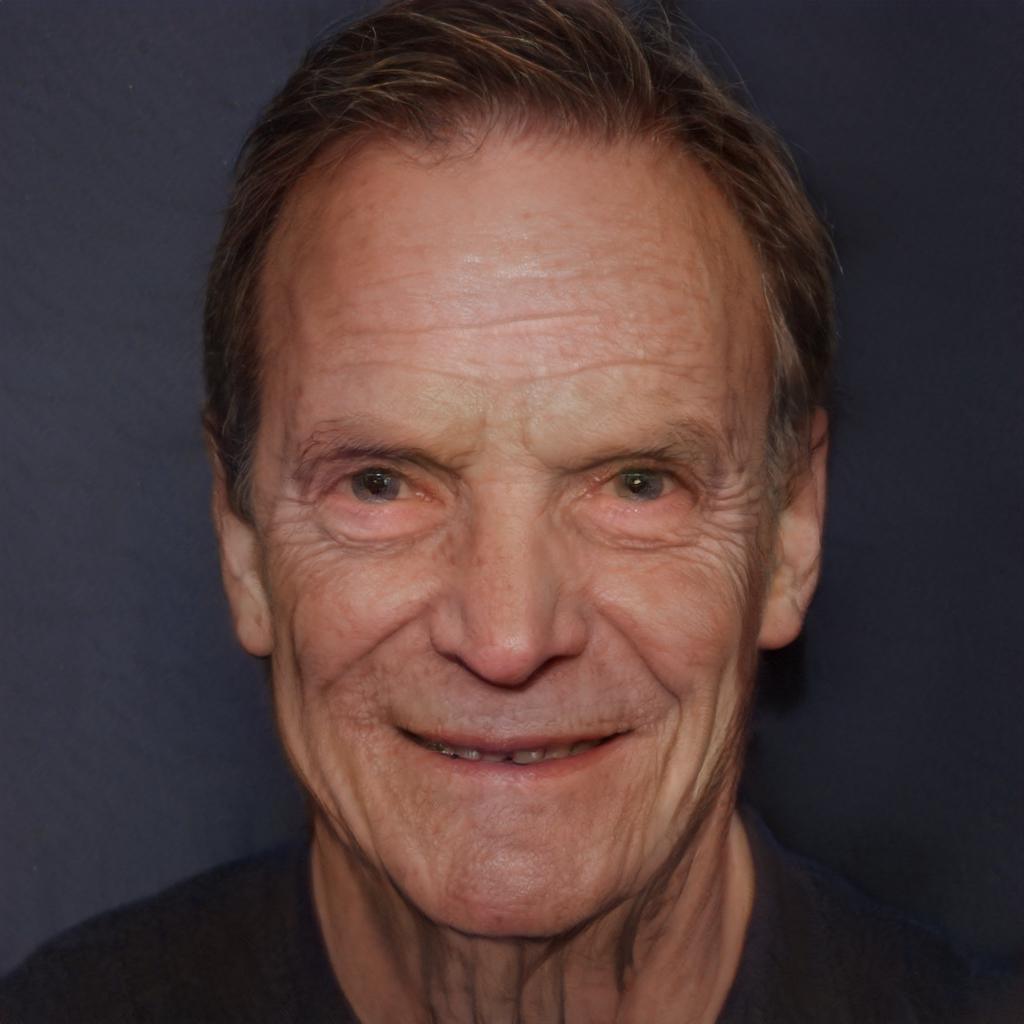} 
        \tabularnewline
        \tabularnewline
        \includegraphics[width=0.135\textwidth]{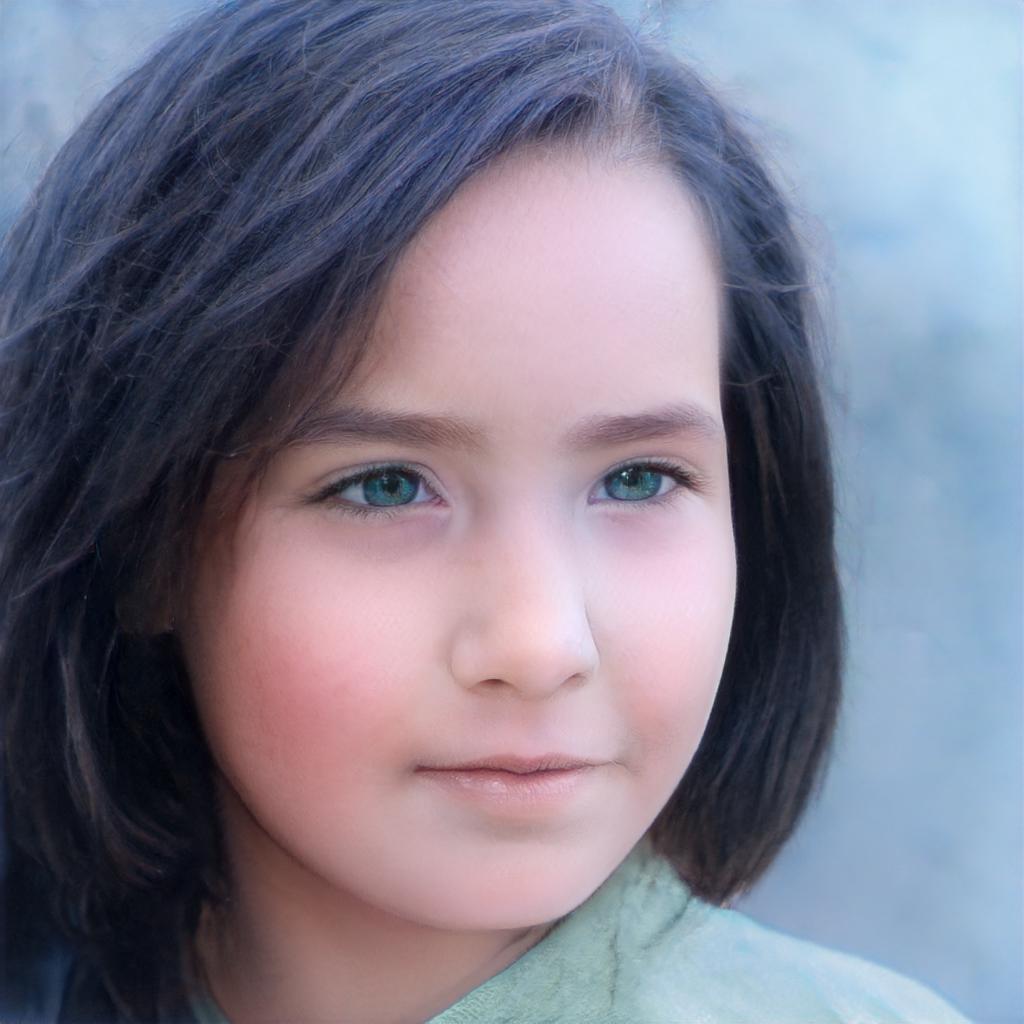} &
        \includegraphics[width=0.135\textwidth]{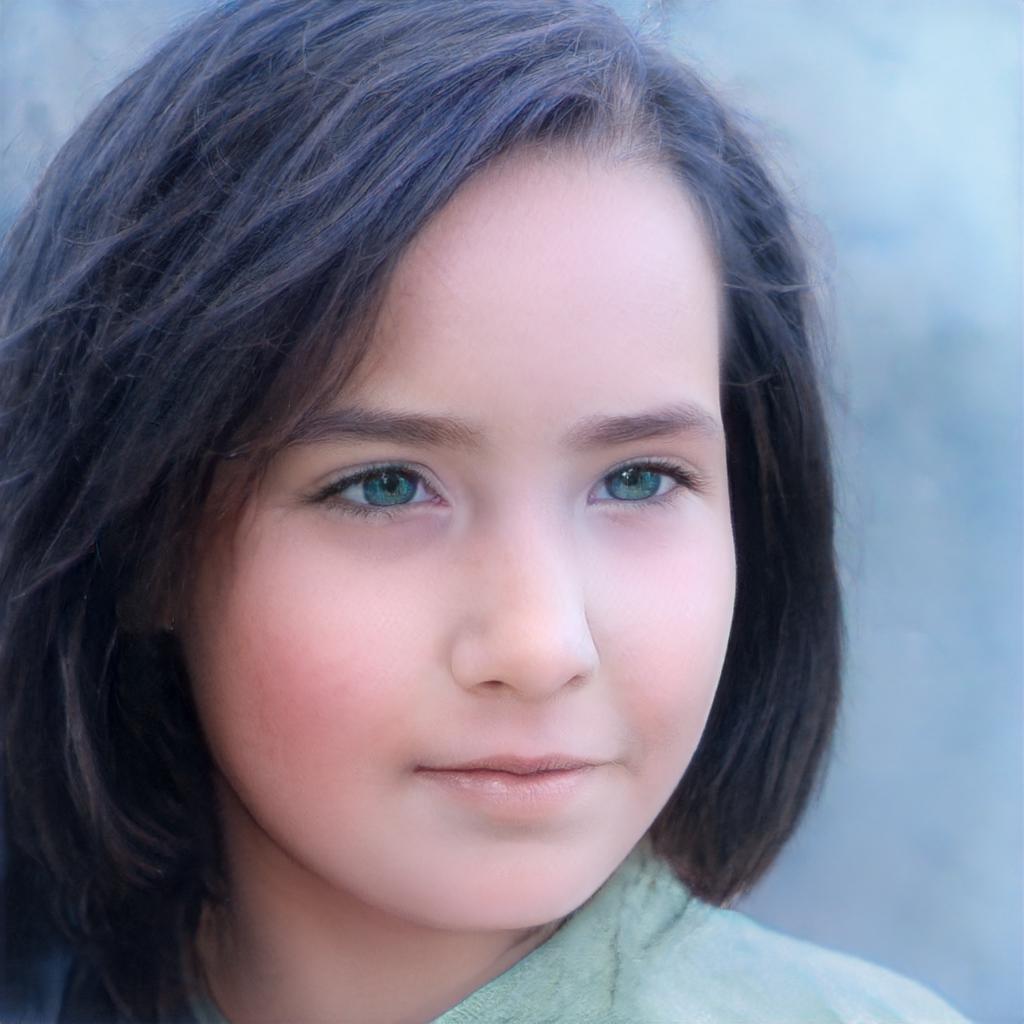} &
        \includegraphics[width=0.135\textwidth]{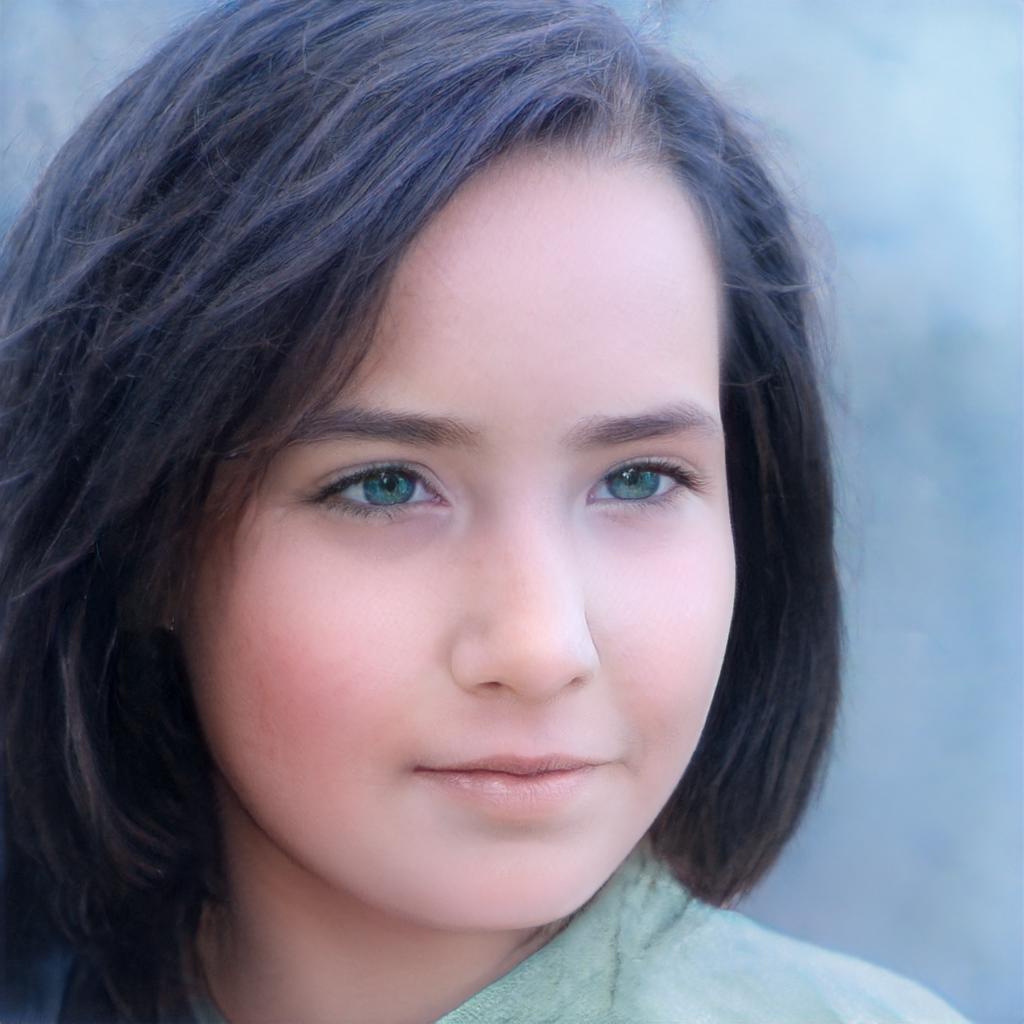} &
        \includegraphics[width=0.135\textwidth]{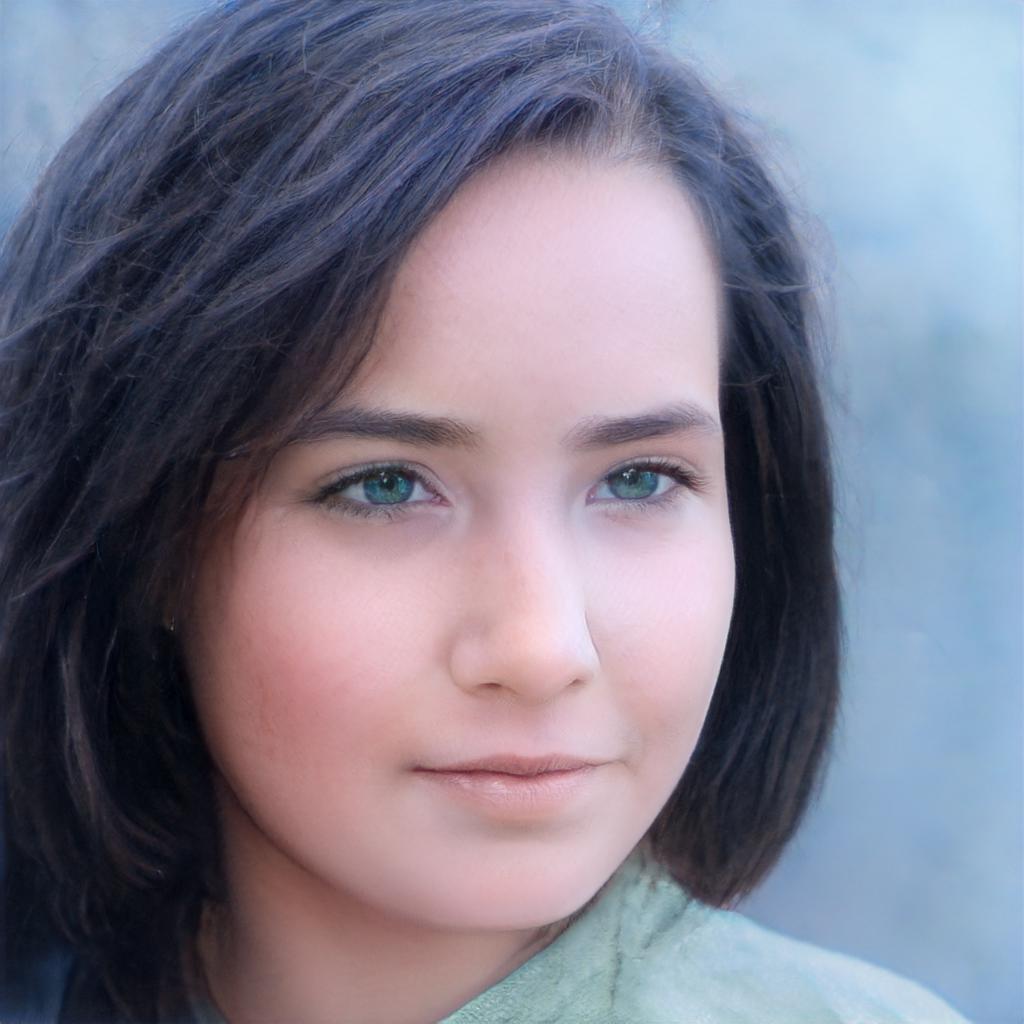} &
        \includegraphics[width=0.135\textwidth]{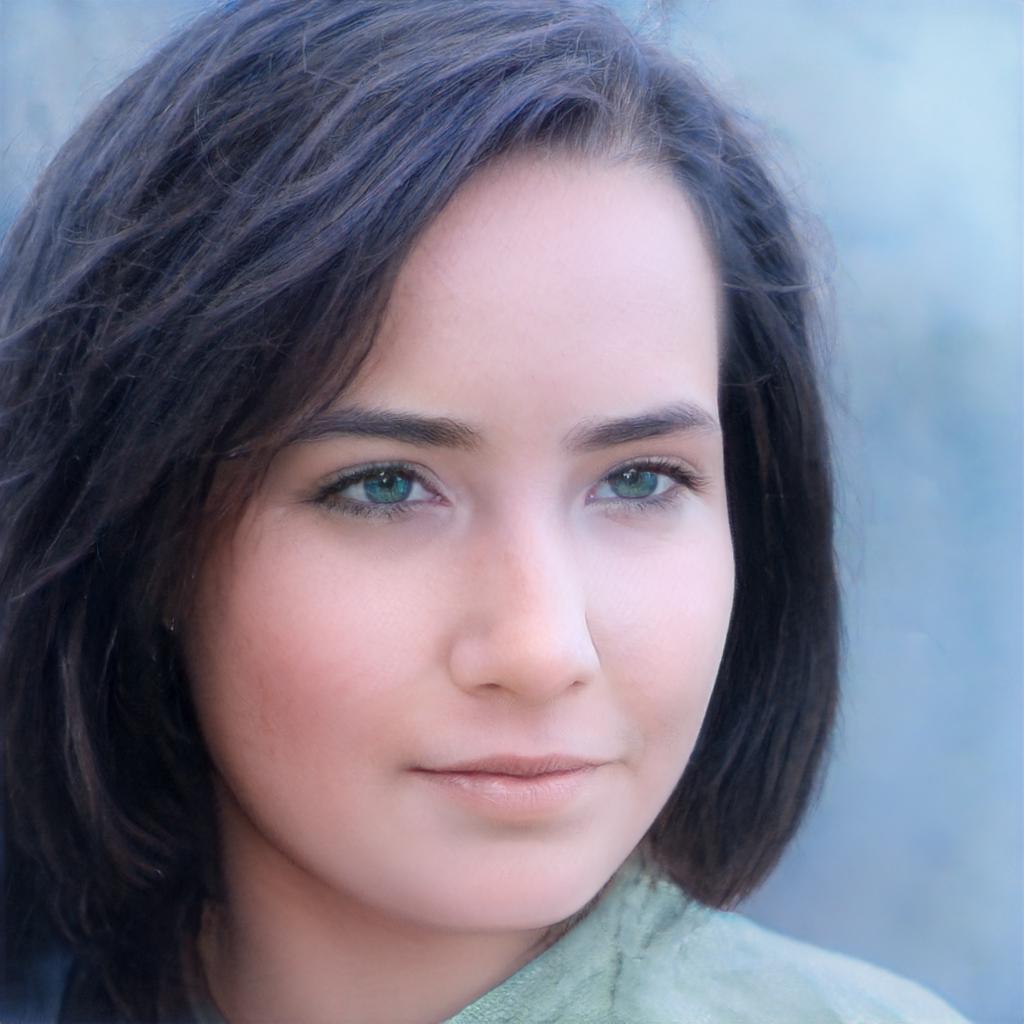} &
        \includegraphics[width=0.135\textwidth]{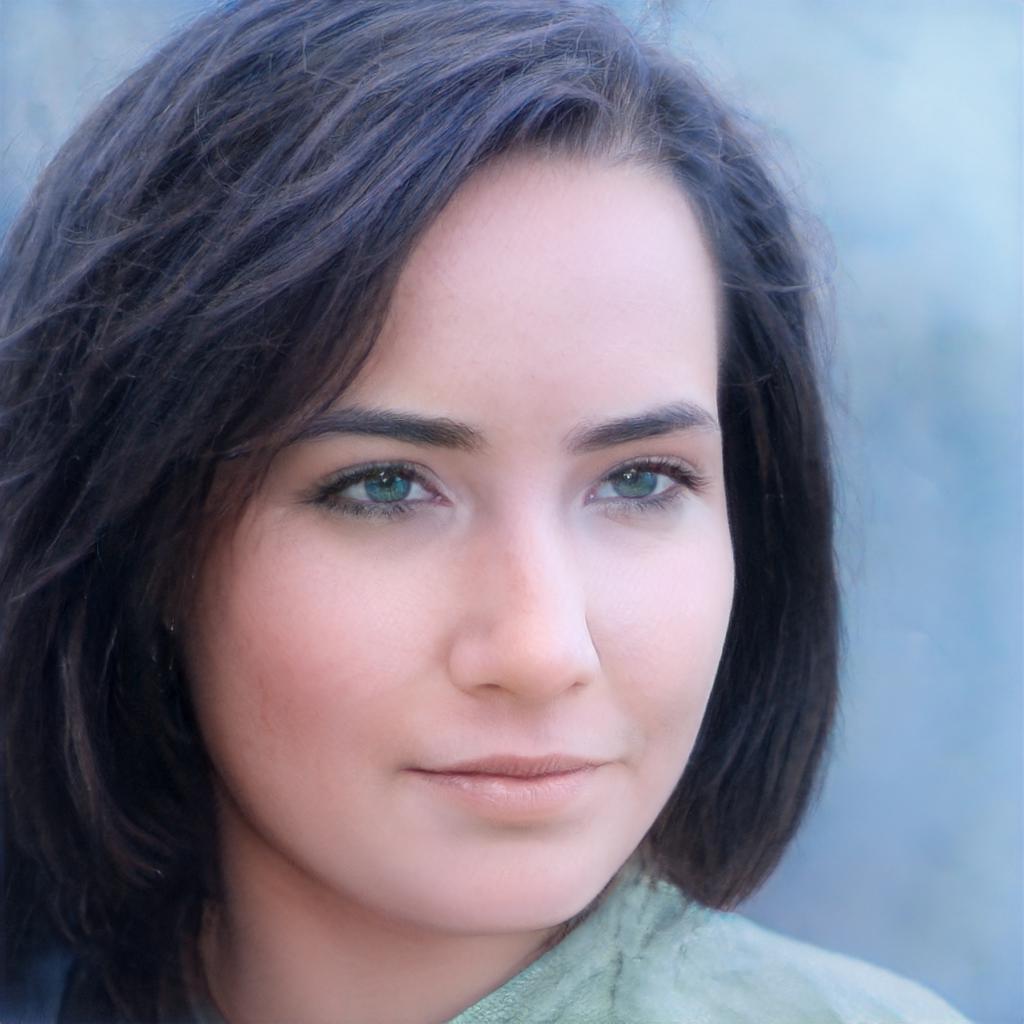} 
        \tabularnewline
        \includegraphics[width=0.135\textwidth]{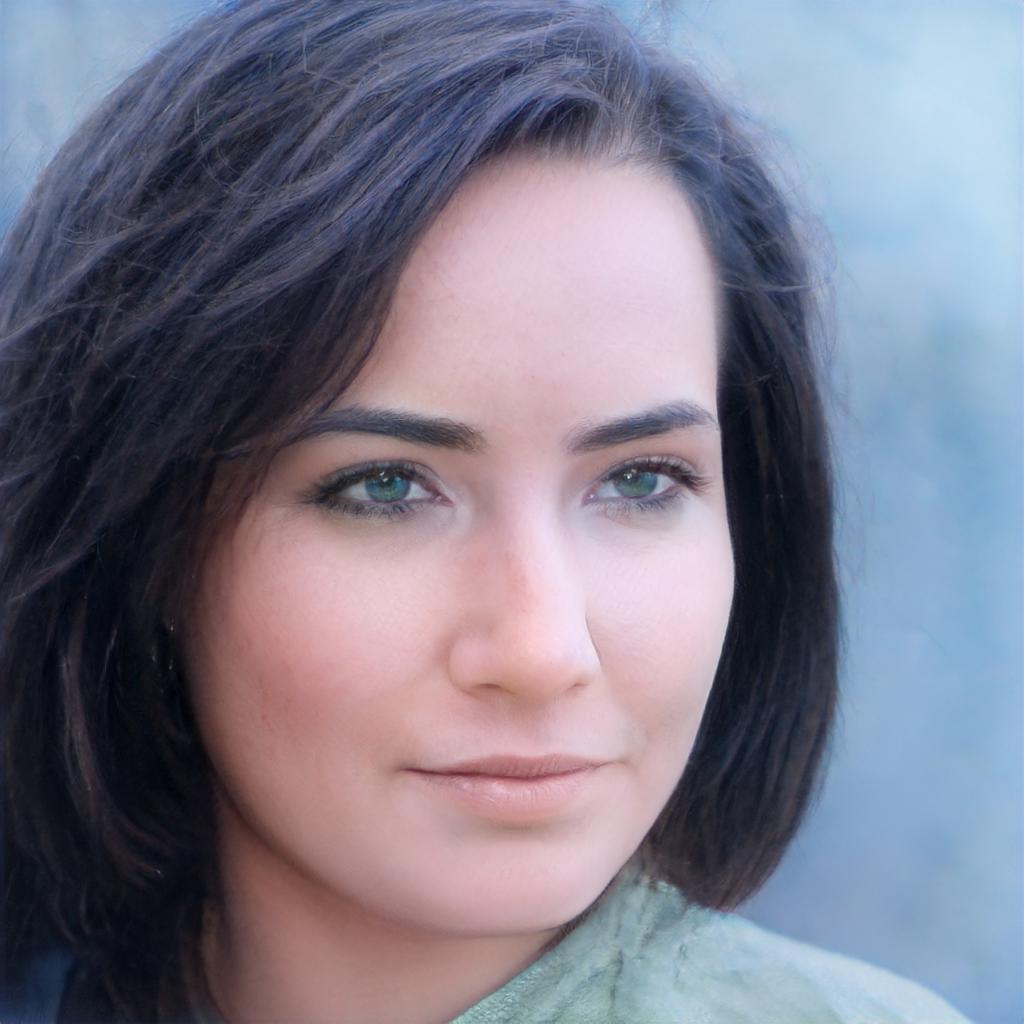} &
        \includegraphics[width=0.135\textwidth]{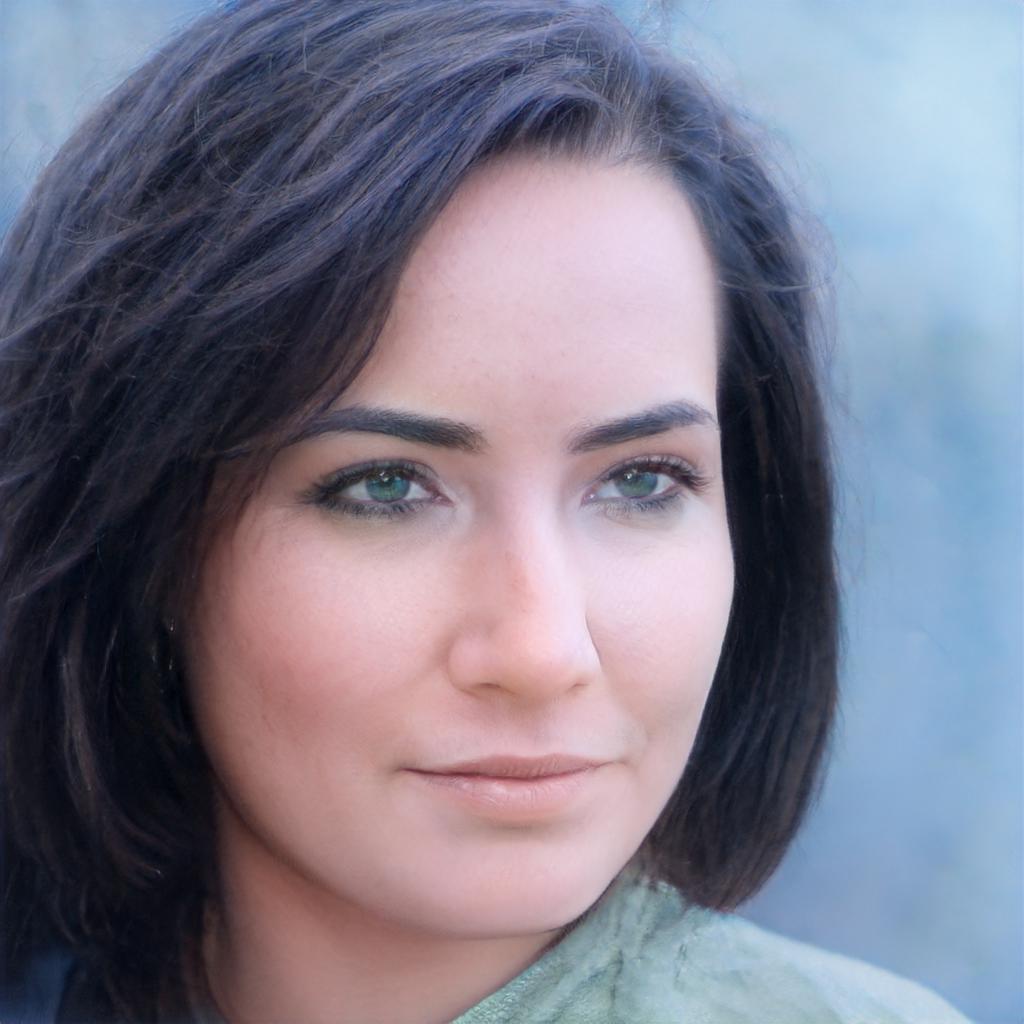} &
        \includegraphics[width=0.135\textwidth]{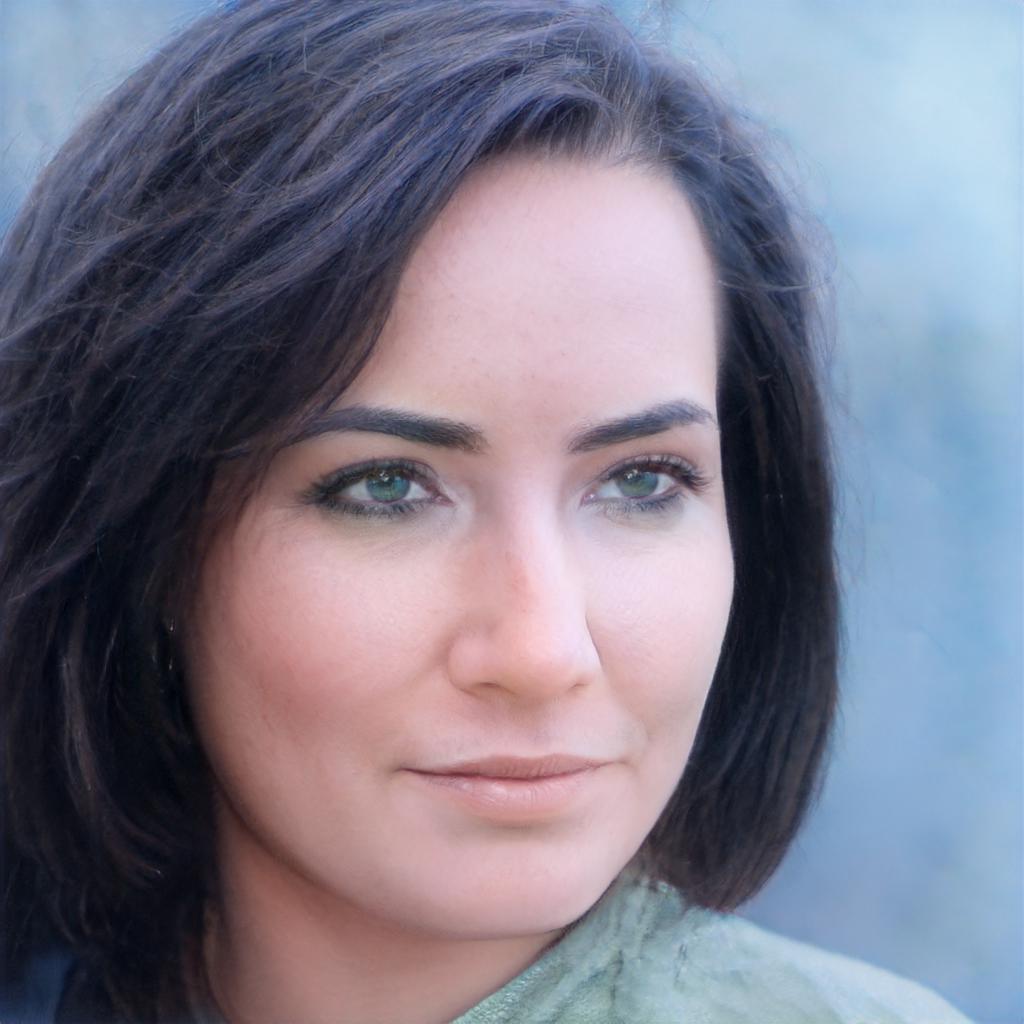} &
        \includegraphics[width=0.135\textwidth]{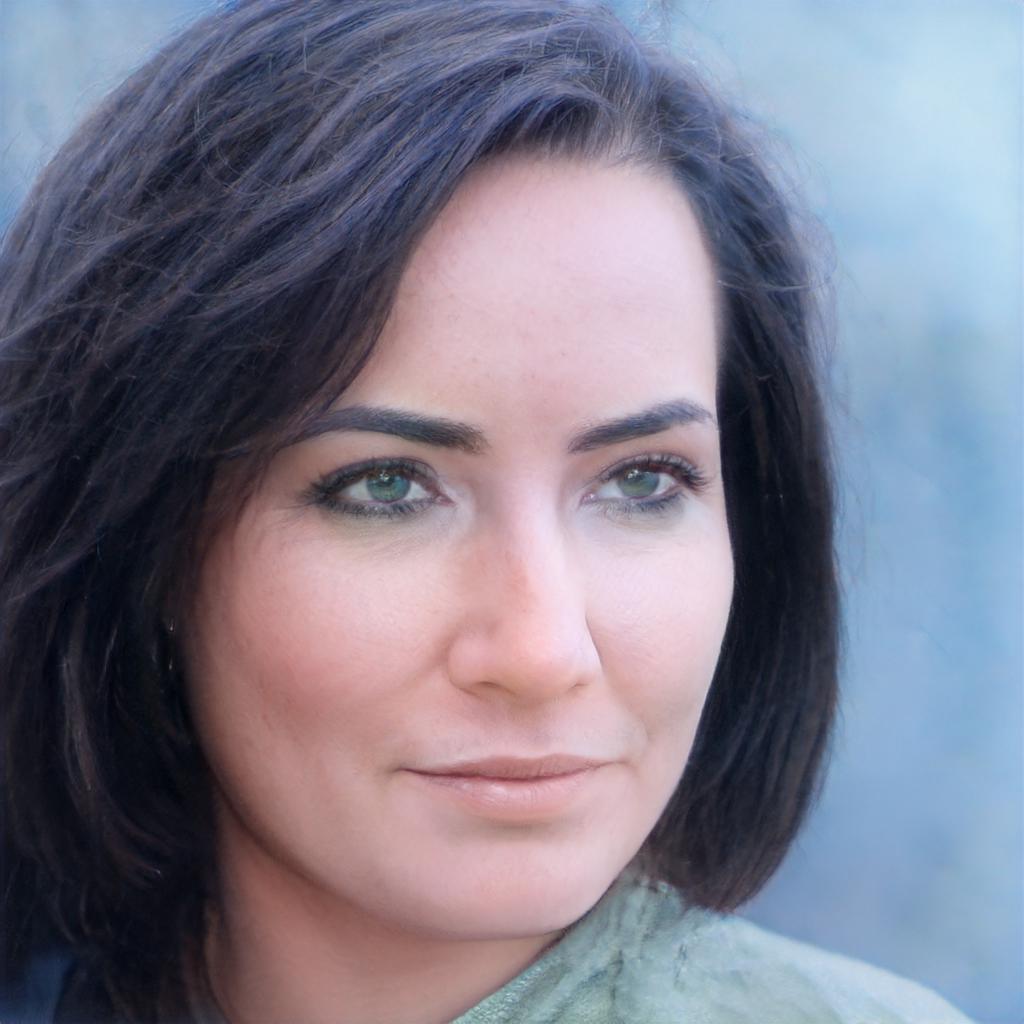} &
        \includegraphics[width=0.135\textwidth]{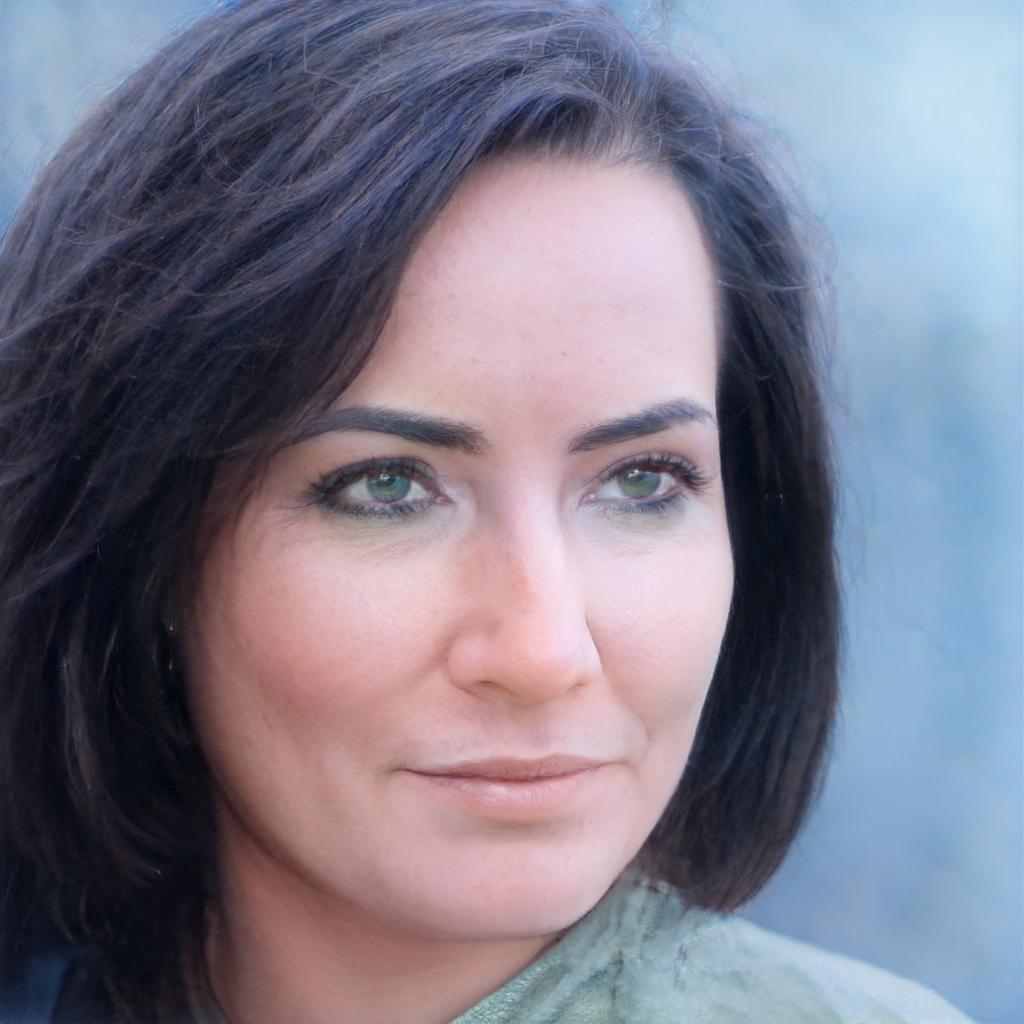} &
        \includegraphics[width=0.135\textwidth]{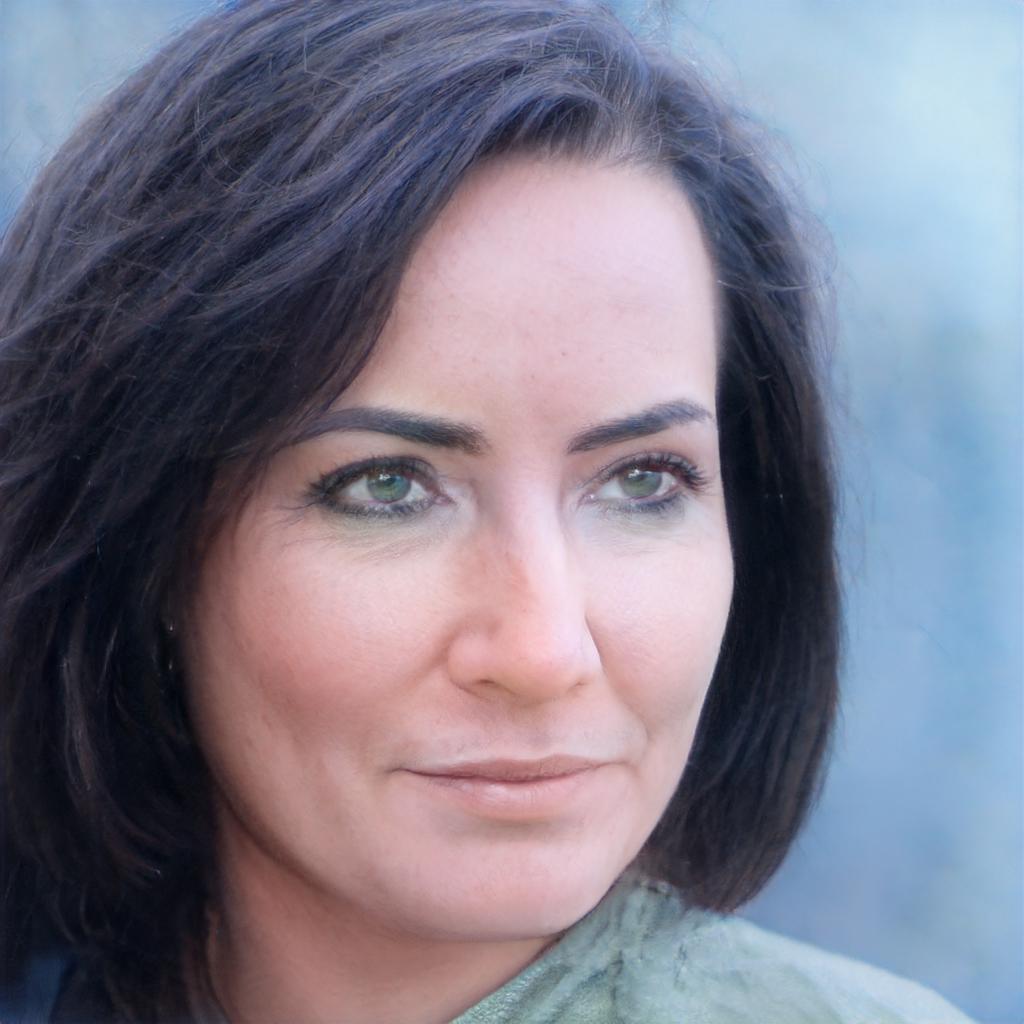} 
        \tabularnewline
        \includegraphics[width=0.135\textwidth]{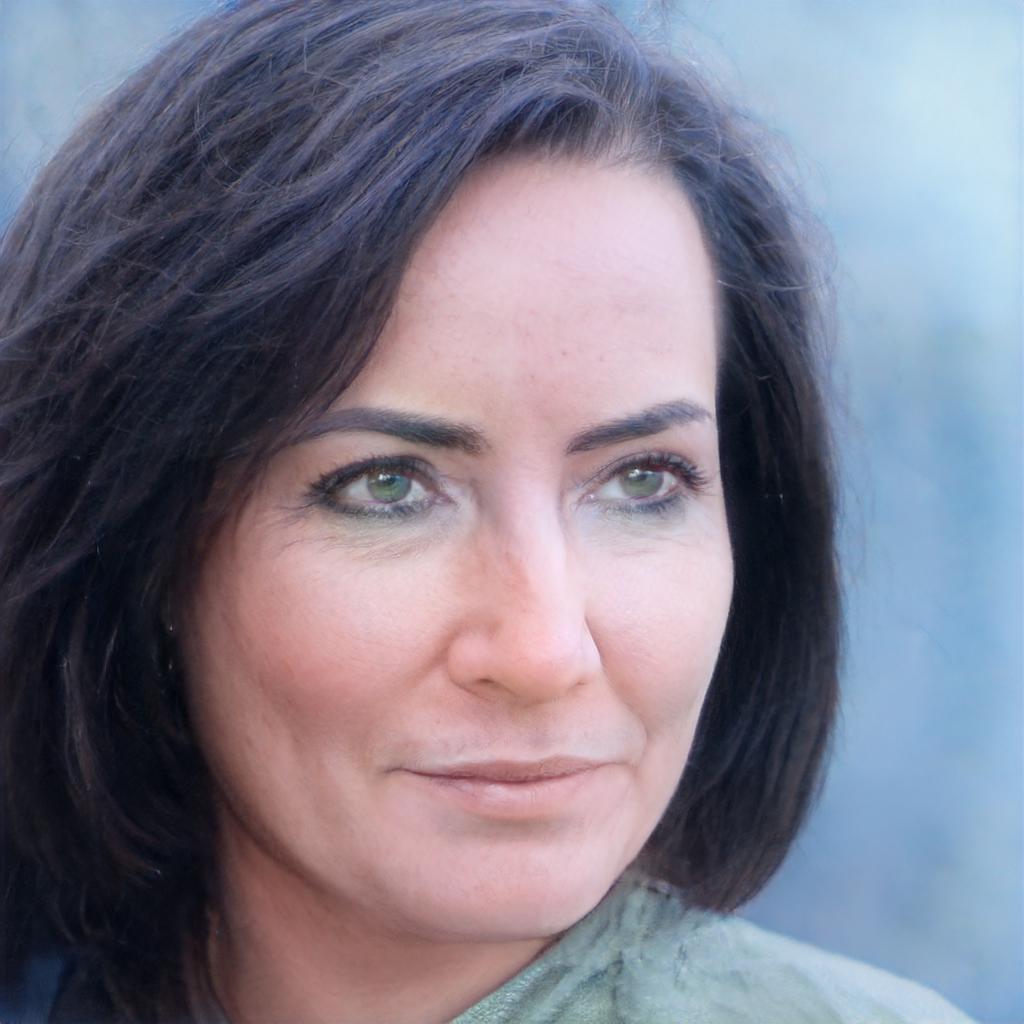} &
        \includegraphics[width=0.135\textwidth]{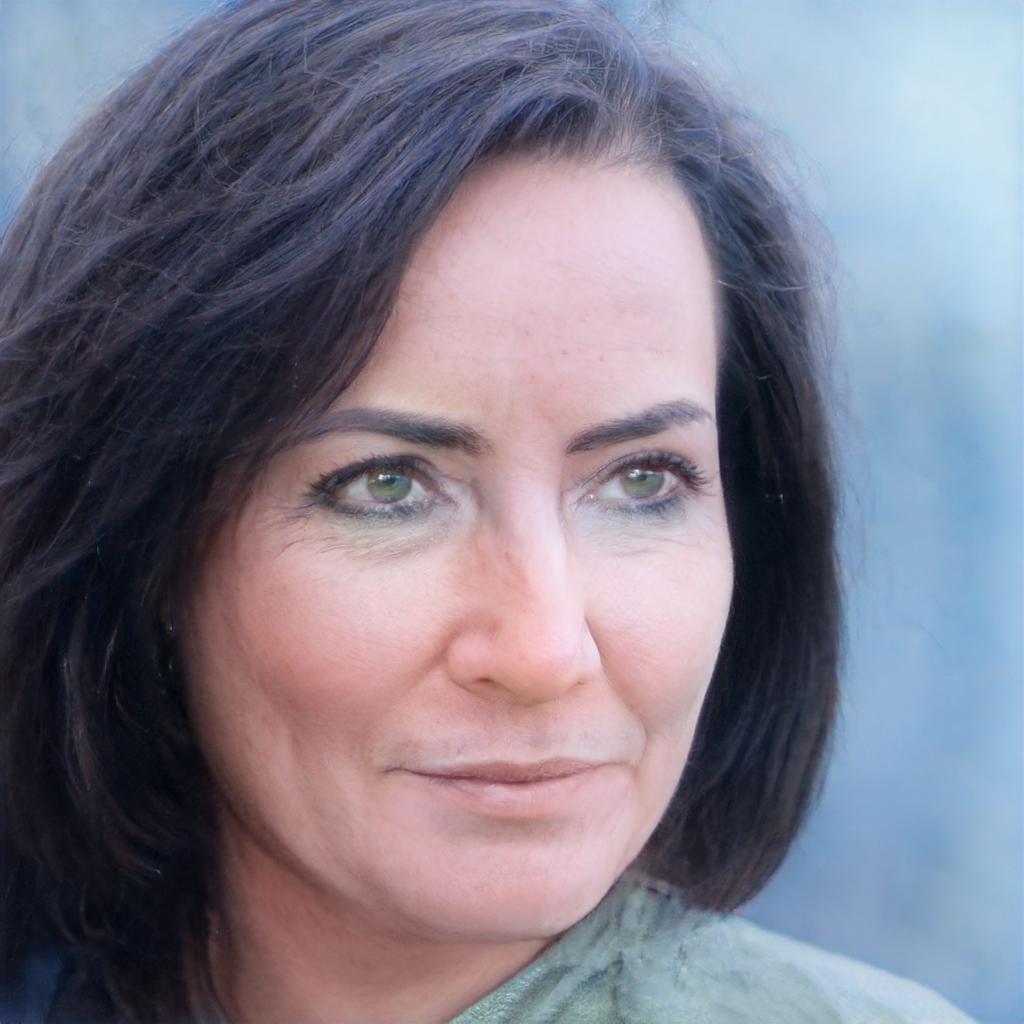} &
        \includegraphics[width=0.135\textwidth]{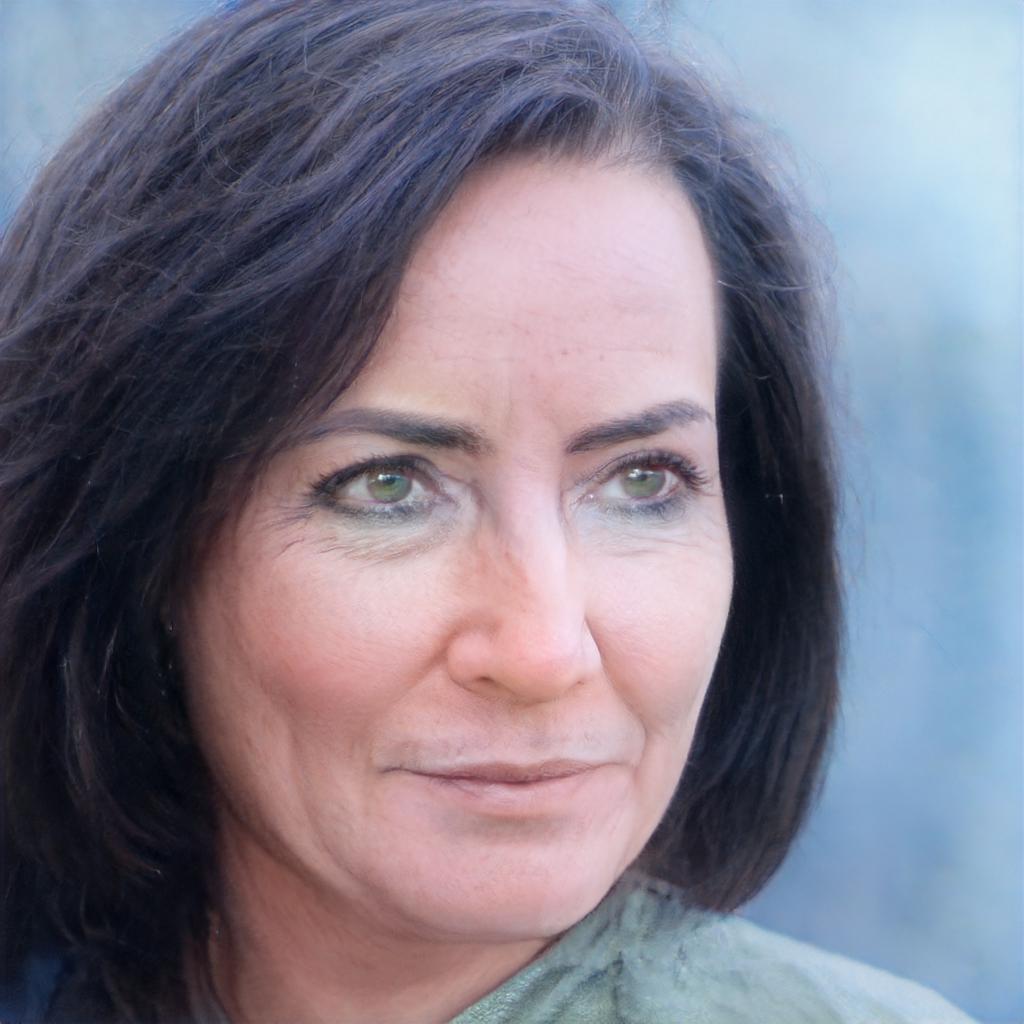} &
        \includegraphics[width=0.135\textwidth]{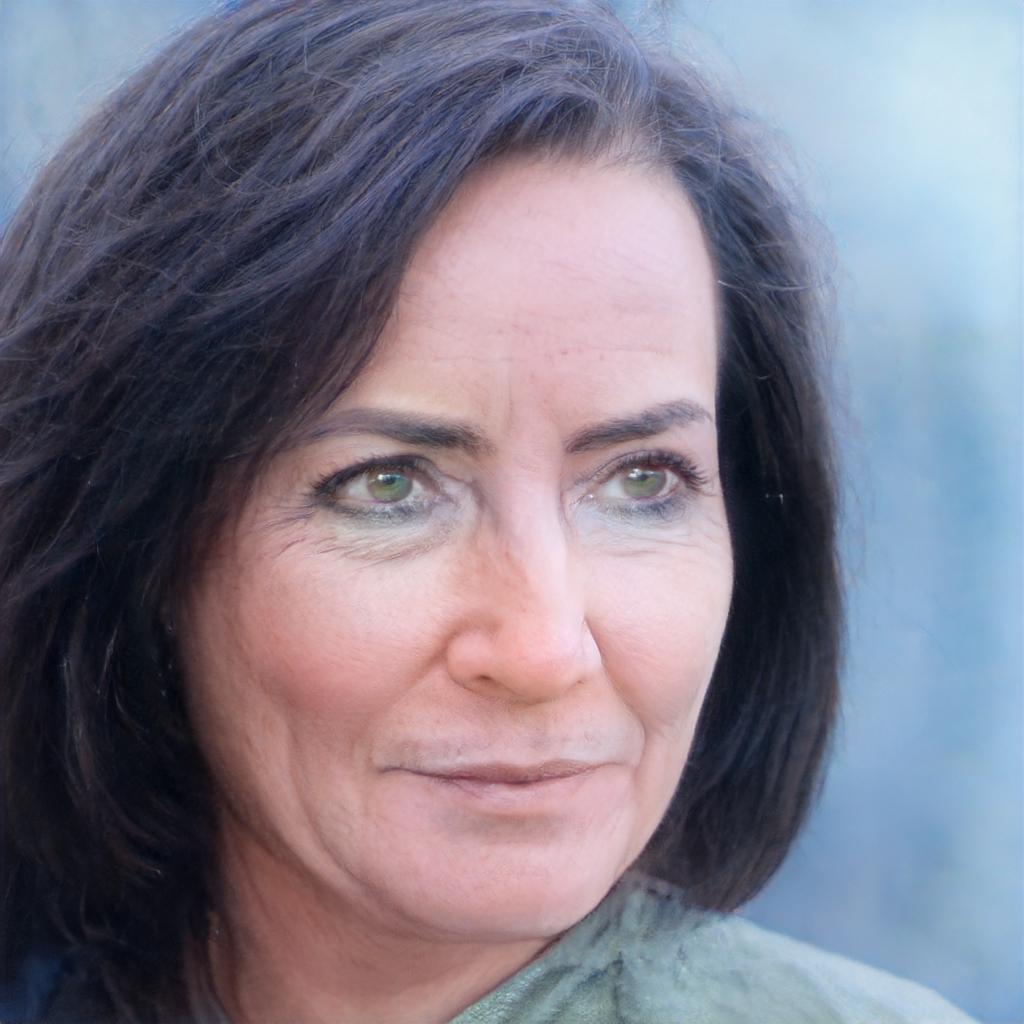} &
        \includegraphics[width=0.135\textwidth]{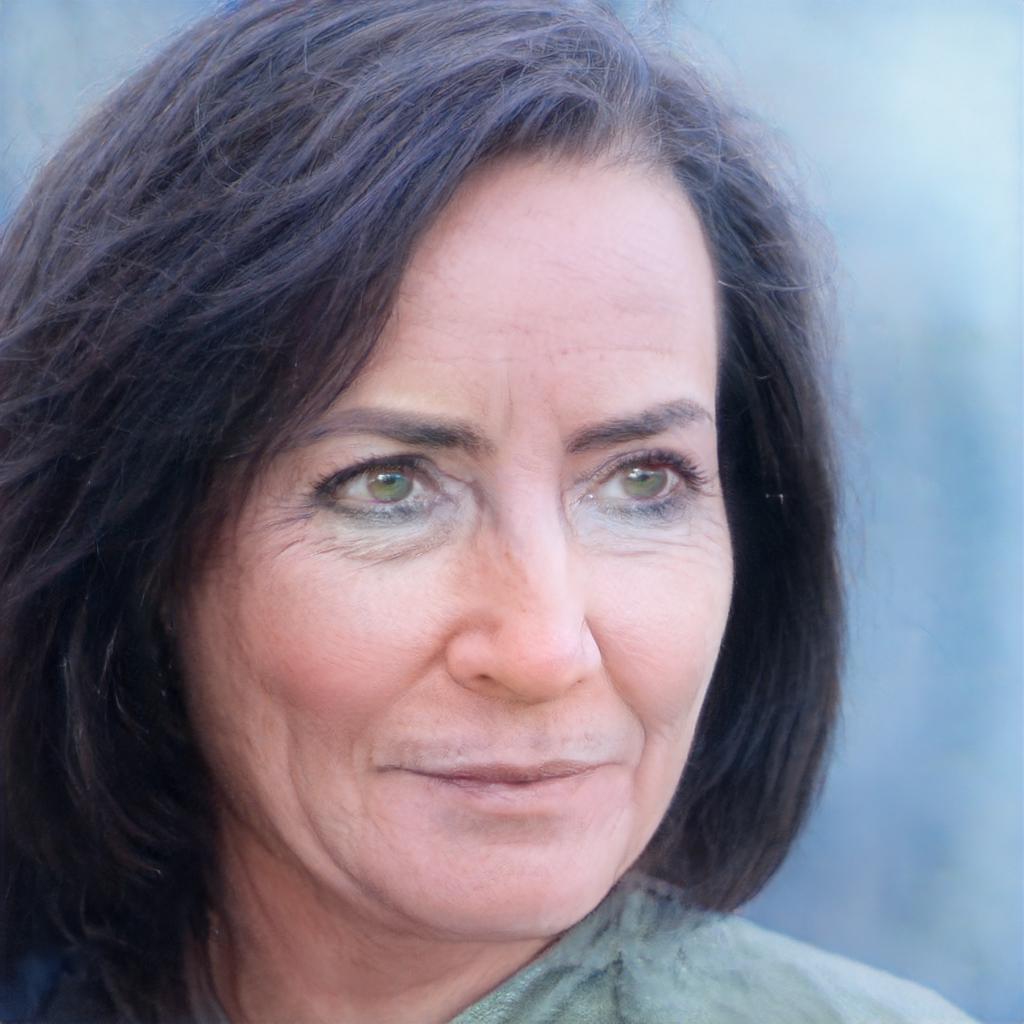} &
        \includegraphics[width=0.135\textwidth]{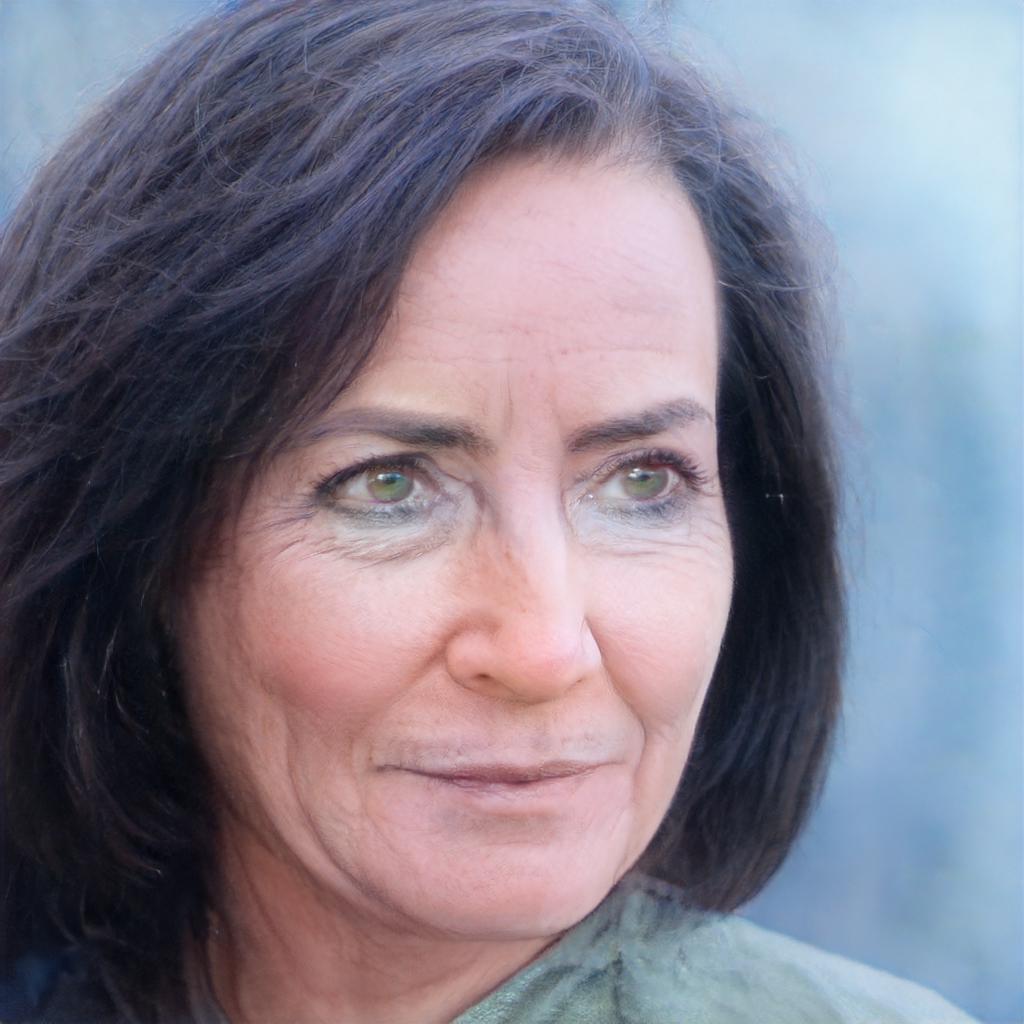} 
        \tabularnewline
        \includegraphics[width=0.135\textwidth]{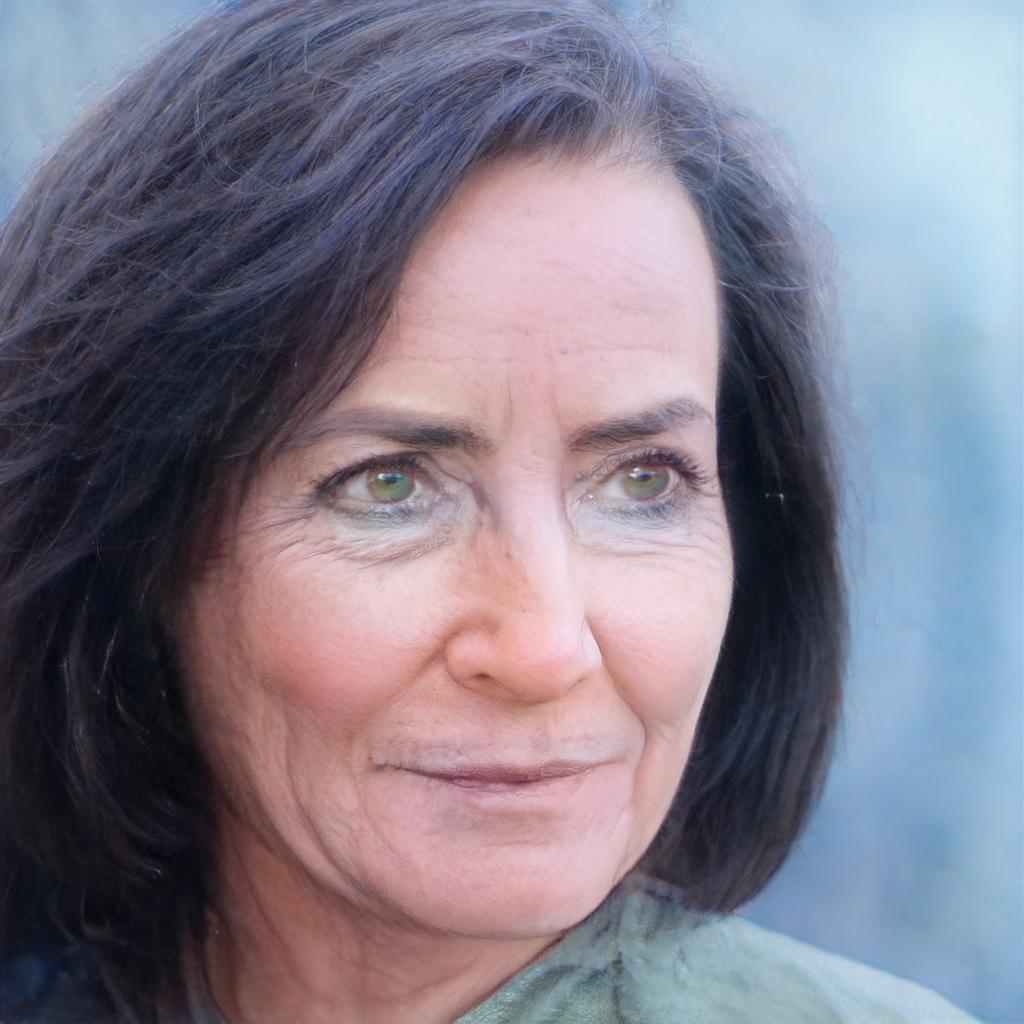} &
        \includegraphics[width=0.135\textwidth]{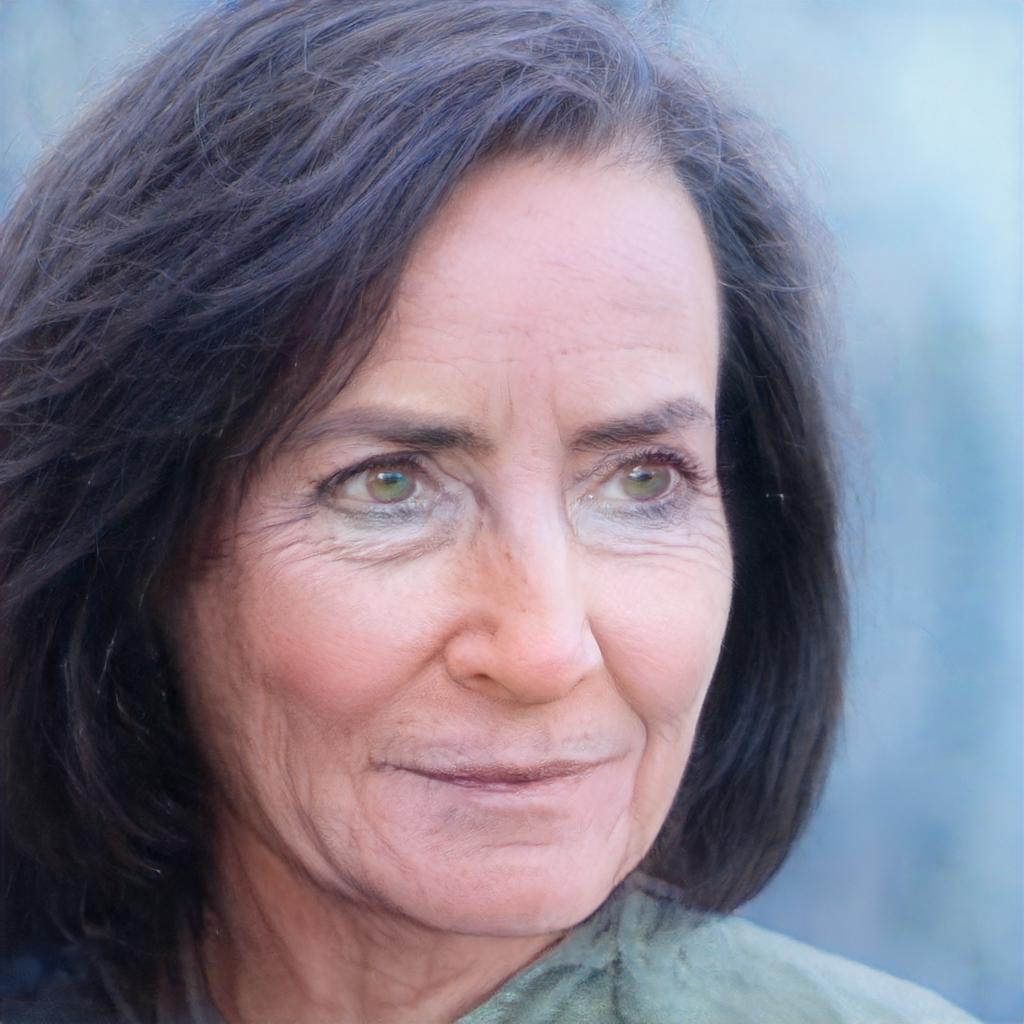} &
        \includegraphics[width=0.135\textwidth]{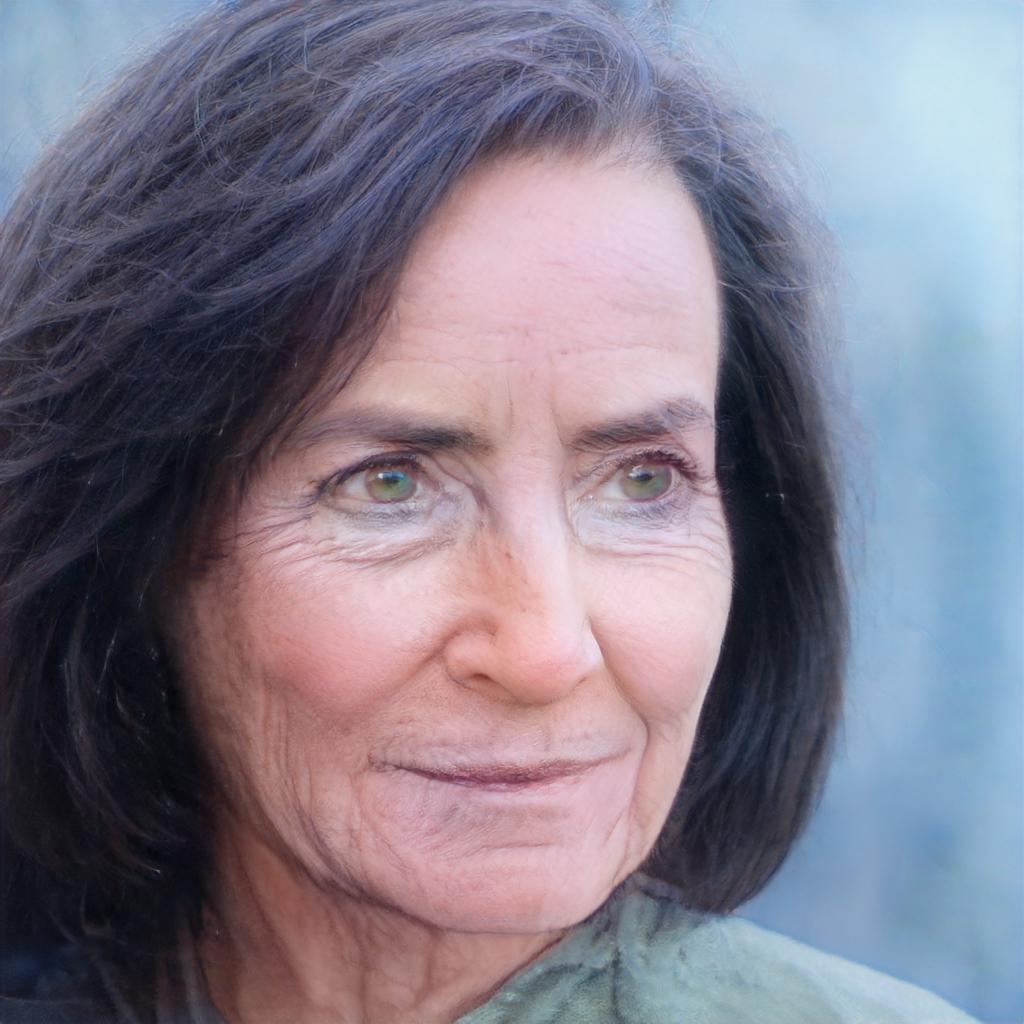} &
        \includegraphics[width=0.135\textwidth]{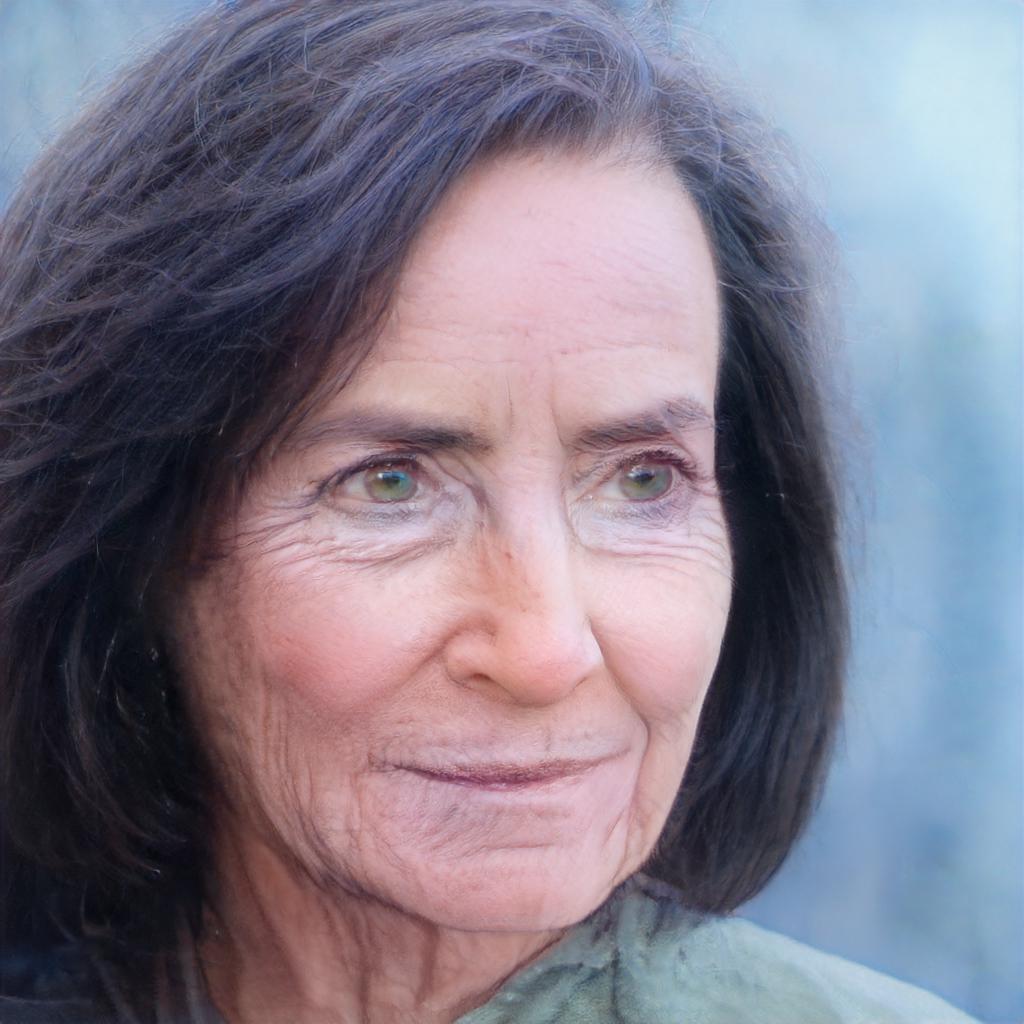} &
        \includegraphics[width=0.135\textwidth]{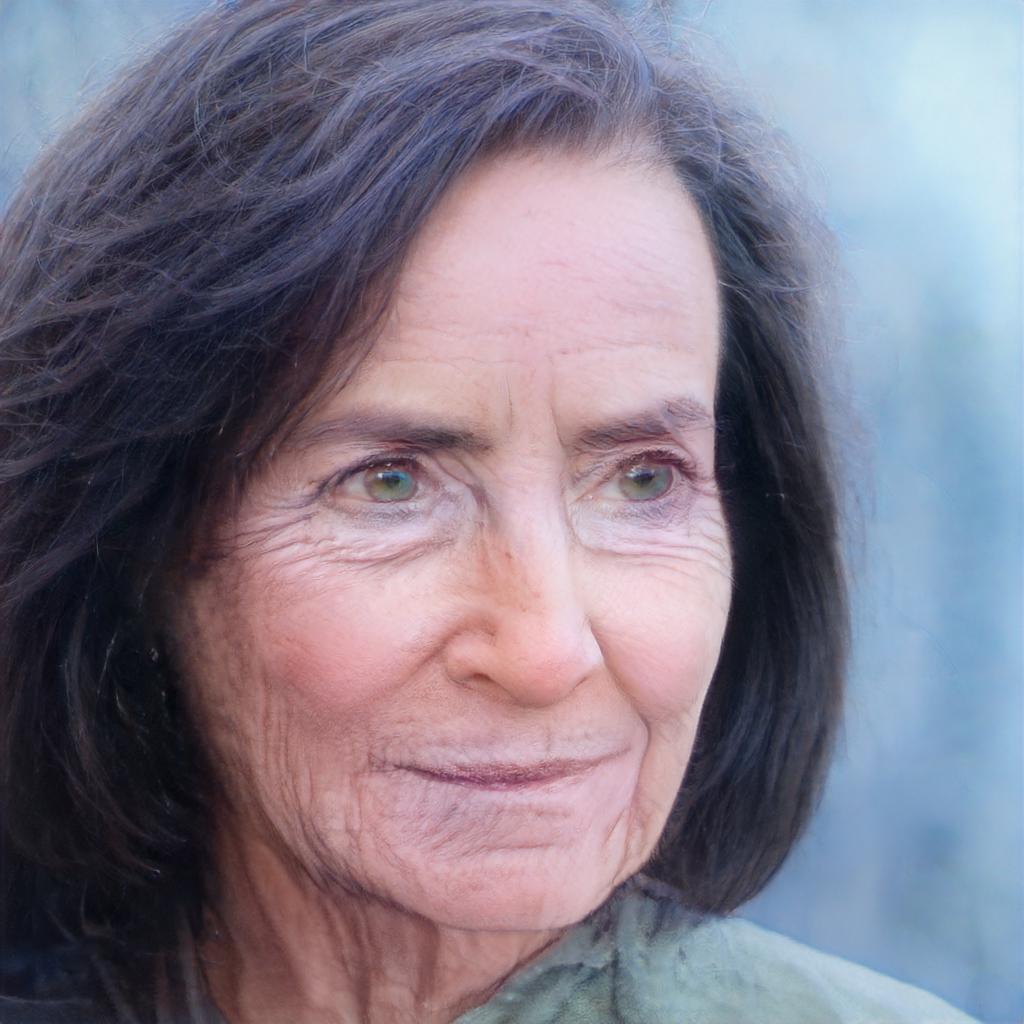} &
        \includegraphics[width=0.135\textwidth]{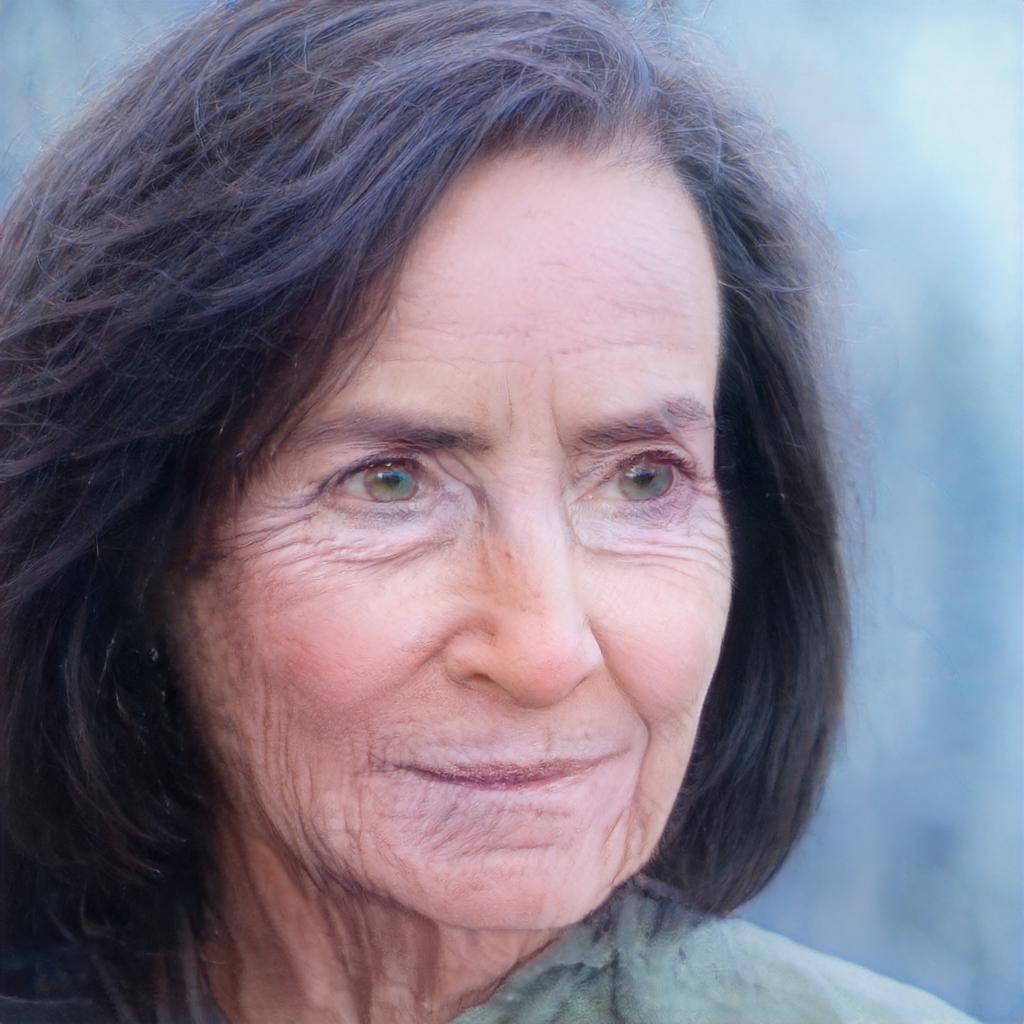} 
        \tabularnewline
    \end{tabular}
    \setlength{\belowcaptionskip}{-10pt}
    \caption{Full lifespan aging results generated using SAM.}
    \label{fig:appendix_full_lifespan}
\end{figure*}